%% file: main.tex
\documentclass{article} 
\usepackage{iclr2027_conference,times}

\input{math_commands.tex}

\usepackage{graphicx}
\usepackage{array}
\usepackage{booktabs}
\usepackage{tabularx} 
\usepackage{multirow}
\usepackage{makecell}
\usepackage{enumitem}
\usepackage[table]{xcolor}
\usepackage{tcolorbox}
\graphicspath{{imgs/}}
\tcbuselibrary{breakable,skins}
\usepackage{placeins} 
\usepackage{pifont} 
\usepackage{CJKutf8} 
\usepackage{algorithm}
\usepackage[noend]{algpseudocode} 

\title{OpenVLHarness: Real World Visual Agentic Tool-Use with Multimodal Memory}

\author{%
\bf Yu Zhou\textsuperscript{*,1}, Cheng-Fu Yang\textsuperscript{*,1}, Rui Sun\textsuperscript{*,1}, Di Wu\textsuperscript{*,1}, Sibo Peng\textsuperscript{1}, Mingyang Zhang\textsuperscript{1}, \\
\bf Zhen Yang\textsuperscript{2}, Ruiyi Zhang\textsuperscript{2}, Nanyun Peng\textsuperscript{1}, Zhe Gan\textsuperscript{2}, Kai-Wei Chang\textsuperscript{1} \\[6pt]
\normalfont \textsuperscript{1}University of California, Los Angeles \qquad \textsuperscript{2}Apple \qquad \textsuperscript{*}Equal contribution \\[16pt]
\normalfont \paperlink{Project page:}{https://openvlharness.github.io}{openvlharness.github.io} \qquad
\paperlink{Code:}{https://github.com/uclanlp/OpenVLHarness}{github.com/uclanlp/OpenVLHarness}
}

\usepackage{xspace}
\newcommand{\ours}{OpenVLHarness\xspace}

\definecolor{rsblue}{HTML}{1F3A93}

\definecolor{my_purple}{HTML}{A02B93}
\definecolor{my_blue}{HTML}{00B0F0}
\definecolor{mygreen}{HTML}{00A64F}
\definecolor{myWhite}{HTML}{FFFFFF}
\definecolor{myGreen}{HTML}{28A745}  
\definecolor{myRed}{HTML}{DC3545} 

\usepackage{hyperref} 

\definecolor{paperlinkblue}{HTML}{0064E0}
\newcommand{\paperlink}[3]{{\hypersetup{pdfborder={0 0 0}}
  \textbf{#1}~\href{#2}{\textcolor{paperlinkblue}{#3}}}}

\newif\ifshowexamples
\showexamplestrue 
\newcommand{\fullexample}[1]{\ifshowexamples Example~\ref{#1}\else see appendix\fi}

\iclrfinalcopy
\begin{document}

\maketitle
\lhead{Preprint}

\vspace{-11pt} 
\input{text/0_abstract}
{\setlength{\intextsep}{0pt}\input{figs/teaser}} 
\input{text/1_intro}
\input{text/3_harness_design}
\input{text/5_experiment}
\input{text/2_related_work}
\input{text/7_conclusion}

\newpage
\input{text/9_acknowledgements} 


\bibliography{custom}
\bibliographystyle{iclr2027_conference}
\newpage
\appendix
\let\oldappsection\section\renewcommand{\section}{\FloatBarrier\oldappsection}
\let\oldappsubsection\subsection\renewcommand{\subsection}{\FloatBarrier\oldappsubsection}
\input{text/appendix}

\end{document}

%% file: math_commands.tex
\usepackage{amsmath,amsfonts,bm}

\def\eqref#1{equation~\ref{#1}}

\def\1{\bm{1}}

\DeclareMathAlphabet{\mathsfit}{\encodingdefault}{\sfdefault}{m}{sl}
\SetMathAlphabet{\mathsfit}{bold}{\encodingdefault}{\sfdefault}{bx}{n}



%% file: text/0_abstract.tex
\begin{abstract}

Multimodal large language models (MLLMs) have made rapid progress on visual reasoning, yet many tasks require evidence that is difficult to obtain through internal reasoning alone: our controlled study on 22 benchmarks shows that simply increasing the inference budget of the base model yields only modest gains. Such tasks demand grounded perception, external knowledge, and numerical measurements obtained through heterogeneous tools, whose outputs must be made interpretable to the model and retained across reasoning steps. We introduce \ours, a multimodal reasoning harness organized around a capability layer and persistent multimodal memory. The capability layer combines task-level tool interfaces with evidence rendering, providing consistent access to heterogeneous backends and converting their outputs into model-readable observations. Persistent multimodal memory retains the corresponding images, structured data, and web references as addressable artifacts throughout each query. A coding interface executes generated programs over stored data, enabling computations that combine results from multiple tools. Together, these mechanisms support an iterative process of acquiring, inspecting, and composing multimodal evidence. Evaluated on 22 benchmarks with eight open and proprietary backbones, \ours improves average scores by 9.7--12.6 points over matched no-tool baselines, including +11.2 with Qwen3-VL-8B, +9.7 with Kimi K3, and +9.9 with GPT-6 Astra. With GPT-6 Luna at medium reasoning effort, \ours outperforms the base model, Codex, and Visual Sketchpad at maximum effort by 7.7--8.5 points while incurring at most 37\% of their model-token cost.

\end{abstract}

%% file: figs/teaser.tex
\begin{figure}[!h]
    \centering
    \includegraphics[width=\linewidth]{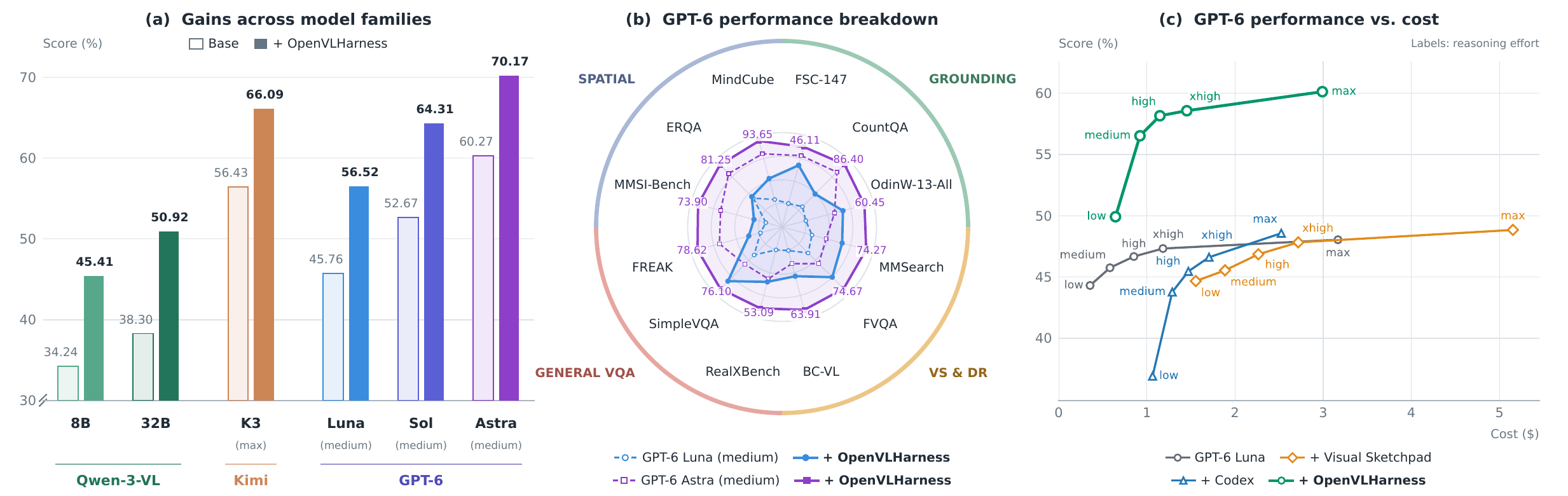}
    \caption{\textbf{Performance of \ours across models and settings.}
    \textbf{(a)} Average score over 22 challenging multimodal benchmarks with and without \ours, across the Qwen, Kimi, and GPT model families.
    \textbf{(b)} Benchmark scores for GPT-6 Luna and GPT-6 Astra at medium reasoning effort, each with and without \ours; radii are rescaled per benchmark, and numbers give the raw scores of GPT-6 Astra (medium) with \ours.
    \textbf{(c)} Average score versus model-token cost (average for 1,000 query rollouts) on 22 benchmarks for GPT-6 Luna alone and with Codex, Visual Sketchpad, or \ours, across all five reasoning-effort settings.}
    \label{fig:teaser}
\end{figure}

%% file: text/1_intro.tex
\section{Introduction}

Multimodal large language models (MLLMs) are increasingly used as reasoning engines over images, documents, webpages, and videos. However, many real-world visual tasks require information that the model cannot reliably infer from the input alone. Answering a question may require localizing numerous tiny objects in dense scenes, retrieving external knowledge about a long-tail entity, or computing quantities from intermediate measurements. Such tasks require additional multimodal evidence acquired through perception, search, and computation. Our inference-time scaling study (Fig.~\ref{fig:teaser}c) reveals diminishing returns at higher reasoning budgets: moving the base model from xhigh to max effort multiplies its model-token spending by 2.7 but improves its average score by only 0.7 points. This raises the question of whether further progress requires better access to task-relevant multimodal evidence, beyond additional reasoning over the initial input.

Visual tool-use agents address this need by connecting MLLMs to perception models, search engines, and code execution \citep{hong2025deepeyesv2, mmsearchr1, deepmmsearchr1, browsecompvl, su2025openthinkimg, wu2025vtool, suris2023vipergpt, gupta2023visual}. Building a general harness around these tools presents two connected challenges. First, heterogeneous backends expose different calling conventions and return outputs such as segmentation masks, depth maps, and retrieved webpages that need to be presented in a form the model can interpret. Second, subsequent reasoning steps may still need access to the underlying artifacts: a rendered depth map supports visual inspection, but a precise measurement requires its numerical values. Effective multimodal tool use therefore requires both consistent tool interfaces with evidence rendering and a mechanism for retaining and reusing the resulting artifacts.

We introduce \ours, a multimodal reasoning harness organized around a \emph{capability layer} and \emph{persistent multimodal memory}. The capability layer combines \emph{task-level tool interfaces} with \emph{model-facing evidence rendering}. Task-level interfaces define operations independently of a particular backend, allowing tools to be replaced without changing the orchestrator's task vocabulary. Evidence rendering converts heterogeneous outputs into observations that combine visualizations, concise descriptions, and compact structured information. For example, visual grounding accepts an image and a textual query, then presents the detected regions through an annotated image and coordinates. Together, these mechanisms let the orchestrator invoke tools consistently and inspect their results in a form suited to multimodal reasoning.

Persistent multimodal memory makes the corresponding artifacts available across turns within a query. Images, structured data, and web references are registered under reusable handles, allowing later calls to operate on earlier evidence. In the grounding example, the orchestrator can inspect the overlay, then reuse the retained masks and a subsequently acquired depth array to measure depth within each region. A Python Coding Agent supports such compositions: the orchestrator specifies a computational objective in natural language, and a code-generation session backed by the same backbone model writes a program that executes over the stored data. The capability layer makes tool results interpretable; memory makes their underlying artifacts addressable and reusable. Together, they support an iterative process of acquiring, inspecting, and computing over multimodal evidence.

We evaluate \ours on 22 benchmarks spanning grounded perception, visual search and deep research, general VQA and hallucination, and spatial reasoning, using eight open and proprietary MLLM backbones of different scales (Fig.~\ref{fig:teaser}a). Across these settings, \ours consistently improves average performance over the base MLLMs by 9.7--12.6 percentage points and outperforms existing visual tool-use baselines (Fig.~\ref{fig:teaser}c). Scaling experiments on 22 benchmarks further show that \ours also benefits from greater reasoning effort (Fig.~\ref{fig:teaser}c), suggesting that evidence acquisition and inference-time scaling are complementary. 
We conduct ablation studies to examine the incremental effects of the capability layer and memory, showing the value of both usable tool outputs and retained multimodal artifacts. Lastly, for users who would like to adapt our harness to a custom domain, we further showcase two coding-agent-driven workflows for specialization: harness search edits prompts, tool descriptions, and output templates, while tool generation composes existing capabilities into new tools. Both operate without updating the underlying MLLM weights.

Our contributions are:
\textbf{(1)} a capability layer that combines task-level tool interfaces with model-facing evidence rendering for heterogeneous perception, search, and computation tools;
\textbf{(2)} persistent multimodal memory that retains addressable artifacts across turns and enables their reuse in subsequent tool calls and memory-grounded computation;
and \textbf{(3)} an evaluation across 22 benchmarks and eight backbones, with component ablations and domain-specialization studies, accompanied by an open-source release of the harness, tools, evaluation protocols, and two optimization workflows.


%% file: text/3_harness_design.tex
\section{Harness Design}
\label{sec:harness_design}

\input{figs/harness_design}

\ours augments an MLLM orchestrator with two complementary mechanisms (Figure~\ref{fig:harness_design}). The \emph{capability layer} combines task-level tool interfaces with model-facing evidence rendering, providing consistent access to tools and observations the orchestrator can inspect (Sec.~\ref{sec:setup_base}). \emph{Persistent multimodal memory} retains the corresponding artifacts and exposes them by reference for reuse across turns within a query (Sec.~\ref{sec:explicit_memory}). At each step, the orchestrator uses the observations and available memory references to select a tool call or produce an answer. A Python Coding Agent supports computation over stored artifacts, connecting outputs from different capabilities into multi-step procedures. The modular interfaces and renderers also support harness specialization through text edits and tool composition (Sec.~\ref{sec:harness_optim}).

\subsection{Capability Layer}
\label{sec:setup_base}

We design the capability layer with two goals: defining how the orchestrator invokes a capability and rendering the resulting evidence for inspection. The first accommodates different backends behind a consistent interface; the second addresses the heterogeneous formats of their outputs.

\paragraph{Task-level tool interfaces.}
We define a capability by the operation it provides, its typed inputs, and the evidence and reusable data it exposes. For example, visual grounding accepts an image reference and a textual query and exposes object locations, associated metadata, and a visualization; a supporting backend may additionally provide segmentation masks. Each backend is wrapped to implement this task-level interface. Visual Grounding can thus use SAM~3 or Grounding DINO, and OCR can use PaddleOCR-VL or DeepSeek-OCR, while preserving the orchestrator's calling convention (Appendix~\ref{sec:appendix_tools}). Backend selection can reflect accuracy, cost, and latency requirements. Our base instantiation includes nine capabilities: five for perception---Zoom-in, Visual Grounding, OCR, Depth Estimation, and Camera Trajectory; three for information seeking---Text Search, Image Search, and Webpage Visit; and a Python Coding Agent for computation over intermediate results. Please refer to Appendix~\ref{sec:appendix_tools} and Table~\ref{tab:appendix_base_tools} for their implementation details.

\paragraph{Model-facing evidence rendering.}
Raw tool outputs often require additional processing to become useful observations. An RLE mask encodes a region precisely, but serializing that encoding into the context does not show the orchestrator which object was segmented. Dense numerical arrays can likewise consume substantial context without providing useful, interpretable evidence.

Each capability therefore includes an output processor that renders its result into a model-facing observation. Depending on the capability, this observation combines visual attachments, concise natural-language descriptions, compact structured information, and references to reusable artifacts. Grounding can return a labeled mask overlay with the number of detected objects and normalized bounding boxes; depth estimation can return a colorized map; and Camera Trajectory can return a visualization of the recovered trajectory. The underlying masks, arrays, and camera parameters are passed to memory for later reuse. Figure~\ref{fig:harness_design} illustrates grounding for ``cabinet'' in \texttt{input\_image\_3}. The capability returns a mask overlay with a description of the five grounded objects and their bounding boxes, allowing the orchestrator to inspect the detections. If a subsequent step requires a depth measurement for each cabinet, this rendered observation supplies the visual context, while the retained masks and a depth array supply the data for computation. The memory mechanism below connects these steps.


\subsection{Persistent Multimodal Memory}
\label{sec:explicit_memory}

The capability layer makes tool results interpretable; persistent multimodal memory makes their underlying artifacts addressable and reusable. A later tool call must be able to identify a particular image, mask, or array and access its contents. Repeating a description of a previous result does not provide this access.

We maintain an explicit artifact registry for each query, with entries retained across its reasoning trajectory. The registry separates an artifact's payload from the handle used to reference it. The orchestrator sees available names and compact descriptors, while the harness retains the underlying values and resolves references when needed. The registry contains three categories:

\begin{itemize}[leftmargin=1.4em]
    \item \textbf{Image artifacts:}
    input, tool-generated, and retrieved images, exposed through handles such as \texttt{input\_image\_3} and \texttt{tool\_generated\_image\_6}. Duplicate images reuse an existing registry entry.
    \item \textbf{Structured data bindings:}
    named values containing boxes, scores, masks, depth arrays, camera parameters, OCR results, and other tool outputs. Each binding has a descriptor identifying its fields and types, including array shapes where applicable.
    \item \textbf{Web references:}
    retrieved URLs exposed through handles such as \texttt{webpage\_url\_1}, allowing later navigation without requiring the model to reproduce the original URL.
\end{itemize}

\paragraph{Registering and reusing artifacts.}
After a successful tool call, the harness registers supported artifacts and associates their handles with the capability's rendered observation. Images are delivered as visual attachments with named references; structured payloads remain accessible through data bindings. An \emph{Environment Update} advertises the available image handles, data bindings, and web references after each tool turn, exposing the current namespace without repeating every artifact's full payload. In the cabinet example, the grounding call adds an overlay image and bindings for the detection data, including bounding boxes and RLE masks. 

When a subsequent call refers to an artifact, the harness resolves its handle to the input required by the receiving tool. An image handle supplies the image for a later visual operation, a web reference supplies a URL for Webpage Visit, and a data binding supplies values for code execution. The orchestrator can thus reuse the grounding and depth results without reconstructing their contents from earlier descriptions. References connect retained evidence across tools and turns. The detailed representation and handles can be found in Appendix~\ref{sec:appendix_memory} and Table~\ref{tab:appendix_memory_artifacts}.

\paragraph{Memory-grounded computation.}
To compute depth within the detected cabinets, the orchestrator delegates a natural-language objective to the Python Coding Agent, such as ``compute the average depth within each detected region.'' The harness supplies descriptors of the available memory bindings to a fresh code-generation context of the same backbone model. The generated program executes with the actual binding values available as named variables, allowing it to combine the grounding masks with the depth array. Execution results and feedback return in language, with supported output artifacts registered for subsequent reuse. The orchestrator can then use the result or revise the objective in a later call.

This mechanism also supports filtering detections, aggregating OCR values, and other computations over intermediate results without requiring the orchestrator to serialize large arrays into its request. The capability layer supplies reusable operations and interpretable observations; memory and code execution connect their outputs into longer reasoning procedures.

\input{figs/qualitative_examples}
\paragraph{Qualitative examples.} Figure~\ref{fig:qualitative} shows three trajectories from our evaluation. In each, the agent chains several capabilities and reuses their outputs through memory: camera poses, depth maps, and grounding masks become bindings for exact 3D computation (a); grounded crops are passed by identifier to zoom-in, OCR, and image search before the answer is verified on the web (b); and masks from two grounding queries are reconciled in code to correct an undercount (c).

%% file: figs/harness_design.tex
\begin{figure}[!t]
    \centering
    \includegraphics[width=\linewidth]{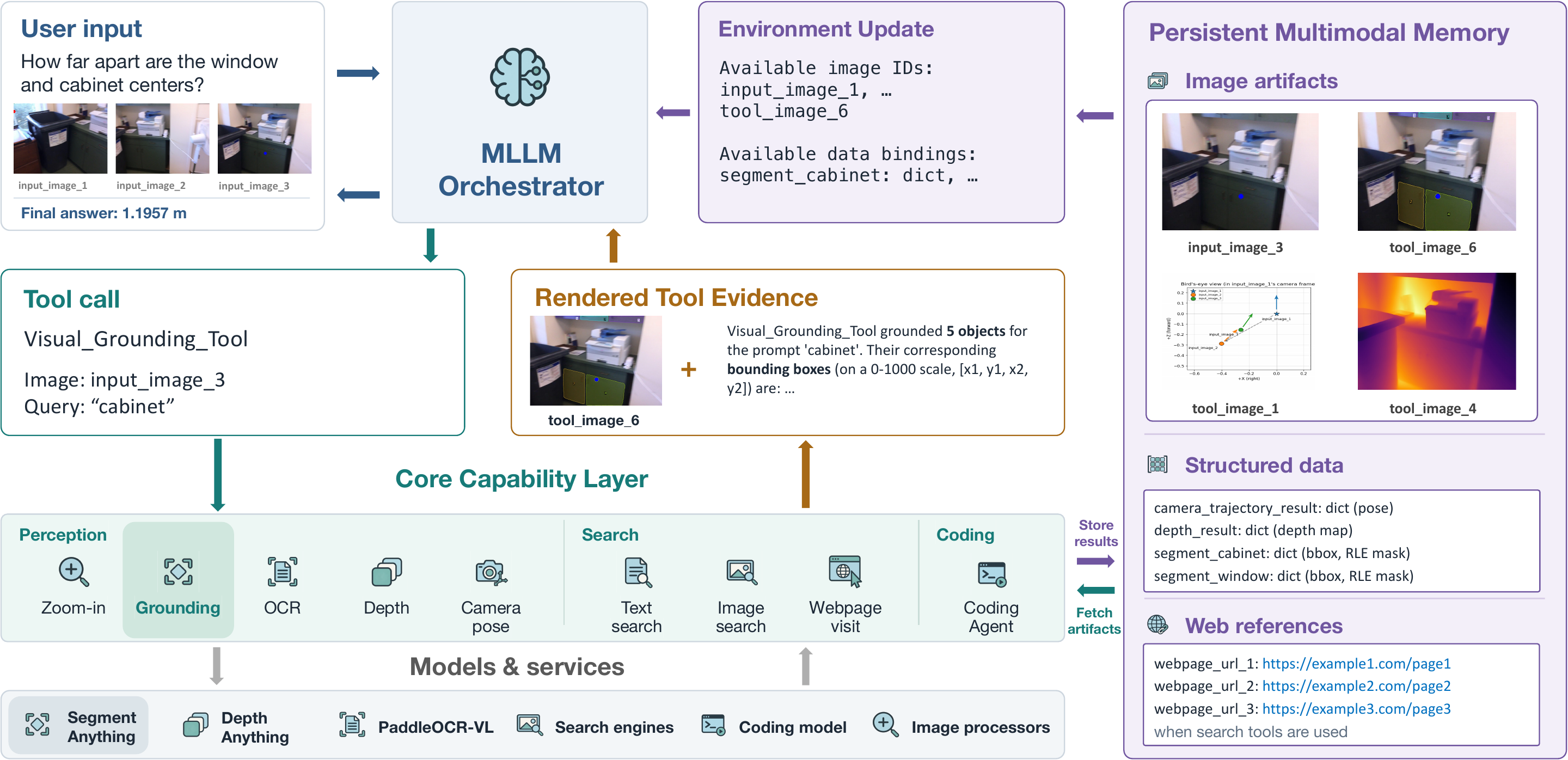}
    \caption{\textbf{Overview of \ours.} The MLLM orchestrator invokes capabilities through typed interfaces. The capability layer connects these interfaces to underlying models and services and renders their outputs into model-facing evidence: here, a labeled mask overlay accompanied by a textual description and normalized bounding boxes. Persistent multimodal memory retains image artifacts, structured data, and web references under reusable handles. An Environment Update advertises the available artifacts to the orchestrator, allowing subsequent calls to reuse earlier evidence, including through memory-grounded computation.}
    \label{fig:harness_design}
\end{figure}

%% file: figs/qualitative_examples.tex
\begin{figure*}[t]
    \centering
    \includegraphics[width=\linewidth]{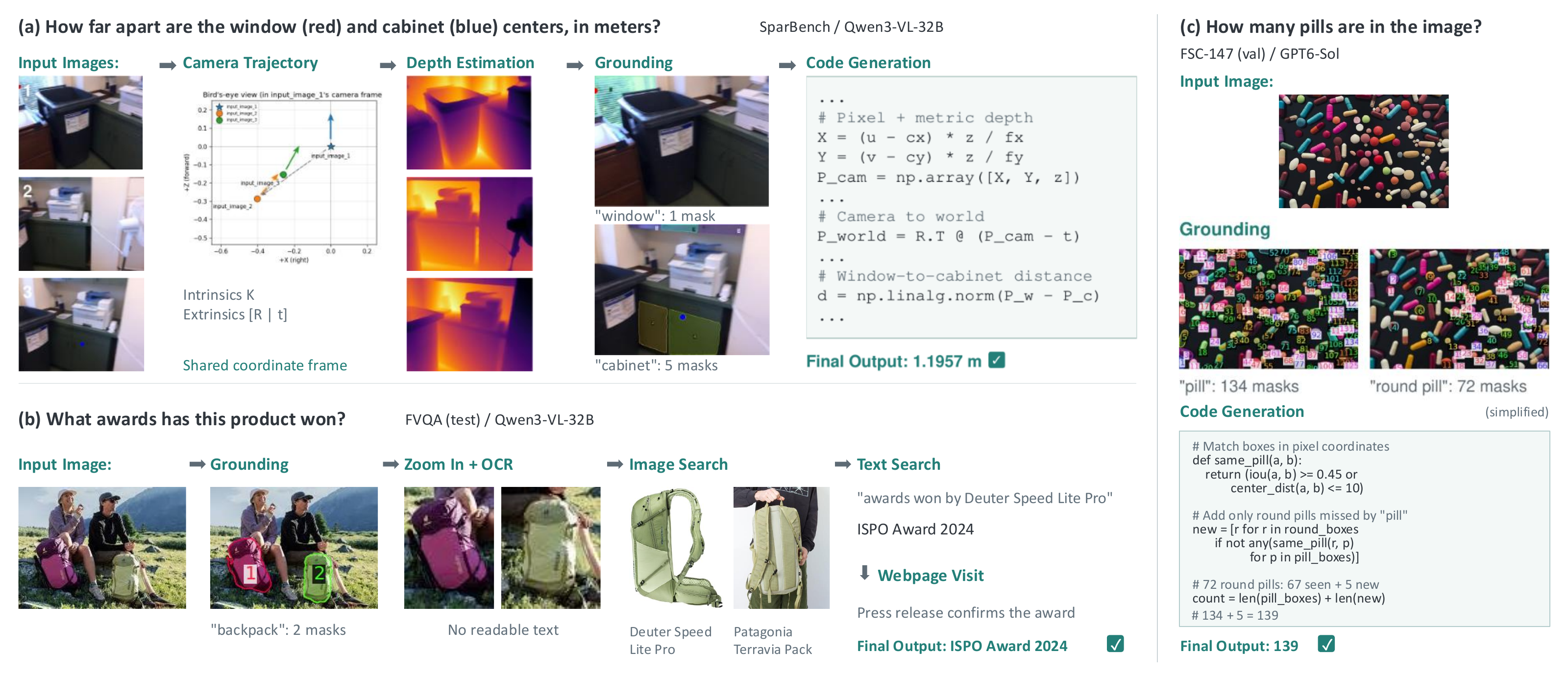}
    \caption{\textbf{Qualitative examples of \ours.}
    \textbf{(a)} Camera poses, depth, and grounding are combined in code to measure a 3D distance (\fullexample{ex:q558}).
    \textbf{(b)} Grounded crops are passed to OCR and image search, and the answer is verified on the web (\fullexample{ex:fvqa}).
    \textbf{(c)} Masks from two grounding queries are reconciled in code to reach the exact count (\fullexample{ex:fsc}).}
    \label{fig:qualitative}
\end{figure*}

%% file: text/5_experiment.tex
\section{Experiments}
\label{sec:experiments}

\subsection{Tasks, Models, and Baselines}

\paragraph{Tasks.} We evaluate \ours across four task domains spanning 22 datasets, as summarized in Table~\ref{tab:harness-combined} (splits, sizes, and scoring in Appendix~\ref{sec:appendix_benchmarks}):
\begin{itemize}[leftmargin=1.4em]
    \item \textbf{Grounded perception.} Five benchmarks targeting object counting and open-vocabulary visual grounding: CountQA \citep{tamarapalli2025countqa}, FSC-147 \citep{ranjan2021learning}, OdinW-13-All \citep{li2022grounded}, FSCD-147 \citep{nguyen2022few}, and FSCD-LVIS \citep{nguyen2022few}.
    \item \textbf{Visual search and deep research.} Six benchmarks drawn from a suite of 11 multimodal information-seeking tasks: FVQA \citep{mmsearchr1}, MMSearch \citep{mmsearch}, MMSearch-Plus \citep{mmsearchplus}, HLE-VL \citep{hle}, BrowseComp-VL \citep{browsecompvl}, and RealX-Bench \citep{hong2025deepeyesv2}; the remaining five are evaluated in Appendix~\ref{sec:appendix_mmdr_results}.
    \item \textbf{General VQA and hallucination.} Six benchmarks covering general visual question answering and hallucination robustness: RealWorldQA \citep{xai2024realworldqa}, SimpleVQA \citep{simplevqa}, HallusionBench \citep{guan2024hallusionbench}, FREAK-VQA \citep{yin2026freak}, FREAK-MCQ \citep{yin2026freak}, and V* \citep{wu2024v}.
    \item \textbf{Spatial reasoning.} Five benchmarks targeting spatial and embodied reasoning: MMSI-Bench \citep{yang2025mmsi}, ERQA \citep{team2025gemini}, EmbSpatial-Bench \citep{du2024embspatial}, SparBench \citep{zhang2026flatland}, and MindCube \citep{wang2025mindcube}.
\end{itemize}

\input{tables/harness_design_combined}

\paragraph{Models and baselines.} We instantiate \ours on four open-source models across different sizes---Qwen2.5-VL-7B-Instruct \citep{qwen25vl}, Qwen3-VL-8B-Instruct \citep{qwen3vl}, Qwen3-VL-32B-Instruct, and Kimi K3~\citep{team2026kimi}---and four proprietary models: GPT-5 \citep{gpt5}, GPT-6 Luna, GPT-6 Sol, and GPT-6 Astra \citep{gpt6}. Per-model sampling parameters can be found in Appendix~\ref{sec:appendix_impl}. Because the search APIs, the agent's tool-routing decisions, and the LLM judge are all stochastic, every Qwen3-VL result, for \ours and for all baselines, is the mean of \emph{three} independent runs with different random seeds, while results with proprietary backbones are single runs because of API cost (standard deviations in Appendix~\ref{sec:appendix_std}). In addition to the no-tool baseline (denoted \textbf{Base}), we compare \ours against two groups of baselines: \textbf{Training-based baselines}: AdaReasoner \citep{song2026adareasoner}, PixelReasoner \citep{wang2025pixel}, and DeepEyesV2 \citep{hong2025deepeyesv2}---harnesses that require fine-tuning the backbone to adapt it to the tool-use framework---evaluated with their released Qwen2.5-VL-7B checkpoints (Appendix Table~\ref{tab:harness-qwen25}; setup in Appendix~\ref{sec:appendix_baselines_trained}). \textbf{Training-free baselines}: Visual Sketchpad \citep{hu2024visual} and OctoTools \citep{lu2025octotools} are visual tool-use harnesses evaluated on the same Qwen3-VL backbones as \ours (detailed setup in Appendix~\ref{sec:appendix_baselines_free}). We additionally evaluate Visual Sketchpad and Codex on GPT-6 Luna.

\subsection{Main Results}

Across the evaluated backbones, \ours improves average performance over the underlying base model and outperforms all existing training-free harnesses (Table~\ref{tab:harness-combined}). On Qwen3-VL-8B, it raises the 22-dataset average from $34.2$ to $45.4$ ($+11.2$), exceeding the strongest training-free baseline, OctoTools ($31.8$), by $13.6$ points; on Qwen3-VL-32B, the average rises from $38.3$ to $50.9$ ($+12.6$). The gains extend to GPT-5 ($44.4\!\to\!55.7$), GPT-6 Sol ($52.7\!\to\!64.3$), and Kimi K3 ($56.4\!\to\!66.1$) (Appendix Table~\ref{tab:harness-kimi}). On the Qwen2.5-VL-7B backbone, \ours averages $33.45$ across the 22 datasets, surpassing the training-based baselines AdaReasoner ($25.29$), PixelReasoner ($28.02$), and DeepEyesV2 ($25.02$), without updating model weights (Appendix Table~\ref{tab:harness-qwen25}).

The GPT-6 results highlight that effective multimodal tool use depends on the harness. With GPT-6 Luna, Codex averages $43.8$, below the no-tool baseline ($45.8$), while \ours reaches $56.5$. This comparison shows that a coding-agent harness does not automatically improve multimodal reasoning. With GPT-6 Astra, \ours improves all 22 benchmarks and raises the average from $60.3$ to $70.2$, the highest of any backbone.

Grounded perception and visual search and deep research benefit substantially from dedicated vision experts and external retrieval, respectively. Gains on general VQA and hallucination and on spatial reasoning depend more on the backbone: Luna and Astra improve on every benchmark in these domains (Luna by less than one point on RealWorldQA, HallusionBench, and EmbSpatial), with average gains of $+6.5$/$+7.5$ on general VQA and hallucination and $+6.5$/$+5.2$ on spatial reasoning, respectively. This pattern is consistent with greater proficiency in using tools and reasoning over their visual evidence. The less consistent gains on Qwen3-VL-8B motivate the failure analysis and domain specialization in Secs.~\ref{sec:failure_analysis} and~\ref{sec:harness_optim}.

\subsection{Can More Reasoning Replace the Harness?}
\label{sec:effort}

\input{tables/effort_domains}

\ours makes several model calls per question, each carrying tool outputs and memory, so it spends more model tokens than prompting the base model directly. To test whether its gains come simply from extra computation, we run Base, Codex, Visual Sketchpad, and \ours with GPT-6 Luna at five reasoning-effort settings, from low to max, on the full evaluation splits of all 22 benchmarks ($26{,}229$ questions per setting). Figure~\ref{fig:teaser}c reports the 22-benchmark averages against model-token cost, and Table~\ref{tab:effort_domains} breaks them down by domain.

Additional effort yields limited gains for the other methods: from low to max effort, Base rises from 44.3 to 48.0 and Visual Sketchpad from 44.7 to 48.8, while Codex, the weakest method at low effort, rises from 36.9 to 48.6. The step from xhigh to max effort alone multiplies their model-token cost by 1.5--2.7. At max effort, all three remain well below \ours at medium effort (56.5), which spends at most 37\% of their model-token cost per question, and \ours at low effort (49.9) is already on par with their max-effort averages. The domain breakdown shows where the harness matters most. On grounded perception and on visual search and deep research, every effort setting of \ours, including low, scores above every setting of the other methods, and at max effort it leads the best of them by 15.6 and 12.4 points. On general VQA and hallucination and on spatial reasoning, the harness benefits from more reasoning: at low effort, \ours trails Base slightly (61.5 vs.\ 61.6 and 54.5 vs.\ 56.5), but from medium effort onward it scores above every setting of every other method in every domain, and at max effort it leads by 6.1 and 7.1 points. Raising the effort of \ours itself lifts its average from 56.5 to 60.1. Acquiring and composing multimodal evidence therefore complements stronger reasoning: additional computation remains useful, but does not eliminate the benefit of the harness within the tested budgets. Costs cover model tokens, including reasoning tokens, and exclude tool execution and judging; Appendix~\ref{sec:appendix_effort_scaling} gives the setup, per-benchmark scores, token usage, and how often Codex executes code.

\subsection{Ablation Study}

Table~\ref{tab:ablation} examines the two mechanisms introduced in Sec.~\ref{sec:harness_design}: the \emph{capability layer}, which combines task-level tool interfaces with model-facing evidence rendering, followed by \emph{persistent multimodal memory}, which makes the resulting artifacts addressable and reusable across turns. Starting from the no-tool baseline, we add raw tool access, then the capability layer, and finally memory, using a textual-memory variant as an intermediate comparison (details in Appendix~\ref{sec:appendix_ablation}).

\input{tables/ablation}

Raw tool access raises the overall average by $+3.8$ points at 8B and $+4.3$ at 32B, with the largest gains in grounded perception. Adding the capability layer restores full tool schemas and renders heterogeneous outputs into concise observations, yielding another $+3.4$ and $+3.2$ points. These gains support the joint value of task-level tool interfaces and model-facing evidence rendering; the comparison does not separate their individual contributions.

Adding textual memory enables structured data bindings, web references, and Environment Updates, but represents tool-generated images only through descriptions, without reusable image handles. This adds $+0.7$ points at 8B and $+0.3$ at 32B. Completing persistent multimodal memory with visual attachments and reusable image handles adds a further $+3.1$ and $+4.5$, bringing the full harness to $47.1$ and $52.5$. The largest gains from this final step occur in visual search and deep research ($+6.2$/$+3.5$), with additional gains in spatial reasoning ($+2.8$/$+3.1$). The results support the method's complementary mechanisms: interpretable tool observations, followed by retained multimodal artifacts that can be revisited and composed across turns.

\subsection{Failure Analysis}
\label{sec:failure_analysis}

We use Kimi K3 to critique $9{,}893$ failed Qwen3-VL-8B trajectories across 22 benchmarks. Given the question, answer, images, trajectory, and toolbox, the critic assesses each tool call and identifies the trajectory's primary failure cause and proposes a fix. Its labels agree with human annotations on $92\%$ of 100 sampled failures. The protocol, coverage, and taxonomy are detailed in Appendix~\ref{sec:appendix_critique}.

Surprisingly, we find that failure patterns differ across domains (Fig.~\ref{fig:failure_analysis}a). Despite substantial overall gains, grounded perception remains challenging in crowded images with more than 25 objects, where missed or merged instances limit exhaustive detection. Such perception errors account for the dominant ``incorrect tool output'' category ($75\%$ of failures in this domain). Failures in visual search and deep research most often involve an omitted tool call ($49\%$), while failures in spatial reasoning and in general VQA and hallucination involve both omitted tool calls and reasoning errors, consistent with our hypothesis that the model has limited proficiency in spatial reasoning and tool use. Failures involving unusable or ignored correct tool outputs are uncommon ($3$--$8\%$ across domains), suggesting that the capability layer generally presents retrieved evidence in a usable form.

Failure causes alone do not determine how to specialize the harness: an omitted call may require clearer instructions or a new composition of tools. We therefore combine the cause and proposed fix into remedy channels (Fig.~\ref{fig:failure_analysis}b). Constructing new task-specific tools can make evidence interpretation explicit (A), enforce systematic verification, filtering of tool results (B), or supply a missing tool composition (C). In contrast, harness text changes (Y)—edits to the system prompt, tool descriptions, and output messages—guide the model to call existing tools when needed and use their results appropriately. The remaining channel (Z) covers cases without a proposed remedy. These diagnostic recommendations guide optimization rather than predict recoverable accuracy, since the critique covers only failed trajectories. Exact mapping rules and per-dataset shares are in Appendix~\ref{sec:appendix_channels}.

\subsection{Harness Optimization}
\label{sec:harness_optim}

\input{figs/failure_analysis}

When adapting \ours to a custom domain, users can use failures on their own data to identify which parts of the harness need refinement. We provide two coding-agent-driven workflows for specializing the general-purpose harness without updating model weights. \textbf{Harness search} edits system prompts, tool descriptions, and evidence-rendering templates, selecting candidates through repeated evaluation. \textbf{Tool generation} implements and refines task-specific tools by composing existing capabilities (algorithms in Appendix~\ref{sec:appendix_optim}). To compare them, we select four datasets with contrasting remedy profiles (Fig.~\ref{fig:failure_analysis}c): harness text changes (Y) dominate ERQA ($71\%$) and RealWorldQA ($56\%$), whereas task-specific tools (A+B+C) dominate SparBench ($75\%$) and FREAK-VQA ($62\%$). We run both workflows on all four datasets with Qwen3-VL-8B, testing whether their relative benefits align with these profiles.

The results follow this distinction (Fig.~\ref{fig:failure_analysis}d). Harness search improves ERQA and RealWorldQA by $+4.1$ and $+3.7$ points, compared with $+0.2$ and $+0.1$ from tool generation. Conversely, tool generation improves SparBench and FREAK-VQA by $+8.2$ and $+4.5$, while harness search yields $-0.5$ and $-1.0$. The generated tools compute metric 3D relations on SparBench and make visibility decisions explicit on FREAK-VQA, matching the missing-composition and evidence-interpretation needs identified by the critique (Table~\ref{tab:appendix_generated_tools}). These results suggest a practical approach to adaptation on custom data: diagnose the prevailing failures, then use harness search to improve the use of existing capabilities or tool generation to add the task-specific operations the domain requires.

%% file: tables/harness_design_combined.tex
\begin{table}[t]
\centering
\setlength{\tabcolsep}{3pt}
\renewcommand{\arraystretch}{1.45}
\setlength{\aboverulesep}{0pt}
\setlength{\belowrulesep}{0pt}
\definecolor{lightblue}{HTML}{A4C2F4}
\definecolor{lightorange}{HTML}{D9EAD3}
\definecolor{lightpink}{HTML}{F5CBCC}
\definecolor{lightcream}{HTML}{FFF3CC}
\definecolor{lightred}{HTML}{F4CCCC}
\newcommand{\effortlabel}[1]{{\renewcommand{\arraystretch}{1}\begin{tabular}[t]{@{}c@{}}\textbf{#1}\\[-1pt]{\footnotesize (medium)}\end{tabular}}}
\resizebox{\textwidth}{!}{%
\begin{tabular}{cl|cccc|cccc|cccc|cc}
\toprule
\multicolumn{2}{c|}{\rule{0pt}{5.5ex}}
 & \multicolumn{4}{c|}{\raisebox{10pt}[0pt][0pt]{\textbf{Qwen3-VL-8B}}}
 & \multicolumn{4}{c|}{\raisebox{10pt}[0pt][0pt]{\textbf{Qwen3-VL-32B}}}
 & \multicolumn{4}{c|}{\raisebox{10pt}[0pt][0pt]{\effortlabel{GPT-6 Luna}}}
 & \multicolumn{2}{c}{\raisebox{10pt}[0pt][0pt]{\effortlabel{GPT-6 Astra}}} \\
\textbf{Domain} & \textbf{Dataset}
 & \textbf{Base} & \textbf{OT} & \textbf{VS} & \textbf{Ours}
 & \textbf{Base} & \textbf{OT} & \textbf{VS} & \textbf{Ours}
 & \textbf{Base} & \textbf{VS} & \textbf{CX} & \textbf{Ours}
 & \textbf{Base} & \textbf{Ours} \\
\midrule
& CountQA         & 32.00 & 29.06 & 19.83 & \textbf{48.56} & 44.70 & 41.75 & 32.85 & \textbf{64.14} & 55.17 & 52.68 & 58.25 & \textbf{66.95} & 82.00 & \textbf{\underline{86.40}} \\
& FSC-147         & 15.32 & 11.26 & 9.41 & \textbf{37.71} & 18.97 & 12.44 & 12.35 & \textbf{39.42} & 28.23 & 30.25 & 28.62 & \textbf{41.52} & 44.01 & \textbf{\underline{46.11}} \\
& OdinW-13-All    & 39.80 & 24.60 & 31.77 & \textbf{50.40} & 42.39 & 44.33 & 36.10 & \textbf{59.13} & 36.74 & 43.84 & 36.14 & \textbf{53.56} & 50.46 & \textbf{\underline{60.45}} \\
& FSCD-147        & 1.79 & 0.93 & 1.41 & \textbf{23.52} & 2.88 & 2.55 & 2.30 & \textbf{28.45} & 4.24 & 6.11 & 2.86 & \textbf{25.74} & 15.49 & \textbf{\underline{30.33}} \\
\multirow{-5}{*}{\rotatebox[origin=c]{90}{\shortstack[c]{Grounded\\Perception}}}
 & FSCD-LVIS       & 1.47 & 1.36 & 2.49 & \textbf{27.37} & 1.79 & 3.91 & 4.57 & \textbf{31.63} & 2.44 & 8.65 & 2.23 & \textbf{30.55} & 10.55 & \textbf{\underline{36.02}} \\
\midrule
& MMSearch        & 14.04 & 9.94 & 14.62 & \textbf{51.46} & 20.47 & 18.13 & 18.71 & \textbf{62.57} & 49.12 & 46.78 & 49.12 & \textbf{64.91} & 57.31 & \textbf{\underline{74.27}} \\
& MMSearch-Plus   & 1.29 & 3.54 & 4.18 & \textbf{14.79} & 2.57 & 3.54 & 5.14 & \textbf{17.68} & 10.93 & 4.50 & 8.68 & \textbf{26.37} & 27.97 & \textbf{\underline{53.70}} \\
& RealX-Bench      & 22.16 & 26.80 & 22.16 & \textbf{34.71} & 24.74 & 26.80 & 29.90 & \textbf{40.21} & 31.96 & 35.05 & 34.02 & \textbf{45.36} & 44.33 & \textbf{\underline{53.09}} \\
& FVQA            & 28.39 & 32.67 & 28.67 & \textbf{55.00} & 32.67 & 35.00 & 33.33 & \textbf{62.67} & 52.89 & 47.72 & 48.06 & \textbf{68.78} & 60.67 & \textbf{\underline{74.67}} \\
& HLE-VL          & 6.36 & 7.58 & 4.24 & \textbf{10.30} & 10.00 & 9.09 & 6.67 & \textbf{11.52} & 27.58 & 20.30 & 22.42 & \textbf{27.88} & 49.09 & \textbf{\underline{56.06}} \\
\multirow{-6}{*}{\rotatebox[origin=c]{90}{\shortstack[c]{Visual Search and\\Deep Research}}}
 & BC-VL           & 21.05 & 15.79 & 10.53 & \textbf{32.33} & 28.82 & 19.55 & 15.04 & \textbf{33.08} & 38.60 & 39.85 & 38.85 & \textbf{51.88} & 46.12 & \textbf{\underline{63.91}} \\
\midrule
& RealWorldQA     & 71.37 & 69.67 & 69.15 & \textbf{72.67} & 78.82 & 76.21 & 74.38 & \textbf{80.52} & 80.65 & 80.26 & 79.35 & \textbf{80.78} & 87.58 & \textbf{\underline{92.16}} \\
& SimpleVQA       & 54.33 & 46.27 & 42.81 & \textbf{70.00} & 59.00 & 51.75 & 49.58 & \textbf{70.33} & 55.46 & 55.06 & 54.47 & \textbf{72.20} & 62.17 & \textbf{\underline{76.10}} \\
& HallusionBench  & 57.76 & 55.66 & 54.01 & \textbf{59.07} & 61.66 & 53.82 & 59.38 & \textbf{63.40} & 67.25 & 66.99 & 63.82 & \textbf{67.76} & 78.75 & \textbf{\underline{83.78}} \\
& FREAK-VQA       & \textbf{44.68} & 38.80 & 33.04 & 37.05 & 44.06 & 45.06 & 37.92 & \textbf{48.19} & 41.68 & 46.06 & 39.17 & \textbf{51.94} & 63.20 & \textbf{\underline{73.34}} \\
& FREAK-MCQ       & 36.42 & 31.39 & 35.31 & \textbf{36.82} & 44.06 & 38.73 & 38.33 & \textbf{45.77} & 49.40 & 51.31 & 44.37 & \textbf{56.34} & 76.26 & \textbf{\underline{83.90}} \\
\multirow{-6}{*}{\rotatebox[origin=c]{90}{\shortstack[c]{General VQA and\\ Hallucination}}}
 & V*              & \textbf{83.24} & 73.30 & 71.73 & \textbf{83.24} & 85.34 & 76.44 & 85.86 & \textbf{88.48} & 85.86 & 85.86 & 80.63 & \textbf{90.05} & 93.72 & \textbf{\underline{97.38}} \\
\midrule
& MMSI-Bench      & 31.60 & 27.90 & 28.10 & \textbf{33.00} & 35.00 & 36.60 & 32.40 & \textbf{38.00} & 38.30 & 40.00 & 38.20 & \textbf{47.70} & 65.20 & \textbf{\underline{73.90}} \\
& ERQA             & \textbf{44.00} & 41.25 & 35.50 & 43.00 & 44.75 & 46.25 & \textbf{48.25} & 48.00 & 59.75 & 59.75 & 54.25 & \textbf{61.00} & 76.50 & \textbf{\underline{81.25}} \\
& EmbSpatial      & 78.37 & 73.71 & 75.74 & \textbf{78.90} & \textbf{80.54} & 76.15 & 78.98 & 80.00 & 78.98 & 77.50 & 75.36 & \textbf{79.48} & 82.99 & \textbf{\underline{85.14}} \\
& SparBench       & 39.55 & 38.32 & 34.63 & \textbf{43.73} & \textbf{46.98} & 45.29 & 44.69 & 46.76 & 52.10 & 46.55 & 46.16 & \textbf{58.31} & 63.68 & \textbf{\underline{68.17}} \\
\multirow{-5}{*}{\rotatebox[origin=c]{90}{\shortstack[c]{Spatial \\ Reasoning}}}
 & MindCube        & 28.36 & 39.71 & 41.44 & \textbf{55.48} & 32.40 & 45.96 & 44.13 & \textbf{60.19} & 59.33 & 57.02 & 58.08 & \textbf{74.42} & 87.79 & \textbf{\underline{93.65}} \\
\midrule
{}& {Average}   & 34.24 & 31.80 & 30.49 & \textbf{45.41} & 38.30 & 36.79 & 35.95 & \textbf{50.92} & 45.76 & 45.55 & 43.78 & \textbf{56.52} & 60.27 & \textbf{\underline{70.17}} \\
\bottomrule
\end{tabular}%
}
\caption{\textbf{Main results across four task domains and 22 datasets.} Results are grouped by MLLM backbone. \textbf{Base} denotes the backbone without tool use; for harness, \textbf{OT}, \textbf{VS}, and \textbf{CX} denote OctoTools, Visual Sketchpad, and Codex, respectively; \textbf{Ours} denotes OpenVLHarness. GPT-6 Sol, GPT-5, and Kimi K3 results are in Table~\ref{tab:harness-kimi}. Bold indicates the best result within each backbone group, and underlining indicates the best result across all shown backbones. Qwen3-VL results are means of three runs (standard deviations in Table~\ref{tab:appendix_std_main}); GPT-6 results are single runs. OdinW-13-All asks for all categories of an image's source dataset in one query.}
\label{tab:harness-combined}
\end{table}

%% file: tables/effort_domains.tex
\begin{table}[t]
\centering
\fontsize{8}{9.6}\selectfont  
\setlength{\tabcolsep}{2pt}
\begin{tabularx}{\textwidth}{@{}l@{\hspace{5pt}}|*{4}{>{\centering\arraybackslash}X}|*{4}{>{\centering\arraybackslash}X}|*{4}{>{\centering\arraybackslash}X}|*{4}{>{\centering\arraybackslash}X}@{}}
\toprule
\multirow{2}{*}{\textbf{Reasoning effort}} & \multicolumn{4}{c|}{\textbf{medium}} & \multicolumn{4}{c|}{\textbf{high}} & \multicolumn{4}{c|}{\textbf{xhigh}} & \multicolumn{4}{c}{\textbf{max}} \\
 & Base & CX & VS & Ours & Base & CX & VS & Ours & Base & CX & VS & Ours & Base & CX & VS & Ours \\
\midrule
Grounding & 25.4 & 25.6 & 28.3 & \textbf{43.7} & 26.5 & 26.3 & 29.3 & \textbf{44.6} & 27.0 & 27.1 & 29.3 & \textbf{45.2} & 27.2 & 30.0 & 29.6 & \textbf{45.6} \\
VS \& DR & 35.2 & 33.5 & 32.4 & \textbf{47.5} & 35.3 & 35.4 & 34.6 & \textbf{50.3} & 37.4 & 36.7 & 35.6 & \textbf{50.6} & 37.8 & 40.3 & 36.9 & \textbf{52.7} \\
General VQA & 63.4 & 60.3 & 64.3 & \textbf{69.8} & 64.4 & 61.8 & 65.1 & \textbf{71.5} & 64.7 & 63.0 & 66.6 & \textbf{71.5} & 64.5 & 63.5 & 66.5 & \textbf{72.6} \\
Spatial & 57.7 & 54.4 & 56.2 & \textbf{64.2} & 59.3 & 57.0 & 57.1 & \textbf{65.3} & 58.7 & 58.4 & 58.5 & \textbf{65.9} & 61.5 & 59.1 & 61.4 & \textbf{68.5} \\
\midrule
\textbf{Average} & 45.8 & 43.8 & 45.5 & \textbf{56.5} & 46.7 & 45.5 & 46.9 & \textbf{58.2} & 47.3 & 46.6 & 47.8 & \textbf{58.6} & 48.0 & 48.6 & 48.8 & \textbf{60.1} \\
\$ / 1k questions & \textbf{0.58} & 1.29 & 1.89 & \underline{0.92} & \textbf{0.85} & 1.47 & 2.27 & \underline{1.15} & \textbf{1.18} & 1.71 & 2.72 & \underline{1.45} & 3.17 & \textbf{2.52} & 5.15 & \underline{2.99} \\
\bottomrule
\end{tabularx}
\caption{\textbf{Reasoning effort versus harness, by domain.} GPT-6 Luna alone (Base) and with Codex (CX), Visual Sketchpad (VS), or \ours (Ours) on the full evaluation splits of our $22$ benchmarks. Rows average each domain's benchmarks: Grounding (grounded perception, $5$), VS~\& DR (visual search and deep research, $6$), General VQA (general VQA and hallucination, $6$), and Spatial (spatial reasoning, $5$). \emph{\$ / 1k questions} is the model-token cost at Standard-tier prices, including reasoning tokens and excluding tool execution and judging, averaged over the $22$ benchmarks. At each effort, bold marks the best score, and in the cost row the cheapest method, with the second cheapest underlined. Low effort is in Appendix Table~\ref{tab:luna_effort_scaling}.}
\label{tab:effort_domains}
\end{table}

%% file: tables/ablation.tex
\begin{table*}[t]
\centering
\small
\setlength{\tabcolsep}{4.5pt}
\resizebox{\textwidth}{!}{%
\begin{tabular}{l*{10}{c}}
\toprule
\multirow{2}{*}{\textbf{Variant}}
& \multicolumn{2}{c}{\textbf{Grounding}}
& \multicolumn{2}{c}{\textbf{VS \& DR}}
& \multicolumn{2}{c}{\textbf{General VQA}}
& \multicolumn{2}{c}{\textbf{Spatial}}
& \multicolumn{2}{c}{\textbf{Overall}} \\
\cmidrule(lr){2-3}
\cmidrule(lr){4-5}
\cmidrule(lr){6-7}
\cmidrule(lr){8-9}
\cmidrule(lr){10-11}
& \textbf{8B} & \textbf{32B}
& \textbf{8B} & \textbf{32B}
& \textbf{8B} & \textbf{32B}
& \textbf{8B} & \textbf{32B}
& \textbf{8B} & \textbf{32B} \\
\midrule

Baseline (no tool)
& 27.48 & 31.72 & 18.40 & 23.34 & 57.97 & 62.16 & 44.38 & 47.93 & 37.06 & 41.29 \\

+ Tools
& 41.70 & 43.30 & 21.44 & 24.89 & 52.41 & 61.41 & 47.83 & 52.84 & 40.84 & 45.61 \\

+ Capability layer
& 46.73 & 46.20 & 25.97 & 32.74 & 56.05 & 63.63 & 48.16 & 52.63 & 44.23 & 48.80 \\

+ Textual memory
& \textbf{47.27} & 45.74 & 26.92 & 34.43 & 57.54 & 64.84 & 47.98 & 51.46 & 44.93 & 49.12 \\

\textbf{OpenVLHarness}
& 45.46 & \textbf{51.74}
& \textbf{33.10} & \textbf{37.96}
& \textbf{59.81} & \textbf{66.12}
& \textbf{50.82} & \textbf{54.59}
& \textbf{47.14} & \textbf{52.51} \\

\bottomrule
\end{tabular}%
}
\caption{\textbf{Ablation study of OpenVLHarness across model scales.} We progressively add raw tool access, capability layer, textual memory, and multimodal memory to the no-tool baseline. Results are reported across four domains and overall for Qwen3-VL 8B and 32B models, averaged over three runs; standard deviations are provided in Table~\ref{tab:appendix_std_ablation}. Bold marks the best result in each column.}
\label{tab:ablation}
\end{table*}

%% file: figs/failure_analysis.tex
\begin{figure}[t]
\centering
\includegraphics[width=\textwidth]{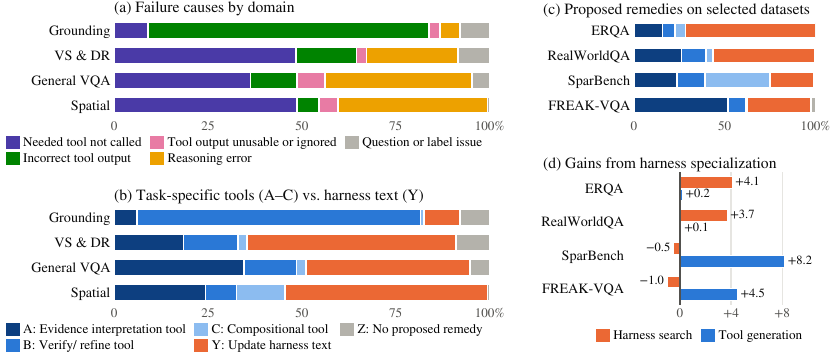}
\caption{\textbf{Failure diagnosis and harness specialization with Qwen3-VL-8B.} We critique failure trajectories using Kimi K3. \textbf{(a)} Distribution of failure causes across four domains. \textbf{(b)} Proposed remedies: grouped into creating task-specific tools (A–C), updating harness text (Y), and no proposed remedy (Z). \textbf{(c)} Four datasets selected for contrasting remedy profiles: Y (orange) dominates ERQA and RealWorldQA, while A–C (blue) dominates SparBench and FREAK-VQA. \textbf{(d)} Gains from harness search and tool generation over the unoptimized harness. Harness search is more effective on datasets dominated by text-level changes, while tool generation yields larger gains when task-specific tools are preferred.}

\label{fig:failure_analysis}
\end{figure}

%% file: text/2_related_work.tex
\section{Related Work}

\paragraph{Visual Tool-Use Baselines.}
Early visual tool-use work treats the problem as program synthesis, compiling a question into a sequential plan that invokes vision tools in a pipeline to produce an answer~\citep{suris2023vipergpt, gupta2023visual, kamali2509neptune, zhao2025pyvision, wu2023visual, yang2023mm, gao2023assistgpt, castrejon2024hammr, zhang2026vipact, zhou2024image, menon2024whiteboard, chen2024genome, you2023idealgpt}; because these plans are fixed in place, they cannot be adjusted based on tool execution results. A later line adopts multi-turn frameworks that interleave reasoning with visual feedback, but depends on brittle, task-specific framework designs that require extensive training to work~\citep{ma2025latte, zhou2025reinforced, song2026adareasoner, wang2025pixel, zheng2025deepeyes, hong2025deepeyesv2, su2025openthinkimg, wu2025vtool, zhao2026pyvision}, and many domain-specific visual tool-use frameworks fall into the same category~\citep{chen2025spacetools, mmsearchr1, deepmmsearchr1, browsecompvl, vdr, mmdeepresearch, yang2026multimodal, du2026towards, wang2025geovista}. Closest to our setting, Visual Sketchpad~\citep{hu2024visual} and OctoTools~\citep{lu2025octotools} are training-free tool-use frameworks built around manually designed tool interfaces and execution strategies. In contrast, we focus on making the harness itself more adaptable, so its evidence processing and tool composition can be specialized as tools, MLLMs, and target tasks evolve.

\paragraph{Working Memory for Tool-Using Agents.} Language agents maintain intermediate information to support reasoning and action across multiple steps. Context management approaches further separate compact model-visible state from more detailed external information, enabling the agent to selectively recover relevant content when needed~\citep{packer2023memgpt, DBLP:conf/icml/LeeCFCF24}. Tool-using agents have also introduced explicit mechanisms for passing intermediate results across actions. ReWOO~\citep{DBLP:journals/corr/abs-2305-18323} represents tool evidence through referenceable variables, while CodeAct~\citep{DBLP:conf/icml/WangCY0L0J24} maintains intermediate results as variables in an executable interpreter. HuggingGPT~\citep{DBLP:conf/nips/0001ST00Z23} and LLMCompiler~\citep{DBLP:conf/icml/KimMTLMKG24} go further in using symbolic references to intermediate tool outputs that are resolved when executing dependent downstream calls. Building on these ideas, \ours extends them to the multimodal setting, maintaining a unified registry of typed artifacts spanning images, structured data, and web evidence. It exposes compact model-facing handles while preserving the underlying artifacts for later inspection, composition, and reuse through modality-appropriate downstream tools.


%% file: text/7_conclusion.tex
\section{Conclusion}

We presented \ours, a general harness for real-world visual agentic tool use built around a capability layer and persistent multimodal memory. The capability layer combines task-level tool interfaces with evidence rendering, enabling the orchestrator to invoke heterogeneous tools and inspect their outputs. Persistent multimodal memory retains the corresponding artifacts for reuse across turns, including computation over stored data through a Python Coding Agent. Together, these mechanisms support reasoning that acquires and composes multimodal evidence. Evaluation across 22 benchmarks and eight backbones demonstrates broad gains, while component ablations show the incremental benefits of the capability layer and multimodal memory. Harness search and tool generation further specialize the system on selected domains without updating the underlying MLLM weights. We release the harness, tools, evaluation protocols, and optimization workflows to support further work on multimodal tool-use agents.

%% file: text/9_acknowledgements.tex
\ificlrfinal
\section*{Acknowledgments}
We sincerely thank Wenbo Hu, Xueqing Wu, and Zixian Ma for helpful discussions and feedback on this work. This work was supported by funding from Apple and U.S. ONR grant N00014-23-1-2780. Chang and Peng are supported in part by a grant from DARPA to the Simons Institute for the Theory of Computing.
\fi


%% file: text/appendix.tex
\section*{Appendix}

\input{text/appendix_tool_use}

\section{Additional Results and Analysis}
\label{sec:appendix_results}

This section reports results with additional backbones (GPT-5, GPT-6 Sol, and Kimi K3), gives details of the ablation in Table~\ref{tab:ablation}, compares \ours with training-based harnesses, analyzes grounded perception across object counts, extends the visual search and deep research results to $11$ information-seeking benchmarks, and tests whether additional reasoning can replace the harness.

\subsection{Additional Backbones}
\label{sec:appendix_kimi}

Table~\ref{tab:harness-kimi} evaluates \ours with GPT-5~\citep{gpt5}, with GPT-6 Sol~\citep{gpt6}, and with Kimi K3~\citep{team2026kimi}, an open-weight mixture-of-experts model with native visual understanding, on the same $22$ datasets as Table~\ref{tab:harness-combined}, using the same tools, memory, prompts, and scoring. With GPT-6 Sol, \ours improves every one of the $22$ benchmarks, raising the average from $52.7$ to $64.3$.

\input{tables/harness_design_kimi}

\subsection{Ablation Study Details}
\label{sec:appendix_ablation}

The ablation in Table~\ref{tab:ablation} isolates the contribution of each harness component by building a strict superset ladder: every rung adds exactly one mechanism to the one below it, so the gap between adjacent rungs measures that mechanism's marginal value.

\paragraph{Variants.} The five rungs are:

\begin{enumerate}[topsep=0.3em,itemsep=0.2em,leftmargin=1.4em]
    \item \textbf{Baseline (no tool).} The backbone model prompted directly with the question image and no tool access, establishing the no-harness floor.
    \item \textbf{+ Tool.} The nine tools are available, but their payloads arrive as raw dumps, their schema is cut to a minimal signature (name, argument names and types, one clause each), and nothing persists between turns: no environment updates, no image labels, no data bindings, no URL references. Tool-generated images are replaced by a short stub.
    \item \textbf{+ Capability layer.} Adds the capability layer of Sec.~\ref{sec:setup_base}: each tool's raw payload is replaced by a per-tool post-processor that condenses it into a concise, model-readable summary (e.g.\ grounding boxes on a 0--1000 grid, reranked search results with inline citations, depth/zoom captions), and the full tool schema is restored so the model is told when to reach for each tool and how to phrase its arguments. No memory is present---nothing is carried across turns.
    \item \textbf{+ Textual memory.} Adds every memory mechanism that is addressed by a \emph{text} handle: per-turn Environment Updates that re-advertise all live state, \texttt{webpage\_url\_N} references that let the agent revisit a URL, and out-of-context data bindings that the Python Coding Agent can read as typed variables. Tool-generated images are replaced by a VLM-written text description (up to 160 tokens) conditioned on the original question. Image references remain absent: tool-produced images cannot be named, stored, or passed to a later tool call.
    \item \textbf{OpenVLHarness (full).} Adds multimodal image memory: every tool-produced image is labeled (\texttt{tool\_generated\_image\_N}), deduplicated by content, and kept addressable. The model selects which image to pass to each tool call by its identifier, enabling compositions such as grounding $\to$ zoom $\to$ OCR $\to$ search.
\end{enumerate}

\paragraph{Implementation.} All five variants share the same nine-tool toolset, backbone weights, and decoding configuration (temperature~$0.7$, top-$p$~$0.8$, top-$k$~$20$, repetition penalty~$1.0$, presence penalty~$1.5$, up to $12{,}800$ generated tokens per turn, $15$ tool-calling iterations, $168$K-token context cap). They differ only in the \emph{ablation preset} applied to the memory stack: a configuration block that switches individual mechanisms (environment updates, image identifiers, per-tool post-processors, data bindings, URL references) on or off. Each preset is a strict superset of the one below it, validated at initialization so a typo fails the run rather than silently selecting the wrong baseline. Table~\ref{tab:ablation_knobs} summarizes the knob settings for each variant.

\begin{table}[!htbp]
\centering
\small
\setlength{\tabcolsep}{4pt}
\begin{tabular}{@{}lcccccc@{}}
\toprule
\textbf{Variant} & \textbf{Tool schema} & \textbf{Post-proc.} & \textbf{Env update} & \textbf{Bindings} & \textbf{URL refs} & \textbf{Image refs} \\
\midrule
Baseline       & ---      & ---    & ---  & ---  & ---  & ---  \\
+ Tool         & minimal  & raw    & off  & off  & off  & off  \\
+ Capability layer & full     & custom & off  & off  & off  & off  \\
+ Textual memory & full     & custom & on   & on   & on   & off  \\
OpenVLHarness  & full     & custom & on   & on   & on   & on   \\
\bottomrule
\end{tabular}
\caption{Ablation knob settings for each variant in Table~\ref{tab:ablation}. ``Post-proc.'' refers to per-tool output processors (custom vs.\ raw dumps). ``Env update'' is the per-turn Environment Update message. ``Bindings'' are out-of-context typed data bindings. ``URL refs'' and ``Image refs'' are symbolic identifiers for web pages and tool-generated images, respectively.}
\label{tab:ablation_knobs}
\end{table}

\paragraph{Evaluation setup.} Each variant is evaluated on 22 datasets spanning all four domains in Table~\ref{tab:harness-combined}. For efficiency, datasets with more than 500 samples are subsampled to 500 examples with a fixed random seed ($42$) shared across all variants, so every rung sees exactly the same instances. All reported numbers are means of three independent runs with different sampling seeds. Domain averages in Table~\ref{tab:ablation} are macro-averages over the datasets in each domain.

\subsection{Training-Based Baselines}
\input{tables/harness_design_qwen25}
Table~\ref{tab:harness-qwen25} compares \ours with the three training-based harnesses on Qwen2.5-VL-7B-Instruct, the base model all three are fine-tuned from (setup in Appendix~\ref{sec:appendix_baselines_trained}). Averaged over the $22$ datasets, \ours reaches $33.45$, ahead of PixelReasoner ($28.02$), AdaReasoner ($25.29$), and DeepEyesV2 ($25.02$), without updating any weights.

\paragraph{Where training helps and where it does not.} The margin is largest on grounded perception ($23.05$ vs.\ at most $10.41$) and on visual search and deep research ($19.91$ vs.\ at most $15.59$). The training-based harnesses are optimized for a fixed tool set and answer style: they rarely produce the dense, well-formed box lists that FSCD and OdinW require, and none of them retrieves web evidence in our setup, so knowledge-intensive search questions fall back on the 7B model's parametric knowledge. On general VQA and hallucination and on spatial reasoning, which resemble their training distributions, the gap narrows ($50.58$ vs.\ $49.32$ and $39.54$ vs.\ $39.37$), and the trained models are best on a few perception-centric datasets such as RealWorldQA and V*, where PixelReasoner's learned zoom-in behavior pays off. Training a model into one tool protocol thus helps in-distribution but transfers poorly, whereas a training-free harness can cover all four domains with the same weights.

\paragraph{The backbone as orchestrator.} Absolute scores with \ours on Qwen2.5-VL-7B are well below those on Qwen3-VL (Table~\ref{tab:harness-combined}). Because the harness relies on the backbone to plan tool calls and to follow output conventions, a weaker orchestrator limits its benefit; on OdinW-13-All, for example, \ours reaches only $7.80$ AP with Qwen2.5-VL-7B versus $50.40$ with Qwen3-VL-8B. The harness therefore benefits directly from stronger backbones, with no retraining when a new backbone is adopted.

\subsection{Grounded Perception Across Object Counts}
\label{sec:appendix_cg_results}

Most counting and grounding benchmarks rarely test models at high object counts, which can make headline scores misleading. To make this concrete, we bin every sample of widely used counting and grounding datasets by its number of target objects (Figures~\ref{fig:appendix_count} and~\ref{fig:appendix_ground}). For counting, TallyQA \citep{acharya2019tallyqa}, CountBenchQA \citep{paiss2023countbench}, and PixmoCount place essentially all samples at ten or fewer objects; only FSC-147 and FSCD-LVIS contribute meaningful mass above $50$ objects. Grounding is even more skewed: referring-expression datasets such as RefCOCO/+/g \citep{kazemzadeh2014referitgame,yu2016refcoco,mao2016refcocog}, Ref-L4 \citep{chen2024refl4}, and GroundingME \citep{li2025groundingme} ground exactly one target per sample, and detection-style benchmarks such as OdinW-13 and LVIS are dominated by a single object per image ($97\%$ and $86\%$ of samples in the $0$--$1$ bucket). Dense, many-object grounding appears only in the FSCD variants. This is why our suite includes FSC-147, FSCD-147, and FSCD-LVIS, and why we break their results down by object count below.

\input{figs/counting_object_counts}
\input{figs/grounding_object_counts}

\paragraph{Results by object count.} Table~\ref{tab:appendix_cg_bucket} breaks down FSC-147 and FSCD-LVIS by the number of ground-truth objects, using the bins defined above. On FSC-147, \ours improves exact-match accuracy in every bucket, with the largest gains between $11$ and $50$ objects (e.g., $5.0 \to 32.9$ at 8B for $21$--$50$), where the base models' accuracy collapses. On FSCD-LVIS, where every image has at least $21$ instances, the base models recover almost none of the instances (F1 $\le 12.5$), whereas \ours reaches F1 above $56$ up to $100$ instances and around $41$ beyond. Accuracy nevertheless falls steeply with count for every configuration: above $100$ objects exact counting remains essentially unsolved, confirming that the high-count regime is where the most headroom lies.

\begin{table}[!htbp]
\centering
\small
\setlength{\tabcolsep}{5pt}
\begin{tabular}{@{}lrcccc@{}}
\toprule
& & \multicolumn{2}{c}{\textbf{Qwen3-VL-8B}} & \multicolumn{2}{c}{\textbf{Qwen3-VL-32B}} \\
\cmidrule(lr){3-4}\cmidrule(lr){5-6}
\textbf{Objects} & \textbf{\# Samples} & Base & \ours & Base & \ours \\
\midrule
\multicolumn{6}{l}{\textit{FSC-147: exact-match accuracy (\%)}} \\
$\le 10$   & 149 & 42.3 & \textbf{73.8} & 64.4 & \textbf{81.9} \\
11--20     & 368 & 28.8 & \textbf{59.2} & 33.4 & \textbf{63.3} \\
21--50     & 383 &  5.0 & \textbf{32.9} &  5.2 & \textbf{34.5} \\
51--100    & 185 &  1.1 & \textbf{9.7}  &  2.2 & \textbf{9.7} \\
$> 100$    & 201 &  3.5 & \textbf{6.5}  &  0.5 & \textbf{1.0} \\
\midrule
\multicolumn{6}{l}{\textit{FSCD-LVIS: box F1 at IoU $0.5$ (\%)}} \\
21--50     & 869 &  8.5 & \textbf{63.2} & 12.5 & \textbf{63.5} \\
51--100    & 220 &  3.4 & \textbf{58.0} &  5.8 & \textbf{56.9} \\
$> 100$    &  92 &  1.0 & \textbf{41.3} &  1.5 & \textbf{40.7} \\
\bottomrule
\end{tabular}
\caption{Grounded perception performance by number of ground-truth objects. FSCD-LVIS is scored with instance-level F1 (predicted boxes matched to ground truth at IoU $\ge 0.5$) because AP is not defined per sample.}
\label{tab:appendix_cg_bucket}
\end{table}

\paragraph{Additional metrics and benchmarks.} Table~\ref{tab:appendix_cg_extra} reports complementary counting metrics and an additional grounding benchmark. \ours lowers the mean absolute error on FSC-147 and raises CountQA relative accuracy, showing that it also reduces large miscounts rather than only fixing near misses. On LVIS (federated val), which queries both present and absent categories over $1{,}203$ classes, \ours more than doubles AP at both scales ($18.1 \to 45.2$ at 8B and $22.9 \to 45.5$ at 32B).

\begin{table}[!htbp]
\centering
\small
\setlength{\tabcolsep}{5pt}
\begin{tabular}{@{}lllccc@{}}
\toprule
& & \textbf{FSC-147} & \multicolumn{2}{c}{\textbf{CountQA}} & \textbf{LVIS} \\
\cmidrule(lr){4-5}
\textbf{Model} & \textbf{Method} & MAE $\downarrow$ & RA@5 $\uparrow$ & RA@10 $\uparrow$ & AP $\uparrow$ \\
\midrule
\multirow{2}{*}{Qwen3-VL-8B}  & Base  & 29.70 & 33.64 & 43.13 & 18.12 \\
                              & \ours & \textbf{23.17} & \textbf{50.00} & \textbf{57.59} & \textbf{45.19} \\
\midrule
\multirow{2}{*}{Qwen3-VL-32B} & Base  & 60.70 & 46.79 & 58.90 & 22.87 \\
                              & \ours & \textbf{31.24} & \textbf{65.51} & \textbf{72.71} & \textbf{45.46} \\
\bottomrule
\end{tabular}
\caption{Complementary counting metrics and LVIS grounding. MAE is the mean absolute counting error; RA@$k$ is the share of predictions within $k\%$ of the ground truth.}
\label{tab:appendix_cg_extra}
\end{table}

\subsection{Extended Visual Search and Deep Research Results}
\label{sec:appendix_mmdr_results}
Table~\ref{tab:appendix_mmdr_extend} extends the search results to $11$ information-seeking benchmarks, including five that are not in Table~\ref{tab:harness-combined}. The agent workflow improves every backbone: the average across the $11$ benchmarks rises from $13.74$ to $29.05$ for Qwen3-VL-8B, from $19.21$ to $34.61$ for Qwen3-VL-32B, and from $31.58$ to $41.80$ for GPT-5. Gains are largest for the open models, whose parametric knowledge is most limited; GPT-5's direct-inference accuracy is additionally depressed by refusals on identity queries (Appendix~\ref{para:appendix_refusal}). With the harness, Qwen3-VL-32B surpasses direct GPT-5 on average ($34.61$ vs.\ $31.58$).

\begin{table}[!htbp]
\centering
\footnotesize
\setlength{\tabcolsep}{2pt}
\resizebox{\textwidth}{!}{%
\begin{tabular}{@{}lcccccccccccc@{}}
\toprule
\textbf{Model} & MMS & MSP & RXB & FVQA & HLE-VL & BC-VL & MM-BC & VDR & VBB & InfoSeek & Enc-VQA & \textit{Avg} \\
\midrule
\multicolumn{13}{c}{\textit{Direct Inference}} \\
\midrule
Qwen3-VL-8B           & 14.04 &  1.29 & 22.16 & 28.39 &  6.36 & 21.05 &  4.94 &  5.40 &  7.69 & 29.20 & 10.61 & 13.74 \\
Qwen3-VL-8B-T         & 12.87 &  3.22 & 22.16 & 25.67 & 10.30 & 33.83 &  7.41 &  8.20 & 11.24 & 27.25 & 18.06 & 16.38 \\
Qwen3-VL-32B          & 20.47 &  2.57 & 24.74 & 32.67 & 10.00 & 28.82 &  6.58 &  9.40 & 10.06 & 33.20 & 32.76 & 19.21 \\
GPT-5                 & 36.84 & 10.93 & 34.54 & 54.67 & 20.00 & 42.11 &  8.64 & 14.20 & 28.40 & 53.67 & 43.33 & 31.58 \\
\midrule
\multicolumn{13}{c}{\textit{Agent Workflow}} \\
\midrule
Qwen3-VL-8B           & 51.46 & 14.79 & 34.71 & 55.00 & 10.30 & 32.33 &  6.00 & 14.40 & 11.24 & 50.67 & 38.67 & 29.05 \\
Qwen3-VL-32B          & 62.57 & 17.68 & 40.21 & 62.67 & 11.52 & 33.08 & 10.29 & 18.80 & 19.53 & 55.33 & 49.00 & 34.61 \\
GPT-5                 & 56.14 & 31.19 & 40.72 & 55.00 & 24.85 & 52.38 & 25.93 & 23.80 & 49.11 & 52.00 & 48.67 & 41.80 \\
\bottomrule
\end{tabular}%
}
\caption{Extended visual search and deep research results on $11$ information-seeking benchmarks, five of which (MM-BC, VDR, VBB, InfoSeek, Enc-VQA) are not in Table~\ref{tab:harness-combined}. Qwen3-VL-8B refers to Qwen3-VL-8B-Instruct and Qwen3-VL-8B-T denotes Qwen3-VL-8B-Thinking. Abbreviations: MM-BC = MM-BrowseComp; MMS = MMSearch; MSP = MMSearch-Plus; BC-VL = BrowseComp-VL; RXB = RealX-Bench; VDR = VDR-Bench; VBB = VisBrowse-Bench; Enc-VQA = EncyclopedicVQA. \emph{Avg} is the unweighted mean across the $11$ benchmarks; per-benchmark sample counts are in Appendix~\ref{sec:appendix_mmdr_bench}.}
\label{tab:appendix_mmdr_extend}
\end{table}

\subsection{Reasoning-Effort Study: Setup and Full Results}
\label{sec:appendix_compute_match}
\label{sec:appendix_effort_scaling}

\ours answers a question over several model calls, and each call carries tool outputs and memory, so it spends more model tokens per question than prompting the base model directly. This raises the question of whether the harness helps simply by spending more compute: a reasoning model given a larger token budget might recover the gain by reasoning for longer. GPT-6 Luna exposes a reasoning-effort setting, which lets us test this directly by raising the effort of every method, including \ours. Section~\ref{sec:effort} and Figure~\ref{fig:teaser}c summarize the results. Table~\ref{tab:luna_effort_scaling} reports domain averages and costs, Table~\ref{tab:luna_effort_tokens} the tokens per question behind the costs, Table~\ref{tab:luna_codex_execution} how often Codex executes code, and Table~\ref{tab:luna_effort_perdataset} the per-benchmark scores.

\paragraph{Setup.} We run Base, Codex, Visual Sketchpad, and \ours with GPT-6 Luna at five reasoning-effort settings (low, medium, high, xhigh, and max) on the full evaluation splits of all $22$ benchmarks of Table~\ref{tab:harness-combined}, that is, $26{,}229$ questions per setting and $524{,}580$ in total. Each method keeps its configuration from Table~\ref{tab:harness-combined} and Appendix~\ref{sec:appendix_baselines_free}, and the medium-effort setting is the configuration reported in Table~\ref{tab:harness-combined}. Apart from the effort, only two settings change. First, Base, Visual Sketchpad, and \ours may produce up to $16{,}000$ output tokens per model call, raised to $64{,}000$ at max effort so that reasoning is not truncated; Codex uses its own default limit. Second, GPT-6 Luna accepts max effort only through the Responses API, so Base switches to that API at max effort; \ours and Codex use it at every setting, and the Chat Completions requests of Visual Sketchpad are translated to it at every setting. The Python Coding Agent of \ours uses the same reasoning effort as the orchestrator. Codex is the Codex CLI agent~\citep[see Appendix~\ref{sec:appendix_baselines_free}]{codex}: every question runs in a fresh session with the image attached and web search disabled to prevent the model from searching the web for the benchmark's gold answer, and model-written code executes in a read-only sandbox enforced with Landlock and seccomp. All settings were run on the Flex service tier.

Cost is the GPT-6 Luna API bill for model tokens: uncached input, cache reads and writes, and output, including hidden reasoning tokens. We price all settings at Standard-tier rates; the Flex tier halves every price and leaves cost ratios unchanged. Costs exclude tool execution (e.g., SAM~3, OCR, and search APIs) and judging, and are averaged over the $22$ benchmarks in the same way as the scores. Scoring follows Table~\ref{tab:harness-combined}; on the search benchmarks, every setting is scored with the GPT-4o judge used in that table. A question whose requests fail on every retry counts as incorrect; this occurred for $134$ of the $524{,}580$ questions (marked $^\dagger$ in Table~\ref{tab:luna_effort_perdataset}).

\input{tables/luna_effort_scaling}

\paragraph{Results.} Raising the effort increases the 22-benchmark average of every method. \ours at medium effort ($56.52$ at \$0.92 per $1{,}000$ questions) exceeds every setting of every other method by at least $7.68$ points; the strongest of those settings is Visual Sketchpad at max effort ($48.84$ at \$5.15). \ours at low effort ($49.92$ at \$0.64) already exceeds the average of every other setting. At low effort, \ours leads Base on grounded perception and on visual search and deep research but trails it on general VQA and hallucination and on spatial reasoning ($61.51$ vs.\ $61.63$; $54.45$ vs.\ $56.54$); from medium effort onward, it leads Base in every domain. At max effort, the gap between \ours ($60.12$ at \$2.99) and the best other method is largest on grounded perception ($15.6$ points) and on visual search and deep research ($12.4$), consistent with the harness supplying evidence that the model cannot infer from the image, and smaller on general VQA and hallucination ($6.1$) and on spatial reasoning ($7.1$). Returns per unit of cost diminish for Base: moving from xhigh to max effort multiplies its cost by $2.7$ and raises its average by $0.72$ points. On FSC-147, no other setting exceeds $38.34$, against $40.59$ for \ours at low effort; on the six search benchmarks, \ours at max effort exceeds the best other setting by $3.0$ to $19.6$ points, and \ours at low effort already exceeds it on $4$ of them. \ours attains the best score of all settings on $20$ of the $22$ benchmarks; the exceptions are HallusionBench (Codex at xhigh effort, $69.89$, against $69.66$ for \ours at max effort) and ERQA (Visual Sketchpad at max effort, $64.75$, against $63.50$ for \ours at max effort).

\paragraph{Token usage.} Most input tokens of \ours, Codex, and Visual Sketchpad are cache reads and writes of the context carried across their model calls, whereas Base sends a single prompt. For Base, raising the effort leaves the input unchanged at about $1{,}800$ tokens per question and mainly increases reasoning tokens, from $301$ to $5{,}852$ per question between low and max effort. For the three harnesses, the input also grows between low and max effort: from about $8{,}000$ to $11{,}800$ tokens per question for Codex, from $20{,}800$ to $24{,}100$ for Visual Sketchpad, and from $14{,}500$ to $24{,}600$ for \ours.

\input{tables/luna_effort_tokens}

\paragraph{Code execution in Codex.} The sandbox permits code execution at every setting, yet GPT-6 Luna rarely uses it: Codex executes at least one command on $0.16\%$ of questions at low effort and $6.17\%$ at max effort (Table~\ref{tab:luna_codex_execution}). Execution concentrates on grounded perception and on visual search and deep research ($12.3\%$ and $9.6\%$ of their questions at max effort) and is almost absent on general VQA and hallucination and on spatial reasoning. On most questions, Codex therefore reduces to the model answering directly within the Codex agent loop and prompt. Its average is at most $2.0$ points below that of Base from medium to xhigh effort and $0.5$ points above it at max effort, but $7.4$ points below it at low effort; this drop is not caused by failed requests, since every question receives a non-empty answer.

\input{tables/luna_codex_execution}

\input{tables/luna_effort_perdataset}

\paragraph{Caveats.} Costs cover model tokens only; the tool backends used by \ours and Visual Sketchpad are not priced. The service tier affects price and latency, not the model. Each cell is a single run.

\subsection{Codex with Open-Source Backbones}
\label{sec:appendix_codex_qwen}

Table~\ref{tab:harness-combined} compares \ours with Codex only for GPT-6 Luna. To test whether that comparison depends on the backbone, we run Codex with Qwen3-VL-8B and Qwen3-VL-32B on all $22$ benchmarks.

\paragraph{Setup.} Codex is the Codex CLI agent~\citep{codex} in code mode, configured as for GPT-6 Luna (Appendix~\ref{sec:appendix_baselines_free}) except that it also appends a short answer-format instruction to each question. Every question runs in a fresh session with the image attached, web search disabled, and medium reasoning effort; the Qwen3-VL Instruct models have no separate reasoning mode, so the effort setting does not change their decoding. Both models are served with vLLM, and each response is capped at $2{,}048$ tokens, the per-turn limit used for \ours. Codex runs model-written code in a read-only sandbox; in these runs it used its default sandbox, which cannot start shell processes on our evaluation host, so Codex could evaluate only simple expressions. These results therefore measure the Codex agent loop and prompt without shell-based tools. The grounded perception benchmarks use the validation splits, as in Table~\ref{tab:harness-combined}. FSC-147 is scored by the last integer in each answer, and an empty answer counts as wrong.

\input{tables/codex_qwen}

\paragraph{Results.} Table~\ref{tab:codex_qwen} reports the results. Averaged over the $22$ benchmarks, Codex scores below Base with both backbones: $31.42$ vs.\ $34.24$ with Qwen3-VL-8B and $36.46$ vs.\ $38.30$ with Qwen3-VL-32B, while \ours reaches $45.41$ and $50.92$. Codex beats Base on $5$ benchmarks with the 8B model and $7$ with the 32B model. The larger gains are on RealX-Bench, FVQA, and MMSearch-Plus, on MindCube ($+1.35$ and $+3.47$ points), and on FREAK-MCQ with the 32B model ($+6.04$); the other gains are below $0.5$ points. The largest losses are on V* ($-14.13$ points with the 8B model), BC-VL ($-7.52$ and $-11.53$), SimpleVQA ($-8.06$ and $-8.88$), and OdinW-13-All ($-6.10$ and $-7.73$). \ours scores above Codex on every benchmark with the 8B model and on $20$ of $22$ with the 32B model. The two exceptions are FREAK-MCQ ($50.10$ vs.\ $45.77$) and SparBench ($46.99$ vs.\ $46.76$). As with GPT-6 Luna, placing a model in a general-purpose coding-agent loop does not by itself reproduce the gains of \ours.

\clearpage
\section{Harness Implementation Details}
\label{sec:appendix_tools}

This section documents the harness of Section~\ref{sec:harness_design}: the nine capabilities of Section~\ref{sec:setup_base}, how their outputs are kept in multimodal memory, and the agent's system prompt.

\paragraph{Capabilities and backends.} Following the generalized capability abstraction of Section~\ref{sec:setup_base}, each capability exposes a fixed interface to the agent: typed arguments in, and a natural-language message plus memory artifacts out. The expert model or service behind the interface is a configurable backend, and switching it changes neither the agent's prompt, the tool schema, nor the memory interface. Table~\ref{tab:appendix_base_tools} lists the capabilities and their backends; all experiments in this paper use the default backend of each capability, shown in bold.

\begin{table}[!htbp]
\centering
\small
\setlength{\tabcolsep}{4pt}
\begin{tabular}{@{}p{0.2\textwidth}p{0.4\textwidth}p{0.34\textwidth}@{}}
\toprule
\textbf{Capability} & \textbf{Interface} & \textbf{Backends} \\
\midrule
\multicolumn{3}{l}{\textit{Search}} \\
Text Search & text query $\rightarrow$ summarized results with cited URLs & \textbf{Serper}; Gemini with Google Search \\
Image Search & image $\rightarrow$ visual matches (entities, URLs) & \textbf{Serper (Google Lens)}; Gemini with Google Search \\
Webpage Visit & URLs and goal $\rightarrow$ goal-relevant evidence & \textbf{Serper} with an LLM extractor \\
\midrule
\multicolumn{3}{l}{\textit{Perception}} \\
Zoom-in & image and box $\rightarrow$ cropped image & \textbf{built-in} \\
Visual Grounding & image and text query $\rightarrow$ boxes, masks, annotated image & \textbf{SAM 3}; Grounding DINO \\
OCR & image $\rightarrow$ transcribed text & \textbf{PaddleOCR-VL}; DeepSeek-OCR \\
Depth Estimation & image $\rightarrow$ depth map & \textbf{Depth Anything 3}; VGGT \\
Camera Trajectory & images $\rightarrow$ camera poses and trajectory plot & \textbf{Depth Anything 3}; VGGT \\
\midrule
\multicolumn{3}{l}{\textit{Compositional computation}} \\
Python Coding Agent & objective $\rightarrow$ program output and new bindings & \textbf{single-shot code generator}; full coding agent \\
\bottomrule
\end{tabular}
\caption{The nine capabilities, their interfaces, and their interchangeable backends. Bold marks the backend used in all experiments.}
\label{tab:appendix_base_tools}
\end{table}

\subsection{Search Tools}

By default, all three search tools route queries through the Serper API\footnote{\url{https://serper.dev/}}; Text Search and Image Search can instead be served by Gemini~\citep{comanici2025gemini25} with Google Search grounding, a costlier but stronger route. The search tools post-process the raw payload with a Qwen3-VL-8B-Instruct model as the VLM summarizer. The summarizer runs at decoding temperature $0$ with a $4096$-token output cap, and each call retrieves the top $10$ raw results before reranking and summarization.

\paragraph{Text Search.} Search the web with intelligent query routing and VLM summarization. Given a text query, the tool automatically classifies the query intent and routes it to the most appropriate Serper endpoint, which has general web, news, academic papers, images, maps, local places, shopping, or patents types, and returns a coherent summary of the top-$10$ raw results with inline citations \texttt{[1], [2], \ldots} followed by a citations block of source titles and URLs. This format supports multi-round search: the agent can read the summary for quick understanding and then visit specific cited URLs through Webpage Visit. The summarizer prompt is shown below.

\begin{tcolorbox}[breakable,title=Text Search summarizer prompt (abridged),label=box:text_search_prompt]
You are a research summarizer. Given raw web search results and a user query, do exactly three things:

\begin{enumerate}[leftmargin=*,noitemsep,topsep=2pt]
\item RERANK the numbered organic results by how directly each helps answer the query.
\item KEEP at most $K$ most useful results; emit their original 1-based ranks as \texttt{kept\_indices}. If none help, emit \texttt{kept\_indices: []}.
\item SUMMARIZE the query-relevant facts from the FULL set of raw results into a concise 3--5 sentence prose paragraph; include named entities, numbers, and dates verbatim.
\end{enumerate}

Knowledge-graph, answer-box, and top-stories blocks are ineligible for \texttt{kept\_indices} but their facts should still surface in \texttt{summary}. Output STRICT JSON with the schema \texttt{\{kept\_indices, summary, no\_results\_justification, suggested\_queries, suggested\_next\_actions\}}; when no result is useful, populate the justification and propose alternative queries or downstream tool calls.
\end{tcolorbox}

\paragraph{Image Search.} Identify the objects, landmarks, artwork, text, products, plants, animals, and other entities depicted in an image, using Serper's reverse-image-search (Google Lens) endpoint. Returns a list of visual matches with titles, source URLs, links, and thumbnails. The agent invokes the tool by passing the runtime identifier of the target image. The tool performs reverse search (image $\rightarrow$ information) on a single image per call; finding images from a text description is instead handled by Text Search with an image-search intent. A separate multimodal summarizer prompt (below) reranks the candidates against the query image and produces the same JSON envelope as Text Search.

\begin{tcolorbox}[breakable,title=Image Search summarizer prompt (abridged),label=box:image_search_prompt]
You are a visual-match summarizer. Given raw reverse-image-search results (and, in the multimodal variant, the query image plus each candidate's thumbnail), do exactly three things:

\begin{enumerate}[leftmargin=*,noitemsep,topsep=2pt]
\item RERANK the numbered matches by how likely each identifies the same subject as the query image.
\item KEEP at most $K$ most useful matches as \texttt{kept\_indices}.
\item SUMMARIZE what the FULL set of raw matches indicates about the query image's likely subject in a concise 3--5 sentence prose paragraph; capture entity names, dates, attributes verbatim.
\end{enumerate}

Output STRICT JSON with \texttt{\{kept\_indices, summary, no\_results\_justification, suggested\_next\_actions\}}; when no match is useful, justify per-match and propose follow-up actions such as cropping with Visual Grounding before re-searching, or switching to Text Search.
\end{tcolorbox}

\paragraph{Webpage Visit.} Visit one or more webpages, convert them to plain text via Serper APIs, and extract goal-relevant content using a local Qwen3-VL-8B-Instruct model. This tool returns per-URL evidence and a summary. The extraction prompt is shown below.

\begin{tcolorbox}[breakable,title=Webpage Visit extraction prompt,label=box:webpage_visit_prompt]
Please process the following webpage content and user goal to extract relevant information:

\textbf{Webpage Content}: \texttt{\{webpage\_content\}}

\textbf{User Goal}: \texttt{\{goal\}}

\textbf{Task Guidelines}
\begin{enumerate}[leftmargin=*,noitemsep,topsep=2pt]
\item \textbf{Content scanning for rationale}: locate the specific sections directly related to the user's goal.
\item \textbf{Key extraction for evidence}: extract the most relevant information; preserve the full original context whenever possible.
\item \textbf{Summary output}: organize the result into a concise paragraph with logical flow that connects the evidence back to the goal.
\end{enumerate}

Output: a single JSON object with keys \texttt{rational}, \texttt{evidence}, and \texttt{summary} (each a string; empty string if no content is available). No code fences, no Markdown, no extra keys.
\end{tcolorbox}

\subsection{Perception Tools}

\paragraph{Zoom-in.} Crop a region of an image given a normalized bounding box on the $[0, 1000]$ scale plus padding, and return the cropped image. Used for closer visual inspection after the agent has already localized a region of interest.

\paragraph{Visual Grounding.} Locate and highlight regions of an image that match a text query, backed by SAM~3 \citep{sam3} by default or by Grounding DINO \citep{liu2023groundingdino}. It returns an annotated image with matched regions highlighted and numbered, along with bounding boxes (and segmentation masks when the backend provides them).

\paragraph{OCR.} Extract readable text from an image, backed by PaddleOCR-VL \citep{paddleocrvl} by default or by DeepSeek-OCR \citep{wei2025deepseekocr}. Returns the transcribed text as a single string. Used when the task is about words, numbers, or printed labels in the image; not for general object localization or segmentation.

\paragraph{Depth Estimation.} Estimate a per-pixel depth map for an input image, backed by Depth Anything 3 \citep{depthanything3} or VGGT \citep{wang2025vggt}. Returns a depth image (closer surfaces brighter, farther darker). Used when the task requires reasoning about relative distance, foreground--background separation, or 3D scene structure cues that flat 2D crops cannot resolve.

\paragraph{Camera Trajectory.} Given an ordered sequence of at least two images of the same scene, estimate each camera's pose relative to the first image using Depth Anything 3 \citep{depthanything3} or VGGT \citep{wang2025vggt}, and translate the result into a form the model can reason over directly. Rather than returning raw geometry alone, the tool reports, for every image, a plain-language summary of translation and rotation with respect to the reference frame (e.g.\ ``salient direction: backward (-5.43); detailed: left (-0.20), down (+0.28), backward (-5.43)''), the accompanying scalars (camera-center offset, forward direction, yaw, pitch), an explicit statement of the coordinate convention ($+X$ right, $+Y$ down, $+Z$ forward, with positive yaw turning right), and a bird's-eye-view rendering of the camera positions on the reference frame's $X$--$Z$ plane. Per-image intrinsics and world-to-camera extrinsics are returned as well for downstream computation. Translations below one meter and rotations below ten degrees are reported as non-salient. Used when a question depends on how the viewpoint moved or rotated across frames, or on multi-view scene structure that no single image conveys; the estimate is metric only up to the model's calibration and requires sufficient scene overlap between the reference image and the other views.

\subsection{Python Coding Agent}

\paragraph{Python Coding Agent.} Run Python over the agent's local computation state and the outputs of previously called tools (bounding boxes, OCR strings, segmentation masks, and so on), returning both standard output and the resulting local variables. The agent describes a computation goal in plain language, and by default a single-shot code-generation session of the same backbone model as the orchestrator writes the snippet using prior tool outputs as named variables before it is executed; the same interface can instead be backed by a full coding agent that inspects and revises its own code before returning. Used for aggregating numbers extracted by OCR, filtering detected regions by bounding-box area, or any cross-tool numerical reasoning the agent cannot complete in its own response.

\subsection{Multimodal Memory Artifacts}
\label{sec:appendix_memory}

Table~\ref{tab:appendix_memory_artifacts} summarizes how each tool's output is retained in the explicit multimodal memory of Section~\ref{sec:explicit_memory} and the handle by which the agent re-addresses it on later turns.
\input{tables/memory_artifacts}

\paragraph{Environment Update example.} After every tool turn, the harness appends an Environment Update message that re-advertises the live memory namespace. The example below is taken verbatim from an FSCD-147 trajectory after two Visual Grounding calls; when web pages have been retrieved, the message additionally lists the registered \texttt{webpage\_url\_N} references.

\begin{tcolorbox}[breakable,title=Environment Update message (example)]
\small
Environment Update (not a new question):\\
Currently available image IDs for tool calls:\\
- input\_image\_1: user\_input\\
- tool\_generated\_image\_1: tool\_output:Visual\_Grounding\_Tool\\
- tool\_generated\_image\_2: tool\_output:Visual\_Grounding\_Tool\\
When a tool needs image input, pass one of these IDs via the *\_image\_id field.
Do NOT invent image file names or paths.

\medskip
Available data bindings (auto-injected into Python\_Coding\_Agent\_Tool):\\
- segment\_finger\_food: dict (keys: pred\_boxes, pred\_scores, pred\_masks, num\_masks, orig\_img\_h, orig\_img\_w)\\
- segment\_small\_pastry\_with\_filling: dict (keys: pred\_boxes, pred\_scores, pred\_masks, num\_masks, orig\_img\_h, orig\_img\_w)
\end{tcolorbox}

\subsection{System Prompt}
\label{sec:appendix_sysprompt}
Here is the system prompt of the agent.
\begin{tcolorbox}[breakable,title=System Prompt]
You are a visual reasoning agent that answers questions about images using a structured Perception-Reasoning-Action framework.

Your goal is to provide accurate, well-reasoned answers. You have access to various tools that compensate for known limitations of vision-language models.

Use native tool-calling for tools. Do not write textual Action/Action Input to call tools.

At most one native tool call is allowed per assistant message/turn.

Use the following format:

Question: the input question you must answer

Observation: your direct perception of the input or the result of the action you took

Thought: you should always think about what to do

Action: the action to take, should be one of the provided tools, or ``Final Answer'' if you have a final answer and do not need to call any more tools.

Action Input: the input to the action, or the final answer if Action is ``Final Answer''

\# Image-input rule:

- Runtime image IDs are provided in Environment Update messages.

- If a tool needs image input, use those IDs in *\_image\_path fields.

- Do NOT invent image file names or paths.

\# Tool-Use Rules

- Call exactly ONE tool per turn. Never include multiple \textless{}tool\_call\textgreater{} blocks.

- Full tool schemas with parameters and types are provided in the \textless{}tools\textgreater{} section below.

- After each tool call, verify whether the result is actually informative for the task.

- If a tool result is empty, missing key values, or non-informative, do NOT guess. Issue another tool call with corrected arguments or use a different tool.

\# Available Tools

You may call ONE tool per turn to assist with the user query.

You are provided with function signatures within \textless{}tools\textgreater{}\textless{}/tools\textgreater{} XML tags:

\textless{}tools\textgreater{}

\{tools\_directory\}

\textless{}/tools\textgreater{}

For each function call, return a json object with function name and arguments within \textless{}tool\_call\textgreater{}\textless{}/tool\_call\textgreater{} XML tags:

\textless{}tool\_call\textgreater{}

\{``name'': \textless{}function-name\textgreater{}, ``arguments'': \textless{}args-json-object\textgreater{}\}

\textless{}/tool\_call\textgreater{}
\end{tcolorbox}

\subsection{Models and Serving}
\label{sec:appendix_impl}

Table~\ref{tab:appendix_models} lists every model used in our experiments and the role it plays.

\begin{table}[!htbp]
\centering
\small
\begin{tabular}{@{}lll@{}}
\toprule
\textbf{Model} & \textbf{Role} & \textbf{Access} \\
\midrule
\multicolumn{3}{l}{\textit{Agent backbones}} \\
Qwen2.5-VL-7B-Instruct & Backbone (vs.\ training-based baselines) & vLLM \\
Qwen3-VL-8B-Instruct & Backbone & vLLM \\
Qwen3-VL-32B-Instruct & Backbone & vLLM \\
Qwen3-VL-8B-Thinking & Direct-inference reference (search) & vLLM \\
GPT-5 & Backbone & OpenAI API \\
GPT-6 Luna, Sol, Astra & Backbones & OpenAI API \\
\midrule
\multicolumn{3}{l}{\textit{Evaluation}} \\
GPT-4o & Judge (visual search and deep research) & OpenAI API \\
Qwen3-VL-8B-Instruct & Judge / answer extractor (other domains) & vLLM \\
\midrule
\multicolumn{3}{l}{\textit{Tool backends}} \\
Qwen3-VL-8B-Instruct & Webpage Visit extractor, search summarizer & vLLM \\
(= orchestrator backbone) & Python Coding Agent & (= backbone serving) \\
SAM 3 & Visual Grounding & local server \\
PaddleOCR-VL & OCR & local server \\
Depth Anything 3, VGGT & Depth Estimation, Camera Trajectory & local server \\
\bottomrule
\end{tabular}
\caption{Models used in our experiments.}
\label{tab:appendix_models}
\end{table}

\paragraph{Open-source backbones.} All Qwen models (Qwen2.5-VL~\citep{qwen25vl} and Qwen3-VL~\citep{qwen3vl}) use the public Hugging Face weights, served in bf16 with vLLM~\citep{kwon2023efficient} behind an OpenAI-compatible endpoint with native tool calling. We sample at temperature $0.7$, top-$p$ $0.8$, top-$k$ $20$, repetition penalty $1.0$, and presence penalty $1.5$, with up to $12{,}800$ generated tokens per turn. The agent runs for at most fifteen tool-calling iterations per query, and its context is capped at 168K tokens; when a trajectory approaches the cap, older turns are compacted while the latest Environment Update and the most recent turn are kept verbatim.

\paragraph{Proprietary backbones.} We call GPT-5~\citep{gpt5}, GPT-6 Luna, GPT-6 Sol, and GPT-6 Astra~\citep{gpt6} through the OpenAI API with native function calling, provider-default sampling, and up to $16{,}000$ output tokens per turn. GPT-5 uses the pinned snapshot \texttt{gpt-5-2025-08-07}, and the three GPT-6 models (\texttt{gpt-6-luna}, \texttt{gpt-6-sol}, and \texttt{gpt-6-astra}) use medium reasoning effort. All four use the same tools, multimodal memory, system prompt, and iteration cap as the open-source backbones.

\paragraph{Tool backends.} Four tools call a language model internally: Text Search and Image Search use it to rerank and summarize the retrieved results, Webpage Visit uses it to extract goal-relevant content from a fetched page, and the Python Coding Agent uses it to write code for the agent's objective. The three search-related tools use Qwen3-VL-8B-Instruct with greedy decoding, independent of the agent backbone. The Python Coding Agent instead routes to a fresh context of the same backbone model as the orchestrator, so that the code generator's capability scales with the backbone. Every capability uses its default backend (Table~\ref{tab:appendix_base_tools}). Perception tools are backed by SAM 3~\citep{sam3} (Visual Grounding), PaddleOCR-VL~\citep{paddleocrvl} (OCR), and Depth Anything 3~\citep{depthanything3} or VGGT~\citep{wang2025vggt} (Depth Estimation and Camera Trajectory), each served locally; search tools query the Serper API.

\paragraph{Judges.} We use GPT-4o~\citep{gpt4o} with the Vision-DeepResearch judge prompt (Appendix~\ref{para:appendix_judge}) for the visual search and deep research benchmarks, and Qwen3-VL-8B-Instruct for the remaining LLM-judged benchmarks; both judges decode greedily. On the counting benchmarks the judge only extracts the final integer from free-form responses, and correctness is decided by exact match. Grounding benchmarks need no judge and are scored with COCO-style box AP.

\section{Benchmarks and Evaluation Protocol}
\label{sec:appendix_benchmarks}

Table~\ref{tab:appendix_bench_overview} summarizes the $22$ benchmarks of Table~\ref{tab:harness-combined}. The following subsections describe each domain's benchmarks, scoring, and prompts.

\begin{table}[!htbp]
\centering
\small
\setlength{\tabcolsep}{4pt}
\begin{tabular}{@{}lllrll@{}}
\toprule
\textbf{Domain} & \textbf{Benchmark} & \textbf{Split} & \textbf{\# Samples} & \textbf{Format} & \textbf{Scoring} \\
\midrule
\multirow{5}{*}{\shortstack[l]{Grounded\\Perception}}
 & CountQA & test & 1{,}528 & integer & exact match \\
 & FSC-147 & val & 1{,}286 & integer & exact match \\
 & OdinW-13-All & val & 4{,}938 & labeled boxes & box AP \\
 & FSCD-147 & val & 1{,}286 & boxes & box AP \\
 & FSCD-LVIS & val & 1{,}181 & boxes & box AP \\
\midrule
\multirow{6}{*}{\shortstack[l]{Visual Search and\\Deep Research}}
 & MMSearch & end-to-end & 171 & open-ended & GPT-4o judge \\
 & MMSearch-Plus & full & 311 & open-ended & GPT-4o judge \\
 & RealX-Bench & full & 194 & open-ended & GPT-4o judge \\
 & FVQA & test & 1{,}800 & open-ended & GPT-4o judge \\
 & HLE-VL & full & 330 & open-ended & GPT-4o judge \\
 & BrowseComp-VL & full & 399 & open-ended & GPT-4o judge \\
\midrule
\multirow{6}{*}{\shortstack[l]{General VQA and\\Hallucination}}
 & RealWorldQA & full & 765 & multiple choice & option match \\
 & SimpleVQA & full & 2{,}025 & open-ended & LLM judge \\
 & HallusionBench & full & 951 & yes/no & exact match \\
 & FREAK-VQA & full & 799 & open-ended & LLM judge \\
 & FREAK-MCQ & full & 994 & multiple choice & option match \\
 & V* & full & 191 & multiple choice & option match \\
\midrule
\multirow{5}{*}{\shortstack[l]{Spatial\\Reasoning}}
 & MMSI-Bench & full & 1{,}000 & multi-image MC & option match \\
 & ERQA & full & 400 & multiple choice & option match \\
 & EmbSpatial-Bench & full & 3{,}640 & multiple choice & option match \\
 & SparBench & tiny & 1{,}000 & MC and numeric & option match / MRA \\
 & MindCube & tiny & 1{,}040 & multi-image MC & option match \\
\bottomrule
\end{tabular}
\caption{The $22$ benchmarks of Table~\ref{tab:harness-combined}. ``LLM judge'' is Qwen3-VL-8B-Instruct; MRA is mean relative accuracy for numeric answers.}
\label{tab:appendix_bench_overview}
\end{table}

\paragraph{Sampling Parameters.} All benchmarks use the same agent configuration. The model is served locally via vLLM~\citep{kwon2023efficient} and sampled with temperature $0.7$, top-$p$ $0.8$, top-$k$ $20$, repetition penalty $1.0$, and presence penalty $1.5$, with a cap of $12{,}800$ tokens per model call and a $168$K-token context window. The agent uses native tool calls and is allowed at most $15$ tool-calling iterations per question; a trajectory that exhausts this budget, or that fails after three API retries, is scored incorrect. Web-page observations are capped at $20$ environment updates per trajectory.


Because decoding is stochastic, all Qwen3-VL results are means of three runs with different random seeds; Appendix~\ref{sec:appendix_std} describes the protocol and reports standard deviations.

\paragraph{Scoring.} Multiple-choice and yes/no answers are matched against the gold option after a rule-based parser extracts the choice from the response; when the parser fails, Qwen3-VL-8B-Instruct extracts the chosen option, without judging correctness. Open-ended answers are graded by an LLM judge: GPT-4o for visual search and deep research (Appendix~\ref{sec:appendix_mmdr_bench}) and Qwen3-VL-8B-Instruct for the other domains. Grounded perception benchmarks are scored without a judge (Appendix~\ref{sec:appendix_cg_bench}). All methods and backbones on a benchmark share the same prompt and scorer.

\subsection{Grounded Perception}
\label{sec:appendix_cg_bench}
\label{sec:appendix_cg_eval}

\paragraph{FSC-147.} A class-agnostic counting benchmark \citep{ranjan2021learning} covering 147 categories with a point annotation for every object. Scenes are dense (median 27 objects, up to 2{,}092). We evaluate on the 1{,}286-image val split and ask ``How many \emph{category} are there in the image?''. Unlike the original few-shot protocol, we do not provide exemplar boxes, so the task is zero-shot.

\paragraph{CountQA.} A benchmark \citep{tamarapalli2025countqa} of real-world counting questions designed to stress MLLMs with cluttered, occluded, and densely packed scenes. We use all 1{,}528 questions of the test split.

\paragraph{OdinW-13-All.} Built from the thirteen detection datasets of Object Detection in the Wild (OdinW-13) \citep{li2022grounded}, spanning aerial and underwater imagery, egocentric hands, thermal images, vehicles, and everyday objects. Unlike the common protocol of querying one category at a time, each image is posed as a single query that lists every category of its source dataset (for example, all $20$ PascalVOC classes), hence the suffix \emph{-All}, and the model must return a box with a category label for every instance of every listed category. A missed or mislabeled category therefore costs AP within the same response, and our numbers are not directly comparable to results obtained with one query per category. We use the 4{,}938 val images and report AP averaged over the thirteen datasets.

\paragraph{FSCD-147 and FSCD-LVIS.} The few-shot counting-and-detection benchmarks of \citet{nguyen2022few}, which annotate a box for every instance: FSCD-147 re-annotates FSC-147, and FSCD-LVIS draws dense scenes (21--401 instances) from LVIS \citep{gupta2019lvis}. We use the val splits (1{,}286 and 1{,}181 images), query one category per image, and again withhold the exemplar boxes.

\paragraph{LVIS (federated).} In addition to the five benchmarks of Table~\ref{tab:harness-combined}, Appendix~\ref{sec:appendix_cg_results} reports grounding on the LVIS val set \citep{gupta2019lvis} (19{,}809 images, 1{,}203 categories) under its federated protocol, which queries both categories present in an image and verified-absent ones, so a model must also learn to return an empty list.

\paragraph{Counting metrics.} Counting benchmarks are scored by exact-match accuracy between the predicted and ground-truth integers. A rule-based parser extracts the integer (or number word) from the response; for long free-form responses, Qwen3-VL-8B-Instruct first extracts the final answer with the prompt below, and correctness is still decided by exact match. We additionally report mean absolute error and, for CountQA, relative accuracy within $5\%$ and $10\%$ of the ground truth (Appendix~\ref{sec:appendix_cg_results}).

\paragraph{Grounding metrics.} Grounding benchmarks are scored with COCO box AP averaged over IoU thresholds $0.50$:$0.95$, computed with the official COCO evaluation code. Predicted boxes are parsed from the response (JSON, with a lenient fallback for lists of four numbers) and converted from the $0$--$1000$ grid to pixel coordinates. For OdinW-13-All, predicted labels are matched to the category names, AP is computed per source dataset, and the thirteen values are averaged. FSCD-147, FSCD-LVIS, and LVIS are evaluated per queried category. No LLM judge is involved.

\paragraph{Task prompts.} All methods receive the same task prompts, shown below with category names as placeholders.

\begin{tcolorbox}[breakable,title=Counting prompts]
\textbf{FSC-147.} How many \{category\} are there in the image? Your final answer should be a single integer.

\medskip
\textbf{CountQA.} \{original question\} Your final answer should be a single integer.
\end{tcolorbox}

\begin{tcolorbox}[breakable,title=Grounding prompts]
\textbf{OdinW-13-All.} Locate every instance that belongs to the following categories: `\{category$_1$\}, \{category$_2$\}, \ldots'. Report the bounding boxes in JSON format as a list of objects, one object per detected instance, where each object has a ``bbox\_2d'' key with [x1, y1, x2, y2] and a ``label'' key with the category name exactly as written above. Coordinates must be normalised on a 0-1000 scale, with (0,0) as the top-left corner and (1000,1000) as the bottom-right corner of the image. [\ldots] If none of the listed categories is present, output an empty list [].

\medskip
\textbf{FSCD-147, FSCD-LVIS, and LVIS.} Locate all instances of `\{category\}' in this image. For each instance, output a bounding box as [x1, y1, x2, y2] where coordinates are normalised on a 0-1000 scale (with (0,0) as the top-left corner and (1000,1000) as the bottom-right corner). Your answer should be a list of lists. If there are no instances of `\{category\}' in the image, output an empty list [] as your answer.
\end{tcolorbox}

\begin{tcolorbox}[breakable,title=Answer extraction prompt (abridged)]
You are an AI assistant who will help me to extract the final answer from a model response to a visual question. You are provided with a question, the ground truth answer, and the model's response. Your task is to extract ONLY the final short answer from the response, matching the format of the ground truth (e.g.\ if the ground truth is a number, extract a number [\ldots]). Do NOT evaluate whether the answer is correct. Just extract it.

If the response contains reasoning or explanation, ignore it and only return the final answer. [\ldots three worked examples \ldots]

Question: \{question\} \quad Ground truth: \{answer\} \quad Response: \{response\} \quad Extracted answer:
\end{tcolorbox}

\subsection{Visual Search and Deep Research}
\label{sec:appendix_mmdr_bench}
\label{sec:appendix_mmdr_eval}


\paragraph{FVQA.} A fact-based VQA dataset \citep{mmsearchr1} that pairs images with questions whose answers require commonsense or world-knowledge facts not directly observable in the image. Following Vision-DeepResearch~\citep{vdr}, we evaluate on the official test split of 1{,}800 samples; all samples are search-required.

\paragraph{MMSearch.} A multimodal search benchmark \citep{mmsearch} of 300 manually curated questions across 14 subfields, designed to assess large multimodal models as multimodal search engines. We evaluate on the End-to-End sub-task, which requires the agent to perform query understanding, retrieval, and answering in a single pipeline. To align with other works for fair comparison, we only use the multimodal subset, which contains 171 samples.

\paragraph{MMSearch-Plus.} A harder, provenance-aware extension \citep{mmsearchplus} that emphasizes localized visual reasoning, verification under noisy or conflicting retrieval, and multi-step text--image search; each item provides 1--5 images and one or more valid answers. We use the full set of 311 samples for evaluation.

\paragraph{BrowseComp-VL.} The vision--language extension of BrowseComp \citep{browsecomp} introduced by WebWatcher \citep{browsecompvl}, with cross-modal information-seeking questions spanning entertainment, humanities, technology, natural science, and other domains. We use all 399 samples for evaluation.

\paragraph{HLE-VL.} The multimodal subset of Humanity's Last Exam \citep{hle}, drawn from frontier-knowledge questions across mathematics, the natural sciences, and the humanities. We use all 330 samples for evaluation.

\paragraph{RealX-Bench.} Introduced by DeepEyesV2 \citep{hong2025deepeyesv2}, it contains expert-verified instances that require integrating fine-grained visual perception, dynamic multi-hop web retrieval, and multi-step logical reasoning. This benchmark contains perception and search types, which makes it easy to attribute gains to visual versus search tools. We use all 194 samples for evaluation.

\paragraph{Extended benchmarks.} The following five benchmarks appear only in the extended search results (Appendix~\ref{sec:appendix_mmdr_results}, Table~\ref{tab:appendix_mmdr_extend}). They use the same judge and prompt as the six benchmarks above.

\paragraph{MM-BrowseComp.} Hand-crafted multimodal browsing questions \citep{mmbrowsecomp} across several sub-tasks, where the answer-relevant content is hard to find. There are two versions of MM-BrowseComp. We use the newest version, which has 243 multimodal tasks.

\paragraph{VDR-Bench.} The benchmark released alongside Vision-DeepResearch \citep{vdr}: long-horizon multimodal deep-research instances that interleave many reasoning turns, designed to evaluate sustained multi-step browsing under heavy real-world noise. We use its test-mini split of 500 samples.

\paragraph{VisBrowse-Bench.} A benchmark of visual-native search instances \citep{visbrowsebench} that require cross-validation of multimodal evidence via interleaved text--image retrieval and joint reasoning, including cases where the answer is not present in any text snippet. We use the full 169 samples for evaluation.

\paragraph{InfoSeek.} A visual information-seeking benchmark \citep{infoseek} of fine-grained questions about entities depicted in images (e.g., species, models, landmarks). The official test split has no public answers, so we evaluate on the 2{,}000-sample test subset used by MMSearch-R1 \citep{mmsearchr1}.

\paragraph{EncyclopedicVQA.} A benchmark \citep{encyclopedicvqa} of questions about detailed properties of fine-grained categories. We draw a random subset of 2{,}000 samples from the test split. 40 of its Google Landmarks v2 images are no longer publicly available, which leaves 1{,}960 samples for evaluation.

\paragraph{No retrieval shortcut.} Across all visual search and deep research benchmarks, we did not observe cases in which the search tools surface the gold answer verbatim from indexed pages. This reduces the likelihood that gains are driven by trivial retrieval of the gold answer.

\paragraph{Judge prompt.}\label{para:appendix_judge} To align with recent multimodal deep-research work \citep{vdr}, we adopt the same LLM judge and use GPT-4o as the judge model, applying the prompt below uniformly across all search-related benchmarks.
\begin{tcolorbox}[breakable,title=LLM Judge Prompt]
You are an impartial judge evaluating whether a deep research report contains the correct answer.

[Question] \{question\}

[Correct Answer] \{correct\_answer\}

[Deep Research Report] \{response\}

Task: Determine if the deep research report contains the correct answer anywhere in its content.

Instructions:

1. Read through the entire research report carefully

2. Look for the correct answer anywhere in the report (it may be embedded in paragraphs, tables, or sections)

3. Check if the information in the report is consistent with the correct answer

4. The answer does NOT need to be in a specific format or labelled as ``final answer''

5. Provide your reasoning

6. Answer with ``yes'' if the report contains the correct answer, ``no'' if it doesn't or contradicts it

Output format: correct: [yes/no] reasoning: [your explanation]
\end{tcolorbox}


\paragraph{Refusals on identity queries.}\label{para:appendix_refusal} Some of our evaluation benchmarks contain queries that reference a real person, such as ``who is the person in this photograph,'' ``what is the affiliation of this researcher''. Proprietary models are increasingly trained to decline such queries under their privacy and identification policies, even when the gold answer is a public-figure fact already widely indexed on the open web. We observe this for the GPT models in our setting, which return a refusal in place of an answer, which the judge counts as incorrect. Therefore, in some of the benchmarks, the GPT performance is affected to some extent.

\subsection{General VQA and Hallucination}
\label{sec:appendix_vqa_bench}

This domain tests whether tool use helps, or at least does not hurt, on everyday visual questions and on questions designed to provoke hallucination.

\paragraph{RealWorldQA.} Real-world photographs, many taken from vehicles, with multiple-choice questions about spatial layout and scene understanding \citep{xai2024realworldqa}; we use all 765 questions.

\paragraph{SimpleVQA.} A factuality benchmark \citep{simplevqa} of short, knowledge-bound questions about images. We use all 2{,}025 questions and grade answers as correct or incorrect with an LLM judge.

\paragraph{HallusionBench.} Yes/no questions built from paired original and edited images and charts to expose language hallucination and visual illusion \citep{guan2024hallusionbench}. Of the $1{,}129$ questions in the released set, we evaluate the $951$ that are grounded in an image. Free-form responses are first mapped to yes/no by an LLM extraction pass before scoring. The benchmark defines three accuracies: \emph{aAcc}, the fraction of individual questions answered correctly; \emph{qAcc}, the fraction of question groups for which \emph{every} question is correct; and \emph{fAcc}, the fraction of figures for which every associated question is correct. The latter two are deliberately strict, since a model that answers from language priors tends to succeed on the original image and fail on its edited counterpart. We report the benchmark's \emph{Overall} score, the unweighted mean of the three.

\paragraph{FREAK-VQA and FREAK-MCQ.} FREAK \citep{yin2026freak} asks about counter-commonsense images, where answering from priors rather than the pixels leads to hallucination, across detection, counting, attribute, position, and OCR questions. It comes in an open-ended format (799 questions, graded by an LLM judge) and a four-option multiple-choice format (994 questions) that includes ``the correct answer is not listed''.

\paragraph{V*.} Questions about small attributes and relative positions of objects in high-resolution images \citep{wu2024v}, where the relevant detail is easy to miss at the model's input resolution; we use all 191 questions.

\subsection{Spatial Reasoning}
\label{sec:appendix_spatial_bench}

The spatial reasoning benchmarks are multiple-choice, and several of them provide multiple views of a scene, which exercises the harness's handling of several input images. We follow the evaluation split used by the EASI leaderboard~\citep{easi2025} and use its official implementation in the VLMEvalKit repository~\citep{duan2024vlmevalkit}.

\paragraph{MMSI-Bench.} Multi-image spatial-intelligence questions \citep{yang2025mmsi} spanning positional relations, attributes, motion, and multi-step reasoning across views; we use the 1{,}000 questions in the standard (non-circular) setting.

\paragraph{ERQA.} The embodied reasoning benchmark released with Gemini Robotics \citep{team2025gemini}, with 400 multiple-choice questions on spatial reasoning, trajectories, actions, and state estimation, some with several images.

\paragraph{EmbSpatial-Bench.} Spatial-relation questions posed from an egocentric viewpoint in embodied 3D scenes \citep{du2024embspatial}; we use all 3{,}640 questions.

\paragraph{SparBench.} Spatial perception and reasoning tasks over single and multiple views \citep{zhang2026flatland}. We use the tiny split: $20$ tasks $\times$ $50$ questions $=1{,}000$ questions. Tasks are scored by type. Eleven multiple-choice tasks (spatial relations, spatial imagination, position matching, camera-motion inference) are scored by option-match accuracy. Eight numeric tasks (depth and distance prediction, object-centric and object-to-object, single- and multi-view) are scored by mean relative accuracy (MRA), which averages the indicator that the relative error falls below a threshold, sweeping the threshold from $0.5$ to $0.95$ in steps of $0.05$, so partially accurate magnitude estimates receive partial credit. One task, view-change inference, uses the benchmark's own view-change metric. We report the \emph{Overall} score: the unweighted mean of the $20$ per-task metrics, so each task contributes equally regardless of its scoring type and regardless of the differing scales of accuracy and MRA. 

\paragraph{MindCube.} Questions that require building a mental model of a scene from a few partial views, covering ``among'', ``around'', and rotation settings \citep{wang2025mindcube}; we use the 1{,}040-question MindCube-Tiny split.

\input{text/appendix_optim}

\ifshowexamples
\clearpage
\input{text/appendix_examples}

\fi

\section{Baseline Reproduction}
\label{sec:appendix_baselines}

We re-run every baseline ourselves on our benchmark suite using each method's official implementation (Table~\ref{tab:appendix_baselines}). All baselines receive exactly the same question text as \ours, including each dataset's answer-format instruction (e.g., ``a single integer'', or a JSON list of boxes on a $0$--$1000$ scale for grounding); the only addition is a short answer-format instruction that Codex appends for Qwen3-VL backbones (Appendix~\ref{sec:appendix_baselines_free}). Their predictions are scored with the same evaluation pipeline and judges. Boxes that a baseline returns in a different coordinate frame, such as pixel coordinates of the model's resized input, are converted to our normalized format before scoring. Every baseline is evaluated on the full evaluation split. On Qwen3-VL, OctoTools and Visual Sketchpad follow the same three-run protocol as \ours (Appendix~\ref{sec:appendix_std}); runs with proprietary backbones and the training-based baselines are single runs. A request that fails or times out counts as incorrect.

\begin{table}[!htbp]
\centering
\small
\setlength{\tabcolsep}{4pt}
\begin{tabular}{@{}lllc@{}}
\toprule
\textbf{Method} & \textbf{Backbone} & \textbf{Tools} & \textbf{Search} \\
\midrule
\multicolumn{4}{l}{\textit{Training-free}} \\
OctoTools & Qwen3-VL-8B/32B & Generalist solution generator & \texttimes \\
Visual Sketchpad & Qwen3-VL-8B/32B, GPT-6 Luna & Detection, SoM, depth, zoom, overlay & \texttimes \\
Codex & Qwen3-VL-8B/32B, GPT-6 Luna & Code execution (GPT-6 Luna only; see text) & \texttimes \\
\midrule
\multicolumn{4}{l}{\textit{Training-based (released checkpoints)}} \\
AdaReasoner & AdaReasoner-7B & Point, OCR, crop, drawing & \texttimes \\
PixelReasoner & PixelReasoner-RL-v1 (7B) & Zoom-in, frame selection & \texttimes \\
DeepEyesV2 & DeepEyesV2-7B & Code execution & \texttimes \\
\midrule
\ours & Matched to baseline & Nine tools (Table~\ref{tab:appendix_base_tools}) & \checkmark \\
\bottomrule
\end{tabular}
\caption{Configurations of the reproduced baselines. \emph{Search} marks access to a web-search tool. All baselines except Codex decode greedily; Codex uses its own request settings.}
\label{tab:appendix_baselines}
\end{table}

\subsection{Training-Free Harnesses}
\label{sec:appendix_baselines_free}

The training-free harnesses run on the same model endpoints as \ours, so any difference reflects the harness rather than the model.

\paragraph{OctoTools.} OctoTools~\citep{lu2025octotools} is a planner--executor framework in which a planner analyzes the query and proposes the next step, an executor turns it into a tool command, and a verifier decides whether to stop. We use the official implementation and route every language-model role, including the planner, executor, verifier, and the Generalist Solution Generator tool, to the Qwen3-VL backbone. We use its base toolset, the Generalist Solution Generator, with a budget of ten steps and greedy decoding. Because OctoTools accepts a single image per query, multi-image inputs are tiled into one composite with labeled panels (``Image~1'', ``Image~2'', \ldots). The final answer is taken from its direct-output mode.

\paragraph{Visual Sketchpad.} Visual Sketchpad~\citep{hu2024visual} prompts the model to write Python that calls vision tools and draws on the image, in a ReAct-style loop. We use the official implementation with the Qwen3-VL backbone as its single planning agent and its vision-task toolset: open-vocabulary detection with Grounding DINO~\citep{liu2023groundingdino}, sliding-window detection, Set-of-Mark segmentation~\citep{yang2023setofmark}, depth estimation with Depth Anything~\citep{yang2024depthanything}, zoom-in, and image overlay, all served with the original expert models. The agent may take up to ten replies with greedy decoding, receives multiple images as separate inputs, and ends with an \texttt{ANSWER:} line from which we read the prediction. For grounding tasks, it is instructed to convert its detector's boxes to our output format.

\paragraph{Codex.} Codex~\citep{codex} (v0.155.1) is OpenAI's general-purpose coding agent. We run it non-interactively (\texttt{codex exec}) in code mode, with a fresh, ephemeral session for every question, the question's images attached as inputs, and its final agent message taken as the prediction. Web search, browser use, apps, plugins, image generation, and sub-agents are disabled, and model-written code runs in a read-only sandbox. With GPT-6 Luna, Codex calls the OpenAI Responses API at medium reasoning effort and receives the question text unchanged. Codex's default sandbox cannot start shell processes on our evaluation host. For the GPT-6 Luna runs, we therefore enforce the same read-only policy with Linux Landlock and seccomp, under which model-written code executes (Table~\ref{tab:luna_codex_execution} reports how often); the Qwen3-VL runs used the default sandbox and could evaluate only simple expressions (Appendix~\ref{sec:appendix_codex_qwen}). With Qwen3-VL-8B and Qwen3-VL-32B, it calls the same vLLM endpoints as \ours through a small proxy that exposes Codex's code tool as a standard function tool, which our vLLM server otherwise fails to stream, and each response is capped at $2{,}048$ tokens. For the Qwen3-VL runs, Codex also appends a short instruction to the question: answer directly, give the option letter for multiple-choice questions, and use the image as evidence. 
A question whose request fails or exceeds the time limit counts as incorrect; the limit is $300$ seconds with Qwen3-VL and $1{,}800$ seconds with GPT-6 Luna. With Qwen3-VL, LLM-judged benchmarks, including the search benchmarks, are scored with the Qwen3-VL-8B judge; with GPT-6 Luna, scoring follows Table~\ref{tab:harness-combined}, so the search benchmarks are scored with the GPT-4o judge.


\subsection{Training-Based Harnesses}
\label{sec:appendix_baselines_trained}

For the training-based baselines, the tool-use behavior lives in the fine-tuned weights, so we evaluate each method's released checkpoint inside its authors' inference loop rather than re-training it. We followed a two-stage reproduction procedure. First, we executed each repository's own reproduction workflow---including its released environment, model checkpoint, prompt and image preprocessing, tool implementations, and agent loop---until the baseline ran correctly on its documented benchmarks. Second, we added a lightweight HTTP serving layer around that verified path. The serving layer standardizes request transport, process lifecycle, and input normalization, while the model-specific reasoning logic remains in the released implementation. Each service accepts a question and one or more ordered images, returns a final prediction and an agent trajectory, and keeps the model and tool processes resident across examples.

All hosted baselines are evaluated through a shared external harness. For each benchmark example, the harness invokes the dataset's prompt builder, converts the resulting text and ordered images into a common request, and records the returned prediction. The harness controls outer concurrency and, after inference, writes the prediction file expected by the benchmark evaluator and invokes the dataset's native \texttt{evaluate} method using the same configured language-model judge where required. Crucially, scoring is performed entirely outside the baseline services: no service inspects ground truth, selects an evaluator, or computes accuracy. This separation ensures that all baselines are scored through a single consistent protocol. We compare them against \ours on Qwen2.5-VL-7B-Instruct, the base model all three were fine-tuned from (Table~\ref{tab:harness-qwen25}).

\paragraph{AdaReasoner.} AdaReasoner~\citep{song2026adareasoner} trains the model to adaptively compose tools. We use the AdaReasoner-7B checkpoint and the authors' \texttt{BaseToolInferencer}, which exposes pointing (backed by Molmo~\citep{deitke2025molmo}), OCR, cropping, and drawing and path-planning tools through \texttt{<tool\_call>} blocks, for up to six rounds. AdaReasoner is the only baseline requiring two Conda environments: the model/vLLM stack and the GPU PaddleOCR tool workers have incompatible tested dependencies. The launcher starts a tool controller and dedicated Point, OCR, and Crop workers alongside the model server.

\paragraph{PixelReasoner.} PixelReasoner~\citep{wang2025pixel} is trained to reason in pixel space. We use the PixelReasoner-RL-v1 checkpoint and initialize the released Ray/vLLM engines, multimodal processor, and function-call prompt once at startup. The service retains the released tool parser and pixel-space operations (normalized cropping and frame selection). Inference proceeds at the granularity of model/tool rounds: ready requests are batched across vLLM engines, generated tool calls are executed, resulting crops or selected frames are appended to the conversation, and unfinished requests are requeued. The final answer is read from \texttt{\textbackslash boxed\{\}}. The documented image-resolution budgets (\texttt{MIN\_PIXELS}=401408, \texttt{MAX\_PIXELS}=4014080) are preserved as they affect how much visual information reaches the model.

\paragraph{DeepEyesV2.} DeepEyesV2~\citep{hong2025deepeyesv2} is trained for agentic multimodal reasoning with code execution. We use the released DeepEyesV2-7B checkpoint with the agent mode of the authors' vendored VLMEvalKit \texttt{VLLMAPIWrapper}. A single launcher supervises the text-search adapter, Python code service, one or more OpenAI-compatible vLLM model replicas, and the public inference API. The wrapper controls model calls, Python execution, retrieval turns, stopping rules, and answer extraction. All ordered input images are bound in the Python sandbox as \texttt{image\_1}, \texttt{image\_2}, \ldots\ so that generated code can address every input.

\paragraph{Fair-comparison controls.} For a defensible comparison, the following are held fixed or explicitly reported across all baselines: dataset version, split, and prompt builder; model checkpoint and quantization; image preprocessing and resolution limits; context window, per-call generation cap, temperature, and maximum agent/tool rounds; enabled tools and external search backend; concurrency and failure-retry policy; judge model, judge prompt, and judge retry count; and whether failures are scored as incorrect or retried.

\section{Run-to-Run Variability}
\label{sec:appendix_std}

Agentic evaluation is stochastic: the backbone samples at a non-zero temperature, the harness makes further model calls inside its tools, and live search results change over time. This section describes how we account for this variability and quantifies it for the Qwen3-VL results in the main paper.

\paragraph{Protocol.} Every Qwen3-VL number in the main paper, including the Base, OctoTools, Visual Sketchpad, and \ours columns of Table~\ref{tab:harness-combined} and every entry of Table~\ref{tab:ablation}, is the mean of three independent runs. Each run uses a different random seed, which controls both the backbone's sampling and every stochastic component of the harness, such as the model calls made inside the Webpage Visit and Python Coding Agent tools. All other factors are held fixed across runs: the benchmark instances, prompts, tool backends, decoding parameters, iteration and context limits, and the scorers. Judges and answer extractors decode greedily, so the variation we measure comes from the agent rather than from scoring. We report all three runs of every configuration; no run is discarded or re-run, and a request that fails after retries is scored as incorrect in the run where it occurs. Search tools query the live web and cannot be seeded, so for search benchmarks the variation also reflects changes in retrieved content between runs. Results with proprietary backbones (GPT-5, GPT-6 Luna, GPT-6 Sol, and GPT-6 Astra) are single runs because of API cost.

\paragraph{Reported statistics.} Tables~\ref{tab:appendix_std_main} and~\ref{tab:appendix_std_ablation} report, for each entry, the standard deviation of the score across the three runs. For averages over datasets, we first compute the average within each run and then take the standard deviation of these three averages; this is smaller than the per-dataset deviations because independent fluctuations partly cancel. With only three runs, each standard deviation is itself a coarse estimate and should be read as indicating the scale of run-to-run variation rather than an exact value.

\input{tables/std_main}

\paragraph{Main results.} Per-dataset standard deviations are small relative to the differences between methods (Table~\ref{tab:appendix_std_main}). For \ours, most are below $2$ points; the largest occur on search benchmarks with live retrieval, such as MMSearch ($171$ questions, where one question is worth $0.58$ points) and FVQA with Qwen3-VL-8B. \ours varies somewhat more than Base, as expected for multi-turn trajectories in which an early difference in a tool call can change the rest of the trajectory. The $22$-dataset averages are very stable, with standard deviations of at most $0.35$ points. The gain of \ours over Base in this average ($+11.2$ at 8B and $+12.6$ at 32B) is therefore about $29$ and $44$ times the combined standard deviation of the two averages, so the improvement cannot be attributed to run-to-run variation.

\input{tables/std_ablation}

\paragraph{Ablation.} The overall averages in Table~\ref{tab:ablation} have standard deviations between $0.2$ and $0.35$ points for every variant and backbone (Table~\ref{tab:appendix_std_ablation}). Adding tools, the capability layer, and multimodal memory each changes the overall average by several points, far beyond this variation. The textual-memory step ($+0.7$ at 8B and $+0.3$ at 32B) is comparable to run-to-run variation, which is consistent with our conclusion that text descriptions of images are a weak substitute for retaining the images themselves.

\paragraph{Single-run results.} Proprietary backbones are evaluated once, so their per-dataset numbers carry run-to-run variation of the scale measured above. Their conclusions, however, rest on large differences: \ours improves the $22$-dataset average by $+10.8$ points with GPT-6 Luna, $+11.6$ with GPT-6 Sol, and $+9.9$ with GPT-6 Astra, more than twenty-five times the largest average-level standard deviation we observe with Qwen3-VL.

\section{Additional Related Work}
\label{sec:appendix_related}
\paragraph{Multimodal search agents.} Our work targets multimodal search, where a query is resolved by orchestrating reasoning, perception, and live web retrieval across many turns. A growing collection of benchmarks frames this task: MMSearch \citep{mmsearch} couples a multimodal query with a curated search engine; MMSearch-Plus \citep{mmsearchplus} adds provenance-aware verification; BrowseComp-VL \citep{browsecompvl} and MM-BrowseComp \citep{mmbrowsecomp} raise cross-modal information-seeking difficulty; VisBrowse-Bench \citep{visbrowsebench} targets visual-native search; and RealX-Bench \citep{hong2025deepeyesv2} couples fine-grained perception with multi-hop retrieval. Beyond pure search, the landscape extends to embodied tasks that integrate web information with physical action \citep{hong2506embodied}. Recent browsing agents tackle these benchmarks via hand-engineered tool inventories \citep{browsecompvl,vdr,mmdeepresearch,opensearchvl,mmsearchr1,deepmmsearchr1}, supervised fine-tuning or reinforcement learning that teaches tool use, or memory learned over a fixed inventory \citep{xiao2025toolmem,hasegawa2025tama}. Our contribution is orthogonal: rather than freeze the inventory, we \emph{generate} it.

\paragraph{Grounded perception.} Class-agnostic counting was popularized by FSC-147~\citep{ranjan2021learning}, which asks for the number of objects of an arbitrary category; specialized counters such as CounTR~\citep{liu2022countr}, LOCA~\citep{djukic2023loca}, and CountGD~\citep{amininaieni2024countgd} regress density maps or detections, and FSCD~\citep{nguyen2022few} extends the task to predicting a box for every instance. Open-vocabulary detectors and segmenters, including GLIP~\citep{li2022grounded}, Grounding DINO~\citep{liu2023groundingdino}, and SAM~3~\citep{sam3}, localize arbitrary text-specified categories, and MLLMs have begun to count by pointing~\citep{deitke2025molmo}. Nevertheless, MLLMs remain unreliable once scenes contain more than a handful of objects~\citep{paiss2023countbench,tamarapalli2025countqa}. Tool-use systems such as ViperGPT~\citep{suris2023vipergpt}, VisProg~\citep{gupta2023visual}, and Visual Sketchpad~\citep{hu2024visual} let a language model call detectors and count their outputs programmatically. \ours follows this direction but keeps grounding outputs as persistent evidence: boxes, masks, and scores are stored as typed bindings that the Python Coding Agent can filter, deduplicate, and aggregate across turns, instead of a count that is read once from text.

\paragraph{Spatial reasoning.} Spatial reasoning benchmarks test models beyond static single-image relations. EmbSpatial-Bench~\citep{du2024embspatial} and ERQA~\citep{team2025gemini} target spatial and embodied reasoning in simulated and real-world scenes; SparBench~\citep{zhang2026flatland} covers diverse single- and multi-view spatial tasks; and MindCube~\citep{wang2025mindcube} and MMSI-Bench~\citep{yang2025mmsi} emphasize reasoning across multiple or limited views. In parallel, VGGT~\citep{wang2025vggt} and subsequent VGGT-style geometric foundation models~\citep{hu2026g} recover explicit scene geometry, including depth, camera pose, point maps, and tracks from visual observations. OpenVLHarness treats such geometric models as callable capabilities rather than requiring the MLLM itself to internalize 3D reconstruction, allowing their outputs to be composed with grounding and other persistent evidence during reasoning.

\paragraph{Experience learning and harness optimization.}
Learning from past experience has a long history in reinforcement learning, exemplified by Hindsight Experience Replay~\citep{andrychowicz2017hindsight}, which relabels failed rollouts as successes, STaR~\citep{zelikman2022star}, which bootstraps a reasoner on its own correct rationales, and ARMAP~\citep{chen2025scaling}, which learns a reward model from an agent's own trajectories to guide its planning. A newer line treats memory as a learning signal, distilling reusable knowledge from cross-sample working memory, though these methods often require additional training or model-weight updates~\citep{packer2023memgpt, yan2025memory, zhou2025mem1, wang2024agent, yang2025verbalized, ouyang2026reasoningbank, zhang2025memevolve, yan2025general, ren2026aligning, shi2026experientialrl, hong2026learning, jiang2026memory}. A third line instead optimizes prompts and harnesses without any model-weight updates, using LLM- and coding-agent-driven optimizers that rewrite prompts, pipelines, and harness code from observed failures~\citep{khattab2024dspy, textgrad, gepa, lee2026meta, yuksekgonul2026learning, openevolve}, but to date this is done purely in the text-only prompt and harness domain.







%% file: text/appendix_tool_use.tex
\section{Tool-Use Behavior}
\label{sec:appendix_tool_use}

Table~\ref{tab:harness-combined} reports the accuracy of \ours. This section characterizes the behavior
that produces it: how frequently the agent invokes a tool, which capabilities it invokes, and how those
invocations relate to accuracy relative to the same model without tools. The analysis covers five base
models --- Qwen3-VL-8B, Qwen3-VL-32B, GPT-6~Luna, GPT-6~Sol, and GPT-6~Astra, the GPT-6 models at medium reasoning effort --- and reads every rollout of the \ours
runs behind Tables~\ref{tab:harness-combined} and~\ref{tab:harness-kimi}, approximately $123{,}000$ rollouts containing $224{,}000$
tool calls. All counts are exact rather than sampled. A question is \emph{engaged} if the agent emitted
at least one tool call on it, regardless of whether the call succeeded.

\paragraph{Measurement protocol.} Section~\ref{sec:appendix_tool_value} compares \ours against the
\textbf{Base} columns of
Tables~\ref{tab:harness-combined} and~\ref{tab:harness-kimi} on a per-question basis: both arms answer the same questions and are
scored with the same dataset metric, and differences are averaged over paired rows. This design supports
statements about individual capabilities, since the rows on which a given tool was invoked can be
isolated and both arms evaluated on that subset. The resulting quantities are correlational rather than
causal: the agent selects the rows on which it invokes a tool, and the remaining capabilities are
available on those rows as well. Each per-tool measurement is therefore reported alongside a control,
namely the same paired difference computed on the rows that did not invoke the tool. A capability whose
two values coincide is not, on its own, responsible for the observed gain. The control is a complement
set, so its composition depends on how often the capability is invoked: for a capability used on a small
fraction of questions the control approximates the model's overall gain, whereas for one used on the
majority of questions the control is a small and atypical remainder.

\subsection{Tool-Call Volume and Engagement}
\label{sec:appendix_tool_demand}

Two quantities characterize how heavily a model uses the tools, and they vary independently: the number
of calls a question costs, and the fraction of questions answered with no call.
Table~\ref{tab:appendix_tool_profile} reports them per model, and Figure~\ref{fig:appendix_tool_demand}
resolves them by domain.

\input{figs/tool_use_demand}

\begin{table}[!htbp]
\centering
\small
\input{tables/tool_use_profile}
\caption{\textbf{Tool-use profile of \ours across five base models.} Calls per question averages over all
questions, including those answered without tools; calls per engaged question conditions on questions
with at least one call. Messages count transcript turns, including the Environment Update emitted after
each call. Repeat calls are byte-identical to an earlier call in the same rollout; failed calls are those
the tool server returned as errors.}
\label{tab:appendix_tool_profile}
\end{table}

\paragraph{Volume and engagement.} Engagement varies with model scale within a family but not across
families. The smaller models leave a substantial fraction of questions untouched (Qwen3-VL-8B $28.9\%$,
GPT-6~Luna $25.3\%$), whereas the larger models rarely do ($6.2\%$, $4.8\%$ and $2.6\%$). Call volume does not
order the models by accuracy: GPT-6~Luna issues the fewest calls per question of the five and attains
higher accuracy than either Qwen model in Table~\ref{tab:harness-combined}. By domain, grounded perception
engages all five models on nearly every question, while spatial reasoning and general VQA and
hallucination account for most of the unengaged questions in the smaller models.

\paragraph{Reliability and context cost.} Fewer than $1\%$ of calls return a tool error for any model, so
execution reliability is not a limiting factor. Repeated calls, defined as byte-identical invocations
within one rollout, are more frequent than errors but remain a small fraction of traffic and do not scale
with model size. The dominant cost is context: a rollout occupies between $8$ and $14$ transcript turns,
between two and $3.5$ of which are Environment Updates, one emitted at the start and one after each
call.

\subsection{Distribution of Tool Calls}
\label{sec:appendix_tool_mix}

\input{figs/tool_use_mix}

\input{figs/tool_use_pies}

\paragraph{Aggregate distribution.} Visual Grounding accounts for the largest share of calls in all five
models, ranging from $45\%$ for GPT-6~Sol to $63\%$ for Qwen3-VL-32B. The models differ mainly in the
second-ranked capability. GPT-6~Sol and GPT-6~Astra allocate $20\%$ and $18\%$ of their calls to Zoom-in, against $5$--$9\%$ for the
other models, corresponding to a ground-then-inspect pattern. The GPT-6 models allocate a larger share to
web retrieval ($17\%$ to Text Search for Luna, $13\%$ for Sol, and $9\%$ for Astra), concentrated in the visual search and deep research domain. Depth Estimation, Camera
Trajectory and the Python Coding Agent are small in aggregate and concentrated in the spatial reasoning datasets.

\paragraph{Distribution by domain.} Figure~\ref{fig:appendix_tool_pies} resolves the same budget by
domain. Grounded perception is dominated by Visual Grounding for every model, at $90$--$94\%$ of
domain calls, with the exception of GPT-6~Sol and GPT-6~Astra, which allocate roughly a quarter of their calls in that
domain to Zoom-in. Visual search and deep research is the only domain in which the three web capabilities dominate. Spatial
reasoning is the one domain in which four capabilities carry comparable load, combining grounding, camera
trajectory, depth and the coding agent, and general VQA and hallucination divides between grounding and inspection.

\subsection{Effect of Tool Use on Accuracy}
\label{sec:appendix_tool_value}

\begin{table}[!htbp]
\centering
\small
\input{tables/tool_use_impact}
\caption{\textbf{\ours against the base model, paired per question.} Rescued counts questions answered
incorrectly by the base model and correctly by \ours; broken counts the converse. Calls per net gain
divides the total number of tool calls issued on these questions by the net number of questions whose
outcome changed.}
\label{tab:appendix_tool_impact}
\end{table}

\paragraph{Aggregate effect.} All five models improve under \ours, by $+6.2$ (GPT-6~Astra) to $+10.1$
(Qwen3-VL-8B) points. The net figure is the balance of two considerably larger flows: between $2{,}336$
and $4{,}649$ questions per model are corrected, while between $700$ and $2{,}146$ are degraded.
Normalizing by the tool calls spent, the cost of a net corrected question ranges from $11.7$ calls
(GPT-6~Luna) to $28.1$ calls (Qwen3-VL-32B), the latter combining the highest call volume with one of the
smaller gains.

\paragraph{Null control.} On questions where the agent issued no tool call, \ours reduces to the base
model with a longer system prompt, so the expected paired difference is zero. The measured values are
$-0.2$ (GPT-6~Sol), $-0.5$ (GPT-6~Luna), $-1.6$ (Qwen3-VL-32B) and $-1.8$ (Qwen3-VL-8B) points, and $+1.8$ for
GPT-6~Astra, which leaves only $2.6\%$ of questions without a call, so its value rests on $614$ questions. The
proximity to zero supports the interpretation of the remaining differences as effects of tool use rather than of
the prompt or the evaluation protocol, and the two larger negative values, both from the Qwen3-VL models, indicate
a small cost to carrying the tool interface when it is not used.

\input{figs/tool_use_value}

\paragraph{Per-capability effects.} Figure~\ref{fig:appendix_tool_value} reports, for each capability
and model, the paired difference on the questions that invoked it.
Table~\ref{tab:appendix_tool_tools} gives the underlying counts, the share of questions each capability
was invoked on, the accuracies of both arms, and the control, that is the same difference on the
questions that did not invoke the capability. The two retrieval
capabilities exhibit the largest separation from their controls in every model: rows invoking Image
Search improve by $18.9$ to $27.1$ points, while the corresponding control values lie between $-0.6$ and
$+6.6$. Visual Grounding is the dominant factor for the smaller models --- Qwen3-VL-8B improves by $14.3$
points on rows that ground and declines by $20.1$ points on rows that do not --- and flattens for the
larger models, which invoke it on nearly all questions. Camera Trajectory is positive for all five
models. Depth Estimation, OCR, Zoom-in and the Python Coding Agent are at or below their controls, apart
from Zoom-in for GPT-6~Luna ($+11.9$ against $+8.6$) and GPT-6~Astra ($+9.4$ against $+5.2$), which
is consistent with their being invoked on the questions a model cannot resolve from the image alone
rather than with their being ineffective in isolation.

\begin{table}[!htbp]
\centering
\small
\setlength{\tabcolsep}{4pt}
\resizebox{\columnwidth}{!}{%
\input{tables/tool_use_tools}}
\caption{\textbf{Per-capability paired comparison, corresponding to
Figure~\ref{fig:appendix_tool_value}.} For each capability and model: the questions whose rollout invoked
it at least once, the calls issued on them, the accuracy of the base model and of \ours on exactly those
questions, their difference, and the same difference computed on the questions that did not invoke the
capability. Share is the fraction of paired questions that invoked it, which determines how closely the
final column approximates the model's overall gain. A question that invoked three capabilities is counted
once under each.}
\label{tab:appendix_tool_tools}
\end{table}

\subsection{Capability Composition}
\label{sec:appendix_tool_shape}

\paragraph{Composition across capabilities.} Rollouts frequently combine several capabilities rather
than invoking one in isolation. Counting the number of \emph{distinct} capabilities per rollout, among
engaged rollouts $31.9\%$ (Qwen3-VL-32B), $26.8\%$ (GPT-6~Sol) and $26.0\%$ (GPT-6~Astra) invoke two or more, against $20.7\%$
and $11.6\%$ for the smaller models, and the longest rollouts combine four or five. Combinations that
cross capability families are rarer, $4$--$16\%$ of engaged rollouts, and consist almost entirely of a
perception capability followed by the Python Coding Agent: ground or measure first, then compute over the
resulting quantities. This is the pattern rewarded by the counting, grounding and spatial datasets, and
it is not expressible in a harness that exposes a single capability.

%% file: figs/tool_use_demand.tex
\begin{figure}[!htbp]
    \centering
    \includegraphics[width=\linewidth]{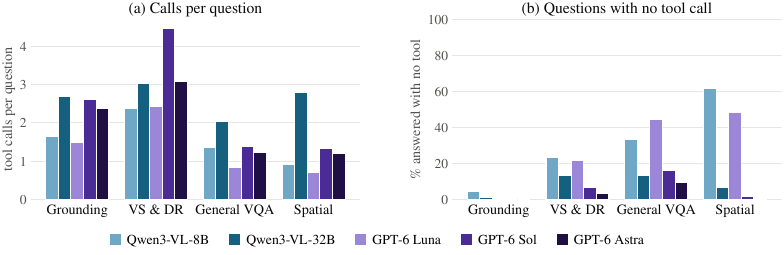}
    \caption{\textbf{Tool-call volume and engagement by domain.} \textbf{(a)} Tool calls per question,
    averaged over all questions in the domain. \textbf{(b)} Fraction of questions answered with no tool
    call. Engagement varies with model scale within a family and not across families, and call volume
    does not order the models by accuracy.}
    \label{fig:appendix_tool_demand}
\end{figure}

%% file: tables/tool_use_profile.tex
\begin{tabular}{@{}lrrrrrr@{}}
\toprule
\textbf{Model} & \textbf{Calls /} & \textbf{Calls / engaged} & \textbf{No tool} & \textbf{Messages /} & \textbf{Repeat} & \textbf{Failed} \\
 & \textbf{question} & \textbf{question} & \textbf{(\%)} & \textbf{question} & \textbf{calls (\%)} & \textbf{calls (\%)} \\
\midrule
Qwen3-VL-8B & 1.43 & 2.00 & 28.9 & 9.7 & 1.92 & 0.59 \\
Qwen3-VL-32B & 2.59 & 2.76 & 6.2 & 14.3 & 0.54 & 0.59 \\
GPT-6 Luna & 1.25 & 1.67 & 25.3 & 7.7 & 0.38 & 0.30 \\
GPT-6 Sol & 2.07 & 2.18 & 4.8 & 10.2 & 1.95 & 0.59 \\
GPT-6 Astra & 1.80 & 1.84 & 2.6 & 9.4 & 0.01 & 0.17 \\
\bottomrule
\end{tabular}

%% file: figs/tool_use_mix.tex
\begin{figure}[!htbp]
    \centering
    \includegraphics[width=\linewidth]{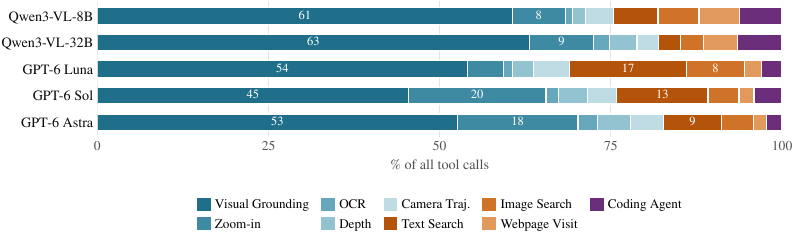}
    \caption{\textbf{Distribution of tool calls across capabilities.} All calls issued by each model,
    normalized to 100\%. Visual Grounding is the largest consumer for every model; the models differ
    mainly in the second-ranked capability.}
    \label{fig:appendix_tool_mix}
\end{figure}

%% file: figs/tool_use_pies.tex
\begin{figure}[!htbp]
    \centering
    \includegraphics[width=0.92\linewidth]{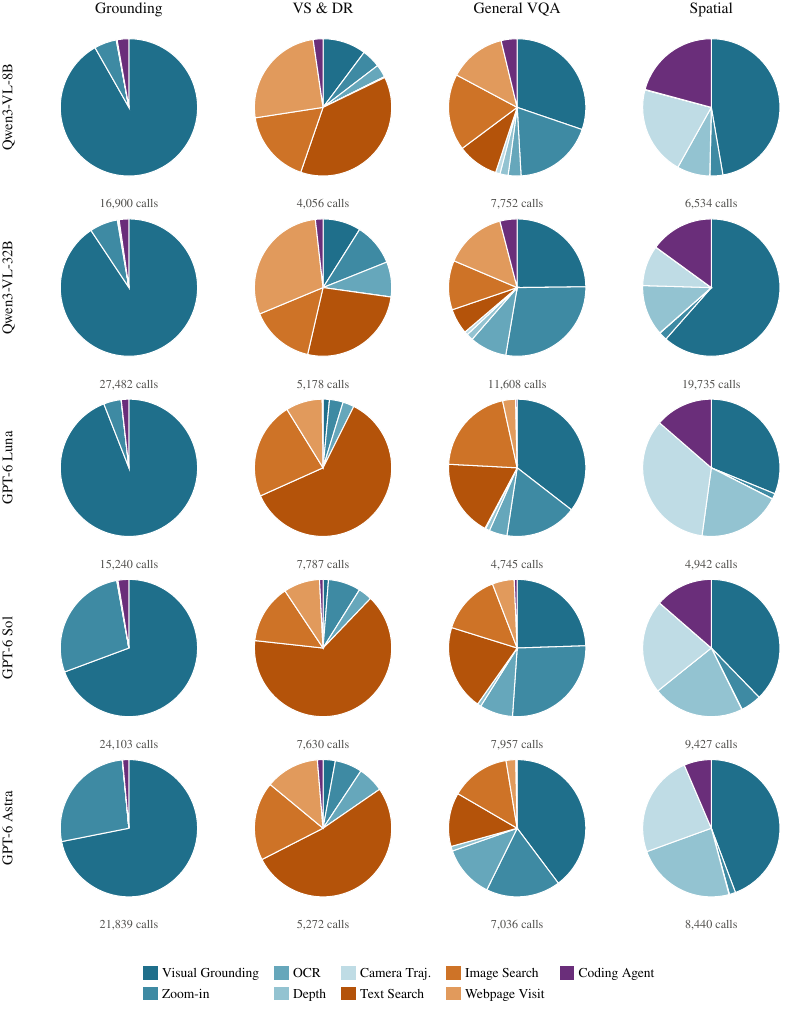}
    \caption{\textbf{Distribution of tool calls by domain.} Calls issued in each domain as a share of
    that domain's calls, with the call total below each panel. Colors are consistent across panels.
    Grounded perception is dominated by Visual Grounding for every model except GPT-6~Sol and GPT-6~Astra; visual
    search and deep research is the only domain dominated by the web capabilities; spatial reasoning is the only domain in which four
    capabilities carry comparable load.}
    \label{fig:appendix_tool_pies}
\end{figure}

%% file: tables/tool_use_impact.tex
\begin{tabular}{@{}lrrrrrrr@{}}
\toprule
\textbf{Model} & \textbf{Rows} & \textbf{Base} & \textbf{\ours} & $\Delta$ & \textbf{Rescued} & \textbf{Broken} & \textbf{Calls per} \\
 & & & & & & & \textbf{net gain} \\
\midrule
Qwen3-VL-8B & 24{,}410 & 50.02 & 60.09 & $+$10.07 & 4{,}649 & 1{,}967 & 12.9 \\
Qwen3-VL-32B & 24{,}393 & 53.73 & 62.17 & $+$8.44 & 4{,}381 & 2{,}146 & 28.1 \\
GPT-6 Luna & 26{,}226 & 57.88 & 67.67 & $+$9.79 & 4{,}191 & 1{,}382 & 11.7 \\
GPT-6 Sol & 23{,}700 & 64.26 & 72.67 & $+$8.41 & 3{,}153 & 1{,}025 & 23.1 \\
GPT-6 Astra & 23{,}700 & 70.74 & 76.98 & $+$6.24 & 2{,}336 & 700 & 26.0 \\
\bottomrule
\end{tabular}

%% file: figs/tool_use_value.tex
\begin{figure}[!htbp]
    \centering
    \includegraphics[width=\linewidth]{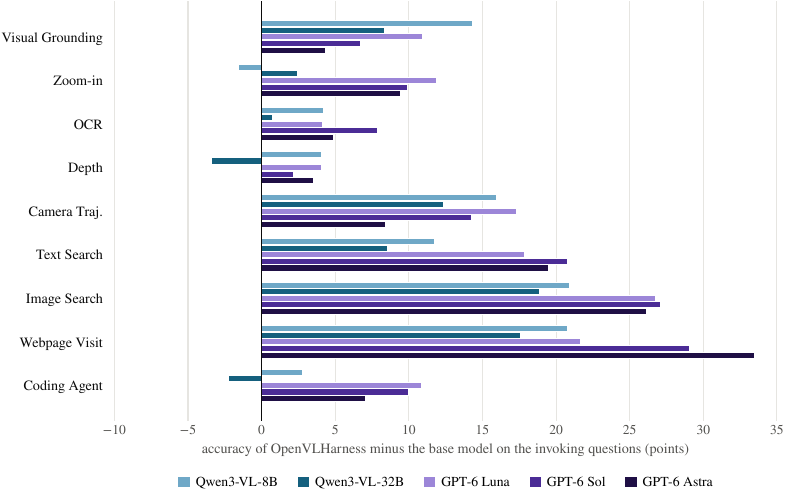}
    \caption{\textbf{Per-capability paired comparison against the base model.} For each capability and
    model, the accuracy of \ours minus that of the base model on the questions whose rollout invoked that
    capability. The corresponding difference on the questions that did not invoke it, which serves as the
    control, is given in the last column of Table~\ref{tab:appendix_tool_tools} together with the share
    of questions each capability was invoked on.}
    \label{fig:appendix_tool_value}
\end{figure}

%% file: tables/tool_use_tools.tex
\begin{tabular}{@{}llrrrrrrr@{}}
\toprule
\textbf{Capability} & \textbf{Model} & \textbf{Questions} & \textbf{Share} & \textbf{Calls} & \textbf{Base} & \textbf{\ours} & $\Delta$ & $\Delta$ \textbf{elsewhere} \\
\midrule
\multirow{5}{*}{Visual Grounding} & Qwen-8B & 13{,}103 & 54\% & 21{,}286 & 47.7 & 62.0 & $+$14.3 & $-$20.1 \\
 & Qwen-32B & 17{,}284 & 71\% & 40{,}057 & 56.1 & 64.5 & $+$8.3 & $+$0.2 \\
 & Luna & 13{,}216 & 50\% & 17{,}670 & 56.5 & 67.4 & $+$10.9 & $+$2.7 \\
 & Sol & 14{,}369 & 61\% & 22{,}313 & 67.5 & 74.3 & $+$6.7 & $+$2.2 \\
 & Astra & 15{,}488 & 65\% & 22{,}385 & 73.6 & 77.9 & $+$4.3 & $+$5.1 \\
\addlinespace[2pt]
\multirow{5}{*}{Zoom-in} & Qwen-8B & 1{,}703 & 7\% & 2{,}714 & 40.3 & 38.8 & $-$1.6 & $+$4.8 \\
 & Qwen-32B & 3{,}214 & 13\% & 5{,}805 & 45.1 & 47.6 & $+$2.4 & $+$7.8 \\
 & Luna & 1{,}551 & 6\% & 1{,}745 & 34.8 & 46.7 & $+$11.9 & $+$8.6 \\
 & Sol & 6{,}025 & 25\% & 9{,}866 & 48.4 & 58.2 & $+$9.9 & $+$9.4 \\
 & Astra & 5{,}351 & 23\% & 7{,}500 & 51.6 & 61.1 & $+$9.4 & $+$5.2 \\
\addlinespace[2pt]
\multirow{5}{*}{OCR} & Qwen-8B & 312 & 1\% & 343 & 50.6 & 54.8 & $+$4.2 & $+$7.5 \\
 & Qwen-32B & 1{,}259 & 5\% & 1{,}433 & 60.8 & 61.5 & $+$0.8 & $+$8.4 \\
 & Luna & 378 & 1\% & 414 & 41.3 & 45.4 & $+$4.1 & $+$12.7 \\
 & Sol & 863 & 4\% & 891 & 62.4 & 70.2 & $+$7.9 & $+$12.0 \\
 & Astra & 1{,}195 & 5\% & 1{,}201 & 73.9 & 78.8 & $+$4.8 & $+$10.7 \\
\addlinespace[2pt]
\multirow{5}{*}{Depth} & Qwen-8B & 598 & 2\% & 663 & 44.1 & 48.2 & $+$4.1 & $+$4.0 \\
 & Qwen-32B & 2{,}187 & 9\% & 2{,}546 & 59.6 & 56.2 & $-$3.4 & $+$4.5 \\
 & Luna & 868 & 3\% & 1{,}018 & 54.9 & 59.0 & $+$4.1 & $+$4.4 \\
 & Sol & 1{,}854 & 8\% & 2{,}101 & 63.1 & 65.2 & $+$2.1 & $+$6.9 \\
 & Astra & 1{,}933 & 8\% & 2{,}071 & 65.2 & 68.8 & $+$3.5 & $+$4.6 \\
\addlinespace[2pt]
\multirow{5}{*}{Camera Traj.} & Qwen-8B & 1{,}444 & 6\% & 1{,}457 & 34.6 & 50.6 & $+$15.9 & $+$3.0 \\
 & Qwen-32B & 1{,}925 & 8\% & 1{,}975 & 39.0 & 51.4 & $+$12.3 & $+$4.2 \\
 & Luna & 1{,}683 & 6\% & 1{,}699 & 45.5 & 62.8 & $+$17.3 & $+$0.4 \\
 & Sol & 1{,}969 & 8\% & 2{,}087 & 56.0 & 70.3 & $+$14.2 & $+$3.5 \\
 & Astra & 1{,}991 & 8\% & 2{,}034 & 73.2 & 81.6 & $+$8.4 & $+$3.6 \\
\addlinespace[2pt]
\multirow{5}{*}{Text Search} & Qwen-8B & 1{,}238 & 5\% & 2{,}048 & 26.1 & 37.9 & $+$11.7 & $+$8.7 \\
 & Qwen-32B & 1{,}353 & 6\% & 1{,}826 & 34.9 & 43.5 & $+$8.6 & $+$9.5 \\
 & Luna & 2{,}009 & 8\% & 5{,}599 & 36.9 & 54.8 & $+$17.9 & $+$9.7 \\
 & Sol & 1{,}801 & 8\% & 6{,}506 & 39.3 & 60.1 & $+$20.8 & $+$9.3 \\
 & Astra & 1{,}556 & 7\% & 3{,}639 & 49.7 & 69.2 & $+$19.5 & $+$8.3 \\
\addlinespace[2pt]
\multirow{5}{*}{Image Search} & Qwen-8B & 1{,}885 & 8\% & 1{,}937 & 32.3 & 53.2 & $+$20.9 & $-$0.6 \\
 & Qwen-32B & 1{,}922 & 8\% & 1{,}974 & 35.6 & 54.5 & $+$18.9 & $+$1.3 \\
 & Luna & 2{,}604 & 10\% & 2{,}769 & 34.6 & 61.4 & $+$26.7 & $+$2.8 \\
 & Sol & 1{,}923 & 8\% & 2{,}236 & 37.6 & 64.7 & $+$27.1 & $+$6.6 \\
 & Astra & 1{,}766 & 7\% & 1{,}968 & 47.9 & 74.1 & $+$26.1 & $+$5.1 \\
\addlinespace[2pt]
\multirow{5}{*}{Webpage Visit} & Qwen-8B & 1{,}215 & 5\% & 1{,}936 & 22.1 & 42.9 & $+$20.7 & $+$7.7 \\
 & Qwen-32B & 2{,}177 & 9\% & 3{,}014 & 34.1 & 51.7 & $+$17.6 & $+$0.7 \\
 & Luna & 698 & 3\% & 815 & 19.8 & 41.4 & $+$21.6 & $+$12.6 \\
 & Sol & 834 & 4\% & 1{,}079 & 26.5 & 55.5 & $+$29.1 & $+$12.2 \\
 & Astra & 655 & 3\% & 829 & 28.9 & 62.3 & $+$33.4 & $+$10.3 \\
\addlinespace[2pt]
\multirow{5}{*}{Coding Agent} & Qwen-8B & 1{,}200 & 5\% & 2{,}196 & 37.1 & 39.9 & $+$2.8 & $+$8.8 \\
 & Qwen-32B & 3{,}318 & 14\% & 4{,}121 & 57.2 & 55.0 & $-$2.3 & $+$6.6 \\
 & Luna & 780 & 3\% & 979 & 37.1 & 48.0 & $+$10.9 & $+$9.5 \\
 & Sol & 906 & 4\% & 2{,}018 & 37.5 & 47.4 & $+$10.0 & $+$7.3 \\
 & Astra & 857 & 4\% & 959 & 44.0 & 51.0 & $+$7.0 & $+$8.3 \\
\bottomrule
\end{tabular}

%% file: tables/harness_design_kimi.tex
\begin{table}[!htbp]
\centering
\setlength{\tabcolsep}{5pt}
\renewcommand{\arraystretch}{1.25}
\setlength{\aboverulesep}{0pt}
\setlength{\belowrulesep}{0pt}
\begin{tabular}{cl|>{\centering\arraybackslash}p{1.25cm}>{\centering\arraybackslash}p{1.25cm}|>{\centering\arraybackslash}p{1.25cm}>{\centering\arraybackslash}p{1.25cm}|>{\centering\arraybackslash}p{1.25cm}>{\centering\arraybackslash}p{1.25cm}}
\toprule
\multicolumn{2}{c|}{} & \multicolumn{2}{c|}{\textbf{GPT-5}} & \multicolumn{2}{c|}{{\renewcommand{\arraystretch}{1}\begin{tabular}[t]{@{}c@{}}\textbf{GPT-6 Sol}\\[-1pt]{\footnotesize (medium)}\end{tabular}}} & \multicolumn{2}{c}{{\renewcommand{\arraystretch}{1}\begin{tabular}[t]{@{}c@{}}\textbf{Kimi K3}\\[-1pt]{\footnotesize (max)}\end{tabular}}} \\
\textbf{Domain} & \textbf{Dataset} & \textbf{Base} & \textbf{Ours} & \textbf{Base} & \textbf{Ours} & \textbf{Base} & \textbf{Ours} \\
\midrule
& CountQA         & 38.48 & \textbf{58.70} & 73.00 & \textbf{76.00} & 71.27 & \textbf{73.10} \\
& FSC-147         & 22.16 & \textbf{42.92} & 38.72 & \textbf{44.40} & 36.78 & \textbf{42.07} \\
& OdinW-13-All    & 46.27 & \textbf{63.12} & 40.86 & \textbf{57.88} & 44.35 & \textbf{56.91} \\
& FSCD-147        & 0.16 & \textbf{28.09} & 8.36 & \textbf{29.42} & 7.88 & \textbf{29.11} \\
\multirow{-5}{*}{\rotatebox[origin=c]{90}{\shortstack[c]{Grounded\\Perception}}}
 & FSCD-LVIS       & 2.04 & \textbf{33.22} & 5.17 & \textbf{35.22} & 7.15 & \textbf{33.60} \\
\midrule
& MMSearch        & 36.84 & \textbf{56.14} & 55.56 & \textbf{77.78} & 67.25 & \textbf{86.55} \\
& MMSearch-Plus   & 10.93 & \textbf{31.19} & 16.72 & \textbf{45.66} & 23.47 & \textbf{55.31} \\
& RealX-Bench      & 34.54 & \textbf{40.72} & 36.08 & \textbf{51.55} & 51.03 & \textbf{65.98} \\
& FVQA            & 54.67 & \textbf{55.00} & 58.67 & \textbf{71.00} & 70.00 & \textbf{79.33} \\
& HLE-VL          & 20.00 & \textbf{24.85} & 36.67 & \textbf{43.33} & 41.52 & \textbf{46.67} \\
\multirow{-6}{*}{\rotatebox[origin=c]{90}{\shortstack[c]{Visual Search and\\Deep Research}}}
 & BC-VL           & 42.11 & \textbf{52.38} & 43.86 & \textbf{58.90} & 51.63 & \textbf{64.66} \\
\midrule
& RealWorldQA     & 82.80 & \textbf{83.13} & 84.58 & \textbf{87.45} & 83.8 & \textbf{86.9} \\
& SimpleVQA       & 66.33 & \textbf{68.33} & 60.30 & \textbf{76.40} & 81.33 & \textbf{82.33} \\
& HallusionBench  & 62.95 & \textbf{66.07} & 72.77 & \textbf{75.27} & 69.6 & \textbf{74.9} \\
& FREAK-VQA       & 45.93 & \textbf{51.44} & 54.94 & \textbf{64.08} & 52.6 & \textbf{59.6} \\
& FREAK-MCQ       & 44.37 & \textbf{52.82} & 61.77 & \textbf{70.32} & 52.9 & \textbf{63.1} \\
\multirow{-6}{*}{\rotatebox[origin=c]{90}{\shortstack[c]{General VQA and\\ Hallucination}}}
 & V*              & 72.80 & \textbf{84.81} & 91.10 & \textbf{93.19} & 90.6 & \textbf{92.1} \\
\midrule
& MMSI-Bench      & 41.20 & \textbf{49.80} & 49.80 & \textbf{57.40} & 46.5 & \textbf{57.7} \\
& ERQA            & \textbf{60.00} & 59.00 & 66.50 & \textbf{71.25} & 63.8 & \textbf{66.8} \\
& EmbSpatial      & 81.29 & \textbf{82.47} & 80.60 & \textbf{82.53} & 83.7 & \textbf{83.9} \\
& SparBench       & 50.12 & \textbf{60.09} & 55.36 & \textbf{62.66} & 60.5 & \textbf{63.8} \\
\multirow{-5}{*}{\rotatebox[origin=c]{90}{\shortstack[c]{Spatial \\ Reasoning}}}
 & MindCube        & 60.09 & \textbf{80.00} & 67.40 & \textbf{83.17} & 83.8 & \textbf{89.6} \\
\midrule
{}& {Average}  & 44.37 & \textbf{55.65} & 52.67 & \textbf{64.31} & 56.43 & \textbf{66.09} \\
\bottomrule
\end{tabular}
\caption{Results of \ours with GPT-5~\citep{gpt5}, GPT-6 Sol~\citep{gpt6}, and Kimi K3~\citep{team2026kimi} as the backbone, on the same $22$ datasets and four task domains as Table~\ref{tab:harness-combined}. Base: the backbone prompted directly; Ours: the backbone with \ours. Bold marks the better of Base and Ours for each backbone.}
\label{tab:harness-kimi}
\end{table}

%% file: tables/harness_design_qwen25.tex
\begin{table}[!htbp]
\centering
\setlength{\tabcolsep}{5pt}
\renewcommand{\arraystretch}{1.3}
\setlength{\aboverulesep}{0pt}
\setlength{\belowrulesep}{0pt}
\definecolor{avgrow}{HTML}{F2F2F2}
\newcommand{\gc}[1]{\textcolor{gray}{#1}}
\resizebox{\textwidth}{!}{%
\begin{tabular}{cl|ccc|c}
\toprule
 & & \multicolumn{3}{c|}{\gc{\textbf{Training-based}}} & \textbf{Training-free} \\
\textbf{Domain} & \textbf{Dataset}  & \gc{\textbf{AdaReasoner}} & \gc{\textbf{PixelReasoner}} & \gc{\textbf{DeepEyesV2}} & \textbf{OpenVLHarness} \\
\midrule
& CountQA        & \gc{36.98} & \gc{21.66} & \gc{21.73} & \textbf{51.37} \\
& FSC-147        & \gc{1.34} & \gc{9.75} & \gc{21.26} & \textbf{31.60} \\
& OdinW-13-All   & \gc{0.00} & \gc{\textbf{10.24}} & \gc{8.97} & 7.80 \\
& FSCD-147       & \gc{0.00} & \gc{0.04} & \gc{0.06} & \textbf{10.30} \\
& FSCD-LVIS      & \gc{0.00} & \gc{0.00} & \gc{0.05} & \textbf{14.20} \\
\multirow{-6}{*}{\rotatebox[origin=c]{90}{\shortstack[c]{Grounded\\Perception}}}
 & \cellcolor{avgrow}\textit{Average} & \cellcolor{avgrow}\gc{7.66} & \cellcolor{avgrow}\gc{8.34} & \cellcolor{avgrow}\gc{10.41} & \cellcolor{avgrow}\textbf{23.05} \\
\midrule
& MMSearch       & \gc{13.45} & \gc{12.87} & \gc{18.13} & \textbf{27.68} \\
& MMSearch-Plus  & \gc{0.96} & \gc{1.93} & \gc{1.61} & \textbf{3.86} \\
& RealX-Bench     & \gc{11.86} & \gc{16.49} & \gc{17.53} & \textbf{22.51} \\
& FVQA           & \gc{21.67} & \gc{26.33} & \gc{29.67} & \textbf{40.33} \\
& HLE-VL         & \gc{\textbf{10.00}} & \gc{6.06} & \gc{3.03} & 7.27 \\
& BC-VL          & \gc{18.05} & \gc{18.30} & \gc{\textbf{23.56}} & 17.79 \\
\multirow{-7}{*}{\rotatebox[origin=c]{90}{\shortstack[c]{Visual Search and\\Deep Research}}}
 & \cellcolor{avgrow}\textit{Average} & \cellcolor{avgrow}\gc{12.66} & \cellcolor{avgrow}\gc{13.66} & \cellcolor{avgrow}\gc{15.59} & \cellcolor{avgrow}\textbf{19.91} \\
\midrule
& RealWorldQA    & \gc{63.52} & \gc{\textbf{70.33}} & \gc{66.67} & 66.14 \\
& SimpleVQA      & \gc{35.90} & \gc{41.68} & \gc{42.12} & \textbf{46.17} \\
& HallusionBench & \gc{37.43} & \gc{34.69} & \gc{18.04} & \textbf{46.33} \\
& FREAK-VQA      & \gc{27.16} & \gc{35.42} & \gc{23.90} & \textbf{36.67} \\
& FREAK-MCQ      & \gc{30.08} & \gc{30.78} & \gc{26.96} & \textbf{33.30} \\
& V*             & \gc{65.97} & \gc{\textbf{83.01}} & \gc{81.15} & 74.87 \\
\multirow{-7}{*}{\rotatebox[origin=c]{90}{\shortstack[c]{General VQA and\\ Hallucination}}}
 & \cellcolor{avgrow}\textit{Average} & \cellcolor{avgrow}\gc{43.34} & \cellcolor{avgrow}\gc{49.32} & \cellcolor{avgrow}\gc{43.14} & \cellcolor{avgrow}\textbf{50.58} \\
\midrule
& MMSI-Bench     & \gc{20.10} & \gc{28.10} & \gc{9.30} & \textbf{29.80} \\
& ERQA           & \gc{26.75} & \gc{34.25} & \gc{32.50} & \textbf{37.00} \\
& EmbSpatial     & \gc{\textbf{72.01}} & \gc{71.84} & \gc{68.93} & 66.70 \\
& SparBench      & \gc{31.62} & \gc{30.33} & \gc{30.44} & \textbf{31.90} \\
& MindCube       & \gc{31.54} & \gc{\textbf{32.31}} & \gc{4.80} & \textbf{32.31} \\
\multirow{-6}{*}{\rotatebox[origin=c]{90}{\shortstack[c]{Spatial \\ Reasoning}}}
 & \cellcolor{avgrow}\textit{Average} & \cellcolor{avgrow}\gc{36.40} & \cellcolor{avgrow}\gc{39.37} & \cellcolor{avgrow}\gc{29.19} & \cellcolor{avgrow}\textbf{39.54} \\
\midrule
\rowcolor{avgrow}
\multicolumn{2}{l|}{\textbf{Overall average}} & \gc{25.29} & \gc{28.02} & \gc{25.02} & \textbf{33.45} \\
\bottomrule
\end{tabular}%
}
\caption{Training-based baselines on the Qwen2.5-VL-7B-Instruct backbone. AdaReasoner, PixelReasoner, and DeepEyesV2 fine-tune the backbone to adapt it to their tool-use harness; \ours is training-free and uses the same weights out of the box. Training-based columns are gray; shaded rows are per-domain averages. Bold marks the best result in each row.}
\label{tab:harness-qwen25}
\end{table}

%% file: figs/counting_object_counts.tex
\begin{figure}[!htbp]
    \centering
    \includegraphics[width=0.95\linewidth]{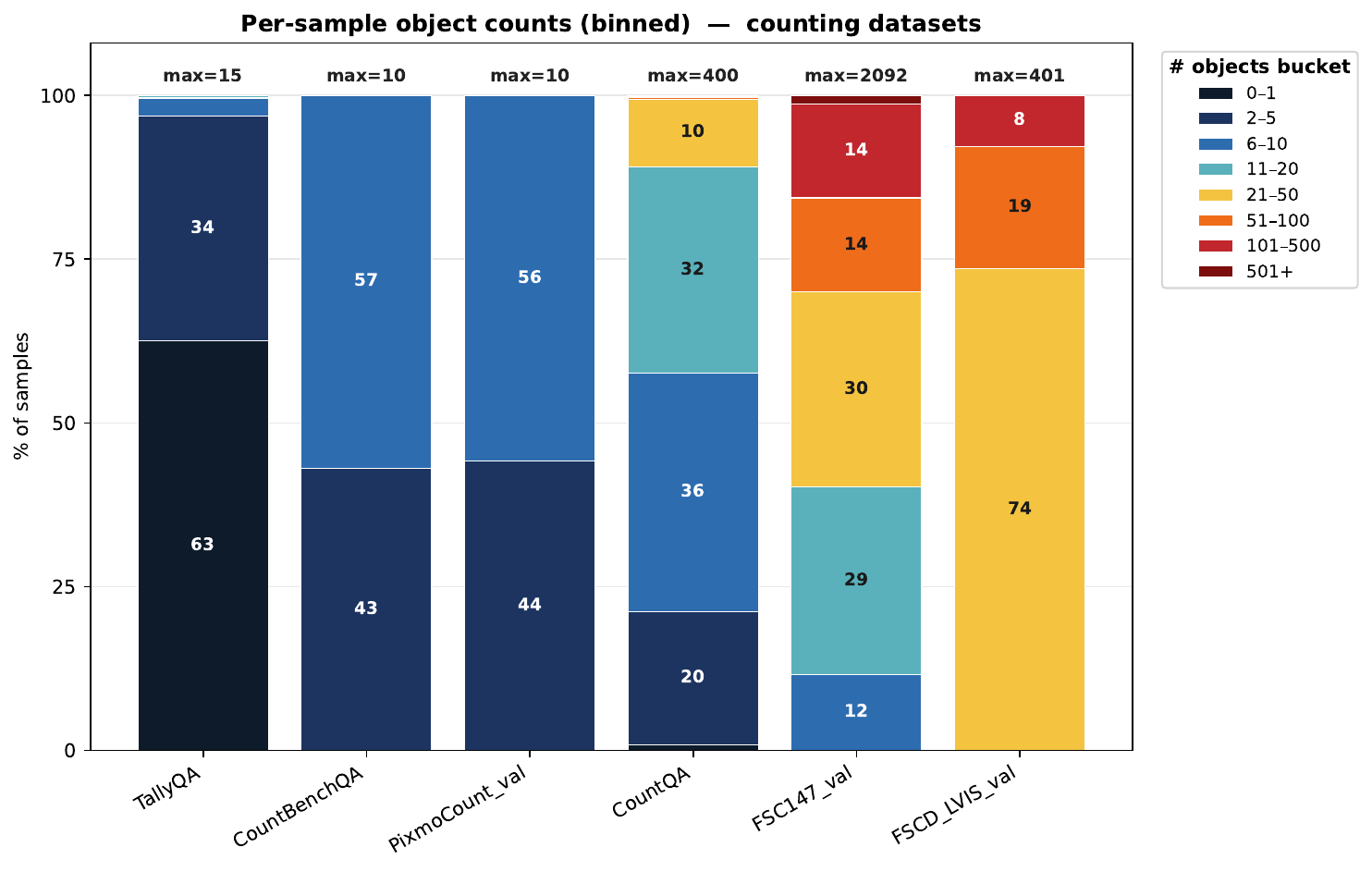}
    \caption{\textbf{Object-count distribution across counting benchmarks.}
    Each bar shows the percentage of samples falling into binned object-count
    buckets, with the per-dataset maximum count annotated above. Widely used
    benchmarks such as TallyQA, CountBenchQA, and PixmoCount are dominated by
    low counts: nearly all of their samples contain ten or fewer objects (and a
    majority of TallyQA fall in the $0$--$1$ bucket), capping at $10$--$15$. Only
    FSC-147 and FSCD-LVIS contain substantial mass in the high-count regime
    ($>50$ objects), reaching into the hundreds or thousands. This skew means
    accuracy on standard counting benchmarks largely reflects performance on
    small object counts and reveals little about high-count visual measurement.}
    \label{fig:appendix_count}
\end{figure}

%% file: figs/grounding_object_counts.tex
\begin{figure}[!htbp]
    \centering
    \includegraphics[width=0.95\linewidth]{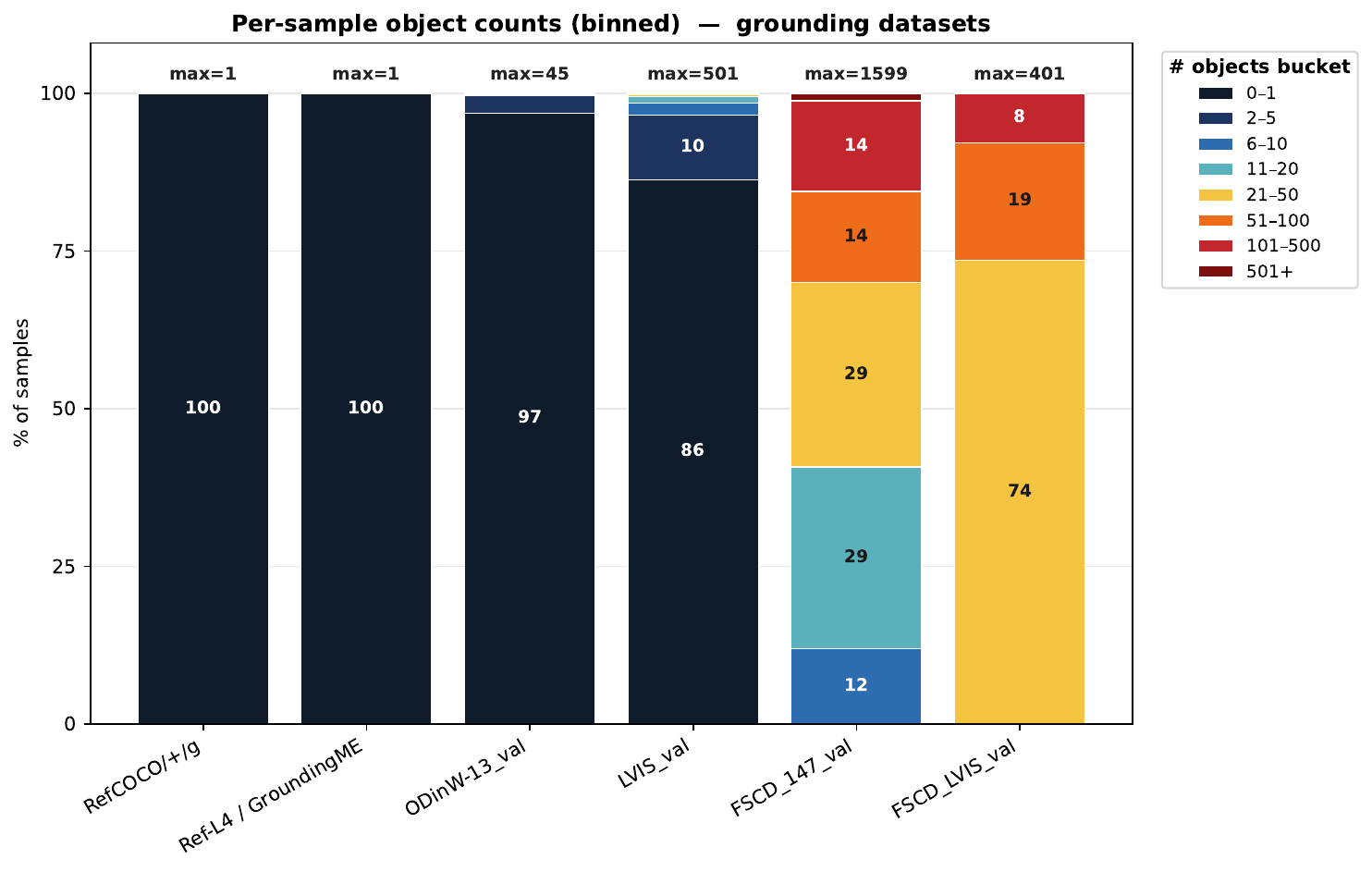}
    \caption{\textbf{Object-count distribution across grounding benchmarks.}
    Each bar shows the percentage of samples falling into binned object-count
    buckets, with the per-dataset maximum count annotated above. Referring-expression benchmarks such as RefCOCO/+/g and Ref-L4/GroundingME ground a
    single target per sample ($100\%$ in the $0$--$1$ bucket, $\text{max}=1$),
    and even detection-style sets like OdinW-13 ($97\%$) and LVIS ($86\%$) are
    overwhelmingly concentrated at one object per image. Only the FSCD variants
    exercise dense, many-object grounding. As with counting, this distribution
    implies that progress on common grounding benchmarks is measured almost
    entirely in the single- or few-object regime.}
    \label{fig:appendix_ground}
\end{figure}

%% file: tables/luna_effort_scaling.tex
\begin{table}[!htbp]
\centering
\small
\setlength{\tabcolsep}{4pt}
\begin{tabular}{@{}ll|cccc|c|r@{}}
\toprule
& & \multicolumn{4}{c|}{\textbf{Domain average}} & & \\
\textbf{Method} & \textbf{Effort} & Grounding & VS \& DR & General VQA & Spatial & \textbf{Average} & \textbf{\$ / 1k} \\
& & (5) & (6) & (6) & (5) & (22) & \\
\midrule
\multirow{5}{*}{Base} & low & 24.28 & 33.49 & 61.63 & 56.54 & 44.31 & 0.36 \\
 & medium & 25.36 & 35.18 & 63.38 & 57.69 & 45.76 & 0.58 \\
 & high & 26.54 & 35.27 & 64.37 & 59.27 & 46.68 & 0.85 \\
 & xhigh & 27.01 & 37.36 & 64.72 & 58.73 & 47.33 & 1.18 \\
 & max & 27.24 & 37.80 & 64.46 & 61.45 & 48.04 & 3.17 \\
\midrule
\multirow{5}{*}{Codex} & low & 20.29 & 26.13 & 54.66 & 45.21 & 36.92 & 1.07 \\
 & medium & 25.62 & 33.53 & 60.30 & 54.41 & 43.78 & 1.29 \\
 & high & 26.29 & 35.44 & 61.81 & 57.04 & 45.46 & 1.47 \\
 & xhigh & 27.11 & 36.65 & 63.02 & 58.39 & 46.61 & 1.71 \\
 & max & 29.98 & 40.30 & 63.54 & 59.13 & 48.58 & 2.52 \\
\midrule
\multirow{5}{*}{Visual Sketchpad} & low & 24.87 & 32.07 & 64.80 & 55.44 & 44.67 & 1.55 \\
 & medium & 28.31 & 32.37 & 64.26 & 56.16 & 45.55 & 1.89 \\
 & high & 29.33 & 34.61 & 65.14 & 57.14 & 46.86 & 2.27 \\
 & xhigh & 29.27 & 35.60 & 66.61 & 58.54 & 47.83 & 2.72 \\
 & max & 29.56 & 36.87 & 66.45 & 61.38 & 48.84 & 5.15 \\
\midrule
\multirow{5}{*}{\ours} & low & 42.05 & 41.11 & 61.51 & 54.45 & 49.92 & 0.64 \\
 & medium & 43.66 & 47.53 & 69.84 & 64.18 & 56.52 & 0.92 \\
 & high & 44.56 & 50.26 & 71.46 & 65.27 & 58.16 & 1.15 \\
 & xhigh & 45.25 & 50.59 & 71.53 & 65.91 & 58.57 & 1.45 \\
 & max & \textbf{45.63} & \textbf{52.71} & \textbf{72.59} & \textbf{68.53} & \textbf{60.12} & 2.99 \\
\bottomrule
\end{tabular}
\caption{GPT-6 Luna across five reasoning-effort settings for each method, evaluated on the full evaluation splits of all $22$ benchmarks of Table~\ref{tab:harness-combined} ($26{,}229$ questions per setting). Domain averages are unweighted means over the benchmarks of each domain (number of benchmarks in parentheses); per-benchmark scores are in Table~\ref{tab:luna_effort_perdataset}. \emph{\$ / 1k} is the model-token cost per $1{,}000$ questions at Standard-tier prices, including reasoning tokens and excluding tool execution and judging, averaged over the $22$ benchmarks. Bold marks the best score in each score column. Each cell is a single run.}
\label{tab:luna_effort_scaling}
\end{table}

%% file: tables/luna_effort_tokens.tex
\begin{table}[!htbp]
\centering
\small
\setlength{\tabcolsep}{3pt}
\begin{tabular}{@{}llrr|rrrr|rr@{}}
\toprule
& & & & \multicolumn{4}{c|}{\textbf{Input tokens}} & \multicolumn{2}{c}{\textbf{Output tokens}} \\
\textbf{Method} & \textbf{Effort} & \textbf{Score} & \textbf{\$/1k} & Uncached & Cache read & Cache write & \textbf{Total} & Reasoning & \textbf{Total} \\
\midrule
\multirow{5}{*}{Base} & low & 44.31 & 0.36 & 1{,}815 & 30 & 0 & 1{,}845 & 301 & 348 \\
 & medium & 45.76 & 0.58 & 1{,}793 & 52 & 0 & 1{,}845 & 750 & 804 \\
 & high & 46.68 & 0.85 & 1{,}795 & 50 & 0 & 1{,}845 & 1{,}283 & 1{,}345 \\
 & xhigh & 47.33 & 1.18 & 1{,}796 & 49 & 0 & 1{,}845 & 1{,}938 & 2{,}006 \\
 & max & 48.04 & 3.17 & 216 & 69 & 1{,}557 & 1{,}842 & 5{,}852 & 5{,}902 \\
\midrule
\multirow{5}{*}{Codex} & low & 36.92 & 1.07 & 3 & 114 & 7{,}926 & 8{,}043 & 104 & 146 \\
 & medium & 43.78 & 1.29 & 8 & 450 & 7{,}846 & 8{,}304 & 556 & 604 \\
 & high & 45.46 & 1.47 & 10 & 928 & 7{,}870 & 8{,}809 & 900 & 954 \\
 & xhigh & 46.61 & 1.71 & 14 & 1{,}625 & 7{,}976 & 9{,}614 & 1{,}321 & 1{,}382 \\
 & max & 48.58 & 2.52 & 30 & 3{,}501 & 8{,}272 & 11{,}803 & 2{,}835 & 2{,}906 \\
\midrule
\multirow{5}{*}{Visual Sketchpad} & low & 44.67 & 1.55 & 7 & 12{,}143 & 8{,}638 & 20{,}788 & 450 & 706 \\
 & medium & 45.55 & 1.89 & 10 & 12{,}996 & 8{,}689 & 21{,}695 & 1{,}076 & 1{,}339 \\
 & high & 46.86 & 2.27 & 19 & 13{,}847 & 8{,}876 & 22{,}742 & 1{,}749 & 2{,}030 \\
 & xhigh & 47.83 & 2.72 & 15 & 13{,}608 & 9{,}084 & 22{,}708 & 2{,}581 & 2{,}889 \\
 & max & 48.84 & 5.15 & 15 & 14{,}724 & 9{,}399 & 24{,}138 & 7{,}284 & 7{,}657 \\
\midrule
\multirow{5}{*}{\ours} & low & 49.92 & 0.64 & 27 & 10{,}804 & 3{,}635 & 14{,}466 & 56 & 159 \\
 & medium & 56.52 & 0.92 & 39 & 13{,}423 & 3{,}896 & 17{,}358 & 468 & 594 \\
 & high & 58.16 & 1.15 & 47 & 14{,}843 & 4{,}093 & 18{,}983 & 825 & 967 \\
 & xhigh & 58.57 & 1.45 & 43 & 16{,}643 & 4{,}321 & 21{,}007 & 1{,}316 & 1{,}485 \\
 & max & 60.12 & 2.99 & 57 & 19{,}685 & 4{,}808 & 24{,}550 & 4{,}118 & 4{,}374 \\
\bottomrule
\end{tabular}
\caption{Model tokens per question behind Table~\ref{tab:luna_effort_scaling}, summed over every model call made for a question and averaged over the $22$ benchmarks. Output totals include the reasoning tokens. For \ours, the calls of the Python Coding Agent are logged per setting rather than per question and are apportioned evenly across questions; they account for $0.4\%$ of its tokens and $2.5\%$ of its model-token cost at low effort, rising to $3.6\%$ and $14.7\%$ at max effort.}
\label{tab:luna_effort_tokens}
\end{table}

%% file: tables/luna_codex_execution.tex
\begin{table}[!htbp]
\centering
\small
\setlength{\tabcolsep}{4pt}
\resizebox{\linewidth}{!}{%
\begin{tabular}{@{}lrrr|rrrr|rrr@{}}
\toprule
& & \multicolumn{2}{c|}{\textbf{Executes code}} & \multicolumn{4}{c|}{\textbf{Executes code, by domain (\%)}} & & & \\
\textbf{Effort} & \textbf{Questions} & $n$ & \% & Grounding & VS \& DR & General VQA & Spatial & \textbf{Commands} & \textbf{Per exec.} & \textbf{Score} \\
\midrule
low & 26{,}229 & 43 & 0.16 & 0.00 & 1.34 & 0.00 & 0.00 & 126 & 2.9 & 36.92 \\
medium & 26{,}229 & 106 & 0.40 & 0.30 & 2.31 & 0.02 & 0.00 & 679 & 6.4 & 43.78 \\
high & 26{,}229 & 221 & 0.84 & 1.11 & 3.21 & 0.09 & 0.00 & 2{,}051 & 9.3 & 45.46 \\
xhigh & 26{,}229 & 457 & 1.74 & 2.88 & 4.77 & 0.16 & 0.01 & 4{,}953 & 10.8 & 46.61 \\
max & 26{,}229 & 1{,}619 & 6.17 & 12.33 & 9.61 & 0.89 & 0.00 & 18{,}753 & 11.6 & 48.58 \\
\bottomrule
\end{tabular}}
\caption{How often Codex executes code with GPT-6 Luna in the effort study. A question counts as executing code if its Codex session contains at least one command execution. \emph{Commands} is the total number of executed commands, and \emph{Per exec.} the mean number of commands per question that executes code. \emph{Score} is the $22$-benchmark average from Table~\ref{tab:luna_effort_scaling}.}
\label{tab:luna_codex_execution}
\end{table}

%% file: tables/luna_effort_perdataset.tex
\begin{table}[!p]
\centering
\scriptsize
\setlength{\tabcolsep}{3pt}
\begin{tabular}{@{}l|ccccc|ccccc@{}}
\toprule
& \multicolumn{5}{c|}{\textbf{Base}} & \multicolumn{5}{c}{\textbf{Codex}} \\
\textbf{Dataset} & low & med & high & xhigh & max & low & med & high & xhigh & max \\
\midrule
CountQA & 52.03 & 55.17 & 58.05 & 59.03 & 62.04 & 44.90 & 58.25 & 59.42 & 59.69 & 62.24 \\
FSC-147 & 28.62 & 28.23 & 29.32 & 29.78 & 29.94 & 22.63 & 28.62 & 29.55 & 31.88 & 38.34 \\
OdinW-13-All & 35.67 & 36.74 & 38.18 & 38.80 & 36.55 & 30.66 & 36.14 & 36.78 & 37.50 & 38.03 \\
FSCD-147 & 3.06 & 4.24 & 4.57 & 4.79 & 4.82 & 1.88 & 2.86 & 3.41 & 4.04 & 7.27 \\
FSCD-LVIS & 2.05 & 2.44 & 2.56 & 2.66 & 2.84 & 1.40 & 2.23 & 2.31 & 2.43 & 4.05 \\
\midrule
MMSearch & 49.12 & 49.12 & 51.46 & 52.05 & 53.80 & 40.35 & 49.12 & 52.63 & 52.63 & 58.48 \\
MMSearch-Plus & 7.72 & 10.93 & 9.32 & 10.93 & 10.93$^{\dagger}$ & 5.14 & 8.68 & 10.29 & 11.25 & 12.22 \\
RealX-Bench & 31.44 & 31.96 & 30.41 & 35.57 & 32.99 & 23.71 & 34.02 & 32.47 & 32.99 & 37.11 \\
FVQA & 49.78 & 52.89 & 52.72 & 53.50 & 54.67$^{\dagger}$ & 39.22 & 48.06 & 49.50 & 50.89 & 54.50 \\
HLE-VL & 23.03$^{\dagger}$ & 27.58$^{\dagger}$ & 27.58$^{\dagger}$ & 30.00$^{\dagger}$ & 33.03$^{\dagger}$ & 15.76 & 22.42 & 27.88 & 28.79 & 33.64 \\
BC-VL & 39.85 & 38.60 & 40.10 & 42.11 & 41.35$^{\dagger}$ & 32.58 & 38.85 & 39.85 & 43.36 & 45.86 \\
\midrule
RealWorldQA & 77.78 & 80.65 & 81.44 & 80.52 & 81.31 & 72.29 & 79.35 & 79.87 & 81.18 & 80.39 \\
SimpleVQA & 55.56 & 55.46 & 57.58 & 57.73 & 58.27$^{\dagger}$ & 48.49 & 54.47 & 55.16 & 56.10 & 58.67 \\
HallusionBench & 59.84 & 67.25 & 68.39 & 68.02 & 68.88 & 56.65 & 63.82 & 65.81 & \textbf{69.89} & 69.66 \\
FREAK-VQA & 40.55 & 41.68 & 42.05 & 43.55 & 41.68 & 41.05 & 39.17 & 39.80 & 39.17 & 39.92 \\
FREAK-MCQ & 48.09 & 49.40 & 50.91 & 48.99 & 49.70 & 38.83 & 44.37 & 46.48 & 47.48 & 47.28 \\
V* & 87.96 & 85.86 & 85.86 & 89.53 & 86.91 & 70.68 & 80.63 & 83.77 & 84.29 & 85.34 \\
\midrule
MMSI-Bench & 39.00 & 38.30 & 43.00 & 42.50 & 45.10 & 33.70 & 38.20 & 39.80 & 40.70 & 44.00 \\
ERQA & 59.50 & 59.75 & 61.25 & 58.25 & 59.75 & 41.00 & 54.25 & 58.00 & 57.00 & 56.75 \\
EmbSpatial & 77.58 & 78.98 & 79.64 & 79.70 & 80.05 & 68.30 & 75.36 & 76.95 & 78.43 & 78.98 \\
SparBench & 49.80 & 52.10 & 51.59 & 52.24 & 56.68 & 39.39 & 46.16 & 49.60 & 51.41 & 52.00 \\
MindCube & 56.83 & 59.33 & 60.87 & 60.96 & 65.67 & 43.65 & 58.08 & 60.87 & 64.42 & 63.94 \\
\midrule
\textit{Average} & 44.31 & 45.76 & 46.68 & 47.33 & 48.04 & 36.92 & 43.78 & 45.46 & 46.61 & 48.58 \\
\textit{\$ / 1k questions} & 0.36 & 0.58 & 0.85 & 1.18 & 3.17 & 1.07 & 1.29 & 1.47 & 1.71 & 2.52 \\
\bottomrule
\end{tabular}
\\[6pt]
\begin{tabular}{@{}l|ccccc|ccccc@{}}
\toprule
& \multicolumn{5}{c|}{\textbf{Visual Sketchpad}} & \multicolumn{5}{c}{\textbf{\ours}} \\
\textbf{Dataset} & low & med & high & xhigh & max & low & med & high & xhigh & max \\
\midrule
CountQA & 50.26 & 52.68 & 53.53 & 53.93$^{\dagger}$ & 56.54$^{\dagger}$ & 58.97 & 66.95 & 68.85 & 69.37 & \textbf{70.16} \\
FSC-147 & 29.16 & 30.25$^{\dagger}$ & 30.33$^{\dagger}$ & 31.88$^{\dagger}$ & 33.67$^{\dagger}$ & 40.59 & 41.52 & 41.37 & 41.45 & \textbf{42.07} \\
OdinW-13-All & 38.79$^{\dagger}$ & 43.84$^{\dagger}$ & 47.89$^{\dagger}$ & 48.50$^{\dagger}$ & 43.88$^{\dagger}$ & 54.45 & 53.56 & 55.29 & \textbf{57.29} & 55.67 \\
FSCD-147 & 2.89 & 6.11$^{\dagger}$ & 6.24 & 6.93$^{\dagger}$ & 7.51$^{\dagger}$ & 26.09 & 25.74 & 26.72 & 26.98 & \textbf{28.38} \\
FSCD-LVIS & 3.23 & 8.65 & 8.66$^{\dagger}$ & 5.09$^{\dagger}$ & 6.18$^{\dagger}$ & 30.17 & 30.55 & 30.58 & 31.14 & \textbf{31.88} \\
\midrule
MMSearch & 49.12 & 46.78 & 51.46 & 52.05 & 52.05 & 66.67 & 64.91 & 69.59 & 71.35 & \textbf{71.93} \\
MMSearch-Plus & 6.43 & 4.50 & 7.07 & 5.79$^{\dagger}$ & 8.68 & 16.08 & 26.37 & 26.37 & 28.62 & \textbf{31.83} \\
RealX-Bench & 30.41 & 35.05 & 31.44 & 34.54 & 35.57 & 31.96 & 45.36 & \textbf{50.00} & 47.94 & 49.48 \\
FVQA & 46.17 & 47.72 & 49.44 & 51.50 & 52.56$^{\dagger}$ & 61.61 & 68.78 & 69.33 & 70.56 & \textbf{71.72} \\
HLE-VL & 20.91$^{\dagger}$ & 20.30$^{\dagger}$ & 26.67$^{\dagger}$ & 26.36$^{\dagger}$ & 28.48$^{\dagger}$ & 21.21$^{\dagger}$ & 27.88$^{\dagger}$ & 33.64$^{\dagger}$ & 33.94$^{\dagger}$ & \textbf{36.67}$^{\dagger}$ \\
BC-VL & 39.35 & 39.85$^{\dagger}$ & 41.60 & 43.36$^{\dagger}$ & 43.86$^{\dagger}$ & 49.12 & 51.88 & 52.63 & 51.13 & \textbf{54.64} \\
\midrule
RealWorldQA & 80.92 & 80.26 & 80.39 & 81.57 & 82.61 & 73.86 & 80.78 & 81.96 & 82.22 & \textbf{83.01} \\
SimpleVQA & 55.11 & 55.06 & 54.91$^{\dagger}$ & 56.15 & 56.15$^{\dagger}$ & 63.95 & 72.20 & 72.99 & 73.28 & \textbf{74.77} \\
HallusionBench & 65.91 & 66.99 & 67.74 & 68.87 & 67.89 & 61.60 & 67.76 & 68.91 & 68.60 & 69.66 \\
FREAK-VQA & 46.06 & 46.06 & 46.43 & 49.69 & 48.69$^{\dagger}$ & 45.93 & 51.94 & 54.19 & 53.69 & \textbf{55.19} \\
FREAK-MCQ & 52.31 & 51.31 & 53.42 & 53.32 & 55.94 & 44.16 & 56.34 & 58.55 & 58.75 & \textbf{60.26} \\
V* & 88.48 & 85.86$^{\dagger}$ & 87.96 & 90.05 & 87.43 & 79.58 & 90.05 & 92.15 & \textbf{92.67} & \textbf{92.67} \\
\midrule
MMSI-Bench & 39.60$^{\dagger}$ & 40.00$^{\dagger}$ & 43.00 & 41.40 & 45.70$^{\dagger}$ & 36.60 & 47.70 & 48.80 & 52.60 & \textbf{57.10} \\
ERQA & 60.50 & 59.75 & 58.25 & 62.25 & \textbf{64.75} & 49.50 & 61.00 & 61.75 & 60.50 & 63.50 \\
EmbSpatial & 77.75 & 77.50 & 77.03$^{\dagger}$ & 78.43$^{\dagger}$ & 79.15 & 71.65 & 79.48 & 80.66 & 80.30 & \textbf{81.95} \\
SparBench & 43.31 & 46.55 & 50.21$^{\dagger}$ & 51.76 & 53.44$^{\dagger}$ & 48.85 & 58.31 & 59.47 & 60.56 & \textbf{62.13} \\
MindCube & 56.06 & 57.02 & 57.21$^{\dagger}$ & 58.85$^{\dagger}$ & 63.85$^{\dagger}$ & 65.67 & 74.42 & 75.67 & 75.58 & \textbf{77.98} \\
\midrule
\textit{Average} & 44.67 & 45.55 & 46.86 & 47.83 & 48.84 & 49.92 & 56.52 & 58.16 & 58.57 & 60.12 \\
\textit{\$ / 1k questions} & 1.55 & 1.89 & 2.27 & 2.72 & 5.15 & 0.64 & 0.92 & 1.15 & 1.45 & 2.99 \\
\bottomrule
\end{tabular}
\caption{Per-benchmark scores behind Table~\ref{tab:luna_effort_scaling}: GPT-6 Luna at five reasoning-effort settings for each method, on the full evaluation split of every benchmark. Scoring follows Table~\ref{tab:harness-combined}. Bold marks the best score in each benchmark row across all $20$ settings. $^\dagger$\,Requests for some questions failed on every retry ($134$ questions in $55$ cells; at most $13$ in a single cell). These count as incorrect. \emph{\$ / 1k questions} is the model-token cost at Standard-tier prices, averaged over the $22$ benchmarks. Each cell is a single run.}
\label{tab:luna_effort_perdataset}
\end{table}

%% file: tables/codex_qwen.tex
\begin{table}[!htbp]
\centering
\footnotesize
\setlength{\tabcolsep}{4pt}
\resizebox{\columnwidth}{!}{%
\begin{tabular}{@{}ll|ccc|ccc@{}}
\toprule
& & \multicolumn{3}{c|}{\textbf{Qwen3-VL-8B}} & \multicolumn{3}{c}{\textbf{Qwen3-VL-32B}} \\
\textbf{Domain} & \textbf{Dataset} & Base & Codex & \ours & Base & Codex & \ours \\
\midrule
\multirow{5}{*}{\shortstack[l]{Grounding}}
 & CountQA        & 32.00 & 29.97 & \textbf{48.56} & 44.70 & 39.73 & \textbf{64.14} \\
 & FSC-147        & 15.32 & 13.53 & \textbf{37.71} & 18.97 & 16.41 & \textbf{39.42} \\
 & OdinW-13-All   & 39.80 & 33.70 & \textbf{50.40} & 42.39 & 34.66 & \textbf{59.13} \\
 & FSCD-147       & 1.79 & 0.50 & \textbf{23.52} & 2.88 & 0.50 & \textbf{28.45} \\
 & FSCD-LVIS      & 1.47 & 0.78 & \textbf{27.37} & 1.79 & 2.24 & \textbf{31.63} \\
\midrule
\multirow{6}{*}{\shortstack[l]{VS \& DR}}
 & MMSearch       & 14.04 & 12.28 & \textbf{51.46} & 20.47 & 17.54 & \textbf{62.57} \\
 & MMSearch-Plus  & 1.29 & 2.57 & \textbf{14.79} & 2.57 & 7.72 & \textbf{17.68} \\
 & RealX-Bench     & 22.16 & 28.35 & \textbf{34.71} & 24.74 & 30.41 & \textbf{40.21} \\
 & FVQA           & 28.39 & 33.00 & \textbf{55.00} & 32.67 & 36.33 & \textbf{62.67} \\
 & HLE-VL         & 6.36 & 5.76 & \textbf{10.30} & 10.00 & 7.27 & \textbf{11.52} \\
 & BC-VL          & 21.05 & 13.53 & \textbf{32.33} & 28.82 & 17.29 & \textbf{33.08} \\
\midrule
\multirow{6}{*}{\shortstack[l]{General VQA}}
 & RealWorldQA    & 71.37 & 67.58 & \textbf{72.67} & 78.82 & 76.08 & \textbf{80.52} \\
 & SimpleVQA      & 54.33 & 46.27 & \textbf{70.00} & 59.00 & 50.12 & \textbf{70.33} \\
 & HallusionBench & 57.76 & 52.77 & \textbf{59.07} & 61.66 & 59.80 & \textbf{63.40} \\
 & FREAK-VQA      & \textbf{44.68} & 34.42 & 37.05 & 44.06 & 43.43 & \textbf{48.19} \\
 & FREAK-MCQ      & 36.42 & 34.51 & \textbf{36.82} & 44.06 & \textbf{50.10} & 45.77 \\
 & V*             & \textbf{83.24} & 69.11 & \textbf{83.24} & 85.34 & 79.06 & \textbf{88.48} \\
\midrule
\multirow{5}{*}{\shortstack[l]{Spatial}}
 & MMSI-Bench     & 31.60 & 27.40 & \textbf{33.00} & 35.00 & 31.30 & \textbf{38.00} \\
 & ERQA           & \textbf{44.00} & 41.25 & 43.00 & 44.75 & 41.75 & \textbf{48.00} \\
 & EmbSpatial     & 78.37 & 74.51 & \textbf{78.90} & \textbf{80.54} & 77.53 & 80.00 \\
 & SparBench      & 39.55 & 39.82 & \textbf{43.73} & 46.98 & \textbf{46.99} & 46.76 \\
 & MindCube       & 28.36 & 29.71 & \textbf{55.48} & 32.40 & 35.87 & \textbf{60.19} \\
\midrule
\multicolumn{2}{@{}l|}{\textit{Average}} & 34.24 & 31.42 & \textbf{45.41} & 38.30 & 36.46 & \textbf{50.92} \\
\bottomrule
\end{tabular}%
}
\caption{Codex with Qwen3-VL backbones on the 22 benchmarks of Table~\ref{tab:harness-combined}. Base and \ours are copied from that table. Codex runs the Codex CLI agent in code mode at medium reasoning effort with web search disabled. Bold marks the best result within each model group. Averages are taken over all 22 rows. Each Codex cell is a single run.}
\label{tab:codex_qwen}
\end{table}

%% file: tables/memory_artifacts.tex
\begin{table}[!htbp]
\centering
\footnotesize
\setlength{\tabcolsep}{4pt}
\renewcommand{\arraystretch}{1.15}
\newcommand{\idtt}[1]{\texttt{#1}}
\resizebox{\textwidth}{!}{%
\begin{tabular}{@{}>{\raggedright\arraybackslash}p{0.215\textwidth}>{\raggedright\arraybackslash}p{0.135\textwidth}>{\raggedright\arraybackslash}p{0.40\textwidth}>{\raggedright\arraybackslash}p{0.20\textwidth}@{}}
\toprule
\textbf{Artifact} & \textbf{Source tool} & \textbf{Stored representation in memory} & \textbf{Reusable handle} \\
\midrule
\multicolumn{4}{@{}l}{\textit{Image artifacts (returned to the model as a native image attachment + an ID)}} \\
Input image & (question) & Registered verbatim & \idtt{input\_image\_\textit{n}} \\
Cropped region & Zoom-in & Cropped image, deduplicated by content & \idtt{tool\_generated\_\allowbreak image\_\textit{n}} \\
Grounded / mask overlay & Visual Grounding & Annotated image with numbered regions & \idtt{tool\_generated\_\allowbreak image\_\textit{n}} \\
Colorized depth map & Depth Estimation & Rendered depth visualization & \idtt{tool\_generated\_\allowbreak image\_\textit{n}} \\
Camera-trajectory plot & Camera Trajectory & Bird's-eye-view render of the estimated camera path & \idtt{tool\_generated\_\allowbreak image\_\textit{n}} \\
Retrieved thumbnails & Image / Text Search & Image bytes fetched lazily on first reuse & \idtt{tool\_generated\_\allowbreak image\_\textit{n}} \\
\midrule
\multicolumn{4}{@{}l}{\textit{Structured data bindings (kept out of the transcript; exposed as typed, named variables)}} \\
Boxes, scores, labels & Visual Grounding & Pixel/normalized boxes, confidences, labels, image size & named binding (e.g.\ \idtt{grounding\_\textit{query}}) \\
Segmentation masks & Visual Grounding & Per-region masks + scores + counts & named binding \\
Metric depth array & Depth Estimation & Per-pixel depth (float array) + shape + colormap & named binding \\
Camera poses & Camera Trajectory & Per-image intrinsic and extrinsic matrices & named binding \\
OCR results & OCR & Recognized text, per-token boxes, confidences & named binding \\
\midrule
\multicolumn{4}{@{}l}{\textit{Web references (URLs replaced by short tokens in retrieved text)}} \\
Search-result pages & Text / Image Search & URL + domain + provenance & \idtt{webpage\_url\_\textit{n}} \\
\midrule
\multicolumn{4}{@{}l}{\textit{Code outputs (returned as text; emitted images are registered as image artifacts)}} \\
Code outputs & Python Coding Agent & Printed output + any emitted images & text + image handle(s) \\
\bottomrule
\end{tabular}%
}
\caption{Tool-generated artifacts retained in persistent multimodal memory, grouped by how they are represented and re-addressed. Heavy payloads (image bytes, numeric arrays) are stored once and replaced in the transcript by a compact handle; the orchestrator reuses an artifact on a later turn by citing its handle, which the harness resolves back to the underlying data.}
\label{tab:appendix_memory_artifacts}
\end{table}

%% file: text/appendix_optim.tex
\section{Failure Analysis and Harness Optimization Details}
\label{sec:appendix_optim_all}

This appendix describes how failure diagnosis guides specialization of \ours on a user's data. We detail the critique protocol, the mapping from diagnoses to remedy channels, and the two workflows that modify harness text or add task-specific tools without updating model weights. We then provide examples of the resulting edits and tools.

\subsection{Critique Protocol}
\label{sec:appendix_critique}

\paragraph{Failure selection and coverage.} We use Kimi K3 to critique failures of the unoptimized \ours with Qwen3-VL-8B, obtaining $9{,}893$ completed critiques across 22 benchmarks. Failures follow each benchmark's judge verdict. For OdinW-13-All, FSCD-147, and FSCD-LVIS, which report box AP without per-image verdicts, we select images whose predicted boxes have F1 below $0.5$ at IoU $0.5$. Table~\ref{tab:appendix_critique_per_dataset} reports the completed coverage.

\paragraph{Critic inputs and outputs.} Each critique includes the question, reference and predicted answers, full trajectory, input and tool-generated images, and toolbox from the run's configuration. Tool descriptions and argument schemas are supplied with called tools marked, so the critic can assess omissions against the capabilities actually available. For each call, it labels information usefulness (\emph{yes}, \emph{partial}, or \emph{no}) and correctness (\emph{correct}, \emph{wrong}, or \emph{unverifiable}). It also returns one primary failure cause, a proposed fix type and sketch, supporting transcript excerpts, and confidence, all in a structured JSON record. Each critique uses approximately $3$k prompt tokens and $0.7$--$2$k completion tokens.

\paragraph{Taxonomy.} The critic separately identifies where the trajectory failed and what change could address it (Table~\ref{tab:appendix_taxonomy}). A \emph{build-tool} fix requires a new callable, including wrappers that verify, filter, or translate outputs and compositions of existing capabilities. A \emph{text-only} fix changes instructions for using existing tools. These labels are subsequently combined into remedy channels; they do not directly determine which workflow to use.

\begin{table}[!htbp]
\centering
\small
\setlength{\tabcolsep}{4pt}
\begin{tabular}{@{}p{0.27\textwidth}p{0.69\textwidth}@{}}
\toprule
\textbf{Label} & \textbf{Meaning} \\
\midrule
\multicolumn{2}{@{}l}{\textit{Root cause (exactly one)}} \\
Tool not called & The needed evidence was never gathered, although a tool in the agent's toolbox could have supplied it. \\
Tool output wrong & A tool was called but returned wrong or missing information (missed or hallucinated boxes, a bad crop, a wrong transcription). \\
Tool output unusable & A tool returned correct information in a form the model could not use (raw arrays, uninterpreted coordinates, an overload of detections). \\
Tool output ignored & A tool returned correct, plainly usable information that the model did not incorporate or contradicted. \\
Reasoning error & The evidence was adequate and used; the error is in reasoning, arithmetic, comparison, or answer format. \\
Label or question issue & The question is ambiguous or the reference answer looks wrong; the agent's answer is defensible. \\
\midrule
\multicolumn{2}{@{}l}{\textit{Fix type}} \\
Build tool & A new callable is needed: a wrapper around a base tool, a combination of tools, or an output-translation layer. \\
Text only & Only a tool description or the agent prompt changes. \\
None & Nothing above would plausibly help. \\
\bottomrule
\end{tabular}
\caption{Critique taxonomy. Each failure receives one primary cause and one proposed fix type. Unusable and ignored outputs remain separate during critique and are combined for reporting in Fig.~\ref{fig:failure_analysis} and Table~\ref{tab:appendix_critique_per_dataset}.}
\label{tab:appendix_taxonomy}
\end{table}

\paragraph{Human agreement.} Human annotators independently labeled 100 randomly sampled critiqued failures, with an average agreement of $92\%$ with the critic. The abridged prompt below specifies the evidence requirements and fixed labels.

\begin{tcolorbox}[breakable,title=Critic prompt (abridged)]
\small
You are a strict, evidence-driven critic of a vision-language tool-using agent. You are shown ONE trajectory where the agent answered a visual question wrongly, together with the correct answer and the catalog of tools the agent could call. [\ldots] The single question your critique must answer cleanly is: does fixing this family of failures require NEW TOOL CODE the agent can call, or only a TEXT change? [\ldots]

\textbf{How to judge each tool call.} For every tool the agent called, decide \texttt{info\_useful} (did the returned information bear on answering this question?) and \texttt{info\_correct} (was that information actually right, judged against the image content, the question, and the correct answer?). If a grounding tool reports 3 boxes but the correct answer implies 5 objects, that output is wrong. If you cannot tell, say unverifiable; do not guess.

\textbf{Root cause.} Choose exactly one of \texttt{tool\_not\_called}, \texttt{tool\_output\_wrong}, \texttt{tool\_output\_unusable\_form}, \texttt{tool\_output\_ignored}, \texttt{reasoning\_error}, \texttt{label\_or\_question\_issue}. \texttt{tool\_not\_called} must name a tool from the agent's toolbox that would have supplied the missing evidence; if no listed tool could have, describe the missing capability in the fix sketch instead. [\ldots]

\textbf{Rules.} Ground every claim in the transcript: quote the tool output or the agent's text. Never assume a tool was right just because the agent believed it. Do not propose memorizing answers or matching question wording; a fix sketch must describe a general capability or change. Set confidence low when the evidence is thin. Return ONLY a JSON object.
\end{tcolorbox}

\subsection{Remedy Channels and Per-Dataset Results}
\label{sec:appendix_channels}

Each failed trajectory receives two labels from the critic: a \emph{root cause}, describing where it failed, and a \emph{fix type}, indicating whether the proposed remedy requires new tool code, a text change, or neither. We determine its remedy channel jointly from this pair, using the tool-call assessments where needed. For example, an omitted call paired with a build-tool fix maps to tool composition (C), whereas the same cause paired with a text-only fix maps to harness text (Y).

The channels distinguish \emph{adding task-specific tools} (A+B+C) from \emph{refining harness text} (Y). New tools can make evidence interpretation explicit, refine outputs, or compose existing capabilities. Harness text comprises the system prompt, tool descriptions, and evidence-rendering templates; editing it guides when existing tools are called and how their results are used. We first assign cases with fix type \emph{none} or a question/reference-answer issue to Z, then map the remaining cause--fix pairs as follows:
\begin{itemize}[topsep=0.3em,itemsep=0.2em,leftmargin=1.4em]
    \item \textbf{A: evidence interpretation tools.} An ignored-output or reasoning-error cause paired with a build-tool fix. We also include text-only fixes for ignored outputs, or for reasoning errors when at least one call returned correct, at least partially useful evidence. These tools turn available evidence into explicit task-level conclusions that the model can more readily use.
    \item \textbf{B: output verification and refinement tools.} A wrong- or unusable-output cause paired with a build-tool fix, motivating systematic verification, filtering, or translation.
    \item \textbf{C: tool composition.} An omitted-call cause paired with a build-tool fix, motivating a task-specific callable that combines existing capabilities into the required operation.
    \item \textbf{Y: harness text for tool-use guidance.} The remaining causes paired with a text-only fix.
    \item \textbf{Z: no proposed remedy.} Cases excluded above because no fix is proposed or the question/reference answer is problematic.
\end{itemize}
The motivation for A is that correct, useful evidence may still require interpretation before it supports an answer. A reasoning error despite such evidence can therefore indicate difficulty interpreting or incorporating it, as can an ignored output. We include these cases even when the critic proposes a text change, to test whether a tool that makes the relevant conclusion explicit can address the difficulty. This motivates a candidate remedy rather than establishing that interpretation caused every such failure.

Table~\ref{tab:appendix_critique_per_dataset} reports the channel shares among critiqued failures. The final column illustrates why an omitted call alone does not identify the remedy: the critic proposes new tool code for $70\%$ of such failures on SparBench, but only $11\%$ on ERQA. The four datasets in Sec.~\ref{sec:harness_optim} span contrasting profiles: Y accounts for $71\%$ of ERQA and $56\%$ of RealWorldQA failures, while A+B+C account for $75\%$ of SparBench and $62\%$ of FREAK-VQA failures. These shares guide the choice of workflow; they do not predict accuracy gains. Critiques of failures alone cannot establish whether an edit harms previously correct answers or whether the orchestrator will use a generated tool.

\begin{table}[!htbp]
\centering
\small
\setlength{\tabcolsep}{3pt}
\resizebox{\textwidth}{!}{%
\begin{tabular}{llr|ccccc|ccccc|cc|c}
\toprule
& & & \multicolumn{5}{c|}{\textbf{Root cause (\%)}} & \multicolumn{5}{c|}{\textbf{Remedy channel (\%)}} & & & \\
\textbf{Domain} & \textbf{Dataset} & \textbf{\#Fail.} & \textbf{NC} & \textbf{Wrong} & \textbf{Unus./Ign.} & \textbf{Reas.} & \textbf{Label} & \textbf{A} & \textbf{B} & \textbf{C} & \textbf{Y} & \textbf{Z} & \textbf{Tool} & \textbf{Text} & \textbf{NC$\to$build} \\
\midrule
\multirow{5}{*}{Grounding} & CountQA & 751 & 32 & 53 & 4 & 10 & 1 & 12 & 54 & 2 & 31 & 1 & \textbf{68} & 31 & 8 \\
 & FSC-147 & 803 & 9 & 89 & 1 & 2 & 0 & 2 & 89 & 2 & 7 & 0 & \textbf{93} & 7 & 27 \\
 & OdinW-13-All & 1{,}016 & 4 & 56 & 2 & 15 & 23 & 12 & 56 & 0 & 9 & 23 & \textbf{68} & 9 & 7 \\
 & FSCD-147 & 90 & 0 & 79 & 6 & 0 & 16 & 3 & 81 & 0 & 0 & 16 & \textbf{84} & 0 & 0 \\
 & FSCD-LVIS & 370 & 0 & 98 & 1 & 0 & 0 & 1 & 98 & 0 & 0 & 0 & \textbf{100} & 0 & 0 \\
\midrule
\multirow{6}{*}{VS \& DR} & MMSearch & 80 & 49 & 15 & 0 & 21 & 15 & 20 & 14 & 0 & 51 & 15 & 34 & \textbf{51} & 0 \\
 & MMSearch-Plus & 280 & 62 & 13 & 4 & 20 & 1 & 19 & 12 & 0 & 67 & 1 & 31 & \textbf{67} & 1 \\
 & RealX-Bench & 125 & 44 & 29 & 4 & 14 & 9 & 10 & 29 & 6 & 47 & 9 & 44 & \textbf{47} & 13 \\
 & FVQA & 753 & 38 & 21 & 5 & 21 & 15 & 21 & 18 & 0 & 46 & 15 & 39 & \textbf{46} & 1 \\
 & HLE-VL & 302 & 42 & 14 & 1 & 41 & 3 & 16 & 13 & 8 & 57 & 6 & 37 & \textbf{57} & 20 \\
 & BC-VL & 265 & 56 & 5 & 3 & 29 & 7 & 25 & 3 & 0 & 66 & 7 & 27 & \textbf{66} & 0 \\
\midrule
\multirow{6}{*}{General VQA} & RealWorldQA & 209 & 50 & 8 & 11 & 32 & 0 & 26 & 13 & 4 & 56 & 0 & 44 & \textbf{56} & 8 \\
 & SimpleVQA & 770 & 14 & 18 & 11 & 42 & 14 & 44 & 18 & 1 & 23 & 15 & \textbf{62} & 23 & 9 \\
 & HallusionBench & 258 & 38 & 9 & 4 & 41 & 8 & 15 & 10 & 8 & 57 & 9 & 33 & \textbf{57} & 21 \\
 & FREAK-VQA & 503 & 32 & 9 & 9 & 49 & 1 & 52 & 10 & 1 & 35 & 2 & \textbf{62} & 35 & 2 \\
 & FREAK-MCQ & 628 & 47 & 8 & 5 & 36 & 4 & 36 & 8 & 1 & 50 & 4 & 46 & \textbf{50} & 2 \\
 & V* & 32 & 38 & 22 & 6 & 34 & 0 & 34 & 25 & 0 & 41 & 0 & \textbf{59} & 41 & 0 \\
\midrule
\multirow{5}{*}{Spatial} & MMSI-Bench & 660 & 41 & 9 & 6 & 44 & 0 & 36 & 10 & 9 & 44 & 0 & \textbf{55} & 44 & 23 \\
 & ERQA & 228 & 52 & 5 & 5 & 38 & 0 & 16 & 7 & 6 & 71 & 0 & 29 & \textbf{71} & 11 \\
 & EmbSpatial & 765 & 75 & 3 & 4 & 18 & 1 & 6 & 6 & 11 & 76 & 1 & 23 & \textbf{76} & 15 \\
 & SparBench & 542 & 51 & 8 & 10 & 30 & 1 & 24 & 15 & 36 & 24 & 1 & \textbf{75} & 24 & 70 \\
 & MindCube & 463 & 25 & 3 & 2 & 70 & 0 & 40 & 3 & 3 & 54 & 0 & 46 & \textbf{54} & 10 \\
\bottomrule
\end{tabular}%
}
\caption{Per-dataset failure analysis of \ours on Qwen3-VL-8B. \#Fail.: critiqued failures. Root causes: NC = tool not called, Wrong = tool output wrong, Unus./Ign. = tool output unusable or ignored, Reas. = reasoning error, Label = label or question issue. Remedies: A = evidence interpretation, B = output verification/refinement, C = tool composition, Y = harness text, Z = no proposed remedy. Tool = A+B+C and Text = Y; the larger share is bold. Percentages are rounded independently. NC$\to$build: share of tool-not-called failures the critic marks as needing a new tool. On the box-AP grounding benchmarks, a failure is an image with F1 below $0.5$ at IoU $0.5$.}
\label{tab:appendix_critique_per_dataset}
\end{table}

\subsection{Optimization Algorithms}
\label{sec:appendix_optim}

Both workflows use a coding agent (Codex with GPT-5.5) to inspect trajectories and modify a scratch copy of the harness. Candidates are evaluated by running the Qwen3-VL-8B orchestrator and applying the benchmark judge. Harness search changes instructions and evidence presentation; tool generation adds callable operations built from existing capabilities. Both keep model weights fixed.

\paragraph{Harness search.} Algorithm~\ref{alg:harness_search} maintains a pool of harness states represented as file overlays on the base repository. The editable components are the system prompt and tool-use policy, tool descriptions, and model-facing output templates. For each parent state, an analysis agent inspects failed trajectories and may test hypotheses on up to 20 tasks with five trials each. A proposer then produces $M$ distinct patches without instance-specific rules. Candidates include individual patches and compatible combinations of up to $K$ patches. Each candidate is evaluated with $S$ samples per instance using paired sampling seeds. The top $N$ parents and children by mean search accuracy survive, with older states winning exact ties. A campaign runbook records proposals and scores for subsequent rounds.

We use $N{=}4$, $M{=}4$, $K{=}4$, $S{=}3$, and three rounds. ERQA uses 100 samples and is evaluated on a disjoint 300-sample validation split for assessing leading candidates. Similarly, RealWorldQA uses 100 samples and is reported on the disjoint 665 validation split. The first-round analysis can optionally use an earlier critique report, provided its claims are checked against the current trajectories. Thus, failure diagnosis can guide the choice of workflow without being a required input to the search algorithm.

\begin{algorithm}[h]
\caption{Harness search}
\label{alg:harness_search}
\begin{algorithmic}[1]
\Require base harness $h_0$, search split $\mathcal{D}$, optional validation split $\mathcal{V}$, pool size $N$, proposals per parent $M$, max combination size $K$, samples per instance $S$, rounds $R$
\State $\mathcal{P} \gets \{h_0\}$; \ $\mathrm{score}(h_0) \gets \mathrm{Eval}(h_0, \mathcal{D}, S)$
\For{$r = 1, \dots, R$}
    \State $\mathcal{C} \gets \emptyset$
    \For{each parent $p \in \mathcal{P}$}
        \State $a \gets \textsc{Analyze}(p, \text{failures of } p \text{ on } \mathcal{D}, \text{runbook})$ \Comment{coding agent; may use the playground}
        \State $\{\delta_1,\dots,\delta_M\} \gets \textsc{Propose}(p, a)$ \Comment{text-only patches to the harness}
        \For{each non-empty $I \subseteq \{1,\dots,M\}$ with $|I| \le K$}
            \State $c \gets p \oplus \{\delta_i\}_{i \in I}$; \ skip if the patches conflict
            \State $\mathrm{score}(c) \gets \mathrm{Eval}(c, \mathcal{D}, S)$; \ $\mathcal{C} \gets \mathcal{C} \cup \{c\}$
        \EndFor
    \EndFor
    \State $\mathcal{P} \gets \mathrm{Top}_N(\mathcal{P} \cup \mathcal{C}, \mathrm{score})$ \Comment{older state wins exact ties}
    \State evaluate new members of $\mathcal{P}$ on $\mathcal{V}$ if supplied; \ \textsc{Bookkeep}(runbook, $\mathcal{C}$)
\EndFor
\State \Return $\mathcal{P}$
\end{algorithmic}
\end{algorithm}

\paragraph{Tool generation.} Algorithm~\ref{alg:tool_generation} groups source failures into at most $D$ candidate tool directions, each describing a reusable operation and the failures it should address. Supporting examples are divided equally into tuning and held-out subsets. A tool-writing agent implements each candidate by composing existing capabilities; it may invoke tools, execute code, and query the base model. For up to $T$ rounds, the candidate is registered in the scratch harness and evaluated on the tuning subset. A separate diagnosis agent classifies its use as not called, execution failed, evidence inadequate, evidence misinterpreted, or used correctly, and the tool-writing agent revises the implementation accordingly. Dataset- or instance-specific answer rules and lookup tables are prohibited.

The final gate requires complete scoring, at least two recovered failures and positive net lift across all supporting examples, and at least one recovered failure with positive net lift on the held-out half, whose trajectories were not shown to the tool-writing agent. Because these examples are selected from baseline failures, this gate measures recovery rather than regressions on initially correct answers. Accepted tools are collected into a bank and evaluated together with the original harness on the target dataset. We use $D{\le}5$, $T{=}5$, and 50 examples per direction on SparBench and 30 on FREAK-VQA. For both SparBench and FREAK-VQA, we generate tools on data disjoint from the standard evaluation splits and evaluate the resulting harness on the same splits used in our main experiments.

\begin{algorithm}[h]
\caption{Tool generation}
\label{alg:tool_generation}
\begin{algorithmic}[1]
\Require harness $h$ with primitive tools, source failures $\mathcal{F}$, max directions $D$, forge rounds $T$
\State $\{(d_j, \mathcal{S}_j)\}_{j \le D} \gets \textsc{Analyze}(\mathcal{F})$ \Comment{tool directions with supporting samples}
\State $\mathcal{B} \gets \emptyset$
\For{each direction $(d_j, \mathcal{S}_j)$}
    \State split $\mathcal{S}_j$ into tuning half $\mathcal{S}_j^{\mathrm{tune}}$ and held-out half $\mathcal{S}_j^{\mathrm{held}}$
    \State $\tau \gets \textsc{Forge}(d_j, h)$ \Comment{coding agent writes a tool composing primitive tools}
    \For{$t = 1, \dots, T$}
        \State $\rho \gets \mathrm{Run}(h \cup \{\tau\}, \mathcal{S}_j^{\mathrm{tune}})$; \ $\kappa \gets \textsc{Diagnose}(\rho)$ \Comment{how the agent used $\tau$}
        \State \textbf{if} $\kappa$ requests an early stop \textbf{then break}
        \State $\tau \gets \textsc{Forge}(d_j, h, \tau, \rho, \kappa)$
    \EndFor
    \State $(\mathrm{fix}, \mathrm{broke}) \gets \mathrm{Lift}(h \cup \{\tau\}, \mathcal{S}_j)$; \ $(\mathrm{fix}', \mathrm{broke}') \gets \mathrm{Lift}(h \cup \{\tau\}, \mathcal{S}_j^{\mathrm{held}})$
    \If{all examples scored \textbf{and} $\mathrm{fix} \ge 2$ \textbf{and} $\mathrm{fix} > \mathrm{broke}$ \textbf{and} $\mathrm{fix}' \ge 1$ \textbf{and} $\mathrm{fix}' > \mathrm{broke}'$}
        \State $\mathcal{B} \gets \mathcal{B} \cup \{\tau\}$
    \EndIf
\EndFor
\State \Return $h \cup \mathcal{B}$
\end{algorithmic}
\end{algorithm}

\subsection{Optimized Harnesses and Generated Tools}
\label{sec:appendix_optim_results}

\paragraph{Harness text edits.} The RealWorldQA example below combines edits from three search rounds: five system-prompt additions and one addition to the Visual Grounding output template. They guide the orchestrator to inspect small road-scene details, verify counts, and match its visual conclusion to the answer options. The output-template edit illustrates how evidence rendering supports adaptation: it prompts verification of grounded candidates before they are used as counts or relation evidence.

\begin{tcolorbox}[breakable,title=Edits in a top-scoring RealWorldQA harness (additions only)]
\small
\textbf{System prompt, when to use visual tools.} For road-scene details such as traffic lights, traffic signs, arrows, lane markings, and whether a driver must stop, use visual tools before answering unless the relevant mark or signal is unmistakably large and clear. Before answering that such a detail is absent, not visible, or has a particular color/direction, ground the likely object or region and inspect the original pixels with zoom if the region is small or distant.

\textbf{System prompt, using tool results.} For count questions with numeric options, compare the filtered candidate count with the option values. If the count is outside the listed values or directly adjacent to a listed value, do one missed-object check using a broader/simple phrase or one targeted zoom of the most uncertain region, then choose the listed option best supported by all evidence.

For counts of printed letters, digits, or symbols, use OCR or a text-focused zoom when available before deciding the count. Unless the question explicitly says ``standalone'', count occurrences inside longer strings as well as isolated characters.

\textbf{System prompt, final answer.} For multiple-choice questions, before writing \texttt{<answer>}, explicitly bind your visual conclusion to the option table. The option letter and option text you output must match each other and match the conclusion in your thinking; if your conclusion is not one of the options, re-check the evidence instead of choosing a nearby option.

If the listed options do not include a none/unknown/cannot-determine choice, never use an unlisted absence answer such as ``none of the above'', ``not visible'', or ``cannot tell''. If your draft conclusion is that the needed cue is absent, do one targeted visual check when a tool can resolve it, then answer with the listed option best supported by the evidence.

\textbf{Visual Grounding output template} (appended after the list of boxes). Treat these as candidate regions, not a verified final count. Before answering count, presence, or relation questions, compare each candidate with the original image and remove duplicates, mirror/reflection copies, partial regions, off-target detections, and objects that fail the question's location or context constraint.
\end{tcolorbox}

\paragraph{Task-specific tools.} Table~\ref{tab:appendix_generated_tools} lists the five accepted tools: two from SparBench failures and three from FREAK-MCQ failures. A fourth FREAK direction, localized text and symbol reading, encountered an execution error before completing the gate. The SparBench tools package depth, camera intrinsics, and cross-view geometry into metric 3D operations, addressing the composition needs of channel C. In the diagnosed SparBench failures, the critic judged $87\%$ of Python Coding Agent calls incorrect, motivating reusable implementations of these computations. The FREAK tools make visibility, layout, and counting decisions explicit, addressing the evidence-interpretation needs of channel A. The visibility verifier recovered the most gate examples among the three FREAK tools; the table reports generation-stage recovery, not a per-tool benchmark ablation.

\begin{table}[!htbp]
\centering
\small
\setlength{\tabcolsep}{4pt}
\newcommand{\tn}[1]{{\footnotesize\ttfamily #1}}
\begin{tabular}{@{}>{\raggedright\arraybackslash}p{0.25\textwidth}>{\raggedright\arraybackslash}p{0.53\textwidth}cc@{}}
\toprule
\textbf{Tool} & \textbf{What it does} & \textbf{Gate} & \textbf{Held-out} \\
\midrule
\multicolumn{4}{@{}l}{\textit{Generated from SparBench failures}} \\
\tn{image\_points\_to\_\allowbreak metric\_3d} & Lifts localized image points or box centers into camera coordinates using depth and inverse intrinsics, and returns pairwise distances, distances from the observer, and nearest-point rankings in meters, with a warning when a point lies on a depth edge. & 30/50 & 13/25 \\
\addlinespace
\tn{camera\_center\_\allowbreak image\_projector} & Estimates camera poses across ordered views and projects another view's camera center into a reference image, returning its image coordinates and depth together with an annotated image. & 40/50 & 18/25 \\
\midrule
\multicolumn{4}{@{}l}{\textit{Generated from FREAK-MCQ failures}} \\
\tn{target\_visibility\_\allowbreak absence\_verifier} & After grounding or zooming, checks whether a small object, part, or marking (a handle, wheel, blade, knob, \ldots) is actually visible, and returns an explicit visible or not-visible status instead of a raw detection. & 12/30 & 7/15 \\
\addlinespace
\tn{object\_layout\_\allowbreak relation\_resolver} & Converts grounded boxes of objects or parts into explicit layout relations: which end or side holds a part, immediate neighbors, row order, left-versus-right counts, and symmetry. & 4/30 & 3/15 \\
\addlinespace
\tn{visible\_part\_\allowbreak count\_verifier} & Checks candidate boxes of repeated parts (legs, keys, spokes, \ldots), merges duplicates, and rejects off-target boxes, returning the visible count rather than the typical one. & 2/30 & 1/15 \\
\bottomrule
\end{tabular}
\caption{Accepted task-specific tools and recovery on their supporting examples. \emph{Gate}: recovered failures / all examples for that direction. \emph{Held-out}: recovered failures / examples withheld from the tool-writing agent. The held-out subset is included in the gate total. These counts measure recovery on selected failures, not accuracy on the full target benchmark.}
\label{tab:appendix_generated_tools}
\end{table}

Together, these examples illustrate two ways to adapt the same general-purpose harness to custom data: refine the instructions and rendered evidence when existing capabilities suffice, or add reusable operations when failures call for explicit interpretation, output refinement, or tool composition.

%% file: text/appendix_examples.tex

\definecolor{exPercept}{HTML}{1F6F8B}   
\definecolor{exSearch}{HTML}{B4530A}    
\definecolor{exCompute}{HTML}{6A2E7A}   
\definecolor{exGood}{HTML}{137752}
\definecolor{exBad}{HTML}{B42318}
\definecolor{exInk}{HTML}{15181C}
\definecolor{exMute}{HTML}{5F6873}
\definecolor{exPanel}{HTML}{F2F3F5}
\definecolor{exHarness}{HTML}{FBF7EC}
\definecolor{exLine}{HTML}{D5D9E0}
\definecolor{exGuide}{HTML}{2E6B47}     
\definecolor{exGuideBack}{HTML}{EEF6F0}
\newcounter{exnum}

\newcommand{\exTitle}[2]{\par\refstepcounter{exnum}\label{#1}%
  \begin{tcolorbox}[enhanced, colback=exInk, colframe=exInk, coltext=white, arc=1.5pt,
    left=6pt, right=6pt, top=4pt, bottom=4pt, before skip=0pt, after skip=8pt,
    fontupper=\bfseries\normalsize]Example~\theexnum\quad #2\end{tcolorbox}}
\newcommand{\exTool}[2]{\tcbox[on line, colback=#1, colframe=#1, coltext=white,
  boxsep=0pt, left=3pt, right=3pt, top=1.5pt, bottom=1.5pt, arc=1pt,
  fontupper=\sffamily\bfseries\scriptsize]{#2}}
\newcommand{\toolVG}{\exTool{exPercept}{Visual Grounding}}
\newcommand{\toolZoom}{\exTool{exPercept}{Zoom In}}
\newcommand{\toolDepth}{\exTool{exPercept}{Depth Estimation}}
\newcommand{\toolCam}{\exTool{exPercept}{Camera Trajectory}}
\newcommand{\toolOCR}{\exTool{exPercept}{OCR}}
\newcommand{\toolIS}{\exTool{exSearch}{Image Search}}
\newcommand{\toolTS}{\exTool{exSearch}{Text Search}}
\newcommand{\toolWV}{\exTool{exSearch}{Webpage Visit}}
\newcommand{\toolCode}{\exTool{exCompute}{Python Coding Agent}}
\newenvironment{exstep}[3]{\par\medskip\noindent
  {\color{exLine}\rule{\linewidth}{0.6pt}}\par\nopagebreak\vspace{2pt}\noindent
  {\color{exMute}\small Example~\theexnum\ \textperiodcentered\ }\ifx\relax#1\relax\else{\large\bfseries Step #1}\hspace{0.8em}\fi#2\par\nopagebreak\vspace{2pt}}{\par}
\newcommand{\exLabel}[2]{\par\smallskip\noindent{\color{exMute}\sffamily\scriptsize\bfseries\MakeUppercase{#1}}%
  \ifx\relax#2\relax\else{\color{exMute}\scriptsize\ \ (#2)}\fi\par\nopagebreak\vspace{1pt}}
\newcommand{\exVerbBack}{exPanel}
\newenvironment{exverb}{\begin{tcolorbox}[breakable, colback=\exVerbBack, colframe=\exVerbBack,
  boxrule=0pt, arc=1pt, left=4pt, right=4pt, top=3pt, bottom=3pt, before skip=1pt, after skip=3pt,
  fontupper=\ttfamily\scriptsize\raggedright]}{\end{tcolorbox}}
\newenvironment{exharness}{\par\begingroup\renewcommand{\exVerbBack}{exHarness}}{\endgroup\par}
\newcommand{\exSayBlock}[1]{\begin{tcolorbox}[colback=white, colframe=exInk, boxrule=0pt, leftrule=2pt,
  arc=0pt, left=5pt, right=4pt, top=2pt, bottom=2pt, before skip=1pt, after skip=3pt,
  fontupper=\small\itshape]#1\end{tcolorbox}}
\newcommand{\exCut}[1]{\par\noindent{\color{exBad}\scriptsize\itshape [#1]}\par}
\newcommand{\exRemark}[1]{\par\smallskip\noindent{\color{exMute}\sffamily\scriptsize\bfseries NOTE}\hspace{0.4em}{\color{exMute}\footnotesize #1}\par}
\newtcolorbox{exnote}{colback=exPanel, colframe=exPanel, boxrule=0pt, arc=1pt,
  left=6pt, right=6pt, top=4pt, bottom=4pt, before skip=10pt, after skip=4pt, fontupper=\small}
\newcommand{\exRight}{{\color{exGood}\ding{51}}}
\newcommand{\exWrong}{{\color{exBad}\ding{55}}}
\newcommand{\exOverview}[1]{\begin{tcolorbox}[enhanced, breakable, colback=exGuideBack, colframe=exGuide,
  boxrule=0.5pt, arc=1.5pt, left=7pt, right=7pt, top=5pt, bottom=5pt, before skip=0pt, after skip=8pt,
  fontupper=\small, title={Overview\ \ {\normalfont\footnotesize(our summary)}}, fonttitle=\sffamily\bfseries\small,
  coltitle=white, colbacktitle=exGuide, toptitle=1pt, bottomtitle=1pt]#1\end{tcolorbox}}
\newcommand{\exWalk}[1]{\begin{tcolorbox}[enhanced, colback=exGuideBack, colframe=exGuideBack, boxrule=0pt,
  leftrule=2.5pt, colframe=exGuide, arc=0pt, left=6pt, right=5pt, top=3pt, bottom=3pt, before skip=2pt, after skip=4pt,
  fontupper=\small]{\color{exGuide}\sffamily\bfseries\scriptsize WHAT HAPPENS}\hspace{0.5em}#1\end{tcolorbox}}
\newcommand{\exEnd}{\par\vspace{6pt}\noindent{\color{exInk}\rule{\linewidth}{1.2pt}}\par\nopagebreak\noindent
  \hfill{\color{exMute}\sffamily\scriptsize\bfseries END OF EXAMPLE~\theexnum}\par}
\newcommand{\exTrajHead}{\par\bigskip\noindent{\large\bfseries With \ours: full trajectory}\par\nopagebreak\smallskip}

\section{Qualitative Examples}
\label{sec:appendix_examples}

This section shows seventeen complete \ours trajectories, ordered as the datasets appear in Table~\ref{tab:harness-combined}: grounded perception (Examples~\ref{ex:fsc}--\ref{ex:fruit}), visual search and deep research (Examples~\ref{ex:smsp161}--\ref{ex:bcvl}), general VQA and hallucination (Examples~\ref{ex:qrwqa77}--\ref{ex:vstar}) and spatial reasoning (Examples~\ref{ex:mmsi}--\ref{ex:q558}). Eleven come from GPT-6 Sol and GPT-6 Luna, taken from the runs behind Table~\ref{tab:harness-combined}; six come from \ours runs of Qwen3-VL-32B. Each example starts on a new page under a dark title bar and ends with an ``End of Example'' rule, and every step header names its example. Each example is laid out the same way:

\begin{itemize}[leftmargin=1.2em,itemsep=1pt,topsep=2pt]
\item \textbf{Overview and step notes.} Each example opens with a \colorbox{exGuideBack}{green overview box} that explains in plain words what the task asks, why it is hard without tools, and how the agent solves it. For the proprietary GPT-6 models, each step then opens with a short green ``What happens'' note that says what the call does and why it matters. We add these notes only for proprietary models, whose reasoning traces we cannot access: the API returns the tool calls and any visible text, but not the model's reasoning, and GPT-6 often calls a tool without writing any text at all. The open Qwen3-VL models expose their reasoning, which they write in \texttt{<thinking>} tags before each call and which we show verbatim, so their examples have no step notes. The overviews and notes are our descriptions, written to guide the reader; everything else in the trajectory is verbatim.
\item \textbf{Inputs, question and results.} The input images and the question exactly as sent to the model, then a table comparing the answer with and without the harness against the ground truth. For grounded perception examples, the table also shows what SAM~3, the model behind Visual Grounding, returns when queried alone, where the trajectory makes that measurable. For some Qwen3-VL examples no run of the same model without tools is available, and the table says so.
\item \textbf{Without tools.} The same model's full response to the same question with no tools, where such a run exists.
\item \textbf{Full trajectory with \ours.} Every tool call in order. Each step shows the tool, colour-coded by family (\textcolor{exPercept}{\textbf{perception}}, \textcolor{exSearch}{\textbf{search}}, \textcolor{exCompute}{\textbf{computation}}); the complete arguments the model sent; the text the model wrote alongside the call, when there is any; the tool's output, including the complete code the coding agent wrote and its execution output; and every image the tool returned, which the model sees in its next turn. For grounding examples, whose final answer is a list of coordinates, we add a panel drawing the final boxes on the input image, marked as ours.
\item \textbf{Harness messages.} On a \colorbox{exHarness}{cream background}: the environment updates through which the multimodal memory tells the model what it currently holds (Appendix~\ref{sec:appendix_memory}), including the data bindings that pass tool results to the coding agent.
\item \textbf{What to notice.} A closing note on how the harness contributed.
\end{itemize}

All model text, arguments and tool outputs are verbatim, generated directly from the trajectory files. To keep each example readable we shorten in five ways, and every cut is marked in red where it happens. Search results show their first two entries, with the number omitted. Coding-agent execution outputs longer than 70 lines keep their first 25 and last 10 lines; the code itself is always complete. A coding-agent run that failed with an error shows the error but not its code. Other tool outputs longer than 1{,}500 characters are cut, with the remaining length stated; one such output in Example~\ref{ex:church} is a webpage summary that loops, repeating the same sentences for over 128{,}000 characters. After the first environment update, which is shown in full, each update is summarized as the memory entries it adds, with any new data bindings kept verbatim. We also show escaped newlines inside JSON strings as line breaks, and emoji in search results as [emoji] or [flag]. In Example~\ref{ex:mmsp} the question and the model's reasoning are in Chinese; the overview translates the question. The system prompt is given in Appendix~\ref{sec:appendix_sysprompt}.

The examples were chosen by hand to illustrate how the harness composes tools, and are not a random sample. Not every one is a clean win, and the overviews say so where it matters: in Example~\ref{ex:fruit} the no-tool answer fails on coordinate format; in Example~\ref{ex:smsp161} the perception step does not contribute; and in Examples~\ref{ex:q558}, \ref{ex:qmmsi7} and~\ref{ex:bcvl} the final answer is correct but only partly supported by the trajectory's evidence.

\clearpage
\exTitle{ex:fsc}{Dataset: FSC-147\qquad Model: GPT-6 Sol}
\input{text/appendix_examples_steps/fsc1256_header}
\exLabel{Results}{}
\begin{center}\small
\begin{tabular}{@{}lr@{}}
\toprule
SAM~3 alone (query ``pill'') & 134 \exWrong \\
SAM~3 alone (query ``round pill'') & 72 \exWrong \\
GPT-6 Sol, no tools & 137 \exWrong \\
GPT-6 Sol + \ours & \textbf{139} \exRight \\
\midrule
Ground truth & 139 \\
\bottomrule
\end{tabular}
\end{center}
\exTrajHead
\input{text/appendix_examples_steps/fsc1256}
\begin{exnote}
\textbf{What to notice.} SAM~3 finds 134 pills with the query ``pill'' and 72 with ``round pill''; the two sets overlap, and neither is the answer. The model asks the coding agent to merge them. The first run reports zero matches and a union of zero (Step~3), which cannot be right, since both queries find the same pills. The model rewrites the objective: match by box overlap or centre distance only, with no mask decoding. The second run (Step~4) finds five round pills missing from the first set, and the model answers $134 + 5 = 139$, the exact count. Without tools the model says 137. The detections reach the code as data bindings held in memory, so the model can recover from a bad run by restating the objective instead of re-detecting.
\end{exnote}
\exEnd

\clearpage
\exTitle{ex:voc331}{Dataset: OdinW-13-All (PascalVOC subset)\qquad Model: GPT-6 Sol}
\input{text/appendix_examples_steps/voc331_header}
\exLabel{Results}{per-image counts from the benchmark evaluator; F1 at IoU $\ge 0.5$; ground-truth categories: person, bottle, dining table, chair}
\begin{center}\small
\begin{tabular}{@{}lrrrr@{}}
\toprule
& TP & FP & FN & F1 \\
\midrule
GPT-6 Sol, no tools & 16 & 5 & 7 & 0.73 \\
GPT-6 Sol + \ours & \textbf{22} & 1 & 1 & \textbf{0.96} \\
\midrule
Ground-truth boxes & \multicolumn{4}{c}{23} \\
\bottomrule
\end{tabular}
\end{center}
\exTrajHead
\input{text/appendix_examples_steps/voc331}
\exRemark{Panel drawn by us from the final answer: the 23 predicted boxes on the input image, coloured by label.}
\begin{center}\includegraphics[width=0.62\linewidth,height=0.3\textheight,keepaspectratio]{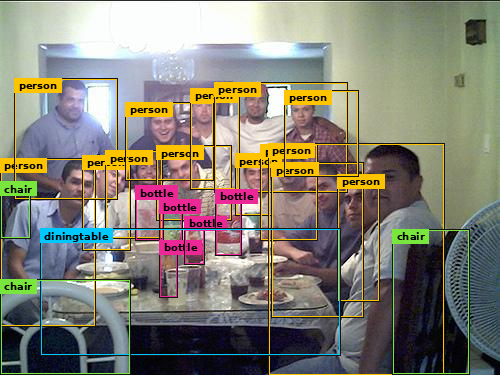}\end{center}
\begin{exnote}
\textbf{What to notice.} After grounding the four categories present, the model zooms on the crowded table and issues narrower queries for what the first pass missed: ``table'', ``man'', ``person in dark shirt'', and finally ``man behind the man in white shirt'' for a partly hidden person. The last query is a referring expression, not a category name; SAM~3 can resolve it, and the harness lets the model use that ability. The final answer matches 22 of the 23 ground-truth objects with one false positive.
\end{exnote}
\exEnd

\clearpage
\exTitle{ex:fruit}{Dataset: FSCD-147\qquad Model: GPT-6 Sol}
\input{text/appendix_examples_steps/fruit_header}
\exLabel{Results}{F1 at IoU $\ge 0.5$ against the 53 ground-truth boxes}
\begin{center}\small
\begin{tabular}{@{}lrrr@{}}
\toprule
& boxes & F1 & mean IoU \\
\midrule
SAM~3 alone (query ``package of fresh cut fruit'') & 25 & 0.64 \exWrong & -- \\
SAM~3 alone (query ``fruit cup'') & 57 & 0.95 \exWrong & -- \\
GPT-6 Sol, no tools & 53 & 0.00 \exWrong & -- \\
GPT-6 Sol + \ours & \textbf{53} & \textbf{1.00} \exRight & \textbf{0.87} \\
\midrule
Ground truth & 53 & & \\
\bottomrule
\end{tabular}
\end{center}
\exRemark{The no-tool answer gives its boxes in pixel coordinates (the image is 731$\times$384) instead of the requested 0--1000 scale, so every box misses. Rescaled to that scale, its boxes would score F1 1.00 with mean IoU 0.81. Here the harness's gain is mostly the answer's coordinate frame, plus tighter boxes.}
\exTrajHead
\input{text/appendix_examples_steps/fruit}
\exRemark{Panel drawn by us from the final answer: the 53 predicted boxes on the input image.}
\begin{center}\includegraphics[width=0.62\linewidth,height=0.3\textheight,keepaspectratio]{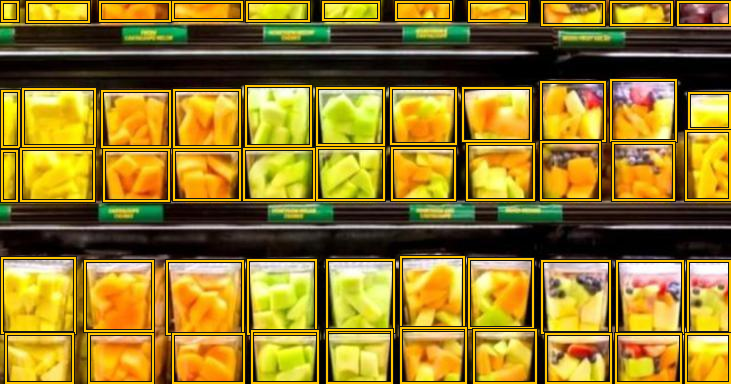}\end{center}
\begin{exnote}
\textbf{What to notice.} The category name as asked, ``package of fresh cut fruit'', finds only 25 candidates. The model rephrases it as ``fruit cup'', which finds 57, more than the true 53. It zooms on two regions and settles on 53 boxes that match every ground-truth package. The trajectory does not state how the final four candidates were dropped. As the note above says, the no-tool model's score of zero comes from its coordinate format, not from missing the objects.
\end{exnote}
\exEnd

\clearpage
\exTitle{ex:smsp161}{Dataset: MMSearch-Plus\qquad Model: GPT-6 Luna}
\input{text/appendix_examples_steps/smsp161_header}
\exLabel{Results}{correctness from the GPT-4o judge}
\begin{center}\small
\begin{tabular}{@{}l>{\raggedright\arraybackslash}p{0.55\linewidth}@{}}
\toprule
GPT-6 Luna, no tools & Miu Miu; 2024; the Miu Miu Spring/Summer 2025 show \exWrong \\
GPT-6 Luna + \ours & \textbf{the Hermès show, Paris Fashion Week Fall/Winter 2025} \exRight \\
\midrule
Ground truth & StreetStyle Winter 2025 Hermes \\
\bottomrule
\end{tabular}
\end{center}
\exTrajHead
\input{text/appendix_examples_steps/smsp161}
\begin{exnote}
\textbf{What to notice.} When reverse image search fails, the agent turns the photograph into words and searches with them, which finds the exact street-style gallery the photo comes from. The zoom and OCR in the middle of the trajectory read an unrelated sign, and the agent is not misled by it: the answer rests on the gallery page.
\end{exnote}
\exEnd

\clearpage
\exTitle{ex:mmsp}{Dataset: MMSearch-Plus\qquad Model: Qwen3-VL-32B}
\input{text/appendix_examples_steps/mmsp_header}
\exLabel{Results}{no run of this model without tools is available for this question}
\begin{center}\small
\begin{tabular}{@{}lr@{}}
\toprule
Qwen3-VL-32B + \ours & \textbf{\begin{CJK}{UTF8}{gbsn}9年\end{CJK}} (9 years) \exRight \\
\midrule
Ground truth & 9 \\
\bottomrule
\end{tabular}
\end{center}
\exTrajHead
\input{text/appendix_examples_steps/mmsp}
\begin{exnote}
\textbf{What to notice.} Each image is identified by its own reverse image search, and each event is dated from a web page, so the answer needs two independent chains of evidence and one subtraction. When a page fetch fails with a server error (Step~4), the harness reports the failure and suggests alternatives, and the model recovers with a text search instead of guessing the date.
\end{exnote}
\exEnd

\clearpage
\exTitle{ex:srx140}{Dataset: RealX-Bench\qquad Model: GPT-6 Luna}
\input{text/appendix_examples_steps/srx140_header}
\exLabel{Results}{correctness from the GPT-4o judge}
\begin{center}\small
\begin{tabular}{@{}l>{\raggedright\arraybackslash}p{0.55\linewidth}@{}}
\toprule
GPT-6 Luna, no tools & AHC \exWrong \\
GPT-6 Luna + \ours & \textbf{MYONS (\begin{CJK}{UTF8}{gbsn}弥\CJKfamily{bsmi}玥\CJKfamily{gbsn}泉\end{CJK})} \exRight \\
\midrule
Ground truth & Miyue Spring \\
\bottomrule
\end{tabular}
\end{center}
\exTrajHead
\input{text/appendix_examples_steps/srx140}
\begin{exnote}
\textbf{What to notice.} When the whole-image search fails, the agent does not keep searching blindly: it returns to the image, uses grounding to isolate the target among similar bottles, and turns a tight crop into the search query. Perception and search alternate several times, and the product is identified only once the crop is tight enough for the reverse search to match it.
\end{exnote}
\exEnd

\clearpage
\exTitle{ex:srx2}{Dataset: RealX-Bench\qquad Model: GPT-6 Sol}
\input{text/appendix_examples_steps/srx2_header}
\exLabel{Results}{correctness from the GPT-4o judge}
\begin{center}\small
\begin{tabular}{@{}l>{\raggedright\arraybackslash}p{0.55\linewidth}@{}}
\toprule
GPT-6 Sol, no tools & Dior \exWrong \\
GPT-6 Sol + \ours & \textbf{KATO-KATO} \exRight \\
\midrule
Ground truth & KATO \\
\bottomrule
\end{tabular}
\end{center}
\exTrajHead
\input{text/appendix_examples_steps/srx2}
\begin{exnote}
\textbf{What to notice.} With no readable text, the agent searches by what it can see (the faceted cap), states and rejects a brand hypothesis, and returns to reverse image search on its own crop. The final targeted search lets it check the candidate brand against the cap it saw, rather than accept the name on trust.
\end{exnote}
\exEnd

\clearpage
\exTitle{ex:fvqa}{Dataset: FVQA\qquad Model: Qwen3-VL-32B}
\input{text/appendix_examples_steps/fvqa_header}
\exLabel{Results}{no run of this model without tools is available for this question}
\begin{center}\small
\begin{tabular}{@{}l>{\raggedright\arraybackslash}p{0.55\linewidth}@{}}
\toprule
Qwen3-VL-32B + \ours & \textbf{The Deuter Speed Lite Pro backpack has won the ISPO Award 2024} \exRight \\
\midrule
Ground truth & ISPO Award 2024 \\
\bottomrule
\end{tabular}
\end{center}
\exTrajHead
\input{text/appendix_examples_steps/fvqa}
\begin{exnote}
\textbf{What to notice.} The answer depends on first naming a product that no text in the image names. Perception tools narrow the photo down to the right object (grounding, zoom), OCR fails, and the model switches to reverse image search on its own crop, which identifies the product. It then does not stop at the first page that lacks the answer: it searches for the award directly and verifies it at the source.
\end{exnote}
\exEnd

\clearpage
\exTitle{ex:bcvl}{Dataset: BrowseComp-VL\qquad Model: Qwen3-VL-32B}
\input{text/appendix_examples_steps/bcvl_header}
\exLabel{Results}{no run of this model without tools is available for this question}
\begin{center}\small
\begin{tabular}{@{}l>{\raggedright\arraybackslash}p{0.55\linewidth}@{}}
\toprule
Qwen3-VL-32B + \ours & \textbf{Yunnan, Guizhou, Sichuan, and Guangxi provinces in southwestern China} \exRight \\
\midrule
Ground truth & Sichuan, eastern Yunnan, northern Guangxi, and Guizhou \\
\bottomrule
\end{tabular}
\end{center}
\exTrajHead
\input{text/appendix_examples_steps/bcvl}
\begin{exnote}
\textbf{What to notice.} The most useful part of this trajectory is the self-correction in Steps~3 and~4: the model states a specific hypothesis, looks it up, finds that the species is native to Java and Sumatra rather than China, and discards it. The weakness is the ending: the province list it answers with came from its own earlier guess, and the retrieved pages support only part of it.
\end{exnote}
\exEnd

\clearpage
\exTitle{ex:qrwqa77}{Dataset: RealWorldQA\qquad Model: Qwen3-VL-32B}
\input{text/appendix_examples_steps/qrwqa77_header}
\exLabel{Results}{}
\begin{center}\small
\begin{tabular}{@{}lr@{}}
\toprule
Qwen3-VL-32B, no tools & A.\ Yes \exWrong \\
Qwen3-VL-32B + \ours & \textbf{B.\ No} \exRight \\
\midrule
Ground truth & B \\
\bottomrule
\end{tabular}
\end{center}
\exTrajHead
\input{text/appendix_examples_steps/qrwqa77}
\begin{exnote}
\textbf{What to notice.} The question compares distances of two kinds of object along a street, which is easy to misjudge in a single photograph. The harness turns it into arithmetic: depth estimation gives metric depth, grounding gives the crosswalk and every car, and a few lines of code compare them. The detections and the depth map reach the code as data bindings held in memory, \texttt{segment\_crosswalk}, \texttt{segment\_cars} and \texttt{depth\_result}.
\end{exnote}
\exEnd

\clearpage
\exTitle{ex:church}{Dataset: SimpleVQA\qquad Model: GPT-6 Sol}
\input{text/appendix_examples_steps/church_header}
\exLabel{Results}{}
\begin{center}\small
\begin{tabular}{@{}l>{\raggedright\arraybackslash}p{0.5\linewidth}@{}}
\toprule
GPT-6 Sol, no tools & Cathedral of Our Lady in Antwerp, Belgium \exWrong \\
GPT-6 Sol + \ours & \textbf{St.\ Augustine's Parish Church in Perchtoldsdorf, Austria} \exRight \\
\midrule
Ground truth & Pfarrkirche hl.\ Augustinus, Perchtoldsdorf \\
\bottomrule
\end{tabular}
\end{center}
\exTrajHead
\input{text/appendix_examples_steps/church}
\begin{exnote}
\textbf{What to notice.} Searching the whole photograph returns look-alike Gothic interiors, and eleven calls of text search, page visits and a wide crop do not narrow them down; the model tries Antwerp, Münster and Eichstätt along the way. Cropping the altarpiece (Step~12) turns the reverse image search into a specific one: its first result is an archive photo of this exact altar painting (Step~13). A text search on that lead then confirms the church (Step~14). Crop, reverse image search and text verification come from different tool families, composed through images and web pages the harness keeps addressable in memory.
\end{exnote}
\exEnd

\clearpage
\exTitle{ex:vstar}{Dataset: V*\qquad Model: GPT-6 Sol}
\input{text/appendix_examples_steps/vstar_header}
\exLabel{Results}{}
\begin{center}\small
\begin{tabular}{@{}lr@{}}
\toprule
GPT-6 Sol, no tools & A.\ right \exWrong \\
GPT-6 Sol + \ours & \textbf{B.\ left} \exRight \\
\midrule
Ground truth & B \\
\bottomrule
\end{tabular}
\end{center}
\exTrajHead
\input{text/appendix_examples_steps/vstar}
\begin{exnote}
\textbf{What to notice.} The two people are a few pixels tall in the full image. Grounding them there returns almost the same tiny box for both queries (Steps~1 and~2), so it cannot separate them. The model zooms three times, each time closer, and grounds again on the final crop (Steps~6 and~7). There the woman with a backpack gets her own box, and the beanie query returns a second person to her right, so the model answers that she is on the left. The beanie query also returns the woman's box, so this last step still needs the model's judgement.
\end{exnote}
\exEnd

\clearpage
\exTitle{ex:mmsi}{Dataset: MMSI-Bench\qquad Model: GPT-6 Sol}
\input{text/appendix_examples_steps/mmsi_header}
\exLabel{Results}{}
\begin{center}\small
\begin{tabular}{@{}lr@{}}
\toprule
GPT-6 Sol, no tools & D (12 throw pillows) \exWrong \\
GPT-6 Sol + \ours & \textbf{C (16 throw pillows)} \exRight \\
\midrule
Ground truth & C \\
\bottomrule
\end{tabular}
\end{center}
\exTrajHead
\input{text/appendix_examples_steps/mmsi}
\begin{exnote}
\textbf{What to notice.} Of the four options, two are pillow counts, so the question reduces to counting throw pillows across seven views of one room, including partly covered ones. The model grounds ``throw pillows'' in the four views that show seating (11, 3, 3 and 1 detections, 18 in total), then zooms on the two halves of the sofa in view~5. It answers 16, which requires recognizing that some pillows appear in more than one view. That cross-view deduplication is implicit: the model does not write it out.
\end{exnote}
\exEnd

\clearpage
\exTitle{ex:qmmsi7}{Dataset: MMSI-Bench\qquad Model: Qwen3-VL-32B}
\input{text/appendix_examples_steps/qmmsi7_header}
\exLabel{Results}{no-tool answer from a separate run of the same model, final answer only}
\begin{center}\small
\begin{tabular}{@{}lr@{}}
\toprule
Qwen3-VL-32B, no tools & D (the piano is longer) \exWrong \\
Qwen3-VL-32B + \ours & \textbf{A (the table under the mirror is longer)} \exRight \\
\midrule
Ground truth & A \\
\bottomrule
\end{tabular}
\end{center}
\exTrajHead
\input{text/appendix_examples_steps/qmmsi7}
\begin{exnote}
\textbf{What to notice.} The model composes grounding, depth and code to turn a comparison between two photographs into a metric one, and it states the plan before the first call. The example also shows a failure mode the harness does not prevent: the model picks the wrong object among the grounding results, and nothing later in the chain checks that choice.
\end{exnote}
\exEnd

\clearpage
\exTitle{ex:spar}{Dataset: SparBench\qquad Model: GPT-6 Sol}
\input{text/appendix_examples_steps/spar_header}
\exLabel{Results}{MRA: SparBench's mean relative accuracy over thresholds 0.50--0.95}
\begin{center}\small
\begin{tabular}{@{}lrr@{}}
\toprule
 & answer & MRA \\
\midrule
GPT-6 Sol, no tools & 4.0 m & 0.00 \exWrong \\
GPT-6 Sol + \ours & \textbf{2.0 m} & \textbf{0.90} \exRight \\
\midrule
Ground truth & 1.9 m & \\
\bottomrule
\end{tabular}
\end{center}
\exTrajHead
\input{text/appendix_examples_steps/spar}
\begin{exnote}
\textbf{What to notice.} The two points are never visible in the same image, so the distance cannot be read off any one view. The harness supplies the three pieces a metric answer needs: camera poses relating the views (Step~1), metric depth in each view (Steps~2 and~3), and code that back-projects both points into a shared frame (Step~4). The model's own contribution is to locate the two markers in pixels and to state the computation. Tool results pass to the coding agent as named variables held in memory, such as \texttt{camera\_trajectory\_result} and \texttt{depth\_result\_2}, not as text the model must copy.
\end{exnote}
\exEnd

\clearpage
\exTitle{ex:cab}{Dataset: SparBench\qquad Model: GPT-6 Luna}
\input{text/appendix_examples_steps/cab_header}
\exLabel{Results}{MRA: SparBench's mean relative accuracy over thresholds 0.50--0.95}
\begin{center}\small
\begin{tabular}{@{}lrr@{}}
\toprule
 & answer & MRA \\
\midrule
GPT-6 Luna, no tools & 2.0 m & 0.00 \exWrong \\
GPT-6 Luna + \ours & \textbf{2.4} & \textbf{1.00} \exRight \\
\midrule
Ground truth & 2.4 & \\
\bottomrule
\end{tabular}
\end{center}
\exTrajHead
\input{text/appendix_examples_steps/cab}
\begin{exnote}
\textbf{What to notice.} The question asks for the distance from the observer of the first view, but the cabinet is marked only in the third view, taken about 0.7\,m behind and 0.5\,m to the left (Step~1). Depth read at the marker in view~3 is 2.16\,m (Step~4), which is the distance from the wrong camera. The final code run unprojects the marker into view~3's camera frame and moves it into the observer's frame with the camera-trajectory extrinsics, giving 2.40\,m (Step~5). A run that failed on a missing import (Step~3) is simply retried with a narrower objective.
\end{exnote}
\exEnd

\clearpage
\exTitle{ex:q558}{Dataset: SparBench\qquad Model: Qwen3-VL-32B}
\input{text/appendix_examples_steps/q558_header}
\exLabel{Results}{MRA: SparBench's mean relative accuracy over thresholds 0.50--0.95; no-tool answer from a separate run of the same model, final answer only}
\begin{center}\small
\begin{tabular}{@{}lrr@{}}
\toprule
& answer & MRA \\
\midrule
Qwen3-VL-32B, no tools & 1.7 & 0.00 \exWrong \\
Qwen3-VL-32B + \ours & \textbf{1.1957} & \textbf{0.90} \exRight \\
\midrule
Ground truth & 1.1 & \\
\bottomrule
\end{tabular}
\end{center}
\exTrajHead
\input{text/appendix_examples_steps/q558}
\begin{exnote}
\textbf{What to notice.} The tool chain is the one a metric cross-view distance needs: camera poses, depth in each view, grounding to find the objects, and code to combine them, and the model plans it in its own words before the first call. The weak link is the last step: the generated code misreads the box format and measures from the wrong cabinet, so the close agreement with the ground truth should not be read as a successful measurement.
\end{exnote}
\exEnd

%% file: text/appendix_examples_steps/fsc1256_header.tex
\exOverview{\textbf{Task.} Count the pills in a photograph of about 140 pills of many colours and two shapes, oblong capsules and small round tablets, scattered on a dark surface. 

\textbf{Why it is hard.} There are too many to count reliably by eye, and a single detector query does not reliably find both shapes. 

\textbf{How the agent solves it.} It runs two grounding queries, one general and one for round pills, then asks the coding agent to merge the two sets of detections without counting any pill twice. The first merge fails silently; the agent notices the impossible result and rewrites the merge, which then finds the five pills the first query missed.}
\exLabel{Input images}{}
\begin{center}\includegraphics[width=0.55\linewidth,height=0.28\textheight,keepaspectratio]{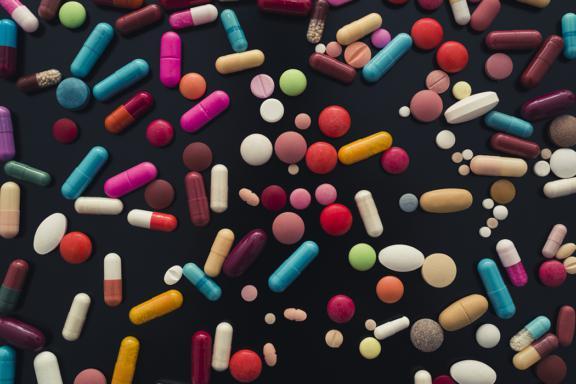}\\[2pt]
{\scriptsize\color{exMute} input\_image\_1}\end{center}
\exLabel{Question}{as sent to the model}
\begin{exverb}
Question: How many pill are there in the image? Your final answer should be a single integer.
\end{exverb}
\exLabel{Without tools}{GPT-6 Sol's full response to the same question}
\begin{exverb}
137
\end{exverb}

%% file: text/appendix_examples_steps/fsc1256.tex
\begin{exharness}
\exLabel{Harness}{environment update, shown in full}
\begin{exverb}
Environment Update (not a new question):\\{}
Currently available image IDs for tool calls:\\{}
- input\_\allowbreak{}image\_\allowbreak{}1: user\_\allowbreak{}input\\{}
When a tool needs image input,\allowbreak{} pass one of these IDs via the *\_\allowbreak{}image\_\allowbreak{}id field.\\{}
Do NOT invent image file names or paths.
\end{exverb}
\end{exharness}
\begin{exstep}{1}{\toolVG}{}
\exWalk{The agent starts with the general query ``pill''. SAM~3 returns 134 detections, drawn as numbered masks on the image.}
\exLabel{Arguments}{}
\begin{exverb}
\{\\{}
~~"\allowbreak{}query"\allowbreak{}: "\allowbreak{}pill"\allowbreak{},\allowbreak{}\\{}
~~"\allowbreak{}image\_\allowbreak{}id"\allowbreak{}: "\allowbreak{}input\_\allowbreak{}image\_\allowbreak{}1"\allowbreak{}\\{}
\}
\end{exverb}
\exLabel{Tool output}{}
\begin{exverb}
Visual\_\allowbreak{}Grounding\_\allowbreak{}Tool grounded 134 objects for the prompt \textquotesingle{}pill\textquotesingle{}. Their corresponding bounding boxes (on a 0-1000 scale,\allowbreak{} [x1,\allowbreak{} y1,\allowbreak{} x2,\allowbreak{} y2]) are: \textasciigrave{}\textasciigrave{}\textasciigrave{}json\\{}
[\\{}
~~~~\{"\allowbreak{}bbox\_\allowbreak{}2d"\allowbreak{}: [0,\allowbreak{} 901,\allowbreak{} 26,\allowbreak{} 999],\allowbreak{} "\allowbreak{}label"\allowbreak{}: "\allowbreak{}pill"\allowbreak{}\},\allowbreak{}\\{}
~~~~\{"\allowbreak{}bbox\_\allowbreak{}2d"\allowbreak{}: [0,\allowbreak{} 52,\allowbreak{} 29,\allowbreak{} 204],\allowbreak{} "\allowbreak{}label"\allowbreak{}: "\allowbreak{}pill"\allowbreak{}\},\allowbreak{}\\{}
~~~~\{"\allowbreak{}bbox\_\allowbreak{}2d"\allowbreak{}: [0,\allowbreak{} 825,\allowbreak{} 78,\allowbreak{} 913],\allowbreak{} "\allowbreak{}label"\allowbreak{}: "\allowbreak{}pill"\allowbreak{}\},\allowbreak{}\\{}
~~~~\{"\allowbreak{}bbox\_\allowbreak{}2d"\allowbreak{}: [0,\allowbreak{} 278,\allowbreak{} 25,\allowbreak{} 417],\allowbreak{} "\allowbreak{}label"\allowbreak{}: "\allowbreak{}pill"\allowbreak{}\},\allowbreak{}\\{}
~~~~\{"\allowbreak{}bbox\_\allowbreak{}2d"\allowbreak{}: [0,\allowbreak{} 674,\allowbreak{} 49,\allowbreak{} 818],\allowbreak{} "\allowbreak{}label"\allowbreak{}: "\allowbreak{}pill"\allowbreak{}\},\allowbreak{}\\{}
~~~~\{"\allowbreak{}bbox\_\allowbreak{}2d"\allowbreak{}: [0,\allowbreak{} 473,\allowbreak{} 34,\allowbreak{} 622],\allowbreak{} "\allowbreak{}label"\allowbreak{}: "\allowbreak{}pill"\allowbreak{}\},\allowbreak{}\\{}
~~~~\{"\allowbreak{}bbox\_\allowbreak{}2d"\allowbreak{}: [1,\allowbreak{} 0,\allowbreak{} 66,\allowbreak{} 84],\allowbreak{} "\allowbreak{}label"\allowbreak{}: "\allowbreak{}pill"\allowbreak{}\},\allowbreak{}\\{}
~~~~\{"\allowbreak{}bbox\_\allowbreak{}2d"\allowbreak{}: [7,\allowbreak{} 418,\allowbreak{} 89,\allowbreak{} 485],\allowbreak{} "\allowbreak{}label"\allowbreak{}: "\allowbreak{}pill"\allowbreak{}\},\allowbreak{}\\{}
~~~~\{"\allowbreak{}bbox\_\allowbreak{}2d"\allowbreak{}: [29,\allowbreak{} 179,\allowbreak{} 107,\allowbreak{} 242],\allowbreak{} "\allowbreak{}label"\allowbreak{}: "\allowbreak{}pill"\allowbreak{}\},\allowbreak{}\\{}
~~~~\{"\allowbreak{}bbox\_\allowbreak{}2d"\allowbreak{}: [58,\allowbreak{} 554,\allowbreak{} 116,\allowbreak{} 663],\allowbreak{} "\allowbreak{}label"\allowbreak{}: "\allowbreak{}pill"\allowbreak{}\},\allowbreak{}\\{}
~~~~\{"\allowbreak{}bbox\_\allowbreak{}2d"\allowbreak{}: [59,\allowbreak{} 915,\allowbreak{} 109,\allowbreak{} 1000],\allowbreak{} "\allowbreak{}label"\allowbreak{}: "\allowbreak{}pill"\allowbreak{}\},\allowbreak{}\\{}
~~~~\{"\allowbreak{}bbox\_\allowbreak{}2d"\allowbreak{}: [65,\allowbreak{} 0,\allowbreak{} 106,\allowbreak{} 53],\allowbreak{} "\allowbreak{}label"\allowbreak{}: "\allowbreak{}pill"\allowbreak{}\},\allowbreak{}\\{}
~~~~\{"\allowbreak{}bbox\_\allowbreak{}2d"\allowbreak{}: [90,\allowbreak{} 307,\allowbreak{} 135,\allowbreak{} 375],\allowbreak{} "\allowbreak{}label"\allowbreak{}: "\allowbreak{}pill"\allowbreak{}\},\allowbreak{}\\{}
~~~~\{"\allowbreak{}bbox\_\allowbreak{}2d"\allowbreak{}: [97,\allowbreak{} 202,\allowbreak{} 156,\allowbreak{} 287],\allowbreak{} "\allowbreak{}label"\allowbreak{}: "\allowbreak{}pill"\allowbreak{}\},\allowbreak{}\\{}
~~~~\{"\allowbreak{}bbox\_\allowbreak{}2d"\allowbreak{}: [102,\allowbreak{} 4,\allowbreak{} 149,\allowbreak{} 75],\allowbreak{} "\allowbreak{}label"\allowbreak{}: "\allowbreak{}pill"\allowbreak{}\},\allowbreak{}\\{}
~~~~\{"\allowbreak{}bbox\_\allowbreak{}2d"\allowbreak{}: [103,\allowbreak{} 603,\allowbreak{} 159,\allowbreak{} 688],\allowbreak{} "\allowbreak{}label"\allowbreak{}: "\allowbreak{}pill"\allowbreak{}\},\allowbreak{}\\{}
~~~~\{"\allowbreak{}bbox\_\allowbreak{}2d"\allowbreak{}: [106,\allowbreak{} 380,\allowbreak{} 189,\allowbreak{} 520],\allowbreak{} "\allowbreak{}label"\allowbreak{}: "\allowbreak{}pill"\allowbreak{}\},\allowbreak{}\\{}
~~~~\{"\allowbreak{}bbox\_\allowbreak{}2d"\allowbreak{}: [107,\allowbreak{} 762,\allowbreak{} 157,\allowbreak{} 889],\allowbreak{} "\allowbreak{}label"\allowbreak{}: "\allowbreak{}pill"\allowbreak{}\},\allowbreak{}\\{}
~~~~\{"\allowbreak{}bbox\_\allowbreak{}2d"\allowbreak{}: [122,\allowbreak{} 59,\allowbreak{} 198,\allowbreak{} 204],\allowbreak{} "\allowbreak{}label"\allowbreak{}: "\allowbreak{}pill"\allowbreak{}\},\allowbreak{}\\{}
~~~~\{"\allowbreak{}bbox\_\allowbreak{}2d"\allowbreak{}: [128,\allowbreak{} 960,\allowbreak{} 184,\allowbreak{} 1000],\allowbreak{} "\allowbreak{}label"\allowbreak{}: "\allowbreak{}pill"\allowbreak{}\},\allowbreak{}\\{}
~~~~\{"\allowbreak{}bbox\_\allowbreak{}2d"\allowbreak{}: [130,\allowbreak{} 513,\allowbreak{} 214,\allowbreak{} 559],\allowbreak{} "\allowbreak{}label"\allowbreak{}: "\allowbreak{}pill"\allowbreak{}\},\allowbreak{}\\{}
~~~~\{"\allowbreak{}bbox\_\allowbreak{}2d"\allowbreak{}: [161,\allowbreak{} 153,\allowbreak{} 260,\allowbreak{} 264],\allowbreak{} "\allowbreak{}label"\allowbreak{}: "\allowbreak{}pill"\allowbreak{}\},\allowbreak{}\\{}
~~~~\{"\allowbreak{}bbox\_\allowbreak{}2d"\allowbreak{}: [163,\allowbreak{} 0,\allowbreak{} 249,\allowbreak{} 23],\allowbreak{} "\allowbreak{}label"\allowbreak{}: "\allowbreak{}pill"\allowbreak{}\},\allowbreak{}\\{}
~~~~\{"\allowbreak{}bbox\_\allowbreak{}2d"\allowbreak{}: [180,\allowbreak{} 658,\allowbreak{} 212,\allowbreak{} 799],\allowbreak{} "\allowbreak{}label"\allowbreak{}: "\allowbreak{}pill"\allowbreak{}\},\allowbreak{}\\{}
~~~~\{"\allowbreak{}bbox\_\allowbreak{}2d"\allowbreak{}: [181,\allowbreak{} 421,\allowbreak{} 273,\allowbreak{} 515],\allowbreak{} "\allowbreak{}label"\allowbreak{}: "\allowbreak{}pill"\allowbreak{}\},\allowbreak{}\\{}
~~~~\{"\allowbreak{}bbox\_\allowbreak{}2d"\allowbreak{}: [182,\allowbreak{} 237,\allowbreak{} 267,\allowbreak{} 348],\allowbreak{} "\allowbreak{}label"\allowbreak{}: "\allowbreak{}pill"\allowbreak{}\},\allowbreak{}
\end{exverb}
\exCut{The remaining 5,835 characters of this output are not shown (7,324 in total).}
\begin{center}\includegraphics[width=0.45\linewidth,height=0.22\textheight,keepaspectratio]{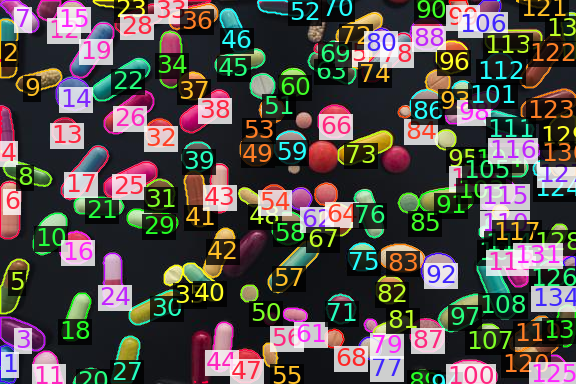}\\[2pt]
{\scriptsize\color{exMute} Image returned with this output.}\end{center}
\begin{exharness}
\exLabel{Harness}{environment update, summarized: new memory entries; new data bindings verbatim}
\begin{exverb}
Memory update: registers 1 new image in memory (tool\_\allowbreak{}generated\_\allowbreak{}image\_\allowbreak{}1,\allowbreak{} from Visual\_\allowbreak{}Grounding\_\allowbreak{}Tool).\\{}
Available data bindings (auto-injected into Python\_\allowbreak{}Coding\_\allowbreak{}Agent\_\allowbreak{}Tool):\\{}
- segment\_\allowbreak{}pill: dict (keys: pred\_\allowbreak{}boxes,\allowbreak{} pred\_\allowbreak{}scores,\allowbreak{} pred\_\allowbreak{}masks,\allowbreak{} num\_\allowbreak{}masks,\allowbreak{} orig\_\allowbreak{}img\_\allowbreak{}h,\allowbreak{} orig\_\allowbreak{}img\_\allowbreak{}w)
\end{exverb}
\end{exharness}
\end{exstep}
\begin{exstep}{2}{\toolVG}{}
\exWalk{A second query, ``round pill'', targets the small round tablets that a general query tends to miss. It returns 72 detections, most of which overlap with those of Step~1.}
\exLabel{Arguments}{}
\begin{exverb}
\{\\{}
~~"\allowbreak{}query"\allowbreak{}: "\allowbreak{}round pill"\allowbreak{},\allowbreak{}\\{}
~~"\allowbreak{}image\_\allowbreak{}id"\allowbreak{}: "\allowbreak{}input\_\allowbreak{}image\_\allowbreak{}1"\allowbreak{}\\{}
\}
\end{exverb}
\exLabel{Tool output}{}
\begin{exverb}
Visual\_\allowbreak{}Grounding\_\allowbreak{}Tool grounded 72 objects for the prompt \textquotesingle{}round pill\textquotesingle{}. Their corresponding bounding boxes (on a 0-1000 scale,\allowbreak{} [x1,\allowbreak{} y1,\allowbreak{} x2,\allowbreak{} y2]) are: \textasciigrave{}\textasciigrave{}\textasciigrave{}json\\{}
[\\{}
~~~~\{"\allowbreak{}bbox\_\allowbreak{}2d"\allowbreak{}: [91,\allowbreak{} 307,\allowbreak{} 136,\allowbreak{} 374],\allowbreak{} "\allowbreak{}label"\allowbreak{}: "\allowbreak{}round pill"\allowbreak{}\},\allowbreak{}\\{}
~~~~\{"\allowbreak{}bbox\_\allowbreak{}2d"\allowbreak{}: [98,\allowbreak{} 202,\allowbreak{} 156,\allowbreak{} 288],\allowbreak{} "\allowbreak{}label"\allowbreak{}: "\allowbreak{}round pill"\allowbreak{}\},\allowbreak{}\\{}
~~~~\{"\allowbreak{}bbox\_\allowbreak{}2d"\allowbreak{}: [102,\allowbreak{} 3,\allowbreak{} 150,\allowbreak{} 76],\allowbreak{} "\allowbreak{}label"\allowbreak{}: "\allowbreak{}round pill"\allowbreak{}\},\allowbreak{}\\{}
~~~~\{"\allowbreak{}bbox\_\allowbreak{}2d"\allowbreak{}: [104,\allowbreak{} 603,\allowbreak{} 160,\allowbreak{} 688],\allowbreak{} "\allowbreak{}label"\allowbreak{}: "\allowbreak{}round pill"\allowbreak{}\},\allowbreak{}\\{}
~~~~\{"\allowbreak{}bbox\_\allowbreak{}2d"\allowbreak{}: [128,\allowbreak{} 959,\allowbreak{} 185,\allowbreak{} 1000],\allowbreak{} "\allowbreak{}label"\allowbreak{}: "\allowbreak{}round pill"\allowbreak{}\},\allowbreak{}\\{}
~~~~\{"\allowbreak{}bbox\_\allowbreak{}2d"\allowbreak{}: [250,\allowbreak{} 467,\allowbreak{} 303,\allowbreak{} 547],\allowbreak{} "\allowbreak{}label"\allowbreak{}: "\allowbreak{}round pill"\allowbreak{}\},\allowbreak{}\\{}
~~~~\{"\allowbreak{}bbox\_\allowbreak{}2d"\allowbreak{}: [253,\allowbreak{} 310,\allowbreak{} 302,\allowbreak{} 384],\allowbreak{} "\allowbreak{}label"\allowbreak{}: "\allowbreak{}round pill"\allowbreak{}\},\allowbreak{}\\{}
~~~~\{"\allowbreak{}bbox\_\allowbreak{}2d"\allowbreak{}: [284,\allowbreak{} 691,\allowbreak{} 317,\allowbreak{} 743],\allowbreak{} "\allowbreak{}label"\allowbreak{}: "\allowbreak{}round pill"\allowbreak{}\},\allowbreak{}\\{}
~~~~\{"\allowbreak{}bbox\_\allowbreak{}2d"\allowbreak{}: [311,\allowbreak{} 190,\allowbreak{} 358,\allowbreak{} 262],\allowbreak{} "\allowbreak{}label"\allowbreak{}: "\allowbreak{}round pill"\allowbreak{}\},\allowbreak{}\\{}
~~~~\{"\allowbreak{}bbox\_\allowbreak{}2d"\allowbreak{}: [316,\allowbreak{} 370,\allowbreak{} 368,\allowbreak{} 446],\allowbreak{} "\allowbreak{}label"\allowbreak{}: "\allowbreak{}round pill"\allowbreak{}\},\allowbreak{}\\{}
~~~~\{"\allowbreak{}bbox\_\allowbreak{}2d"\allowbreak{}: [398,\allowbreak{} 911,\allowbreak{} 455,\allowbreak{} 998],\allowbreak{} "\allowbreak{}label"\allowbreak{}: "\allowbreak{}round pill"\allowbreak{}\},\allowbreak{}\\{}
~~~~\{"\allowbreak{}bbox\_\allowbreak{}2d"\allowbreak{}: [415,\allowbreak{} 491,\allowbreak{} 450,\allowbreak{} 538],\allowbreak{} "\allowbreak{}label"\allowbreak{}: "\allowbreak{}round pill"\allowbreak{}\},\allowbreak{}\\{}
~~~~\{"\allowbreak{}bbox\_\allowbreak{}2d"\allowbreak{}: [418,\allowbreak{} 742,\allowbreak{} 448,\allowbreak{} 787],\allowbreak{} "\allowbreak{}label"\allowbreak{}: "\allowbreak{}round pill"\allowbreak{}\},\allowbreak{}\\{}
~~~~\{"\allowbreak{}bbox\_\allowbreak{}2d"\allowbreak{}: [418,\allowbreak{} 348,\allowbreak{} 474,\allowbreak{} 431],\allowbreak{} "\allowbreak{}label"\allowbreak{}: "\allowbreak{}round pill"\allowbreak{}\},\allowbreak{}\\{}
~~~~\{"\allowbreak{}bbox\_\allowbreak{}2d"\allowbreak{}: [434,\allowbreak{} 192,\allowbreak{} 476,\allowbreak{} 258],\allowbreak{} "\allowbreak{}label"\allowbreak{}: "\allowbreak{}round pill"\allowbreak{}\},\allowbreak{}\\{}
~~~~\{"\allowbreak{}bbox\_\allowbreak{}2d"\allowbreak{}: [451,\allowbreak{} 256,\allowbreak{} 494,\allowbreak{} 320],\allowbreak{} "\allowbreak{}label"\allowbreak{}: "\allowbreak{}round pill"\allowbreak{}\},\allowbreak{}\\{}
~~~~\{"\allowbreak{}bbox\_\allowbreak{}2d"\allowbreak{}: [453,\allowbreak{} 482,\allowbreak{} 497,\allowbreak{} 551],\allowbreak{} "\allowbreak{}label"\allowbreak{}: "\allowbreak{}round pill"\allowbreak{}\},\allowbreak{}\\{}
~~~~\{"\allowbreak{}bbox\_\allowbreak{}2d"\allowbreak{}: [456,\allowbreak{} 893,\allowbreak{} 480,\allowbreak{} 956],\allowbreak{} "\allowbreak{}label"\allowbreak{}: "\allowbreak{}round pill"\allowbreak{}\},\allowbreak{}\\{}
~~~~\{"\allowbreak{}bbox\_\allowbreak{}2d"\allowbreak{}: [458,\allowbreak{} 815,\allowbreak{} 479,\allowbreak{} 845],\allowbreak{} "\allowbreak{}label"\allowbreak{}: "\allowbreak{}round pill"\allowbreak{}\},\allowbreak{}\\{}
~~~~\{"\allowbreak{}bbox\_\allowbreak{}2d"\allowbreak{}: [472,\allowbreak{} 552,\allowbreak{} 529,\allowbreak{} 637],\allowbreak{} "\allowbreak{}label"\allowbreak{}: "\allowbreak{}round pill"\allowbreak{}\},\allowbreak{}\\{}
~~~~\{"\allowbreak{}bbox\_\allowbreak{}2d"\allowbreak{}: [476,\allowbreak{} 342,\allowbreak{} 533,\allowbreak{} 427],\allowbreak{} "\allowbreak{}label"\allowbreak{}: "\allowbreak{}round pill"\allowbreak{}\},\allowbreak{}\\{}
~~~~\{"\allowbreak{}bbox\_\allowbreak{}2d"\allowbreak{}: [484,\allowbreak{} 179,\allowbreak{} 532,\allowbreak{} 251],\allowbreak{} "\allowbreak{}label"\allowbreak{}: "\allowbreak{}round pill"\allowbreak{}\},\allowbreak{}
\end{exverb}
\exCut{The remaining 3,087 characters of this output are not shown (4,539 in total).}
\begin{center}\includegraphics[width=0.45\linewidth,height=0.22\textheight,keepaspectratio]{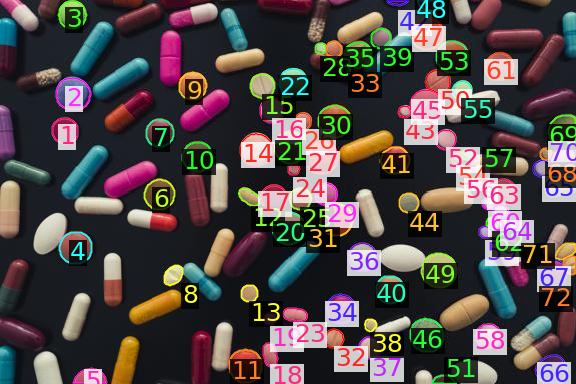}\\[2pt]
{\scriptsize\color{exMute} Image returned with this output.}\end{center}
\begin{exharness}
\exLabel{Harness}{environment update, summarized: new memory entries; new data bindings verbatim}
\begin{exverb}
Memory update: registers 1 new image in memory (tool\_\allowbreak{}generated\_\allowbreak{}image\_\allowbreak{}2,\allowbreak{} from Visual\_\allowbreak{}Grounding\_\allowbreak{}Tool).\\{}
- segment\_\allowbreak{}round\_\allowbreak{}pill: dict (keys: pred\_\allowbreak{}boxes,\allowbreak{} pred\_\allowbreak{}scores,\allowbreak{} pred\_\allowbreak{}masks,\allowbreak{} num\_\allowbreak{}masks,\allowbreak{} orig\_\allowbreak{}img\_\allowbreak{}h,\allowbreak{} orig\_\allowbreak{}img\_\allowbreak{}w)
\end{exverb}
\end{exharness}
\end{exstep}
\begin{exstep}{3}{\toolCode}{}
\exWalk{The agent asks the coding agent to merge the two sets. Both sets are available to the code as data bindings (\texttt{segment\_pill} and \texttt{segment\_round\_pill}), so no coordinates need to be copied by hand. The run completes but reports zero matches and a union of zero, which cannot be right, since both queries detect the same pills.}
\exLabel{Arguments}{}
\begin{exverb}
\{\\{}
~~"\allowbreak{}objective"\allowbreak{}: "\allowbreak{}Match the 134 detected pills against round-pill detections using mask overlap or bounding box intersection to count round pills absent from original pill detections,\allowbreak{} listing locations of missing ones and total union count."\allowbreak{}\\{}
\}
\end{exverb}
\exLabel{Tool output}{}
\begin{exverb}
Coding agent generated and executed the following Python code:\\{}
\textasciigrave{}\textasciigrave{}\textasciigrave{}python\\{}
import numpy as np\\{}
from PIL import Image,\allowbreak{} ImageDraw\\{}
import base64\\{}
import io\\{}
import json\\{}
\mbox{}\\{}
\# Decode masks to numpy arrays for intersection computation\\{}
def decode\_\allowbreak{}mask(mask\_\allowbreak{}str):\\{}
~~~~\# Each mask is a base64-encoded string representing a segmentation mask\\{}
~~~~\# We\textquotesingle{}ll decode it to a numpy array of shape (H,\allowbreak{} W) with 0/\allowbreak{}1 values\\{}
~~~~\# The format is assumed to be a run-length encoded or compressed binary mask\\{}
~~~~\# For simplicity,\allowbreak{} we\textquotesingle{}ll treat it as a base64-encoded PNG-like mask\\{}
~~~~try:\\{}
~~~~~~~~\# Decode base64 string to bytes\\{}
~~~~~~~~mask\_\allowbreak{}bytes =\allowbreak{} base64.b64decode(mask\_\allowbreak{}str)\\{}
~~~~~~~~\# Create a PIL Image from the bytes\\{}
~~~~~~~~img =\allowbreak{} Image.open(io.BytesIO(mask\_\allowbreak{}bytes))\\{}
~~~~~~~~\# Convert to numpy array\\{}
~~~~~~~~mask\_\allowbreak{}array =\allowbreak{} np.array(img)\\{}
~~~~~~~~\# Ensure binary mask (0 or 1)\\{}
~~~~~~~~mask\_\allowbreak{}array =\allowbreak{} (mask\_\allowbreak{}array \textgreater{} 0).astype(np.uint8)\\{}
~~~~~~~~return mask\_\allowbreak{}array\\{}
~~~~except Exception:\\{}
~~~~~~~~\# If decoding fails,\allowbreak{} return None\\{}
~~~~~~~~return None\\{}
\mbox{}\\{}
\# Compute IoU (Intersection over Union) between two masks\\{}
def compute\_\allowbreak{}iou(mask1,\allowbreak{} mask2):\\{}
~~~~if mask1 is None or mask2 is None:\\{}
~~~~~~~~return 0.0\\{}
~~~~\# Ensure masks are binary and same shape\\{}
~~~~mask1 =\allowbreak{} (mask1 \textgreater{} 0).astype(np.uint8)\\{}
~~~~mask2 =\allowbreak{} (mask2 \textgreater{} 0).astype(np.uint8)\\{}
~~~~\# Compute intersection and union\\{}
~~~~intersection =\allowbreak{} np.logical\_\allowbreak{}and(mask1,\allowbreak{} mask2).sum()\\{}
~~~~union =\allowbreak{} np.logical\_\allowbreak{}or(mask1,\allowbreak{} mask2).sum()\\{}
~~~~if union =\allowbreak{}=\allowbreak{} 0:\\{}
~~~~~~~~return 0.0\\{}
~~~~return intersection /\allowbreak{} union\\{}
\mbox{}\\{}
\# Compute IoU between two bounding boxes\\{}
def compute\_\allowbreak{}bbox\_\allowbreak{}iou(box1,\allowbreak{} box2):\\{}
~~~~\# box format: [x1,\allowbreak{} y1,\allowbreak{} x2,\allowbreak{} y2] (top-left,\allowbreak{} bottom-right)\\{}
~~~~x1\_\allowbreak{}1,\allowbreak{} y1\_\allowbreak{}1,\allowbreak{} x2\_\allowbreak{}1,\allowbreak{} y2\_\allowbreak{}1 =\allowbreak{} box1\\{}
~~~~x1\_\allowbreak{}2,\allowbreak{} y1\_\allowbreak{}2,\allowbreak{} x2\_\allowbreak{}2,\allowbreak{} y2\_\allowbreak{}2 =\allowbreak{} box2\\{}
\mbox{}\\{}
~~~~\# Compute intersection\\{}
~~~~x\_\allowbreak{}left =\allowbreak{} max(x1\_\allowbreak{}1,\allowbreak{} x1\_\allowbreak{}2)\\{}
~~~~y\_\allowbreak{}top =\allowbreak{} max(y1\_\allowbreak{}1,\allowbreak{} y1\_\allowbreak{}2)\\{}
~~~~x\_\allowbreak{}right =\allowbreak{} min(x2\_\allowbreak{}1,\allowbreak{} x2\_\allowbreak{}2)\\{}
~~~~y\_\allowbreak{}bottom =\allowbreak{} min(y2\_\allowbreak{}1,\allowbreak{} y2\_\allowbreak{}2)\\{}
\mbox{}\\{}
~~~~if x\_\allowbreak{}right \textless{}=\allowbreak{} x\_\allowbreak{}left or y\_\allowbreak{}bottom \textless{}=\allowbreak{} y\_\allowbreak{}top:\\{}
~~~~~~~~return 0.0\\{}
\mbox{}\\{}
~~~~intersection\_\allowbreak{}area =\allowbreak{} (x\_\allowbreak{}right - x\_\allowbreak{}left) * (y\_\allowbreak{}bottom - y\_\allowbreak{}top)\\{}
~~~~\# Compute union\\{}
~~~~area1 =\allowbreak{} (x2\_\allowbreak{}1 - x1\_\allowbreak{}1) * (y2\_\allowbreak{}1 - y1\_\allowbreak{}1)\\{}
~~~~area2 =\allowbreak{} (x2\_\allowbreak{}2 - x1\_\allowbreak{}2) * (y2\_\allowbreak{}2 - y1\_\allowbreak{}2)\\{}
~~~~union\_\allowbreak{}area =\allowbreak{} area1 + area2 - intersection\_\allowbreak{}area\\{}
\mbox{}\\{}
~~~~return intersection\_\allowbreak{}area /\allowbreak{} union\_\allowbreak{}area if union\_\allowbreak{}area \textgreater{} 0 else 0.0\\{}
\mbox{}\\{}
\# Get mask dimensions from original image\\{}
H,\allowbreak{} W =\allowbreak{} segment\_\allowbreak{}pill["\allowbreak{}orig\_\allowbreak{}img\_\allowbreak{}h"\allowbreak{}],\allowbreak{} segment\_\allowbreak{}pill["\allowbreak{}orig\_\allowbreak{}img\_\allowbreak{}w"\allowbreak{}]\\{}
\mbox{}\\{}
\# Decode all masks for both datasets\\{}
decoded\_\allowbreak{}masks\_\allowbreak{}pill =\allowbreak{} [decode\_\allowbreak{}mask(mask) for mask in segment\_\allowbreak{}pill["\allowbreak{}pred\_\allowbreak{}masks"\allowbreak{}]]\\{}
decoded\_\allowbreak{}masks\_\allowbreak{}round\_\allowbreak{}pill =\allowbreak{} [decode\_\allowbreak{}mask(mask) for mask in segment\_\allowbreak{}round\_\allowbreak{}pill["\allowbreak{}pred\_\allowbreak{}masks"\allowbreak{}]]\\{}
\mbox{}\\{}
\# Initialize lists to store matched and unmatched round pills\\{}
matched\_\allowbreak{}round\_\allowbreak{}pills =\allowbreak{} []\\{}
unmatched\_\allowbreak{}round\_\allowbreak{}pills =\allowbreak{} []\\{}
\mbox{}\\{}
\# For each round pill,\allowbreak{} find the best match among original pills using IoU\\{}
for i,\allowbreak{} round\_\allowbreak{}mask in enumerate(decoded\_\allowbreak{}masks\_\allowbreak{}round\_\allowbreak{}pill):\\{}
~~~~best\_\allowbreak{}iou =\allowbreak{} 0.0\\{}
~~~~best\_\allowbreak{}match\_\allowbreak{}idx =\allowbreak{} -1\\{}
~~~~best\_\allowbreak{}bbox\_\allowbreak{}iou =\allowbreak{} 0.0\\{}
~~~~best\_\allowbreak{}bbox\_\allowbreak{}idx =\allowbreak{} -1\\{}
\mbox{}\\{}
~~~~for j,\allowbreak{} pill\_\allowbreak{}mask in enumerate(decoded\_\allowbreak{}masks\_\allowbreak{}pill):\\{}
~~~~~~~~if pill\_\allowbreak{}mask is not None and round\_\allowbreak{}mask is not None:\\{}
~~~~~~~~~~~~mask\_\allowbreak{}iou =\allowbreak{} compute\_\allowbreak{}iou(pill\_\allowbreak{}mask,\allowbreak{} round\_\allowbreak{}mask)\\{}
~~~~~~~~~~~~bbox\_\allowbreak{}iou =\allowbreak{} compute\_\allowbreak{}bbox\_\allowbreak{}iou(segment\_\allowbreak{}round\_\allowbreak{}pill["\allowbreak{}pred\_\allowbreak{}boxes"\allowbreak{}][i],\allowbreak{} segment\_\allowbreak{}pill["\allowbreak{}pred\_\allowbreak{}boxes"\allowbreak{}][j])\\{}
~~~~~~~~~~~~\# Use a combined score or choose the higher one\\{}
~~~~~~~~~~~~combined\_\allowbreak{}score =\allowbreak{} (mask\_\allowbreak{}iou + bbox\_\allowbreak{}iou) /\allowbreak{} 2.0\\{}
~~~~~~~~~~~~if combined\_\allowbreak{}score \textgreater{} best\_\allowbreak{}iou:\\{}
~~~~~~~~~~~~~~~~best\_\allowbreak{}iou =\allowbreak{} combined\_\allowbreak{}score\\{}
~~~~~~~~~~~~~~~~best\_\allowbreak{}match\_\allowbreak{}idx =\allowbreak{} j\\{}
~~~~~~~~~~~~~~~~best\_\allowbreak{}bbox\_\allowbreak{}iou =\allowbreak{} bbox\_\allowbreak{}iou\\{}
~~~~~~~~~~~~~~~~best\_\allowbreak{}bbox\_\allowbreak{}idx =\allowbreak{} j\\{}
\mbox{}\\{}
~~~~if best\_\allowbreak{}iou \textgreater{} 0.5:  \# Threshold for matching\\{}
~~~~~~~~matched\_\allowbreak{}round\_\allowbreak{}pills.append((i,\allowbreak{} best\_\allowbreak{}match\_\allowbreak{}idx,\allowbreak{} best\_\allowbreak{}iou,\allowbreak{} best\_\allowbreak{}bbox\_\allowbreak{}iou))\\{}
~~~~else:\\{}
~~~~~~~~unmatched\_\allowbreak{}round\_\allowbreak{}pills.append((i,\allowbreak{} segment\_\allowbreak{}round\_\allowbreak{}pill["\allowbreak{}pred\_\allowbreak{}boxes"\allowbreak{}][i],\allowbreak{} segment\_\allowbreak{}round\_\allowbreak{}pill["\allowbreak{}pred\_\allowbreak{}scores"\allowbreak{}][i]))\\{}
\mbox{}\\{}
\# Count total union of detected pills (unique pills from both sets)\\{}
all\_\allowbreak{}masks =\allowbreak{} decoded\_\allowbreak{}masks\_\allowbreak{}pill + decoded\_\allowbreak{}masks\_\allowbreak{}round\_\allowbreak{}pill\\{}
unique\_\allowbreak{}masks =\allowbreak{} []\\{}
for mask in all\_\allowbreak{}masks:\\{}
~~~~if mask is not None:\\{}
~~~~~~~~\# Check if mask is already in unique\_\allowbreak{}masks\\{}
~~~~~~~~if not any(np.array\_\allowbreak{}equal(mask,\allowbreak{} existing\_\allowbreak{}mask) for existing\_\allowbreak{}mask in unique\_\allowbreak{}masks):\\{}
~~~~~~~~~~~~unique\_\allowbreak{}masks.append(mask)\\{}
\mbox{}\\{}
total\_\allowbreak{}union\_\allowbreak{}count =\allowbreak{} len(unique\_\allowbreak{}masks)\\{}
\mbox{}\\{}
\# Output results\\{}
print(f"\allowbreak{}Total round pills detected: \{len(segment\_\allowbreak{}round\_\allowbreak{}pill[\textquotesingle{}pred\_\allowbreak{}masks\textquotesingle{}])\}"\allowbreak{})\\{}
print(f"\allowbreak{}Total original pills detected: \{len(segment\_\allowbreak{}pill[\textquotesingle{}pred\_\allowbreak{}masks\textquotesingle{}])\}"\allowbreak{})\\{}
print(f"\allowbreak{}Matched round pills: \{len(matched\_\allowbreak{}round\_\allowbreak{}pills)\}"\allowbreak{})\\{}
print(f"\allowbreak{}Unmatched round pills: \{len(unmatched\_\allowbreak{}round\_\allowbreak{}pills)\}"\allowbreak{})\\{}
print(f"\allowbreak{}Total union count: \{total\_\allowbreak{}union\_\allowbreak{}count\}"\allowbreak{})\\{}
\mbox{}\\{}
\# List locations of unmatched round pills\\{}
print("\allowbreak{}\\{}
Unmatched round pills (location and score):"\allowbreak{})\\{}
for idx,\allowbreak{} box,\allowbreak{} score in unmatched\_\allowbreak{}round\_\allowbreak{}pills:\\{}
~~~~print(f"\allowbreak{}  Index: \{idx\},\allowbreak{} BBox: \{box\},\allowbreak{} Score: \{score\}"\allowbreak{})\\{}
\mbox{}\\{}
\# Optional: Print matched pairs\\{}
print("\allowbreak{}\\{}
Matched round pills (round\_\allowbreak{}idx,\allowbreak{} original\_\allowbreak{}idx,\allowbreak{} IoU,\allowbreak{} bbox\_\allowbreak{}iou):"\allowbreak{})\\{}
for round\_\allowbreak{}idx,\allowbreak{} orig\_\allowbreak{}idx,\allowbreak{} iou,\allowbreak{} bbox\_\allowbreak{}iou in matched\_\allowbreak{}round\_\allowbreak{}pills:\\{}
~~~~print(f"\allowbreak{}  Round \{round\_\allowbreak{}idx\} -\textgreater{} Original \{orig\_\allowbreak{}idx\},\allowbreak{} IoU: \{iou:.4f\},\allowbreak{} BBox IoU: \{bbox\_\allowbreak{}iou:.4f\}"\allowbreak{})\\{}
\textasciigrave{}\textasciigrave{}\textasciigrave{}\\{}
\mbox{}\\{}
\mbox{}\\{}
\mbox{}\\{}
Execution output:\\{}
Total round pills detected: 72\\{}
Total original pills detected: 134\\{}
Matched round pills: 0\\{}
Unmatched round pills: 72\\{}
Total union count: 0\\{}
\mbox{}\\{}
Unmatched round pills (location and score):\\{}
~~Index: 0,\allowbreak{} BBox: [52.31292724609375,\allowbreak{} 117.94264221191406,\allowbreak{} 78.2960433959961,\allowbreak{} 143.74490356445312],\allowbreak{} Score: 0.921875\\{}
~~Index: 1,\allowbreak{} BBox: [56.460731506347656,\allowbreak{} 77.62562561035156,\allowbreak{} 90.11420440673828,\allowbreak{} 110.55953979492188],\allowbreak{} Score: 0.91015625\\{}
~~Index: 2,\allowbreak{} BBox: [58.77238082885742,\allowbreak{} 1.2981762886047363,\allowbreak{} 86.19441986083984,\allowbreak{} 29.0463809967041],\allowbreak{} Score: 0.9140625\\{}
~~Index: 3,\allowbreak{} BBox: [59.73981857299805,\allowbreak{} 231.58026123046875,\allowbreak{} 91.96317291259766,\allowbreak{} 264.26812744140625],\allowbreak{} Score: 0.921875\\{}
~~Index: 4,\allowbreak{} BBox: [73.73127746582031,\allowbreak{} 368.23492431640625,\allowbreak{} 106.33198547363281,\allowbreak{} 384.2291259765625],\allowbreak{} Score: 0.85546875\\{}
~~Index: 5,\allowbreak{} BBox: [144.1370849609375,\allowbreak{} 179.3444061279297,\allowbreak{} 174.76834106445312,\allowbreak{} 209.95547485351562],\allowbreak{} Score: 0.9296875\\{}
~~Index: 6,\allowbreak{} BBox: [145.93304443359375,\allowbreak{} 119.18267822265625,\allowbreak{} 173.69546508789062,\allowbreak{} 147.4763641357422],\allowbreak{} Score: 0.9140625\\{}
~~Index: 7,\allowbreak{} BBox: [163.64910888671875,\allowbreak{} 265.3359680175781,\allowbreak{} 182.84002685546875,\allowbreak{} 285.3944396972656],\allowbreak{} Score: 0.875\\{}
~~Index: 8,\allowbreak{} BBox: [178.857666015625,\allowbreak{} 72.8042221069336,\allowbreak{} 206.23037719726562,\allowbreak{} 100.41905212402344],\allowbreak{} Score: 0.93359375\\{}
~~Index: 9,\allowbreak{} BBox: [182.19520568847656,\allowbreak{} 142.0789031982422,\allowbreak{} 212.25015258789062,\allowbreak{} 171.2662811279297],\allowbreak{} Score: 0.9375\\{}
~~Index: 10,\allowbreak{} BBox: [229.33837890625,\allowbreak{} 349.857177734375,\allowbreak{} 262.0947265625,\allowbreak{} 383.17645263671875],\allowbreak{} Score: 0.89453125\\{}
~~Index: 11,\allowbreak{} BBox: [238.81997680664062,\allowbreak{} 188.43115234375,\allowbreak{} 259.37640380859375,\allowbreak{} 206.6243896484375],\allowbreak{} Score: 0.69140625\\{}
~~Index: 12,\allowbreak{} BBox: [240.49752807617188,\allowbreak{} 284.8372802734375,\allowbreak{} 258.06903076171875,\allowbreak{} 302.1544494628906],\allowbreak{} Score: 0.89453125\\{}
~~Index: 13,\allowbreak{} BBox: [240.52491760253906,\allowbreak{} 133.4823760986328,\allowbreak{} 273.088134765625,\allowbreak{} 165.65768432617188],\allowbreak{} Score: 0.93359375\\{}
~~Index: 14,\allowbreak{} BBox: [250.11485290527344,\allowbreak{} 73.80632019042969,\allowbreak{} 274.3455810546875,\allowbreak{} 98.89035034179688],\allowbreak{} Score: 0.93359375\\{}
~~Index: 15,\allowbreak{} BBox: [259.9403076171875,\allowbreak{} 98.40367126464844,\allowbreak{} 284.60345458984375,\allowbreak{} 122.7780990600586],\allowbreak{} Score: 0.9296875\\{}
~~Index: 16,\allowbreak{} BBox: [260.7716369628906,\allowbreak{} 185.05271911621094,\allowbreak{} 286.3943176269531,\allowbreak{} 211.45742797851562],\allowbreak{} Score: 0.91015625
\end{exverb}
\exCut{47 lines of execution output not shown here (82 in total).}
\begin{exverb}
~~Index: 64,\allowbreak{} BBox: [534.013671875,\allowbreak{} 160.11773681640625,\allowbreak{} 550.9052734375,\allowbreak{} 176.45880126953125],\allowbreak{} Score: 0.80859375\\{}
~~Index: 65,\allowbreak{} BBox: [535.5726318359375,\allowbreak{} 355.00115966796875,\allowbreak{} 566.8264770507812,\allowbreak{} 384.34881591796875],\allowbreak{} Score: 0.859375\\{}
~~Index: 66,\allowbreak{} BBox: [538.7619018554688,\allowbreak{} 260.182861328125,\allowbreak{} 566.955322265625,\allowbreak{} 287.40594482421875],\allowbreak{} Score: 0.89453125\\{}
~~Index: 67,\allowbreak{} BBox: [540.86328125,\allowbreak{} 147.92442321777344,\allowbreak{} 552.2364501953125,\allowbreak{} 159.67457580566406],\allowbreak{} Score: 0.8203125\\{}
~~Index: 68,\allowbreak{} BBox: [548.8511962890625,\allowbreak{} 114.78163146972656,\allowbreak{} 577.1387939453125,\allowbreak{} 147.60922241210938],\allowbreak{} Score: 0.890625\\{}
~~Index: 69,\allowbreak{} BBox: [549.4615478515625,\allowbreak{} 148.23785400390625,\allowbreak{} 577.478515625,\allowbreak{} 178.4287872314453],\allowbreak{} Score: 0.87890625\\{}
~~Index: 70,\allowbreak{} BBox: [555.9362182617188,\allowbreak{} 242.7733154296875,\allowbreak{} 575.7222290039062,\allowbreak{} 259.68731689453125],\allowbreak{} Score: 0.703125\\{}
~~Index: 71,\allowbreak{} BBox: [563.8709106445312,\allowbreak{} 256.3594665527344,\allowbreak{} 577.0241088867188,\allowbreak{} 274.3338317871094],\allowbreak{} Score: 0.78125\\{}
\mbox{}\\{}
Matched round pills (round\_\allowbreak{}idx,\allowbreak{} original\_\allowbreak{}idx,\allowbreak{} IoU,\allowbreak{} bbox\_\allowbreak{}iou):
\end{exverb}
\end{exstep}
\begin{exstep}{4}{\toolCode}{}
\exWalk{The agent restates the objective more narrowly: compare boxes only, by overlap or centre distance, and do not decode masks. This time the code finds five round detections that match nothing in the first set.}
\exLabel{Arguments}{}
\begin{exverb}
\{\\{}
~~"\allowbreak{}objective"\allowbreak{}: "\allowbreak{}Use ONLY the pred\_\allowbreak{}boxes pixel coordinate arrays for the original 134 pill detections and the 72 round pill detections. Match round boxes to original boxes by intersection-over-union \textgreater{}=\allowbreak{}0.45 or center proximity; report numbered round detections that are missing from original and original detections with major overlaps (duplicates). No mask decoding."\allowbreak{}\\{}
\}
\end{exverb}
\exLabel{Tool output}{}
\begin{exverb}
Coding agent generated and executed the following Python code:\\{}
\textasciigrave{}\textasciigrave{}\textasciigrave{}python\\{}
import numpy as np\\{}
\mbox{}\\{}
\# Extract pred\_\allowbreak{}boxes for original pills and round pills\\{}
original\_\allowbreak{}boxes =\allowbreak{} segment\_\allowbreak{}pill["\allowbreak{}pred\_\allowbreak{}boxes"\allowbreak{}]\\{}
round\_\allowbreak{}boxes =\allowbreak{} segment\_\allowbreak{}round\_\allowbreak{}pill["\allowbreak{}pred\_\allowbreak{}boxes"\allowbreak{}]\\{}
\mbox{}\\{}
\# Convert to numpy arrays for easier computation\\{}
original\_\allowbreak{}boxes =\allowbreak{} np.array(original\_\allowbreak{}boxes)\\{}
round\_\allowbreak{}boxes =\allowbreak{} np.array(round\_\allowbreak{}boxes)\\{}
\mbox{}\\{}
\# Helper function to compute IoU\\{}
def iou(box1,\allowbreak{} box2):\\{}
~~~~\# box format: [x1,\allowbreak{} y1,\allowbreak{} x2,\allowbreak{} y2] (assuming top-left and bottom-right)\\{}
~~~~x1\_\allowbreak{}1,\allowbreak{} y1\_\allowbreak{}1,\allowbreak{} x2\_\allowbreak{}1,\allowbreak{} y2\_\allowbreak{}1 =\allowbreak{} box1\\{}
~~~~x1\_\allowbreak{}2,\allowbreak{} y1\_\allowbreak{}2,\allowbreak{} x2\_\allowbreak{}2,\allowbreak{} y2\_\allowbreak{}2 =\allowbreak{} box2\\{}
\mbox{}\\{}
~~~~\# Compute intersection\\{}
~~~~x\_\allowbreak{}int =\allowbreak{} max(0,\allowbreak{} min(x2\_\allowbreak{}1,\allowbreak{} x2\_\allowbreak{}2) - max(x1\_\allowbreak{}1,\allowbreak{} x1\_\allowbreak{}2))\\{}
~~~~y\_\allowbreak{}int =\allowbreak{} max(0,\allowbreak{} min(y2\_\allowbreak{}1,\allowbreak{} y2\_\allowbreak{}2) - max(y1\_\allowbreak{}1,\allowbreak{} y1\_\allowbreak{}2))\\{}
~~~~intersection =\allowbreak{} x\_\allowbreak{}int * y\_\allowbreak{}int\\{}
\mbox{}\\{}
~~~~\# Compute union\\{}
~~~~area1 =\allowbreak{} (x2\_\allowbreak{}1 - x1\_\allowbreak{}1) * (y2\_\allowbreak{}1 - y1\_\allowbreak{}1)\\{}
~~~~area2 =\allowbreak{} (x2\_\allowbreak{}2 - x1\_\allowbreak{}2) * (y2\_\allowbreak{}2 - y1\_\allowbreak{}2)\\{}
~~~~union =\allowbreak{} area1 + area2 - intersection\\{}
\mbox{}\\{}
~~~~if union =\allowbreak{}=\allowbreak{} 0:\\{}
~~~~~~~~return 0.0\\{}
~~~~return intersection /\allowbreak{} union\\{}
\mbox{}\\{}
\# Helper function to compute center proximity (Euclidean distance between centers)\\{}
def center\_\allowbreak{}proximity(box1,\allowbreak{} box2):\\{}
~~~~x1\_\allowbreak{}1,\allowbreak{} y1\_\allowbreak{}1,\allowbreak{} x2\_\allowbreak{}1,\allowbreak{} y2\_\allowbreak{}1 =\allowbreak{} box1\\{}
~~~~x1\_\allowbreak{}2,\allowbreak{} y1\_\allowbreak{}2,\allowbreak{} x2\_\allowbreak{}2,\allowbreak{} y2\_\allowbreak{}2 =\allowbreak{} box2\\{}
~~~~center1\_\allowbreak{}x,\allowbreak{} center1\_\allowbreak{}y =\allowbreak{} (x1\_\allowbreak{}1 + x2\_\allowbreak{}1) /\allowbreak{} 2,\allowbreak{} (y1\_\allowbreak{}1 + y2\_\allowbreak{}1) /\allowbreak{} 2\\{}
~~~~center2\_\allowbreak{}x,\allowbreak{} center2\_\allowbreak{}y =\allowbreak{} (x1\_\allowbreak{}2 + x2\_\allowbreak{}2) /\allowbreak{} 2,\allowbreak{} (y1\_\allowbreak{}2 + y2\_\allowbreak{}2) /\allowbreak{} 2\\{}
~~~~return np.sqrt((center1\_\allowbreak{}x - center2\_\allowbreak{}x)**2 + (center1\_\allowbreak{}y - center2\_\allowbreak{}y)**2)\\{}
\mbox{}\\{}
\# Threshold for matching: IoU \textgreater{}=\allowbreak{} 0.45 OR center proximity \textless{}=\allowbreak{} 10 pixels (arbitrary,\allowbreak{} but reasonable for visual matching)\\{}
MATCH\_\allowbreak{}THRESHOLD\_\allowbreak{}IOU =\allowbreak{} 0.45\\{}
MATCH\_\allowbreak{}THRESHOLD\_\allowbreak{}CENTER =\allowbreak{} 10.0\\{}
\mbox{}\\{}
\# Initialize lists to track matches and unmatched detections\\{}
matched\_\allowbreak{}round\_\allowbreak{}indices =\allowbreak{} set()\\{}
matched\_\allowbreak{}original\_\allowbreak{}indices =\allowbreak{} set()\\{}
unmatched\_\allowbreak{}round\_\allowbreak{}indices =\allowbreak{} []\\{}
unmatched\_\allowbreak{}original\_\allowbreak{}indices =\allowbreak{} []\\{}
\mbox{}\\{}
\# For each round pill detection,\allowbreak{} try to match with original pill detections\\{}
for round\_\allowbreak{}idx,\allowbreak{} round\_\allowbreak{}box in enumerate(round\_\allowbreak{}boxes):\\{}
~~~~matched =\allowbreak{} False\\{}
~~~~for orig\_\allowbreak{}idx,\allowbreak{} orig\_\allowbreak{}box in enumerate(original\_\allowbreak{}boxes):\\{}
~~~~~~~~iou\_\allowbreak{}val =\allowbreak{} iou(round\_\allowbreak{}box,\allowbreak{} orig\_\allowbreak{}box)\\{}
~~~~~~~~center\_\allowbreak{}dist =\allowbreak{} center\_\allowbreak{}proximity(round\_\allowbreak{}box,\allowbreak{} orig\_\allowbreak{}box)\\{}
\mbox{}\\{}
~~~~~~~~if iou\_\allowbreak{}val \textgreater{}=\allowbreak{} MATCH\_\allowbreak{}THRESHOLD\_\allowbreak{}IOU or center\_\allowbreak{}dist \textless{}=\allowbreak{} MATCH\_\allowbreak{}THRESHOLD\_\allowbreak{}CENTER:\\{}
~~~~~~~~~~~~\# Found a match\\{}
~~~~~~~~~~~~matched =\allowbreak{} True\\{}
~~~~~~~~~~~~matched\_\allowbreak{}round\_\allowbreak{}indices.add(round\_\allowbreak{}idx)\\{}
~~~~~~~~~~~~matched\_\allowbreak{}original\_\allowbreak{}indices.add(orig\_\allowbreak{}idx)\\{}
~~~~~~~~~~~~break\\{}
\mbox{}\\{}
~~~~if not matched:\\{}
~~~~~~~~unmatched\_\allowbreak{}round\_\allowbreak{}indices.append(round\_\allowbreak{}idx)\\{}
\mbox{}\\{}
\# Find original detections that have major overlaps (duplicates) - i.e.,\allowbreak{} matched with multiple round pills\\{}
duplicate\_\allowbreak{}original\_\allowbreak{}indices =\allowbreak{} []\\{}
for orig\_\allowbreak{}idx in matched\_\allowbreak{}original\_\allowbreak{}indices:\\{}
~~~~count =\allowbreak{} 0\\{}
~~~~for round\_\allowbreak{}idx in range(len(round\_\allowbreak{}boxes)):\\{}
~~~~~~~~if round\_\allowbreak{}idx not in matched\_\allowbreak{}round\_\allowbreak{}indices:\\{}
~~~~~~~~~~~~continue\\{}
~~~~~~~~iou\_\allowbreak{}val =\allowbreak{} iou(original\_\allowbreak{}boxes[orig\_\allowbreak{}idx],\allowbreak{} round\_\allowbreak{}boxes[round\_\allowbreak{}idx])\\{}
~~~~~~~~center\_\allowbreak{}dist =\allowbreak{} center\_\allowbreak{}proximity(original\_\allowbreak{}boxes[orig\_\allowbreak{}idx],\allowbreak{} round\_\allowbreak{}boxes[round\_\allowbreak{}idx])\\{}
~~~~~~~~if iou\_\allowbreak{}val \textgreater{}=\allowbreak{} MATCH\_\allowbreak{}THRESHOLD\_\allowbreak{}IOU or center\_\allowbreak{}dist \textless{}=\allowbreak{} MATCH\_\allowbreak{}THRESHOLD\_\allowbreak{}CENTER:\\{}
~~~~~~~~~~~~count +=\allowbreak{} 1\\{}
~~~~if count \textgreater{} 1:\\{}
~~~~~~~~duplicate\_\allowbreak{}original\_\allowbreak{}indices.append(orig\_\allowbreak{}idx)\\{}
\mbox{}\\{}
\# Output results\\{}
print("\allowbreak{}Round detections missing from original:"\allowbreak{},\allowbreak{} unmatched\_\allowbreak{}round\_\allowbreak{}indices)\\{}
print("\allowbreak{}Original detections with major overlaps (duplicates):"\allowbreak{},\allowbreak{} duplicate\_\allowbreak{}original\_\allowbreak{}indices)\\{}
\textasciigrave{}\textasciigrave{}\textasciigrave{}\\{}
\mbox{}\\{}
\mbox{}\\{}
\mbox{}\\{}
Execution output:\\{}
Round detections missing from original: [23,\allowbreak{} 25,\allowbreak{} 26,\allowbreak{} 40,\allowbreak{} 41]\\{}
Original detections with major overlaps (duplicates): []
\end{exverb}
\end{exstep}
\begin{exstep}{}{\exTool{exInk}{Final answer}}{}
\exWalk{The agent adds the five missing pills to the 134 from Step~1 and answers 139, the exact count.}
\exLabel{Model}{}
\begin{exverb}
\textless{}answer\textgreater{}139\textless{}/\allowbreak{}answer\textgreater{}
\end{exverb}
\end{exstep}

%% file: text/appendix_examples_steps/voc331_header.tex
\exOverview{\textbf{Task.} Detect every object of the twenty PascalVOC categories in a photograph of fourteen men posing around a dinner table, and return labelled boxes. 

\textbf{Why it is hard.} The people overlap heavily, some are visible only as a face between two others, and the table is covered with bottles and plates. 

\textbf{How the agent solves it.} After a first pass over the four categories present, it zooms on the crowded table and issues narrower queries, ending with a referring expression for one partly hidden man. The final answer matches 22 of the 23 ground-truth objects.}
\exLabel{Input images}{}
\begin{center}\includegraphics[width=0.55\linewidth,height=0.28\textheight,keepaspectratio]{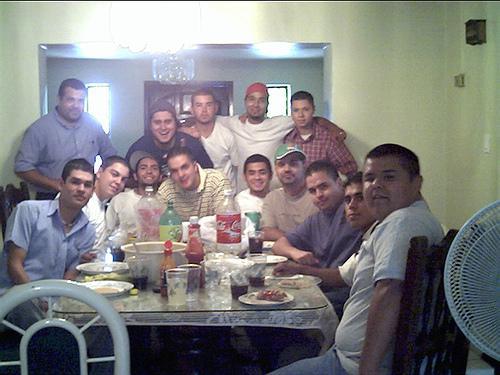}\\[2pt]
{\scriptsize\color{exMute} input\_image\_1}\end{center}
\exLabel{Question}{as sent to the model}
\begin{exverb}
Question: Locate every instance that belongs to the following categories: \textquotesingle{}aeroplane,\allowbreak{} bicycle,\allowbreak{} bird,\allowbreak{} boat,\allowbreak{} bottle,\allowbreak{} bus,\allowbreak{} car,\allowbreak{} cat,\allowbreak{} chair,\allowbreak{} cow,\allowbreak{} diningtable,\allowbreak{} dog,\allowbreak{} horse,\allowbreak{} motorbike,\allowbreak{} person,\allowbreak{} pottedplant,\allowbreak{} sheep,\allowbreak{} sofa,\allowbreak{} train,\allowbreak{} tvmonitor\textquotesingle{}. Report the bounding boxes in JSON format as a list of objects,\allowbreak{} one object per detected instance,\allowbreak{} where each object has a "\allowbreak{}bbox\_\allowbreak{}2d"\allowbreak{} key with [x1,\allowbreak{} y1,\allowbreak{} x2,\allowbreak{} y2] and a "\allowbreak{}label"\allowbreak{} key with the category name exactly as written above. Coordinates must be normalised on a 0-1000 scale,\allowbreak{} with (0,\allowbreak{}0) as the top-left corner and (1000,\allowbreak{}1000) as the bottom-right corner of the image. For example: [\{"\allowbreak{}bbox\_\allowbreak{}2d"\allowbreak{}: [x1,\allowbreak{} y1,\allowbreak{} x2,\allowbreak{} y2],\allowbreak{} "\allowbreak{}label"\allowbreak{}: "\allowbreak{}aeroplane"\allowbreak{}\},\allowbreak{} \{"\allowbreak{}bbox\_\allowbreak{}2d"\allowbreak{}: [x1,\allowbreak{} y1,\allowbreak{} x2,\allowbreak{} y2],\allowbreak{} "\allowbreak{}label"\allowbreak{}: "\allowbreak{}bicycle"\allowbreak{}\}]. If none of the listed categories is present,\allowbreak{} output an empty list [].
\end{exverb}
\exLabel{Without tools}{GPT-6 Sol's full response to the same question}
\begin{exverb}
[\\{}
~~\{"\allowbreak{}bbox\_\allowbreak{}2d"\allowbreak{}:[28,\allowbreak{}205,\allowbreak{}238,\allowbreak{}565],\allowbreak{}"\allowbreak{}label"\allowbreak{}:"\allowbreak{}person"\allowbreak{}\},\allowbreak{}\\{}
~~\{"\allowbreak{}bbox\_\allowbreak{}2d"\allowbreak{}:[4,\allowbreak{}421,\allowbreak{}208,\allowbreak{}757],\allowbreak{}"\allowbreak{}label"\allowbreak{}:"\allowbreak{}person"\allowbreak{}\},\allowbreak{}\\{}
~~\{"\allowbreak{}bbox\_\allowbreak{}2d"\allowbreak{}:[162,\allowbreak{}413,\allowbreak{}282,\allowbreak{}651],\allowbreak{}"\allowbreak{}label"\allowbreak{}:"\allowbreak{}person"\allowbreak{}\},\allowbreak{}\\{}
~~\{"\allowbreak{}bbox\_\allowbreak{}2d"\allowbreak{}:[218,\allowbreak{}397,\allowbreak{}330,\allowbreak{}637],\allowbreak{}"\allowbreak{}label"\allowbreak{}:"\allowbreak{}person"\allowbreak{}\},\allowbreak{}\\{}
~~\{"\allowbreak{}bbox\_\allowbreak{}2d"\allowbreak{}:[250,\allowbreak{}269,\allowbreak{}376,\allowbreak{}520],\allowbreak{}"\allowbreak{}label"\allowbreak{}:"\allowbreak{}person"\allowbreak{}\},\allowbreak{}\\{}
~~\{"\allowbreak{}bbox\_\allowbreak{}2d"\allowbreak{}:[348,\allowbreak{}232,\allowbreak{}464,\allowbreak{}456],\allowbreak{}"\allowbreak{}label"\allowbreak{}:"\allowbreak{}person"\allowbreak{}\},\allowbreak{}\\{}
~~\{"\allowbreak{}bbox\_\allowbreak{}2d"\allowbreak{}:[438,\allowbreak{}216,\allowbreak{}588,\allowbreak{}501],\allowbreak{}"\allowbreak{}label"\allowbreak{}:"\allowbreak{}person"\allowbreak{}\},\allowbreak{}\\{}
~~\{"\allowbreak{}bbox\_\allowbreak{}2d"\allowbreak{}:[552,\allowbreak{}243,\allowbreak{}714,\allowbreak{}501],\allowbreak{}"\allowbreak{}label"\allowbreak{}:"\allowbreak{}person"\allowbreak{}\},\allowbreak{}\\{}
~~\{"\allowbreak{}bbox\_\allowbreak{}2d"\allowbreak{}:[314,\allowbreak{}381,\allowbreak{}464,\allowbreak{}635],\allowbreak{}"\allowbreak{}label"\allowbreak{}:"\allowbreak{}person"\allowbreak{}\},\allowbreak{}\\{}
~~\{"\allowbreak{}bbox\_\allowbreak{}2d"\allowbreak{}:[438,\allowbreak{}408,\allowbreak{}564,\allowbreak{}627],\allowbreak{}"\allowbreak{}label"\allowbreak{}:"\allowbreak{}person"\allowbreak{}\},\allowbreak{}\\{}
~~\{"\allowbreak{}bbox\_\allowbreak{}2d"\allowbreak{}:[508,\allowbreak{}387,\allowbreak{}636,\allowbreak{}680],\allowbreak{}"\allowbreak{}label"\allowbreak{}:"\allowbreak{}person"\allowbreak{}\},\allowbreak{}\\{}
~~\{"\allowbreak{}bbox\_\allowbreak{}2d"\allowbreak{}:[538,\allowbreak{}432,\allowbreak{}738,\allowbreak{}720],\allowbreak{}"\allowbreak{}label"\allowbreak{}:"\allowbreak{}person"\allowbreak{}\},\allowbreak{}\\{}
~~\{"\allowbreak{}bbox\_\allowbreak{}2d"\allowbreak{}:[678,\allowbreak{}451,\allowbreak{}758,\allowbreak{}688],\allowbreak{}"\allowbreak{}label"\allowbreak{}:"\allowbreak{}person"\allowbreak{}\},\allowbreak{}\\{}
~~\{"\allowbreak{}bbox\_\allowbreak{}2d"\allowbreak{}:[524,\allowbreak{}379,\allowbreak{}892,\allowbreak{}997],\allowbreak{}"\allowbreak{}label"\allowbreak{}:"\allowbreak{}person"\allowbreak{}\},\allowbreak{}\\{}
~~\{"\allowbreak{}bbox\_\allowbreak{}2d"\allowbreak{}:[138,\allowbreak{}632,\allowbreak{}698,\allowbreak{}997],\allowbreak{}"\allowbreak{}label"\allowbreak{}:"\allowbreak{}diningtable"\allowbreak{}\},\allowbreak{}\\{}
~~\{"\allowbreak{}bbox\_\allowbreak{}2d"\allowbreak{}:[0,\allowbreak{}741,\allowbreak{}272,\allowbreak{}997],\allowbreak{}"\allowbreak{}label"\allowbreak{}:"\allowbreak{}chair"\allowbreak{}\},\allowbreak{}\\{}
~~\{"\allowbreak{}bbox\_\allowbreak{}2d"\allowbreak{}:[810,\allowbreak{}613,\allowbreak{}908,\allowbreak{}997],\allowbreak{}"\allowbreak{}label"\allowbreak{}:"\allowbreak{}chair"\allowbreak{}\},\allowbreak{}\\{}
~~\{"\allowbreak{}bbox\_\allowbreak{}2d"\allowbreak{}:[264,\allowbreak{}501,\allowbreak{}318,\allowbreak{}643],\allowbreak{}"\allowbreak{}label"\allowbreak{}:"\allowbreak{}bottle"\allowbreak{}\},\allowbreak{}\\{}
~~\{"\allowbreak{}bbox\_\allowbreak{}2d"\allowbreak{}:[318,\allowbreak{}515,\allowbreak{}374,\allowbreak{}685],\allowbreak{}"\allowbreak{}label"\allowbreak{}:"\allowbreak{}bottle"\allowbreak{}\},\allowbreak{}\\{}
~~\{"\allowbreak{}bbox\_\allowbreak{}2d"\allowbreak{}:[366,\allowbreak{}560,\allowbreak{}412,\allowbreak{}685],\allowbreak{}"\allowbreak{}label"\allowbreak{}:"\allowbreak{}bottle"\allowbreak{}\},\allowbreak{}\\{}
~~\{"\allowbreak{}bbox\_\allowbreak{}2d"\allowbreak{}:[428,\allowbreak{}507,\allowbreak{}484,\allowbreak{}677],\allowbreak{}"\allowbreak{}label"\allowbreak{}:"\allowbreak{}bottle"\allowbreak{}\}\\{}
]
\end{exverb}

%% file: text/appendix_examples_steps/voc331.tex
\begin{exharness}
\exLabel{Harness}{environment update, shown in full}
\begin{exverb}
Environment Update (not a new question):\\{}
Currently available image IDs for tool calls:\\{}
- input\_\allowbreak{}image\_\allowbreak{}1: user\_\allowbreak{}input\\{}
When a tool needs image input,\allowbreak{} pass one of these IDs via the *\_\allowbreak{}image\_\allowbreak{}id field.\\{}
Do NOT invent image file names or paths.
\end{exverb}
\end{exharness}
\begin{exstep}{1}{\toolVG}{}
\exWalk{A first pass on ``person'' returns 13 people.}
\exLabel{Arguments}{}
\begin{exverb}
\{\\{}
~~"\allowbreak{}query"\allowbreak{}: "\allowbreak{}person"\allowbreak{},\allowbreak{}\\{}
~~"\allowbreak{}image\_\allowbreak{}id"\allowbreak{}: "\allowbreak{}input\_\allowbreak{}image\_\allowbreak{}1"\allowbreak{}\\{}
\}
\end{exverb}
\exLabel{Tool output}{}
\begin{exverb}
Visual\_\allowbreak{}Grounding\_\allowbreak{}Tool grounded 13 objects for the prompt \textquotesingle{}person\textquotesingle{}. Their corresponding bounding boxes (on a 0-1000 scale,\allowbreak{} [x1,\allowbreak{} y1,\allowbreak{} x2,\allowbreak{} y2]) are: \textasciigrave{}\textasciigrave{}\textasciigrave{}json\\{}
[\\{}
~~~~\{"\allowbreak{}bbox\_\allowbreak{}2d"\allowbreak{}: [0,\allowbreak{} 422,\allowbreak{} 191,\allowbreak{} 871],\allowbreak{} "\allowbreak{}label"\allowbreak{}: "\allowbreak{}person"\allowbreak{}\},\allowbreak{}\\{}
~~~~\{"\allowbreak{}bbox\_\allowbreak{}2d"\allowbreak{}: [29,\allowbreak{} 208,\allowbreak{} 234,\allowbreak{} 533],\allowbreak{} "\allowbreak{}label"\allowbreak{}: "\allowbreak{}person"\allowbreak{}\},\allowbreak{}\\{}
~~~~\{"\allowbreak{}bbox\_\allowbreak{}2d"\allowbreak{}: [165,\allowbreak{} 414,\allowbreak{} 260,\allowbreak{} 667],\allowbreak{} "\allowbreak{}label"\allowbreak{}: "\allowbreak{}person"\allowbreak{}\},\allowbreak{}\\{}
~~~~\{"\allowbreak{}bbox\_\allowbreak{}2d"\allowbreak{}: [211,\allowbreak{} 402,\allowbreak{} 324,\allowbreak{} 636],\allowbreak{} "\allowbreak{}label"\allowbreak{}: "\allowbreak{}person"\allowbreak{}\},\allowbreak{}\\{}
~~~~\{"\allowbreak{}bbox\_\allowbreak{}2d"\allowbreak{}: [251,\allowbreak{} 272,\allowbreak{} 426,\allowbreak{} 480],\allowbreak{} "\allowbreak{}label"\allowbreak{}: "\allowbreak{}person"\allowbreak{}\},\allowbreak{}\\{}
~~~~\{"\allowbreak{}bbox\_\allowbreak{}2d"\allowbreak{}: [313,\allowbreak{} 389,\allowbreak{} 462,\allowbreak{} 589],\allowbreak{} "\allowbreak{}label"\allowbreak{}: "\allowbreak{}person"\allowbreak{}\},\allowbreak{}\\{}
~~~~\{"\allowbreak{}bbox\_\allowbreak{}2d"\allowbreak{}: [381,\allowbreak{} 236,\allowbreak{} 478,\allowbreak{} 505],\allowbreak{} "\allowbreak{}label"\allowbreak{}: "\allowbreak{}person"\allowbreak{}\},\allowbreak{}\\{}
~~~~\{"\allowbreak{}bbox\_\allowbreak{}2d"\allowbreak{}: [428,\allowbreak{} 219,\allowbreak{} 695,\allowbreak{} 512],\allowbreak{} "\allowbreak{}label"\allowbreak{}: "\allowbreak{}person"\allowbreak{}\},\allowbreak{}\\{}
~~~~\{"\allowbreak{}bbox\_\allowbreak{}2d"\allowbreak{}: [470,\allowbreak{} 410,\allowbreak{} 545,\allowbreak{} 574],\allowbreak{} "\allowbreak{}label"\allowbreak{}: "\allowbreak{}person"\allowbreak{}\},\allowbreak{}\\{}
~~~~\{"\allowbreak{}bbox\_\allowbreak{}2d"\allowbreak{}: [521,\allowbreak{} 384,\allowbreak{} 635,\allowbreak{} 642],\allowbreak{} "\allowbreak{}label"\allowbreak{}: "\allowbreak{}person"\allowbreak{}\},\allowbreak{}\\{}
~~~~\{"\allowbreak{}bbox\_\allowbreak{}2d"\allowbreak{}: [536,\allowbreak{} 382,\allowbreak{} 888,\allowbreak{} 1000],\allowbreak{} "\allowbreak{}label"\allowbreak{}: "\allowbreak{}person"\allowbreak{}\},\allowbreak{}\\{}
~~~~\{"\allowbreak{}bbox\_\allowbreak{}2d"\allowbreak{}: [543,\allowbreak{} 434,\allowbreak{} 727,\allowbreak{} 846],\allowbreak{} "\allowbreak{}label"\allowbreak{}: "\allowbreak{}person"\allowbreak{}\},\allowbreak{}\\{}
~~~~\{"\allowbreak{}bbox\_\allowbreak{}2d"\allowbreak{}: [569,\allowbreak{} 240,\allowbreak{} 717,\allowbreak{} 470],\allowbreak{} "\allowbreak{}label"\allowbreak{}: "\allowbreak{}person"\allowbreak{}\}\\{}
]\textasciigrave{}\textasciigrave{}\textasciigrave{}. The object numbers correspond to the labeled masks rendered in the following image,\allowbreak{} which you can refer to as tool\_\allowbreak{}generated\_\allowbreak{}image\_\allowbreak{}1.
\end{exverb}
\begin{center}\includegraphics[width=0.45\linewidth,height=0.22\textheight,keepaspectratio]{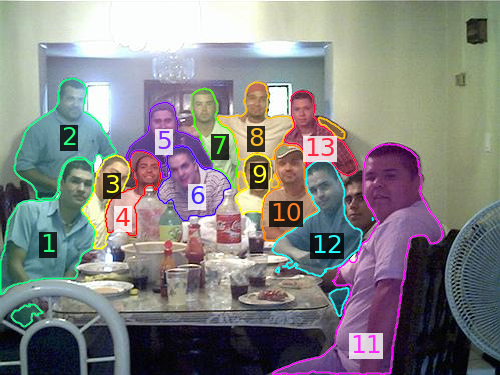}\\[2pt]
{\scriptsize\color{exMute} Image returned with this output.}\end{center}
\begin{exharness}
\exLabel{Harness}{environment update, summarized: new memory entries; new data bindings verbatim}
\begin{exverb}
Memory update: registers 1 new image in memory (tool\_\allowbreak{}generated\_\allowbreak{}image\_\allowbreak{}1,\allowbreak{} from Visual\_\allowbreak{}Grounding\_\allowbreak{}Tool).\\{}
Available data bindings (auto-injected into Python\_\allowbreak{}Coding\_\allowbreak{}Agent\_\allowbreak{}Tool):\\{}
- segment\_\allowbreak{}person: dict (keys: pred\_\allowbreak{}boxes,\allowbreak{} pred\_\allowbreak{}scores,\allowbreak{} pred\_\allowbreak{}masks,\allowbreak{} num\_\allowbreak{}masks,\allowbreak{} orig\_\allowbreak{}img\_\allowbreak{}h,\allowbreak{} orig\_\allowbreak{}img\_\allowbreak{}w)
\end{exverb}
\end{exharness}
\end{exstep}
\begin{exstep}{2}{\toolVG}{}
\exWalk{Grounding ``dining table'' returns the table.}
\exLabel{Arguments}{}
\begin{exverb}
\{\\{}
~~"\allowbreak{}query"\allowbreak{}: "\allowbreak{}dining table"\allowbreak{},\allowbreak{}\\{}
~~"\allowbreak{}image\_\allowbreak{}id"\allowbreak{}: "\allowbreak{}input\_\allowbreak{}image\_\allowbreak{}1"\allowbreak{}\\{}
\}
\end{exverb}
\exLabel{Tool output}{}
\begin{exverb}
Visual\_\allowbreak{}Grounding\_\allowbreak{}Tool grounded 1 objects for the prompt \textquotesingle{}dining table\textquotesingle{}. Their corresponding bounding boxes (on a 0-1000 scale,\allowbreak{} [x1,\allowbreak{} y1,\allowbreak{} x2,\allowbreak{} y2]) are: \textasciigrave{}\textasciigrave{}\textasciigrave{}json\\{}
[\\{}
~~~~\{"\allowbreak{}bbox\_\allowbreak{}2d"\allowbreak{}: [108,\allowbreak{} 722,\allowbreak{} 677,\allowbreak{} 942],\allowbreak{} "\allowbreak{}label"\allowbreak{}: "\allowbreak{}dining table"\allowbreak{}\}\\{}
]\textasciigrave{}\textasciigrave{}\textasciigrave{}. The object numbers correspond to the labeled masks rendered in the following image,\allowbreak{} which you can refer to as tool\_\allowbreak{}generated\_\allowbreak{}image\_\allowbreak{}2.
\end{exverb}
\begin{center}\includegraphics[width=0.45\linewidth,height=0.22\textheight,keepaspectratio]{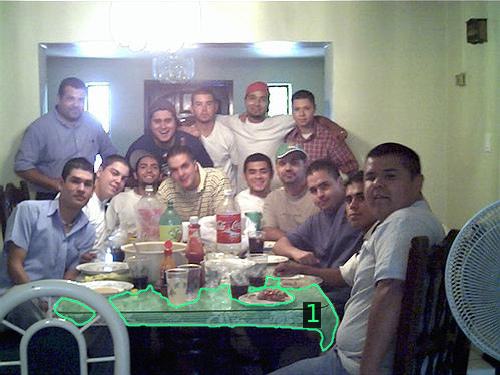}\\[2pt]
{\scriptsize\color{exMute} Image returned with this output.}\end{center}
\begin{exharness}
\exLabel{Harness}{environment update, summarized: new memory entries; new data bindings verbatim}
\begin{exverb}
Memory update: registers 1 new image in memory (tool\_\allowbreak{}generated\_\allowbreak{}image\_\allowbreak{}2,\allowbreak{} from Visual\_\allowbreak{}Grounding\_\allowbreak{}Tool).\\{}
- segment\_\allowbreak{}dining\_\allowbreak{}table: dict (keys: pred\_\allowbreak{}boxes,\allowbreak{} pred\_\allowbreak{}scores,\allowbreak{} pred\_\allowbreak{}masks,\allowbreak{} num\_\allowbreak{}masks,\allowbreak{} orig\_\allowbreak{}img\_\allowbreak{}h,\allowbreak{} orig\_\allowbreak{}img\_\allowbreak{}w)
\end{exverb}
\end{exharness}
\end{exstep}
\begin{exstep}{3}{\toolVG}{}
\exWalk{Grounding ``bottle'' returns five bottles on the table.}
\exLabel{Arguments}{}
\begin{exverb}
\{\\{}
~~"\allowbreak{}query"\allowbreak{}: "\allowbreak{}bottle"\allowbreak{},\allowbreak{}\\{}
~~"\allowbreak{}image\_\allowbreak{}id"\allowbreak{}: "\allowbreak{}input\_\allowbreak{}image\_\allowbreak{}1"\allowbreak{}\\{}
\}
\end{exverb}
\exLabel{Tool output}{}
\begin{exverb}
Visual\_\allowbreak{}Grounding\_\allowbreak{}Tool grounded 5 objects for the prompt \textquotesingle{}bottle\textquotesingle{}. Their corresponding bounding boxes (on a 0-1000 scale,\allowbreak{} [x1,\allowbreak{} y1,\allowbreak{} x2,\allowbreak{} y2]) are: \textasciigrave{}\textasciigrave{}\textasciigrave{}json\\{}
[\\{}
~~~~\{"\allowbreak{}bbox\_\allowbreak{}2d"\allowbreak{}: [271,\allowbreak{} 495,\allowbreak{} 327,\allowbreak{} 642],\allowbreak{} "\allowbreak{}label"\allowbreak{}: "\allowbreak{}bottle"\allowbreak{}\},\allowbreak{}\\{}
~~~~\{"\allowbreak{}bbox\_\allowbreak{}2d"\allowbreak{}: [318,\allowbreak{} 533,\allowbreak{} 364,\allowbreak{} 645],\allowbreak{} "\allowbreak{}label"\allowbreak{}: "\allowbreak{}bottle"\allowbreak{}\},\allowbreak{}\\{}
~~~~\{"\allowbreak{}bbox\_\allowbreak{}2d"\allowbreak{}: [319,\allowbreak{} 640,\allowbreak{} 354,\allowbreak{} 794],\allowbreak{} "\allowbreak{}label"\allowbreak{}: "\allowbreak{}bottle"\allowbreak{}\},\allowbreak{}\\{}
~~~~\{"\allowbreak{}bbox\_\allowbreak{}2d"\allowbreak{}: [369,\allowbreak{} 576,\allowbreak{} 409,\allowbreak{} 705],\allowbreak{} "\allowbreak{}label"\allowbreak{}: "\allowbreak{}bottle"\allowbreak{}\},\allowbreak{}\\{}
~~~~\{"\allowbreak{}bbox\_\allowbreak{}2d"\allowbreak{}: [431,\allowbreak{} 504,\allowbreak{} 483,\allowbreak{} 681],\allowbreak{} "\allowbreak{}label"\allowbreak{}: "\allowbreak{}bottle"\allowbreak{}\}\\{}
]\textasciigrave{}\textasciigrave{}\textasciigrave{}. The object numbers correspond to the labeled masks rendered in the following image,\allowbreak{} which you can refer to as tool\_\allowbreak{}generated\_\allowbreak{}image\_\allowbreak{}3.
\end{exverb}
\begin{center}\includegraphics[width=0.45\linewidth,height=0.22\textheight,keepaspectratio]{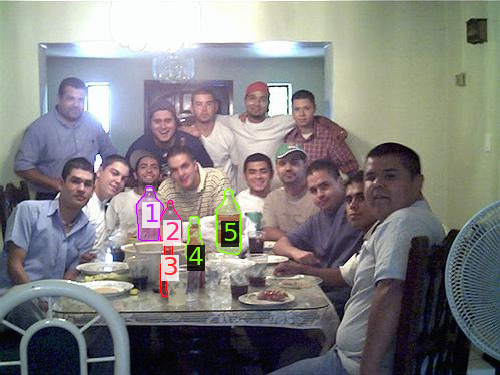}\\[2pt]
{\scriptsize\color{exMute} Image returned with this output.}\end{center}
\begin{exharness}
\exLabel{Harness}{environment update, summarized: new memory entries; new data bindings verbatim}
\begin{exverb}
Memory update: registers 1 new image in memory (tool\_\allowbreak{}generated\_\allowbreak{}image\_\allowbreak{}3,\allowbreak{} from Visual\_\allowbreak{}Grounding\_\allowbreak{}Tool).\\{}
- segment\_\allowbreak{}bottle: dict (keys: pred\_\allowbreak{}boxes,\allowbreak{} pred\_\allowbreak{}scores,\allowbreak{} pred\_\allowbreak{}masks,\allowbreak{} num\_\allowbreak{}masks,\allowbreak{} orig\_\allowbreak{}img\_\allowbreak{}h,\allowbreak{} orig\_\allowbreak{}img\_\allowbreak{}w)
\end{exverb}
\end{exharness}
\end{exstep}
\begin{exstep}{4}{\toolVG}{}
\exWalk{Grounding ``chair'' returns three chairs.}
\exLabel{Arguments}{}
\begin{exverb}
\{\\{}
~~"\allowbreak{}query"\allowbreak{}: "\allowbreak{}chair"\allowbreak{},\allowbreak{}\\{}
~~"\allowbreak{}image\_\allowbreak{}id"\allowbreak{}: "\allowbreak{}input\_\allowbreak{}image\_\allowbreak{}1"\allowbreak{}\\{}
\}
\end{exverb}
\exLabel{Tool output}{}
\begin{exverb}
Visual\_\allowbreak{}Grounding\_\allowbreak{}Tool grounded 3 objects for the prompt \textquotesingle{}chair\textquotesingle{}. Their corresponding bounding boxes (on a 0-1000 scale,\allowbreak{} [x1,\allowbreak{} y1,\allowbreak{} x2,\allowbreak{} y2]) are: \textasciigrave{}\textasciigrave{}\textasciigrave{}json\\{}
[\\{}
~~~~\{"\allowbreak{}bbox\_\allowbreak{}2d"\allowbreak{}: [0,\allowbreak{} 745,\allowbreak{} 260,\allowbreak{} 999],\allowbreak{} "\allowbreak{}label"\allowbreak{}: "\allowbreak{}chair"\allowbreak{}\},\allowbreak{}\\{}
~~~~\{"\allowbreak{}bbox\_\allowbreak{}2d"\allowbreak{}: [0,\allowbreak{} 485,\allowbreak{} 61,\allowbreak{} 635],\allowbreak{} "\allowbreak{}label"\allowbreak{}: "\allowbreak{}chair"\allowbreak{}\},\allowbreak{}\\{}
~~~~\{"\allowbreak{}bbox\_\allowbreak{}2d"\allowbreak{}: [785,\allowbreak{} 612,\allowbreak{} 939,\allowbreak{} 999],\allowbreak{} "\allowbreak{}label"\allowbreak{}: "\allowbreak{}chair"\allowbreak{}\}\\{}
]\textasciigrave{}\textasciigrave{}\textasciigrave{}. The object numbers correspond to the labeled masks rendered in the following image,\allowbreak{} which you can refer to as tool\_\allowbreak{}generated\_\allowbreak{}image\_\allowbreak{}4.
\end{exverb}
\begin{center}\includegraphics[width=0.45\linewidth,height=0.22\textheight,keepaspectratio]{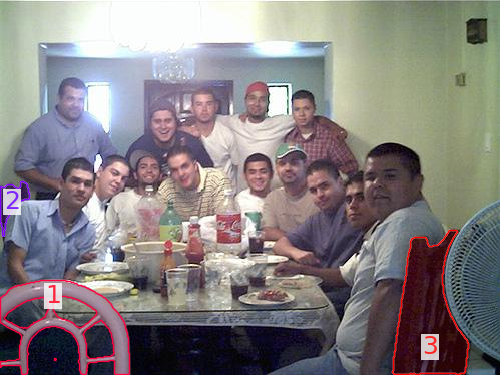}\\[2pt]
{\scriptsize\color{exMute} Image returned with this output.}\end{center}
\begin{exharness}
\exLabel{Harness}{environment update, summarized: new memory entries; new data bindings verbatim}
\begin{exverb}
Memory update: registers 1 new image in memory (tool\_\allowbreak{}generated\_\allowbreak{}image\_\allowbreak{}4,\allowbreak{} from Visual\_\allowbreak{}Grounding\_\allowbreak{}Tool).\\{}
- segment\_\allowbreak{}chair: dict (keys: pred\_\allowbreak{}boxes,\allowbreak{} pred\_\allowbreak{}scores,\allowbreak{} pred\_\allowbreak{}masks,\allowbreak{} num\_\allowbreak{}masks,\allowbreak{} orig\_\allowbreak{}img\_\allowbreak{}h,\allowbreak{} orig\_\allowbreak{}img\_\allowbreak{}w)
\end{exverb}
\end{exharness}
\end{exstep}
\begin{exstep}{5}{\toolZoom}{}
\exWalk{A zoom on the crowded right half of the table, where people overlap most.}
\exLabel{Arguments}{}
\begin{exverb}
\{\\{}
~~"\allowbreak{}image\_\allowbreak{}id"\allowbreak{}: "\allowbreak{}input\_\allowbreak{}image\_\allowbreak{}1"\allowbreak{},\allowbreak{}\\{}
~~"\allowbreak{}coords"\allowbreak{}: [450,\allowbreak{} 350,\allowbreak{} 900,\allowbreak{} 810],\allowbreak{}\\{}
~~"\allowbreak{}padding"\allowbreak{}: 50\\{}
\}
\end{exverb}
\exLabel{Tool output}{}
\begin{exverb}
Tool \textasciigrave{}Zoom\_\allowbreak{}In\_\allowbreak{}Tool\textasciigrave{} (tool\_\allowbreak{}call\_\allowbreak{}id=\allowbreak{}call\_\allowbreak{}GCCULV9QqGvr09iIPPIpWTM2) returned:\\{}
Zoomed in image rendered below; refer to it as tool\_\allowbreak{}generated\_\allowbreak{}image\_\allowbreak{}5.
\end{exverb}
\begin{center}\includegraphics[width=0.45\linewidth,height=0.22\textheight,keepaspectratio]{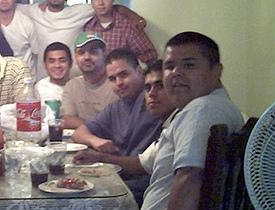}\\[2pt]
{\scriptsize\color{exMute} Image returned with this output.}\end{center}
\begin{exharness}
\exLabel{Harness}{environment update, summarized: new memory entries; new data bindings verbatim}
\begin{exverb}
Memory update: registers 1 new image in memory (tool\_\allowbreak{}generated\_\allowbreak{}image\_\allowbreak{}5,\allowbreak{} from Zoom\_\allowbreak{}In\_\allowbreak{}Tool).
\end{exverb}
\end{exharness}
\end{exstep}
\begin{exstep}{6}{\toolVG}{}
\exWalk{The agent re-grounds the table with the shorter query ``table''.}
\exLabel{Arguments}{}
\begin{exverb}
\{\\{}
~~"\allowbreak{}query"\allowbreak{}: "\allowbreak{}table"\allowbreak{},\allowbreak{}\\{}
~~"\allowbreak{}image\_\allowbreak{}id"\allowbreak{}: "\allowbreak{}input\_\allowbreak{}image\_\allowbreak{}1"\allowbreak{}\\{}
\}
\end{exverb}
\exLabel{Tool output}{}
\begin{exverb}
Visual\_\allowbreak{}Grounding\_\allowbreak{}Tool grounded 1 objects for the prompt \textquotesingle{}table\textquotesingle{}. Their corresponding bounding boxes (on a 0-1000 scale,\allowbreak{} [x1,\allowbreak{} y1,\allowbreak{} x2,\allowbreak{} y2]) are: \textasciigrave{}\textasciigrave{}\textasciigrave{}json\\{}
[\\{}
~~~~\{"\allowbreak{}bbox\_\allowbreak{}2d"\allowbreak{}: [109,\allowbreak{} 741,\allowbreak{} 677,\allowbreak{} 940],\allowbreak{} "\allowbreak{}label"\allowbreak{}: "\allowbreak{}table"\allowbreak{}\}\\{}
]\textasciigrave{}\textasciigrave{}\textasciigrave{}. The object numbers correspond to the labeled masks rendered in the following image,\allowbreak{} which you can refer to as tool\_\allowbreak{}generated\_\allowbreak{}image\_\allowbreak{}6.
\end{exverb}
\begin{center}\includegraphics[width=0.45\linewidth,height=0.22\textheight,keepaspectratio]{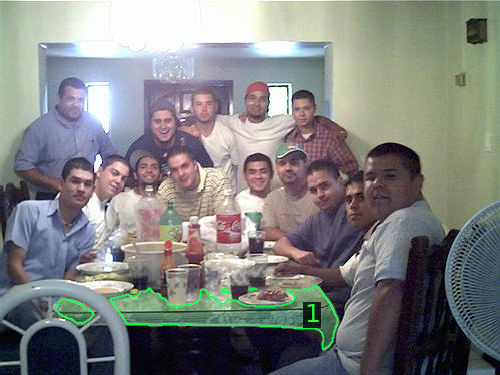}\\[2pt]
{\scriptsize\color{exMute} Image returned with this output.}\end{center}
\begin{exharness}
\exLabel{Harness}{environment update, summarized: new memory entries; new data bindings verbatim}
\begin{exverb}
Memory update: registers 1 new image in memory (tool\_\allowbreak{}generated\_\allowbreak{}image\_\allowbreak{}6,\allowbreak{} from Visual\_\allowbreak{}Grounding\_\allowbreak{}Tool).\\{}
- segment\_\allowbreak{}table: dict (keys: pred\_\allowbreak{}boxes,\allowbreak{} pred\_\allowbreak{}scores,\allowbreak{} pred\_\allowbreak{}masks,\allowbreak{} num\_\allowbreak{}masks,\allowbreak{} orig\_\allowbreak{}img\_\allowbreak{}h,\allowbreak{} orig\_\allowbreak{}img\_\allowbreak{}w)
\end{exverb}
\end{exharness}
\end{exstep}
\begin{exstep}{7}{\toolVG}{}
\exWalk{The query ``man'' is an alternative wording for the people, and again returns 13.}
\exLabel{Arguments}{}
\begin{exverb}
\{\\{}
~~"\allowbreak{}query"\allowbreak{}: "\allowbreak{}man"\allowbreak{},\allowbreak{}\\{}
~~"\allowbreak{}image\_\allowbreak{}id"\allowbreak{}: "\allowbreak{}input\_\allowbreak{}image\_\allowbreak{}1"\allowbreak{}\\{}
\}
\end{exverb}
\exLabel{Tool output}{}
\begin{exverb}
Visual\_\allowbreak{}Grounding\_\allowbreak{}Tool grounded 13 objects for the prompt \textquotesingle{}man\textquotesingle{}. Their corresponding bounding boxes (on a 0-1000 scale,\allowbreak{} [x1,\allowbreak{} y1,\allowbreak{} x2,\allowbreak{} y2]) are: \textasciigrave{}\textasciigrave{}\textasciigrave{}json\\{}
[\\{}
~~~~\{"\allowbreak{}bbox\_\allowbreak{}2d"\allowbreak{}: [0,\allowbreak{} 422,\allowbreak{} 190,\allowbreak{} 870],\allowbreak{} "\allowbreak{}label"\allowbreak{}: "\allowbreak{}man"\allowbreak{}\},\allowbreak{}\\{}
~~~~\{"\allowbreak{}bbox\_\allowbreak{}2d"\allowbreak{}: [29,\allowbreak{} 209,\allowbreak{} 234,\allowbreak{} 533],\allowbreak{} "\allowbreak{}label"\allowbreak{}: "\allowbreak{}man"\allowbreak{}\},\allowbreak{}\\{}
~~~~\{"\allowbreak{}bbox\_\allowbreak{}2d"\allowbreak{}: [166,\allowbreak{} 415,\allowbreak{} 259,\allowbreak{} 667],\allowbreak{} "\allowbreak{}label"\allowbreak{}: "\allowbreak{}man"\allowbreak{}\},\allowbreak{}\\{}
~~~~\{"\allowbreak{}bbox\_\allowbreak{}2d"\allowbreak{}: [212,\allowbreak{} 402,\allowbreak{} 324,\allowbreak{} 636],\allowbreak{} "\allowbreak{}label"\allowbreak{}: "\allowbreak{}man"\allowbreak{}\},\allowbreak{}\\{}
~~~~\{"\allowbreak{}bbox\_\allowbreak{}2d"\allowbreak{}: [251,\allowbreak{} 272,\allowbreak{} 426,\allowbreak{} 478],\allowbreak{} "\allowbreak{}label"\allowbreak{}: "\allowbreak{}man"\allowbreak{}\},\allowbreak{}\\{}
~~~~\{"\allowbreak{}bbox\_\allowbreak{}2d"\allowbreak{}: [313,\allowbreak{} 389,\allowbreak{} 462,\allowbreak{} 589],\allowbreak{} "\allowbreak{}label"\allowbreak{}: "\allowbreak{}man"\allowbreak{}\},\allowbreak{}\\{}
~~~~\{"\allowbreak{}bbox\_\allowbreak{}2d"\allowbreak{}: [381,\allowbreak{} 236,\allowbreak{} 478,\allowbreak{} 504],\allowbreak{} "\allowbreak{}label"\allowbreak{}: "\allowbreak{}man"\allowbreak{}\},\allowbreak{}\\{}
~~~~\{"\allowbreak{}bbox\_\allowbreak{}2d"\allowbreak{}: [430,\allowbreak{} 219,\allowbreak{} 694,\allowbreak{} 512],\allowbreak{} "\allowbreak{}label"\allowbreak{}: "\allowbreak{}man"\allowbreak{}\},\allowbreak{}\\{}
~~~~\{"\allowbreak{}bbox\_\allowbreak{}2d"\allowbreak{}: [470,\allowbreak{} 410,\allowbreak{} 545,\allowbreak{} 573],\allowbreak{} "\allowbreak{}label"\allowbreak{}: "\allowbreak{}man"\allowbreak{}\},\allowbreak{}\\{}
~~~~\{"\allowbreak{}bbox\_\allowbreak{}2d"\allowbreak{}: [521,\allowbreak{} 384,\allowbreak{} 636,\allowbreak{} 642],\allowbreak{} "\allowbreak{}label"\allowbreak{}: "\allowbreak{}man"\allowbreak{}\},\allowbreak{}\\{}
~~~~\{"\allowbreak{}bbox\_\allowbreak{}2d"\allowbreak{}: [537,\allowbreak{} 382,\allowbreak{} 887,\allowbreak{} 1000],\allowbreak{} "\allowbreak{}label"\allowbreak{}: "\allowbreak{}man"\allowbreak{}\},\allowbreak{}\\{}
~~~~\{"\allowbreak{}bbox\_\allowbreak{}2d"\allowbreak{}: [543,\allowbreak{} 432,\allowbreak{} 726,\allowbreak{} 845],\allowbreak{} "\allowbreak{}label"\allowbreak{}: "\allowbreak{}man"\allowbreak{}\},\allowbreak{}\\{}
~~~~\{"\allowbreak{}bbox\_\allowbreak{}2d"\allowbreak{}: [570,\allowbreak{} 240,\allowbreak{} 716,\allowbreak{} 470],\allowbreak{} "\allowbreak{}label"\allowbreak{}: "\allowbreak{}man"\allowbreak{}\}\\{}
]\textasciigrave{}\textasciigrave{}\textasciigrave{}. The object numbers correspond to the labeled masks rendered in the following image,\allowbreak{} which you can refer to as tool\_\allowbreak{}generated\_\allowbreak{}image\_\allowbreak{}7.
\end{exverb}
\begin{center}\includegraphics[width=0.45\linewidth,height=0.22\textheight,keepaspectratio]{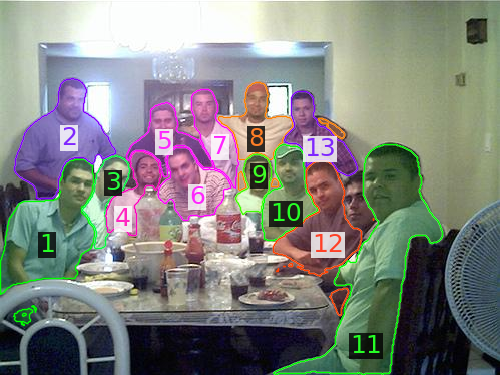}\\[2pt]
{\scriptsize\color{exMute} Image returned with this output.}\end{center}
\begin{exharness}
\exLabel{Harness}{environment update, summarized: new memory entries; new data bindings verbatim}
\begin{exverb}
Memory update: registers 1 new image in memory (tool\_\allowbreak{}generated\_\allowbreak{}image\_\allowbreak{}7,\allowbreak{} from Visual\_\allowbreak{}Grounding\_\allowbreak{}Tool).\\{}
- segment\_\allowbreak{}man: dict (keys: pred\_\allowbreak{}boxes,\allowbreak{} pred\_\allowbreak{}scores,\allowbreak{} pred\_\allowbreak{}masks,\allowbreak{} num\_\allowbreak{}masks,\allowbreak{} orig\_\allowbreak{}img\_\allowbreak{}h,\allowbreak{} orig\_\allowbreak{}img\_\allowbreak{}w)
\end{exverb}
\end{exharness}
\end{exstep}
\begin{exstep}{8}{\toolVG}{}
\exWalk{A narrow query for one specific person, ``person in dark shirt'', returns a single match.}
\exLabel{Arguments}{}
\begin{exverb}
\{\\{}
~~"\allowbreak{}query"\allowbreak{}: "\allowbreak{}person in dark shirt"\allowbreak{},\allowbreak{}\\{}
~~"\allowbreak{}image\_\allowbreak{}id"\allowbreak{}: "\allowbreak{}input\_\allowbreak{}image\_\allowbreak{}1"\allowbreak{}\\{}
\}
\end{exverb}
\exLabel{Tool output}{}
\begin{exverb}
Visual\_\allowbreak{}Grounding\_\allowbreak{}Tool grounded 1 objects for the prompt \textquotesingle{}person in dark shirt\textquotesingle{}. Their corresponding bounding boxes (on a 0-1000 scale,\allowbreak{} [x1,\allowbreak{} y1,\allowbreak{} x2,\allowbreak{} y2]) are: \textasciigrave{}\textasciigrave{}\textasciigrave{}json\\{}
[\\{}
~~~~\{"\allowbreak{}bbox\_\allowbreak{}2d"\allowbreak{}: [251,\allowbreak{} 273,\allowbreak{} 426,\allowbreak{} 483],\allowbreak{} "\allowbreak{}label"\allowbreak{}: "\allowbreak{}person in dark shirt"\allowbreak{}\}\\{}
]\textasciigrave{}\textasciigrave{}\textasciigrave{}. The object numbers correspond to the labeled masks rendered in the following image,\allowbreak{} which you can refer to as tool\_\allowbreak{}generated\_\allowbreak{}image\_\allowbreak{}8.
\end{exverb}
\begin{center}\includegraphics[width=0.45\linewidth,height=0.22\textheight,keepaspectratio]{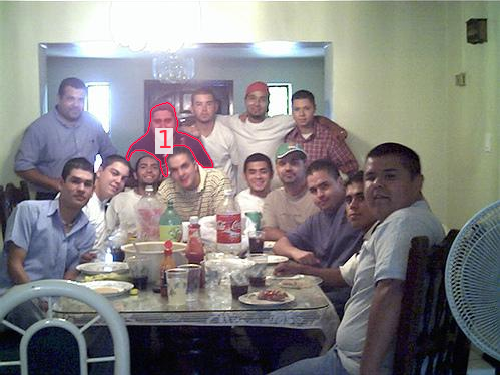}\\[2pt]
{\scriptsize\color{exMute} Image returned with this output.}\end{center}
\begin{exharness}
\exLabel{Harness}{environment update, summarized: new memory entries; new data bindings verbatim}
\begin{exverb}
Memory update: registers 1 new image in memory (tool\_\allowbreak{}generated\_\allowbreak{}image\_\allowbreak{}8,\allowbreak{} from Visual\_\allowbreak{}Grounding\_\allowbreak{}Tool).\\{}
- segment\_\allowbreak{}person\_\allowbreak{}in\_\allowbreak{}dark\_\allowbreak{}shirt: dict (keys: pred\_\allowbreak{}boxes,\allowbreak{} pred\_\allowbreak{}scores,\allowbreak{} pred\_\allowbreak{}masks,\allowbreak{} num\_\allowbreak{}masks,\allowbreak{} orig\_\allowbreak{}img\_\allowbreak{}h,\allowbreak{} orig\_\allowbreak{}img\_\allowbreak{}w)
\end{exverb}
\end{exharness}
\end{exstep}
\begin{exstep}{9}{\toolZoom}{}
\exWalk{A second, tighter zoom on a group of people on the right side of the table.}
\exLabel{Arguments}{}
\begin{exverb}
\{\\{}
~~"\allowbreak{}image\_\allowbreak{}id"\allowbreak{}: "\allowbreak{}input\_\allowbreak{}image\_\allowbreak{}1"\allowbreak{},\allowbreak{}\\{}
~~"\allowbreak{}coords"\allowbreak{}: [630,\allowbreak{} 400,\allowbreak{} 785,\allowbreak{} 730],\allowbreak{}\\{}
~~"\allowbreak{}padding"\allowbreak{}: 50\\{}
\}
\end{exverb}
\exLabel{Tool output}{}
\begin{exverb}
Tool \textasciigrave{}Zoom\_\allowbreak{}In\_\allowbreak{}Tool\textasciigrave{} (tool\_\allowbreak{}call\_\allowbreak{}id=\allowbreak{}call\_\allowbreak{}vl7F9q9d7GQQotr47vmOtha0) returned:\\{}
Zoomed in image rendered below; refer to it as tool\_\allowbreak{}generated\_\allowbreak{}image\_\allowbreak{}9.
\end{exverb}
\begin{center}\includegraphics[width=0.45\linewidth,height=0.22\textheight,keepaspectratio]{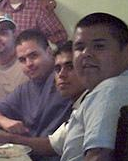}\\[2pt]
{\scriptsize\color{exMute} Image returned with this output.}\end{center}
\begin{exharness}
\exLabel{Harness}{environment update, summarized: new memory entries; new data bindings verbatim}
\begin{exverb}
Memory update: registers 1 new image in memory (tool\_\allowbreak{}generated\_\allowbreak{}image\_\allowbreak{}9,\allowbreak{} from Zoom\_\allowbreak{}In\_\allowbreak{}Tool).
\end{exverb}
\end{exharness}
\end{exstep}
\begin{exstep}{10}{\toolVG}{}
\exWalk{The last query is a referring expression, ``man behind the man in white shirt'', aimed at one partly hidden person. SAM~3 accepts such phrases, and the harness lets the agent use them when a category name is not enough.}
\exLabel{Arguments}{}
\begin{exverb}
\{\\{}
~~"\allowbreak{}query"\allowbreak{}: "\allowbreak{}man behind the man in white shirt"\allowbreak{},\allowbreak{}\\{}
~~"\allowbreak{}image\_\allowbreak{}id"\allowbreak{}: "\allowbreak{}input\_\allowbreak{}image\_\allowbreak{}1"\allowbreak{}\\{}
\}
\end{exverb}
\exLabel{Tool output}{}
\begin{exverb}
Visual\_\allowbreak{}Grounding\_\allowbreak{}Tool grounded 7 objects for the prompt \textquotesingle{}man behind the man in white shirt\textquotesingle{}. Their corresponding bounding boxes (on a 0-1000 scale,\allowbreak{} [x1,\allowbreak{} y1,\allowbreak{} x2,\allowbreak{} y2]) are: \textasciigrave{}\textasciigrave{}\textasciigrave{}json\\{}
[\\{}
~~~~\{"\allowbreak{}bbox\_\allowbreak{}2d"\allowbreak{}: [165,\allowbreak{} 414,\allowbreak{} 259,\allowbreak{} 667],\allowbreak{} "\allowbreak{}label"\allowbreak{}: "\allowbreak{}man behind the man in white shirt"\allowbreak{}\},\allowbreak{}\\{}
~~~~\{"\allowbreak{}bbox\_\allowbreak{}2d"\allowbreak{}: [211,\allowbreak{} 402,\allowbreak{} 324,\allowbreak{} 637],\allowbreak{} "\allowbreak{}label"\allowbreak{}: "\allowbreak{}man behind the man in white shirt"\allowbreak{}\},\allowbreak{}\\{}
~~~~\{"\allowbreak{}bbox\_\allowbreak{}2d"\allowbreak{}: [381,\allowbreak{} 235,\allowbreak{} 478,\allowbreak{} 506],\allowbreak{} "\allowbreak{}label"\allowbreak{}: "\allowbreak{}man behind the man in white shirt"\allowbreak{}\},\allowbreak{}\\{}
~~~~\{"\allowbreak{}bbox\_\allowbreak{}2d"\allowbreak{}: [430,\allowbreak{} 219,\allowbreak{} 694,\allowbreak{} 513],\allowbreak{} "\allowbreak{}label"\allowbreak{}: "\allowbreak{}man behind the man in white shirt"\allowbreak{}\},\allowbreak{}\\{}
~~~~\{"\allowbreak{}bbox\_\allowbreak{}2d"\allowbreak{}: [470,\allowbreak{} 409,\allowbreak{} 545,\allowbreak{} 573],\allowbreak{} "\allowbreak{}label"\allowbreak{}: "\allowbreak{}man behind the man in white shirt"\allowbreak{}\},\allowbreak{}\\{}
~~~~\{"\allowbreak{}bbox\_\allowbreak{}2d"\allowbreak{}: [522,\allowbreak{} 384,\allowbreak{} 635,\allowbreak{} 642],\allowbreak{} "\allowbreak{}label"\allowbreak{}: "\allowbreak{}man behind the man in white shirt"\allowbreak{}\},\allowbreak{}\\{}
~~~~\{"\allowbreak{}bbox\_\allowbreak{}2d"\allowbreak{}: [537,\allowbreak{} 382,\allowbreak{} 887,\allowbreak{} 1000],\allowbreak{} "\allowbreak{}label"\allowbreak{}: "\allowbreak{}man behind the man in white shirt"\allowbreak{}\}\\{}
]\textasciigrave{}\textasciigrave{}\textasciigrave{}. The object numbers correspond to the labeled masks rendered in the following image,\allowbreak{} which you can refer to as tool\_\allowbreak{}generated\_\allowbreak{}image\_\allowbreak{}10.
\end{exverb}
\begin{center}\includegraphics[width=0.45\linewidth,height=0.22\textheight,keepaspectratio]{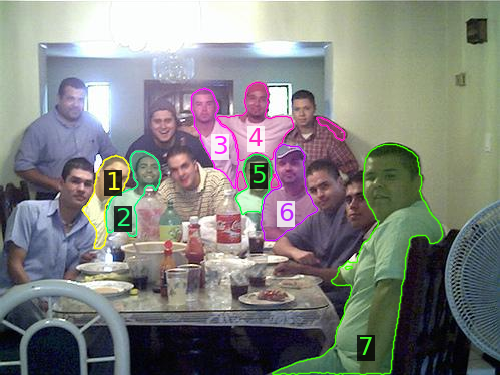}\\[2pt]
{\scriptsize\color{exMute} Image returned with this output.}\end{center}
\begin{exharness}
\exLabel{Harness}{environment update, summarized: new memory entries; new data bindings verbatim}
\begin{exverb}
Memory update: registers 1 new image in memory (tool\_\allowbreak{}generated\_\allowbreak{}image\_\allowbreak{}10,\allowbreak{} from Visual\_\allowbreak{}Grounding\_\allowbreak{}Tool).\\{}
- segment\_\allowbreak{}man\_\allowbreak{}behind\_\allowbreak{}the\_\allowbreak{}man\_\allowbreak{}in\_\allowbreak{}white\_\allowbreak{}sh: dict (keys: pred\_\allowbreak{}boxes,\allowbreak{} pred\_\allowbreak{}scores,\allowbreak{} pred\_\allowbreak{}masks,\allowbreak{} num\_\allowbreak{}masks,\allowbreak{} orig\_\allowbreak{}img\_\allowbreak{}h,\allowbreak{} orig\_\allowbreak{}img\_\allowbreak{}w)
\end{exverb}
\end{exharness}
\end{exstep}
\begin{exstep}{}{\exTool{exInk}{Final answer}}{}
\exWalk{The agent answers with 23 labelled boxes; the evaluator matches 22 of them to ground truth, with one false positive and one miss.}
\exLabel{Model}{}
\begin{exverb}
\textless{}answer\textgreater{}[\{"\allowbreak{}bbox\_\allowbreak{}2d"\allowbreak{}:[0,\allowbreak{}422,\allowbreak{}191,\allowbreak{}871],\allowbreak{}"\allowbreak{}label"\allowbreak{}:"\allowbreak{}person"\allowbreak{}\},\allowbreak{}\{"\allowbreak{}bbox\_\allowbreak{}2d"\allowbreak{}:[29,\allowbreak{}208,\allowbreak{}234,\allowbreak{}533],\allowbreak{}"\allowbreak{}label"\allowbreak{}:"\allowbreak{}person"\allowbreak{}\},\allowbreak{}\{"\allowbreak{}bbox\_\allowbreak{}2d"\allowbreak{}:[165,\allowbreak{}414,\allowbreak{}260,\allowbreak{}667],\allowbreak{}"\allowbreak{}label"\allowbreak{}:"\allowbreak{}person"\allowbreak{}\},\allowbreak{}\{"\allowbreak{}bbox\_\allowbreak{}2d"\allowbreak{}:[211,\allowbreak{}402,\allowbreak{}324,\allowbreak{}636],\allowbreak{}"\allowbreak{}label"\allowbreak{}:"\allowbreak{}person"\allowbreak{}\},\allowbreak{}\{"\allowbreak{}bbox\_\allowbreak{}2d"\allowbreak{}:[251,\allowbreak{}272,\allowbreak{}426,\allowbreak{}480],\allowbreak{}"\allowbreak{}label"\allowbreak{}:"\allowbreak{}person"\allowbreak{}\},\allowbreak{}\{"\allowbreak{}bbox\_\allowbreak{}2d"\allowbreak{}:[313,\allowbreak{}389,\allowbreak{}462,\allowbreak{}589],\allowbreak{}"\allowbreak{}label"\allowbreak{}:"\allowbreak{}person"\allowbreak{}\},\allowbreak{}\{"\allowbreak{}bbox\_\allowbreak{}2d"\allowbreak{}:[381,\allowbreak{}236,\allowbreak{}478,\allowbreak{}505],\allowbreak{}"\allowbreak{}label"\allowbreak{}:"\allowbreak{}person"\allowbreak{}\},\allowbreak{}\{"\allowbreak{}bbox\_\allowbreak{}2d"\allowbreak{}:[428,\allowbreak{}219,\allowbreak{}695,\allowbreak{}512],\allowbreak{}"\allowbreak{}label"\allowbreak{}:"\allowbreak{}person"\allowbreak{}\},\allowbreak{}\{"\allowbreak{}bbox\_\allowbreak{}2d"\allowbreak{}:[470,\allowbreak{}410,\allowbreak{}545,\allowbreak{}574],\allowbreak{}"\allowbreak{}label"\allowbreak{}:"\allowbreak{}person"\allowbreak{}\},\allowbreak{}\{"\allowbreak{}bbox\_\allowbreak{}2d"\allowbreak{}:[521,\allowbreak{}384,\allowbreak{}635,\allowbreak{}642],\allowbreak{}"\allowbreak{}label"\allowbreak{}:"\allowbreak{}person"\allowbreak{}\},\allowbreak{}\{"\allowbreak{}bbox\_\allowbreak{}2d"\allowbreak{}:[543,\allowbreak{}434,\allowbreak{}727,\allowbreak{}846],\allowbreak{}"\allowbreak{}label"\allowbreak{}:"\allowbreak{}person"\allowbreak{}\},\allowbreak{}\{"\allowbreak{}bbox\_\allowbreak{}2d"\allowbreak{}:[675,\allowbreak{}465,\allowbreak{}759,\allowbreak{}803],\allowbreak{}"\allowbreak{}label"\allowbreak{}:"\allowbreak{}person"\allowbreak{}\},\allowbreak{}\{"\allowbreak{}bbox\_\allowbreak{}2d"\allowbreak{}:[536,\allowbreak{}382,\allowbreak{}888,\allowbreak{}1000],\allowbreak{}"\allowbreak{}label"\allowbreak{}:"\allowbreak{}person"\allowbreak{}\},\allowbreak{}\{"\allowbreak{}bbox\_\allowbreak{}2d"\allowbreak{}:[569,\allowbreak{}240,\allowbreak{}717,\allowbreak{}470],\allowbreak{}"\allowbreak{}label"\allowbreak{}:"\allowbreak{}person"\allowbreak{}\},\allowbreak{}\{"\allowbreak{}bbox\_\allowbreak{}2d"\allowbreak{}:[271,\allowbreak{}495,\allowbreak{}327,\allowbreak{}642],\allowbreak{}"\allowbreak{}label"\allowbreak{}:"\allowbreak{}bottle"\allowbreak{}\},\allowbreak{}\{"\allowbreak{}bbox\_\allowbreak{}2d"\allowbreak{}:[318,\allowbreak{}533,\allowbreak{}364,\allowbreak{}645],\allowbreak{}"\allowbreak{}label"\allowbreak{}:"\allowbreak{}bottle"\allowbreak{}\},\allowbreak{}\{"\allowbreak{}bbox\_\allowbreak{}2d"\allowbreak{}:[319,\allowbreak{}640,\allowbreak{}354,\allowbreak{}794],\allowbreak{}"\allowbreak{}label"\allowbreak{}:"\allowbreak{}bottle"\allowbreak{}\},\allowbreak{}\{"\allowbreak{}bbox\_\allowbreak{}2d"\allowbreak{}:[369,\allowbreak{}576,\allowbreak{}409,\allowbreak{}705],\allowbreak{}"\allowbreak{}label"\allowbreak{}:"\allowbreak{}bottle"\allowbreak{}\},\allowbreak{}\{"\allowbreak{}bbox\_\allowbreak{}2d"\allowbreak{}:[431,\allowbreak{}504,\allowbreak{}483,\allowbreak{}681],\allowbreak{}"\allowbreak{}label"\allowbreak{}:"\allowbreak{}bottle"\allowbreak{}\},\allowbreak{}\{"\allowbreak{}bbox\_\allowbreak{}2d"\allowbreak{}:[80,\allowbreak{}610,\allowbreak{}680,\allowbreak{}947],\allowbreak{}"\allowbreak{}label"\allowbreak{}:"\allowbreak{}diningtable"\allowbreak{}\},\allowbreak{}\{"\allowbreak{}bbox\_\allowbreak{}2d"\allowbreak{}:[0,\allowbreak{}745,\allowbreak{}260,\allowbreak{}999],\allowbreak{}"\allowbreak{}label"\allowbreak{}:"\allowbreak{}chair"\allowbreak{}\},\allowbreak{}\{"\allowbreak{}bbox\_\allowbreak{}2d"\allowbreak{}:[0,\allowbreak{}485,\allowbreak{}61,\allowbreak{}635],\allowbreak{}"\allowbreak{}label"\allowbreak{}:"\allowbreak{}chair"\allowbreak{}\},\allowbreak{}\{"\allowbreak{}bbox\_\allowbreak{}2d"\allowbreak{}:[785,\allowbreak{}612,\allowbreak{}939,\allowbreak{}999],\allowbreak{}"\allowbreak{}label"\allowbreak{}:"\allowbreak{}chair"\allowbreak{}\}]\textless{}/\allowbreak{}answer\textgreater{}
\end{exverb}
\end{exstep}

%% file: text/appendix_examples_steps/fruit_header.tex
\exOverview{\textbf{Task.} Box every ``package of fresh cut fruit'' on a supermarket shelf that holds 53 plastic fruit cups in rows. 

\textbf{Why it is hard.} The category name as given is unusual; a detector queried with it finds only part of the shelf. 

\textbf{How the agent solves it.} It rephrases the category as ``fruit cup'', which finds every cup, checks two regions with zooms, and returns 53 boxes that all match. Note that the no-tool answer scores zero only because it used pixel coordinates instead of the requested 0--1000 scale (see the note under the results).}
\exLabel{Input images}{}
\begin{center}\includegraphics[width=0.55\linewidth,height=0.28\textheight,keepaspectratio]{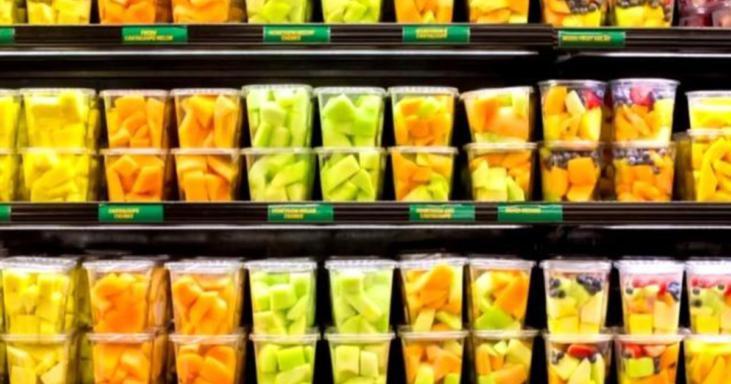}\\[2pt]
{\scriptsize\color{exMute} input\_image\_1}\end{center}
\exLabel{Question}{as sent to the model}
\begin{exverb}
Question: Locate all instances of \textquotesingle{}package of fresh cut fruit\textquotesingle{} in this image. For each instance,\allowbreak{} output a bounding box as [x1,\allowbreak{} y1,\allowbreak{} x2,\allowbreak{} y2] where coordinates are normalised on a 0-1000 scale (with (0,\allowbreak{}0) as the top-left corner and (1000,\allowbreak{}1000) as the bottom-right corner). Your answer should be a list of lists. If there are no instances of \textquotesingle{}package of fresh cut fruit\textquotesingle{} in the image,\allowbreak{} output an empty list [] as your answer.
\end{exverb}
\exLabel{Without tools}{GPT-6 Sol's full response to the same question}
\begin{exverb}
[[0,\allowbreak{}0,\allowbreak{}24,\allowbreak{}24],\allowbreak{}[24,\allowbreak{}0,\allowbreak{}95,\allowbreak{}24],\allowbreak{}[98,\allowbreak{}0,\allowbreak{}171,\allowbreak{}24],\allowbreak{}[173,\allowbreak{}0,\allowbreak{}244,\allowbreak{}24],\allowbreak{}[247,\allowbreak{}0,\allowbreak{}315,\allowbreak{}24],\allowbreak{}[319,\allowbreak{}0,\allowbreak{}392,\allowbreak{}24],\allowbreak{}[395,\allowbreak{}0,\allowbreak{}465,\allowbreak{}24],\allowbreak{}[470,\allowbreak{}0,\allowbreak{}538,\allowbreak{}24],\allowbreak{}[542,\allowbreak{}0,\allowbreak{}611,\allowbreak{}24],\allowbreak{}[612,\allowbreak{}0,\allowbreak{}679,\allowbreak{}24],\allowbreak{}[680,\allowbreak{}0,\allowbreak{}731,\allowbreak{}24],\allowbreak{}[0,\allowbreak{}88,\allowbreak{}22,\allowbreak{}151],\allowbreak{}[22,\allowbreak{}88,\allowbreak{}101,\allowbreak{}151],\allowbreak{}[104,\allowbreak{}88,\allowbreak{}175,\allowbreak{}151],\allowbreak{}[175,\allowbreak{}88,\allowbreak{}246,\allowbreak{}151],\allowbreak{}[244,\allowbreak{}85,\allowbreak{}319,\allowbreak{}151],\allowbreak{}[318,\allowbreak{}86,\allowbreak{}391,\allowbreak{}151],\allowbreak{}[392,\allowbreak{}86,\allowbreak{}464,\allowbreak{}151],\allowbreak{}[463,\allowbreak{}87,\allowbreak{}538,\allowbreak{}151],\allowbreak{}[539,\allowbreak{}79,\allowbreak{}612,\allowbreak{}149],\allowbreak{}[611,\allowbreak{}78,\allowbreak{}685,\allowbreak{}149],\allowbreak{}[687,\allowbreak{}89,\allowbreak{}731,\allowbreak{}149],\allowbreak{}[0,\allowbreak{}149,\allowbreak{}22,\allowbreak{}204],\allowbreak{}[22,\allowbreak{}149,\allowbreak{}101,\allowbreak{}204],\allowbreak{}[103,\allowbreak{}149,\allowbreak{}175,\allowbreak{}204],\allowbreak{}[175,\allowbreak{}149,\allowbreak{}246,\allowbreak{}204],\allowbreak{}[245,\allowbreak{}149,\allowbreak{}319,\allowbreak{}204],\allowbreak{}[318,\allowbreak{}149,\allowbreak{}391,\allowbreak{}204],\allowbreak{}[391,\allowbreak{}149,\allowbreak{}464,\allowbreak{}204],\allowbreak{}[464,\allowbreak{}147,\allowbreak{}538,\allowbreak{}204],\allowbreak{}[539,\allowbreak{}145,\allowbreak{}612,\allowbreak{}204],\allowbreak{}[611,\allowbreak{}144,\allowbreak{}685,\allowbreak{}204],\allowbreak{}[686,\allowbreak{}145,\allowbreak{}731,\allowbreak{}204],\allowbreak{}[0,\allowbreak{}255,\allowbreak{}84,\allowbreak{}341],\allowbreak{}[84,\allowbreak{}257,\allowbreak{}166,\allowbreak{}341],\allowbreak{}[165,\allowbreak{}258,\allowbreak{}248,\allowbreak{}341],\allowbreak{}[247,\allowbreak{}258,\allowbreak{}324,\allowbreak{}341],\allowbreak{}[326,\allowbreak{}258,\allowbreak{}400,\allowbreak{}341],\allowbreak{}[400,\allowbreak{}258,\allowbreak{}469,\allowbreak{}340],\allowbreak{}[469,\allowbreak{}258,\allowbreak{}536,\allowbreak{}340],\allowbreak{}[540,\allowbreak{}258,\allowbreak{}615,\allowbreak{}341],\allowbreak{}[616,\allowbreak{}258,\allowbreak{}686,\allowbreak{}341],\allowbreak{}[686,\allowbreak{}260,\allowbreak{}731,\allowbreak{}341],\allowbreak{}[0,\allowbreak{}335,\allowbreak{}84,\allowbreak{}384],\allowbreak{}[84,\allowbreak{}335,\allowbreak{}166,\allowbreak{}384],\allowbreak{}[166,\allowbreak{}335,\allowbreak{}248,\allowbreak{}384],\allowbreak{}[248,\allowbreak{}334,\allowbreak{}324,\allowbreak{}384],\allowbreak{}[326,\allowbreak{}334,\allowbreak{}400,\allowbreak{}384],\allowbreak{}[400,\allowbreak{}332,\allowbreak{}469,\allowbreak{}384],\allowbreak{}[469,\allowbreak{}332,\allowbreak{}536,\allowbreak{}384],\allowbreak{}[540,\allowbreak{}332,\allowbreak{}615,\allowbreak{}384],\allowbreak{}[616,\allowbreak{}332,\allowbreak{}686,\allowbreak{}384],\allowbreak{}[686,\allowbreak{}332,\allowbreak{}731,\allowbreak{}384]]
\end{exverb}

%% file: text/appendix_examples_steps/fruit.tex
\begin{exharness}
\exLabel{Harness}{environment update, shown in full}
\begin{exverb}
Environment Update (not a new question):\\{}
Currently available image IDs for tool calls:\\{}
- input\_\allowbreak{}image\_\allowbreak{}1: user\_\allowbreak{}input\\{}
When a tool needs image input,\allowbreak{} pass one of these IDs via the *\_\allowbreak{}image\_\allowbreak{}id field.\\{}
Do NOT invent image file names or paths.
\end{exverb}
\end{exharness}
\begin{exstep}{1}{\toolVG}{}
\exWalk{Grounding the category exactly as asked returns only 25 candidates, missing much of the shelf.}
\exLabel{Arguments}{}
\begin{exverb}
\{\\{}
~~"\allowbreak{}query"\allowbreak{}: "\allowbreak{}package of fresh cut fruit"\allowbreak{},\allowbreak{}\\{}
~~"\allowbreak{}image\_\allowbreak{}id"\allowbreak{}: "\allowbreak{}input\_\allowbreak{}image\_\allowbreak{}1"\allowbreak{}\\{}
\}
\end{exverb}
\exLabel{Tool output}{}
\begin{exverb}
Visual\_\allowbreak{}Grounding\_\allowbreak{}Tool grounded 25 objects for the prompt \textquotesingle{}package of fresh cut fruit\textquotesingle{}. Their corresponding bounding boxes (on a 0-1000 scale,\allowbreak{} [x1,\allowbreak{} y1,\allowbreak{} x2,\allowbreak{} y2]) are: \textasciigrave{}\textasciigrave{}\textasciigrave{}json\\{}
[\\{}
~~~~\{"\allowbreak{}bbox\_\allowbreak{}2d"\allowbreak{}: [337,\allowbreak{} 673,\allowbreak{} 434,\allowbreak{} 873],\allowbreak{} "\allowbreak{}label"\allowbreak{}: "\allowbreak{}package of fresh cut fruit"\allowbreak{}\},\allowbreak{}\\{}
~~~~\{"\allowbreak{}bbox\_\allowbreak{}2d"\allowbreak{}: [343,\allowbreak{} 859,\allowbreak{} 437,\allowbreak{} 1000],\allowbreak{} "\allowbreak{}label"\allowbreak{}: "\allowbreak{}package of fresh cut fruit"\allowbreak{}\},\allowbreak{}\\{}
~~~~\{"\allowbreak{}bbox\_\allowbreak{}2d"\allowbreak{}: [431,\allowbreak{} 226,\allowbreak{} 527,\allowbreak{} 383],\allowbreak{} "\allowbreak{}label"\allowbreak{}: "\allowbreak{}package of fresh cut fruit"\allowbreak{}\},\allowbreak{}\\{}
~~~~\{"\allowbreak{}bbox\_\allowbreak{}2d"\allowbreak{}: [445,\allowbreak{} 855,\allowbreak{} 538,\allowbreak{} 1000],\allowbreak{} "\allowbreak{}label"\allowbreak{}: "\allowbreak{}package of fresh cut fruit"\allowbreak{}\},\allowbreak{}\\{}
~~~~\{"\allowbreak{}bbox\_\allowbreak{}2d"\allowbreak{}: [445,\allowbreak{} 674,\allowbreak{} 541,\allowbreak{} 869],\allowbreak{} "\allowbreak{}label"\allowbreak{}: "\allowbreak{}package of fresh cut fruit"\allowbreak{}\},\allowbreak{}\\{}
~~~~\{"\allowbreak{}bbox\_\allowbreak{}2d"\allowbreak{}: [532,\allowbreak{} 225,\allowbreak{} 626,\allowbreak{} 381],\allowbreak{} "\allowbreak{}label"\allowbreak{}: "\allowbreak{}package of fresh cut fruit"\allowbreak{}\},\allowbreak{}\\{}
~~~~\{"\allowbreak{}bbox\_\allowbreak{}2d"\allowbreak{}: [535,\allowbreak{} 381,\allowbreak{} 625,\allowbreak{} 526],\allowbreak{} "\allowbreak{}label"\allowbreak{}: "\allowbreak{}package of fresh cut fruit"\allowbreak{}\},\allowbreak{}\\{}
~~~~\{"\allowbreak{}bbox\_\allowbreak{}2d"\allowbreak{}: [546,\allowbreak{} 663,\allowbreak{} 637,\allowbreak{} 863],\allowbreak{} "\allowbreak{}label"\allowbreak{}: "\allowbreak{}package of fresh cut fruit"\allowbreak{}\},\allowbreak{}\\{}
~~~~\{"\allowbreak{}bbox\_\allowbreak{}2d"\allowbreak{}: [547,\allowbreak{} 850,\allowbreak{} 638,\allowbreak{} 1000],\allowbreak{} "\allowbreak{}label"\allowbreak{}: "\allowbreak{}package of fresh cut fruit"\allowbreak{}\},\allowbreak{}\\{}
~~~~\{"\allowbreak{}bbox\_\allowbreak{}2d"\allowbreak{}: [631,\allowbreak{} 225,\allowbreak{} 729,\allowbreak{} 372],\allowbreak{} "\allowbreak{}label"\allowbreak{}: "\allowbreak{}package of fresh cut fruit"\allowbreak{}\},\allowbreak{}\\{}
~~~~\{"\allowbreak{}bbox\_\allowbreak{}2d"\allowbreak{}: [637,\allowbreak{} 371,\allowbreak{} 733,\allowbreak{} 527],\allowbreak{} "\allowbreak{}label"\allowbreak{}: "\allowbreak{}package of fresh cut fruit"\allowbreak{}\},\allowbreak{}\\{}
~~~~\{"\allowbreak{}bbox\_\allowbreak{}2d"\allowbreak{}: [640,\allowbreak{} 669,\allowbreak{} 731,\allowbreak{} 859],\allowbreak{} "\allowbreak{}label"\allowbreak{}: "\allowbreak{}package of fresh cut fruit"\allowbreak{}\},\allowbreak{}\\{}
~~~~\{"\allowbreak{}bbox\_\allowbreak{}2d"\allowbreak{}: [648,\allowbreak{} 855,\allowbreak{} 735,\allowbreak{} 1000],\allowbreak{} "\allowbreak{}label"\allowbreak{}: "\allowbreak{}package of fresh cut fruit"\allowbreak{}\},\allowbreak{}\\{}
~~~~\{"\allowbreak{}bbox\_\allowbreak{}2d"\allowbreak{}: [738,\allowbreak{} 208,\allowbreak{} 829,\allowbreak{} 372],\allowbreak{} "\allowbreak{}label"\allowbreak{}: "\allowbreak{}package of fresh cut fruit"\allowbreak{}\},\allowbreak{}\\{}
~~~~\{"\allowbreak{}bbox\_\allowbreak{}2d"\allowbreak{}: [739,\allowbreak{} 367,\allowbreak{} 824,\allowbreak{} 526],\allowbreak{} "\allowbreak{}label"\allowbreak{}: "\allowbreak{}package of fresh cut fruit"\allowbreak{}\},\allowbreak{}\\{}
~~~~\{"\allowbreak{}bbox\_\allowbreak{}2d"\allowbreak{}: [740,\allowbreak{} 674,\allowbreak{} 837,\allowbreak{} 871],\allowbreak{} "\allowbreak{}label"\allowbreak{}: "\allowbreak{}package of fresh cut fruit"\allowbreak{}\},\allowbreak{}\\{}
~~~~\{"\allowbreak{}bbox\_\allowbreak{}2d"\allowbreak{}: [745,\allowbreak{} 864,\allowbreak{} 838,\allowbreak{} 1000],\allowbreak{} "\allowbreak{}label"\allowbreak{}: "\allowbreak{}package of fresh cut fruit"\allowbreak{}\},\allowbreak{}
\end{exverb}
\exCut{The remaining 745 characters of this output are not shown (2,199 in total).}
\begin{center}\includegraphics[width=0.45\linewidth,height=0.22\textheight,keepaspectratio]{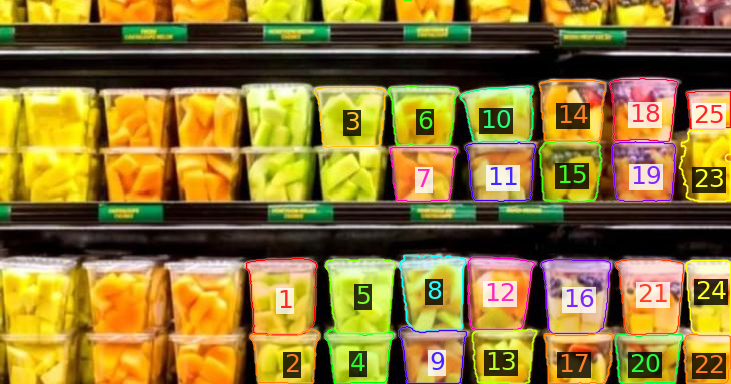}\\[2pt]
{\scriptsize\color{exMute} Image returned with this output.}\end{center}
\begin{exharness}
\exLabel{Harness}{environment update, summarized: new memory entries; new data bindings verbatim}
\begin{exverb}
Memory update: registers 1 new image in memory (tool\_\allowbreak{}generated\_\allowbreak{}image\_\allowbreak{}1,\allowbreak{} from Visual\_\allowbreak{}Grounding\_\allowbreak{}Tool).\\{}
Available data bindings (auto-injected into Python\_\allowbreak{}Coding\_\allowbreak{}Agent\_\allowbreak{}Tool):\\{}
- segment\_\allowbreak{}package\_\allowbreak{}of\_\allowbreak{}fresh\_\allowbreak{}cut\_\allowbreak{}fruit: dict (keys: pred\_\allowbreak{}boxes,\allowbreak{} pred\_\allowbreak{}scores,\allowbreak{} pred\_\allowbreak{}masks,\allowbreak{} num\_\allowbreak{}masks,\allowbreak{} orig\_\allowbreak{}img\_\allowbreak{}h,\allowbreak{} orig\_\allowbreak{}img\_\allowbreak{}w)
\end{exverb}
\end{exharness}
\end{exstep}
\begin{exstep}{2}{\toolVG}{}
\exWalk{The agent rephrases the query as ``fruit cup''. SAM~3 now returns 57 candidates covering the whole shelf, slightly more than the 53 cups.}
\exLabel{Arguments}{}
\begin{exverb}
\{\\{}
~~"\allowbreak{}query"\allowbreak{}: "\allowbreak{}fruit cup"\allowbreak{},\allowbreak{}\\{}
~~"\allowbreak{}image\_\allowbreak{}id"\allowbreak{}: "\allowbreak{}input\_\allowbreak{}image\_\allowbreak{}1"\allowbreak{}\\{}
\}
\end{exverb}
\exLabel{Tool output}{}
\begin{exverb}
Visual\_\allowbreak{}Grounding\_\allowbreak{}Tool grounded 57 objects for the prompt \textquotesingle{}fruit cup\textquotesingle{}. Their corresponding bounding boxes (on a 0-1000 scale,\allowbreak{} [x1,\allowbreak{} y1,\allowbreak{} x2,\allowbreak{} y2]) are: \textasciigrave{}\textasciigrave{}\textasciigrave{}json\\{}
[\\{}
~~~~\{"\allowbreak{}bbox\_\allowbreak{}2d"\allowbreak{}: [0,\allowbreak{} 234,\allowbreak{} 28,\allowbreak{} 387],\allowbreak{} "\allowbreak{}label"\allowbreak{}: "\allowbreak{}fruit cup"\allowbreak{}\},\allowbreak{}\\{}
~~~~\{"\allowbreak{}bbox\_\allowbreak{}2d"\allowbreak{}: [0,\allowbreak{} 0,\allowbreak{} 26,\allowbreak{} 59],\allowbreak{} "\allowbreak{}label"\allowbreak{}: "\allowbreak{}fruit cup"\allowbreak{}\},\allowbreak{}\\{}
~~~~\{"\allowbreak{}bbox\_\allowbreak{}2d"\allowbreak{}: [0,\allowbreak{} 390,\allowbreak{} 25,\allowbreak{} 527],\allowbreak{} "\allowbreak{}label"\allowbreak{}: "\allowbreak{}fruit cup"\allowbreak{}\},\allowbreak{}\\{}
~~~~\{"\allowbreak{}bbox\_\allowbreak{}2d"\allowbreak{}: [1,\allowbreak{} 668,\allowbreak{} 105,\allowbreak{} 871],\allowbreak{} "\allowbreak{}label"\allowbreak{}: "\allowbreak{}fruit cup"\allowbreak{}\},\allowbreak{}\\{}
~~~~\{"\allowbreak{}bbox\_\allowbreak{}2d"\allowbreak{}: [6,\allowbreak{} 867,\allowbreak{} 102,\allowbreak{} 1000],\allowbreak{} "\allowbreak{}label"\allowbreak{}: "\allowbreak{}fruit cup"\allowbreak{}\},\allowbreak{}\\{}
~~~~\{"\allowbreak{}bbox\_\allowbreak{}2d"\allowbreak{}: [26,\allowbreak{} 385,\allowbreak{} 130,\allowbreak{} 527],\allowbreak{} "\allowbreak{}label"\allowbreak{}: "\allowbreak{}fruit cup"\allowbreak{}\},\allowbreak{}\\{}
~~~~\{"\allowbreak{}bbox\_\allowbreak{}2d"\allowbreak{}: [29,\allowbreak{} 229,\allowbreak{} 132,\allowbreak{} 385],\allowbreak{} "\allowbreak{}label"\allowbreak{}: "\allowbreak{}fruit cup"\allowbreak{}\},\allowbreak{}\\{}
~~~~\{"\allowbreak{}bbox\_\allowbreak{}2d"\allowbreak{}: [35,\allowbreak{} 0,\allowbreak{} 124,\allowbreak{} 60],\allowbreak{} "\allowbreak{}label"\allowbreak{}: "\allowbreak{}fruit cup"\allowbreak{}\},\allowbreak{}\\{}
~~~~\{"\allowbreak{}bbox\_\allowbreak{}2d"\allowbreak{}: [83,\allowbreak{} 664,\allowbreak{} 120,\allowbreak{} 861],\allowbreak{} "\allowbreak{}label"\allowbreak{}: "\allowbreak{}fruit cup"\allowbreak{}\},\allowbreak{}\\{}
~~~~\{"\allowbreak{}bbox\_\allowbreak{}2d"\allowbreak{}: [106,\allowbreak{} 866,\allowbreak{} 128,\allowbreak{} 1000],\allowbreak{} "\allowbreak{}label"\allowbreak{}: "\allowbreak{}fruit cup"\allowbreak{}\},\allowbreak{}\\{}
~~~~\{"\allowbreak{}bbox\_\allowbreak{}2d"\allowbreak{}: [114,\allowbreak{} 676,\allowbreak{} 212,\allowbreak{} 867],\allowbreak{} "\allowbreak{}label"\allowbreak{}: "\allowbreak{}fruit cup"\allowbreak{}\},\allowbreak{}\\{}
~~~~\{"\allowbreak{}bbox\_\allowbreak{}2d"\allowbreak{}: [120,\allowbreak{} 865,\allowbreak{} 212,\allowbreak{} 1000],\allowbreak{} "\allowbreak{}label"\allowbreak{}: "\allowbreak{}fruit cup"\allowbreak{}\},\allowbreak{}\\{}
~~~~\{"\allowbreak{}bbox\_\allowbreak{}2d"\allowbreak{}: [135,\allowbreak{} 0,\allowbreak{} 234,\allowbreak{} 59],\allowbreak{} "\allowbreak{}label"\allowbreak{}: "\allowbreak{}fruit cup"\allowbreak{}\},\allowbreak{}\\{}
~~~~\{"\allowbreak{}bbox\_\allowbreak{}2d"\allowbreak{}: [138,\allowbreak{} 234,\allowbreak{} 234,\allowbreak{} 386],\allowbreak{} "\allowbreak{}label"\allowbreak{}: "\allowbreak{}fruit cup"\allowbreak{}\},\allowbreak{}\\{}
~~~~\{"\allowbreak{}bbox\_\allowbreak{}2d"\allowbreak{}: [140,\allowbreak{} 386,\allowbreak{} 233,\allowbreak{} 527],\allowbreak{} "\allowbreak{}label"\allowbreak{}: "\allowbreak{}fruit cup"\allowbreak{}\},\allowbreak{}\\{}
~~~~\{"\allowbreak{}bbox\_\allowbreak{}2d"\allowbreak{}: [166,\allowbreak{} 663,\allowbreak{} 234,\allowbreak{} 856],\allowbreak{} "\allowbreak{}label"\allowbreak{}: "\allowbreak{}fruit cup"\allowbreak{}\},\allowbreak{}\\{}
~~~~\{"\allowbreak{}bbox\_\allowbreak{}2d"\allowbreak{}: [207,\allowbreak{} 858,\allowbreak{} 230,\allowbreak{} 1000],\allowbreak{} "\allowbreak{}label"\allowbreak{}: "\allowbreak{}fruit cup"\allowbreak{}\},\allowbreak{}\\{}
~~~~\{"\allowbreak{}bbox\_\allowbreak{}2d"\allowbreak{}: [227,\allowbreak{} 676,\allowbreak{} 335,\allowbreak{} 873],\allowbreak{} "\allowbreak{}label"\allowbreak{}: "\allowbreak{}fruit cup"\allowbreak{}\},\allowbreak{}\\{}
~~~~\{"\allowbreak{}bbox\_\allowbreak{}2d"\allowbreak{}: [234,\allowbreak{} 866,\allowbreak{} 336,\allowbreak{} 1000],\allowbreak{} "\allowbreak{}label"\allowbreak{}: "\allowbreak{}fruit cup"\allowbreak{}\},\allowbreak{}\\{}
~~~~\{"\allowbreak{}bbox\_\allowbreak{}2d"\allowbreak{}: [235,\allowbreak{} 0,\allowbreak{} 333,\allowbreak{} 57],\allowbreak{} "\allowbreak{}label"\allowbreak{}: "\allowbreak{}fruit cup"\allowbreak{}\},\allowbreak{}\\{}
~~~~\{"\allowbreak{}bbox\_\allowbreak{}2d"\allowbreak{}: [236,\allowbreak{} 230,\allowbreak{} 331,\allowbreak{} 387],\allowbreak{} "\allowbreak{}label"\allowbreak{}: "\allowbreak{}fruit cup"\allowbreak{}\},\allowbreak{}\\{}
~~~~\{"\allowbreak{}bbox\_\allowbreak{}2d"\allowbreak{}: [236,\allowbreak{} 386,\allowbreak{} 331,\allowbreak{} 526],\allowbreak{} "\allowbreak{}label"\allowbreak{}: "\allowbreak{}fruit cup"\allowbreak{}\},\allowbreak{}\\{}
~~~~\{"\allowbreak{}bbox\_\allowbreak{}2d"\allowbreak{}: [332,\allowbreak{} 385,\allowbreak{} 432,\allowbreak{} 526],\allowbreak{} "\allowbreak{}label"\allowbreak{}: "\allowbreak{}fruit cup"\allowbreak{}\},\allowbreak{}
\end{exverb}
\exCut{The remaining 2,101 characters of this output are not shown (3,568 in total).}
\begin{center}\includegraphics[width=0.45\linewidth,height=0.22\textheight,keepaspectratio]{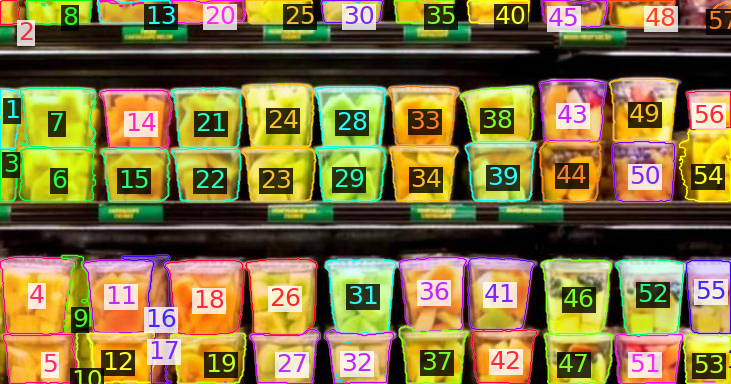}\\[2pt]
{\scriptsize\color{exMute} Image returned with this output.}\end{center}
\begin{exharness}
\exLabel{Harness}{environment update, summarized: new memory entries; new data bindings verbatim}
\begin{exverb}
Memory update: registers 1 new image in memory (tool\_\allowbreak{}generated\_\allowbreak{}image\_\allowbreak{}2,\allowbreak{} from Visual\_\allowbreak{}Grounding\_\allowbreak{}Tool).\\{}
- segment\_\allowbreak{}fruit\_\allowbreak{}cup: dict (keys: pred\_\allowbreak{}boxes,\allowbreak{} pred\_\allowbreak{}scores,\allowbreak{} pred\_\allowbreak{}masks,\allowbreak{} num\_\allowbreak{}masks,\allowbreak{} orig\_\allowbreak{}img\_\allowbreak{}h,\allowbreak{} orig\_\allowbreak{}img\_\allowbreak{}w)
\end{exverb}
\end{exharness}
\end{exstep}
\begin{exstep}{3}{\toolZoom}{}
\exWalk{A zoom on the lower left of the shelf, to check the cups there.}
\exLabel{Arguments}{}
\begin{exverb}
\{\\{}
~~"\allowbreak{}image\_\allowbreak{}id"\allowbreak{}: "\allowbreak{}input\_\allowbreak{}image\_\allowbreak{}1"\allowbreak{},\allowbreak{}\\{}
~~"\allowbreak{}coords"\allowbreak{}: [0,\allowbreak{} 640,\allowbreak{} 343,\allowbreak{} 1000],\allowbreak{}\\{}
~~"\allowbreak{}padding"\allowbreak{}: 50\\{}
\}
\end{exverb}
\exLabel{Tool output}{}
\begin{exverb}
Tool \textasciigrave{}Zoom\_\allowbreak{}In\_\allowbreak{}Tool\textasciigrave{} (tool\_\allowbreak{}call\_\allowbreak{}id=\allowbreak{}call\_\allowbreak{}7rYFu8RXWlhnpwbcoWjvJPR2) returned:\\{}
Zoomed in image rendered below; refer to it as tool\_\allowbreak{}generated\_\allowbreak{}image\_\allowbreak{}3.
\end{exverb}
\begin{center}\includegraphics[width=0.45\linewidth,height=0.22\textheight,keepaspectratio]{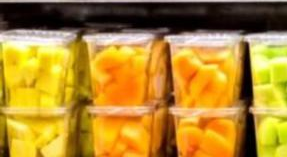}\\[2pt]
{\scriptsize\color{exMute} Image returned with this output.}\end{center}
\begin{exharness}
\exLabel{Harness}{environment update, summarized: new memory entries; new data bindings verbatim}
\begin{exverb}
Memory update: registers 1 new image in memory (tool\_\allowbreak{}generated\_\allowbreak{}image\_\allowbreak{}3,\allowbreak{} from Zoom\_\allowbreak{}In\_\allowbreak{}Tool).
\end{exverb}
\end{exharness}
\end{exstep}
\begin{exstep}{4}{\toolZoom}{}
\exWalk{A zoom on the strip along the top edge, where partly cut-off cups sit.}
\exLabel{Arguments}{}
\begin{exverb}
\{\\{}
~~"\allowbreak{}image\_\allowbreak{}id"\allowbreak{}: "\allowbreak{}input\_\allowbreak{}image\_\allowbreak{}1"\allowbreak{},\allowbreak{}\\{}
~~"\allowbreak{}coords"\allowbreak{}: [0,\allowbreak{} 0,\allowbreak{} 1000,\allowbreak{} 85],\allowbreak{}\\{}
~~"\allowbreak{}padding"\allowbreak{}: 50\\{}
\}
\end{exverb}
\exLabel{Tool output}{}
\begin{exverb}
Tool \textasciigrave{}Zoom\_\allowbreak{}In\_\allowbreak{}Tool\textasciigrave{} (tool\_\allowbreak{}call\_\allowbreak{}id=\allowbreak{}call\_\allowbreak{}BAISDAjjqrtvjY61E2BI4mEj) returned:\\{}
Zoomed in image rendered below; refer to it as tool\_\allowbreak{}generated\_\allowbreak{}image\_\allowbreak{}4.
\end{exverb}
\begin{center}\includegraphics[width=0.45\linewidth,height=0.22\textheight,keepaspectratio]{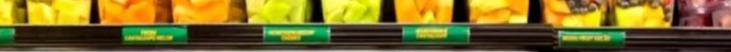}\\[2pt]
{\scriptsize\color{exMute} Image returned with this output.}\end{center}
\begin{exharness}
\exLabel{Harness}{environment update, summarized: new memory entries; new data bindings verbatim}
\begin{exverb}
Memory update: registers 1 new image in memory (tool\_\allowbreak{}generated\_\allowbreak{}image\_\allowbreak{}4,\allowbreak{} from Zoom\_\allowbreak{}In\_\allowbreak{}Tool).
\end{exverb}
\end{exharness}
\end{exstep}
\begin{exstep}{}{\exTool{exInk}{Final answer}}{}
\exWalk{The agent returns 53 boxes, and all 53 match a ground-truth cup. The trajectory does not say how the agent chose 53 of the 57 candidates.}
\exLabel{Model}{}
\begin{exverb}
\textless{}answer\textgreater{}[[0,\allowbreak{}0,\allowbreak{}26,\allowbreak{}59],\allowbreak{}[35,\allowbreak{}0,\allowbreak{}124,\allowbreak{}60],\allowbreak{}[135,\allowbreak{}0,\allowbreak{}234,\allowbreak{}59],\allowbreak{}[235,\allowbreak{}0,\allowbreak{}333,\allowbreak{}57],\allowbreak{}[336,\allowbreak{}0,\allowbreak{}425,\allowbreak{}58],\allowbreak{}[440,\allowbreak{}0,\allowbreak{}521,\allowbreak{}57],\allowbreak{}[539,\allowbreak{}0,\allowbreak{}619,\allowbreak{}56],\allowbreak{}[640,\allowbreak{}0,\allowbreak{}723,\allowbreak{}59],\allowbreak{}[740,\allowbreak{}0,\allowbreak{}829,\allowbreak{}68],\allowbreak{}[834,\allowbreak{}0,\allowbreak{}921,\allowbreak{}68],\allowbreak{}[925,\allowbreak{}0,\allowbreak{}1000,\allowbreak{}68],\allowbreak{}[0,\allowbreak{}234,\allowbreak{}28,\allowbreak{}387],\allowbreak{}[29,\allowbreak{}229,\allowbreak{}132,\allowbreak{}385],\allowbreak{}[138,\allowbreak{}234,\allowbreak{}234,\allowbreak{}386],\allowbreak{}[236,\allowbreak{}230,\allowbreak{}331,\allowbreak{}387],\allowbreak{}[334,\allowbreak{}220,\allowbreak{}428,\allowbreak{}384],\allowbreak{}[432,\allowbreak{}226,\allowbreak{}527,\allowbreak{}381],\allowbreak{}[533,\allowbreak{}225,\allowbreak{}626,\allowbreak{}380],\allowbreak{}[631,\allowbreak{}225,\allowbreak{}729,\allowbreak{}372],\allowbreak{}[738,\allowbreak{}209,\allowbreak{}830,\allowbreak{}371],\allowbreak{}[834,\allowbreak{}204,\allowbreak{}927,\allowbreak{}371],\allowbreak{}[940,\allowbreak{}237,\allowbreak{}1000,\allowbreak{}339],\allowbreak{}[0,\allowbreak{}390,\allowbreak{}25,\allowbreak{}527],\allowbreak{}[26,\allowbreak{}385,\allowbreak{}130,\allowbreak{}527],\allowbreak{}[140,\allowbreak{}386,\allowbreak{}233,\allowbreak{}527],\allowbreak{}[236,\allowbreak{}386,\allowbreak{}331,\allowbreak{}526],\allowbreak{}[332,\allowbreak{}385,\allowbreak{}432,\allowbreak{}526],\allowbreak{}[432,\allowbreak{}382,\allowbreak{}528,\allowbreak{}526],\allowbreak{}[534,\allowbreak{}381,\allowbreak{}625,\allowbreak{}525],\allowbreak{}[637,\allowbreak{}370,\allowbreak{}733,\allowbreak{}526],\allowbreak{}[738,\allowbreak{}367,\allowbreak{}823,\allowbreak{}525],\allowbreak{}[836,\allowbreak{}368,\allowbreak{}924,\allowbreak{}526],\allowbreak{}[937,\allowbreak{}336,\allowbreak{}1000,\allowbreak{}525],\allowbreak{}[1,\allowbreak{}668,\allowbreak{}105,\allowbreak{}871],\allowbreak{}[114,\allowbreak{}676,\allowbreak{}212,\allowbreak{}867],\allowbreak{}[227,\allowbreak{}676,\allowbreak{}335,\allowbreak{}873],\allowbreak{}[337,\allowbreak{}673,\allowbreak{}434,\allowbreak{}873],\allowbreak{}[445,\allowbreak{}674,\allowbreak{}541,\allowbreak{}868],\allowbreak{}[546,\allowbreak{}666,\allowbreak{}637,\allowbreak{}862],\allowbreak{}[640,\allowbreak{}670,\allowbreak{}731,\allowbreak{}859],\allowbreak{}[740,\allowbreak{}675,\allowbreak{}836,\allowbreak{}871],\allowbreak{}[842,\allowbreak{}676,\allowbreak{}936,\allowbreak{}872],\allowbreak{}[939,\allowbreak{}677,\allowbreak{}1000,\allowbreak{}870],\allowbreak{}[6,\allowbreak{}867,\allowbreak{}102,\allowbreak{}1000],\allowbreak{}[120,\allowbreak{}865,\allowbreak{}212,\allowbreak{}1000],\allowbreak{}[234,\allowbreak{}866,\allowbreak{}336,\allowbreak{}1000],\allowbreak{}[343,\allowbreak{}859,\allowbreak{}437,\allowbreak{}1000],\allowbreak{}[445,\allowbreak{}855,\allowbreak{}538,\allowbreak{}1000],\allowbreak{}[547,\allowbreak{}851,\allowbreak{}639,\allowbreak{}1000],\allowbreak{}[647,\allowbreak{}856,\allowbreak{}735,\allowbreak{}1000],\allowbreak{}[745,\allowbreak{}866,\allowbreak{}838,\allowbreak{}1000],\allowbreak{}[841,\allowbreak{}866,\allowbreak{}934,\allowbreak{}1000],\allowbreak{}[936,\allowbreak{}867,\allowbreak{}1000,\allowbreak{}1000]]\textless{}/\allowbreak{}answer\textgreater{}
\end{exverb}
\end{exstep}

%% file: text/appendix_examples_steps/smsp161_header.tex
\exOverview{\textbf{Task.} A street photograph from Paris Fashion Week shows a woman in a beige jacket and an orange scarf (faces are blurred in the benchmark image). Which brand's show was she attending, and which event and year was it? 

\textbf{Why it is hard.} Nothing in the image names the show; without tools the model guesses a Miu Miu show in 2024. 

\textbf{How the agent solves it.} A reverse image search finds no match, so the agent turns the photo into a text description and runs an image search with it. The top result is a street-style gallery titled ``What they wore to Hermès'' at Paris Fashion Week Fall/Winter 2025, and a page visit confirms the event and season. 

\textbf{Caveat.} The perception step contributes nothing: the agent zooms on what it takes for a shopping bag, but the OCR reads a Paris litter-bag slogan (``Ensemble rendons Paris propre''). The answer comes from search alone.}
\exLabel{Input images}{}
\begin{center}\includegraphics[width=0.55\linewidth,height=0.28\textheight,keepaspectratio]{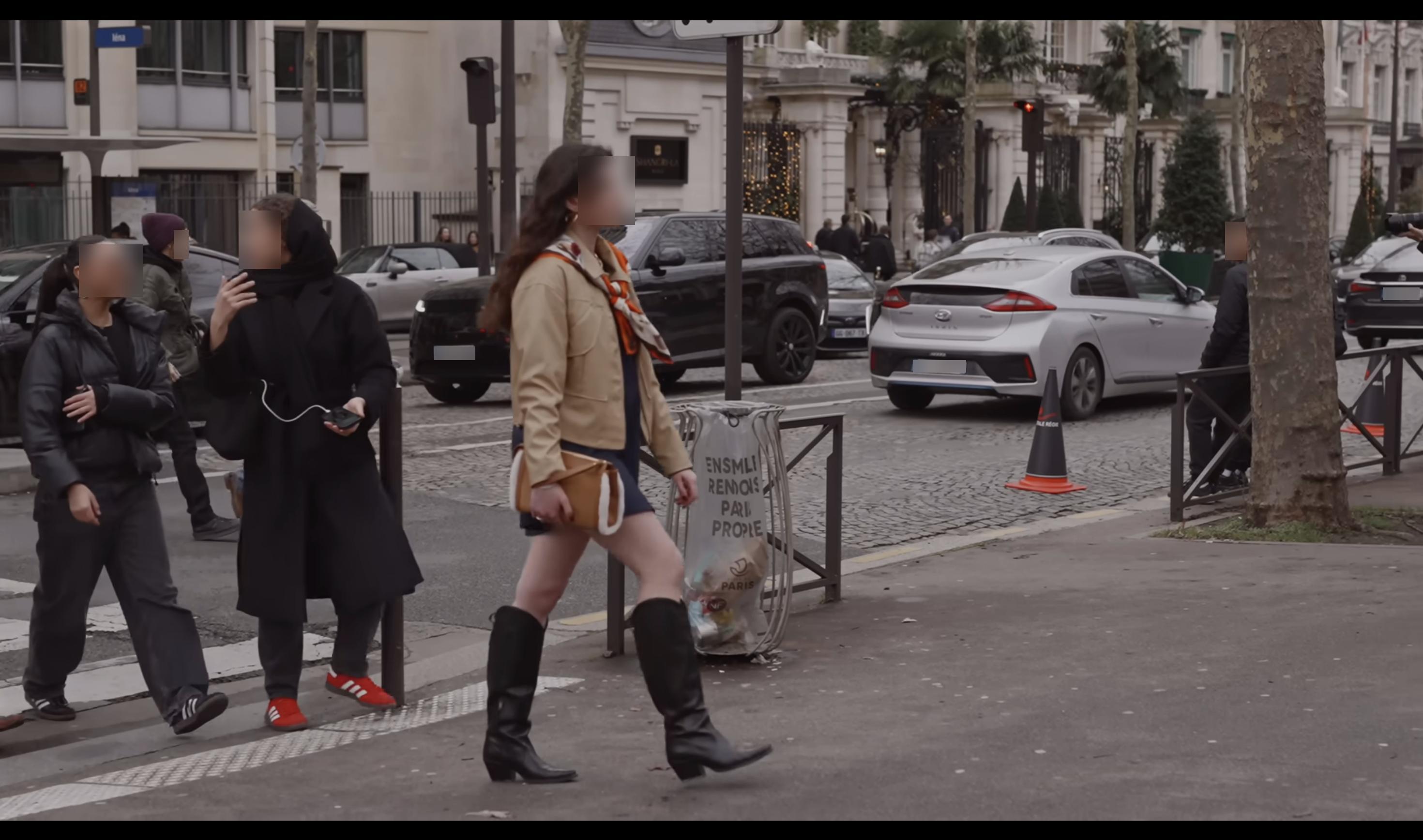}\\[2pt]
{\scriptsize\color{exMute} input\_image\_1}\end{center}
\exLabel{Question}{as sent to the model}
\begin{exverb}
Question: In Paris Fashion Week,\allowbreak{}  what brand did the person in the image represent? In which year did the event occur,\allowbreak{} and what is the event\textquotesingle{}s name?
\end{exverb}
\exLabel{Without tools}{GPT-6 Luna's full response to the same question}
\begin{exverb}
The brand is **Miu Miu**. The event took place in **2024** and was the **Miu Miu Spring/\allowbreak{}Summer 2025 show** during Paris Fashion Week.
\end{exverb}

%% file: text/appendix_examples_steps/smsp161.tex
\begin{exharness}
\exLabel{Harness}{environment update, shown in full}
\begin{exverb}
Environment Update (not a new question):\\{}
Currently available image IDs for tool calls:\\{}
- input\_\allowbreak{}image\_\allowbreak{}1: user\_\allowbreak{}input\\{}
When a tool needs image input,\allowbreak{} pass one of these IDs via the *\_\allowbreak{}image\_\allowbreak{}id field.\\{}
Do NOT invent image file names or paths.
\end{exverb}
\end{exharness}
\begin{exstep}{1}{\toolIS}{}
\exWalk{Reverse image search on the photograph returns no relevant matches.}
\exLabel{Arguments}{}
\begin{exverb}
\{\\{}
~~"\allowbreak{}image\_\allowbreak{}id"\allowbreak{}: "\allowbreak{}input\_\allowbreak{}image\_\allowbreak{}1"\allowbreak{}\\{}
\}
\end{exverb}
\exLabel{Model}{text emitted with this call}
\exSayBlock{I’ll use reverse-image search to identify the street-style photo or event source,\allowbreak{} which may reveal the represented brand and event details.}
\exLabel{Tool output}{}
\begin{exverb}
Tool \textasciigrave{}Image\_\allowbreak{}Search\_\allowbreak{}Tool\textasciigrave{} (tool\_\allowbreak{}call\_\allowbreak{}id=\allowbreak{}call\_\allowbreak{}x1uncz6TEwkbHACyGoWvL0AL) returned:\\{}
\{"\allowbreak{}ok"\allowbreak{}: true,\allowbreak{} "\allowbreak{}result"\allowbreak{}: "\allowbreak{}[Image Search — Visual Matches]\\{}
NO RELEVANT MATCHES\\{}
\mbox{}\\{}
Why each raw match was rejected:\\{}
All 10 candidates show studio product photos of jackets,\allowbreak{} which visually mismatch the QUERY image\textquotesingle{}s street scene. Their text metadata correctly describes the products but is irrelevant to the QUERY image\textquotesingle{}s context. No candidate image matches the QUERY image\textquotesingle{}s subject or environment.\\{}
\mbox{}\\{}
Suggested next actions:\\{}
- Image\_\allowbreak{}Search\_\allowbreak{}Tool(image\_\allowbreak{}id=\allowbreak{}input\_\allowbreak{}image\_\allowbreak{}1)\\{}
- Text\_\allowbreak{}Search\_\allowbreak{}Tool(query=\allowbreak{}\textquotesingle{}Paris street scene with woman in beige jacket\textquotesingle{},\allowbreak{} search\_\allowbreak{}type=\allowbreak{}images)"\allowbreak{},\allowbreak{} "\allowbreak{}error"\allowbreak{}: null\}
\end{exverb}
\end{exstep}
\begin{exstep}{2}{\toolTS}{}
\exWalk{The agent describes the outfit in words and runs an image search with that description. The top result is a street-style gallery from outside the Hermès show at Paris Fashion Week Fall/Winter 2025.}
\exLabel{Arguments}{}
\begin{exverb}
\{\\{}
~~"\allowbreak{}query"\allowbreak{}: "\allowbreak{}Paris Fashion Week street style woman beige jacket orange scarf paper shopping bag Paris black boots"\allowbreak{},\allowbreak{}\\{}
~~"\allowbreak{}search\_\allowbreak{}type"\allowbreak{}: "\allowbreak{}images"\allowbreak{}\\{}
\}
\end{exverb}
\exLabel{Tool output}{}
\begin{exverb}
Tool \textasciigrave{}Text\_\allowbreak{}Search\_\allowbreak{}Tool\textasciigrave{} (tool\_\allowbreak{}call\_\allowbreak{}id=\allowbreak{}call\_\allowbreak{}eGfMUoOe3PcNico2aru6bQVC) returned:\\{}
Image refs: tool\_\allowbreak{}generated\_\allowbreak{}image\_\allowbreak{}1,\allowbreak{} tool\_\allowbreak{}generated\_\allowbreak{}image\_\allowbreak{}2,\allowbreak{} tool\_\allowbreak{}generated\_\allowbreak{}image\_\allowbreak{}3\\{}
[Search Results — Type: images]\\{}
1. Paris Fashion Week FW 2025 - What they Wore to Hermes - FunkyForty\\{}
~~~Image: [tool\_\allowbreak{}generated\_\allowbreak{}image\_\allowbreak{}1]\\{}
~~~Source: FunkyForty   [webpage\_\allowbreak{}url\_\allowbreak{}1]\\{}
~~~Size: 1707x2560\\{}
2. Showgoers Embraced the Colors of Fall for Day 5 of Paris Fashion Week -  Fashionista\\{}
~~~Image: [tool\_\allowbreak{}generated\_\allowbreak{}image\_\allowbreak{}2]\\{}
~~~Source: Fashionista   [webpage\_\allowbreak{}url\_\allowbreak{}2]\\{}
~~~Size: 2560x3840
\end{exverb}
\exCut{1 more results not shown (3 in total).}
\begin{center}
\begin{minipage}[t]{0.475\linewidth}\centering\includegraphics[width=\linewidth,height=0.16\textheight,keepaspectratio]{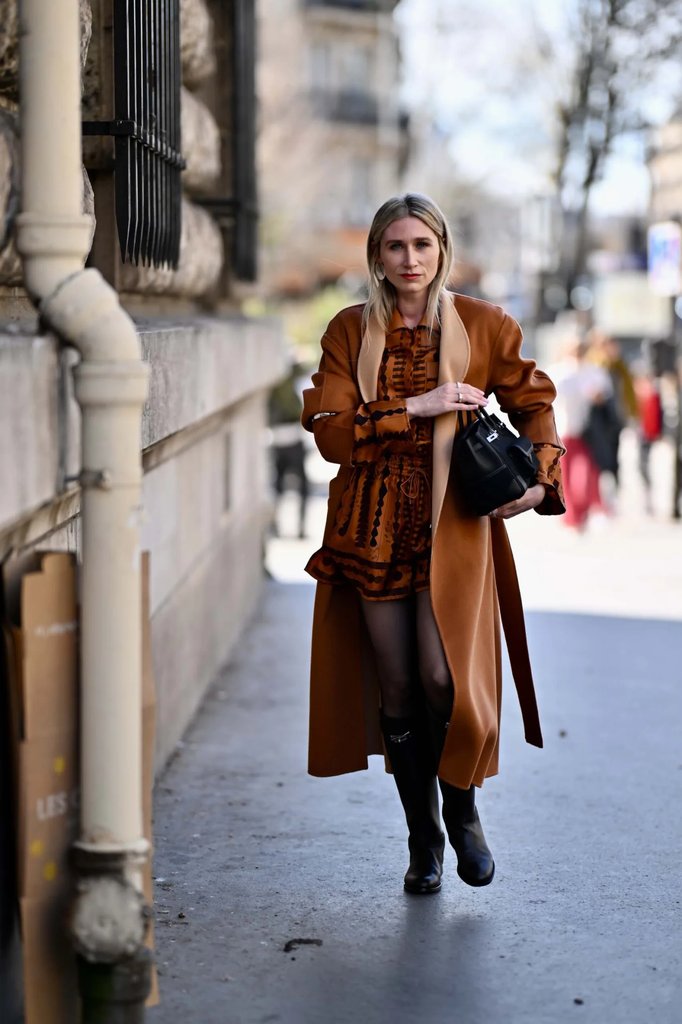}\end{minipage}\hfill\begin{minipage}[t]{0.475\linewidth}\centering\includegraphics[width=\linewidth,height=0.16\textheight,keepaspectratio]{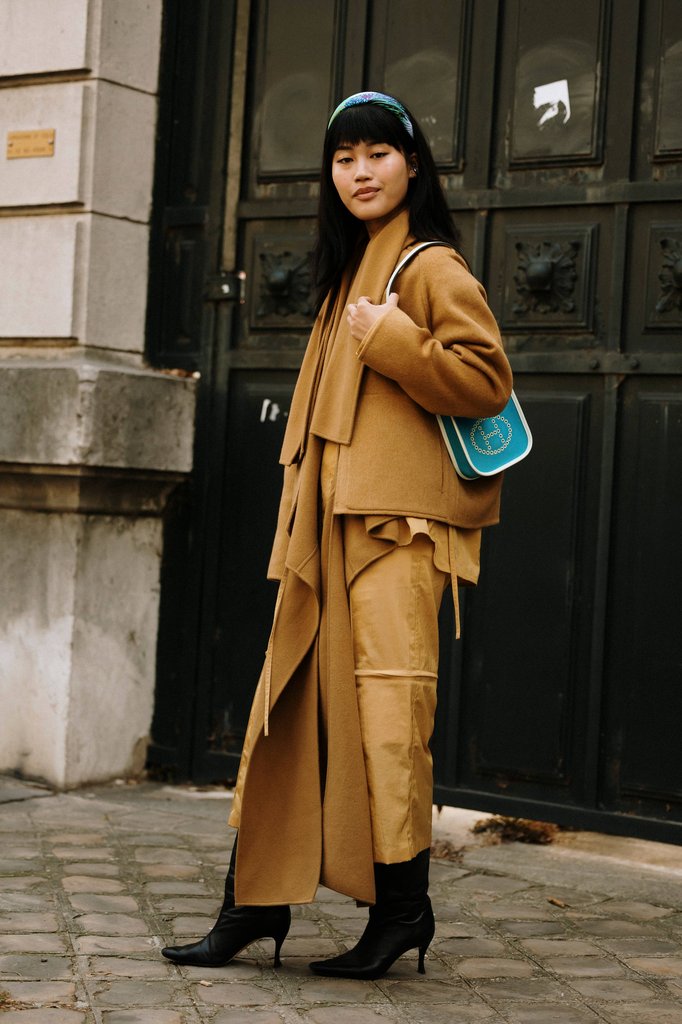}\end{minipage}
\\[1pt]{\scriptsize\color{exMute} Images returned with this output.}
\end{center}
\begin{exharness}
\exLabel{Harness}{environment update, summarized: new memory entries; new data bindings verbatim}
\begin{exverb}
Memory update: registers 3 new images in memory (tool\_\allowbreak{}generated\_\allowbreak{}image\_\allowbreak{}1 to tool\_\allowbreak{}generated\_\allowbreak{}image\_\allowbreak{}3,\allowbreak{} from Text\_\allowbreak{}Search\_\allowbreak{}Tool); registers 3 new webpage URLs in memory (webpage\_\allowbreak{}url\_\allowbreak{}1 to webpage\_\allowbreak{}url\_\allowbreak{}3,\allowbreak{} from Text\_\allowbreak{}Search\_\allowbreak{}Tool).
\end{exverb}
\end{exharness}
\end{exstep}
\begin{exstep}{3}{\toolZoom}{}
\exWalk{A zoom on the lettering behind the woman, which the agent takes to be her shopping bag.}
\exLabel{Arguments}{}
\begin{exverb}
\{\\{}
~~"\allowbreak{}image\_\allowbreak{}id"\allowbreak{}: "\allowbreak{}input\_\allowbreak{}image\_\allowbreak{}1"\allowbreak{},\allowbreak{}\\{}
~~"\allowbreak{}coords"\allowbreak{}: [468,\allowbreak{} 477,\allowbreak{} 551,\allowbreak{} 772],\allowbreak{}\\{}
~~"\allowbreak{}padding"\allowbreak{}: 70\\{}
\}
\end{exverb}
\exLabel{Model}{text emitted with this call}
\exSayBlock{I’ll inspect the lettering on the woman’s shopping bag; it could provide a clue to the brand associated with her.}
\exLabel{Tool output}{}
\begin{exverb}
Tool \textasciigrave{}Zoom\_\allowbreak{}In\_\allowbreak{}Tool\textasciigrave{} (tool\_\allowbreak{}call\_\allowbreak{}id=\allowbreak{}call\_\allowbreak{}7WEex9earQgiebwQJQxRu5lz) returned:\\{}
Zoomed in image rendered below; refer to it as tool\_\allowbreak{}generated\_\allowbreak{}image\_\allowbreak{}4.
\end{exverb}
\begin{center}\includegraphics[width=0.45\linewidth,height=0.22\textheight,keepaspectratio]{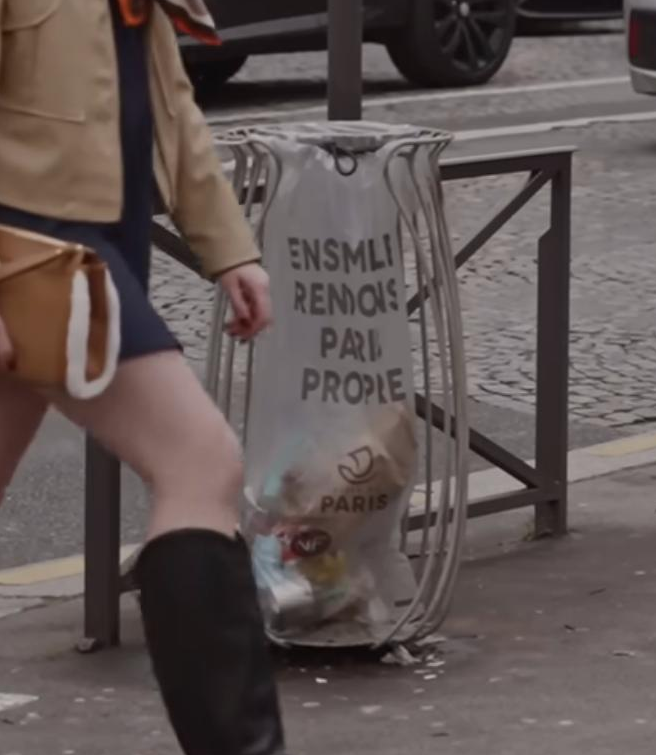}\\[2pt]
{\scriptsize\color{exMute} Image returned with this output.}\end{center}
\begin{exharness}
\exLabel{Harness}{environment update, summarized: new memory entries; new data bindings verbatim}
\begin{exverb}
Memory update: registers 1 new image in memory (tool\_\allowbreak{}generated\_\allowbreak{}image\_\allowbreak{}4,\allowbreak{} from Zoom\_\allowbreak{}In\_\allowbreak{}Tool).
\end{exverb}
\end{exharness}
\end{exstep}
\begin{exstep}{4}{\toolOCR}{}
\exWalk{OCR reads the lettering, a municipal slogan on a Paris litter bag, not a brand.}
\exLabel{Arguments}{}
\begin{exverb}
\{\\{}
~~"\allowbreak{}image\_\allowbreak{}id"\allowbreak{}: "\allowbreak{}tool\_\allowbreak{}generated\_\allowbreak{}image\_\allowbreak{}4"\allowbreak{}\\{}
\}
\end{exverb}
\exLabel{Tool output}{}
\begin{exverb}
OCR\_\allowbreak{}Tool transcribed the following text from the image:\\{}
\mbox{}\\{}
ENSMLI\\{}
RENOIS\\{}
PARI\\{}
PROPIE\\{}
PARIS
\end{exverb}
\end{exstep}
\begin{exstep}{5}{\toolWV}{}
\exWalk{A visit to the gallery page confirms the event: street style at the Hermès show, Paris Fashion Week Fall/Winter 2025.}
\exLabel{Arguments}{}
\begin{exverb}
\{\\{}
~~"\allowbreak{}urls"\allowbreak{}: ["\allowbreak{}webpage\_\allowbreak{}url\_\allowbreak{}1"\allowbreak{}],\allowbreak{}\\{}
~~"\allowbreak{}goal"\allowbreak{}: "\allowbreak{}Find whether the matching woman is shown and identify her outfit brand/\allowbreak{}represented brand,\allowbreak{} the Paris Fashion Week year and event name/\allowbreak{}season."\allowbreak{}\\{}
\}
\end{exverb}
\exLabel{Model}{text emitted with this call}
\exSayBlock{The person was attending the Hermès show. The event was Paris Fashion Week Fall/\allowbreak{}Winter 2025.}
\exLabel{Tool output}{}
\begin{exverb}
Tool \textasciigrave{}Webpage\_\allowbreak{}Visit\_\allowbreak{}Tool\textasciigrave{} (tool\_\allowbreak{}call\_\allowbreak{}id=\allowbreak{}call\_\allowbreak{}ax7zd7wE1WPQ3f3IsEkgRq9v) returned:\\{}
\{"\allowbreak{}ok"\allowbreak{}: true,\allowbreak{} "\allowbreak{}result"\allowbreak{}: "\allowbreak{}The useful information in https:/\allowbreak{}/\allowbreak{}funkyforty.com/\allowbreak{}paris-fashion-week-fw-2025-what-they-wore-to-hermes/\allowbreak{} for user goal Find whether the matching woman is shown and identify her outfit brand/\allowbreak{}represented brand,\allowbreak{} the Paris Fashion Week year and event name/\allowbreak{}season. as follows: \\{}
\mbox{}\\{}
Evidence in page: \\{}
Paris Fashion Week FW 2025 - What they Wore to Hermes. The article discusses street style at Hermes during Paris Fashion Week FW 2025,\allowbreak{} including styling ideas and bag trends,\allowbreak{} with photos taken by Yvonne Aeberhard Stutz. The event is clearly identified as Paris Fashion Week Fall/\allowbreak{}Winter 2025.\\{}
\mbox{}\\{}
Summary: \\{}
The webpage documents street style at Paris Fashion Week Fall/\allowbreak{}Winter 2025,\allowbreak{} specifically focusing on what attendees wore to the Hermes show. While it does not explicitly name a specific woman,\allowbreak{} it references the event,\allowbreak{} season,\allowbreak{} and brand (Hermes) as requested in the goal.\\{}
\mbox{}\\{}
"\allowbreak{},\allowbreak{} "\allowbreak{}error"\allowbreak{}: null\}
\end{exverb}
\end{exstep}

%% file: text/appendix_examples_steps/mmsp_header.tex
\exOverview{\textbf{Task.} The question, in Chinese, asks: how many years apart are the famous Great-Depression-era event commemorated in the first image and the event shown in the second? The first image is a rusted steel monument in a park; the second is a black-and-white photograph of a fight. 

\textbf{How the agent solves it.} It identifies each image with reverse image search and dates each event from the web: the Ford Hunger March of 7 March 1932, and the Ford Strikers Riot of 3 April 1941. One page visit fails with a server error, and the agent recovers with a text search. It answers 9 years. The model reasons in Chinese throughout.}
\exLabel{Input images}{}
\begin{center}
\begin{minipage}[t]{0.475\linewidth}\centering\includegraphics[width=\linewidth,height=0.22\textheight,keepaspectratio]{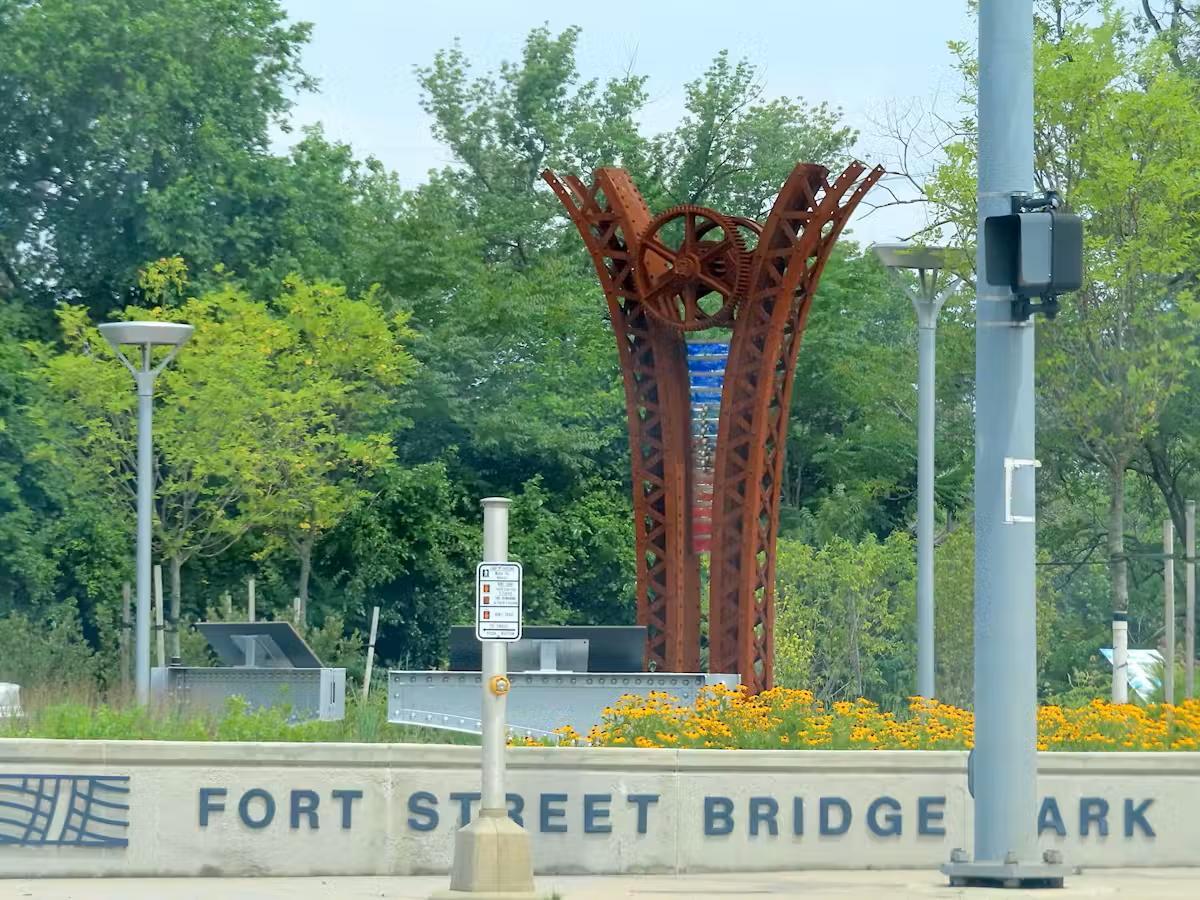}\\[1pt]{\scriptsize\color{exMute} input\_image\_1}\end{minipage}\hfill\begin{minipage}[t]{0.475\linewidth}\centering\includegraphics[width=\linewidth,height=0.22\textheight,keepaspectratio]{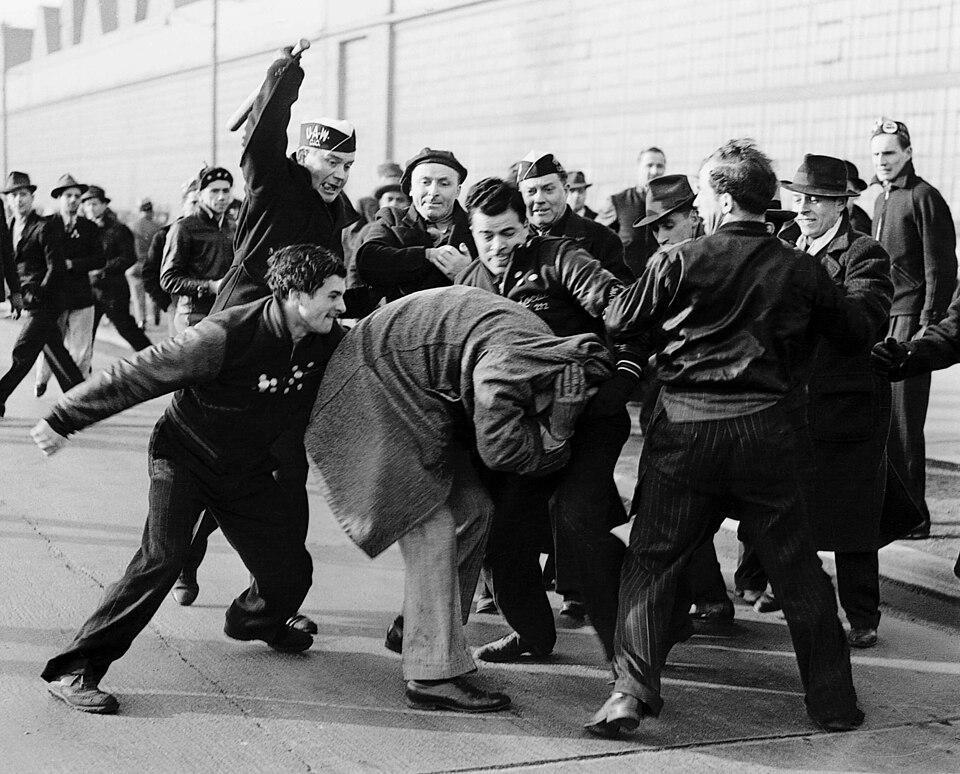}\\[1pt]{\scriptsize\color{exMute} input\_image\_2}\end{minipage}
\end{center}
\exLabel{Question}{as sent to the model}
\begin{exverb}
\begin{CJK}{UTF8}{gbsn}Question: 图1中发生在大萧条时代的著名历史事件，和图2中的事件相隔多少年？\end{CJK}
\end{exverb}

%% file: text/appendix_examples_steps/mmsp.tex
\begin{exharness}
\exLabel{Harness}{environment update, shown in full}
\begin{exverb}
Environment Update (not a new question):\\{}
Currently available image IDs for tool calls:\\{}
- input\_\allowbreak{}image\_\allowbreak{}1: user\_\allowbreak{}input\\{}
- input\_\allowbreak{}image\_\allowbreak{}2: user\_\allowbreak{}input\\{}
When a tool needs image input,\allowbreak{} pass one of these IDs via the *\_\allowbreak{}image\_\allowbreak{}id field.\\{}
Do NOT invent image file names or paths.
\end{exverb}
\end{exharness}
\begin{exstep}{1}{\toolIS}{}
\exLabel{Model}{text emitted with this call}
\exSayBlock{\begin{CJK}{UTF8}{gbsn}\textless{}thinking\textgreater{}我需要先识别图1中的地点和相关的历史事件。从图1的墙上的文字“FORT STREET BRIDGE PARK”可以推断出这是位于美国密歇根州底特律市的一个公园。这个公园与底特律在工业时代的历史有关。然后，我需要确定图2中的历史事件是什么，并计算两个事件之间的时间间隔。

首先，我会使用图像搜索工具来确认图1中的雕塑或地标是否与某个特定的历史事件相关联。
\textless{}/\allowbreak{}thinking\textgreater{}\end{CJK}}
\exLabel{Arguments}{}
\begin{exverb}
\{\\{}
~~"\allowbreak{}image\_\allowbreak{}id"\allowbreak{}: "\allowbreak{}input\_\allowbreak{}image\_\allowbreak{}1"\allowbreak{}\\{}
\}
\end{exverb}
\exLabel{Tool output}{}
\begin{exverb}
Tool \textasciigrave{}Image\_\allowbreak{}Search\_\allowbreak{}Tool\textasciigrave{} (tool\_\allowbreak{}call\_\allowbreak{}id=\allowbreak{}chatcmpl-tool-89abdf6fe5e37463) returned:\\{}
Image refs: tool\_\allowbreak{}generated\_\allowbreak{}image\_\allowbreak{}1,\allowbreak{} tool\_\allowbreak{}generated\_\allowbreak{}image\_\allowbreak{}2,\allowbreak{} tool\_\allowbreak{}generated\_\allowbreak{}image\_\allowbreak{}3\\{}
[Image Search — Visual Matches]\\{}
1. Fort Street Bridge Park chronicles labor and environmental history\\{}
~~~Source: planetdetroit.org/\allowbreak{}2024/\allowbreak{}03/\allowbreak{}fort-street-bridge-park   [webpage\_\allowbreak{}url\_\allowbreak{}1]\\{}
~~~Image: [tool\_\allowbreak{}generated\_\allowbreak{}image\_\allowbreak{}1]\\{}
2. MotorCities - Fort Street Bridge Interpretive Park\\{}
~~~Source: motorcities.org/\allowbreak{}fortstreet   [webpage\_\allowbreak{}url\_\allowbreak{}2]\\{}
~~~Image: [tool\_\allowbreak{}generated\_\allowbreak{}image\_\allowbreak{}2]
\end{exverb}
\exCut{1 more results not shown (3 in total).}
\begin{center}
\begin{minipage}[t]{0.312\linewidth}\centering\includegraphics[width=\linewidth,height=0.16\textheight,keepaspectratio]{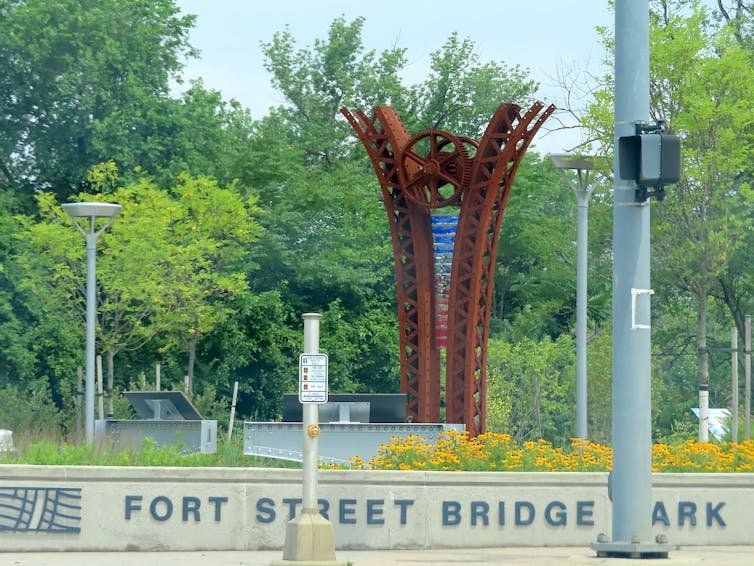}\end{minipage}\hfill\begin{minipage}[t]{0.312\linewidth}\centering\includegraphics[width=\linewidth,height=0.16\textheight,keepaspectratio]{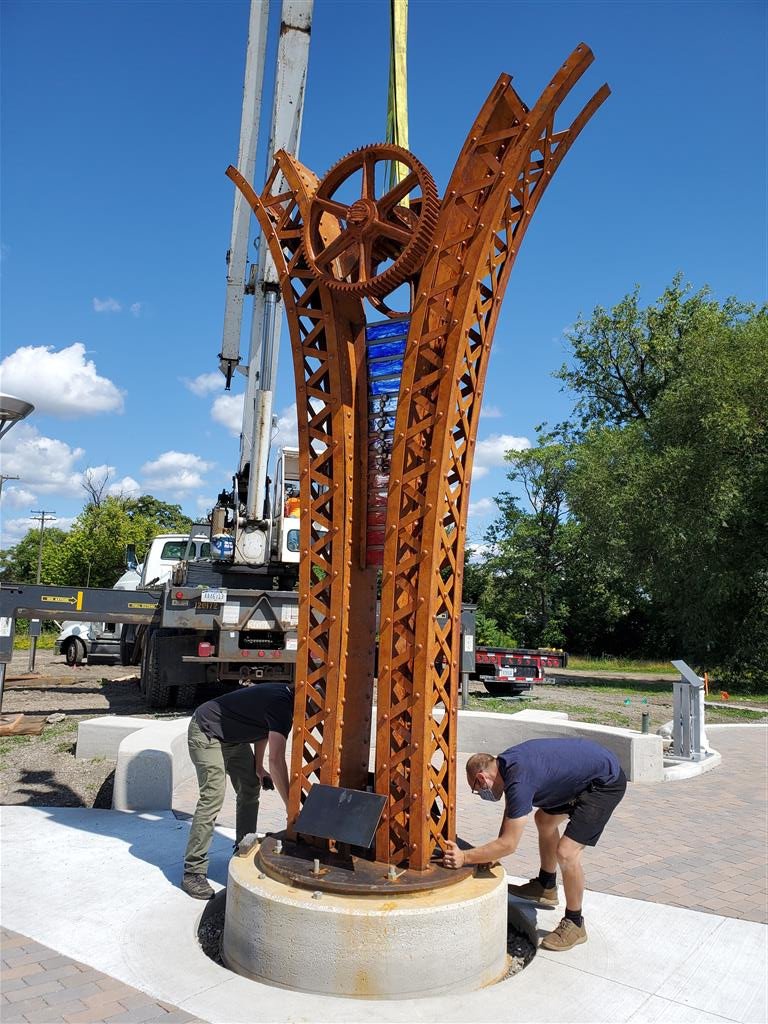}\end{minipage}\hfill\begin{minipage}[t]{0.312\linewidth}\centering\includegraphics[width=\linewidth,height=0.16\textheight,keepaspectratio]{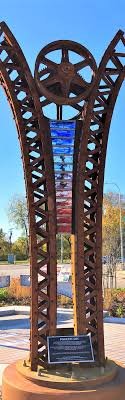}\end{minipage}
\\[1pt]{\scriptsize\color{exMute} Images returned with this output.}
\end{center}
\begin{exharness}
\exLabel{Harness}{environment update, summarized: new memory entries; new data bindings verbatim}
\begin{exverb}
Memory update: registers 3 new images in memory (tool\_\allowbreak{}generated\_\allowbreak{}image\_\allowbreak{}1 to tool\_\allowbreak{}generated\_\allowbreak{}image\_\allowbreak{}3,\allowbreak{} from Image\_\allowbreak{}Search\_\allowbreak{}Tool); registers 3 new webpage URLs in memory (webpage\_\allowbreak{}url\_\allowbreak{}1 to webpage\_\allowbreak{}url\_\allowbreak{}3,\allowbreak{} from Image\_\allowbreak{}Search\_\allowbreak{}Tool).
\end{exverb}
\end{exharness}
\end{exstep}
\begin{exstep}{2}{\toolWV}{}
\exLabel{Model}{text emitted with this call}
\exSayBlock{\begin{CJK}{UTF8}{gbsn}\textless{}thinking\textgreater{}根据图像搜索结果，图1显示的是位于底特律的Fort Street Bridge Park的雕塑。该公园纪念了底特律的劳工和环境历史。为了找到与大萧条时期相关的具体历史事件，我将访问其中一个来源网页以获取更多信息。
\textless{}/\allowbreak{}thinking\textgreater{}\end{CJK}}
\exLabel{Arguments}{}
\begin{exverb}
\{\\{}
~~"\allowbreak{}urls"\allowbreak{}: ["\allowbreak{}webpage\_\allowbreak{}url\_\allowbreak{}1"\allowbreak{}],\allowbreak{}\\{}
~~"\allowbreak{}goal"\allowbreak{}: "\allowbreak{}Find out which historical event during the Great Depression is commemorated at Fort Street Bridge Park."\allowbreak{}\\{}
\}
\end{exverb}
\exLabel{Tool output}{}
\begin{exverb}
Tool \textasciigrave{}Webpage\_\allowbreak{}Visit\_\allowbreak{}Tool\textasciigrave{} (tool\_\allowbreak{}call\_\allowbreak{}id=\allowbreak{}chatcmpl-tool-bbf896e936550bcb) returned:\\{}
\{"\allowbreak{}ok"\allowbreak{}: true,\allowbreak{} "\allowbreak{}result"\allowbreak{}: "\allowbreak{}The useful information in https:/\allowbreak{}/\allowbreak{}planetdetroit.org/\allowbreak{}2024/\allowbreak{}03/\allowbreak{}fort-street-bridge-park/\allowbreak{} for user goal Find out which historical event during the Great Depression is commemorated at Fort Street Bridge Park. as follows: \textbackslash{}n\textbackslash{}nEvidence in page: \textbackslash{}nSigns explain why this sculpture and park are here: to honor the memory of protesters who met on this very spot on March 7,\allowbreak{} 1932,\allowbreak{} before marching up Miller Road to the massive Ford Rouge River Complex located in the adjacent city of Dearborn. ... This crowd of several thousand marched up the road on one of the coldest days of winter. They were greeted at the Dearborn border with clouds of tear gas,\allowbreak{} jets of cold water and a shower of bullets. It was then that the Ford Hunger March became the Ford Massacre.\textbackslash{}n\textbackslash{}nSummary: \textbackslash{}nFort Street Bridge Park commemorates the Ford Hunger March of March 7,\allowbreak{} 1932,\allowbreak{} a pivotal labor protest during the Great Depression in which unemployed Ford workers marched to demand better working conditions and an end to discrimination,\allowbreak{} only to be violently suppressed by Ford security and Dearborn police — an event that became known as the Ford Massacre.\textbackslash{}n\textbackslash{}n"\allowbreak{},\allowbreak{} "\allowbreak{}error"\allowbreak{}: null\}
\end{exverb}
\end{exstep}
\begin{exstep}{3}{\toolIS}{}
\exLabel{Model}{text emitted with this call}
\exSayBlock{\begin{CJK}{UTF8}{gbsn}\textless{}thinking\textgreater{}现在我知道了图1中的事件是1932年的福特饥饿游行（Ford Hunger March），也被称为福特大屠杀（Ford Massacre）。接下来，我需要分析图2中的事件并确定其发生时间。图2是一张黑白照片，描绘了一群人正在打斗，其中一人举着棍子。这看起来像一个历史性的暴力事件。我将使用图像搜索工具来识别这张图片所代表的具体事件。
\textless{}/\allowbreak{}thinking\textgreater{}\end{CJK}}
\exLabel{Arguments}{}
\begin{exverb}
\{\\{}
~~"\allowbreak{}image\_\allowbreak{}id"\allowbreak{}: "\allowbreak{}input\_\allowbreak{}image\_\allowbreak{}2"\allowbreak{}\\{}
\}
\end{exverb}
\exLabel{Tool output}{}
\begin{exverb}
Tool \textasciigrave{}Image\_\allowbreak{}Search\_\allowbreak{}Tool\textasciigrave{} (tool\_\allowbreak{}call\_\allowbreak{}id=\allowbreak{}chatcmpl-tool-924377b36e8efdaf) returned:\\{}
Image refs: tool\_\allowbreak{}generated\_\allowbreak{}image\_\allowbreak{}4,\allowbreak{} tool\_\allowbreak{}generated\_\allowbreak{}image\_\allowbreak{}5,\allowbreak{} tool\_\allowbreak{}generated\_\allowbreak{}image\_\allowbreak{}6\\{}
[Image Search — Visual Matches]\\{}
1. Ford Strikers Riot - A black and white photo of a group of men fighting - PICRYL - Public Domain Media Search Engine Public Domain Search\\{}
~~~Source: picryl.com/\allowbreak{}media/\allowbreak{}ford-strikers-riot-a4a7fb   [webpage\_\allowbreak{}url\_\allowbreak{}4]\\{}
~~~Image: [tool\_\allowbreak{}generated\_\allowbreak{}image\_\allowbreak{}4]\\{}
2. Moments The Pulitzer Prize Photographs 1942-1977 Sheryle John Leekley Vtg Book - very-good · Whatnot: Shop,\allowbreak{} Sell,\allowbreak{} Connect\\{}
~~~Source: whatnot.com/\allowbreak{}listing/\allowbreak{}TGlzdGluZ05vZGU6NzAzNzkzMTEy   [webpage\_\allowbreak{}url\_\allowbreak{}5]\\{}
~~~Image: [tool\_\allowbreak{}generated\_\allowbreak{}image\_\allowbreak{}5]
\end{exverb}
\exCut{1 more results not shown (3 in total).}
\begin{center}
\begin{minipage}[t]{0.312\linewidth}\centering\includegraphics[width=\linewidth,height=0.16\textheight,keepaspectratio]{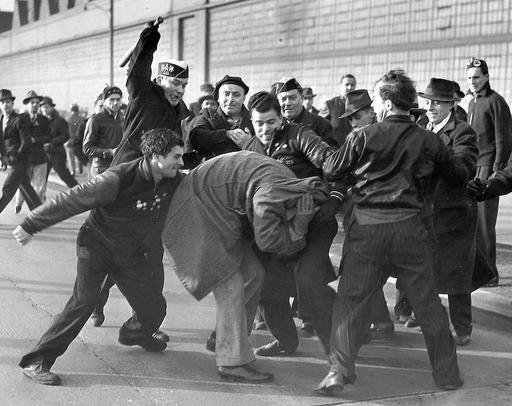}\end{minipage}\hfill\begin{minipage}[t]{0.312\linewidth}\centering\includegraphics[width=\linewidth,height=0.16\textheight,keepaspectratio]{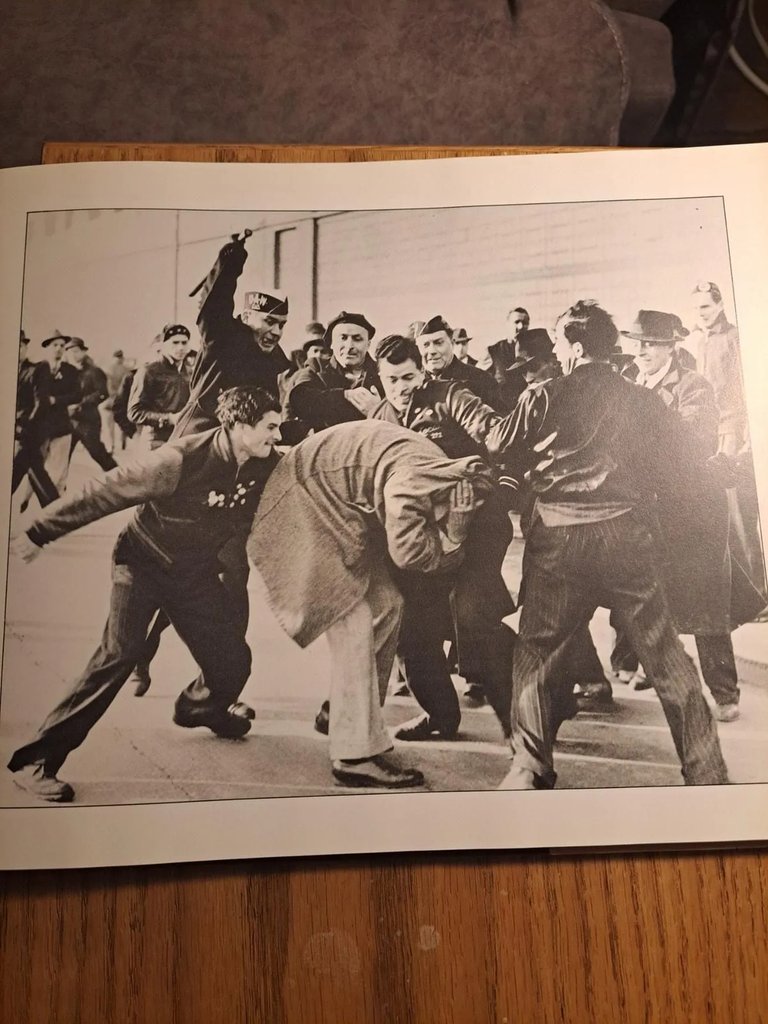}\end{minipage}\hfill\begin{minipage}[t]{0.312\linewidth}\centering\includegraphics[width=\linewidth,height=0.16\textheight,keepaspectratio]{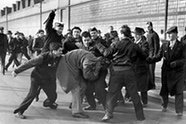}\end{minipage}
\\[1pt]{\scriptsize\color{exMute} Images returned with this output.}
\end{center}
\begin{exharness}
\exLabel{Harness}{environment update, summarized: new memory entries; new data bindings verbatim}
\begin{exverb}
Memory update: registers 3 new images in memory (tool\_\allowbreak{}generated\_\allowbreak{}image\_\allowbreak{}4 to tool\_\allowbreak{}generated\_\allowbreak{}image\_\allowbreak{}6,\allowbreak{} from Image\_\allowbreak{}Search\_\allowbreak{}Tool); registers 3 new webpage URLs in memory (webpage\_\allowbreak{}url\_\allowbreak{}4 to webpage\_\allowbreak{}url\_\allowbreak{}6,\allowbreak{} from Image\_\allowbreak{}Search\_\allowbreak{}Tool).
\end{exverb}
\end{exharness}
\end{exstep}
\begin{exstep}{4}{\toolWV}{}
\exLabel{Model}{text emitted with this call}
\exSayBlock{\begin{CJK}{UTF8}{gbsn}\textless{}thinking\textgreater{}图像搜索结果显示，图2中的事件被标识为“福特罢工者骚乱”（Ford Strikers Riot）。为了准确确定这一事件的发生时间，我需要查阅更详细的信息。我将访问PICRYL网站上提供的链接，因为它是直接描述该事件的可靠来源。
\textless{}/\allowbreak{}thinking\textgreater{}\end{CJK}}
\exLabel{Arguments}{}
\begin{exverb}
\{\\{}
~~"\allowbreak{}urls"\allowbreak{}: ["\allowbreak{}webpage\_\allowbreak{}url\_\allowbreak{}4"\allowbreak{}],\allowbreak{}\\{}
~~"\allowbreak{}goal"\allowbreak{}: "\allowbreak{}Determine the exact date of the Ford Strikers Riot shown in the image."\allowbreak{}\\{}
\}
\end{exverb}
\exLabel{Tool output}{}
\begin{exverb}
Tool \textasciigrave{}Webpage\_\allowbreak{}Visit\_\allowbreak{}Tool\textasciigrave{} (tool\_\allowbreak{}call\_\allowbreak{}id=\allowbreak{}chatcmpl-tool-a3db75968c467ffc) returned:\\{}
\{"\allowbreak{}ok"\allowbreak{}: true,\allowbreak{} "\allowbreak{}result"\allowbreak{}: "\allowbreak{}The useful information in https:/\allowbreak{}/\allowbreak{}picryl.com/\allowbreak{}media/\allowbreak{}ford-strikers-riot-a4a7fb for user goal Determine the exact date of the Ford Strikers Riot shown in the image. as follows: \textbackslash{}n\textbackslash{}nEvidence in page: \textbackslash{}nFailed to fetch https:/\allowbreak{}/\allowbreak{}picryl.com/\allowbreak{}media/\allowbreak{}ford-strikers-riot-a4a7fb (HTTP 500,\allowbreak{} transient) after 3 attempts Detail: 500 Server Error: Internal Server Error for url: https:/\allowbreak{}/\allowbreak{}scrape.serper.dev/\allowbreak{}. Suggestion: source temporarily unavailable; try again later or pick a different URL.\textbackslash{}n\textbackslash{}nSummary: \textbackslash{}nFETCH FAILED.\textbackslash{}n\textbackslash{}nReason: Failed to fetch https:/\allowbreak{}/\allowbreak{}picryl.com/\allowbreak{}media/\allowbreak{}ford-strikers-riot-a4a7fb (HTTP 500,\allowbreak{} transient) after 3 attempts Detail: 500 Server Error: Internal Server Error for url: https:/\allowbreak{}/\allowbreak{}scrape.serper.dev/\allowbreak{}. Suggestion: source temporarily unavailable; try again later or pick a different URL.\textbackslash{}nDiagnosis: Transient failure (5xx /\allowbreak{} 429 /\allowbreak{} timeout) after 3 retries. The source MAY be back online in a minute; retrying this URL once more after a short pause is reasonable. If it fails again,\allowbreak{} switch to a different source.\textbackslash{}n\textbackslash{}nSuggested next actions for the agent:\textbackslash{}n\textbackslash{}n  - Tool switches (the agent may also re-call this same Webpage\_\allowbreak{}Visit\_\allowbreak{}Tool on a DIFFERENT URL — nothing is excluded just because the current URL failed):\textbackslash{}n    - Webpage\_\allowbreak{}Visit\_\allowbreak{}Tool(urls=\allowbreak{}[\textless{}a different webpage\_\allowbreak{}url\_\allowbreak{}N\textgreater{}],\allowbreak{} goal=\allowbreak{}...)\textbackslash{}n        Try a different URL from the same search results. If the prior search had multiple results and only th
\end{exverb}
\exCut{The remaining 541 characters of this output are not shown (2,041 in total).}
\end{exstep}
\begin{exstep}{5}{\toolTS}{}
\exLabel{Model}{text emitted with this call}
\exSayBlock{\begin{CJK}{UTF8}{gbsn}\textless{}thinking\textgreater{}由于无法访问PICRYL网站，我需要尝试另一个来源来确定福特罢工者骚乱的确切日期。我将使用文本搜索工具来查找关于“福特罢工者骚乱”的信息，特别是它的发生时间。
\textless{}/\allowbreak{}thinking\textgreater{}\end{CJK}}
\exLabel{Arguments}{}
\begin{exverb}
\{\\{}
~~"\allowbreak{}query"\allowbreak{}: "\allowbreak{}Ford Strikers Riot date"\allowbreak{},\allowbreak{}\\{}
~~"\allowbreak{}search\_\allowbreak{}type"\allowbreak{}: "\allowbreak{}search"\allowbreak{}\\{}
\}
\end{exverb}
\exLabel{Tool output}{}
\begin{exverb}
Tool \textasciigrave{}Text\_\allowbreak{}Search\_\allowbreak{}Tool\textasciigrave{} (tool\_\allowbreak{}call\_\allowbreak{}id=\allowbreak{}chatcmpl-tool-9b84597577bb0a7f) returned:\\{}
\{"\allowbreak{}ok"\allowbreak{}: true,\allowbreak{} "\allowbreak{}result"\allowbreak{}: "\allowbreak{}[Search Results — Type: search]\textbackslash{}n1. File:Ford Strikers Riot.jpg - Wikimedia Commons\textbackslash{}n   Source: commons.wikimedia.org/\allowbreak{}wiki/\allowbreak{}File:Ford\_\allowbreak{}Strikers\_\allowbreak{}Riot.jpg   [tool\_\allowbreak{}generated\_\allowbreak{}image\_\allowbreak{}7]\textbackslash{}n   Date,\allowbreak{} Taken on 3 April 1941. Source,\allowbreak{} The Detroit News,\allowbreak{} The legacy of Henry ... File:Ford Strikers Riot.png · Category:1941-04-03. File usage on other wikis.\textbackslash{}n2. Unemployed Detroit auto workers conduct Hunger March to protest ...\textbackslash{}n   Source: nvdatabase.swarthmore.edu/\allowbreak{}content/\allowbreak{}unemployed-detroit-auto-workers-conduct-hun...   [webpage\_\allowbreak{}url\_\allowbreak{}7]\textbackslash{}n   On 1 April 1941,\allowbreak{} Ford workers conducted a walkout in order to protest the firing of eight union members. Because of the strike,\allowbreak{} the River Rouge plant was ...\textbackslash{}n3. Ford Strikers Riot - Wikipedia\textbackslash{}n   Source: en.wikipedia.org/\allowbreak{}wiki/\allowbreak{}Ford\_\allowbreak{}Strikers\_\allowbreak{}Riot   [webpage\_\allowbreak{}url\_\allowbreak{}8]\textbackslash{}n   Ford Strikers Riot is a 1941 photograph which shows a strikebreaker getting beaten by United Auto Workers (UAW) strikers.\textbackslash{}n\textbackslash{}n[Summary]\textbackslash{}nThe Ford Strikers Riot occurred on April 3,\allowbreak{} 1941,\allowbreak{} as documented in a photograph taken by Milton Brooks and published by The Detroit News,\allowbreak{} capturing a strikebreaker being beaten by United Auto Workers strikers. A strike began on April 1,\allowbreak{} 1941,\allowbreak{} after Ford fired eight union members,\allowbreak{} leading to a walkout and escalating tensions. The strike lasted about ten days,\allowbreak{} culminating in violence at the Ford River Rouge Plant,\allowbreak{} with strikers using around 2,\allowbreak{}000 vehicles to block access to the plant. The event is also referenced in Wikipedia and Wikimedia Commons as a pivotal moment in American labor history,\allowbreak{} with the photograph later winning a Pulitzer Prize for Photography."\allowbreak{},\allowbreak{} "\allowbreak{}error"\allowbreak{}: null\}
\end{exverb}
\begin{exharness}
\exLabel{Harness}{environment update, summarized: new memory entries; new data bindings verbatim}
\begin{exverb}
Memory update: registers 1 new image in memory (tool\_\allowbreak{}generated\_\allowbreak{}image\_\allowbreak{}7,\allowbreak{} from Text\_\allowbreak{}Search\_\allowbreak{}Tool); registers 2 new webpage URLs in memory (webpage\_\allowbreak{}url\_\allowbreak{}7 to webpage\_\allowbreak{}url\_\allowbreak{}8,\allowbreak{} from Text\_\allowbreak{}Search\_\allowbreak{}Tool).
\end{exverb}
\end{exharness}
\end{exstep}
\begin{exstep}{}{\exTool{exInk}{Final answer}}{}
\exLabel{Model}{}
\exSayBlock{\begin{CJK}{UTF8}{gbsn}\textless{}thinking\textgreater{}现在我已经知道：

- 图1中的事件是**福特饥饿游行（Ford Hunger March）**，发生在 **1932年3月7日**。
- 图2中的事件是**福特罢工者骚乱（Ford Strikers Riot）**，发生在 **1941年4月3日**。

因此，这两个事件之间的年份差可以通过简单的减法计算得出：1941 - 1932 =\allowbreak{} 9年。

所以，图1和图2中的事件相隔9年。
\textless{}/\allowbreak{}thinking\textgreater{}

\textless{}answer\textgreater{}9年\textless{}/\allowbreak{}answer\textgreater{}\end{CJK}}
\end{exstep}

%% file: text/appendix_examples_steps/srx140_header.tex
\exOverview{\textbf{Task.} A photograph shows a crowded skincare shelf. Which brand does the item at the far right of the top row belong to? The item is a tall white spray can whose label is far too small to read. 

\textbf{Why it is hard.} Nothing legible identifies the product; without tools the model guesses AHC. 

\textbf{How the agent solves it.} Zooms and OCR on the top shelf return only book spines and dates, and a reverse search of the whole shelf returns generic ``shelfie'' boards. The agent goes back to the image: it zooms on the bottles again, grounds them to see where each one sits, crops tightly, and reverse-searches the crop, which matches Miyuequan (Miyue Spring) hydrating spray.}
\exLabel{Input images}{}
\begin{center}\includegraphics[width=0.55\linewidth,height=0.28\textheight,keepaspectratio]{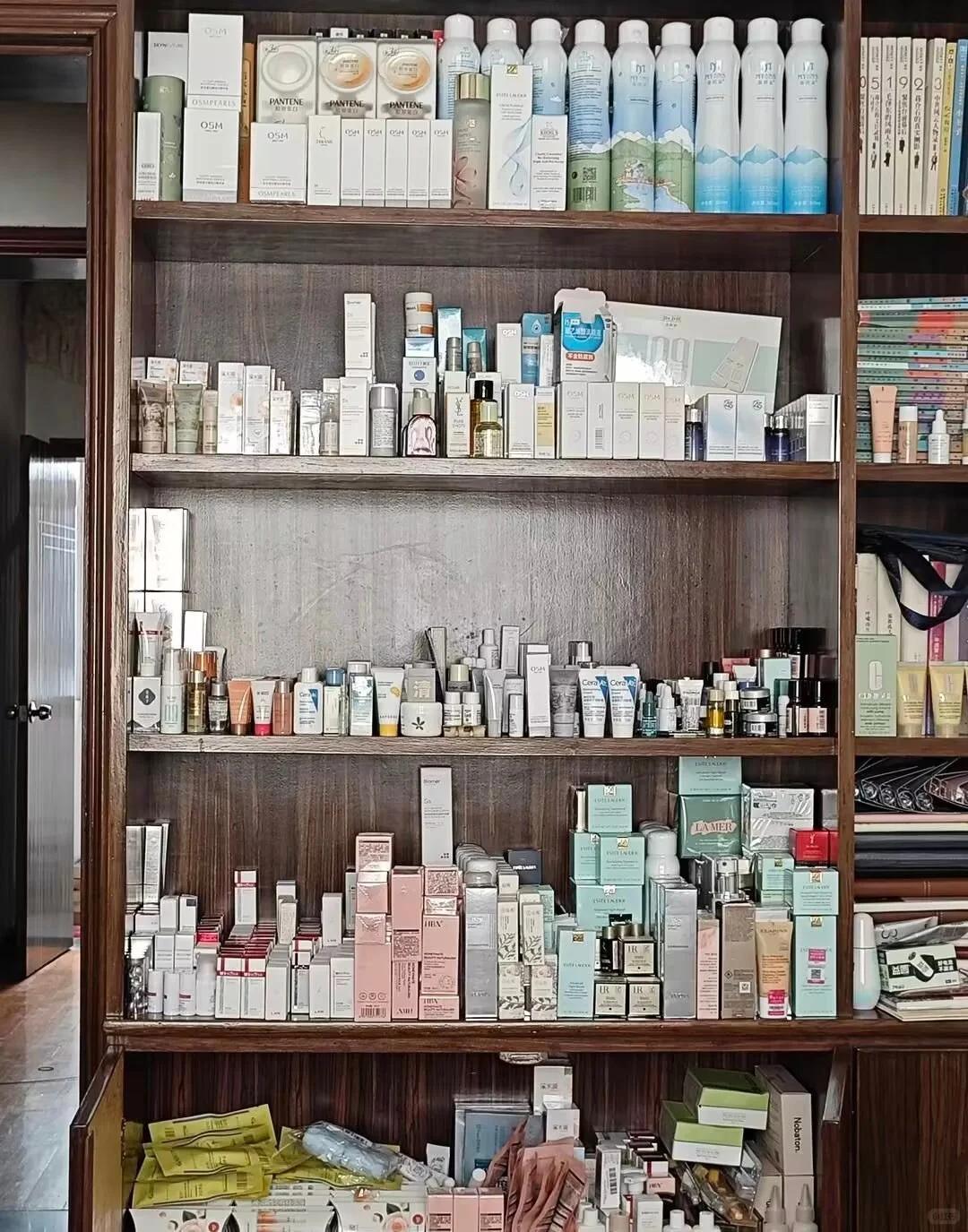}\\[2pt]
{\scriptsize\color{exMute} input\_image\_1}\end{center}
\exLabel{Question}{as sent to the model}
\begin{exverb}
Question: Which brand does the item on the far right in the top row belong to?
\end{exverb}
\exLabel{Without tools}{GPT-6 Luna's full response to the same question}
\begin{exverb}
The item on the far right in the top row appears to be from AHC.
\end{exverb}

%% file: text/appendix_examples_steps/srx140.tex
\begin{exharness}
\exLabel{Harness}{environment update, shown in full}
\begin{exverb}
Environment Update (not a new question):\\{}
Currently available image IDs for tool calls:\\{}
- input\_\allowbreak{}image\_\allowbreak{}1: user\_\allowbreak{}input\\{}
When a tool needs image input,\allowbreak{} pass one of these IDs via the *\_\allowbreak{}image\_\allowbreak{}id field.\\{}
Do NOT invent image file names or paths.
\end{exverb}
\end{exharness}
\begin{exstep}{1}{\toolZoom}{}
\exWalk{A first zoom on the top-right corner of the shelf.}
\exLabel{Arguments}{}
\begin{exverb}
\{\\{}
~~"\allowbreak{}image\_\allowbreak{}id"\allowbreak{}: "\allowbreak{}input\_\allowbreak{}image\_\allowbreak{}1"\allowbreak{},\allowbreak{}\\{}
~~"\allowbreak{}coords"\allowbreak{}: [812,\allowbreak{} 0,\allowbreak{} 872,\allowbreak{} 178],\allowbreak{}\\{}
~~"\allowbreak{}padding"\allowbreak{}: 70\\{}
\}
\end{exverb}
\exLabel{Tool output}{}
\begin{exverb}
Tool \textasciigrave{}Zoom\_\allowbreak{}In\_\allowbreak{}Tool\textasciigrave{} (tool\_\allowbreak{}call\_\allowbreak{}id=\allowbreak{}call\_\allowbreak{}aS30iH3kdrzpsXHTthJAJebA) returned:\\{}
Zoomed in image rendered below; refer to it as tool\_\allowbreak{}generated\_\allowbreak{}image\_\allowbreak{}1.
\end{exverb}
\begin{center}\includegraphics[width=0.45\linewidth,height=0.22\textheight,keepaspectratio]{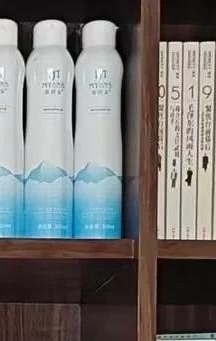}\\[2pt]
{\scriptsize\color{exMute} Image returned with this output.}\end{center}
\begin{exharness}
\exLabel{Harness}{environment update, summarized: new memory entries; new data bindings verbatim}
\begin{exverb}
Memory update: registers 1 new image in memory (tool\_\allowbreak{}generated\_\allowbreak{}image\_\allowbreak{}1,\allowbreak{} from Zoom\_\allowbreak{}In\_\allowbreak{}Tool).
\end{exverb}
\end{exharness}
\end{exstep}
\begin{exstep}{2}{\toolOCR}{}
\exWalk{OCR on the crop reads only a date, from something other than the spray can.}
\exLabel{Arguments}{}
\begin{exverb}
\{\\{}
~~"\allowbreak{}image\_\allowbreak{}id"\allowbreak{}: "\allowbreak{}tool\_\allowbreak{}generated\_\allowbreak{}image\_\allowbreak{}1"\allowbreak{}\\{}
\}
\end{exverb}
\exLabel{Tool output}{}
\begin{exverb}
\begin{CJK}{UTF8}{gbsn}OCR\_\allowbreak{}Tool transcribed the following text from the image:\\{}
\mbox{}\\{}
1982年10月1日\end{CJK}
\end{exverb}
\end{exstep}
\begin{exstep}{3}{\toolZoom}{}
\exWalk{A zoom further to the right, along the edge of the shelf.}
\exLabel{Arguments}{}
\begin{exverb}
\{\\{}
~~"\allowbreak{}image\_\allowbreak{}id"\allowbreak{}: "\allowbreak{}input\_\allowbreak{}image\_\allowbreak{}1"\allowbreak{},\allowbreak{}\\{}
~~"\allowbreak{}coords"\allowbreak{}: [962,\allowbreak{} 18,\allowbreak{} 999,\allowbreak{} 176],\allowbreak{}\\{}
~~"\allowbreak{}padding"\allowbreak{}: 70\\{}
\}
\end{exverb}
\exLabel{Tool output}{}
\begin{exverb}
Tool \textasciigrave{}Zoom\_\allowbreak{}In\_\allowbreak{}Tool\textasciigrave{} (tool\_\allowbreak{}call\_\allowbreak{}id=\allowbreak{}call\_\allowbreak{}FP7fQd52QBGnF4r6XelMVkCY) returned:\\{}
Zoomed in image rendered below; refer to it as tool\_\allowbreak{}generated\_\allowbreak{}image\_\allowbreak{}2.
\end{exverb}
\begin{center}\includegraphics[width=0.45\linewidth,height=0.22\textheight,keepaspectratio]{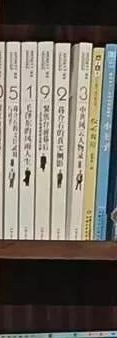}\\[2pt]
{\scriptsize\color{exMute} Image returned with this output.}\end{center}
\begin{exharness}
\exLabel{Harness}{environment update, summarized: new memory entries; new data bindings verbatim}
\begin{exverb}
Memory update: registers 1 new image in memory (tool\_\allowbreak{}generated\_\allowbreak{}image\_\allowbreak{}2,\allowbreak{} from Zoom\_\allowbreak{}In\_\allowbreak{}Tool).
\end{exverb}
\end{exharness}
\end{exstep}
\begin{exstep}{4}{\toolOCR}{}
\exWalk{OCR there reads the titles of books next to the shelf, not the product.}
\exLabel{Arguments}{}
\begin{exverb}
\{\\{}
~~"\allowbreak{}image\_\allowbreak{}id"\allowbreak{}: "\allowbreak{}tool\_\allowbreak{}generated\_\allowbreak{}image\_\allowbreak{}2"\allowbreak{}\\{}
\}
\end{exverb}
\exLabel{Tool output}{}
\begin{exverb}
\begin{CJK}{UTF8}{bsmi}OCR\_\allowbreak{}Tool transcribed the following text from the image:\\{}
\mbox{}\\{}
3 中國公物業\\{}
2 符合的直文例語\\{}
9 製作例語\\{}
1 毛優的風趣人生\end{CJK}
\end{exverb}
\end{exstep}
\begin{exstep}{5}{\toolIS}{}
\exWalk{A reverse search of the whole photograph returns generic Pinterest boards of beauty shelves.}
\exLabel{Arguments}{}
\begin{exverb}
\{\\{}
~~"\allowbreak{}image\_\allowbreak{}id"\allowbreak{}: "\allowbreak{}input\_\allowbreak{}image\_\allowbreak{}1"\allowbreak{}\\{}
\}
\end{exverb}
\exLabel{Tool output}{}
\begin{exverb}
Tool \textasciigrave{}Image\_\allowbreak{}Search\_\allowbreak{}Tool\textasciigrave{} (tool\_\allowbreak{}call\_\allowbreak{}id=\allowbreak{}call\_\allowbreak{}Whh5HfzIXfILrrmtcCajNDKj) returned:\\{}
Image refs: tool\_\allowbreak{}generated\_\allowbreak{}image\_\allowbreak{}3,\allowbreak{} tool\_\allowbreak{}generated\_\allowbreak{}image\_\allowbreak{}4,\allowbreak{} tool\_\allowbreak{}generated\_\allowbreak{}image\_\allowbreak{}5\\{}
[Image Search — Visual Matches]\\{}
1. Discover 130 Beauty Shelf and shelf full of beauty products ideas \textbar{} beauty products on a shelf,\allowbreak{} skincare shelfie ideas,\allowbreak{} korean skincare products on shelf and more\\{}
~~~Source: pinterest.com/\allowbreak{}honeymaesh/\allowbreak{}beauty-shelf   [webpage\_\allowbreak{}url\_\allowbreak{}1]\\{}
~~~Image: [tool\_\allowbreak{}generated\_\allowbreak{}image\_\allowbreak{}3]\\{}
2. I\textquotesingle{}m a Recent Dermstore Convert—Here Are the 21 Items I Would Buy in Bulk\\{}
~~~Source: es.pinterest.com/\allowbreak{}pin/\allowbreak{}266064290476051096   [webpage\_\allowbreak{}url\_\allowbreak{}2]\\{}
~~~Image: [tool\_\allowbreak{}generated\_\allowbreak{}image\_\allowbreak{}4]
\end{exverb}
\exCut{1 more results not shown (3 in total).}
\begin{center}
\begin{minipage}[t]{0.475\linewidth}\centering\includegraphics[width=\linewidth,height=0.16\textheight,keepaspectratio]{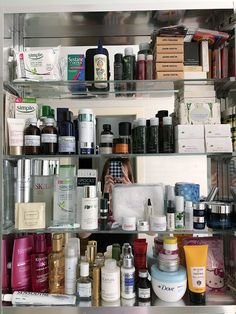}\end{minipage}\hfill\begin{minipage}[t]{0.475\linewidth}\centering\includegraphics[width=\linewidth,height=0.16\textheight,keepaspectratio]{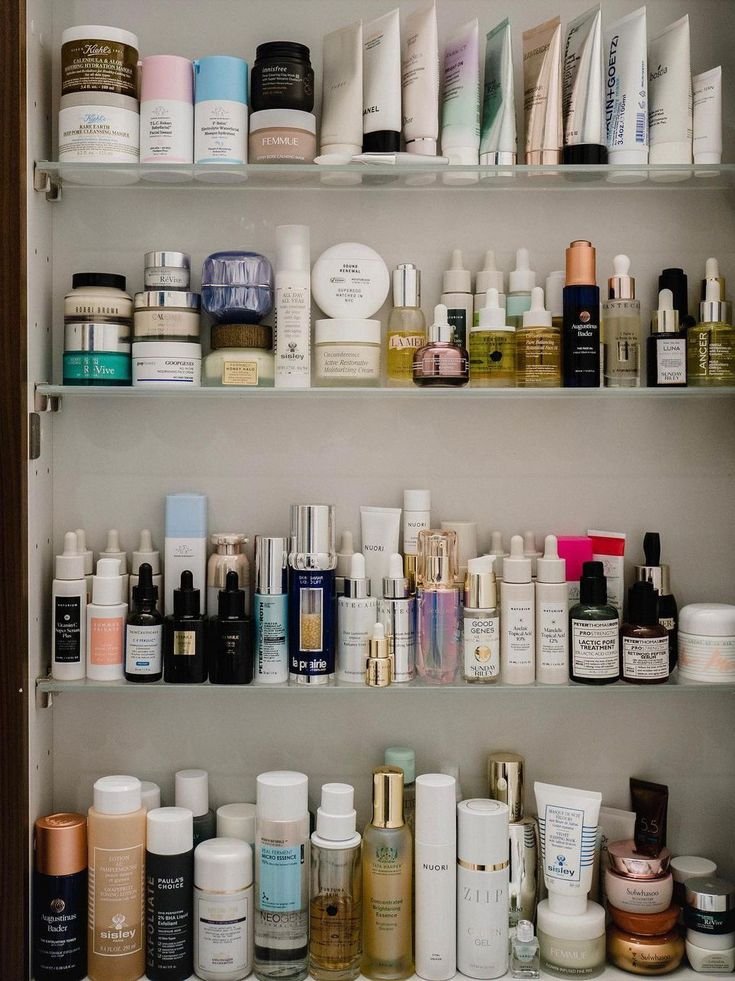}\end{minipage}
\\[1pt]{\scriptsize\color{exMute} Images returned with this output.}
\end{center}
\begin{exharness}
\exLabel{Harness}{environment update, summarized: new memory entries; new data bindings verbatim}
\begin{exverb}
Memory update: registers 3 new images in memory (tool\_\allowbreak{}generated\_\allowbreak{}image\_\allowbreak{}3 to tool\_\allowbreak{}generated\_\allowbreak{}image\_\allowbreak{}5,\allowbreak{} from Image\_\allowbreak{}Search\_\allowbreak{}Tool); registers 3 new webpage URLs in memory (webpage\_\allowbreak{}url\_\allowbreak{}1 to webpage\_\allowbreak{}url\_\allowbreak{}3,\allowbreak{} from Image\_\allowbreak{}Search\_\allowbreak{}Tool).
\end{exverb}
\end{exharness}
\end{exstep}
\begin{exstep}{6}{\toolZoom}{}
\exWalk{The agent zooms back onto the row of spray cans.}
\exLabel{Arguments}{}
\begin{exverb}
\{\\{}
~~"\allowbreak{}image\_\allowbreak{}id"\allowbreak{}: "\allowbreak{}input\_\allowbreak{}image\_\allowbreak{}1"\allowbreak{},\allowbreak{}\\{}
~~"\allowbreak{}coords"\allowbreak{}: [814,\allowbreak{} 26,\allowbreak{} 870,\allowbreak{} 178],\allowbreak{}\\{}
~~"\allowbreak{}padding"\allowbreak{}: 100\\{}
\}
\end{exverb}
\exLabel{Tool output}{}
\begin{exverb}
Tool \textasciigrave{}Zoom\_\allowbreak{}In\_\allowbreak{}Tool\textasciigrave{} (tool\_\allowbreak{}call\_\allowbreak{}id=\allowbreak{}call\_\allowbreak{}NbfAqx1ZRkUL0NVPM4WWiKLl) returned:\\{}
Zoomed in image rendered below; refer to it as tool\_\allowbreak{}generated\_\allowbreak{}image\_\allowbreak{}6.
\end{exverb}
\begin{center}\includegraphics[width=0.45\linewidth,height=0.22\textheight,keepaspectratio]{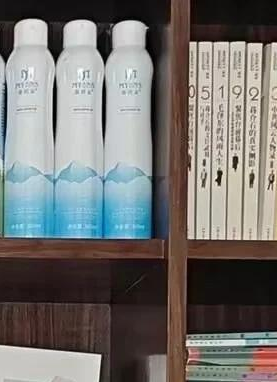}\\[2pt]
{\scriptsize\color{exMute} Image returned with this output.}\end{center}
\begin{exharness}
\exLabel{Harness}{environment update, summarized: new memory entries; new data bindings verbatim}
\begin{exverb}
Memory update: registers 1 new image in memory (tool\_\allowbreak{}generated\_\allowbreak{}image\_\allowbreak{}6,\allowbreak{} from Zoom\_\allowbreak{}In\_\allowbreak{}Tool).
\end{exverb}
\end{exharness}
\end{exstep}
\begin{exstep}{7}{\toolVG}{}
\exWalk{Grounding ``shampoo bottle'' on that crop separates the four tall bottles, so the agent can see which one is at the far right.}
\exLabel{Arguments}{}
\begin{exverb}
\{\\{}
~~"\allowbreak{}query"\allowbreak{}: "\allowbreak{}shampoo bottle"\allowbreak{},\allowbreak{}\\{}
~~"\allowbreak{}image\_\allowbreak{}id"\allowbreak{}: "\allowbreak{}tool\_\allowbreak{}generated\_\allowbreak{}image\_\allowbreak{}6"\allowbreak{}\\{}
\}
\end{exverb}
\exLabel{Tool output}{}
\begin{exverb}
Visual\_\allowbreak{}Grounding\_\allowbreak{}Tool grounded 4 objects for the prompt \textquotesingle{}shampoo bottle\textquotesingle{}. Their corresponding bounding boxes (on a 0-1000 scale,\allowbreak{} [x1,\allowbreak{} y1,\allowbreak{} x2,\allowbreak{} y2]) are: \textasciigrave{}\textasciigrave{}\textasciigrave{}json\\{}
[\\{}
~~~~\{"\allowbreak{}bbox\_\allowbreak{}2d"\allowbreak{}: [0,\allowbreak{} 144,\allowbreak{} 18,\allowbreak{} 615],\allowbreak{} "\allowbreak{}label"\allowbreak{}: "\allowbreak{}shampoo bottle"\allowbreak{}\},\allowbreak{}\\{}
~~~~\{"\allowbreak{}bbox\_\allowbreak{}2d"\allowbreak{}: [16,\allowbreak{} 50,\allowbreak{} 197,\allowbreak{} 621],\allowbreak{} "\allowbreak{}label"\allowbreak{}: "\allowbreak{}shampoo bottle"\allowbreak{}\},\allowbreak{}\\{}
~~~~\{"\allowbreak{}bbox\_\allowbreak{}2d"\allowbreak{}: [198,\allowbreak{} 52,\allowbreak{} 376,\allowbreak{} 623],\allowbreak{} "\allowbreak{}label"\allowbreak{}: "\allowbreak{}shampoo bottle"\allowbreak{}\},\allowbreak{}\\{}
~~~~\{"\allowbreak{}bbox\_\allowbreak{}2d"\allowbreak{}: [376,\allowbreak{} 54,\allowbreak{} 554,\allowbreak{} 624],\allowbreak{} "\allowbreak{}label"\allowbreak{}: "\allowbreak{}shampoo bottle"\allowbreak{}\}\\{}
]\textasciigrave{}\textasciigrave{}\textasciigrave{}. The object numbers correspond to the labeled masks rendered in the following image,\allowbreak{} which you can refer to as tool\_\allowbreak{}generated\_\allowbreak{}image\_\allowbreak{}7.
\end{exverb}
\begin{center}\includegraphics[width=0.45\linewidth,height=0.22\textheight,keepaspectratio]{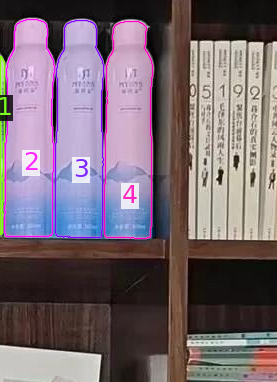}\\[2pt]
{\scriptsize\color{exMute} Image returned with this output.}\end{center}
\begin{exharness}
\exLabel{Harness}{environment update, summarized: new memory entries; new data bindings verbatim}
\begin{exverb}
Memory update: registers 1 new image in memory (tool\_\allowbreak{}generated\_\allowbreak{}image\_\allowbreak{}7,\allowbreak{} from Visual\_\allowbreak{}Grounding\_\allowbreak{}Tool).\\{}
Available data bindings (auto-injected into Python\_\allowbreak{}Coding\_\allowbreak{}Agent\_\allowbreak{}Tool):\\{}
- segment\_\allowbreak{}shampoo\_\allowbreak{}bottle: dict (keys: pred\_\allowbreak{}boxes,\allowbreak{} pred\_\allowbreak{}scores,\allowbreak{} pred\_\allowbreak{}masks,\allowbreak{} num\_\allowbreak{}masks,\allowbreak{} orig\_\allowbreak{}img\_\allowbreak{}h,\allowbreak{} orig\_\allowbreak{}img\_\allowbreak{}w)
\end{exverb}
\end{exharness}
\end{exstep}
\begin{exstep}{8}{\toolZoom}{}
\exWalk{A tighter zoom on the rightmost bottles.}
\exLabel{Arguments}{}
\begin{exverb}
\{\\{}
~~"\allowbreak{}image\_\allowbreak{}id"\allowbreak{}: "\allowbreak{}input\_\allowbreak{}image\_\allowbreak{}1"\allowbreak{},\allowbreak{}\\{}
~~"\allowbreak{}coords"\allowbreak{}: [804,\allowbreak{} 37,\allowbreak{} 872,\allowbreak{} 176],\allowbreak{}\\{}
~~"\allowbreak{}padding"\allowbreak{}: 120\\{}
\}
\end{exverb}
\exLabel{Tool output}{}
\begin{exverb}
Tool \textasciigrave{}Zoom\_\allowbreak{}In\_\allowbreak{}Tool\textasciigrave{} (tool\_\allowbreak{}call\_\allowbreak{}id=\allowbreak{}call\_\allowbreak{}azos4cSBGPDQ4vfi9c2sTq93) returned:\\{}
Zoomed in image rendered below; refer to it as tool\_\allowbreak{}generated\_\allowbreak{}image\_\allowbreak{}8.
\end{exverb}
\begin{center}\includegraphics[width=0.45\linewidth,height=0.22\textheight,keepaspectratio]{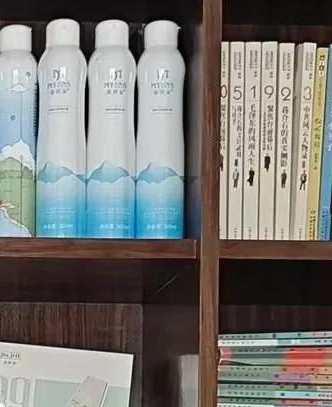}\\[2pt]
{\scriptsize\color{exMute} Image returned with this output.}\end{center}
\begin{exharness}
\exLabel{Harness}{environment update, summarized: new memory entries; new data bindings verbatim}
\begin{exverb}
Memory update: registers 1 new image in memory (tool\_\allowbreak{}generated\_\allowbreak{}image\_\allowbreak{}8,\allowbreak{} from Zoom\_\allowbreak{}In\_\allowbreak{}Tool).
\end{exverb}
\end{exharness}
\end{exstep}
\begin{exstep}{9}{\toolZoom}{}
\exWalk{A final tight crop around the target can.}
\exLabel{Arguments}{}
\begin{exverb}
\{\\{}
~~"\allowbreak{}image\_\allowbreak{}id"\allowbreak{}: "\allowbreak{}input\_\allowbreak{}image\_\allowbreak{}1"\allowbreak{},\allowbreak{}\\{}
~~"\allowbreak{}coords"\allowbreak{}: [805,\allowbreak{} 25,\allowbreak{} 875,\allowbreak{} 175],\allowbreak{}\\{}
~~"\allowbreak{}padding"\allowbreak{}: 100\\{}
\}
\end{exverb}
\exLabel{Tool output}{}
\begin{exverb}
Tool \textasciigrave{}Zoom\_\allowbreak{}In\_\allowbreak{}Tool\textasciigrave{} (tool\_\allowbreak{}call\_\allowbreak{}id=\allowbreak{}call\_\allowbreak{}Nq6FPXZJ2SVXf38UvzL0sBlU) returned:\\{}
Zoomed in image rendered below; refer to it as tool\_\allowbreak{}generated\_\allowbreak{}image\_\allowbreak{}9.
\end{exverb}
\begin{center}\includegraphics[width=0.45\linewidth,height=0.22\textheight,keepaspectratio]{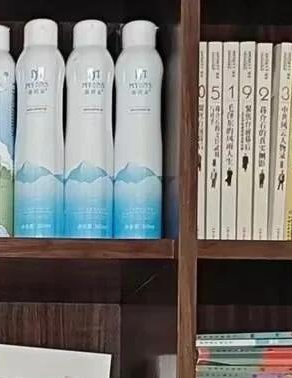}\\[2pt]
{\scriptsize\color{exMute} Image returned with this output.}\end{center}
\begin{exharness}
\exLabel{Harness}{environment update, summarized: new memory entries; new data bindings verbatim}
\begin{exverb}
Memory update: registers 1 new image in memory (tool\_\allowbreak{}generated\_\allowbreak{}image\_\allowbreak{}9,\allowbreak{} from Zoom\_\allowbreak{}In\_\allowbreak{}Tool).
\end{exverb}
\end{exharness}
\end{exstep}
\begin{exstep}{10}{\toolOCR}{}
\exWalk{OCR on the tight crop again reads only a date.}
\exLabel{Arguments}{}
\begin{exverb}
\{\\{}
~~"\allowbreak{}image\_\allowbreak{}id"\allowbreak{}: "\allowbreak{}tool\_\allowbreak{}generated\_\allowbreak{}image\_\allowbreak{}9"\allowbreak{}\\{}
\}
\end{exverb}
\exLabel{Tool output}{}
\begin{exverb}
\begin{CJK}{UTF8}{gbsn}OCR\_\allowbreak{}Tool transcribed the following text from the image:\\{}
\mbox{}\\{}
2015年1月1日\end{CJK}
\end{exverb}
\end{exstep}
\begin{exstep}{11}{\toolIS}{}
\exWalk{Reverse image search on the tight crop. Its second match is Miyuequan Hydrating Spray, the product on the shelf.}
\exLabel{Arguments}{}
\begin{exverb}
\{\\{}
~~"\allowbreak{}image\_\allowbreak{}id"\allowbreak{}: "\allowbreak{}tool\_\allowbreak{}generated\_\allowbreak{}image\_\allowbreak{}9"\allowbreak{}\\{}
\}
\end{exverb}
\exLabel{Model}{text emitted with this call}
\exSayBlock{\begin{CJK}{UTF8}{gbsn}MYONS (弥\CJKfamily{bsmi}玥\CJKfamily{gbsn}泉)\end{CJK}}
\exLabel{Tool output}{}
\begin{exverb}
Tool \textasciigrave{}Image\_\allowbreak{}Search\_\allowbreak{}Tool\textasciigrave{} (tool\_\allowbreak{}call\_\allowbreak{}id=\allowbreak{}call\_\allowbreak{}zqaG64IKUZe4rIVRkEfufdxn) returned:\\{}
Image refs: tool\_\allowbreak{}generated\_\allowbreak{}image\_\allowbreak{}10,\allowbreak{} tool\_\allowbreak{}generated\_\allowbreak{}image\_\allowbreak{}11,\allowbreak{} tool\_\allowbreak{}generated\_\allowbreak{}image\_\allowbreak{}12\\{}
[Image Search — Visual Matches]\\{}
1. Amazon.com: YILIAN Face Mist Natural Facial Spray with HA,\allowbreak{} Soothing and Calming Sensitive Skin,\allowbreak{} 10.14 fl. oz.\\{}
~~~Source: amazon.com/\allowbreak{}clp/\allowbreak{}B0DTT9SV1Y   [webpage\_\allowbreak{}url\_\allowbreak{}4]\\{}
~~~Image: [tool\_\allowbreak{}generated\_\allowbreak{}image\_\allowbreak{}10]\\{}
2. Fast delivery][Hot selling in China] Miyuequan Hydrating Spray Moisturizing for Women Soothing Skin Pre-Makeup Glacier Hot Spring Small Molecule Toner Miyuequan \textbar{} Daraz.com.np\\{}
~~~Source: daraz.com.np/\allowbreak{}products/\allowbreak{}fast-deliveryhot-selling-in-china-miyuequan-hydrating-s...   [webpage\_\allowbreak{}url\_\allowbreak{}5]\\{}
~~~Image: [tool\_\allowbreak{}generated\_\allowbreak{}image\_\allowbreak{}11]
\end{exverb}
\exCut{1 more results not shown (3 in total).}
\begin{center}
\begin{minipage}[t]{0.475\linewidth}\centering\includegraphics[width=\linewidth,height=0.16\textheight,keepaspectratio]{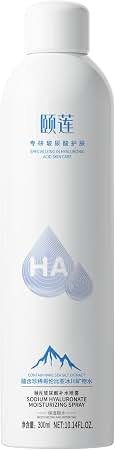}\end{minipage}\hfill\begin{minipage}[t]{0.475\linewidth}\centering\includegraphics[width=\linewidth,height=0.16\textheight,keepaspectratio]{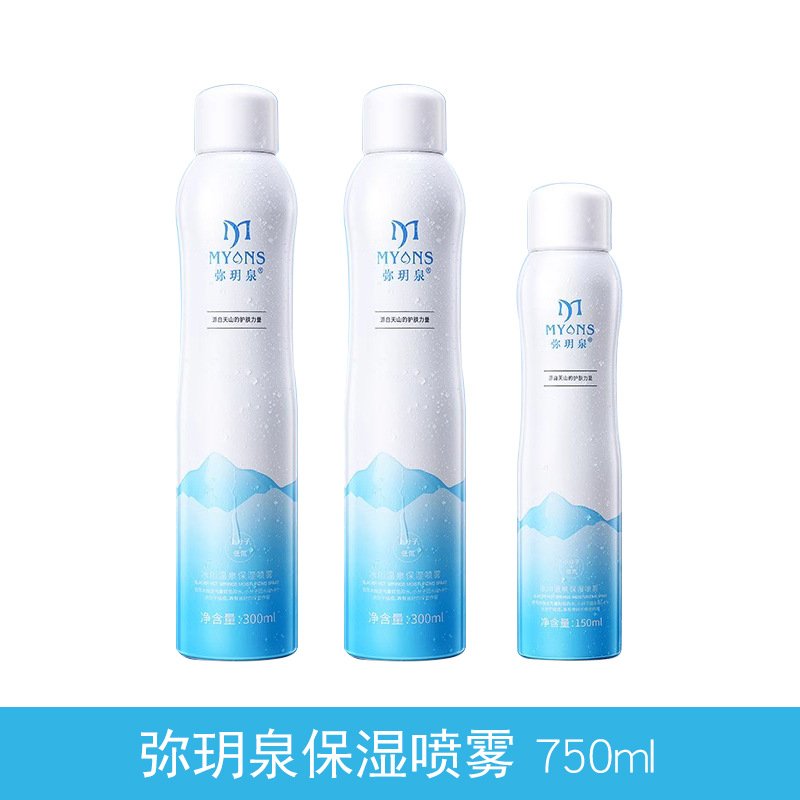}\end{minipage}
\\[1pt]{\scriptsize\color{exMute} Images returned with this output.}
\end{center}
\begin{exharness}
\exLabel{Harness}{environment update, summarized: new memory entries; new data bindings verbatim}
\begin{exverb}
Memory update: registers 3 new images in memory (tool\_\allowbreak{}generated\_\allowbreak{}image\_\allowbreak{}10 to tool\_\allowbreak{}generated\_\allowbreak{}image\_\allowbreak{}12,\allowbreak{} from Image\_\allowbreak{}Search\_\allowbreak{}Tool); registers 3 new webpage URLs in memory (webpage\_\allowbreak{}url\_\allowbreak{}4 to webpage\_\allowbreak{}url\_\allowbreak{}6,\allowbreak{} from Image\_\allowbreak{}Search\_\allowbreak{}Tool).
\end{exverb}
\end{exharness}
\end{exstep}

%% file: text/appendix_examples_steps/srx2_header.tex
\exOverview{\textbf{Task.} A makeup organizer holds brushes and cosmetics. Which brand is the liquid foundation at the far right? Its label is unreadable; its most distinctive feature is a crystal-faceted cap. 

\textbf{Why it is hard.} Without tools the model guesses Dior. 

\textbf{How the agent solves it.} OCR reads nothing, and reverse-searching a crop of the bottle finds no match. The agent then searches by description (``faceted crystal cap''), tries a Jill Stuart hypothesis and drops it, and reverse-searches a tighter crop. That search's summary names the bottle as KATOKATO, and a targeted image search returns a KATO-KATO foundation whose cap matches the crop. 

\textbf{Caveat.} The brand name first appears in the reverse-search tool's own summary of the crop, so the decisive step is partly inside the tool.}
\exLabel{Input images}{}
\begin{center}\includegraphics[width=0.55\linewidth,height=0.28\textheight,keepaspectratio]{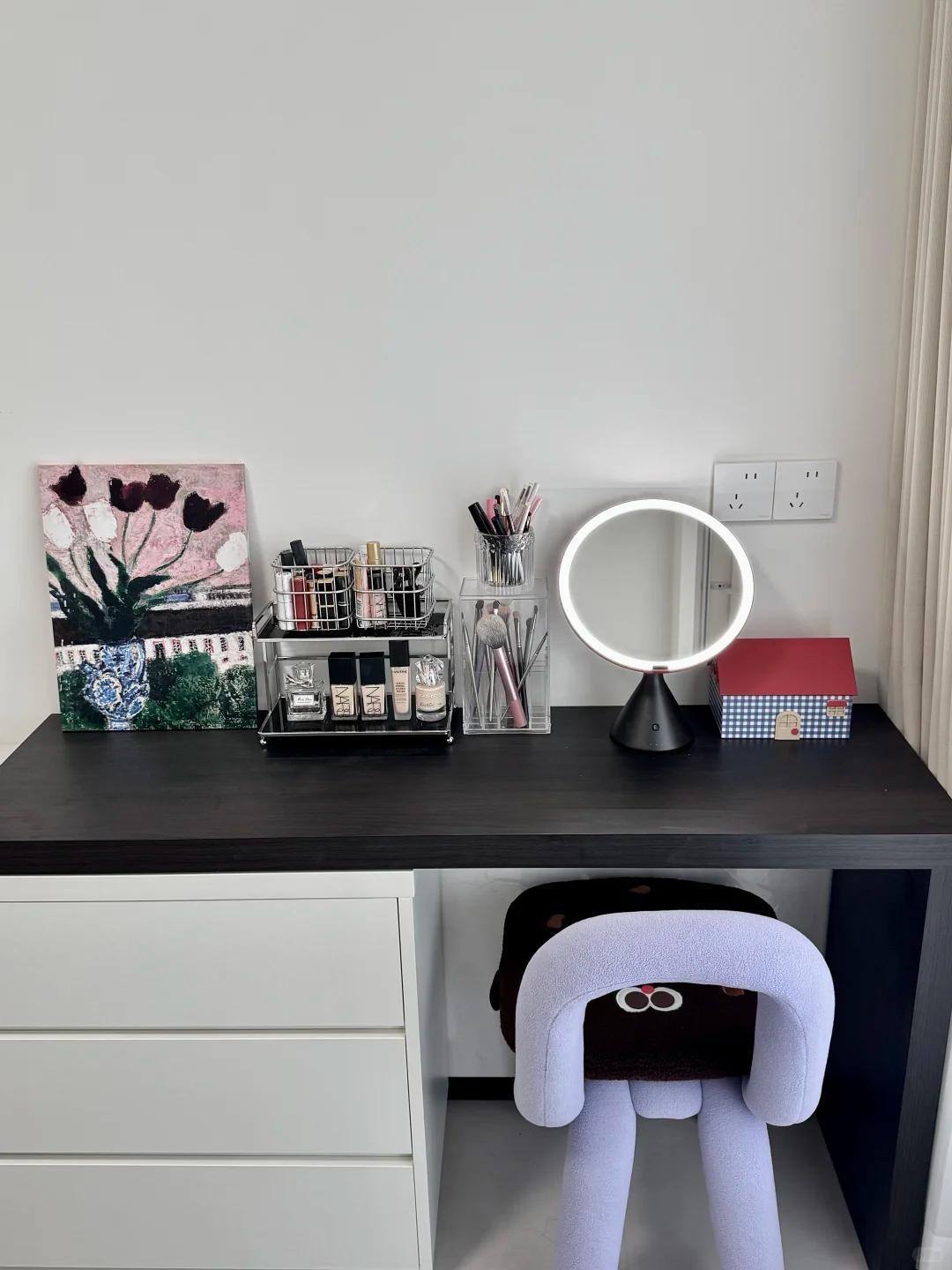}\\[2pt]
{\scriptsize\color{exMute} input\_image\_1}\end{center}
\exLabel{Question}{as sent to the model}
\begin{exverb}
Question: Which brand does the liquid foundation on the far right on the storage shelf in the picture belong to?
\end{exverb}
\exLabel{Without tools}{GPT-6 Sol's full response to the same question}
\begin{exverb}
The liquid foundation on the far right appears to be **Dior**.
\end{exverb}

%% file: text/appendix_examples_steps/srx2.tex
\begin{exharness}
\exLabel{Harness}{environment update, shown in full}
\begin{exverb}
Environment Update (not a new question):\\{}
Currently available image IDs for tool calls:\\{}
- input\_\allowbreak{}image\_\allowbreak{}1: user\_\allowbreak{}input\\{}
When a tool needs image input,\allowbreak{} pass one of these IDs via the *\_\allowbreak{}image\_\allowbreak{}id field.\\{}
Do NOT invent image file names or paths.
\end{exverb}
\end{exharness}
\begin{exstep}{1}{\toolOCR}{}
\exWalk{OCR on the whole photograph finds no readable text.}
\exLabel{Arguments}{}
\begin{exverb}
\{\\{}
~~"\allowbreak{}image\_\allowbreak{}id"\allowbreak{}: "\allowbreak{}input\_\allowbreak{}image\_\allowbreak{}1"\allowbreak{}\\{}
\}
\end{exverb}
\exLabel{Model}{text emitted with this call}
\exSayBlock{I’ll check the label on the rightmost foundation bottle on the shelf.}
\exLabel{Tool output}{}
\begin{exverb}
OCR\_\allowbreak{}Tool did not transcribe any readable text.
\end{exverb}
\end{exstep}
\begin{exstep}{2}{\toolZoom}{}
\exWalk{A zoom on the rightmost bottle.}
\exLabel{Arguments}{}
\begin{exverb}
\{\\{}
~~"\allowbreak{}image\_\allowbreak{}id"\allowbreak{}: "\allowbreak{}input\_\allowbreak{}image\_\allowbreak{}1"\allowbreak{},\allowbreak{}\\{}
~~"\allowbreak{}coords"\allowbreak{}: [423,\allowbreak{} 495,\allowbreak{} 485,\allowbreak{} 585],\allowbreak{}\\{}
~~"\allowbreak{}padding"\allowbreak{}: 70\\{}
\}
\end{exverb}
\exLabel{Model}{text emitted with this call}
\exSayBlock{The label wasn’t readable at full size,\allowbreak{} so I’ll inspect the rightmost bottle more closely.}
\exLabel{Tool output}{}
\begin{exverb}
Tool \textasciigrave{}Zoom\_\allowbreak{}In\_\allowbreak{}Tool\textasciigrave{} (tool\_\allowbreak{}call\_\allowbreak{}id=\allowbreak{}call\_\allowbreak{}zZ2NiO8lBIDogDGCwUzWI6ZD) returned:\\{}
Zoomed in image rendered below; refer to it as tool\_\allowbreak{}generated\_\allowbreak{}image\_\allowbreak{}1.
\end{exverb}
\begin{center}\includegraphics[width=0.45\linewidth,height=0.22\textheight,keepaspectratio]{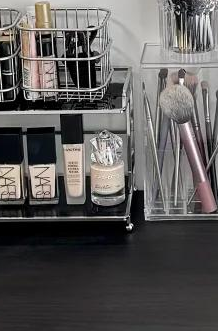}\\[2pt]
{\scriptsize\color{exMute} Image returned with this output.}\end{center}
\begin{exharness}
\exLabel{Harness}{environment update, summarized: new memory entries; new data bindings verbatim}
\begin{exverb}
Memory update: registers 1 new image in memory (tool\_\allowbreak{}generated\_\allowbreak{}image\_\allowbreak{}1,\allowbreak{} from Zoom\_\allowbreak{}In\_\allowbreak{}Tool).
\end{exverb}
\end{exharness}
\end{exstep}
\begin{exstep}{3}{\toolIS}{}
\exWalk{Reverse image search on that crop finds no relevant match.}
\exLabel{Arguments}{}
\begin{exverb}
\{\\{}
~~"\allowbreak{}image\_\allowbreak{}id"\allowbreak{}: "\allowbreak{}tool\_\allowbreak{}generated\_\allowbreak{}image\_\allowbreak{}1"\allowbreak{}\\{}
\}
\end{exverb}
\exLabel{Model}{text emitted with this call}
\exSayBlock{I’ll compare the bottle’s distinctive cap and shape with product images to confirm the brand.}
\exLabel{Tool output}{}
\begin{exverb}
Tool \textasciigrave{}Image\_\allowbreak{}Search\_\allowbreak{}Tool\textasciigrave{} (tool\_\allowbreak{}call\_\allowbreak{}id=\allowbreak{}call\_\allowbreak{}OoscIm9kAfd999QA3dwuSAnZ) returned:\\{}
\{"\allowbreak{}ok"\allowbreak{}: true,\allowbreak{} "\allowbreak{}result"\allowbreak{}: "\allowbreak{}[Image Search — Visual Matches]\\{}
NO RELEVANT MATCHES\\{}
\mbox{}\\{}
Why each raw match was rejected:\\{}
All candidates show single cosmetic products (nail polish or foundation) in isolation,\allowbreak{} while the QUERY image shows a multi-tiered makeup organizer with brushes and cosmetics. The image-text alignment is correct for each candidate but irrelevant to the QUERY image\textquotesingle{}s subject. The text metadata is domain-relevant (beauty products) but fails to match the QUERY image\textquotesingle{}s specific visual subject.\\{}
\mbox{}\\{}
Suggested next actions:\\{}
- Image\_\allowbreak{}Search\_\allowbreak{}Tool(image\_\allowbreak{}id=\allowbreak{}input\_\allowbreak{}image\_\allowbreak{}1)"\allowbreak{},\allowbreak{} "\allowbreak{}error"\allowbreak{}: null\}
\end{exverb}
\end{exstep}
\begin{exstep}{4}{\toolZoom}{}
\exWalk{A tighter zoom on the bottle's label and cap.}
\exLabel{Arguments}{}
\begin{exverb}
\{\\{}
~~"\allowbreak{}image\_\allowbreak{}id"\allowbreak{}: "\allowbreak{}input\_\allowbreak{}image\_\allowbreak{}1"\allowbreak{},\allowbreak{}\\{}
~~"\allowbreak{}coords"\allowbreak{}: [432,\allowbreak{} 511,\allowbreak{} 474,\allowbreak{} 570],\allowbreak{}\\{}
~~"\allowbreak{}padding"\allowbreak{}: 50\\{}
\}
\end{exverb}
\exLabel{Model}{text emitted with this call}
\exSayBlock{I’ll look closer at the bottle’s lettering and cap.}
\exLabel{Tool output}{}
\begin{exverb}
Tool \textasciigrave{}Zoom\_\allowbreak{}In\_\allowbreak{}Tool\textasciigrave{} (tool\_\allowbreak{}call\_\allowbreak{}id=\allowbreak{}call\_\allowbreak{}IaM1kjcmpzK9Jytg2996KDA3) returned:\\{}
Zoomed in image rendered below; refer to it as tool\_\allowbreak{}generated\_\allowbreak{}image\_\allowbreak{}2.
\end{exverb}
\begin{center}\includegraphics[width=0.45\linewidth,height=0.22\textheight,keepaspectratio]{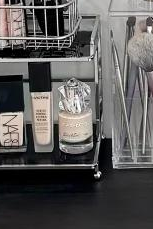}\\[2pt]
{\scriptsize\color{exMute} Image returned with this output.}\end{center}
\begin{exharness}
\exLabel{Harness}{environment update, summarized: new memory entries; new data bindings verbatim}
\begin{exverb}
Memory update: registers 1 new image in memory (tool\_\allowbreak{}generated\_\allowbreak{}image\_\allowbreak{}2,\allowbreak{} from Zoom\_\allowbreak{}In\_\allowbreak{}Tool).
\end{exverb}
\end{exharness}
\end{exstep}
\begin{exstep}{5}{\toolOCR}{}
\exWalk{OCR on the tight crop still reads nothing.}
\exLabel{Arguments}{}
\begin{exverb}
\{\\{}
~~"\allowbreak{}image\_\allowbreak{}id"\allowbreak{}: "\allowbreak{}tool\_\allowbreak{}generated\_\allowbreak{}image\_\allowbreak{}2"\allowbreak{}\\{}
\}
\end{exverb}
\exLabel{Tool output}{}
\begin{exverb}
OCR\_\allowbreak{}Tool did not transcribe any readable text.
\end{exverb}
\end{exstep}
\begin{exstep}{6}{\toolTS}{}
\exWalk{A text search that describes the bottle (faceted silver cap, beige liquid) returns lists of luxury foundations.}
\exLabel{Arguments}{}
\begin{exverb}
\{\\{}
~~"\allowbreak{}query"\allowbreak{}: "\allowbreak{}foundation bottle ornate faceted silver cap rounded glass beige foundation black base luxury brand"\allowbreak{},\allowbreak{}\\{}
~~"\allowbreak{}search\_\allowbreak{}type"\allowbreak{}: "\allowbreak{}images"\allowbreak{}\\{}
\}
\end{exverb}
\exLabel{Tool output}{}
\begin{exverb}
Tool \textasciigrave{}Text\_\allowbreak{}Search\_\allowbreak{}Tool\textasciigrave{} (tool\_\allowbreak{}call\_\allowbreak{}id=\allowbreak{}call\_\allowbreak{}BGJ52afbD0ZHnyo2fgnt4UGG) returned:\\{}
Image refs: tool\_\allowbreak{}generated\_\allowbreak{}image\_\allowbreak{}3,\allowbreak{} tool\_\allowbreak{}generated\_\allowbreak{}image\_\allowbreak{}4,\allowbreak{} tool\_\allowbreak{}generated\_\allowbreak{}image\_\allowbreak{}5\\{}
[Search Results — Type: images]\\{}
1. Top 7 Luxury Foundations: Flawless,\allowbreak{} Long-Lasting \& Easy to Apply\\{}
~~~Image: [tool\_\allowbreak{}generated\_\allowbreak{}image\_\allowbreak{}3]\\{}
~~~Source: Lemon8   [webpage\_\allowbreak{}url\_\allowbreak{}1]\\{}
~~~Size: 1440x1920\\{}
2. Top 7 Luxury Foundations: Flawless,\allowbreak{} Long-Lasting \& Easy to Apply\\{}
~~~Image: [tool\_\allowbreak{}generated\_\allowbreak{}image\_\allowbreak{}4]\\{}
~~~Source: Lemon8   [webpage\_\allowbreak{}url\_\allowbreak{}1]\\{}
~~~Size: 1440x1920
\end{exverb}
\exCut{1 more results not shown (3 in total).}
\begin{center}
\begin{minipage}[t]{0.475\linewidth}\centering\includegraphics[width=\linewidth,height=0.16\textheight,keepaspectratio]{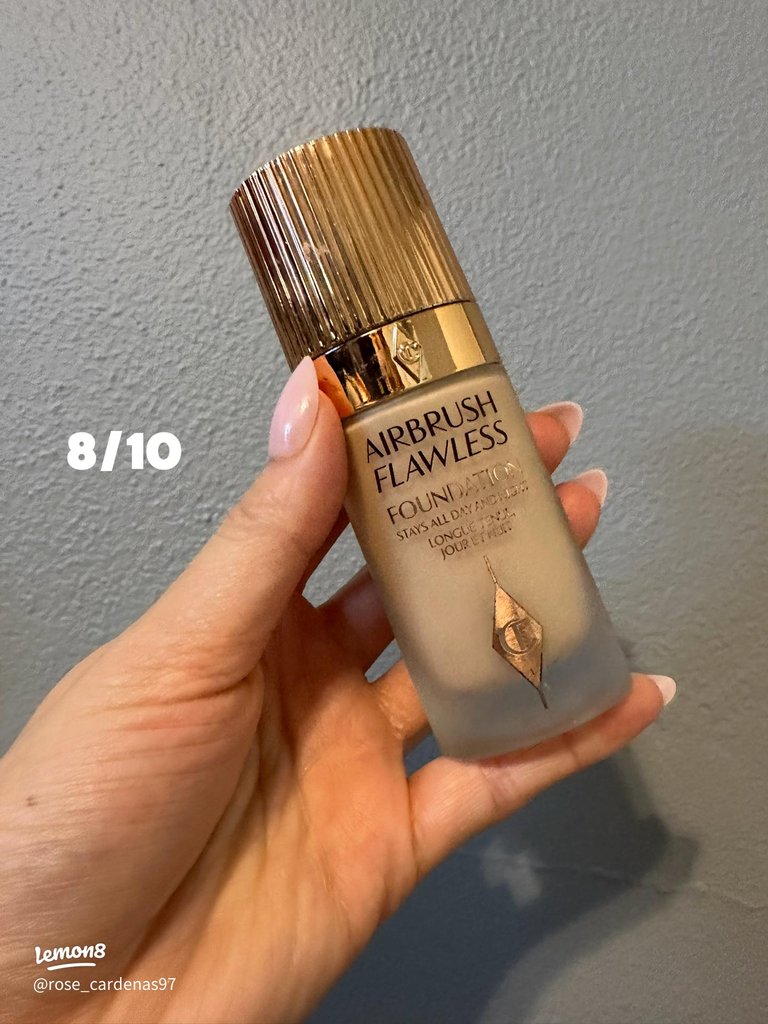}\end{minipage}\hfill\begin{minipage}[t]{0.475\linewidth}\centering\includegraphics[width=\linewidth,height=0.16\textheight,keepaspectratio]{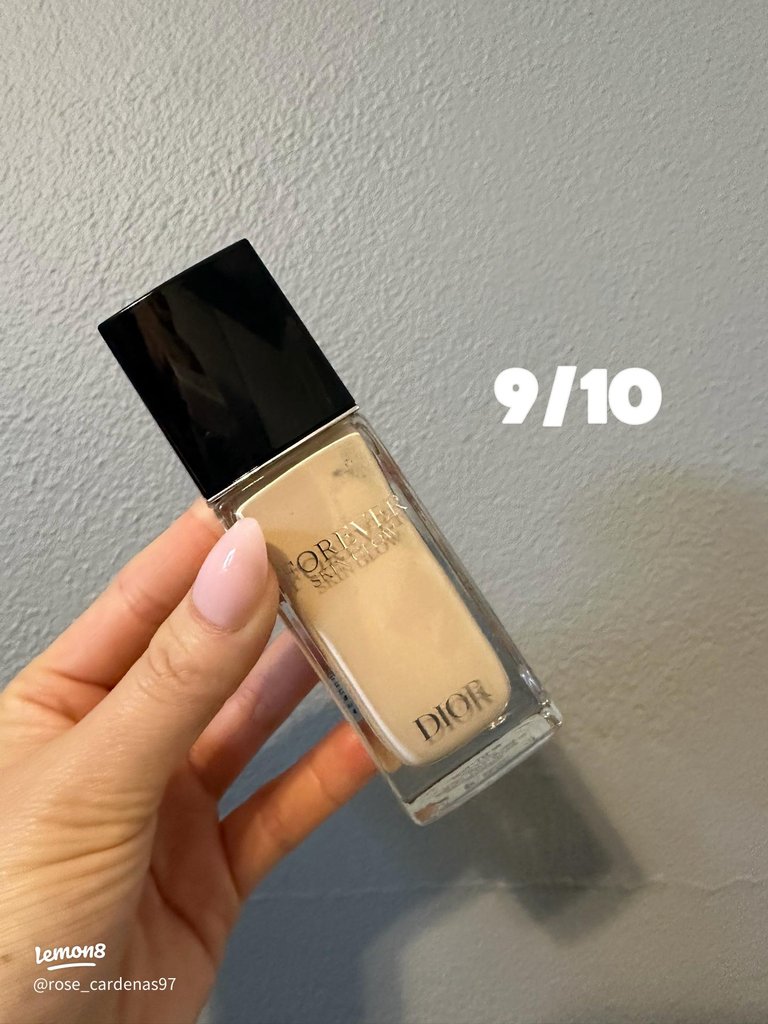}\end{minipage}
\\[1pt]{\scriptsize\color{exMute} Images returned with this output.}
\end{center}
\begin{exharness}
\exLabel{Harness}{environment update, summarized: new memory entries; new data bindings verbatim}
\begin{exverb}
Memory update: registers 3 new images in memory (tool\_\allowbreak{}generated\_\allowbreak{}image\_\allowbreak{}3 to tool\_\allowbreak{}generated\_\allowbreak{}image\_\allowbreak{}5,\allowbreak{} from Text\_\allowbreak{}Search\_\allowbreak{}Tool); registers 1 new webpage URL in memory (webpage\_\allowbreak{}url\_\allowbreak{}1,\allowbreak{} from Text\_\allowbreak{}Search\_\allowbreak{}Tool).
\end{exverb}
\end{exharness}
\end{exstep}
\begin{exstep}{7}{\toolTS}{}
\exWalk{A second descriptive search returns nothing relevant.}
\exLabel{Arguments}{}
\begin{exverb}
\{\\{}
~~"\allowbreak{}query"\allowbreak{}: "\allowbreak{}foundation bottle clear faceted jewel crystal cap silver beige liquid round short bottle black base"\allowbreak{},\allowbreak{}\\{}
~~"\allowbreak{}search\_\allowbreak{}type"\allowbreak{}: "\allowbreak{}images"\allowbreak{}\\{}
\}
\end{exverb}
\exLabel{Tool output}{}
\begin{exverb}
Tool \textasciigrave{}Text\_\allowbreak{}Search\_\allowbreak{}Tool\textasciigrave{} (tool\_\allowbreak{}call\_\allowbreak{}id=\allowbreak{}call\_\allowbreak{}S6nR1Mt5MOIn4aui2uN06iyd) returned:\\{}
\{"\allowbreak{}ok"\allowbreak{}: true,\allowbreak{} "\allowbreak{}result"\allowbreak{}: "\allowbreak{}[Search Results — Type: images]\\{}
NO RELEVANT RESULTS\\{}
\mbox{}\\{}
Why each raw result was rejected:\\{}
All candidates show foundation bottles that are either square,\allowbreak{} cylindrical,\allowbreak{} or lack the specified faceted crystal design,\allowbreak{} silver cap,\allowbreak{} or black base. Image 4 is the closest but lacks the cap color and base detail. Image 9 has beige liquid but a green cap and is not clear or faceted. Image 5 and 8 show frosted bottles,\allowbreak{} not clear or faceted. Image 1,\allowbreak{} 2,\allowbreak{} 3,\allowbreak{} 6,\allowbreak{} 7 show square bottles with black caps,\allowbreak{} not round or faceted. Image 10 shows a tube,\allowbreak{} not a bottle. All candidates fail on at least one key attribute (shape,\allowbreak{} cap color,\allowbreak{} base color,\allowbreak{} faceted crystal design).\\{}
\mbox{}\\{}
Suggested query rewrites:\\{}
- clear faceted glass foundation bottle with black cap and beige liquid\\{}
- jewel crystal foundation bottle round short black base\\{}
- refillable clear glass foundation bottle with silver cap\\{}
\mbox{}\\{}
Suggested next actions:\\{}
- Text\_\allowbreak{}Search\_\allowbreak{}Tool(query=\allowbreak{}"\allowbreak{}clear faceted glass foundation bottle with black cap and beige liquid"\allowbreak{},\allowbreak{} search\_\allowbreak{}type=\allowbreak{}images)\\{}
- Text\_\allowbreak{}Search\_\allowbreak{}Tool(query=\allowbreak{}"\allowbreak{}jewel crystal foundation bottle round short black base"\allowbreak{},\allowbreak{} search\_\allowbreak{}type=\allowbreak{}images)\\{}
- Text\_\allowbreak{}Search\_\allowbreak{}Tool(query=\allowbreak{}"\allowbreak{}refillable clear glass foundation bottle with silver cap"\allowbreak{},\allowbreak{} search\_\allowbreak{}type=\allowbreak{}images)"\allowbreak{},\allowbreak{} "\allowbreak{}error"\allowbreak{}: null\}
\end{exverb}
\end{exstep}
\begin{exstep}{8}{\toolTS}{}
\exWalk{The agent tests a hypothesis, Jill Stuart, whose products have crystal-style caps; the results are concealers, not this bottle.}
\exLabel{Arguments}{}
\begin{exverb}
\{\\{}
~~"\allowbreak{}query"\allowbreak{}: "\allowbreak{}Jill Stuart foundation bottle crystal ornate silver cap round beige"\allowbreak{},\allowbreak{}\\{}
~~"\allowbreak{}search\_\allowbreak{}type"\allowbreak{}: "\allowbreak{}images"\allowbreak{}\\{}
\}
\end{exverb}
\exLabel{Tool output}{}
\begin{exverb}
Tool \textasciigrave{}Text\_\allowbreak{}Search\_\allowbreak{}Tool\textasciigrave{} (tool\_\allowbreak{}call\_\allowbreak{}id=\allowbreak{}call\_\allowbreak{}dE4zv498aQD0an48qVLBJ0jv) returned:\\{}
Image refs: tool\_\allowbreak{}generated\_\allowbreak{}image\_\allowbreak{}6,\allowbreak{} tool\_\allowbreak{}generated\_\allowbreak{}image\_\allowbreak{}7,\allowbreak{} tool\_\allowbreak{}generated\_\allowbreak{}image\_\allowbreak{}8\\{}
[Search Results — Type: images]\\{}
1. Diamond Tips Concealer \textbar{} PRODUCTS \textbar{} JILL STUART Beauty Official Site\\{}
~~~Image: [tool\_\allowbreak{}generated\_\allowbreak{}image\_\allowbreak{}6]\\{}
~~~Source: www.jillstuart-beauty.com   [webpage\_\allowbreak{}url\_\allowbreak{}2]\\{}
~~~Size: 640x640\\{}
2. JILL STUART Base Makeup New Items \textbar{} JILL STUART Beauty Official Site\\{}
~~~Image: [tool\_\allowbreak{}generated\_\allowbreak{}image\_\allowbreak{}7]\\{}
~~~Source: www.jillstuart-beauty.com   [webpage\_\allowbreak{}url\_\allowbreak{}3]\\{}
~~~Size: 750x1090
\end{exverb}
\exCut{1 more results not shown (3 in total).}
\begin{center}
\begin{minipage}[t]{0.475\linewidth}\centering\includegraphics[width=\linewidth,height=0.16\textheight,keepaspectratio]{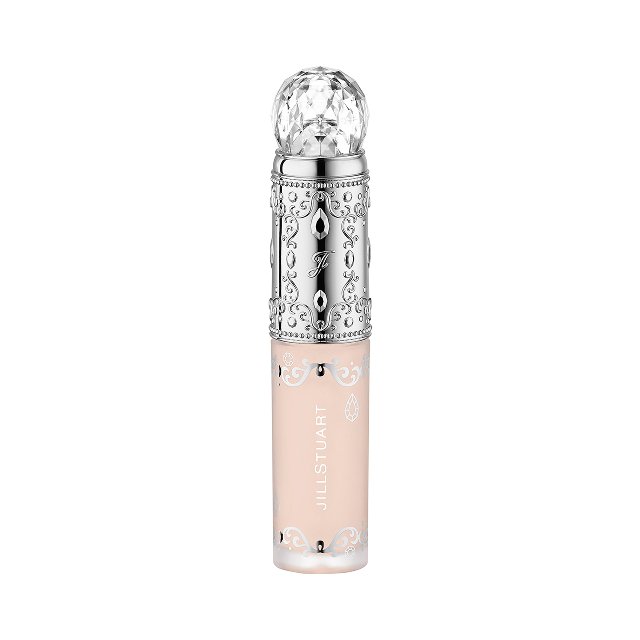}\end{minipage}\hfill\begin{minipage}[t]{0.475\linewidth}\centering\includegraphics[width=\linewidth,height=0.16\textheight,keepaspectratio]{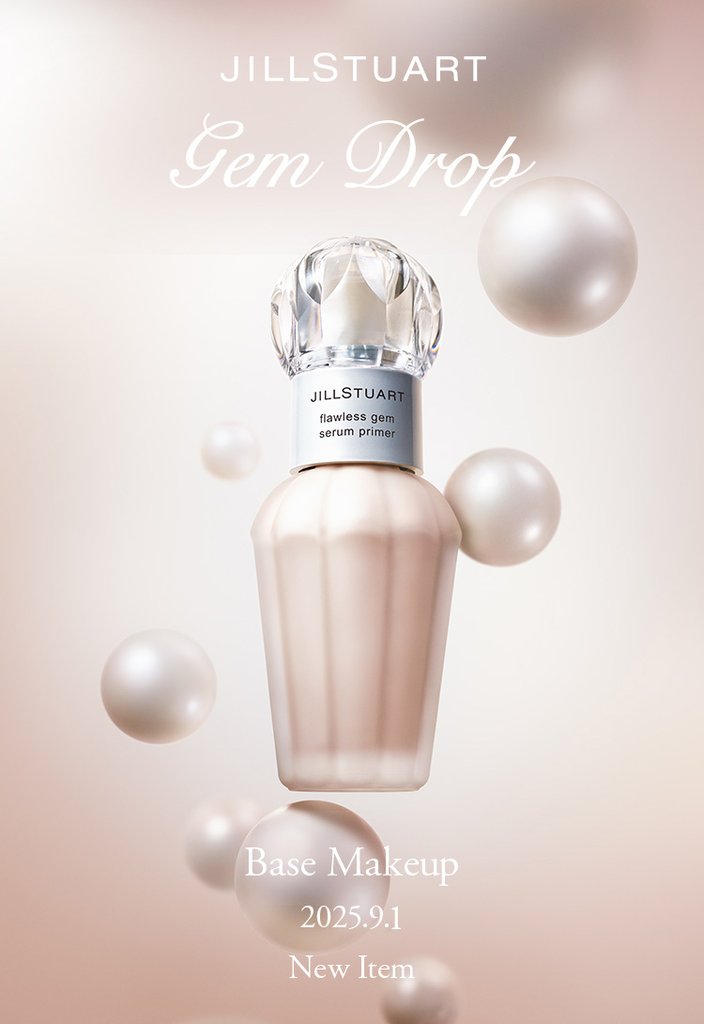}\end{minipage}
\\[1pt]{\scriptsize\color{exMute} Images returned with this output.}
\end{center}
\begin{exharness}
\exLabel{Harness}{environment update, summarized: new memory entries; new data bindings verbatim}
\begin{exverb}
Memory update: registers 3 new images in memory (tool\_\allowbreak{}generated\_\allowbreak{}image\_\allowbreak{}6 to tool\_\allowbreak{}generated\_\allowbreak{}image\_\allowbreak{}8,\allowbreak{} from Text\_\allowbreak{}Search\_\allowbreak{}Tool); registers 3 new webpage URLs in memory (webpage\_\allowbreak{}url\_\allowbreak{}2 to webpage\_\allowbreak{}url\_\allowbreak{}4,\allowbreak{} from Text\_\allowbreak{}Search\_\allowbreak{}Tool).
\end{exverb}
\end{exharness}
\end{exstep}
\begin{exstep}{9}{\toolTS}{}
\exWalk{A second Jill Stuart search does not match the bottle either, so the agent drops the hypothesis.}
\exLabel{Arguments}{}
\begin{exverb}
\{\\{}
~~"\allowbreak{}query"\allowbreak{}: "\allowbreak{}Jill Stuart liquid foundation glass bottle ornate crystal cap black base foundation"\allowbreak{},\allowbreak{}\\{}
~~"\allowbreak{}search\_\allowbreak{}type"\allowbreak{}: "\allowbreak{}images"\allowbreak{}\\{}
\}
\end{exverb}
\exLabel{Tool output}{}
\begin{exverb}
Tool \textasciigrave{}Text\_\allowbreak{}Search\_\allowbreak{}Tool\textasciigrave{} (tool\_\allowbreak{}call\_\allowbreak{}id=\allowbreak{}call\_\allowbreak{}IjxBHLUO1w76MLNjNAudQW3e) returned:\\{}
Image refs: tool\_\allowbreak{}generated\_\allowbreak{}image\_\allowbreak{}9,\allowbreak{} tool\_\allowbreak{}generated\_\allowbreak{}image\_\allowbreak{}10,\allowbreak{} tool\_\allowbreak{}generated\_\allowbreak{}image\_\allowbreak{}11\\{}
[Search Results — Type: images]\\{}
1. JILL STUART Diamond Tips Concealer Cover type 5.3mL Japan\\{}
~~~Image: [tool\_\allowbreak{}generated\_\allowbreak{}image\_\allowbreak{}9]\\{}
~~~Source: eBay   [webpage\_\allowbreak{}url\_\allowbreak{}5]\\{}
~~~Size: 923x1200\\{}
2. JILL STUART Base Makeup New Items \& Limited Items \textbar{} NEW ITEM \textbar{} JILL STUART  Beauty Official Site\\{}
~~~Image: [tool\_\allowbreak{}generated\_\allowbreak{}image\_\allowbreak{}10]\\{}
~~~Source: JILL STUART   [webpage\_\allowbreak{}url\_\allowbreak{}6]\\{}
~~~Size: 650x618
\end{exverb}
\exCut{1 more results not shown (3 in total).}
\begin{center}
\begin{minipage}[t]{0.475\linewidth}\centering\includegraphics[width=\linewidth,height=0.16\textheight,keepaspectratio]{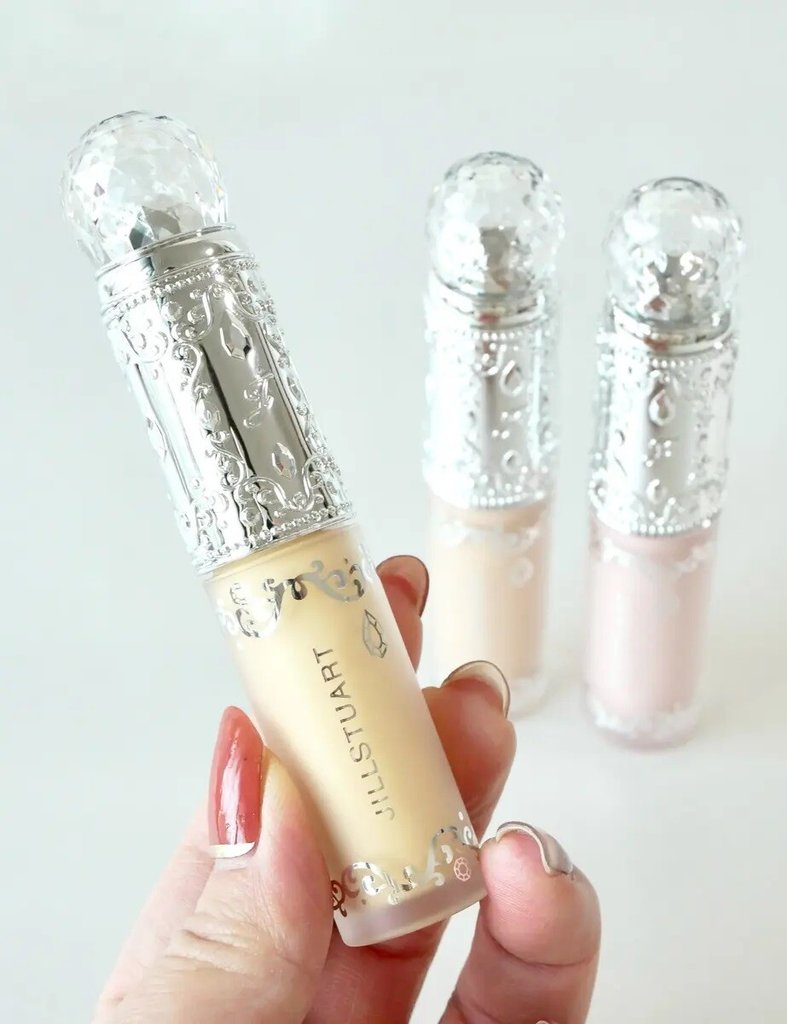}\end{minipage}\hfill\begin{minipage}[t]{0.475\linewidth}\centering\includegraphics[width=\linewidth,height=0.16\textheight,keepaspectratio]{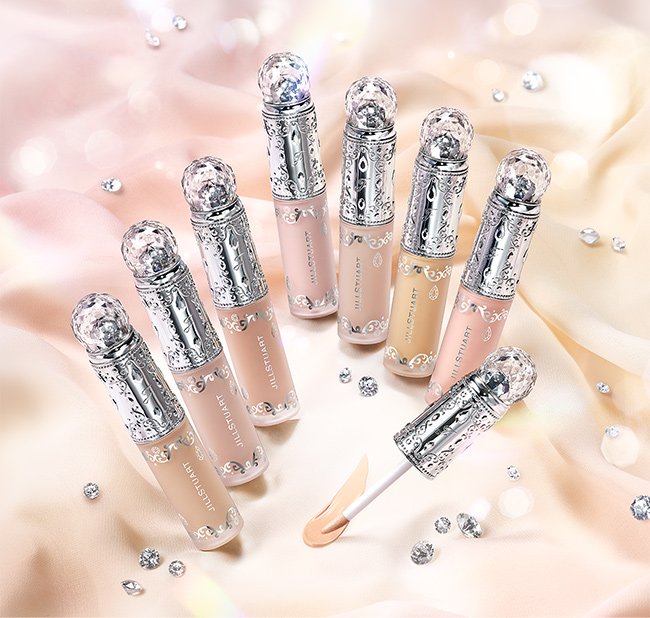}\end{minipage}
\\[1pt]{\scriptsize\color{exMute} Images returned with this output.}
\end{center}
\begin{exharness}
\exLabel{Harness}{environment update, summarized: new memory entries; new data bindings verbatim}
\begin{exverb}
Memory update: registers 3 new images in memory (tool\_\allowbreak{}generated\_\allowbreak{}image\_\allowbreak{}9 to tool\_\allowbreak{}generated\_\allowbreak{}image\_\allowbreak{}11,\allowbreak{} from Text\_\allowbreak{}Search\_\allowbreak{}Tool); registers 3 new webpage URLs in memory (webpage\_\allowbreak{}url\_\allowbreak{}5 to webpage\_\allowbreak{}url\_\allowbreak{}7,\allowbreak{} from Text\_\allowbreak{}Search\_\allowbreak{}Tool).
\end{exverb}
\end{exharness}
\end{exstep}
\begin{exstep}{10}{\toolTS}{}
\exWalk{Another descriptive search returns only empty stock bottles.}
\exLabel{Arguments}{}
\begin{exverb}
\{\\{}
~~"\allowbreak{}query"\allowbreak{}: "\allowbreak{}foundation bottle \textbackslash{}"\allowbreak{}black\textbackslash{}"\allowbreak{} \textbackslash{}"\allowbreak{}crystal\textbackslash{}"\allowbreak{} cap \textbackslash{}"\allowbreak{}silver\textbackslash{}"\allowbreak{} round liquid foundation"\allowbreak{},\allowbreak{}\\{}
~~"\allowbreak{}search\_\allowbreak{}type"\allowbreak{}: "\allowbreak{}images"\allowbreak{}\\{}
\}
\end{exverb}
\exLabel{Tool output}{}
\begin{exverb}
Tool \textasciigrave{}Text\_\allowbreak{}Search\_\allowbreak{}Tool\textasciigrave{} (tool\_\allowbreak{}call\_\allowbreak{}id=\allowbreak{}call\_\allowbreak{}xM5R3uof4FvmZ1yv5tkzAkW6) returned:\\{}
Image refs: tool\_\allowbreak{}generated\_\allowbreak{}image\_\allowbreak{}12,\allowbreak{} tool\_\allowbreak{}generated\_\allowbreak{}image\_\allowbreak{}13,\allowbreak{} tool\_\allowbreak{}generated\_\allowbreak{}image\_\allowbreak{}14\\{}
[Search Results — Type: images]\\{}
1. 30ml Custom Logo Oem Plastic Empty Cosmetic Liquid Foundation Bottle Round  Silver Clear Foundation Stick With Huge Brush Wand - Buy Product on  Alibaba.com\\{}
~~~Image: [tool\_\allowbreak{}generated\_\allowbreak{}image\_\allowbreak{}12]\\{}
~~~Source: Alibaba.com   [webpage\_\allowbreak{}url\_\allowbreak{}8]\\{}
~~~Size: 2268x2268\\{}
2. 30ml Custom Logo Oem Plastic Empty Cosmetic Liquid Foundation Bottle Round  Silver Clear Foundation Stick With Huge Brush Wand - Buy Product on  Alibaba.com\\{}
~~~Image: [tool\_\allowbreak{}generated\_\allowbreak{}image\_\allowbreak{}13]\\{}
~~~Source: Alibaba.com   [webpage\_\allowbreak{}url\_\allowbreak{}8]\\{}
~~~Size: 800x800
\end{exverb}
\exCut{1 more results not shown (3 in total).}
\begin{center}
\begin{minipage}[t]{0.475\linewidth}\centering\includegraphics[width=\linewidth,height=0.16\textheight,keepaspectratio]{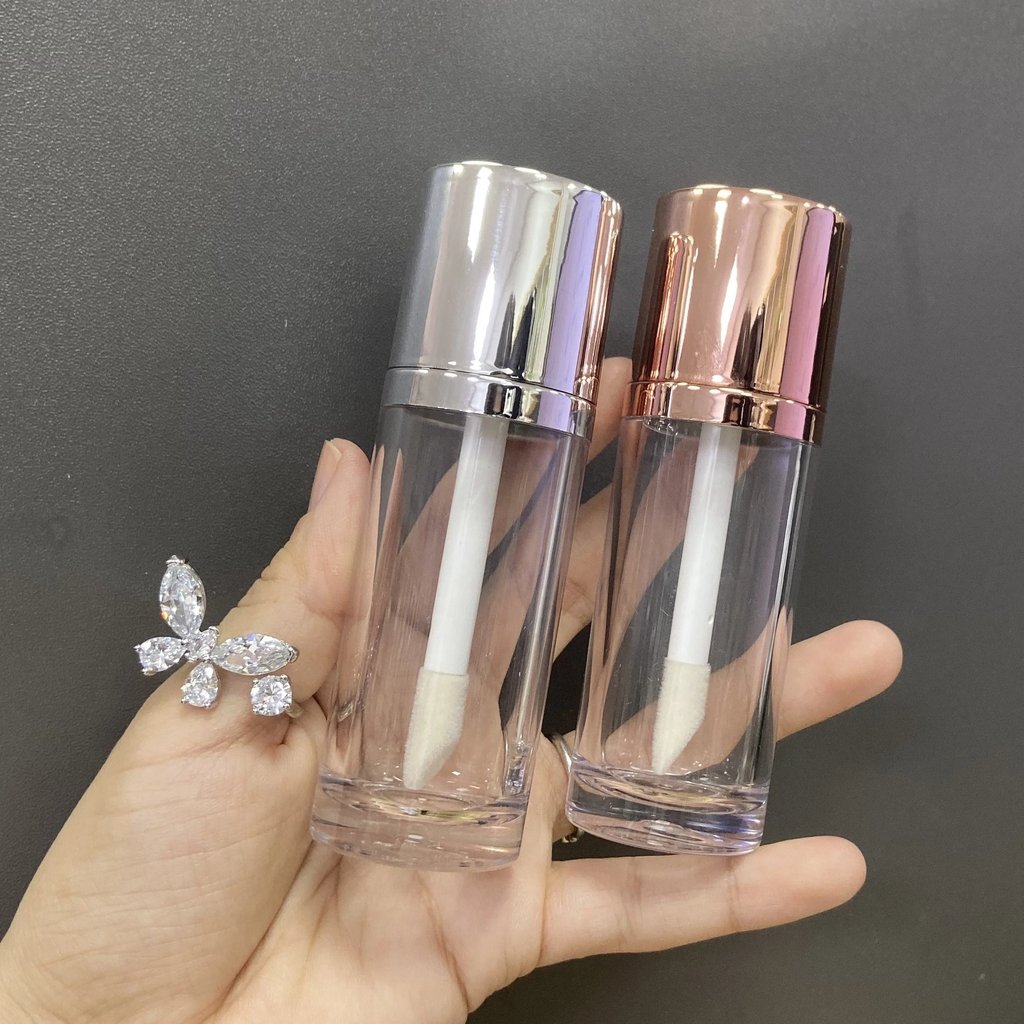}\end{minipage}\hfill\begin{minipage}[t]{0.475\linewidth}\centering\includegraphics[width=\linewidth,height=0.16\textheight,keepaspectratio]{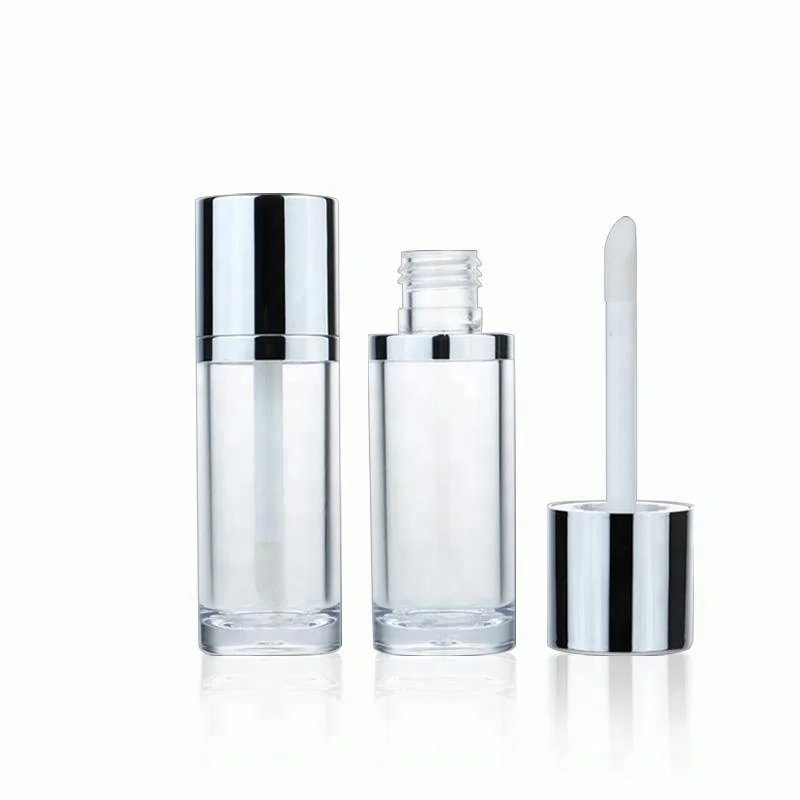}\end{minipage}
\\[1pt]{\scriptsize\color{exMute} Images returned with this output.}
\end{center}
\begin{exharness}
\exLabel{Harness}{environment update, summarized: new memory entries; new data bindings verbatim}
\begin{exverb}
Memory update: registers 3 new images in memory (tool\_\allowbreak{}generated\_\allowbreak{}image\_\allowbreak{}12 to tool\_\allowbreak{}generated\_\allowbreak{}image\_\allowbreak{}14,\allowbreak{} from Text\_\allowbreak{}Search\_\allowbreak{}Tool); registers 2 new webpage URLs in memory (webpage\_\allowbreak{}url\_\allowbreak{}8 to webpage\_\allowbreak{}url\_\allowbreak{}9,\allowbreak{} from Text\_\allowbreak{}Search\_\allowbreak{}Tool).
\end{exverb}
\end{exharness}
\end{exstep}
\begin{exstep}{11}{\toolIS}{}
\exWalk{A reverse search on the tight crop from Step~4. The tool's summary identifies a ZOYA nail polish and ``a KATOKATO foundation bottle''.}
\exLabel{Arguments}{}
\begin{exverb}
\{\\{}
~~"\allowbreak{}image\_\allowbreak{}id"\allowbreak{}: "\allowbreak{}tool\_\allowbreak{}generated\_\allowbreak{}image\_\allowbreak{}2"\allowbreak{}\\{}
\}
\end{exverb}
\exLabel{Tool output}{}
\begin{exverb}
Tool \textasciigrave{}Image\_\allowbreak{}Search\_\allowbreak{}Tool\textasciigrave{} (tool\_\allowbreak{}call\_\allowbreak{}id=\allowbreak{}call\_\allowbreak{}iL45p6mk8iTJWkV0ud1xLoi9) returned:\\{}
Image refs: tool\_\allowbreak{}generated\_\allowbreak{}image\_\allowbreak{}15,\allowbreak{} tool\_\allowbreak{}generated\_\allowbreak{}image\_\allowbreak{}16,\allowbreak{} tool\_\allowbreak{}generated\_\allowbreak{}image\_\allowbreak{}17\\{}
[Image Search — Visual Matches]\\{}
1. Zoya Bridal Bliss Collection (Part II) Swatch and Review - LacquerExpression\\{}
~~~Source: lacquerexpression.com/\allowbreak{}2018/\allowbreak{}06/\allowbreak{}mainstream-swatch-and-review-zoya\_\allowbreak{}28.html   [webpage\_\allowbreak{}url\_\allowbreak{}10]\\{}
~~~Image: [tool\_\allowbreak{}generated\_\allowbreak{}image\_\allowbreak{}15]\\{}
2. Amazon.com: ZOYA Nail Polish,\allowbreak{} Maxine,\allowbreak{} 0.5 fl. oz. : Beauty \& Personal Care\\{}
~~~Source: amazon.com/\allowbreak{}ZOYA-Nail-Polish-Maxine-0-5/\allowbreak{}dp/\allowbreak{}B07HHN2N6C   [webpage\_\allowbreak{}url\_\allowbreak{}11]\\{}
~~~Image: [tool\_\allowbreak{}generated\_\allowbreak{}image\_\allowbreak{}16]
\end{exverb}
\exCut{1 more results not shown (3 in total).}
\begin{center}
\begin{minipage}[t]{0.475\linewidth}\centering\includegraphics[width=\linewidth,height=0.16\textheight,keepaspectratio]{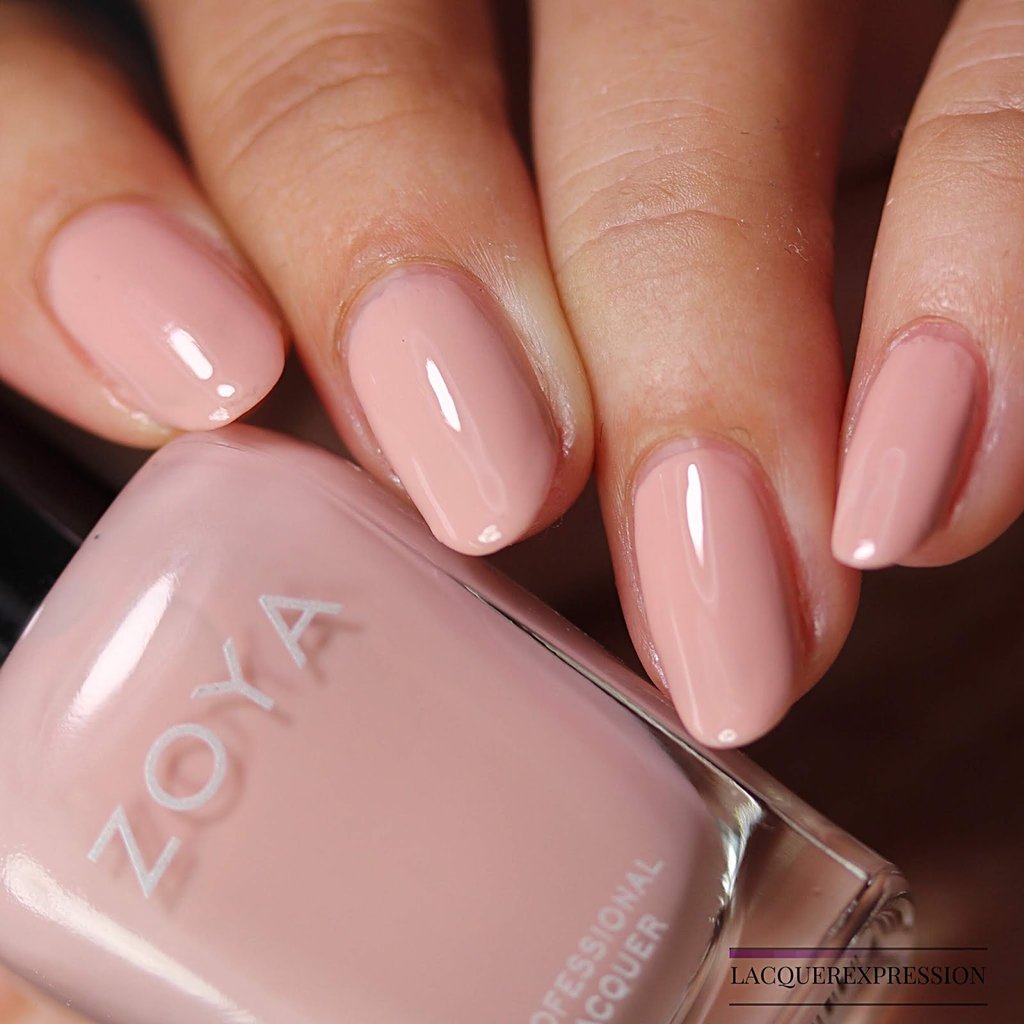}\end{minipage}\hfill\begin{minipage}[t]{0.475\linewidth}\centering\includegraphics[width=\linewidth,height=0.16\textheight,keepaspectratio]{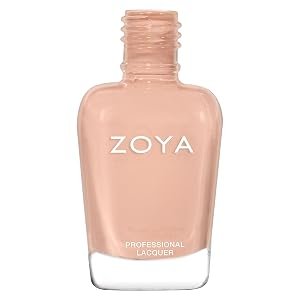}\end{minipage}
\\[1pt]{\scriptsize\color{exMute} Images returned with this output.}
\end{center}
\begin{exharness}
\exLabel{Harness}{environment update, summarized: new memory entries; new data bindings verbatim}
\begin{exverb}
Memory update: registers 3 new images in memory (tool\_\allowbreak{}generated\_\allowbreak{}image\_\allowbreak{}15 to tool\_\allowbreak{}generated\_\allowbreak{}image\_\allowbreak{}17,\allowbreak{} from Image\_\allowbreak{}Search\_\allowbreak{}Tool); registers 3 new webpage URLs in memory (webpage\_\allowbreak{}url\_\allowbreak{}10 to webpage\_\allowbreak{}url\_\allowbreak{}12,\allowbreak{} from Image\_\allowbreak{}Search\_\allowbreak{}Tool).
\end{exverb}
\end{exharness}
\end{exstep}
\begin{exstep}{12}{\toolTS}{}
\exWalk{A targeted image search for KATOKATO foundation returns a KATO-KATO product whose crystal cap matches the crop.}
\exLabel{Arguments}{}
\begin{exverb}
\{\\{}
~~"\allowbreak{}query"\allowbreak{}: "\allowbreak{}\textbackslash{}"\allowbreak{}KATOKATO\textbackslash{}"\allowbreak{} liquid foundation crystal cap bottle"\allowbreak{},\allowbreak{}\\{}
~~"\allowbreak{}search\_\allowbreak{}type"\allowbreak{}: "\allowbreak{}images"\allowbreak{}\\{}
\}
\end{exverb}
\exLabel{Tool output}{}
\begin{exverb}
Tool \textasciigrave{}Text\_\allowbreak{}Search\_\allowbreak{}Tool\textasciigrave{} (tool\_\allowbreak{}call\_\allowbreak{}id=\allowbreak{}call\_\allowbreak{}ILCAxaED23QcWvUq1eDpGdmQ) returned:\\{}
Image refs: tool\_\allowbreak{}generated\_\allowbreak{}image\_\allowbreak{}18\\{}
[Search Results — Type: images]\\{}
1. KATO-KATO – Chic Decent\\{}
~~~Image: [tool\_\allowbreak{}generated\_\allowbreak{}image\_\allowbreak{}18]\\{}
~~~Source: Chic Decent   [webpage\_\allowbreak{}url\_\allowbreak{}13]\\{}
~~~Size: 645x645\\{}
\mbox{}\\{}
[Summary]\\{}
The primary candidate (rank 1) from Chic Decent clearly shows the KATO-KATO liquid foundation with its distinctive crystal cap,\allowbreak{} matching the query exactly. The second candidate (rank 2) displays a KATO-KATO product but it is a \textquotesingle{}Misty Project\textquotesingle{} spray with a black cap,\allowbreak{} not a liquid foundation. The third candidate (rank 3) is a \textquotesingle{}Cloudy to Misty Lightweight Setting Spray\textquotesingle{} also with a black cap,\allowbreak{} confirming it is not the requested foundation. All three sources are from Chic Decent or The Hype Cosmetics Outlet,\allowbreak{} both retailing KATO-KATO products,\allowbreak{} but only the first image correctly depicts the liquid foundation with the crystal cap as specified in the query.\\{}
tool\_\allowbreak{}generated\_\allowbreak{}image\_\allowbreak{}18
\end{exverb}
\begin{center}\includegraphics[width=0.45\linewidth,height=0.22\textheight,keepaspectratio]{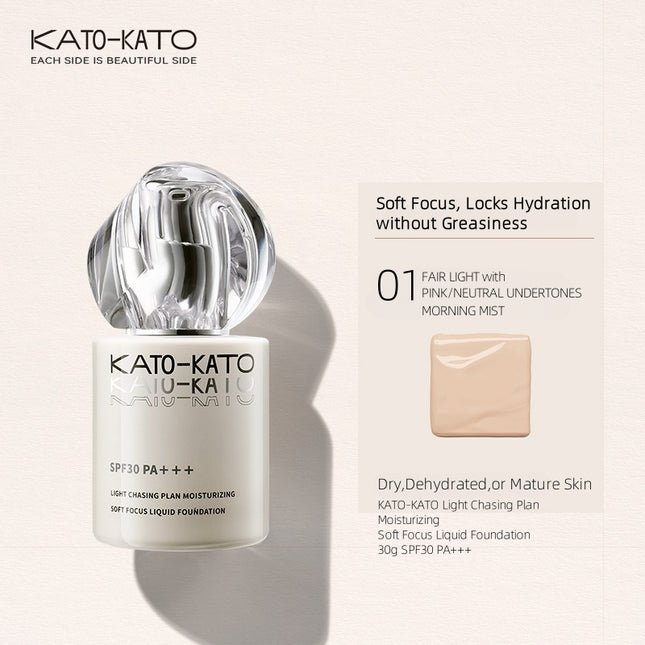}\\[2pt]
{\scriptsize\color{exMute} Image returned with this output.}\end{center}
\begin{exharness}
\exLabel{Harness}{environment update, summarized: new memory entries; new data bindings verbatim}
\begin{exverb}
Memory update: registers 1 new image in memory (tool\_\allowbreak{}generated\_\allowbreak{}image\_\allowbreak{}18,\allowbreak{} from Text\_\allowbreak{}Search\_\allowbreak{}Tool); registers 1 new webpage URL in memory (webpage\_\allowbreak{}url\_\allowbreak{}13,\allowbreak{} from Text\_\allowbreak{}Search\_\allowbreak{}Tool).
\end{exverb}
\end{exharness}
\end{exstep}
\begin{exstep}{}{\exTool{exInk}{Final answer}}{}
\exWalk{The agent answers KATO-KATO, which matches the ground truth, KATO.}
\exLabel{Model}{}
\begin{exverb}
\textless{}answer\textgreater{}KATO-KATO\textless{}/\allowbreak{}answer\textgreater{}
\end{exverb}
\end{exstep}

%% file: text/appendix_examples_steps/fvqa_header.tex
\exOverview{\textbf{Task.} A photograph shows two hikers resting with their backpacks. What awards has this product won? 

\textbf{Why it is hard.} The product is not named anywhere, and the logos on the backpacks are too small to read. 

\textbf{How the agent solves it.} It grounds the two backpacks, tries to read each with zoom and OCR, and, when that fails, reverse-searches the crop of the green backpack, which identifies it as a Deuter Speed Lite Pro. A product page mentions no award, so it searches for awards and verifies the result on the manufacturer's press release: the ISPO Award 2024. Nine calls use six different tools.}
\exLabel{Input images}{}
\begin{center}\includegraphics[width=0.55\linewidth,height=0.28\textheight,keepaspectratio]{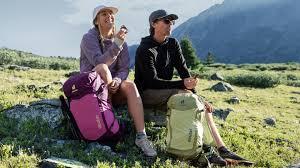}\\[2pt]
{\scriptsize\color{exMute} input\_image\_1}\end{center}
\exLabel{Question}{as sent to the model}
\begin{exverb}
Question: What awards has this product won?
\end{exverb}

%% file: text/appendix_examples_steps/fvqa.tex
\begin{exharness}
\exLabel{Harness}{environment update, shown in full}
\begin{exverb}
Environment Update (not a new question):\\{}
Currently available image IDs for tool calls:\\{}
- input\_\allowbreak{}image\_\allowbreak{}1: user\_\allowbreak{}input\\{}
When a tool needs image input,\allowbreak{} pass one of these IDs via the *\_\allowbreak{}image\_\allowbreak{}id field.\\{}
Do NOT invent image file names or paths.
\end{exverb}
\end{exharness}
\begin{exstep}{1}{\toolVG}{}
\exLabel{Model}{text emitted with this call}
\exSayBlock{\textless{}thinking\textgreater{}The image shows two people sitting outdoors with backpacks,\allowbreak{} likely hiking or camping. There is no clear indication of a specific product being advertised or labeled in the image. Without identifying the exact product,\allowbreak{} it’s impossible to determine any awards it may have won. I need to identify the product first.\textless{}/\allowbreak{}thinking\textgreater{}}
\exLabel{Arguments}{}
\begin{exverb}
\{\\{}
~~"\allowbreak{}query"\allowbreak{}: "\allowbreak{}backpack"\allowbreak{},\allowbreak{}\\{}
~~"\allowbreak{}image\_\allowbreak{}id"\allowbreak{}: "\allowbreak{}input\_\allowbreak{}image\_\allowbreak{}1"\allowbreak{}\\{}
\}
\end{exverb}
\exLabel{Tool output}{}
\begin{exverb}
Visual\_\allowbreak{}Grounding\_\allowbreak{}Tool grounded 2 objects for the prompt \textquotesingle{}backpack\textquotesingle{}. Their corresponding bounding boxes (on a 0-1000 scale,\allowbreak{} [x1,\allowbreak{} y1,\allowbreak{} x2,\allowbreak{} y2]) are: \{\textquotesingle{}object 1\textquotesingle{}: [199.49,\allowbreak{} 423.01,\allowbreak{} 407.82,\allowbreak{} 875.1],\allowbreak{} \textquotesingle{}object 2\textquotesingle{}: [552.17,\allowbreak{} 550.46,\allowbreak{} 684.45,\allowbreak{} 951.58]\}. The object numbers correspond to the labeled masks rendered in the following image,\allowbreak{} which you can refer to as tool\_\allowbreak{}generated\_\allowbreak{}image\_\allowbreak{}1.
\end{exverb}
\begin{center}\includegraphics[width=0.45\linewidth,height=0.22\textheight,keepaspectratio]{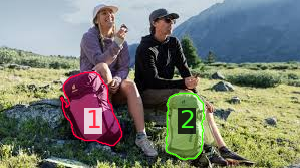}\\[2pt]
{\scriptsize\color{exMute} Image returned with this output.}\end{center}
\begin{exharness}
\exLabel{Harness}{environment update, summarized: new memory entries; new data bindings verbatim}
\begin{exverb}
Memory update: registers 1 new image in memory (tool\_\allowbreak{}generated\_\allowbreak{}image\_\allowbreak{}1,\allowbreak{} from Visual\_\allowbreak{}Grounding\_\allowbreak{}Tool).\\{}
Available data bindings (auto-injected into Python\_\allowbreak{}Coding\_\allowbreak{}Agent\_\allowbreak{}Tool):\\{}
- segment\_\allowbreak{}backpack: dict (keys: pred\_\allowbreak{}boxes,\allowbreak{} pred\_\allowbreak{}scores,\allowbreak{} pred\_\allowbreak{}masks,\allowbreak{} num\_\allowbreak{}masks,\allowbreak{} orig\_\allowbreak{}img\_\allowbreak{}h,\allowbreak{} orig\_\allowbreak{}img\_\allowbreak{}w)
\end{exverb}
\end{exharness}
\end{exstep}
\begin{exstep}{2}{\toolZoom}{}
\exLabel{Model}{text emitted with this call}
\exSayBlock{\textless{}thinking\textgreater{}I have identified two backpacks in the image. To determine which product might have won awards,\allowbreak{} I need to examine them more closely. I\textquotesingle{}ll start by zooming in on the purple backpack (labeled as object 1) to see if there are any visible brand names,\allowbreak{} logos,\allowbreak{} or labels that could help identify the product.\textless{}/\allowbreak{}thinking\textgreater{}}
\exLabel{Arguments}{}
\begin{exverb}
\{\\{}
~~"\allowbreak{}image\_\allowbreak{}id"\allowbreak{}: "\allowbreak{}input\_\allowbreak{}image\_\allowbreak{}1"\allowbreak{},\allowbreak{}\\{}
~~"\allowbreak{}coords"\allowbreak{}: [199,\allowbreak{} 423,\allowbreak{} 408,\allowbreak{} 875],\allowbreak{}\\{}
~~"\allowbreak{}padding"\allowbreak{}: 50\\{}
\}
\end{exverb}
\exLabel{Tool output}{}
\begin{exverb}
Tool \textasciigrave{}Zoom\_\allowbreak{}In\_\allowbreak{}Tool\textasciigrave{} (tool\_\allowbreak{}call\_\allowbreak{}id=\allowbreak{}chatcmpl-tool-92335cf0b5261d44) returned:\\{}
Zoomed in image rendered below; refer to it as tool\_\allowbreak{}generated\_\allowbreak{}image\_\allowbreak{}2.
\end{exverb}
\begin{center}\includegraphics[width=0.45\linewidth,height=0.22\textheight,keepaspectratio]{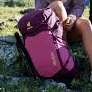}\\[2pt]
{\scriptsize\color{exMute} Image returned with this output.}\end{center}
\begin{exharness}
\exLabel{Harness}{environment update, summarized: new memory entries; new data bindings verbatim}
\begin{exverb}
Memory update: registers 1 new image in memory (tool\_\allowbreak{}generated\_\allowbreak{}image\_\allowbreak{}2,\allowbreak{} from Zoom\_\allowbreak{}In\_\allowbreak{}Tool).
\end{exverb}
\end{exharness}
\end{exstep}
\begin{exstep}{3}{\toolOCR}{}
\exLabel{Model}{text emitted with this call}
\exSayBlock{\textless{}thinking\textgreater{}I\textquotesingle{}ve zoomed in on the purple backpack,\allowbreak{} but the details are still not clear enough to read any text or logos. I\textquotesingle{}ll now attempt to use OCR to extract any text from this zoomed-in image to identify the brand or model of the backpack.\textless{}/\allowbreak{}thinking\textgreater{}}
\exLabel{Arguments}{}
\begin{exverb}
\{\\{}
~~"\allowbreak{}image\_\allowbreak{}id"\allowbreak{}: "\allowbreak{}tool\_\allowbreak{}generated\_\allowbreak{}image\_\allowbreak{}2"\allowbreak{}\\{}
\}
\end{exverb}
\exLabel{Tool output}{}
\begin{exverb}
OCR\_\allowbreak{}Tool did not transcribe any readable text.
\end{exverb}
\end{exstep}
\begin{exstep}{4}{\toolZoom}{}
\exLabel{Model}{text emitted with this call}
\exSayBlock{\textless{}thinking\textgreater{}The OCR tool failed to extract any text from the zoomed-in image of the purple backpack. I will now zoom in on the green backpack (labeled as object 2) to check for any visible branding or labels.\textless{}/\allowbreak{}thinking\textgreater{}}
\exLabel{Arguments}{}
\begin{exverb}
\{\\{}
~~"\allowbreak{}image\_\allowbreak{}id"\allowbreak{}: "\allowbreak{}input\_\allowbreak{}image\_\allowbreak{}1"\allowbreak{},\allowbreak{}\\{}
~~"\allowbreak{}coords"\allowbreak{}: [552,\allowbreak{} 550,\allowbreak{} 684,\allowbreak{} 951],\allowbreak{}\\{}
~~"\allowbreak{}padding"\allowbreak{}: 50\\{}
\}
\end{exverb}
\exLabel{Tool output}{}
\begin{exverb}
Tool \textasciigrave{}Zoom\_\allowbreak{}In\_\allowbreak{}Tool\textasciigrave{} (tool\_\allowbreak{}call\_\allowbreak{}id=\allowbreak{}chatcmpl-tool-bd7bc329779eaddf) returned:\\{}
Zoomed in image rendered below; refer to it as tool\_\allowbreak{}generated\_\allowbreak{}image\_\allowbreak{}3.
\end{exverb}
\begin{center}\includegraphics[width=0.45\linewidth,height=0.22\textheight,keepaspectratio]{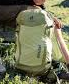}\\[2pt]
{\scriptsize\color{exMute} Image returned with this output.}\end{center}
\begin{exharness}
\exLabel{Harness}{environment update, summarized: new memory entries; new data bindings verbatim}
\begin{exverb}
Memory update: registers 1 new image in memory (tool\_\allowbreak{}generated\_\allowbreak{}image\_\allowbreak{}3,\allowbreak{} from Zoom\_\allowbreak{}In\_\allowbreak{}Tool).
\end{exverb}
\end{exharness}
\end{exstep}
\begin{exstep}{5}{\toolOCR}{}
\exLabel{Model}{text emitted with this call}
\exSayBlock{\textless{}thinking\textgreater{}I\textquotesingle{}ve zoomed in on the green backpack,\allowbreak{} and it appears to have some text or a logo on it. I will now use OCR to extract any readable text from this zoomed-in image to identify the brand or model.\textless{}/\allowbreak{}thinking\textgreater{}}
\exLabel{Arguments}{}
\begin{exverb}
\{\\{}
~~"\allowbreak{}image\_\allowbreak{}id"\allowbreak{}: "\allowbreak{}tool\_\allowbreak{}generated\_\allowbreak{}image\_\allowbreak{}3"\allowbreak{}\\{}
\}
\end{exverb}
\exLabel{Tool output}{}
\begin{exverb}
OCR\_\allowbreak{}Tool transcribed the following text from the image:\\{}
\mbox{}\\{}
-
\end{exverb}
\end{exstep}
\begin{exstep}{6}{\toolIS}{}
\exLabel{Model}{text emitted with this call}
\exSayBlock{\textless{}thinking\textgreater{}Despite zooming in and attempting OCR,\allowbreak{} no readable text was extracted from either backpack. Since the visual inspection hasn\textquotesingle{}t yielded any identifiable brand or product name,\allowbreak{} I will now use an image search to identify the backpacks based on their appearance.\textless{}/\allowbreak{}thinking\textgreater{}}
\exLabel{Arguments}{}
\begin{exverb}
\{\\{}
~~"\allowbreak{}image\_\allowbreak{}id"\allowbreak{}: "\allowbreak{}tool\_\allowbreak{}generated\_\allowbreak{}image\_\allowbreak{}3"\allowbreak{}\\{}
\}
\end{exverb}
\exLabel{Tool output}{}
\begin{exverb}
Tool \textasciigrave{}Image\_\allowbreak{}Search\_\allowbreak{}Tool\textasciigrave{} (tool\_\allowbreak{}call\_\allowbreak{}id=\allowbreak{}chatcmpl-tool-bd7ce98f66d6d010) returned:\\{}
Image refs: tool\_\allowbreak{}generated\_\allowbreak{}image\_\allowbreak{}4,\allowbreak{} tool\_\allowbreak{}generated\_\allowbreak{}image\_\allowbreak{}5,\allowbreak{} tool\_\allowbreak{}generated\_\allowbreak{}image\_\allowbreak{}6\\{}
[Image Search — Visual Matches]\\{}
1. DEUTER Speed Lite Pro 25 Hiking Backpack in Mineral Grove \textbar{} eBay Australia\\{}
~~~Source: ebay.com.au/\allowbreak{}itm/\allowbreak{}388231206979   [webpage\_\allowbreak{}url\_\allowbreak{}1]\\{}
~~~Image: [tool\_\allowbreak{}generated\_\allowbreak{}image\_\allowbreak{}4]\\{}
2. deuter Men\textquotesingle{}s Speed Lite Pro 19 Hiking Backpack : Amazon.ca: Sports \& Outdoors\\{}
~~~Source: amazon.ca/\allowbreak{}deuter-Mens-Speed-Hiking-Backpack/\allowbreak{}dp/\allowbreak{}B0F79T5CY9   [webpage\_\allowbreak{}url\_\allowbreak{}2]\\{}
~~~Image: [tool\_\allowbreak{}generated\_\allowbreak{}image\_\allowbreak{}5]
\end{exverb}
\exCut{1 more results not shown (3 in total).}
\begin{center}
\begin{minipage}[t]{0.312\linewidth}\centering\includegraphics[width=\linewidth,height=0.16\textheight,keepaspectratio]{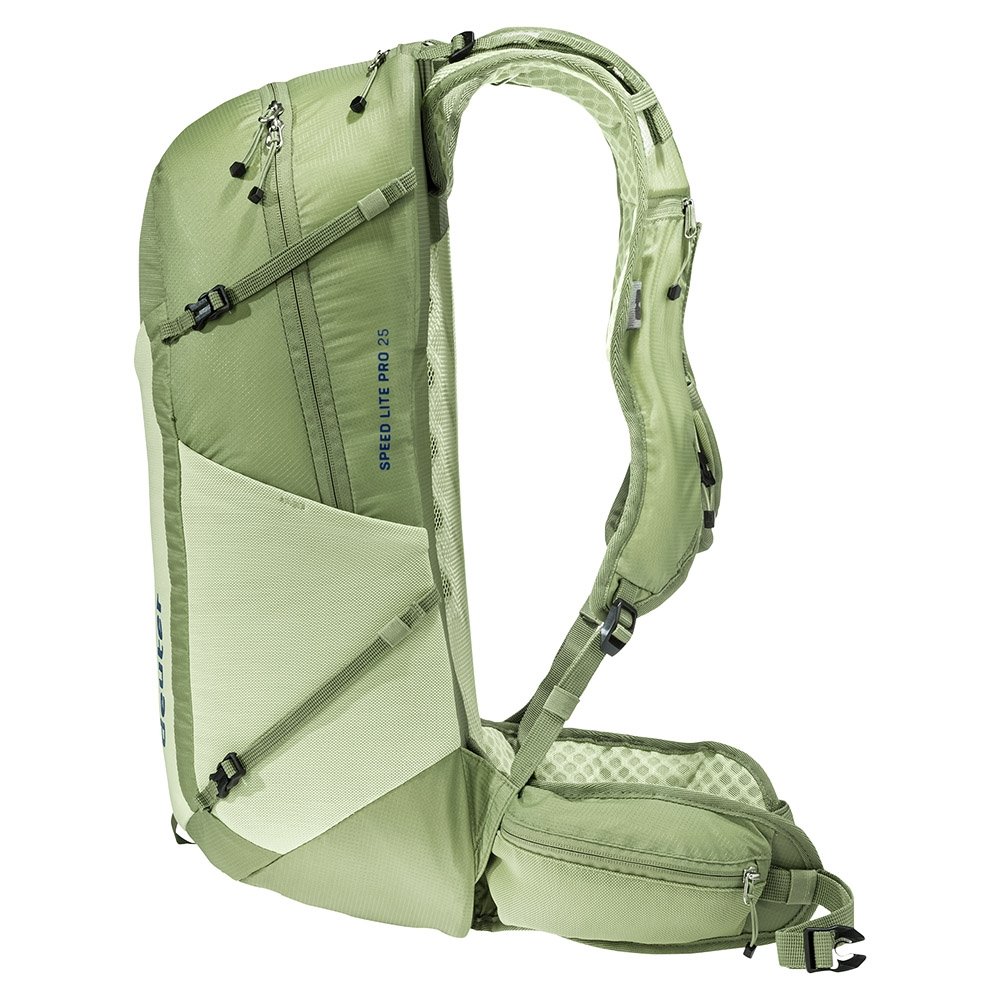}\end{minipage}\hfill\begin{minipage}[t]{0.312\linewidth}\centering\includegraphics[width=\linewidth,height=0.16\textheight,keepaspectratio]{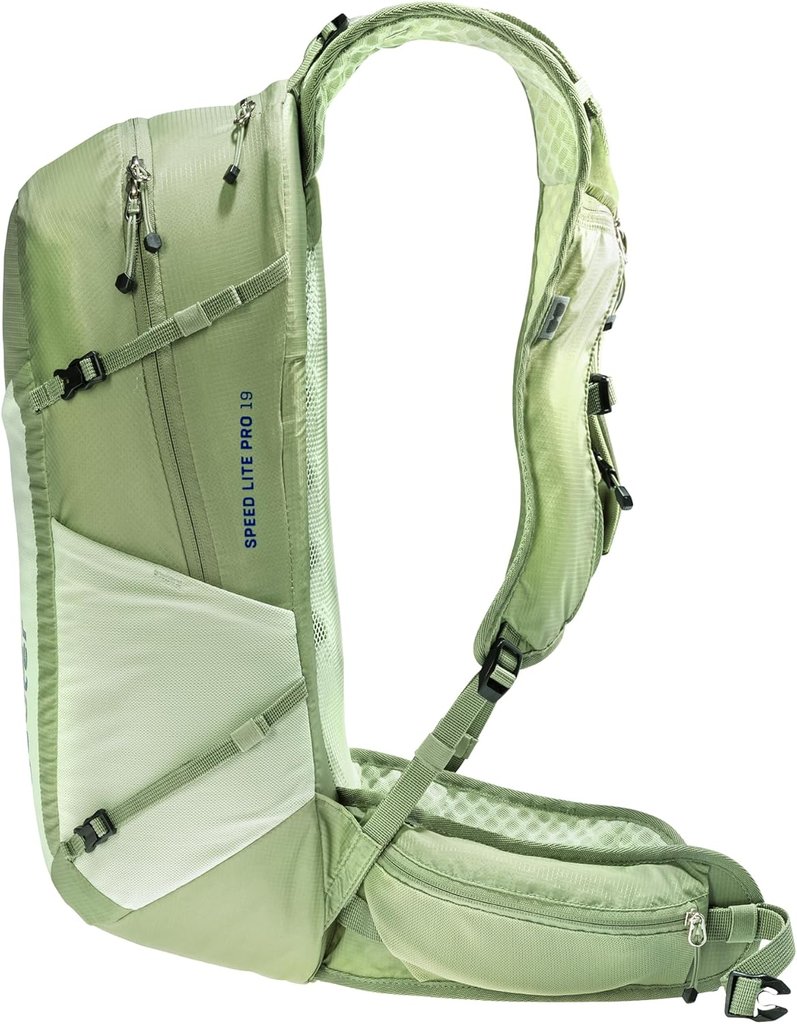}\end{minipage}\hfill\begin{minipage}[t]{0.312\linewidth}\centering\includegraphics[width=\linewidth,height=0.16\textheight,keepaspectratio]{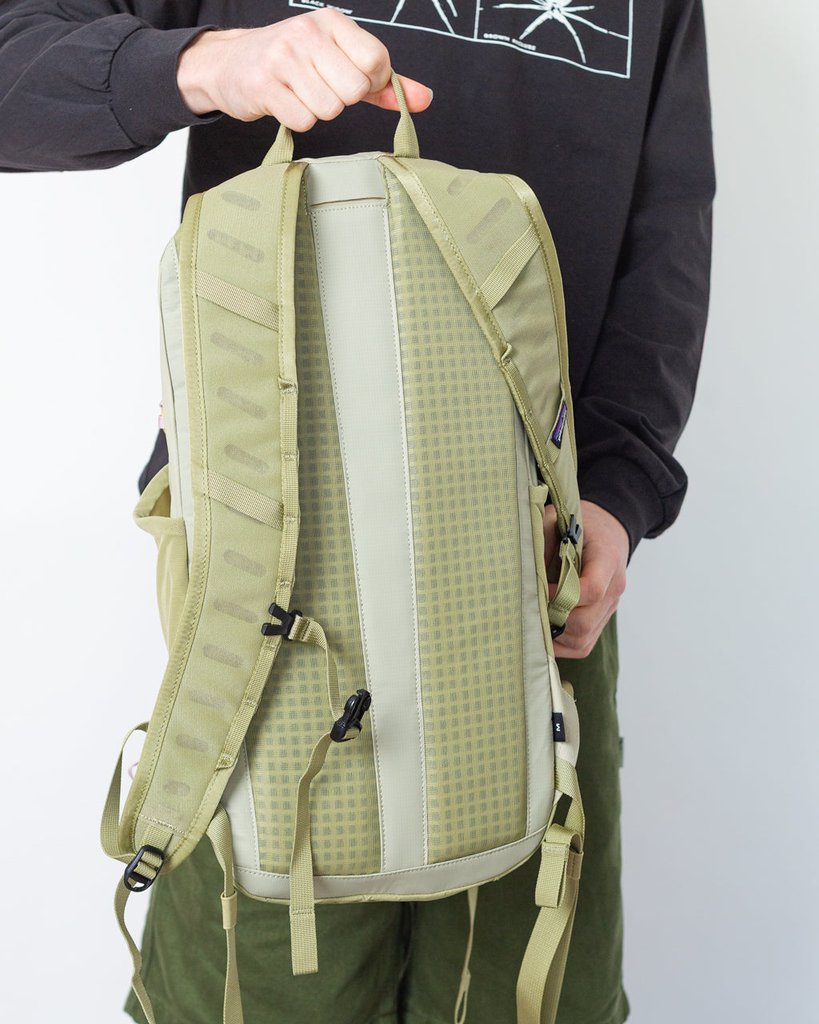}\end{minipage}
\\[1pt]{\scriptsize\color{exMute} Images returned with this output.}
\end{center}
\begin{exharness}
\exLabel{Harness}{environment update, summarized: new memory entries; new data bindings verbatim}
\begin{exverb}
Memory update: registers 3 new images in memory (tool\_\allowbreak{}generated\_\allowbreak{}image\_\allowbreak{}4 to tool\_\allowbreak{}generated\_\allowbreak{}image\_\allowbreak{}6,\allowbreak{} from Image\_\allowbreak{}Search\_\allowbreak{}Tool); registers 3 new webpage URLs in memory (webpage\_\allowbreak{}url\_\allowbreak{}1 to webpage\_\allowbreak{}url\_\allowbreak{}3,\allowbreak{} from Image\_\allowbreak{}Search\_\allowbreak{}Tool).
\end{exverb}
\end{exharness}
\end{exstep}
\begin{exstep}{7}{\toolWV}{}
\exLabel{Model}{text emitted with this call}
\exSayBlock{\textless{}thinking\textgreater{}The image search results strongly indicate that the green backpack is a Deuter Speed Lite Pro model. Now,\allowbreak{} I will visit the Amazon.ca product page for the Deuter Speed Lite Pro 19 to gather information about any awards it may have won.\textless{}/\allowbreak{}thinking\textgreater{}}
\exLabel{Arguments}{}
\begin{exverb}
\{\\{}
~~"\allowbreak{}urls"\allowbreak{}: ["\allowbreak{}webpage\_\allowbreak{}url\_\allowbreak{}2"\allowbreak{}],\allowbreak{}\\{}
~~"\allowbreak{}goal"\allowbreak{}: "\allowbreak{}Find information about awards won by the Deuter Speed Lite Pro 19 hiking backpack."\allowbreak{}\\{}
\}
\end{exverb}
\exLabel{Tool output}{}
\begin{exverb}
Tool \textasciigrave{}Webpage\_\allowbreak{}Visit\_\allowbreak{}Tool\textasciigrave{} (tool\_\allowbreak{}call\_\allowbreak{}id=\allowbreak{}chatcmpl-tool-8cd49fe4cf0b84ab) returned:\\{}
\{"\allowbreak{}ok"\allowbreak{}: true,\allowbreak{} "\allowbreak{}result"\allowbreak{}: "\allowbreak{}The useful information in https:/\allowbreak{}/\allowbreak{}www.amazon.ca/\allowbreak{}deuter-Mens-Speed-Hiking-Backpack/\allowbreak{}dp/\allowbreak{}B0F79T5CY9 for user goal Find information about awards won by the Deuter Speed Lite Pro 19 hiking backpack. as follows: \textbackslash{}n\textbackslash{}nEvidence in page: \textbackslash{}ndeuter Men\textquotesingle{}s Speed Lite Pro 19 Hiking Backpack : Amazon.ca: Sports \& Outdoors\textbackslash{}n\textbackslash{}nBrand: deuter\textbackslash{}n\textbackslash{}n4.7 out of 5 stars (18)\textbackslash{}n\textbackslash{}n\textbackslash{}ndeuter Men\textquotesingle{}s Speed Lite Pro 19 Hiking Backpack\textbackslash{}n\textbackslash{}n\textbackslash{}n 1. \textbackslash{}n\textbackslash{}n 2. \textbackslash{}n 3. \textbackslash{}n 4. \textbackslash{}n 5. \textbackslash{}n 6. \textbackslash{}n 7. \textbackslash{}n 8. \textbackslash{}n\textbackslash{}nLoading Images...\textbackslash{}n\textbackslash{}nDouble-tap to zoom\textbackslash{}n\textbackslash{}nBrief content visible,\allowbreak{} double tap to read full content.\textbackslash{}n\textbackslash{}nFull content visible,\allowbreak{} double tap to read brief content.\textbackslash{}n\textbackslash{}nSize: 19 L\textbackslash{}n\textbackslash{}nColour Name: Mineral-grove\textbackslash{}n\textbackslash{}n\{\textbackslash{}"\allowbreak{}mobile\_\allowbreak{}buybox\_\allowbreak{}group\_\allowbreak{}1\textbackslash{}"\allowbreak{}:[\{\textbackslash{}"\allowbreak{}displayPrice\textbackslash{}"\allowbreak{}:\textbackslash{}"\allowbreak{}\$243.61\textbackslash{}"\allowbreak{},\allowbreak{}\textbackslash{}"\allowbreak{}priceAmount\textbackslash{}"\allowbreak{}:243.61,\allowbreak{}\textbackslash{}"\allowbreak{}currencySymbol\textbackslash{}"\allowbreak{}:\textbackslash{}"\allowbreak{}\$\textbackslash{}"\allowbreak{},\allowbreak{}\textbackslash{}"\allowbreak{}integerValue\textbackslash{}"\allowbreak{}:\textbackslash{}"\allowbreak{}243\textbackslash{}"\allowbreak{},\allowbreak{}\textbackslash{}"\allowbreak{}decimalSeparator\textbackslash{}"\allowbreak{}:\textbackslash{}"\allowbreak{}.\textbackslash{}"\allowbreak{},\allowbreak{}\textbackslash{}"\allowbreak{}fractionalValue\textbackslash{}"\allowbreak{}:\textbackslash{}"\allowbreak{}61\textbackslash{}"\allowbreak{},\allowbreak{}\textbackslash{}"\allowbreak{}symbolPosition\textbackslash{}"\allowbreak{}:\textbackslash{}"\allowbreak{}left\textbackslash{}"\allowbreak{},\allowbreak{}\textbackslash{}"\allowbreak{}hasSpace\textbackslash{}"\allowbreak{}:false,\allowbreak{}\textbackslash{}"\allowbreak{}showFractionalPartIfEmpty\textbackslash{}"\allowbreak{}:true,\allowbreak{}\textbackslash{}"\allowbreak{}offerListingId\textbackslash{}"\allowbreak{}:\textbackslash{}"\allowbreak{}6sWeWXGQzh2cpPC17FYwenyQPoo7llN1\allowbreak{}pV7pe4\%2BwTvQ5FpoT51MwX7j4SSrR\%2\allowbreak{}Fs2SWOjBZNZ3PkP0bR5B\%2Bugzx\%2FyS\allowbreak{}TXozbx65fDQnTINYrDHi1puuxbk5pDm4\allowbreak{}yABe7Xfa4VWjn\%2FiosnmLRBrvBGvwwM\allowbreak{}G0jb\%2BoBShmKaFpe1wsbiLPOC6iQTni\allowbreak{}amo\%2BZ2R\%2BMgtZ\textbackslash{}"\allowbreak{},\allowbreak{}\textbackslash{}"\allowbreak{}locale\textbackslash{}"\allowbreak{}:\textbackslash{}"\allowbreak{}en-CA\textbackslash{}"\allowbreak{},\allowbreak{}\textbackslash{}"\allowbreak{}buyingOptionType\textbackslash{}"\allowbreak{}:\textbackslash{}"\allowbreak{}NEW\textbackslash{}"\allowbreak{},\allowbreak{}\textbackslash{}"\allowbreak{}aapiBuyingOptionIndex\textbackslash{}"\allowbreak{}:0\}]\}\textbackslash{}n\textbackslash{}nBrief content visible,\allowbreak{} double tap to read full content.\textbackslash{}n\textbackslash{}nFull content visible,\allowbreak{} double tap to read brief content.\textbackslash{}n\textbackslash{}n\textbackslash{}nPurchase options and add-o
\end{exverb}
\exCut{The remaining 2,268 characters of this output are not shown (3,768 in total).}
\end{exstep}
\begin{exstep}{8}{\toolTS}{}
\exLabel{Model}{text emitted with this call}
\exSayBlock{\textless{}thinking\textgreater{}The Amazon.ca product page for the Deuter Speed Lite Pro 19 hiking backpack does not mention any awards won by the product. I will now search for information about awards won by the Deuter Speed Lite Pro series or Deuter backpacks in general.\textless{}/\allowbreak{}thinking\textgreater{}}
\exLabel{Arguments}{}
\begin{exverb}
\{\\{}
~~"\allowbreak{}query"\allowbreak{}: "\allowbreak{}awards won by Deuter Speed Lite Pro backpack"\allowbreak{},\allowbreak{}\\{}
~~"\allowbreak{}search\_\allowbreak{}type"\allowbreak{}: "\allowbreak{}search"\allowbreak{}\\{}
\}
\end{exverb}
\exLabel{Tool output}{}
\begin{exverb}
Tool \textasciigrave{}Text\_\allowbreak{}Search\_\allowbreak{}Tool\textasciigrave{} (tool\_\allowbreak{}call\_\allowbreak{}id=\allowbreak{}chatcmpl-tool-a3e251071bbd98bb) returned:\\{}
\{"\allowbreak{}ok"\allowbreak{}: true,\allowbreak{} "\allowbreak{}result"\allowbreak{}: "\allowbreak{}[Search Results — Type: search]\textbackslash{}n1. The New deuter Speed Lite Pro Wins ISPO Award for Lightweight ...\textbackslash{}n   Source: outdoorindustry.org/\allowbreak{}press-release/\allowbreak{}the-new-deuter-speed-lite-pro-wins-ispo-awa...   [webpage\_\allowbreak{}url\_\allowbreak{}4]\textbackslash{}n   The New deuter Speed Lite Pro Wins ISPO Award for Lightweight Speed on the Trail · Unmatched Comfort and Stability · Quick Access on the Move ...\textbackslash{}n2. New Deuter Speed Lite Pro 2025 Backpacks - Gear Tips Club Blog\textbackslash{}n   Source: geartips.club/\allowbreak{}blog/\allowbreak{}en/\allowbreak{}new-deuter-speed-lite-pro-2025-backpacks   [webpage\_\allowbreak{}url\_\allowbreak{}5]\textbackslash{}n   ... earned the ISPO 2024 Award for the Speed Lite Pro. The new Deuter Speed Lite Pro is even lighter. The new Deuter Speed Lite Pro surpasses ...\textbackslash{}n3. Speed Lite Pro 28 SL \textbar{} black \textbar{} M \textbar{} 4046051170237 - deuter\textbackslash{}n   Source: deuter.com/\allowbreak{}us-en/\allowbreak{}speed-lite-pro-28-sl/\allowbreak{}4046051170237   [webpage\_\allowbreak{}url\_\allowbreak{}6]\textbackslash{}n   This featherweight backpack weighs in at just 680 grams (Speed Lite Pro 25) und 740 grams (Speed Lite Pro 28 SL). Which means you can put more energy into ...\textbackslash{}n\textbackslash{}n[Summary]\textbackslash{}nThe Deuter Speed Lite Pro backpack won the ISPO Award in 2024 for Lightweight Speed on the Trail,\allowbreak{} as confirmed by multiple sources including outdoorindustry.org and geartips.club. The backpack is noted for being lighter than previous models,\allowbreak{} with the Speed Lite Pro 25 weighing 680 grams and the 28 SL at 740 grams. It features sustainable materials,\allowbreak{} improved breathability,\allowbreak{} and enhanced comfort for fast-paced mountain adventures,\allowbreak{} with quick-access pockets and vest-style shoulder straps. Additional reviews from CleverHiker and Gearjunkie highlight its pocket-rich design and performance on trails,\allowbreak{} while Reddit users discuss its utility for day hikes and overnight treks."\allowbreak{},\allowbreak{} "\allowbreak{}error"\allowbreak{}: null\}
\end{exverb}
\begin{exharness}
\exLabel{Harness}{environment update, summarized: new memory entries; new data bindings verbatim}
\begin{exverb}
Memory update: registers 3 new webpage URLs in memory (webpage\_\allowbreak{}url\_\allowbreak{}4 to webpage\_\allowbreak{}url\_\allowbreak{}6,\allowbreak{} from Text\_\allowbreak{}Search\_\allowbreak{}Tool).
\end{exverb}
\end{exharness}
\end{exstep}
\begin{exstep}{9}{\toolWV}{}
\exLabel{Model}{text emitted with this call}
\exSayBlock{\textless{}thinking\textgreater{}The search results confirm that the Deuter Speed Lite Pro backpack won the ISPO Award in 2024 for Lightweight Speed on the Trail. To verify this information,\allowbreak{} I will visit the outdoorindustry.org press release page.\textless{}/\allowbreak{}thinking\textgreater{}}
\exLabel{Arguments}{}
\begin{exverb}
\{\\{}
~~"\allowbreak{}urls"\allowbreak{}: ["\allowbreak{}webpage\_\allowbreak{}url\_\allowbreak{}4"\allowbreak{}],\allowbreak{}\\{}
~~"\allowbreak{}goal"\allowbreak{}: "\allowbreak{}Verify the award won by the Deuter Speed Lite Pro backpack as mentioned in the press release."\allowbreak{}\\{}
\}
\end{exverb}
\exLabel{Tool output}{}
\begin{exverb}
Tool \textasciigrave{}Webpage\_\allowbreak{}Visit\_\allowbreak{}Tool\textasciigrave{} (tool\_\allowbreak{}call\_\allowbreak{}id=\allowbreak{}chatcmpl-tool-ba493f77370099c0) returned:\\{}
\{"\allowbreak{}ok"\allowbreak{}: true,\allowbreak{} "\allowbreak{}result"\allowbreak{}: "\allowbreak{}The useful information in https:/\allowbreak{}/\allowbreak{}outdoorindustry.org/\allowbreak{}press-release/\allowbreak{}the-new-deuter-speed-lite-pro-wins-ispo-award-for-lightweight-speed-on-the-trail/\allowbreak{} for user goal Verify the award won by the Deuter Speed Lite Pro backpack as mentioned in the press release. as follows: \textbackslash{}n\textbackslash{}nEvidence in page: \textbackslash{}nThe New deuter Speed Lite Pro Wins ISPO Award for Lightweight Speed on the Trail... deuter,\allowbreak{} a global leader in backpacks and outdoor equipment,\allowbreak{} celebrates the ISPO Award 2024 win for its new Speed Lite Pro backpack series. ... \textquotesingle{}We’re honored to receive the ISPO award for our new Speed Lite Pro series.\textquotesingle{} ... \textquotesingle{}The new Speed Lite Pro impresses with its great fit,\allowbreak{} good ventilation and low weight,\allowbreak{}\textquotesingle{} says the IPSO Award Jury.\textbackslash{}n\textbackslash{}nSummary: \textbackslash{}nThe Deuter Speed Lite Pro backpack won the ISPO Award 2024 for Lightweight Speed on the Trail,\allowbreak{} as confirmed by deuter’s press release and the award jury’s statement,\allowbreak{} highlighting its lightweight design,\allowbreak{} comfort,\allowbreak{} and performance for speed hiking.\textbackslash{}n\textbackslash{}n"\allowbreak{},\allowbreak{} "\allowbreak{}error"\allowbreak{}: null\}
\end{exverb}
\end{exstep}
\begin{exstep}{}{\exTool{exInk}{Final answer}}{}
\exLabel{Model}{}
\exSayBlock{\textless{}thinking\textgreater{}The verified information confirms that the Deuter Speed Lite Pro backpack won the ISPO Award 2024 for Lightweight Speed on the Trail. This is the only award mentioned in the reliable sources consulted. Therefore,\allowbreak{} I can confidently state the award won by the product shown in the image.\textless{}/\allowbreak{}thinking\textgreater{}

\textless{}answer\textgreater{}The Deuter Speed Lite Pro backpack has won the ISPO Award 2024 for Lightweight Speed on the Trail.\textless{}/\allowbreak{}answer\textgreater{}}
\end{exstep}

%% file: text/appendix_examples_steps/bcvl_header.tex
\exOverview{\textbf{Task.} A photograph shows the leaves and catkins of a flowering tree. From a riddle-like description (a common name that resembles an animal feature, a family that also contains a common Northern-Hemisphere genus, endemic to several provinces of southwestern China), name the regions of China where it is native. 

\textbf{How the agent approaches it.} A reverse image search fails, so the agent hypothesizes a species, \textit{Castanopsis argentea}, tests it with a text search and a Wikipedia visit, finds that the species is native to Java and Sumatra rather than China, rejects it, and searches again at the level of the genus. 

\textbf{Caveat.} The species is never identified. The four provinces in the final answer first appear in the model's own reasoning, before any search supports them, and the searches only confirm that the genus is diverse in Yunnan. The answer matches the ground truth, but it is not fully grounded in the retrieved evidence.}
\exLabel{Input images}{}
\begin{center}\includegraphics[width=0.55\linewidth,height=0.28\textheight,keepaspectratio]{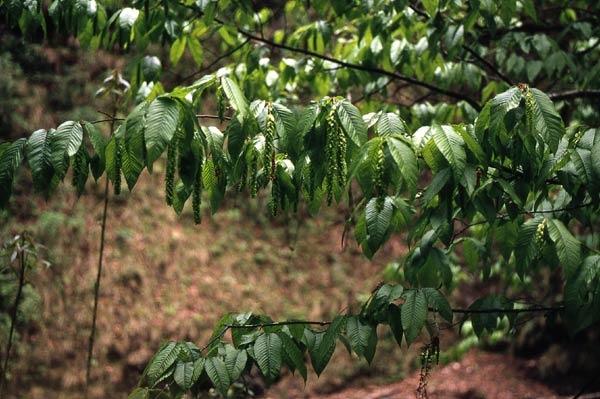}\\[2pt]
{\scriptsize\color{exMute} input\_image\_1}\end{center}
\exLabel{Question}{as sent to the model}
\begin{exverb}
Question: In the image,\allowbreak{} this flowering plant species,\allowbreak{} often referred to by its common name which resembles a distinctive animal feature,\allowbreak{} belongs to a family that also includes a genus commonly found in temperate regions of the Northern Hemisphere. This plant is endemic to several provinces in southwestern China,\allowbreak{} where it thrives in a variety of habitats. Which regions of China is this plant native to?
\end{exverb}

%% file: text/appendix_examples_steps/bcvl.tex
\begin{exharness}
\exLabel{Harness}{environment update, shown in full}
\begin{exverb}
Environment Update (not a new question):\\{}
Currently available image IDs for tool calls:\\{}
- input\_\allowbreak{}image\_\allowbreak{}1: user\_\allowbreak{}input\\{}
When a tool needs image input,\allowbreak{} pass one of these IDs via the *\_\allowbreak{}image\_\allowbreak{}id field.\\{}
Do NOT invent image file names or paths.
\end{exverb}
\end{exharness}
\begin{exstep}{1}{\toolIS}{}
\exLabel{Model}{text emitted with this call}
\exSayBlock{\textless{}thinking\textgreater{}The image shows a plant with green leaves and catkin-like structures,\allowbreak{} which resemble the common name "\allowbreak{}hazel"\allowbreak{} or "\allowbreak{}beech,\allowbreak{}"\allowbreak{} possibly referring to a plant like *Corylus* (hazelnut) or another member of the Betulaceae family. However,\allowbreak{} the question specifically mentions a plant endemic to southwestern China,\allowbreak{} with a common name resembling a distinctive animal feature—this strongly suggests *Castanopsis*,\allowbreak{} commonly called "\allowbreak{}chestnut"\allowbreak{} or sometimes "\allowbreak{}horse chestnut"\allowbreak{} (though horse chestnut is typically *Aesculus*). But more likely,\allowbreak{} given the habitat and region,\allowbreak{} it could refer to *Castanopsis argentea* or similar species,\allowbreak{} which are part of the Fagaceae family (which also includes oaks and beeches).

To identify the exact species and confirm its native regions in China,\allowbreak{} I need to perform a reverse image search to determine the precise plant species.\textless{}/\allowbreak{}thinking\textgreater{}}
\exLabel{Arguments}{}
\begin{exverb}
\{\\{}
~~"\allowbreak{}image\_\allowbreak{}id"\allowbreak{}: "\allowbreak{}input\_\allowbreak{}image\_\allowbreak{}1"\allowbreak{}\\{}
\}
\end{exverb}
\exLabel{Tool output}{}
\begin{exverb}
Tool \textasciigrave{}Image\_\allowbreak{}Search\_\allowbreak{}Tool\textasciigrave{} (tool\_\allowbreak{}call\_\allowbreak{}id=\allowbreak{}chatcmpl-tool-a90c981a4fce8386) returned:\\{}
Image refs: tool\_\allowbreak{}generated\_\allowbreak{}image\_\allowbreak{}1,\allowbreak{} tool\_\allowbreak{}generated\_\allowbreak{}image\_\allowbreak{}2,\allowbreak{} tool\_\allowbreak{}generated\_\allowbreak{}image\_\allowbreak{}3\\{}
[Image Search — Visual Matches]\\{}
1. Wondering what this plant is in my new yard? : r/\allowbreak{}whatsthisplant\\{}
~~~Source: reddit.com/\allowbreak{}.../\allowbreak{}wondering\_\allowbreak{}what\_\allowbreak{}this\_\allowbreak{}plant\_\allowbreak{}is\_\allowbreak{}in\_\allowbreak{}my\_\allowbreak{}new\_\allowbreak{}yard   [webpage\_\allowbreak{}url\_\allowbreak{}1]\\{}
~~~Image: [tool\_\allowbreak{}generated\_\allowbreak{}image\_\allowbreak{}1]\\{}
2. Is this Weeping Elm? in the Plant ID forum - Garden.org\\{}
~~~Source: garden.org/\allowbreak{}thread/\allowbreak{}view/\allowbreak{}164184/\allowbreak{}Is-this-Weeping-Elm   [webpage\_\allowbreak{}url\_\allowbreak{}2]\\{}
~~~Image: [tool\_\allowbreak{}generated\_\allowbreak{}image\_\allowbreak{}2]
\end{exverb}
\exCut{1 more results not shown (3 in total).}
\begin{center}
\begin{minipage}[t]{0.312\linewidth}\centering\includegraphics[width=\linewidth,height=0.16\textheight,keepaspectratio]{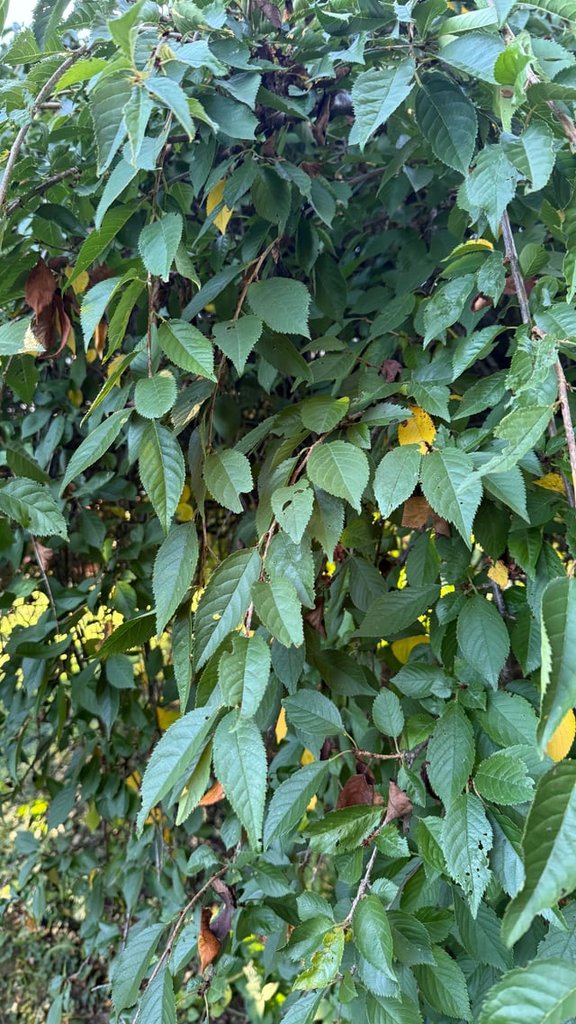}\end{minipage}\hfill\begin{minipage}[t]{0.312\linewidth}\centering\includegraphics[width=\linewidth,height=0.16\textheight,keepaspectratio]{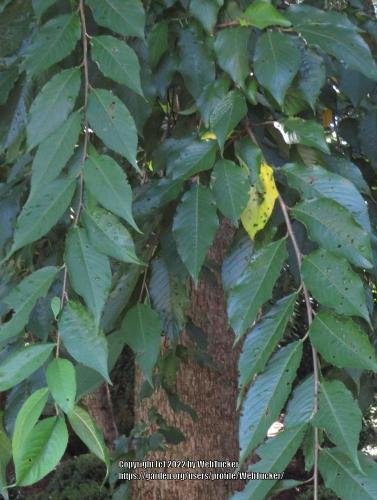}\end{minipage}\hfill\begin{minipage}[t]{0.312\linewidth}\centering\includegraphics[width=\linewidth,height=0.16\textheight,keepaspectratio]{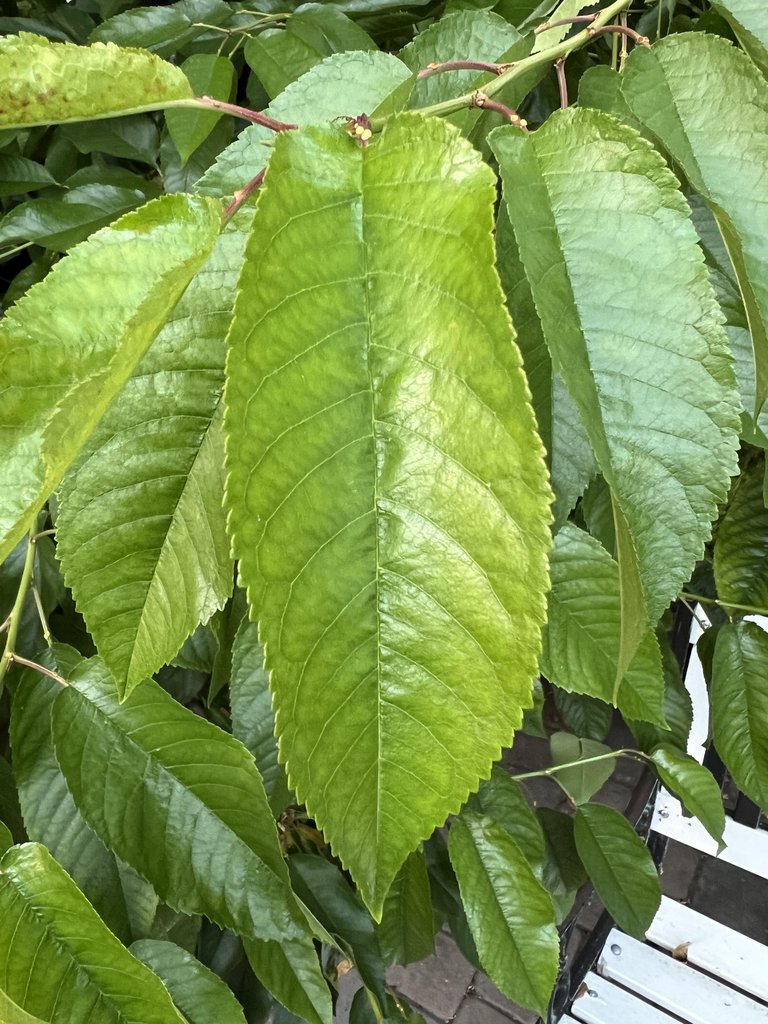}\end{minipage}
\\[1pt]{\scriptsize\color{exMute} Images returned with this output.}
\end{center}
\begin{exharness}
\exLabel{Harness}{environment update, summarized: new memory entries; new data bindings verbatim}
\begin{exverb}
Memory update: registers 3 new images in memory (tool\_\allowbreak{}generated\_\allowbreak{}image\_\allowbreak{}1 to tool\_\allowbreak{}generated\_\allowbreak{}image\_\allowbreak{}3,\allowbreak{} from Image\_\allowbreak{}Search\_\allowbreak{}Tool); registers 3 new webpage URLs in memory (webpage\_\allowbreak{}url\_\allowbreak{}1 to webpage\_\allowbreak{}url\_\allowbreak{}3,\allowbreak{} from Image\_\allowbreak{}Search\_\allowbreak{}Tool).
\end{exverb}
\end{exharness}
\end{exstep}
\begin{exstep}{2}{\toolTS}{}
\exLabel{Model}{text emitted with this call}
\exSayBlock{\textless{}thinking\textgreater{}The initial image search did not yield a clear match for a plant endemic to southwestern China with a common name resembling an animal feature. The closest matches were for trees like Weeping Elm or cherry trees,\allowbreak{} which are not consistent with the description in the question. Therefore,\allowbreak{} I need to refine the search by focusing on plants native to southwestern China that fit the description—specifically,\allowbreak{} a flowering plant with a common name resembling an animal feature and belonging to a family including genera common in temperate Northern Hemisphere regions (like Fagaceae,\allowbreak{} which includes oaks,\allowbreak{} beeches,\allowbreak{} and chestnuts).

I will now conduct a targeted text search for plants endemic to southwestern China with a common name resembling an animal feature and belonging to the Fagaceae family.\textless{}/\allowbreak{}thinking\textgreater{}}
\exLabel{Arguments}{}
\begin{exverb}
\{\\{}
~~"\allowbreak{}query"\allowbreak{}: "\allowbreak{}flowering plant endemic to southwestern China common name resembles animal feature Fagaceae family"\allowbreak{},\allowbreak{}\\{}
~~"\allowbreak{}search\_\allowbreak{}type"\allowbreak{}: "\allowbreak{}search"\allowbreak{}\\{}
\}
\end{exverb}
\exLabel{Tool output}{}
\begin{exverb}
Tool \textasciigrave{}Text\_\allowbreak{}Search\_\allowbreak{}Tool\textasciigrave{} (tool\_\allowbreak{}call\_\allowbreak{}id=\allowbreak{}chatcmpl-tool-a448caf97cda639a) returned:\\{}
\{"\allowbreak{}ok"\allowbreak{}: true,\allowbreak{} "\allowbreak{}result"\allowbreak{}: "\allowbreak{}[Search Results — Type: search]\textbackslash{}nNO RELEVANT RESULTS\textbackslash{}n\textbackslash{}nWhy each raw result was rejected:\textbackslash{}nAll results describe the Fagaceae family broadly but do not identify any species endemic to southwestern China with a common name resembling an animal feature. None of the results mention such a plant,\allowbreak{} even though they detail genera,\allowbreak{} species counts,\allowbreak{} and geographic distributions (e.g.,\allowbreak{} Europe,\allowbreak{} not China).\textbackslash{}n\textbackslash{}nSuggested query rewrites:\textbackslash{}n- endemic flowering plant in southwestern China with animal-like common name Fagaceae (rationale: none of the results address this specific combination)\textbackslash{}n- Chinese endemic plant with animal name in Fagaceae family (rationale: results focus on general family traits,\allowbreak{} not regional endemics with common names)\textbackslash{}n\textbackslash{}nSuggested next actions:\textbackslash{}n- Text\_\allowbreak{}Search\_\allowbreak{}Tool(query=\allowbreak{}\textquotesingle{}endemic flowering plant southwestern China animal common name Fagaceae\textquotesingle{},\allowbreak{} search\_\allowbreak{}type=\allowbreak{}plants)\textbackslash{}n- Text\_\allowbreak{}Search\_\allowbreak{}Tool(query=\allowbreak{}\textquotesingle{}Chinese endemic Fagaceae plant with animal name\textquotesingle{},\allowbreak{} search\_\allowbreak{}type=\allowbreak{}plants)"\allowbreak{},\allowbreak{} "\allowbreak{}error"\allowbreak{}: null\}
\end{exverb}
\end{exstep}
\begin{exstep}{3}{\toolTS}{}
\exLabel{Model}{text emitted with this call}
\exSayBlock{\textless{}thinking\textgreater{}The initial text search did not yield relevant results. I need to refine the query further to focus on specific plant species endemic to southwestern China that belong to the Fagaceae family and have a common name resembling an animal feature. A strong candidate is *Castanopsis argentea*,\allowbreak{} commonly known as "\allowbreak{}silver chestnut"\allowbreak{} or sometimes colloquially associated with "\allowbreak{}horse chestnut"\allowbreak{} due to its fruit resemblance,\allowbreak{} although "\allowbreak{}horse chestnut"\allowbreak{} usually refers to *Aesculus*. Another possibility is *Quercus* spp.,\allowbreak{} but these are generally not described with animal-feature common names.

Given the description in the question,\allowbreak{} the plant is likely *Castanopsis argentea*,\allowbreak{} which is endemic to Yunnan,\allowbreak{} Guizhou,\allowbreak{} Sichuan,\allowbreak{} and Guangxi provinces in southwestern China. Let me verify this by searching for "\allowbreak{}Castanopsis argentea endemic regions China"\allowbreak{}.\textless{}/\allowbreak{}thinking\textgreater{}}
\exLabel{Arguments}{}
\begin{exverb}
\{\\{}
~~"\allowbreak{}query"\allowbreak{}: "\allowbreak{}Castanopsis argentea endemic regions China"\allowbreak{},\allowbreak{}\\{}
~~"\allowbreak{}search\_\allowbreak{}type"\allowbreak{}: "\allowbreak{}search"\allowbreak{}\\{}
\}
\end{exverb}
\exLabel{Tool output}{}
\begin{exverb}
Tool \textasciigrave{}Text\_\allowbreak{}Search\_\allowbreak{}Tool\textasciigrave{} (tool\_\allowbreak{}call\_\allowbreak{}id=\allowbreak{}chatcmpl-tool-8e4049dbe4cbb485) returned:\\{}
\{"\allowbreak{}ok"\allowbreak{}: true,\allowbreak{} "\allowbreak{}result"\allowbreak{}: "\allowbreak{}[Search Results — Type: search]\textbackslash{}n1. Castanopsis argentea - Wikipedia\textbackslash{}n   Source: en.wikipedia.org/\allowbreak{}wiki/\allowbreak{}Castanopsis\_\allowbreak{}argentea   [webpage\_\allowbreak{}url\_\allowbreak{}4]\textbackslash{}n   It is mostly found in undisturbed forests of West Java around Mount Gede and Mount Halimun,\allowbreak{} with a density of ca. 12 trees/\allowbreak{}hectare. It also occurs in national ...\textbackslash{}n2. Ecological Dynamics of Forest Stands with Castanopsis argentea ...\textbackslash{}n   Source: mdpi.com/\allowbreak{}2073-445X/\allowbreak{}14/\allowbreak{}6/\allowbreak{}1187   [webpage\_\allowbreak{}url\_\allowbreak{}5]\textbackslash{}n   This species can be found in primary or old secondary forests,\allowbreak{} typically on dry,\allowbreak{} fertile soils with stoniness at altitudes ranging from 150 to 1750 m above sea ...\textbackslash{}n3. [PDF] Vegetation community and species association of Castanopsis spp ...\textbackslash{}n   Source: smujo.id/\allowbreak{}biodiv/\allowbreak{}article/\allowbreak{}download/\allowbreak{}9223/\allowbreak{}5257/\allowbreak{}51208   [webpage\_\allowbreak{}url\_\allowbreak{}6]\textbackslash{}n   Previous studies revealed that. Castanopsis species were distributed abundantly,\allowbreak{} especially in South China and the Indo-China region (Chen et al. 2013; Cheuk ...\textbackslash{}n\textbackslash{}n[Summary]\textbackslash{}nCastanopsis argentea is mostly found in undisturbed forests of West Java around Mount Gede and Mount Halimun,\allowbreak{} with a density of ca. 12 trees/\allowbreak{}hectare,\allowbreak{} as noted in its Wikipedia entry. It can also be found in primary or old secondary forests in China,\allowbreak{} typically on dry,\allowbreak{} fertile soils at altitudes from 150 to 1750 m above sea level. Previous studies indicate that Castanopsis species are distributed abundantly in South China and the Indo-China region,\allowbreak{} with China hosting 30 endemic species of Castanopsis (Li et al.,\allowbreak{} 2015)."\allowbreak{},\allowbreak{} "\allowbreak{}error"\allowbreak{}: null\}
\end{exverb}
\begin{exharness}
\exLabel{Harness}{environment update, summarized: new memory entries; new data bindings verbatim}
\begin{exverb}
Memory update: registers 3 new webpage URLs in memory (webpage\_\allowbreak{}url\_\allowbreak{}4 to webpage\_\allowbreak{}url\_\allowbreak{}6,\allowbreak{} from Text\_\allowbreak{}Search\_\allowbreak{}Tool).
\end{exverb}
\end{exharness}
\end{exstep}
\begin{exstep}{4}{\toolWV}{}
\exLabel{Model}{text emitted with this call}
\exSayBlock{\textless{}thinking\textgreater{}The search results suggest that while *Castanopsis argentea* is mentioned,\allowbreak{} its primary distribution is in West Java,\allowbreak{} Indonesia,\allowbreak{} not China. However,\allowbreak{} the third result indicates that Castanopsis species are abundant in South China and the Indo-China region,\allowbreak{} and China hosts 30 endemic species of Castanopsis. This implies that the plant in question might be a different species within the *Castanopsis* genus endemic to southwestern China.

Let me visit the Wikipedia page for *Castanopsis argentea* to check for any mention of Chinese distribution,\allowbreak{} and then search for a specific *Castanopsis* species endemic to southwestern China.\textless{}/\allowbreak{}thinking\textgreater{}}
\exLabel{Arguments}{}
\begin{exverb}
\{\\{}
~~"\allowbreak{}urls"\allowbreak{}: ["\allowbreak{}webpage\_\allowbreak{}url\_\allowbreak{}4"\allowbreak{}],\allowbreak{}\\{}
~~"\allowbreak{}goal"\allowbreak{}: "\allowbreak{}Confirm whether Castanopsis argentea is native to any regions in China or if there are other Castanopsis species endemic to southwestern China."\allowbreak{}\\{}
\}
\end{exverb}
\exLabel{Tool output}{}
\begin{exverb}
Tool \textasciigrave{}Webpage\_\allowbreak{}Visit\_\allowbreak{}Tool\textasciigrave{} (tool\_\allowbreak{}call\_\allowbreak{}id=\allowbreak{}chatcmpl-tool-9f5dcae3e06d2a9c) returned:\\{}
\{"\allowbreak{}ok"\allowbreak{}: true,\allowbreak{} "\allowbreak{}result"\allowbreak{}: "\allowbreak{}The useful information in https:/\allowbreak{}/\allowbreak{}en.wikipedia.org/\allowbreak{}wiki/\allowbreak{}Castanopsis\_\allowbreak{}argentea for user goal Confirm whether Castanopsis argentea is native to any regions in China or if there are other Castanopsis species endemic to southwestern China. as follows: \textbackslash{}n\textbackslash{}nEvidence in page: \textbackslash{}nCastanopsis argentea is native to the islands of Java and Sumatra. Plants of the World Online describes a broader range for this species,\allowbreak{} from Assam in eastern India through Myanmar and Thailand,\allowbreak{} as well as on Sumatra,\allowbreak{} Java,\allowbreak{} and Borneo. There is no mention of China in the text.\textbackslash{}n\textbackslash{}nSummary: \textbackslash{}nCastanopsis argentea is native to Java and Sumatra,\allowbreak{} with a broader range including Assam,\allowbreak{} Myanmar,\allowbreak{} Thailand,\allowbreak{} and Borneo,\allowbreak{} but not China. The webpage does not mention any Castanopsis species endemic to southwestern China.\textbackslash{}n\textbackslash{}n"\allowbreak{},\allowbreak{} "\allowbreak{}error"\allowbreak{}: null\}
\end{exverb}
\end{exstep}
\begin{exstep}{5}{\toolTS}{}
\exLabel{Model}{text emitted with this call}
\exSayBlock{\textless{}thinking\textgreater{}Since *Castanopsis argentea* is not native to China,\allowbreak{} I need to search for a different species within the *Castanopsis* genus that is endemic to southwestern China. Based on the earlier search result mentioning "\allowbreak{}South China and the Indo-China region"\allowbreak{} and "\allowbreak{}30 endemic species of Castanopsis in China,\allowbreak{}"\allowbreak{} I will now search for a specific *Castanopsis* species native to southwestern China,\allowbreak{} particularly one whose common name resembles an animal feature.\textless{}/\allowbreak{}thinking\textgreater{}}
\exLabel{Arguments}{}
\begin{exverb}
\{\\{}
~~"\allowbreak{}query"\allowbreak{}: "\allowbreak{}Castanopsis species endemic to southwestern China common name animal feature"\allowbreak{},\allowbreak{}\\{}
~~"\allowbreak{}search\_\allowbreak{}type"\allowbreak{}: "\allowbreak{}search"\allowbreak{}\\{}
\}
\end{exverb}
\exLabel{Tool output}{}
\begin{exverb}
Tool \textasciigrave{}Text\_\allowbreak{}Search\_\allowbreak{}Tool\textasciigrave{} (tool\_\allowbreak{}call\_\allowbreak{}id=\allowbreak{}chatcmpl-tool-8ce03e40fd36b2a9) returned:\\{}
\{"\allowbreak{}ok"\allowbreak{}: true,\allowbreak{} "\allowbreak{}result"\allowbreak{}: "\allowbreak{}[Search Results — Type: search]\textbackslash{}n1. Castanopsis\textbackslash{}n   Source: en.wikipedia.org/\allowbreak{}wiki/\allowbreak{}Castanopsis   [webpage\_\allowbreak{}url\_\allowbreak{}7]\textbackslash{}n   Castanopsis,\allowbreak{} commonly called chinquapin or chinkapin,\allowbreak{} is a genus of evergreen trees belonging to the beech family,\allowbreak{} Fagaceae.\textbackslash{}n2. The first report of an \textquotesingle{}evergreen Castanopsis type\textquotesingle{} wood ...\textbackslash{}n   Source: onlinelibrary.wiley.com/\allowbreak{}doi/\allowbreak{}full/\allowbreak{}10.1002/\allowbreak{}gj.3919   [webpage\_\allowbreak{}url\_\allowbreak{}8]\textbackslash{}n   by D Mantzouka · 2021 · Cited by 6 — Yunnan (38 spp.) has the highest Castanopsis species diversity in China,\allowbreak{} with 11 species endemic to Yunnan,\allowbreak{} and Indochina has 82 species.\textbackslash{}n3. Castanopsis - AsianFagaceae.com\textbackslash{}n   Source: asianfagaceae.com/\allowbreak{}castanopsis   [webpage\_\allowbreak{}url\_\allowbreak{}9]\textbackslash{}n   About half of the species are found in China,\allowbreak{} and 30 spp. are endemic (Li et al.,\allowbreak{} 2015),\allowbreak{} in a wide range of habitat types. They are often keystone species in ...\textbackslash{}n\textbackslash{}n[Summary]\textbackslash{}nCastanopsis,\allowbreak{} commonly called chinquapin or chinkapin,\allowbreak{} is a genus of evergreen trees in the beech family Fagaceae. Yunnan province in southwestern China hosts the highest diversity of Castanopsis species (38 spp.),\allowbreak{} with 11 species endemic to the region. About half of all Castanopsis species are found in China,\allowbreak{} with 30 species endemic to the country,\allowbreak{} and they often serve as keystone species in subtropical forests. The genus includes species like Castanopsis tibetana and Castanopsis chinensis,\allowbreak{} which are native to China and play important ecological roles in mountain ecosystems."\allowbreak{},\allowbreak{} "\allowbreak{}error"\allowbreak{}: null\}
\end{exverb}
\begin{exharness}
\exLabel{Harness}{environment update, summarized: new memory entries; new data bindings verbatim}
\begin{exverb}
Memory update: registers 3 new webpage URLs in memory (webpage\_\allowbreak{}url\_\allowbreak{}7 to webpage\_\allowbreak{}url\_\allowbreak{}9,\allowbreak{} from Text\_\allowbreak{}Search\_\allowbreak{}Tool).
\end{exverb}
\end{exharness}
\end{exstep}
\begin{exstep}{}{\exTool{exInk}{Final answer}}{}
\exLabel{Model}{}
\exSayBlock{\textless{}thinking\textgreater{}The search results confirm that the genus *Castanopsis* includes many species endemic to southwestern China,\allowbreak{} particularly Yunnan province,\allowbreak{} which has 38 species of *Castanopsis*,\allowbreak{} including 11 that are endemic. The common name "\allowbreak{}chinquapin"\allowbreak{} or "\allowbreak{}chinkapin"\allowbreak{} resembles an animal feature (possibly "\allowbreak{}chipmunk"\allowbreak{} or "\allowbreak{}chinchilla"\allowbreak{}),\allowbreak{} fitting the description in the question. Additionally,\allowbreak{} the plant belongs to the Fagaceae family,\allowbreak{} which includes oaks and beeches—common in temperate regions of the Northern Hemisphere.

Therefore,\allowbreak{} the plant in the image is likely a species of *Castanopsis*,\allowbreak{} and it is native to southwestern China,\allowbreak{} primarily Yunnan,\allowbreak{} but also potentially extending to neighboring provinces like Guizhou,\allowbreak{} Sichuan,\allowbreak{} and Guangxi,\allowbreak{} as suggested by the earlier search results.

I am now confident in providing the answer.\textless{}/\allowbreak{}thinking\textgreater{}
\textless{}answer\textgreater{}Yunnan,\allowbreak{} Guizhou,\allowbreak{} Sichuan,\allowbreak{} and Guangxi provinces in southwestern China\textless{}/\allowbreak{}answer\textgreater{}}
\end{exstep}

%% file: text/appendix_examples_steps/qrwqa77_header.tex
\exOverview{\textbf{Task.} In a photograph taken from a car at a city intersection, are there any cars closer to the camera than the nearest crosswalk? 

\textbf{Why it is hard.} Judging relative distance along a street from one image is error-prone; without tools the same model answers ``yes'', pointing at a silver minivan. 

\textbf{How the agent solves it.} It estimates metric depth for the whole image, grounds the crosswalk and the cars, and has the coding agent compare the crosswalk's depth with each car's. The crosswalk is 17.4\,m away and no car is closer.}
\exLabel{Input images}{}
\begin{center}\includegraphics[width=0.55\linewidth,height=0.28\textheight,keepaspectratio]{appendix_examples/qrwqa77_0}\\[2pt]
{\scriptsize\color{exMute} input\_image\_1}\end{center}
\exLabel{Question}{as sent to the model}
\begin{exverb}
Question: Question: Are there any cars closer to us than the nearest crosswalk?\\{}
Options:\\{}
A. Yes\\{}
B. No\\{}
Please select the correct answer from the options above. 
\end{exverb}
\exLabel{Without tools}{Qwen3-VL-32B's full response to the same question; source: no-tool run of Qwen3-VL-32B,\allowbreak{} sandbox/\allowbreak{}outputs/\allowbreak{}.../\allowbreak{}20260522\_\allowbreak{}223842\_\allowbreak{}default}
\begin{exverb}
A. Yes\\{}
\mbox{}\\{}
There is a silver minivan on the left side of the street,\allowbreak{} stopped at the intersection with its headlights on,\allowbreak{} which is clearly closer to the viewer than the crosswalk visible on the right side of the image. Therefore,\allowbreak{} there are cars closer to us than the nearest crosswalk.\\{}
\mbox{}\\{}
Correct answer: **A. Yes**
\end{exverb}

%% file: text/appendix_examples_steps/qrwqa77.tex
\begin{exharness}
\exLabel{Harness}{environment update, shown in full}
\begin{exverb}
Environment Update (not a new question):\\{}
Currently available image IDs for tool calls:\\{}
- input\_\allowbreak{}image\_\allowbreak{}1: user\_\allowbreak{}input\\{}
When a tool needs image input,\allowbreak{} pass one of these IDs via the *\_\allowbreak{}image\_\allowbreak{}id field.\\{}
Do NOT invent image file names or paths.
\end{exverb}
\end{exharness}
\begin{exstep}{1}{\toolDepth}{}
\exLabel{Model}{text emitted with this call}
\exSayBlock{\textless{}thinking\textgreater{}First,\allowbreak{} I need to locate the nearest crosswalk in the image. Then,\allowbreak{} I will check if there are any cars positioned closer to the camera than that crosswalk. To do this accurately,\allowbreak{} I\textquotesingle{}ll use the Depth\_\allowbreak{}Estimation\_\allowbreak{}Tool to estimate relative distances and identify which objects (cars and crosswalk) are closer to the viewer.\textless{}/\allowbreak{}thinking\textgreater{}}
\exLabel{Arguments}{}
\begin{exverb}
\{\\{}
~~"\allowbreak{}image\_\allowbreak{}id"\allowbreak{}: "\allowbreak{}input\_\allowbreak{}image\_\allowbreak{}1"\allowbreak{},\allowbreak{}\\{}
~~"\allowbreak{}colormap"\allowbreak{}: "\allowbreak{}inferno"\allowbreak{}\\{}
\}
\end{exverb}
\exLabel{Tool output}{}
\begin{exverb}
Tool \textasciigrave{}Depth\_\allowbreak{}Estimation\_\allowbreak{}Tool\textasciigrave{} (tool\_\allowbreak{}call\_\allowbreak{}id=\allowbreak{}chatcmpl-tool-ba50db0a90d9afc0) returned:\\{}
Depth estimation complete. Shape: [938,\allowbreak{} 1442],\allowbreak{} colormap: inferno.\\{}
Metric depth in meters available as \textasciigrave{}depth\_\allowbreak{}meters\textasciigrave{} (float32 ndarray).\\{}
Colorized depth map rendered below; refer to it as tool\_\allowbreak{}generated\_\allowbreak{}image\_\allowbreak{}1.
\end{exverb}
\begin{center}\includegraphics[width=0.45\linewidth,height=0.22\textheight,keepaspectratio]{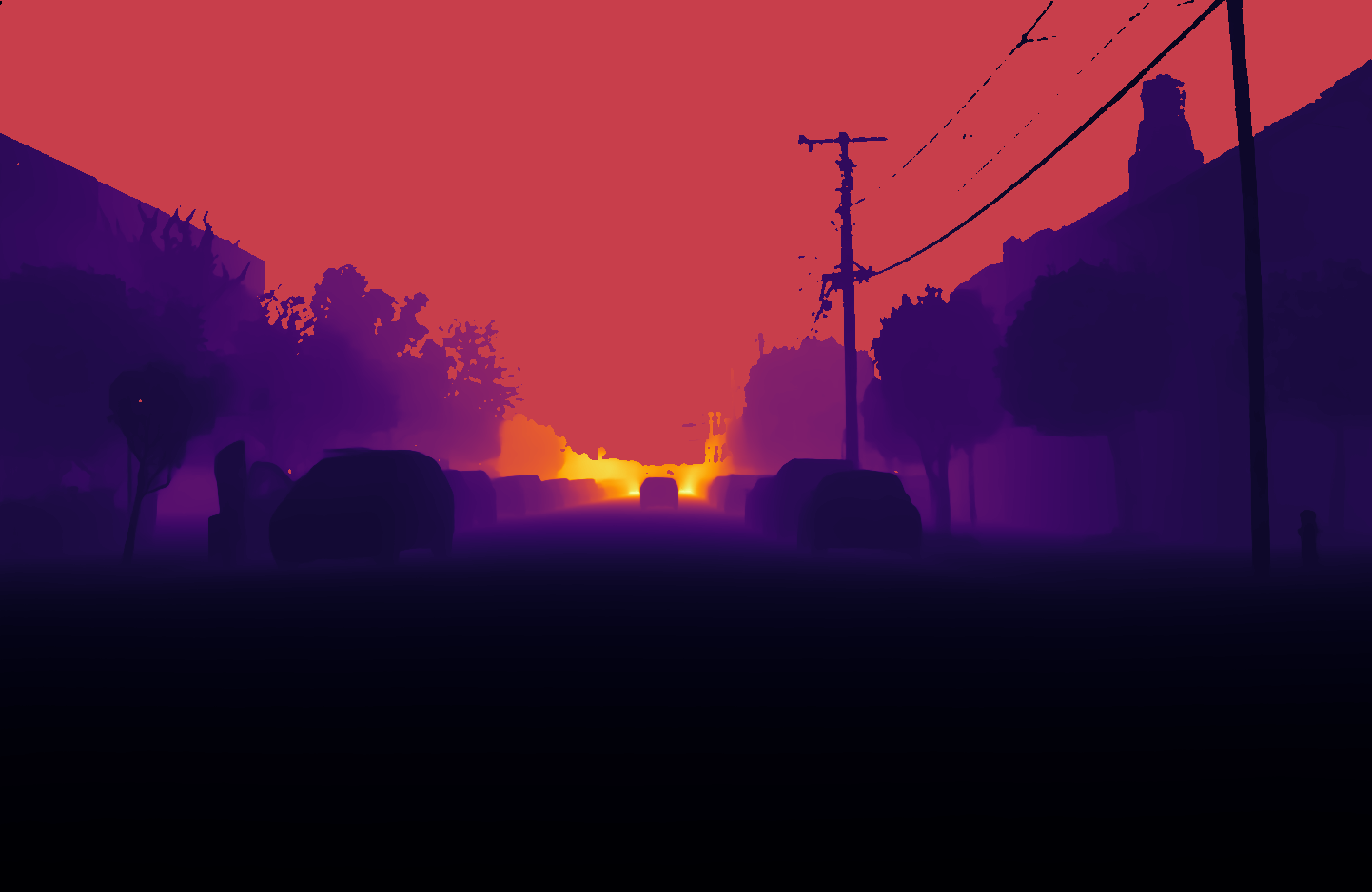}\\[2pt]
{\scriptsize\color{exMute} Image returned with this output.}\end{center}
\begin{exharness}
\exLabel{Harness}{environment update, summarized: new memory entries; new data bindings verbatim}
\begin{exverb}
Memory update: registers 1 new image in memory (tool\_\allowbreak{}generated\_\allowbreak{}image\_\allowbreak{}1,\allowbreak{} from Depth\_\allowbreak{}Estimation\_\allowbreak{}Tool).\\{}
Available data bindings (auto-injected into Python\_\allowbreak{}Coding\_\allowbreak{}Agent\_\allowbreak{}Tool):\\{}
- depth\_\allowbreak{}result: dict (keys: depth\_\allowbreak{}shape,\allowbreak{} colormap,\allowbreak{} size,\allowbreak{} depth\_\allowbreak{}meters)
\end{exverb}
\end{exharness}
\end{exstep}
\begin{exstep}{2}{\toolVG}{}
\exLabel{Model}{text emitted with this call}
\exSayBlock{\textless{}thinking\textgreater{}Now I will locate the crosswalk in the image. The crosswalk is visible on the right side of the road,\allowbreak{} near the bottom of the image. From the depth map (tool\_\allowbreak{}generated\_\allowbreak{}image\_\allowbreak{}1),\allowbreak{} the crosswalk appears in a region with relatively low depth values (dark purple/\allowbreak{}blue),\allowbreak{} indicating it is close to the camera. Next,\allowbreak{} I need to identify any cars that are even closer than this crosswalk. In the original image,\allowbreak{} there is a silver car on the left side of the road,\allowbreak{} just past the stop sign,\allowbreak{} which appears to be very close to the camera. Its position in the depth map also shows a low depth value,\allowbreak{} suggesting it is closer than the crosswalk. To confirm this,\allowbreak{} I will use the Visual\_\allowbreak{}Grounding\_\allowbreak{}Tool to locate both the crosswalk and the cars.\textless{}/\allowbreak{}thinking\textgreater{}}
\exLabel{Arguments}{}
\begin{exverb}
\{\\{}
~~"\allowbreak{}query"\allowbreak{}: "\allowbreak{}crosswalk"\allowbreak{},\allowbreak{}\\{}
~~"\allowbreak{}image\_\allowbreak{}id"\allowbreak{}: "\allowbreak{}input\_\allowbreak{}image\_\allowbreak{}1"\allowbreak{}\\{}
\}
\end{exverb}
\exLabel{Tool output}{}
\begin{exverb}
Visual\_\allowbreak{}Grounding\_\allowbreak{}Tool grounded 1 objects for the prompt \textquotesingle{}crosswalk\textquotesingle{}. Their corresponding bounding boxes (on a 0-1000 scale,\allowbreak{} [x1,\allowbreak{} y1,\allowbreak{} x2,\allowbreak{} y2]) are: \{\textquotesingle{}object 1\textquotesingle{}: [788.82,\allowbreak{} 646.79,\allowbreak{} 1000,\allowbreak{} 744.88]\}. The object numbers correspond to the labeled masks rendered in the following image,\allowbreak{} which you can refer to as tool\_\allowbreak{}generated\_\allowbreak{}image\_\allowbreak{}2.
\end{exverb}
\begin{center}\includegraphics[width=0.45\linewidth,height=0.22\textheight,keepaspectratio]{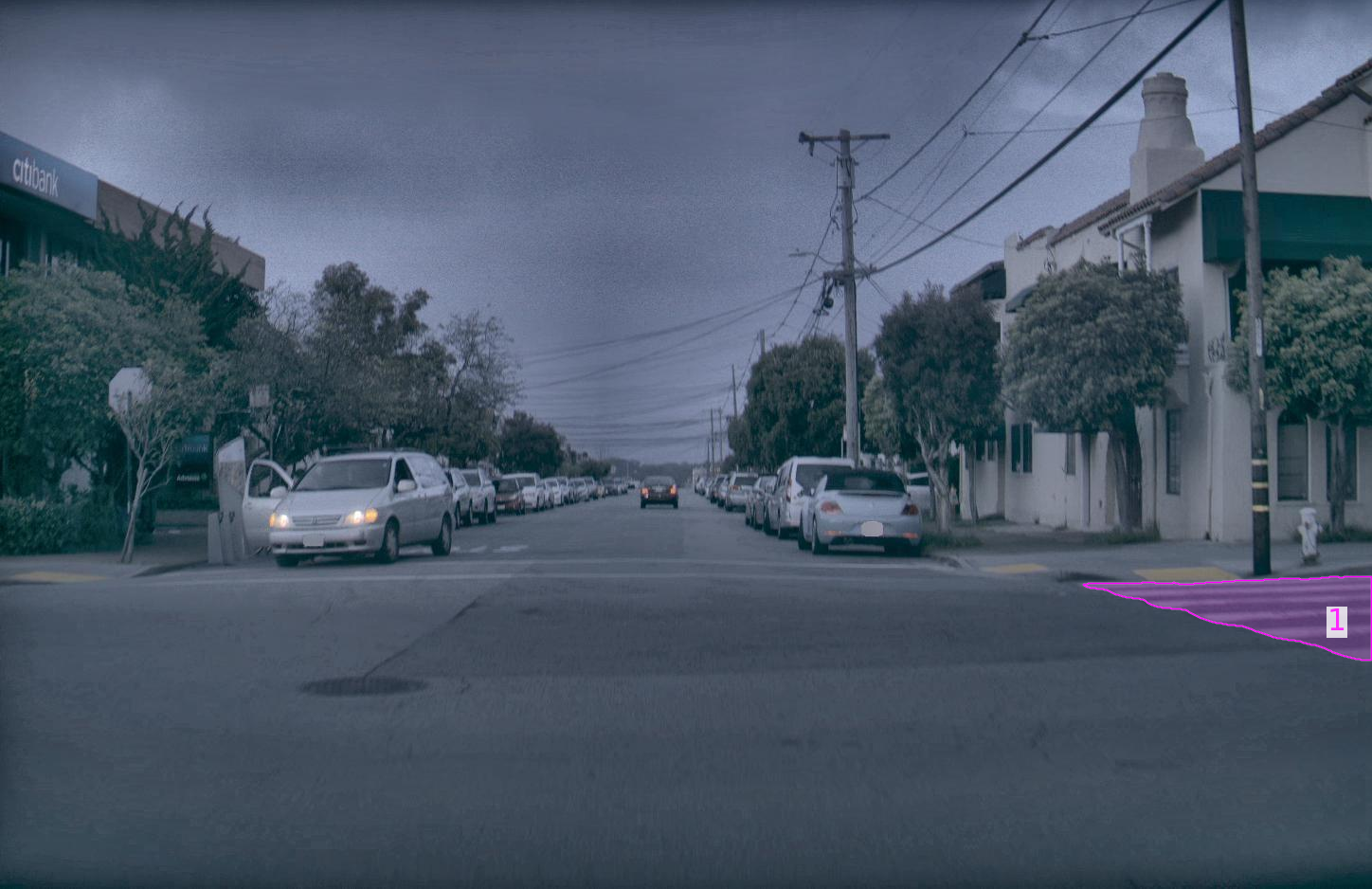}\\[2pt]
{\scriptsize\color{exMute} Image returned with this output.}\end{center}
\begin{exharness}
\exLabel{Harness}{environment update, summarized: new memory entries; new data bindings verbatim}
\begin{exverb}
Memory update: registers 1 new image in memory (tool\_\allowbreak{}generated\_\allowbreak{}image\_\allowbreak{}2,\allowbreak{} from Visual\_\allowbreak{}Grounding\_\allowbreak{}Tool).\\{}
- segment\_\allowbreak{}crosswalk: dict (keys: pred\_\allowbreak{}boxes,\allowbreak{} pred\_\allowbreak{}scores,\allowbreak{} pred\_\allowbreak{}masks,\allowbreak{} num\_\allowbreak{}masks,\allowbreak{} orig\_\allowbreak{}img\_\allowbreak{}h,\allowbreak{} orig\_\allowbreak{}img\_\allowbreak{}w)
\end{exverb}
\end{exharness}
\end{exstep}
\begin{exstep}{3}{\toolVG}{}
\exLabel{Model}{text emitted with this call}
\exSayBlock{\textless{}thinking\textgreater{}Now I will locate the cars in the image using the Visual\_\allowbreak{}Grounding\_\allowbreak{}Tool. I need to identify any cars that are closer to the camera than the crosswalk,\allowbreak{} which is located on the right side of the road near the bottom. The car on the left side of the road appears to be very close to the camera and may be closer than the crosswalk.\textless{}/\allowbreak{}thinking\textgreater{}}
\exLabel{Arguments}{}
\begin{exverb}
\{\\{}
~~"\allowbreak{}query"\allowbreak{}: "\allowbreak{}cars"\allowbreak{},\allowbreak{}\\{}
~~"\allowbreak{}image\_\allowbreak{}id"\allowbreak{}: "\allowbreak{}input\_\allowbreak{}image\_\allowbreak{}1"\allowbreak{}\\{}
\}
\end{exverb}
\exLabel{Tool output}{}
\begin{exverb}
Visual\_\allowbreak{}Grounding\_\allowbreak{}Tool grounded 15 objects for the prompt \textquotesingle{}cars\textquotesingle{}. Their corresponding bounding boxes (on a 0-1000 scale,\allowbreak{} [x1,\allowbreak{} y1,\allowbreak{} x2,\allowbreak{} y2]) are: \{\textquotesingle{}object 1\textquotesingle{}: [176.53,\allowbreak{} 513.87,\allowbreak{} 214.9,\allowbreak{} 620.67],\allowbreak{} \textquotesingle{}object 2\textquotesingle{}: [196.14,\allowbreak{} 505.4,\allowbreak{} 330.53,\allowbreak{} 637.25],\allowbreak{} \textquotesingle{}object 3\textquotesingle{}: [325.42,\allowbreak{} 525.2,\allowbreak{} 345.27,\allowbreak{} 592.08],\allowbreak{} \textquotesingle{}object 4\textquotesingle{}: [335.01,\allowbreak{} 525.61,\allowbreak{} 362.31,\allowbreak{} 590.18],\allowbreak{} \textquotesingle{}object 5\textquotesingle{}: [361.83,\allowbreak{} 532.05,\allowbreak{} 391.95,\allowbreak{} 577.99],\allowbreak{} \textquotesingle{}object 6\textquotesingle{}: [389.67,\allowbreak{} 539.5,\allowbreak{} 403.25,\allowbreak{} 573.49],\allowbreak{} \textquotesingle{}object 7\textquotesingle{}: [401.02,\allowbreak{} 535.78,\allowbreak{} 414.74,\allowbreak{} 568.96],\allowbreak{} \textquotesingle{}object 8\textquotesingle{}: [415.52,\allowbreak{} 537.59,\allowbreak{} 429.71,\allowbreak{} 565.0],\allowbreak{} \textquotesingle{}object 9\textquotesingle{}: [466.41,\allowbreak{} 534.55,\allowbreak{} 494.33,\allowbreak{} 571.55],\allowbreak{} \textquotesingle{}object 10\textquotesingle{}: [515.88,\allowbreak{} 533.44,\allowbreak{} 529.58,\allowbreak{} 567.8],\allowbreak{} \textquotesingle{}object 11\textquotesingle{}: [525.95,\allowbreak{} 530.37,\allowbreak{} 551.8,\allowbreak{} 575.87],\allowbreak{} \textquotesingle{}object 12\textquotesingle{}: [541.83,\allowbreak{} 534.66,\allowbreak{} 565.86,\allowbreak{} 594.67],\allowbreak{} \textquotesingle{}object 13\textquotesingle{}: [558.22,\allowbreak{} 512.4,\allowbreak{} 620.77,\allowbreak{} 606.17],\allowbreak{} \textquotesingle{}object 14\textquotesingle{}: [582.0,\allowbreak{} 524.63,\allowbreak{} 671.94,\allowbreak{} 624.84],\allowbreak{} \textquotesingle{}object 15\textquotesingle{}: [658.99,\allowbreak{} 527.86,\allowbreak{} 679.94,\allowbreak{} 580.83]\}. The object numbers correspond to the labeled masks rendered in the following image,\allowbreak{} which you can refer to as tool\_\allowbreak{}generated\_\allowbreak{}image\_\allowbreak{}3.
\end{exverb}
\begin{center}\includegraphics[width=0.45\linewidth,height=0.22\textheight,keepaspectratio]{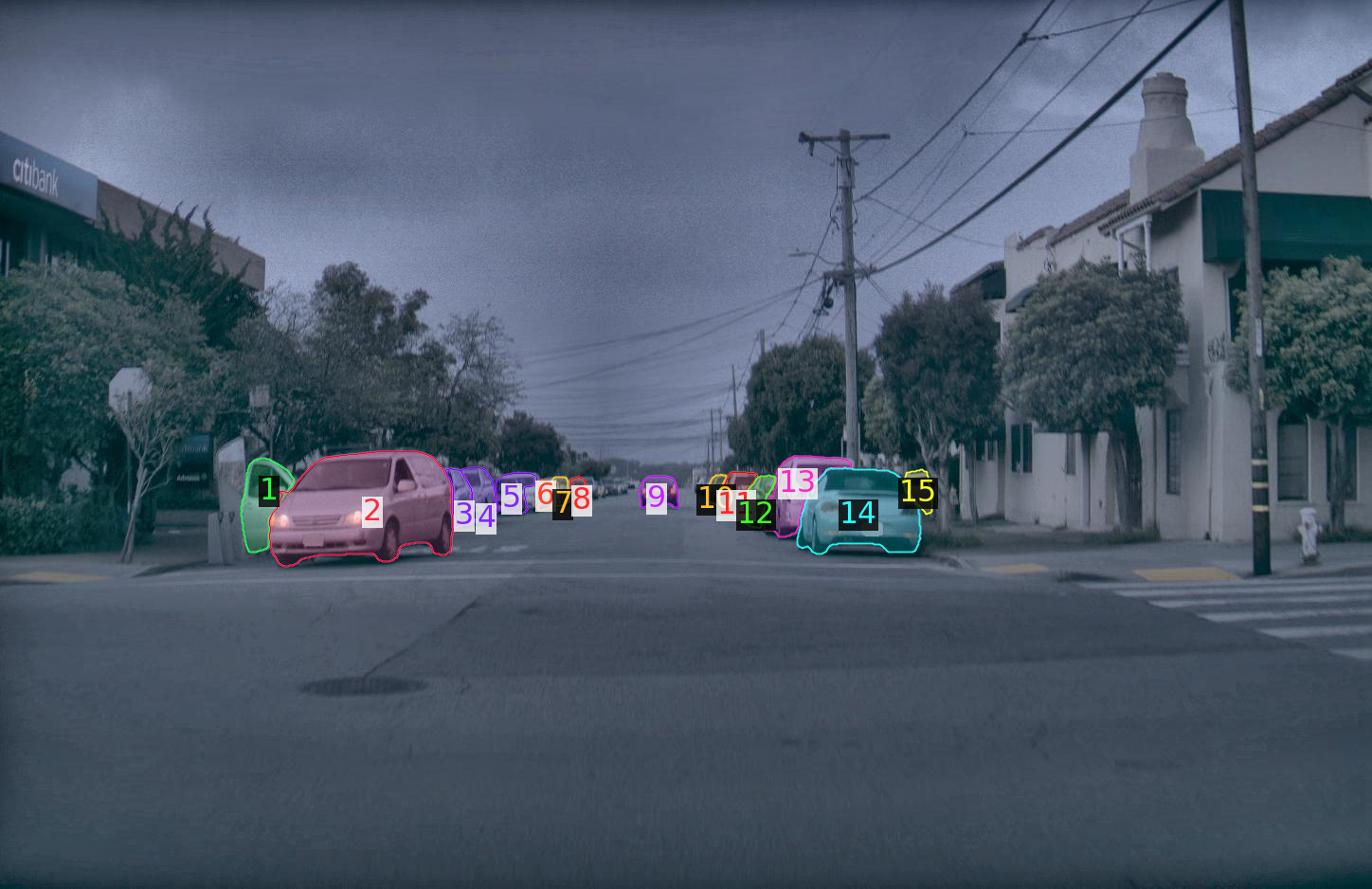}\\[2pt]
{\scriptsize\color{exMute} Image returned with this output.}\end{center}
\begin{exharness}
\exLabel{Harness}{environment update, summarized: new memory entries; new data bindings verbatim}
\begin{exverb}
Memory update: registers 1 new image in memory (tool\_\allowbreak{}generated\_\allowbreak{}image\_\allowbreak{}3,\allowbreak{} from Visual\_\allowbreak{}Grounding\_\allowbreak{}Tool).\\{}
- segment\_\allowbreak{}cars: dict (keys: pred\_\allowbreak{}boxes,\allowbreak{} pred\_\allowbreak{}scores,\allowbreak{} pred\_\allowbreak{}masks,\allowbreak{} num\_\allowbreak{}masks,\allowbreak{} orig\_\allowbreak{}img\_\allowbreak{}h,\allowbreak{} orig\_\allowbreak{}img\_\allowbreak{}w)
\end{exverb}
\end{exharness}
\end{exstep}
\begin{exstep}{4}{\toolCode}{}
\exLabel{Model}{text emitted with this call}
\exSayBlock{\textless{}thinking\textgreater{}Now I will use the Python\_\allowbreak{}Coding\_\allowbreak{}Agent\_\allowbreak{}Tool to compute the average depth at the center of each detected car and compare it with the average depth of the crosswalk. This will allow me to determine if any cars are closer to the camera than the crosswalk.\textless{}/\allowbreak{}thinking\textgreater{}}
\exLabel{Arguments}{}
\begin{exverb}
\{\\{}
~~"\allowbreak{}objective"\allowbreak{}: "\allowbreak{}Compute the average depth at the center of each detected car bounding box and the average depth at the center of the crosswalk bounding box. Compare these depths to determine if any cars are closer to the camera than the crosswalk."\allowbreak{}\\{}
\}
\end{exverb}
\exLabel{Tool output}{}
\begin{exverb}
Coding agent generated and executed the following Python code:\\{}
\textasciigrave{}\textasciigrave{}\textasciigrave{}python\\{}
import numpy as np\\{}
\mbox{}\\{}
\# Extract depth map and its dimensions\\{}
depth\_\allowbreak{}map =\allowbreak{} depth\_\allowbreak{}result["\allowbreak{}depth\_\allowbreak{}meters"\allowbreak{}]\\{}
height,\allowbreak{} width =\allowbreak{} depth\_\allowbreak{}result["\allowbreak{}depth\_\allowbreak{}shape"\allowbreak{}]\\{}
\mbox{}\\{}
\# Extract crosswalk bounding box\\{}
crosswalk\_\allowbreak{}box =\allowbreak{} segment\_\allowbreak{}crosswalk["\allowbreak{}pred\_\allowbreak{}boxes"\allowbreak{}][0]\\{}
crosswalk\_\allowbreak{}center\_\allowbreak{}x =\allowbreak{} (crosswalk\_\allowbreak{}box[0] + crosswalk\_\allowbreak{}box[2]) /\allowbreak{} 2\\{}
crosswalk\_\allowbreak{}center\_\allowbreak{}y =\allowbreak{} (crosswalk\_\allowbreak{}box[1] + crosswalk\_\allowbreak{}box[3]) /\allowbreak{} 2\\{}
\mbox{}\\{}
\# Convert center coordinates to integer (pixel indices)\\{}
crosswalk\_\allowbreak{}center\_\allowbreak{}x\_\allowbreak{}int =\allowbreak{} int(crosswalk\_\allowbreak{}center\_\allowbreak{}x)\\{}
crosswalk\_\allowbreak{}center\_\allowbreak{}y\_\allowbreak{}int =\allowbreak{} int(crosswalk\_\allowbreak{}center\_\allowbreak{}y)\\{}
\mbox{}\\{}
\# Get depth at crosswalk center\\{}
crosswalk\_\allowbreak{}depth =\allowbreak{} depth\_\allowbreak{}map[crosswalk\_\allowbreak{}center\_\allowbreak{}y\_\allowbreak{}int,\allowbreak{} crosswalk\_\allowbreak{}center\_\allowbreak{}x\_\allowbreak{}int]\\{}
\mbox{}\\{}
\# Extract car bounding boxes\\{}
car\_\allowbreak{}boxes =\allowbreak{} segment\_\allowbreak{}cars["\allowbreak{}pred\_\allowbreak{}boxes"\allowbreak{}]\\{}
\mbox{}\\{}
\# Compute average depth at center of each car bounding box\\{}
car\_\allowbreak{}centers\_\allowbreak{}depths =\allowbreak{} []\\{}
for box in car\_\allowbreak{}boxes:\\{}
~~~~center\_\allowbreak{}x =\allowbreak{} (box[0] + box[2]) /\allowbreak{} 2\\{}
~~~~center\_\allowbreak{}y =\allowbreak{} (box[1] + box[3]) /\allowbreak{} 2\\{}
~~~~center\_\allowbreak{}x\_\allowbreak{}int =\allowbreak{} int(center\_\allowbreak{}x)\\{}
~~~~center\_\allowbreak{}y\_\allowbreak{}int =\allowbreak{} int(center\_\allowbreak{}y)\\{}
~~~~car\_\allowbreak{}depth =\allowbreak{} depth\_\allowbreak{}map[center\_\allowbreak{}y\_\allowbreak{}int,\allowbreak{} center\_\allowbreak{}x\_\allowbreak{}int]\\{}
~~~~car\_\allowbreak{}centers\_\allowbreak{}depths.append(car\_\allowbreak{}depth)\\{}
\mbox{}\\{}
\# Compute average depth for all cars\\{}
avg\_\allowbreak{}car\_\allowbreak{}depth =\allowbreak{} np.mean(car\_\allowbreak{}centers\_\allowbreak{}depths)\\{}
\mbox{}\\{}
\# Compare: any car closer than crosswalk?\\{}
cars\_\allowbreak{}closer\_\allowbreak{}than\_\allowbreak{}crosswalk =\allowbreak{} any(car\_\allowbreak{}depth \textless{} crosswalk\_\allowbreak{}depth for car\_\allowbreak{}depth in car\_\allowbreak{}centers\_\allowbreak{}depths)\\{}
\mbox{}\\{}
\# Output results\\{}
print(f"\allowbreak{}Crosswalk center depth: \{crosswalk\_\allowbreak{}depth:.2f\} meters"\allowbreak{})\\{}
print(f"\allowbreak{}Average car center depth: \{avg\_\allowbreak{}car\_\allowbreak{}depth:.2f\} meters"\allowbreak{})\\{}
print(f"\allowbreak{}Any car closer to camera than crosswalk? \{\textquotesingle{}Yes\textquotesingle{} if cars\_\allowbreak{}closer\_\allowbreak{}than\_\allowbreak{}crosswalk else \textquotesingle{}No\textquotesingle{}\}"\allowbreak{})\\{}
\mbox{}\\{}
\# Assign results to variables for result namespace\\{}
crosswalk\_\allowbreak{}center\_\allowbreak{}depth =\allowbreak{} crosswalk\_\allowbreak{}depth\\{}
average\_\allowbreak{}car\_\allowbreak{}center\_\allowbreak{}depth =\allowbreak{} avg\_\allowbreak{}car\_\allowbreak{}depth\\{}
any\_\allowbreak{}car\_\allowbreak{}closer\_\allowbreak{}than\_\allowbreak{}crosswalk =\allowbreak{} cars\_\allowbreak{}closer\_\allowbreak{}than\_\allowbreak{}crosswalk\\{}
\textasciigrave{}\textasciigrave{}\textasciigrave{}\\{}
\mbox{}\\{}
\mbox{}\\{}
\mbox{}\\{}
Execution output:\\{}
Crosswalk center depth: 17.40 meters\\{}
Average car center depth: 54.14 meters\\{}
Any car closer to camera than crosswalk? No
\end{exverb}
\end{exstep}
\begin{exstep}{}{\exTool{exInk}{Final answer}}{}
\exLabel{Model}{}
\exSayBlock{\textless{}thinking\textgreater{}Based on the depth analysis,\allowbreak{} the crosswalk is located at an average depth of 17.40 meters from the camera,\allowbreak{} while the average depth of the cars is 54.14 meters. This indicates that all cars are farther away from the camera than the crosswalk. Therefore,\allowbreak{} there are no cars closer to us than the nearest crosswalk.\textless{}/\allowbreak{}thinking\textgreater{}
\textless{}answer\textgreater{}B. No\textless{}/\allowbreak{}answer\textgreater{}}
\end{exstep}

%% file: text/appendix_examples_steps/church_header.tex
\exOverview{\textbf{Task.} Name the church in a photograph of a dark Gothic interior with a large baroque altar. 

\textbf{Why it is hard.} Many Gothic churches look alike; without tools the model guesses Antwerp Cathedral. 

\textbf{How the agent solves it.} Searching the whole photograph and describing it in text return only look-alike churches for eleven calls. The breakthrough is to crop the altarpiece alone and reverse-search the crop, which finds an archive photo of this exact painting; a final text search confirms the church.}
\exLabel{Input images}{}
\begin{center}\includegraphics[width=0.55\linewidth,height=0.28\textheight,keepaspectratio]{appendix_examples/church_input}\\[2pt]
{\scriptsize\color{exMute} input\_image\_1}\end{center}
\exLabel{Question}{as sent to the model}
\begin{exverb}
Question: What church is depicted in this picture?
\end{exverb}
\exLabel{Without tools}{GPT-6 Sol's full response to the same question}
\begin{exverb}
This appears to be the Cathedral of Our Lady in Antwerp,\allowbreak{} Belgium.
\end{exverb}

%% file: text/appendix_examples_steps/church.tex
\begin{exharness}
\exLabel{Harness}{environment update, shown in full}
\begin{exverb}
Environment Update (not a new question):\\{}
Currently available image IDs for tool calls:\\{}
- input\_\allowbreak{}image\_\allowbreak{}1: user\_\allowbreak{}input\\{}
When a tool needs image input,\allowbreak{} pass one of these IDs via the *\_\allowbreak{}image\_\allowbreak{}id field.\\{}
Do NOT invent image file names or paths.
\end{exverb}
\end{exharness}
\begin{exstep}{1}{\toolIS}{}
\exWalk{Reverse image search on the whole photograph. The matches are other churches with similar interiors.}
\exLabel{Arguments}{}
\begin{exverb}
\{\\{}
~~"\allowbreak{}image\_\allowbreak{}id"\allowbreak{}: "\allowbreak{}input\_\allowbreak{}image\_\allowbreak{}1"\allowbreak{}\\{}
\}
\end{exverb}
\exLabel{Tool output}{}
\begin{exverb}
Tool \textasciigrave{}Image\_\allowbreak{}Search\_\allowbreak{}Tool\textasciigrave{} (tool\_\allowbreak{}call\_\allowbreak{}id=\allowbreak{}call\_\allowbreak{}oH6X4bFwHKAz5nlS5Tjs5Fg8) returned:\\{}
\{"\allowbreak{}ok"\allowbreak{}: true,\allowbreak{} "\allowbreak{}result"\allowbreak{}: "\allowbreak{}[Image Search — Visual Matches]\\{}
1. St Hilda\textquotesingle{}s: “It was like a Cathedral!” – Shrine of the Relic Image of Our Lady of Guadalupe\\{}
~~~Source: relicourladyofguadalupe.wordpress.com/\allowbreak{}.../\allowbreak{}st-hildas-it-was-like-a-cathedral   [webpage\_\allowbreak{}url\_\allowbreak{}1]\\{}
~~~Image: [tool\_\allowbreak{}generated\_\allowbreak{}image\_\allowbreak{}1]\\{}
2. Oh Mary conceived without sin,\allowbreak{} pray for us who have recourse to Thee!\\{}
~~~Source: instagram.com/\allowbreak{}p/\allowbreak{}DRkWXrTj0OL   [webpage\_\allowbreak{}url\_\allowbreak{}2]\\{}
~~~Image: [tool\_\allowbreak{}generated\_\allowbreak{}image\_\allowbreak{}2]
\end{exverb}
\exCut{8 more results not shown (10 in total).}
\begin{exverb}
~~~Image: [tool\_\allowbreak{}generated\_\allowbreak{}image\_\allowbreak{}12]"\allowbreak{},\allowbreak{} "\allowbreak{}error"\allowbreak{}: null\}
\end{exverb}
\begin{exharness}
\exLabel{Harness}{environment update, summarized: new memory entries; new data bindings verbatim}
\begin{exverb}
Memory update: registers 12 new images in memory (tool\_\allowbreak{}generated\_\allowbreak{}image\_\allowbreak{}1 to tool\_\allowbreak{}generated\_\allowbreak{}image\_\allowbreak{}12,\allowbreak{} from Image\_\allowbreak{}Search\_\allowbreak{}Tool); registers 8 new webpage URLs in memory (webpage\_\allowbreak{}url\_\allowbreak{}1 to webpage\_\allowbreak{}url\_\allowbreak{}8,\allowbreak{} from Image\_\allowbreak{}Search\_\allowbreak{}Tool).
\end{exverb}
\end{exharness}
\end{exstep}
\begin{exstep}{2}{\toolWV}{}
\exWalk{The agent visits the top match's page to check it; it describes a different church.}
\exLabel{Arguments}{}
\begin{exverb}
\{\\{}
~~"\allowbreak{}urls"\allowbreak{}: ["\allowbreak{}webpage\_\allowbreak{}url\_\allowbreak{}1"\allowbreak{}],\allowbreak{}\\{}
~~"\allowbreak{}goal"\allowbreak{}: "\allowbreak{}Determine whether the photograph of dark Gothic church with ornate dark wood and gold baroque high altar is shown and identify church location."\allowbreak{}\\{}
\}
\end{exverb}
\exLabel{Tool output}{}
\begin{exverb}
Tool \textasciigrave{}Webpage\_\allowbreak{}Visit\_\allowbreak{}Tool\textasciigrave{} (tool\_\allowbreak{}call\_\allowbreak{}id=\allowbreak{}call\_\allowbreak{}86VJYX7gpuDLXJWAcBH74xsS) returned:\\{}
\{"\allowbreak{}ok"\allowbreak{}: true,\allowbreak{} "\allowbreak{}result"\allowbreak{}: "\allowbreak{}The useful information in https:/\allowbreak{}/\allowbreak{}relicourladyofguadalupe.wordpress.com/\allowbreak{}2017/\allowbreak{}10/\allowbreak{}14/\allowbreak{}st-hildas-it-was-like-a-cathedral/\allowbreak{} for user goal Determine whether the photograph of dark Gothic church with ornate dark wood and gold baroque high altar is shown and identify church location. as follows: \\{}
\mbox{}\\{}
Evidence in page: \\{}
St Hilda’s: “It was like a Cathedral!” – Shrine of the Relic Image of Our Lady of Guadalupe\\{}
\mbox{}\\{}
Father Michael Doherty,\allowbreak{} Parish Priest of St Hilda’s Church,\allowbreak{} Stevenage,\allowbreak{} was clearly delighted with the Pilgrimage Visit of the Relic Image. “It was like being a cathedral for the weekend!” he exclaimed. “St Hilda’s has become the National Shrine of Our Lady of Guadalupe!”\\{}
\mbox{}\\{}
He was standing in his beautiful church with its cathedral-sized sanctuary,\allowbreak{} which was lavishly decorated with flowers – mainly roses for Our Lady,\allowbreak{} but also a number of cheerful sunflowers on the back wall and,\allowbreak{} of course,\allowbreak{} with the Relic Image taking pride of place in the sanctuary.
\end{exverb}
\exCut{The remaining 5,465 characters of this output are not shown (6,528 in total).}
\end{exstep}
\begin{exstep}{3}{\toolTS}{}
\exWalk{A text search that describes the interior in words. The top result is Antwerp Cathedral, the same guess the model makes without tools.}
\exLabel{Arguments}{}
\begin{exverb}
\{\\{}
~~"\allowbreak{}query"\allowbreak{}: "\allowbreak{}church interior gothic stone pointed arch dark carved wooden baroque altar painting virgin mary angels chandeliers"\allowbreak{},\allowbreak{}\\{}
~~"\allowbreak{}search\_\allowbreak{}type"\allowbreak{}: "\allowbreak{}images"\allowbreak{}\\{}
\}
\end{exverb}
\exLabel{Tool output}{}
\begin{exverb}
Tool \textasciigrave{}Text\_\allowbreak{}Search\_\allowbreak{}Tool\textasciigrave{} (tool\_\allowbreak{}call\_\allowbreak{}id=\allowbreak{}call\_\allowbreak{}C0u5rG9aZlse8h5wvlCYCE2n) returned:\\{}
\{"\allowbreak{}ok"\allowbreak{}: true,\allowbreak{} "\allowbreak{}result"\allowbreak{}: "\allowbreak{}[Search Results — Type: images]\\{}
1. Cathedral of Our Lady in Antwerp: Rubens,\allowbreak{} Gothic architecture and UNESCO\\{}
~~~Image: [tool\_\allowbreak{}generated\_\allowbreak{}image\_\allowbreak{}13]\\{}
~~~Source: FromPlaceToPlace.travel   [webpage\_\allowbreak{}url\_\allowbreak{}9]\\{}
~~~Size: 768x1024\\{}
2. Baroque Altar\\{}
~~~Image: [tool\_\allowbreak{}generated\_\allowbreak{}image\_\allowbreak{}14]\\{}
~~~Source: Instagram   [webpage\_\allowbreak{}url\_\allowbreak{}10]\\{}
~~~Size: 640x1136
\end{exverb}
\exCut{8 more results not shown (10 in total).}
\begin{exverb}
~~~Size: 800x533"\allowbreak{},\allowbreak{} "\allowbreak{}error"\allowbreak{}: null\}
\end{exverb}
\begin{exharness}
\exLabel{Harness}{environment update, summarized: new memory entries; new data bindings verbatim}
\begin{exverb}
Memory update: registers 10 new images in memory (tool\_\allowbreak{}generated\_\allowbreak{}image\_\allowbreak{}13 to tool\_\allowbreak{}generated\_\allowbreak{}image\_\allowbreak{}22,\allowbreak{} from Text\_\allowbreak{}Search\_\allowbreak{}Tool); registers 9 new webpage URLs in memory (webpage\_\allowbreak{}url\_\allowbreak{}9 to webpage\_\allowbreak{}url\_\allowbreak{}17,\allowbreak{} from Text\_\allowbreak{}Search\_\allowbreak{}Tool).
\end{exverb}
\end{exharness}
\end{exstep}
\begin{exstep}{4}{\toolZoom}{}
\exWalk{A wide crop of the nave and altar, used as a new image to search with.}
\exLabel{Arguments}{}
\begin{exverb}
\{\\{}
~~"\allowbreak{}image\_\allowbreak{}id"\allowbreak{}: "\allowbreak{}input\_\allowbreak{}image\_\allowbreak{}1"\allowbreak{},\allowbreak{}\\{}
~~"\allowbreak{}coords"\allowbreak{}: [125,\allowbreak{} 155,\allowbreak{} 830,\allowbreak{} 905],\allowbreak{}\\{}
~~"\allowbreak{}padding"\allowbreak{}: 50\\{}
\}
\end{exverb}
\exLabel{Tool output}{}
\begin{exverb}
Tool \textasciigrave{}Zoom\_\allowbreak{}In\_\allowbreak{}Tool\textasciigrave{} (tool\_\allowbreak{}call\_\allowbreak{}id=\allowbreak{}call\_\allowbreak{}pjFOkd8Yz6Q1ZeiGYRJcNHfo) returned:\\{}
Zoomed in image rendered below; refer to it as tool\_\allowbreak{}generated\_\allowbreak{}image\_\allowbreak{}23.
\end{exverb}
\begin{center}\includegraphics[width=0.45\linewidth,height=0.22\textheight,keepaspectratio]{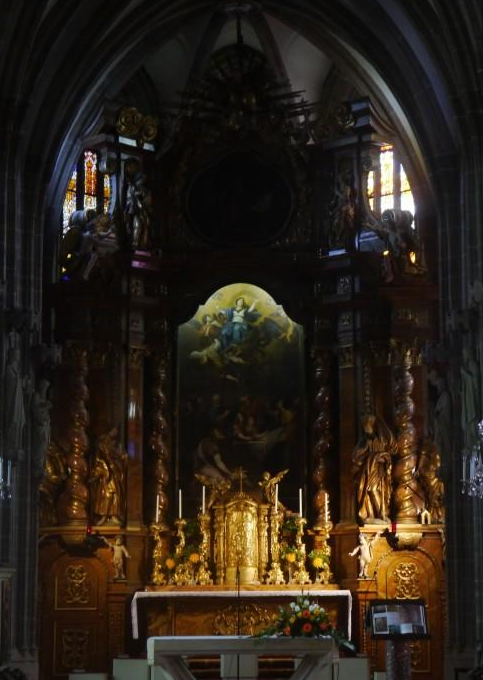}\\[2pt]
{\scriptsize\color{exMute} Crop of the nave and altar.}\end{center}
\begin{exharness}
\exLabel{Harness}{environment update, summarized: new memory entries; new data bindings verbatim}
\begin{exverb}
Memory update: registers 1 new image in memory (tool\_\allowbreak{}generated\_\allowbreak{}image\_\allowbreak{}23,\allowbreak{} from Zoom\_\allowbreak{}In\_\allowbreak{}Tool).
\end{exverb}
\end{exharness}
\end{exstep}
\begin{exstep}{5}{\toolIS}{}
\exWalk{Reverse image search on that crop. The matches are still generic Gothic interiors.}
\exLabel{Arguments}{}
\begin{exverb}
\{\\{}
~~"\allowbreak{}image\_\allowbreak{}id"\allowbreak{}: "\allowbreak{}tool\_\allowbreak{}generated\_\allowbreak{}image\_\allowbreak{}23"\allowbreak{}\\{}
\}
\end{exverb}
\exLabel{Tool output}{}
\begin{exverb}
Tool \textasciigrave{}Image\_\allowbreak{}Search\_\allowbreak{}Tool\textasciigrave{} (tool\_\allowbreak{}call\_\allowbreak{}id=\allowbreak{}call\_\allowbreak{}RJdprFbLE4EF2X3BcNmtd6GK) returned:\\{}
\{"\allowbreak{}ok"\allowbreak{}: true,\allowbreak{} "\allowbreak{}result"\allowbreak{}: "\allowbreak{}[Image Search — Visual Matches]\\{}
1. So beautiful Gothic Architecture!\\{}
~~~Source: facebook.com/\allowbreak{}.../\allowbreak{}10241119283933448   [webpage\_\allowbreak{}url\_\allowbreak{}18]\\{}
~~~Image: [tool\_\allowbreak{}generated\_\allowbreak{}image\_\allowbreak{}24]\\{}
2. This is St. Salvator Church,\allowbreak{} a Roman Catholic Church in the Klementinum,\allowbreak{} in Old Town,\allowbreak{} Prague,\allowbreak{} Czech Republic. As early as 1578 - 1581,\allowbreak{} the Society of Jesus,\allowbreak{} led by Rector Giovanni\\{}
~~~Source: reddit.com/\allowbreak{}.../\allowbreak{}this\_\allowbreak{}is\_\allowbreak{}st\_\allowbreak{}salvator\_\allowbreak{}church\_\allowbreak{}a\_\allowbreak{}roman\_\allowbreak{}catholic   [webpage\_\allowbreak{}url\_\allowbreak{}19]\\{}
~~~Image: [tool\_\allowbreak{}generated\_\allowbreak{}image\_\allowbreak{}25]
\end{exverb}
\exCut{8 more results not shown (10 in total).}
\begin{exverb}
~~~Image: [tool\_\allowbreak{}generated\_\allowbreak{}image\_\allowbreak{}34]"\allowbreak{},\allowbreak{} "\allowbreak{}error"\allowbreak{}: null\}
\end{exverb}
\begin{exharness}
\exLabel{Harness}{environment update, summarized: new memory entries; new data bindings verbatim}
\begin{exverb}
Memory update: registers 11 new images in memory (tool\_\allowbreak{}generated\_\allowbreak{}image\_\allowbreak{}24 to tool\_\allowbreak{}generated\_\allowbreak{}image\_\allowbreak{}34,\allowbreak{} from Image\_\allowbreak{}Search\_\allowbreak{}Tool); registers 9 new webpage URLs in memory (webpage\_\allowbreak{}url\_\allowbreak{}18 to webpage\_\allowbreak{}url\_\allowbreak{}26,\allowbreak{} from Image\_\allowbreak{}Search\_\allowbreak{}Tool).
\end{exverb}
\end{exharness}
\end{exstep}
\begin{exstep}{6}{\toolTS}{}
\exWalk{A text search guessing a specific cathedral, with German keywords. It surfaces Münster Cathedral.}
\exLabel{Arguments}{}
\begin{exverb}
\{\\{}
~~"\allowbreak{}query"\allowbreak{}: "\allowbreak{}\textbackslash{}"\allowbreak{}Stephansdom\textbackslash{}"\allowbreak{} \textbackslash{}"\allowbreak{}altar\textbackslash{}"\allowbreak{} \textbackslash{}"\allowbreak{}Maria\textbackslash{}"\allowbreak{} \textbackslash{}"\allowbreak{}barock\textbackslash{}"\allowbreak{} \textbackslash{}"\allowbreak{}Seitenaltar\textbackslash{}"\allowbreak{} schwarze Säulen"\allowbreak{},\allowbreak{}\\{}
~~"\allowbreak{}search\_\allowbreak{}type"\allowbreak{}: "\allowbreak{}images"\allowbreak{}\\{}
\}
\end{exverb}
\exLabel{Tool output}{}
\begin{exverb}
Tool \textasciigrave{}Text\_\allowbreak{}Search\_\allowbreak{}Tool\textasciigrave{} (tool\_\allowbreak{}call\_\allowbreak{}id=\allowbreak{}call\_\allowbreak{}1zLdigDU9Lxb3Ukttglm3K77) returned:\\{}
\{"\allowbreak{}ok"\allowbreak{}: true,\allowbreak{} "\allowbreak{}result"\allowbreak{}: "\allowbreak{}[Search Results — Type: images]\\{}
1. wdf - wupper digitale fotografie - Münster,\allowbreak{} St. Paulus-Dom\\{}
~~~Image: [tool\_\allowbreak{}generated\_\allowbreak{}image\_\allowbreak{}35]\\{}
~~~Source: wupper digitale fotografie   [webpage\_\allowbreak{}url\_\allowbreak{}27]\\{}
~~~Size: 1000x1500\\{}
2. wdf - wupper digitale fotografie - [\,] Sakralkunst\\{}
~~~Image: [tool\_\allowbreak{}generated\_\allowbreak{}image\_\allowbreak{}36]\\{}
~~~Source: wupper digitale fotografie   [webpage\_\allowbreak{}url\_\allowbreak{}28]\\{}
~~~Size: 764x1021
\end{exverb}
\exCut{8 more results not shown (10 in total).}
\begin{exverb}
~~~Size: 764x1021"\allowbreak{},\allowbreak{} "\allowbreak{}error"\allowbreak{}: null\}
\end{exverb}
\begin{exharness}
\exLabel{Harness}{environment update, summarized: new memory entries; new data bindings verbatim}
\begin{exverb}
Memory update: registers 10 new images in memory (tool\_\allowbreak{}generated\_\allowbreak{}image\_\allowbreak{}35 to tool\_\allowbreak{}generated\_\allowbreak{}image\_\allowbreak{}44,\allowbreak{} from Text\_\allowbreak{}Search\_\allowbreak{}Tool); registers 5 new webpage URLs in memory (webpage\_\allowbreak{}url\_\allowbreak{}27 to webpage\_\allowbreak{}url\_\allowbreak{}31,\allowbreak{} from Text\_\allowbreak{}Search\_\allowbreak{}Tool).
\end{exverb}
\end{exharness}
\end{exstep}
\begin{exstep}{7}{\toolIS}{}
\exWalk{Reverse image search on one of the returned pictures, to test the Münster lead.}
\exLabel{Arguments}{}
\begin{exverb}
\{\\{}
~~"\allowbreak{}image\_\allowbreak{}id"\allowbreak{}: "\allowbreak{}tool\_\allowbreak{}generated\_\allowbreak{}image\_\allowbreak{}35"\allowbreak{}\\{}
\}
\end{exverb}
\exLabel{Tool output}{}
\begin{exverb}
Tool \textasciigrave{}Image\_\allowbreak{}Search\_\allowbreak{}Tool\textasciigrave{} (tool\_\allowbreak{}call\_\allowbreak{}id=\allowbreak{}call\_\allowbreak{}JHAzpfmSFZFnJj1CMWty12ej) returned:\\{}
\{"\allowbreak{}ok"\allowbreak{}: true,\allowbreak{} "\allowbreak{}result"\allowbreak{}: "\allowbreak{}[Image Search — Visual Matches]\\{}
1. wdf - wupper digitale fotografie - Münster,\allowbreak{} St. Paulus-Dom\\{}
~~~Source: kdwupper.de/\allowbreak{}paulusdom-in-münster.html   [webpage\_\allowbreak{}url\_\allowbreak{}32]\\{}
~~~Image: [tool\_\allowbreak{}generated\_\allowbreak{}image\_\allowbreak{}35]\\{}
2. Rochester NY Chapter \textbar{} American Guild of Organists\\{}
~~~Source: rocago.org   [webpage\_\allowbreak{}url\_\allowbreak{}33]\\{}
~~~Image: [tool\_\allowbreak{}generated\_\allowbreak{}image\_\allowbreak{}45]
\end{exverb}
\exCut{8 more results not shown (10 in total).}
\begin{exverb}
~~~Image: [tool\_\allowbreak{}generated\_\allowbreak{}image\_\allowbreak{}54]"\allowbreak{},\allowbreak{} "\allowbreak{}error"\allowbreak{}: null\}
\end{exverb}
\begin{exharness}
\exLabel{Harness}{environment update, summarized: new memory entries; new data bindings verbatim}
\begin{exverb}
Memory update: registers 10 new images in memory (tool\_\allowbreak{}generated\_\allowbreak{}image\_\allowbreak{}45 to tool\_\allowbreak{}generated\_\allowbreak{}image\_\allowbreak{}54,\allowbreak{} from Image\_\allowbreak{}Search\_\allowbreak{}Tool); registers 9 new webpage URLs in memory (webpage\_\allowbreak{}url\_\allowbreak{}32 to webpage\_\allowbreak{}url\_\allowbreak{}40,\allowbreak{} from Image\_\allowbreak{}Search\_\allowbreak{}Tool).
\end{exverb}
\end{exharness}
\end{exstep}
\begin{exstep}{8}{\toolTS}{}
\exWalk{A text search on the Münster lead, which brings up other German cathedrals.}
\exLabel{Arguments}{}
\begin{exverb}
\{\\{}
~~"\allowbreak{}query"\allowbreak{}: "\allowbreak{}\textbackslash{}"\allowbreak{}St. Paulus Dom\textbackslash{}"\allowbreak{} \textbackslash{}"\allowbreak{}altar\textbackslash{}"\allowbreak{} \textbackslash{}"\allowbreak{}Himmelfahrt\textbackslash{}"\allowbreak{} \textbackslash{}"\allowbreak{}barock\textbackslash{}"\allowbreak{}"\allowbreak{},\allowbreak{}\\{}
~~"\allowbreak{}search\_\allowbreak{}type"\allowbreak{}: "\allowbreak{}images"\allowbreak{}\\{}
\}
\end{exverb}
\exLabel{Tool output}{}
\begin{exverb}
Tool \textasciigrave{}Text\_\allowbreak{}Search\_\allowbreak{}Tool\textasciigrave{} (tool\_\allowbreak{}call\_\allowbreak{}id=\allowbreak{}call\_\allowbreak{}juQSJEm9vDllrbjT4ZcCblRU) returned:\\{}
\{"\allowbreak{}ok"\allowbreak{}: true,\allowbreak{} "\allowbreak{}result"\allowbreak{}: "\allowbreak{}[Search Results — Type: images]\\{}
1. Dom zu Eichstätt – Wikipedia\\{}
~~~Image: [tool\_\allowbreak{}generated\_\allowbreak{}image\_\allowbreak{}55]\\{}
~~~Source: Wikipedia   [webpage\_\allowbreak{}url\_\allowbreak{}41]\\{}
~~~Size: 250x333\\{}
2. Monaco di Baviera - Chiesa dello Spirito Santo (1724-1730),\allowbreak{} affreschi della  navata dipinti da Cosmas Damian Asam tra il 1724 e il 1730. Rappresentano  Re Davide con l\textquotesingle{}arpa e la fondazione dell\textquotesingle{}Ospedale\\{}
~~~Image: [tool\_\allowbreak{}generated\_\allowbreak{}image\_\allowbreak{}56]\\{}
~~~Source: Instagram   [webpage\_\allowbreak{}url\_\allowbreak{}42]\\{}
~~~Size: 1440x1920
\end{exverb}
\exCut{8 more results not shown (10 in total).}
\begin{exverb}
~~~Size: 764x1146"\allowbreak{},\allowbreak{} "\allowbreak{}error"\allowbreak{}: null\}
\end{exverb}
\begin{exharness}
\exLabel{Harness}{environment update, summarized: new memory entries; new data bindings verbatim}
\begin{exverb}
Memory update: registers 10 new images in memory (tool\_\allowbreak{}generated\_\allowbreak{}image\_\allowbreak{}55 to tool\_\allowbreak{}generated\_\allowbreak{}image\_\allowbreak{}64,\allowbreak{} from Text\_\allowbreak{}Search\_\allowbreak{}Tool); registers 9 new webpage URLs in memory (webpage\_\allowbreak{}url\_\allowbreak{}41 to webpage\_\allowbreak{}url\_\allowbreak{}49,\allowbreak{} from Text\_\allowbreak{}Search\_\allowbreak{}Tool).
\end{exverb}
\end{exharness}
\end{exstep}
\begin{exstep}{9}{\toolWV}{}
\exWalk{The agent visits a page about Münster Cathedral's altars. The descriptions do not match the photograph, and the page summary loops.}
\exLabel{Arguments}{}
\begin{exverb}
\{\\{}
~~"\allowbreak{}urls"\allowbreak{}: ["\allowbreak{}webpage\_\allowbreak{}url\_\allowbreak{}32"\allowbreak{}],\allowbreak{}\\{}
~~"\allowbreak{}goal"\allowbreak{}: "\allowbreak{}Identify whether this Münster St Paulus Dom contains a tall dark wooden Baroque altarpiece with an image of Mary dressed blue above saint kneeling and gold tabernacle,\allowbreak{} and which altar it is."\allowbreak{}\\{}
\}
\end{exverb}
\exLabel{Tool output}{}
\begin{exverb}
Tool \textasciigrave{}Webpage\_\allowbreak{}Visit\_\allowbreak{}Tool\textasciigrave{} (tool\_\allowbreak{}call\_\allowbreak{}id=\allowbreak{}call\_\allowbreak{}10Se3cWeACLTlacvvK4aENmK) returned:\\{}
\{"\allowbreak{}ok"\allowbreak{}: true,\allowbreak{} "\allowbreak{}result"\allowbreak{}: "\allowbreak{}The useful information in https:/\allowbreak{}/\allowbreak{}www.kdwupper.de/\allowbreak{}paulusdom-in-münster.html for user goal Identify whether this Münster St Paulus Dom contains a tall dark wooden Baroque altarpiece with an image of Mary dressed blue above saint kneeling and gold tabernacle,\allowbreak{} and which altar it is. as follows: \\{}
\mbox{}\\{}
Evidence in page: \\{}
\{\\{}
~~"\allowbreak{}rationale"\allowbreak{}: "\allowbreak{}The user is seeking to identify a specific Baroque altarpiece in the St. Paulus-Dom,\allowbreak{} Münster,\allowbreak{} described as tall,\allowbreak{} dark wooden,\allowbreak{} with an image of Mary dressed blue above a kneeling saint and a gold tabernacle. The webpage content contains detailed descriptions of altars and their features,\allowbreak{} including the Hauptaltar and the Laurentiusaltar,\allowbreak{} which match this description."\allowbreak{},\allowbreak{}
\end{exverb}
\exCut{The remaining 130,110 characters of this output are not shown (130,916 in total).}
\end{exstep}
\begin{exstep}{10}{\toolTS}{}
\exWalk{Another descriptive text search; it returns stock photos of altars.}
\exLabel{Arguments}{}
\begin{exverb}
\{\\{}
~~"\allowbreak{}query"\allowbreak{}: "\allowbreak{}gothic cathedral baroque altar \textbackslash{}"\allowbreak{}Madonna\textbackslash{}"\allowbreak{} \textbackslash{}"\allowbreak{}kneeling\textbackslash{}"\allowbreak{} dark wood gold tabernacle"\allowbreak{},\allowbreak{}\\{}
~~"\allowbreak{}search\_\allowbreak{}type"\allowbreak{}: "\allowbreak{}images"\allowbreak{}\\{}
\}
\end{exverb}
\exLabel{Tool output}{}
\begin{exverb}
Tool \textasciigrave{}Text\_\allowbreak{}Search\_\allowbreak{}Tool\textasciigrave{} (tool\_\allowbreak{}call\_\allowbreak{}id=\allowbreak{}call\_\allowbreak{}vvXGQlwI8M5zkRMwcdjcA3CR) returned:\\{}
\{"\allowbreak{}ok"\allowbreak{}: true,\allowbreak{} "\allowbreak{}result"\allowbreak{}: "\allowbreak{}[Search Results — Type: images]\\{}
1. 676 Gold Tabernacle Stock Photos - Free \& Royalty-Free Stock Photos from  Dreamstime\\{}
~~~Image: [tool\_\allowbreak{}generated\_\allowbreak{}image\_\allowbreak{}65]\\{}
~~~Source: Dreamstime   [webpage\_\allowbreak{}url\_\allowbreak{}50]\\{}
~~~Size: 600x900\\{}
2. 676 Gold Tabernacle Stock Photos - Free \& Royalty-Free Stock Photos from  Dreamstime\\{}
~~~Image: [tool\_\allowbreak{}generated\_\allowbreak{}image\_\allowbreak{}66]\\{}
~~~Source: Dreamstime   [webpage\_\allowbreak{}url\_\allowbreak{}50]\\{}
~~~Size: 600x900
\end{exverb}
\exCut{8 more results not shown (10 in total).}
\begin{exverb}
~~~Size: 600x900"\allowbreak{},\allowbreak{} "\allowbreak{}error"\allowbreak{}: null\}
\end{exverb}
\begin{exharness}
\exLabel{Harness}{environment update, summarized: new memory entries; new data bindings verbatim}
\begin{exverb}
Memory update: registers 10 new images in memory (tool\_\allowbreak{}generated\_\allowbreak{}image\_\allowbreak{}65 to tool\_\allowbreak{}generated\_\allowbreak{}image\_\allowbreak{}74,\allowbreak{} from Text\_\allowbreak{}Search\_\allowbreak{}Tool); registers 8 new webpage URLs in memory (webpage\_\allowbreak{}url\_\allowbreak{}50 to webpage\_\allowbreak{}url\_\allowbreak{}57,\allowbreak{} from Text\_\allowbreak{}Search\_\allowbreak{}Tool).
\end{exverb}
\end{exharness}
\end{exstep}
\begin{exstep}{11}{\toolTS}{}
\exWalk{A German-language descriptive search; still no match.}
\exLabel{Arguments}{}
\begin{exverb}
\{\\{}
~~"\allowbreak{}query"\allowbreak{}: "\allowbreak{}\textbackslash{}"\allowbreak{}gotische\textbackslash{}"\allowbreak{} \textbackslash{}"\allowbreak{}Kirche\textbackslash{}"\allowbreak{} \textbackslash{}"\allowbreak{}barocker\textbackslash{}"\allowbreak{} \textbackslash{}"\allowbreak{}Hochaltar\textbackslash{}"\allowbreak{} \textbackslash{}"\allowbreak{}Rosenkranz\textbackslash{}"\allowbreak{} \textbackslash{}"\allowbreak{}Tabernakel\textbackslash{}"\allowbreak{}"\allowbreak{},\allowbreak{}\\{}
~~"\allowbreak{}search\_\allowbreak{}type"\allowbreak{}: "\allowbreak{}images"\allowbreak{}\\{}
\}
\end{exverb}
\exLabel{Tool output}{}
\begin{exverb}
Tool \textasciigrave{}Text\_\allowbreak{}Search\_\allowbreak{}Tool\textasciigrave{} (tool\_\allowbreak{}call\_\allowbreak{}id=\allowbreak{}call\_\allowbreak{}W4yiz0YOUUZ9djQcKEyENQj6) returned:\\{}
\{"\allowbreak{}ok"\allowbreak{}: true,\allowbreak{} "\allowbreak{}result"\allowbreak{}: "\allowbreak{}[Search Results — Type: images]\\{}
1. verlag-st-peter.at - Die Kirchen von Hopfgarten im Brixental - Pfarrkirche:  Rundgang Altäre\\{}
~~~Image: [tool\_\allowbreak{}generated\_\allowbreak{}image\_\allowbreak{}75]\\{}
~~~Source: Kirchenführer aus dem Verlag St. Peter   [webpage\_\allowbreak{}url\_\allowbreak{}58]\\{}
~~~Size: 400x600\\{}
2. Barockheiligtum mit hochaltar -Fotos und -Bildmaterial in hoher Auflösung –  Alamy\\{}
~~~Image: [tool\_\allowbreak{}generated\_\allowbreak{}image\_\allowbreak{}76]\\{}
~~~Source: Alamy   [webpage\_\allowbreak{}url\_\allowbreak{}59]\\{}
~~~Size: 1300x956
\end{exverb}
\exCut{8 more results not shown (10 in total).}
\begin{exverb}
~~~Size: 400x600"\allowbreak{},\allowbreak{} "\allowbreak{}error"\allowbreak{}: null\}
\end{exverb}
\begin{exharness}
\exLabel{Harness}{environment update, summarized: new memory entries; new data bindings verbatim}
\begin{exverb}
Memory update: registers 10 new images in memory (tool\_\allowbreak{}generated\_\allowbreak{}image\_\allowbreak{}75 to tool\_\allowbreak{}generated\_\allowbreak{}image\_\allowbreak{}84,\allowbreak{} from Text\_\allowbreak{}Search\_\allowbreak{}Tool); registers 2 new webpage URLs in memory (webpage\_\allowbreak{}url\_\allowbreak{}58 to webpage\_\allowbreak{}url\_\allowbreak{}59,\allowbreak{} from Text\_\allowbreak{}Search\_\allowbreak{}Tool).
\end{exverb}
\end{exharness}
\end{exstep}
\begin{exstep}{12}{\toolZoom}{}
\exWalk{The key step: a tight crop of the altarpiece alone, the one detail unique to this church.}
\exLabel{Arguments}{}
\begin{exverb}
\{\\{}
~~"\allowbreak{}image\_\allowbreak{}id"\allowbreak{}: "\allowbreak{}input\_\allowbreak{}image\_\allowbreak{}1"\allowbreak{},\allowbreak{}\\{}
~~"\allowbreak{}coords"\allowbreak{}: [360,\allowbreak{} 440,\allowbreak{} 590,\allowbreak{} 713],\allowbreak{}\\{}
~~"\allowbreak{}padding"\allowbreak{}: 50\\{}
\}
\end{exverb}
\exLabel{Tool output}{}
\begin{exverb}
Tool \textasciigrave{}Zoom\_\allowbreak{}In\_\allowbreak{}Tool\textasciigrave{} (tool\_\allowbreak{}call\_\allowbreak{}id=\allowbreak{}call\_\allowbreak{}o1GC4x3B6pWmL7FYCQuhlMR8) returned:\\{}
Zoomed in image rendered below; refer to it as tool\_\allowbreak{}generated\_\allowbreak{}image\_\allowbreak{}85.
\end{exverb}
\begin{center}\includegraphics[width=0.45\linewidth,height=0.22\textheight,keepaspectratio]{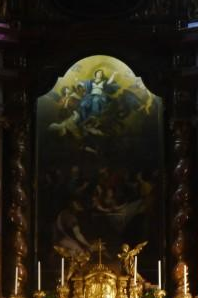}\\[2pt]
{\scriptsize\color{exMute} Crop of the altarpiece alone.}\end{center}
\begin{exharness}
\exLabel{Harness}{environment update, summarized: new memory entries; new data bindings verbatim}
\begin{exverb}
Memory update: registers 1 new image in memory (tool\_\allowbreak{}generated\_\allowbreak{}image\_\allowbreak{}85,\allowbreak{} from Zoom\_\allowbreak{}In\_\allowbreak{}Tool).
\end{exverb}
\end{exharness}
\end{exstep}
\begin{exstep}{13}{\toolIS}{}
\exWalk{Reverse image search on the altarpiece crop. The first result is a Wikimedia Commons photo of the high-altar painting of the parish church in Perchtoldsdorf.}
\exLabel{Arguments}{}
\begin{exverb}
\{\\{}
~~"\allowbreak{}image\_\allowbreak{}id"\allowbreak{}: "\allowbreak{}tool\_\allowbreak{}generated\_\allowbreak{}image\_\allowbreak{}85"\allowbreak{}\\{}
\}
\end{exverb}
\exLabel{Tool output}{}
\begin{exverb}
Tool \textasciigrave{}Image\_\allowbreak{}Search\_\allowbreak{}Tool\textasciigrave{} (tool\_\allowbreak{}call\_\allowbreak{}id=\allowbreak{}call\_\allowbreak{}hPFqkdstj5TH5Bkj2yYmJHFE) returned:\\{}
\{"\allowbreak{}ok"\allowbreak{}: true,\allowbreak{} "\allowbreak{}result"\allowbreak{}: "\allowbreak{}[Image Search — Visual Matches]\\{}
1. File:Perchtoldsdorf Pfarrkirche Hochaltarbild 01.jpg - Wikimedia Commons\\{}
~~~Source: commons.wikimedia.org/\allowbreak{}wiki/\allowbreak{}File:Perchtoldsdorf\_\allowbreak{}Pfarrkirche\_\allowbreak{}Hochaltarbild\_\allowbreak{}01.jpg   [tool\_\allowbreak{}generated\_\allowbreak{}image\_\allowbreak{}86]\\{}
~~~Image: [tool\_\allowbreak{}generated\_\allowbreak{}image\_\allowbreak{}87]\\{}
2. Solum Church - Wikipedia\\{}
~~~Source: en.wikipedia.org/\allowbreak{}wiki/\allowbreak{}Solum\_\allowbreak{}Church   [webpage\_\allowbreak{}url\_\allowbreak{}60]\\{}
~~~Image: [tool\_\allowbreak{}generated\_\allowbreak{}image\_\allowbreak{}88]
\end{exverb}
\exCut{8 more results not shown (10 in total).}
\begin{exverb}
~~~Image: [tool\_\allowbreak{}generated\_\allowbreak{}image\_\allowbreak{}97]"\allowbreak{},\allowbreak{} "\allowbreak{}error"\allowbreak{}: null\}
\end{exverb}
\begin{exharness}
\exLabel{Harness}{environment update, summarized: new memory entries; new data bindings verbatim}
\begin{exverb}
Memory update: registers 12 new images in memory (tool\_\allowbreak{}generated\_\allowbreak{}image\_\allowbreak{}86 to tool\_\allowbreak{}generated\_\allowbreak{}image\_\allowbreak{}97,\allowbreak{} from Image\_\allowbreak{}Search\_\allowbreak{}Tool); registers 8 new webpage URLs in memory (webpage\_\allowbreak{}url\_\allowbreak{}60 to webpage\_\allowbreak{}url\_\allowbreak{}67,\allowbreak{} from Image\_\allowbreak{}Search\_\allowbreak{}Tool).
\end{exverb}
\end{exharness}
\end{exstep}
\begin{exstep}{14}{\toolTS}{}
\exWalk{A text search on that lead confirms it: the parish church of St.\ Augustine in Perchtoldsdorf has a high altar with an Assumption of Mary.}
\exLabel{Arguments}{}
\begin{exverb}
\{\\{}
~~"\allowbreak{}query"\allowbreak{}: "\allowbreak{}Perchtoldsdorf Pfarrkirche Hochaltar Kirche Mariä Himmelfahrt altar"\allowbreak{},\allowbreak{}\\{}
~~"\allowbreak{}search\_\allowbreak{}type"\allowbreak{}: "\allowbreak{}search"\allowbreak{}\\{}
\}
\end{exverb}
\exLabel{Tool output}{}
\begin{exverb}
Tool \textasciigrave{}Text\_\allowbreak{}Search\_\allowbreak{}Tool\textasciigrave{} (tool\_\allowbreak{}call\_\allowbreak{}id=\allowbreak{}call\_\allowbreak{}8aq24tQS1GhdDHIDdzwjZ5Q6) returned:\\{}
\{"\allowbreak{}ok"\allowbreak{}: true,\allowbreak{} "\allowbreak{}result"\allowbreak{}: "\allowbreak{}[Search Results — Type: search]\\{}
1. Pfarrkirche St. Augustinus und Pfarrhof\\{}
~~~Source: perchtoldsdorf.at/\allowbreak{}Pfarrkirche\_\allowbreak{}St\_\allowbreak{}Augustinus\_\allowbreak{}und\_\allowbreak{}Pfarrhof   [webpage\_\allowbreak{}url\_\allowbreak{}68]\\{}
~~~von Perchtoldsdorf im Jahre 1217 zur Pfarrkirche. Der Hochaltar mit tordierten Säulen. Hochaltar mit dem Hauptbild ,,Mariä Himmelfahrt“ und dem hl. Augustinus ...\\{}
2. Pfarrkirche \textbar{} röm.-kath. Pfarre Perchtoldsdorf\\{}
~~~Source: pfarre-perchtoldsdorf.at/\allowbreak{}unsere-pfarre/\allowbreak{}pfarrkirche   [webpage\_\allowbreak{}url\_\allowbreak{}69]\\{}
~~~Neuigkeiten,\allowbreak{} bevorstehende Veranstaltungen und Informationen zu Taufen,\allowbreak{} Trauungen und weiteren kirchlichen Anlässen der röm. kath. Pfarre Perchtoldsdorf.
\end{exverb}
\exCut{8 more results not shown (10 in total).}
\begin{exverb}
~~~Die römisch-katholische Pfarrkirche Perchtoldsdorf steht am Marktplatz in Perchtoldsdorf,\allowbreak{} Bezirk Mödling,\allowbreak{} Niederösterreich,\allowbreak{} im Bereich der ehemaligen Burg."\allowbreak{},\allowbreak{} "\allowbreak{}error"\allowbreak{}: null\}
\end{exverb}
\begin{exharness}
\exLabel{Harness}{environment update, summarized: new memory entries; new data bindings verbatim}
\begin{exverb}
Memory update: registers 9 new webpage URLs in memory (webpage\_\allowbreak{}url\_\allowbreak{}68 to webpage\_\allowbreak{}url\_\allowbreak{}76,\allowbreak{} from Text\_\allowbreak{}Search\_\allowbreak{}Tool).
\end{exverb}
\end{exharness}
\begin{exharness}
\exLabel{Harness}{}
\begin{exverb}
Please provide your best final answer now if enough information is available.
\end{exverb}
\end{exharness}
\end{exstep}
\begin{exstep}{}{\exTool{exInk}{Final answer}}{}
\exWalk{After the harness asks for a final answer, the agent names St.\ Augustine's Parish Church in Perchtoldsdorf, Austria, which the judge accepts.}
\exLabel{Model}{}
\begin{exverb}
\textless{}answer\textgreater{}St. Augustine’s Parish Church in Perchtoldsdorf,\allowbreak{} Austria.\textless{}/\allowbreak{}answer\textgreater{}
\end{exverb}
\end{exstep}

%% file: text/appendix_examples_steps/vstar_header.tex
\exOverview{\textbf{Task.} In a large landscape photograph of a hilltop town, say whether a woman with a backpack is to the left or right of a person wearing a beanie. 

\textbf{Why it is hard.} The two people stand near a bus on a road far below the town and are only a few pixels tall; without tools the model answers ``right''. 

\textbf{How the agent solves it.} Grounding on the full image cannot separate the two people, so the agent zooms in three times, each time closer, and grounds both people again inside the final crop, where they are large enough to tell apart.}
\exLabel{Input images}{}
\begin{center}\includegraphics[width=0.55\linewidth,height=0.28\textheight,keepaspectratio]{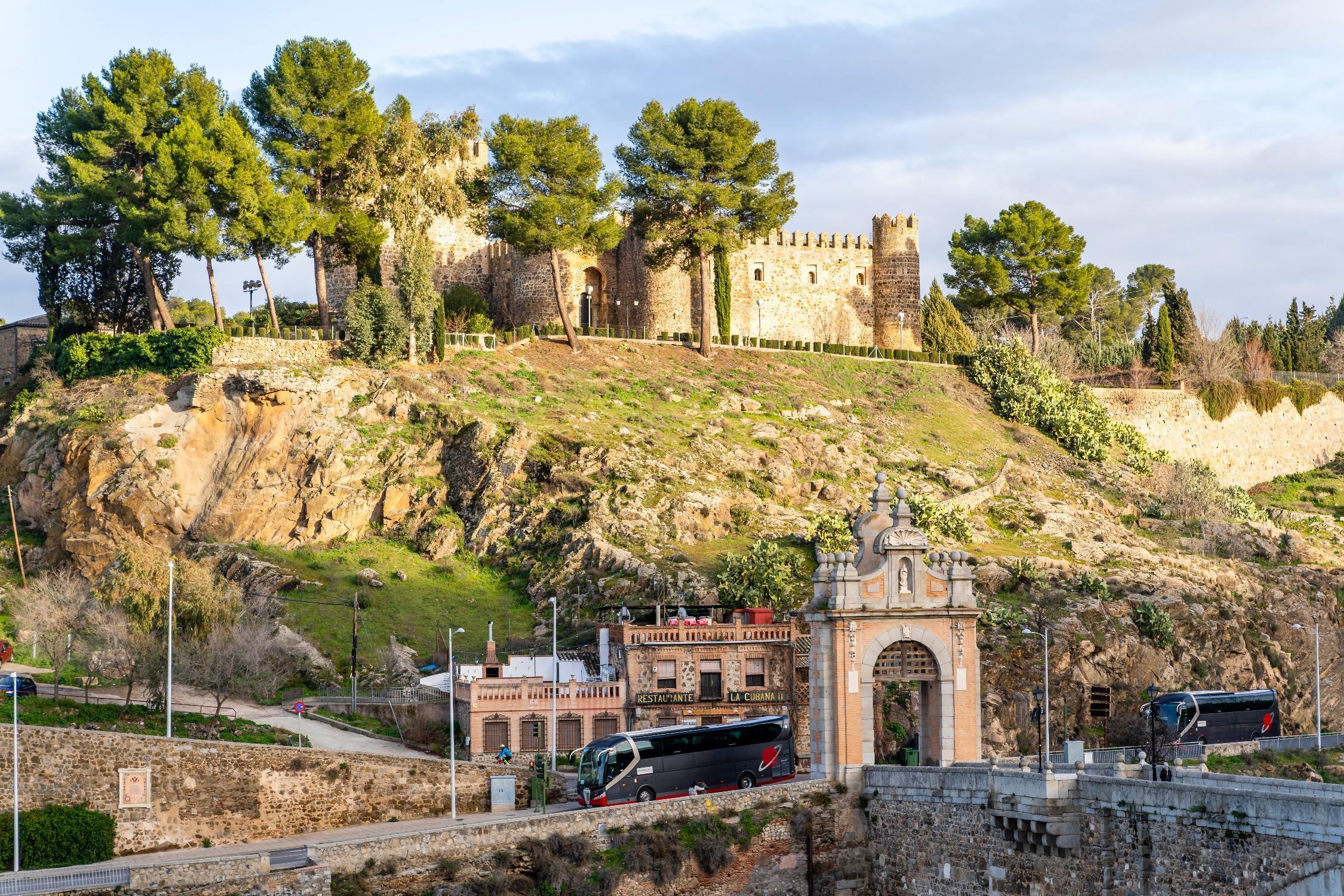}\\[2pt]
{\scriptsize\color{exMute} input\_image\_1}\end{center}
\exLabel{Question}{as sent to the model}
\begin{exverb}
Question: Question: Is the woman with a backpack on the left or right side of the person with a beanie?\\{}
Options:\\{}
A. right\\{}
B. left\\{}
Please select the correct answer from the options above. 
\end{exverb}
\exLabel{Without tools}{GPT-6 Sol's full response to the same question}
\begin{exverb}
A. right
\end{exverb}

%% file: text/appendix_examples_steps/vstar.tex
\begin{exharness}
\exLabel{Harness}{environment update, shown in full}
\begin{exverb}
Environment Update (not a new question):\\{}
Currently available image IDs for tool calls:\\{}
- input\_\allowbreak{}image\_\allowbreak{}1: user\_\allowbreak{}input\\{}
When a tool needs image input,\allowbreak{} pass one of these IDs via the *\_\allowbreak{}image\_\allowbreak{}id field.\\{}
Do NOT invent image file names or paths.
\end{exverb}
\end{exharness}
\begin{exstep}{1}{\toolVG}{}
\exWalk{Grounding ``person wearing beanie'' on the full image returns one tiny box.}
\exLabel{Arguments}{}
\begin{exverb}
\{\\{}
~~"\allowbreak{}query"\allowbreak{}: "\allowbreak{}person wearing beanie"\allowbreak{},\allowbreak{}\\{}
~~"\allowbreak{}image\_\allowbreak{}id"\allowbreak{}: "\allowbreak{}input\_\allowbreak{}image\_\allowbreak{}1"\allowbreak{}\\{}
\}
\end{exverb}
\exLabel{Tool output}{}
\begin{exverb}
Visual\_\allowbreak{}Grounding\_\allowbreak{}Tool grounded 1 objects for the prompt \textquotesingle{}person wearing beanie\textquotesingle{}. Their corresponding bounding boxes (on a 0-1000 scale,\allowbreak{} [x1,\allowbreak{} y1,\allowbreak{} x2,\allowbreak{} y2]) are: \textasciigrave{}\textasciigrave{}\textasciigrave{}json\\{}
[\\{}
~~~~\{"\allowbreak{}bbox\_\allowbreak{}2d"\allowbreak{}: [369,\allowbreak{} 830,\allowbreak{} 381,\allowbreak{} 853],\allowbreak{} "\allowbreak{}label"\allowbreak{}: "\allowbreak{}person wearing beanie"\allowbreak{}\}\\{}
]\textasciigrave{}\textasciigrave{}\textasciigrave{}. The object numbers correspond to the labeled masks rendered in the following image,\allowbreak{} which you can refer to as tool\_\allowbreak{}generated\_\allowbreak{}image\_\allowbreak{}1.
\end{exverb}
\begin{center}\includegraphics[width=0.45\linewidth,height=0.22\textheight,keepaspectratio]{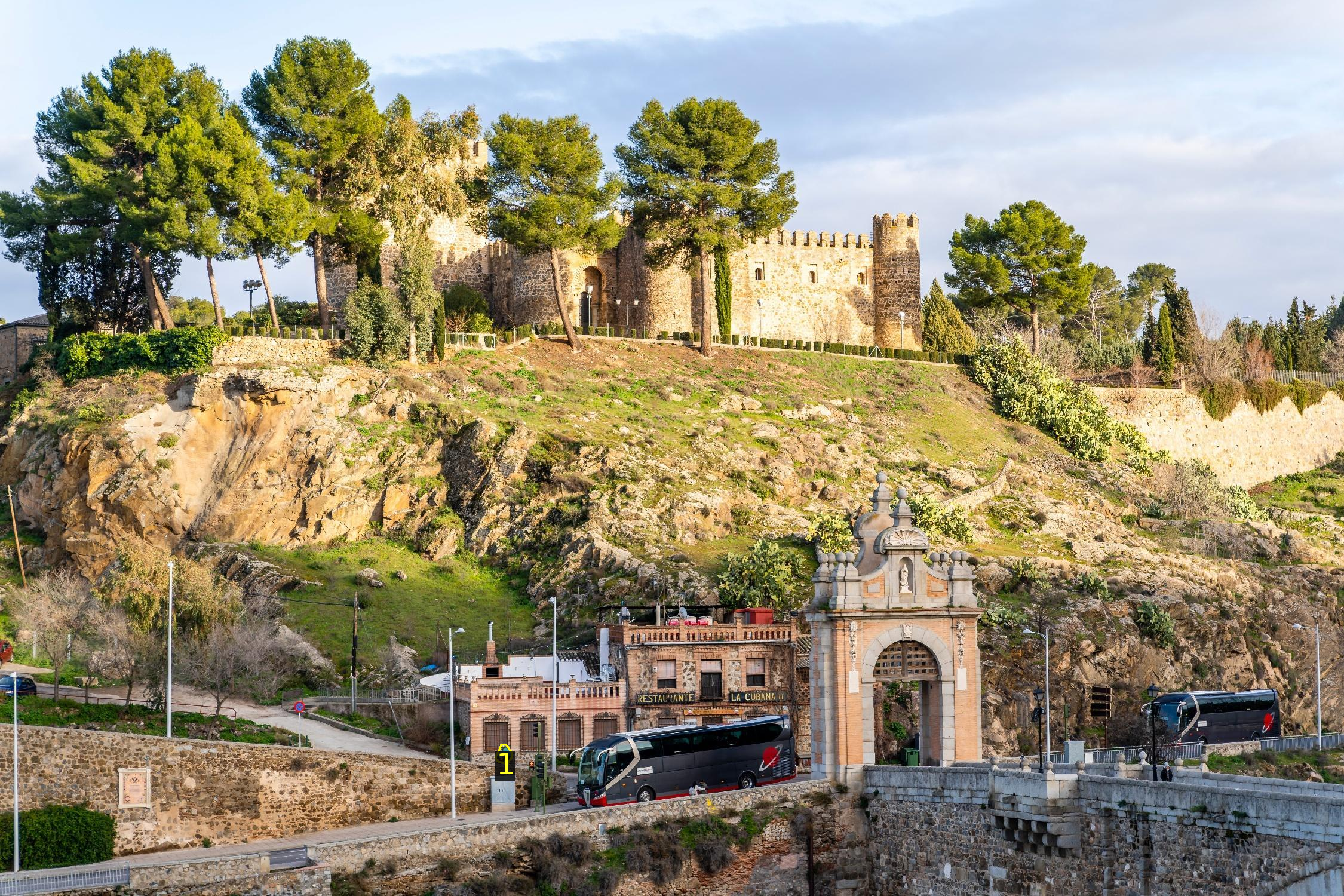}\\[2pt]
{\scriptsize\color{exMute} Image returned with this output.}\end{center}
\begin{exharness}
\exLabel{Harness}{environment update, summarized: new memory entries; new data bindings verbatim}
\begin{exverb}
Memory update: registers 1 new image in memory (tool\_\allowbreak{}generated\_\allowbreak{}image\_\allowbreak{}1,\allowbreak{} from Visual\_\allowbreak{}Grounding\_\allowbreak{}Tool).\\{}
Available data bindings (auto-injected into Python\_\allowbreak{}Coding\_\allowbreak{}Agent\_\allowbreak{}Tool):\\{}
- segment\_\allowbreak{}person\_\allowbreak{}wearing\_\allowbreak{}beanie: dict (keys: pred\_\allowbreak{}boxes,\allowbreak{} pred\_\allowbreak{}scores,\allowbreak{} pred\_\allowbreak{}masks,\allowbreak{} num\_\allowbreak{}masks,\allowbreak{} orig\_\allowbreak{}img\_\allowbreak{}h,\allowbreak{} orig\_\allowbreak{}img\_\allowbreak{}w)
\end{exverb}
\end{exharness}
\end{exstep}
\begin{exstep}{2}{\toolVG}{}
\exWalk{Grounding ``woman with backpack'' returns almost the same tiny box, so at this scale the two people cannot be told apart.}
\exLabel{Arguments}{}
\begin{exverb}
\{\\{}
~~"\allowbreak{}query"\allowbreak{}: "\allowbreak{}woman with backpack"\allowbreak{},\allowbreak{}\\{}
~~"\allowbreak{}image\_\allowbreak{}id"\allowbreak{}: "\allowbreak{}input\_\allowbreak{}image\_\allowbreak{}1"\allowbreak{}\\{}
\}
\end{exverb}
\exLabel{Tool output}{}
\begin{exverb}
Visual\_\allowbreak{}Grounding\_\allowbreak{}Tool grounded 1 objects for the prompt \textquotesingle{}woman with backpack\textquotesingle{}. Their corresponding bounding boxes (on a 0-1000 scale,\allowbreak{} [x1,\allowbreak{} y1,\allowbreak{} x2,\allowbreak{} y2]) are: \textasciigrave{}\textasciigrave{}\textasciigrave{}json\\{}
[\\{}
~~~~\{"\allowbreak{}bbox\_\allowbreak{}2d"\allowbreak{}: [368,\allowbreak{} 831,\allowbreak{} 381,\allowbreak{} 853],\allowbreak{} "\allowbreak{}label"\allowbreak{}: "\allowbreak{}woman with backpack"\allowbreak{}\}\\{}
]\textasciigrave{}\textasciigrave{}\textasciigrave{}. The object numbers correspond to the labeled masks rendered in the following image,\allowbreak{} which you can refer to as tool\_\allowbreak{}generated\_\allowbreak{}image\_\allowbreak{}2.
\end{exverb}
\begin{center}\includegraphics[width=0.45\linewidth,height=0.22\textheight,keepaspectratio]{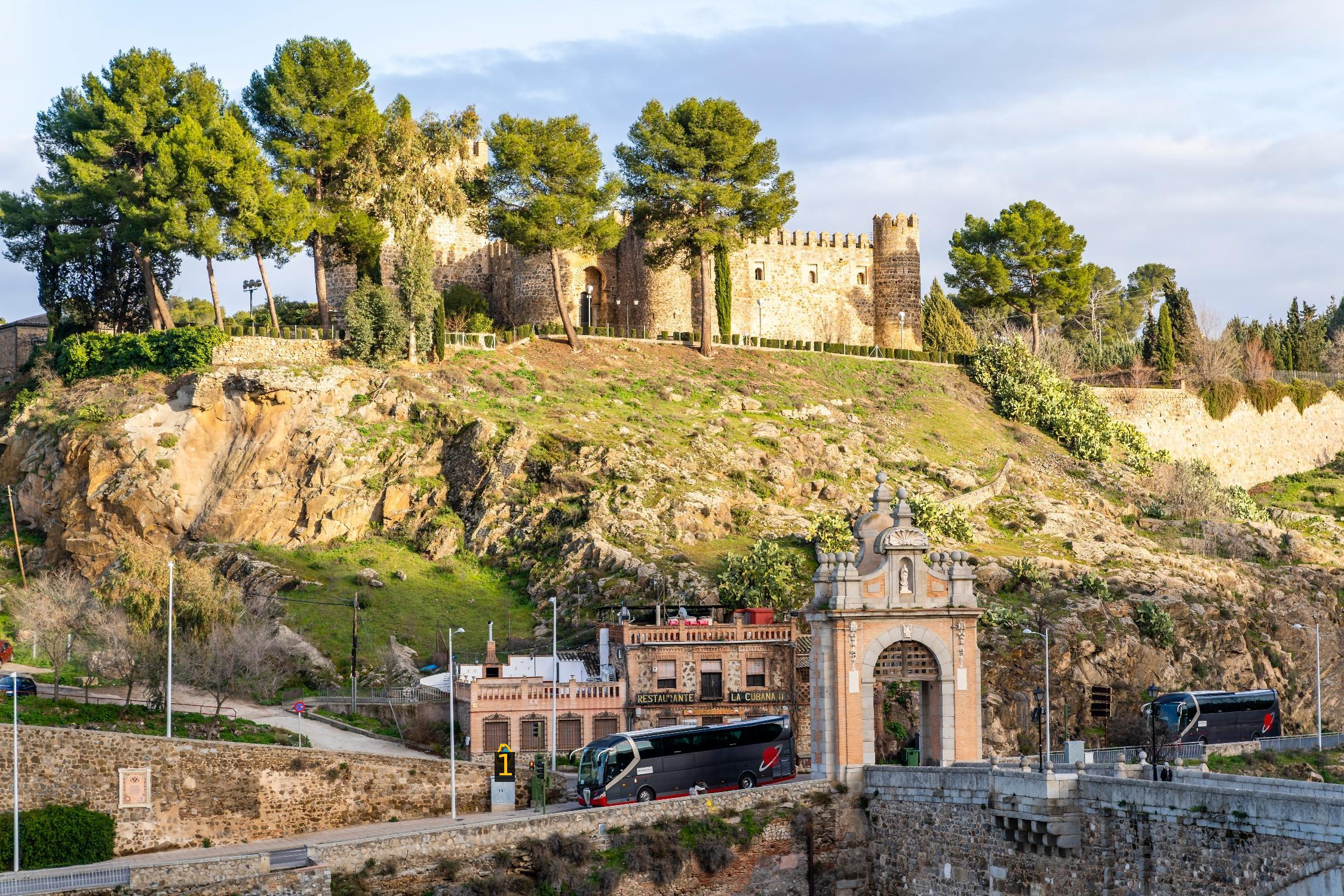}\\[2pt]
{\scriptsize\color{exMute} Image returned with this output.}\end{center}
\begin{exharness}
\exLabel{Harness}{environment update, summarized: new memory entries; new data bindings verbatim}
\begin{exverb}
Memory update: registers 1 new image in memory (tool\_\allowbreak{}generated\_\allowbreak{}image\_\allowbreak{}2,\allowbreak{} from Visual\_\allowbreak{}Grounding\_\allowbreak{}Tool).\\{}
- segment\_\allowbreak{}woman\_\allowbreak{}with\_\allowbreak{}backpack: dict (keys: pred\_\allowbreak{}boxes,\allowbreak{} pred\_\allowbreak{}scores,\allowbreak{} pred\_\allowbreak{}masks,\allowbreak{} num\_\allowbreak{}masks,\allowbreak{} orig\_\allowbreak{}img\_\allowbreak{}h,\allowbreak{} orig\_\allowbreak{}img\_\allowbreak{}w)
\end{exverb}
\end{exharness}
\end{exstep}
\begin{exstep}{3}{\toolZoom}{}
\exWalk{A first zoom on the region around that box.}
\exLabel{Arguments}{}
\begin{exverb}
\{\\{}
~~"\allowbreak{}image\_\allowbreak{}id"\allowbreak{}: "\allowbreak{}input\_\allowbreak{}image\_\allowbreak{}1"\allowbreak{},\allowbreak{}\\{}
~~"\allowbreak{}coords"\allowbreak{}: [350,\allowbreak{} 810,\allowbreak{} 410,\allowbreak{} 875],\allowbreak{}\\{}
~~"\allowbreak{}padding"\allowbreak{}: 100\\{}
\}
\end{exverb}
\exLabel{Tool output}{}
\begin{exverb}
Tool \textasciigrave{}Zoom\_\allowbreak{}In\_\allowbreak{}Tool\textasciigrave{} (tool\_\allowbreak{}call\_\allowbreak{}id=\allowbreak{}call\_\allowbreak{}48KspHqeWAyb0sVqvUkH75he) returned:\\{}
Zoomed in image rendered below; refer to it as tool\_\allowbreak{}generated\_\allowbreak{}image\_\allowbreak{}3.
\end{exverb}
\begin{center}\includegraphics[width=0.45\linewidth,height=0.22\textheight,keepaspectratio]{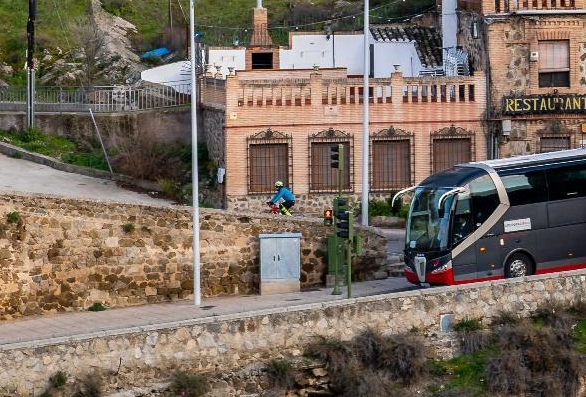}\\[2pt]
{\scriptsize\color{exMute} Image returned with this output.}\end{center}
\begin{exharness}
\exLabel{Harness}{environment update, summarized: new memory entries; new data bindings verbatim}
\begin{exverb}
Memory update: registers 1 new image in memory (tool\_\allowbreak{}generated\_\allowbreak{}image\_\allowbreak{}3,\allowbreak{} from Zoom\_\allowbreak{}In\_\allowbreak{}Tool).
\end{exverb}
\end{exharness}
\end{exstep}
\begin{exstep}{4}{\toolZoom}{}
\exWalk{A second zoom on a nearby region below the bus.}
\exLabel{Arguments}{}
\begin{exverb}
\{\\{}
~~"\allowbreak{}image\_\allowbreak{}id"\allowbreak{}: "\allowbreak{}input\_\allowbreak{}image\_\allowbreak{}1"\allowbreak{},\allowbreak{}\\{}
~~"\allowbreak{}coords"\allowbreak{}: [500,\allowbreak{} 810,\allowbreak{} 605,\allowbreak{} 920],\allowbreak{}\\{}
~~"\allowbreak{}padding"\allowbreak{}: 100\\{}
\}
\end{exverb}
\exLabel{Tool output}{}
\begin{exverb}
Tool \textasciigrave{}Zoom\_\allowbreak{}In\_\allowbreak{}Tool\textasciigrave{} (tool\_\allowbreak{}call\_\allowbreak{}id=\allowbreak{}call\_\allowbreak{}K9cLu6IVuAX39EqWkYJa6WuQ) returned:\\{}
Zoomed in image rendered below; refer to it as tool\_\allowbreak{}generated\_\allowbreak{}image\_\allowbreak{}4.
\end{exverb}
\begin{center}\includegraphics[width=0.45\linewidth,height=0.22\textheight,keepaspectratio]{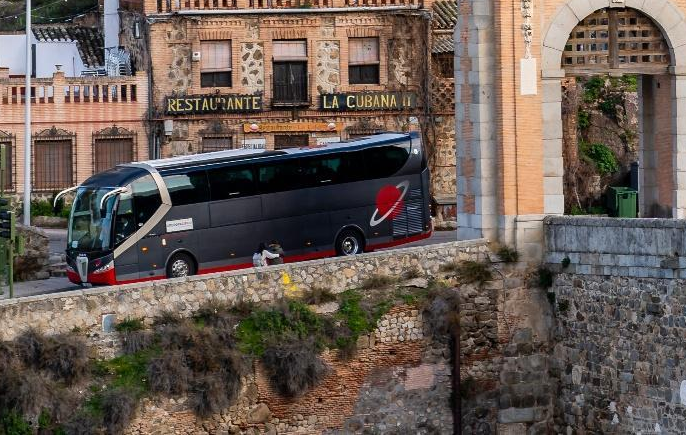}\\[2pt]
{\scriptsize\color{exMute} Image returned with this output.}\end{center}
\begin{exharness}
\exLabel{Harness}{environment update, summarized: new memory entries; new data bindings verbatim}
\begin{exverb}
Memory update: registers 1 new image in memory (tool\_\allowbreak{}generated\_\allowbreak{}image\_\allowbreak{}4,\allowbreak{} from Zoom\_\allowbreak{}In\_\allowbreak{}Tool).
\end{exverb}
\end{exharness}
\end{exstep}
\begin{exstep}{5}{\toolZoom}{}
\exWalk{A third, much tighter zoom on the two figures. This crop is the evidence for the answer.}
\exLabel{Arguments}{}
\begin{exverb}
\{\\{}
~~"\allowbreak{}image\_\allowbreak{}id"\allowbreak{}: "\allowbreak{}input\_\allowbreak{}image\_\allowbreak{}1"\allowbreak{},\allowbreak{}\\{}
~~"\allowbreak{}coords"\allowbreak{}: [508,\allowbreak{} 862,\allowbreak{} 538,\allowbreak{} 895],\allowbreak{}\\{}
~~"\allowbreak{}padding"\allowbreak{}: 50\\{}
\}
\end{exverb}
\exLabel{Tool output}{}
\begin{exverb}
Tool \textasciigrave{}Zoom\_\allowbreak{}In\_\allowbreak{}Tool\textasciigrave{} (tool\_\allowbreak{}call\_\allowbreak{}id=\allowbreak{}call\_\allowbreak{}u8svfarDLYuRGK81UTyQojba) returned:\\{}
Zoomed in image rendered below; refer to it as tool\_\allowbreak{}generated\_\allowbreak{}image\_\allowbreak{}5.
\end{exverb}
\begin{center}\includegraphics[width=0.45\linewidth,height=0.22\textheight,keepaspectratio]{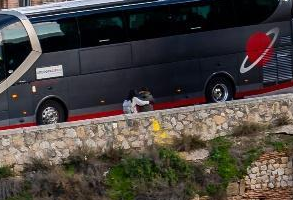}\\[2pt]
{\scriptsize\color{exMute} Image returned with this output.}\end{center}
\begin{exharness}
\exLabel{Harness}{environment update, summarized: new memory entries; new data bindings verbatim}
\begin{exverb}
Memory update: registers 1 new image in memory (tool\_\allowbreak{}generated\_\allowbreak{}image\_\allowbreak{}5,\allowbreak{} from Zoom\_\allowbreak{}In\_\allowbreak{}Tool).
\end{exverb}
\end{exharness}
\end{exstep}
\begin{exstep}{6}{\toolVG}{}
\exWalk{Grounding ``woman with backpack'' inside the crop now returns a clear box for her.}
\exLabel{Arguments}{}
\begin{exverb}
\{\\{}
~~"\allowbreak{}query"\allowbreak{}: "\allowbreak{}woman with backpack"\allowbreak{},\allowbreak{}\\{}
~~"\allowbreak{}image\_\allowbreak{}id"\allowbreak{}: "\allowbreak{}tool\_\allowbreak{}generated\_\allowbreak{}image\_\allowbreak{}5"\allowbreak{}\\{}
\}
\end{exverb}
\exLabel{Tool output}{}
\begin{exverb}
Visual\_\allowbreak{}Grounding\_\allowbreak{}Tool grounded 1 objects for the prompt \textquotesingle{}woman with backpack\textquotesingle{}. Their corresponding bounding boxes (on a 0-1000 scale,\allowbreak{} [x1,\allowbreak{} y1,\allowbreak{} x2,\allowbreak{} y2]) are: \textasciigrave{}\textasciigrave{}\textasciigrave{}json\\{}
[\\{}
~~~~\{"\allowbreak{}bbox\_\allowbreak{}2d"\allowbreak{}: [419,\allowbreak{} 447,\allowbreak{} 511,\allowbreak{} 569],\allowbreak{} "\allowbreak{}label"\allowbreak{}: "\allowbreak{}woman with backpack"\allowbreak{}\}\\{}
]\textasciigrave{}\textasciigrave{}\textasciigrave{}. The object numbers correspond to the labeled masks rendered in the following image,\allowbreak{} which you can refer to as tool\_\allowbreak{}generated\_\allowbreak{}image\_\allowbreak{}6.
\end{exverb}
\begin{center}\includegraphics[width=0.45\linewidth,height=0.22\textheight,keepaspectratio]{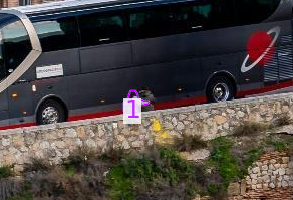}\\[2pt]
{\scriptsize\color{exMute} Image returned with this output.}\end{center}
\begin{exharness}
\exLabel{Harness}{environment update, summarized: new memory entries; new data bindings verbatim}
\begin{exverb}
Memory update: registers 1 new image in memory (tool\_\allowbreak{}generated\_\allowbreak{}image\_\allowbreak{}6,\allowbreak{} from Visual\_\allowbreak{}Grounding\_\allowbreak{}Tool).\\{}
- segment\_\allowbreak{}woman\_\allowbreak{}with\_\allowbreak{}backpack\_\allowbreak{}2: dict (keys: pred\_\allowbreak{}boxes,\allowbreak{} pred\_\allowbreak{}scores,\allowbreak{} pred\_\allowbreak{}masks,\allowbreak{} num\_\allowbreak{}masks,\allowbreak{} orig\_\allowbreak{}img\_\allowbreak{}h,\allowbreak{} orig\_\allowbreak{}img\_\allowbreak{}w)
\end{exverb}
\end{exharness}
\end{exstep}
\begin{exstep}{7}{\toolVG}{}
\exWalk{Grounding ``person with beanie'' in the crop returns two boxes: the woman's own box again, and a second person to her right.}
\exLabel{Arguments}{}
\begin{exverb}
\{\\{}
~~"\allowbreak{}query"\allowbreak{}: "\allowbreak{}person with beanie"\allowbreak{},\allowbreak{}\\{}
~~"\allowbreak{}image\_\allowbreak{}id"\allowbreak{}: "\allowbreak{}tool\_\allowbreak{}generated\_\allowbreak{}image\_\allowbreak{}5"\allowbreak{}\\{}
\}
\end{exverb}
\exLabel{Tool output}{}
\begin{exverb}
Visual\_\allowbreak{}Grounding\_\allowbreak{}Tool grounded 2 objects for the prompt \textquotesingle{}person with beanie\textquotesingle{}. Their corresponding bounding boxes (on a 0-1000 scale,\allowbreak{} [x1,\allowbreak{} y1,\allowbreak{} x2,\allowbreak{} y2]) are: \textasciigrave{}\textasciigrave{}\textasciigrave{}json\\{}
[\\{}
~~~~\{"\allowbreak{}bbox\_\allowbreak{}2d"\allowbreak{}: [419,\allowbreak{} 446,\allowbreak{} 511,\allowbreak{} 569],\allowbreak{} "\allowbreak{}label"\allowbreak{}: "\allowbreak{}person with beanie"\allowbreak{}\},\allowbreak{}\\{}
~~~~\{"\allowbreak{}bbox\_\allowbreak{}2d"\allowbreak{}: [472,\allowbreak{} 435,\allowbreak{} 527,\allowbreak{} 560],\allowbreak{} "\allowbreak{}label"\allowbreak{}: "\allowbreak{}person with beanie"\allowbreak{}\}\\{}
]\textasciigrave{}\textasciigrave{}\textasciigrave{}. The object numbers correspond to the labeled masks rendered in the following image,\allowbreak{} which you can refer to as tool\_\allowbreak{}generated\_\allowbreak{}image\_\allowbreak{}7.
\end{exverb}
\begin{center}\includegraphics[width=0.45\linewidth,height=0.22\textheight,keepaspectratio]{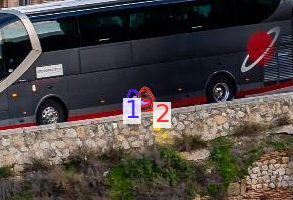}\\[2pt]
{\scriptsize\color{exMute} Image returned with this output.}\end{center}
\begin{exharness}
\exLabel{Harness}{environment update, summarized: new memory entries; new data bindings verbatim}
\begin{exverb}
Memory update: registers 1 new image in memory (tool\_\allowbreak{}generated\_\allowbreak{}image\_\allowbreak{}7,\allowbreak{} from Visual\_\allowbreak{}Grounding\_\allowbreak{}Tool).\\{}
- segment\_\allowbreak{}person\_\allowbreak{}with\_\allowbreak{}beanie: dict (keys: pred\_\allowbreak{}boxes,\allowbreak{} pred\_\allowbreak{}scores,\allowbreak{} pred\_\allowbreak{}masks,\allowbreak{} num\_\allowbreak{}masks,\allowbreak{} orig\_\allowbreak{}img\_\allowbreak{}h,\allowbreak{} orig\_\allowbreak{}img\_\allowbreak{}w)
\end{exverb}
\end{exharness}
\end{exstep}
\begin{exstep}{}{\exTool{exInk}{Final answer}}{}
\exWalk{The agent concludes that the woman is on the left of the person with the beanie, option B. The beanie query also returned the woman's own box, so this last step still relies on the model's reading of the crop.}
\exLabel{Model}{}
\begin{exverb}
\textless{}answer\textgreater{}B. left\textless{}/\allowbreak{}answer\textgreater{}
\end{exverb}
\end{exstep}

%% file: text/appendix_examples_steps/mmsi_header.tex
\exOverview{\textbf{Task.} Seven photographs show one living room from different positions. Of four statements about the room, choose the correct one; two of them give the number of throw pillows, 16 or 12, counting partly covered ones. 

\textbf{Why it is hard.} The pillows are spread across several views, some appear in more than one view, and some are partly hidden; without tools the model picks 12. 

\textbf{How the agent solves it.} It grounds ``throw pillows'' in each of the four views that show seating, zooms on both halves of the main sofa, and chooses the statement with 16 pillows.}
\exLabel{Input images}{}
\begin{center}
\begin{minipage}[t]{0.23\linewidth}\centering\includegraphics[width=\linewidth,height=0.22\textheight,keepaspectratio]{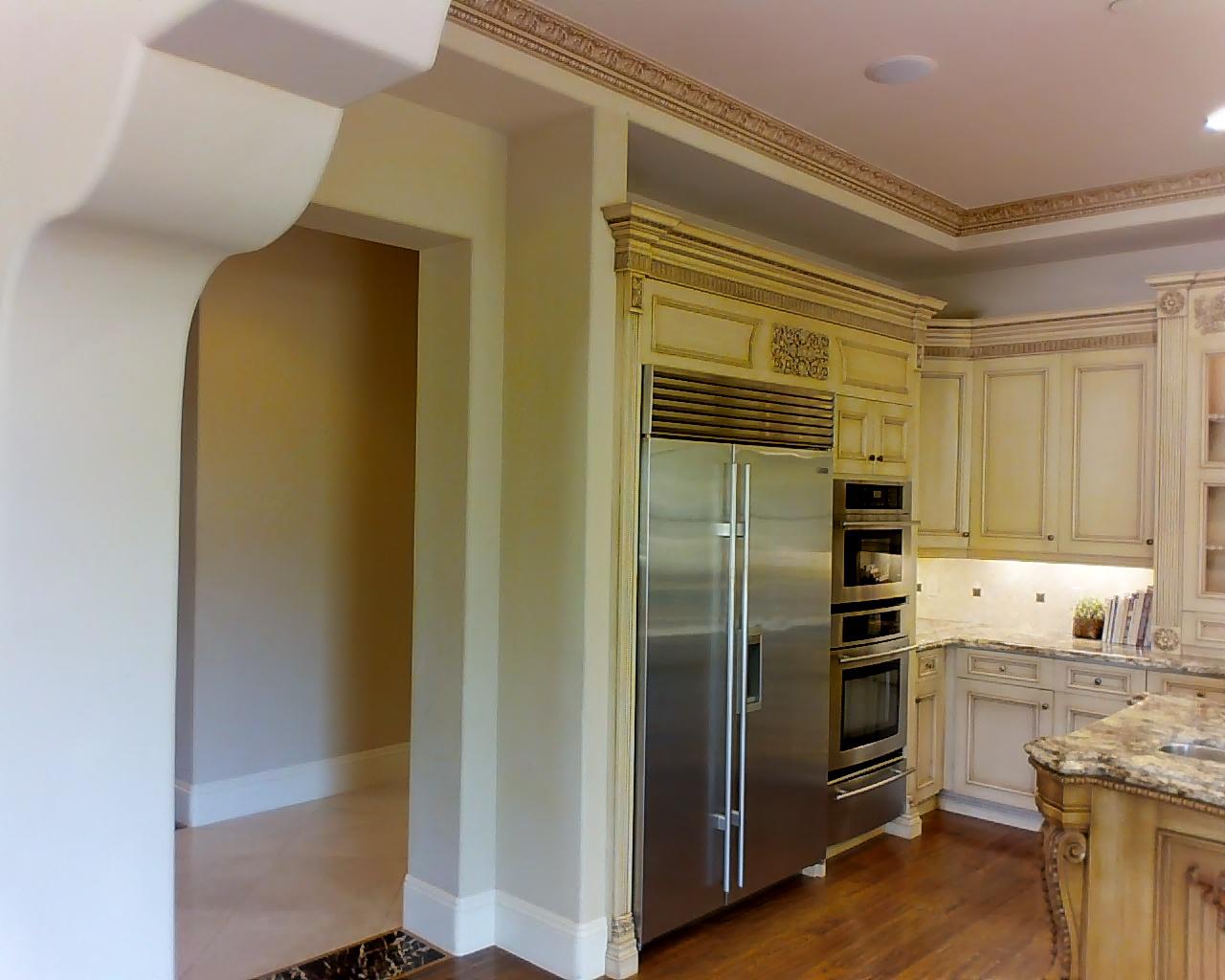}\\[1pt]{\scriptsize\color{exMute} input\_image\_1}\end{minipage}\hfill\begin{minipage}[t]{0.23\linewidth}\centering\includegraphics[width=\linewidth,height=0.22\textheight,keepaspectratio]{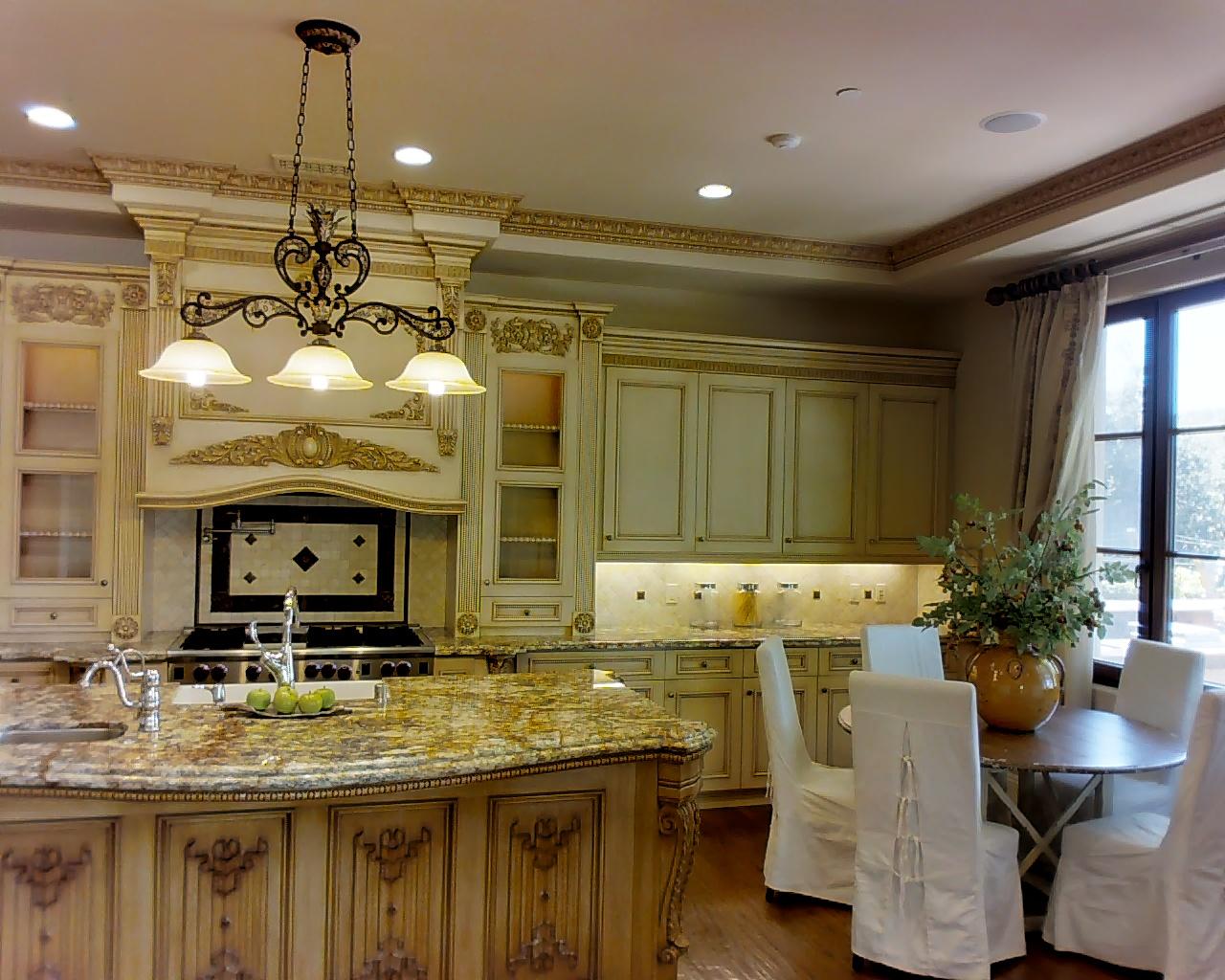}\\[1pt]{\scriptsize\color{exMute} input\_image\_2}\end{minipage}\hfill\begin{minipage}[t]{0.23\linewidth}\centering\includegraphics[width=\linewidth,height=0.22\textheight,keepaspectratio]{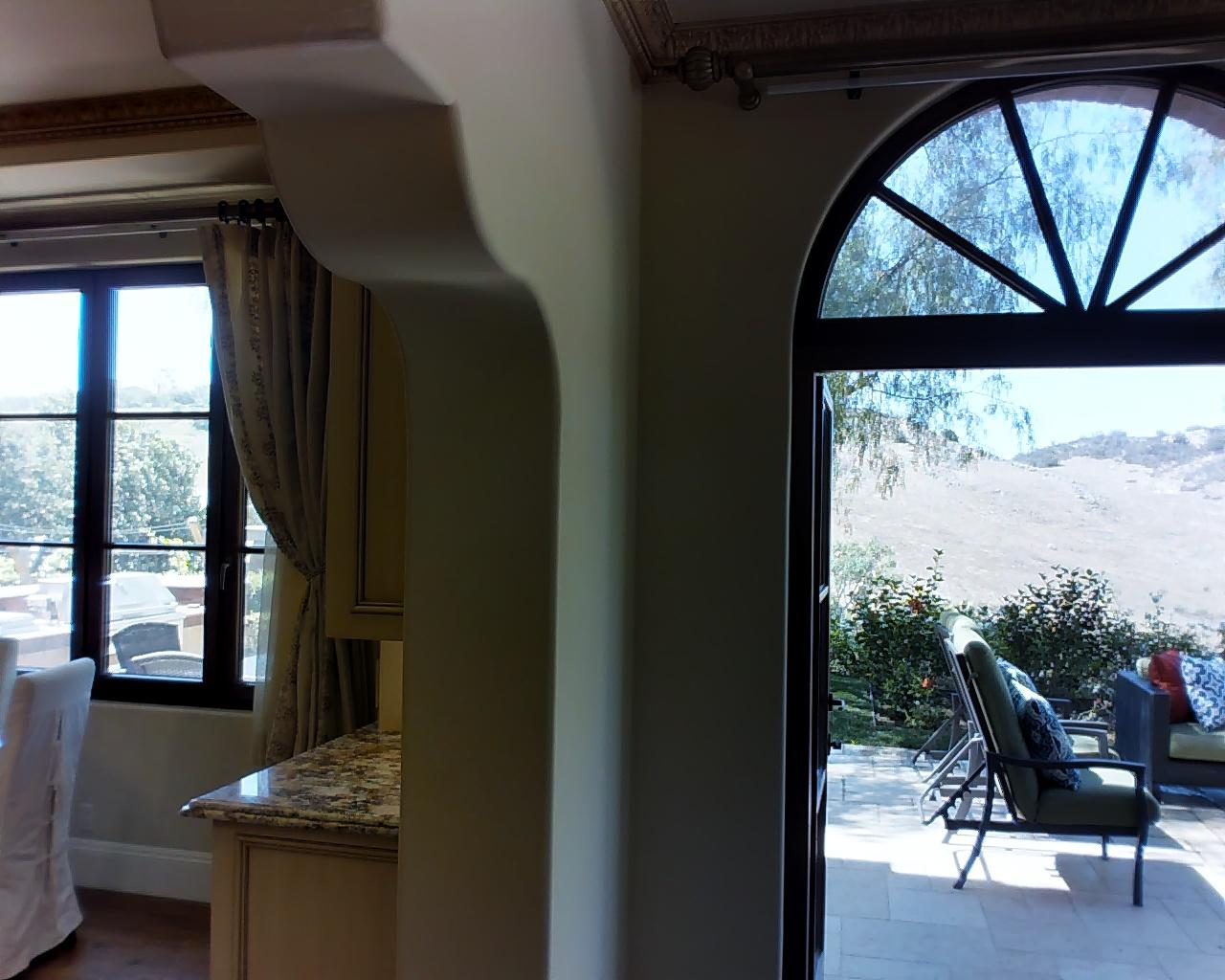}\\[1pt]{\scriptsize\color{exMute} input\_image\_3}\end{minipage}\hfill\begin{minipage}[t]{0.23\linewidth}\centering\includegraphics[width=\linewidth,height=0.22\textheight,keepaspectratio]{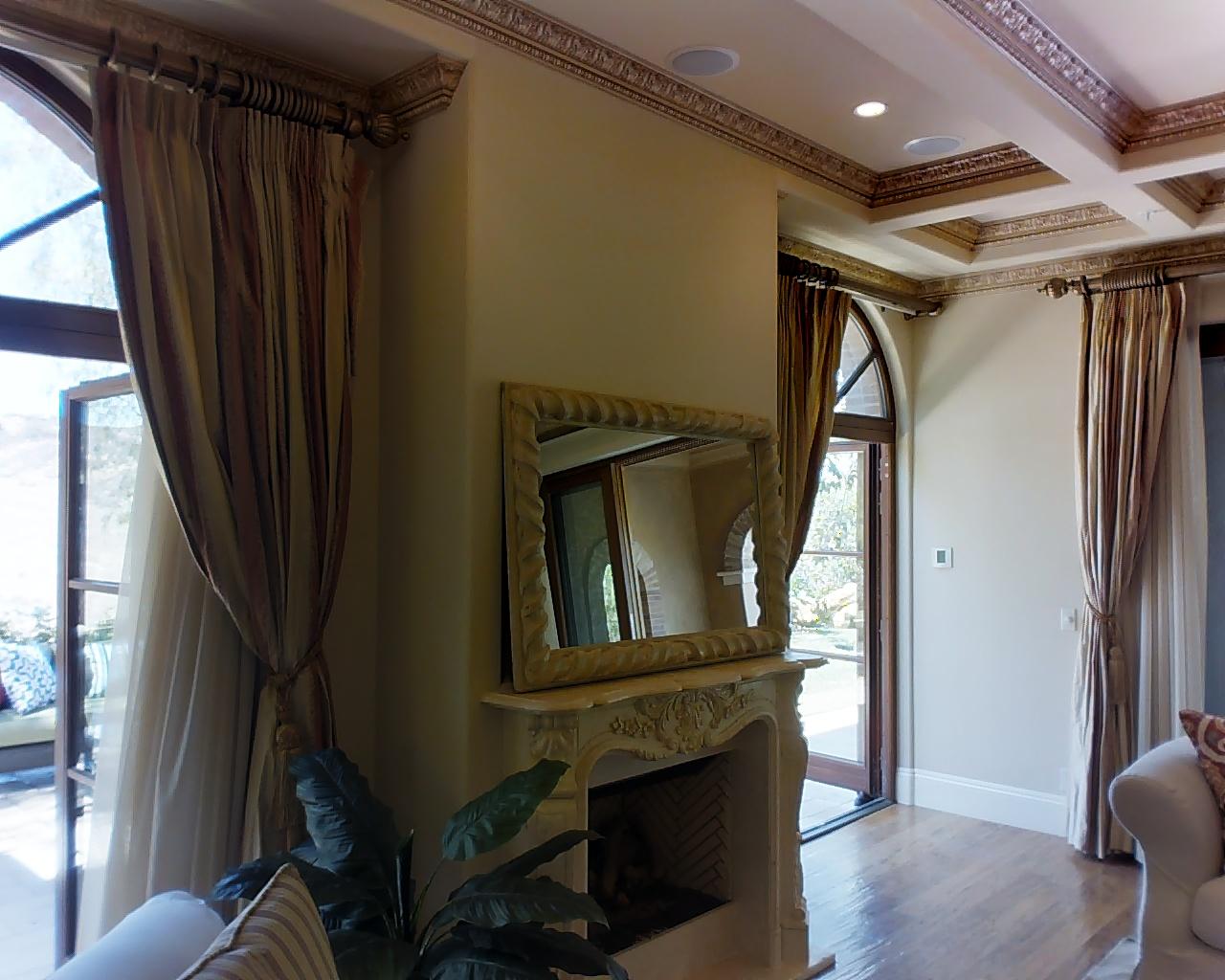}\\[1pt]{\scriptsize\color{exMute} input\_image\_4}\end{minipage}\\[4pt]
\begin{minipage}[t]{0.312\linewidth}\centering\includegraphics[width=\linewidth,height=0.22\textheight,keepaspectratio]{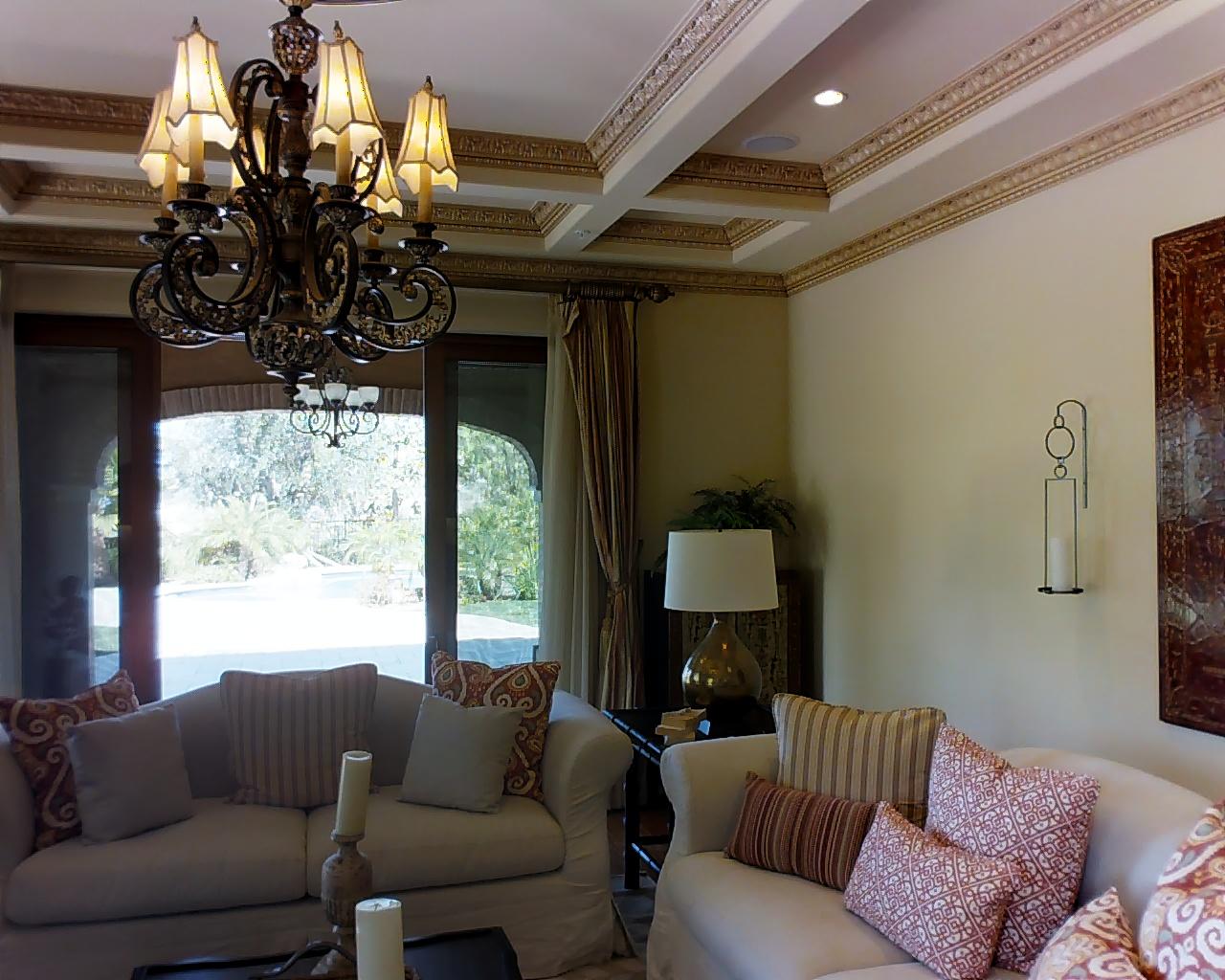}\\[1pt]{\scriptsize\color{exMute} input\_image\_5}\end{minipage}\hfill\begin{minipage}[t]{0.312\linewidth}\centering\includegraphics[width=\linewidth,height=0.22\textheight,keepaspectratio]{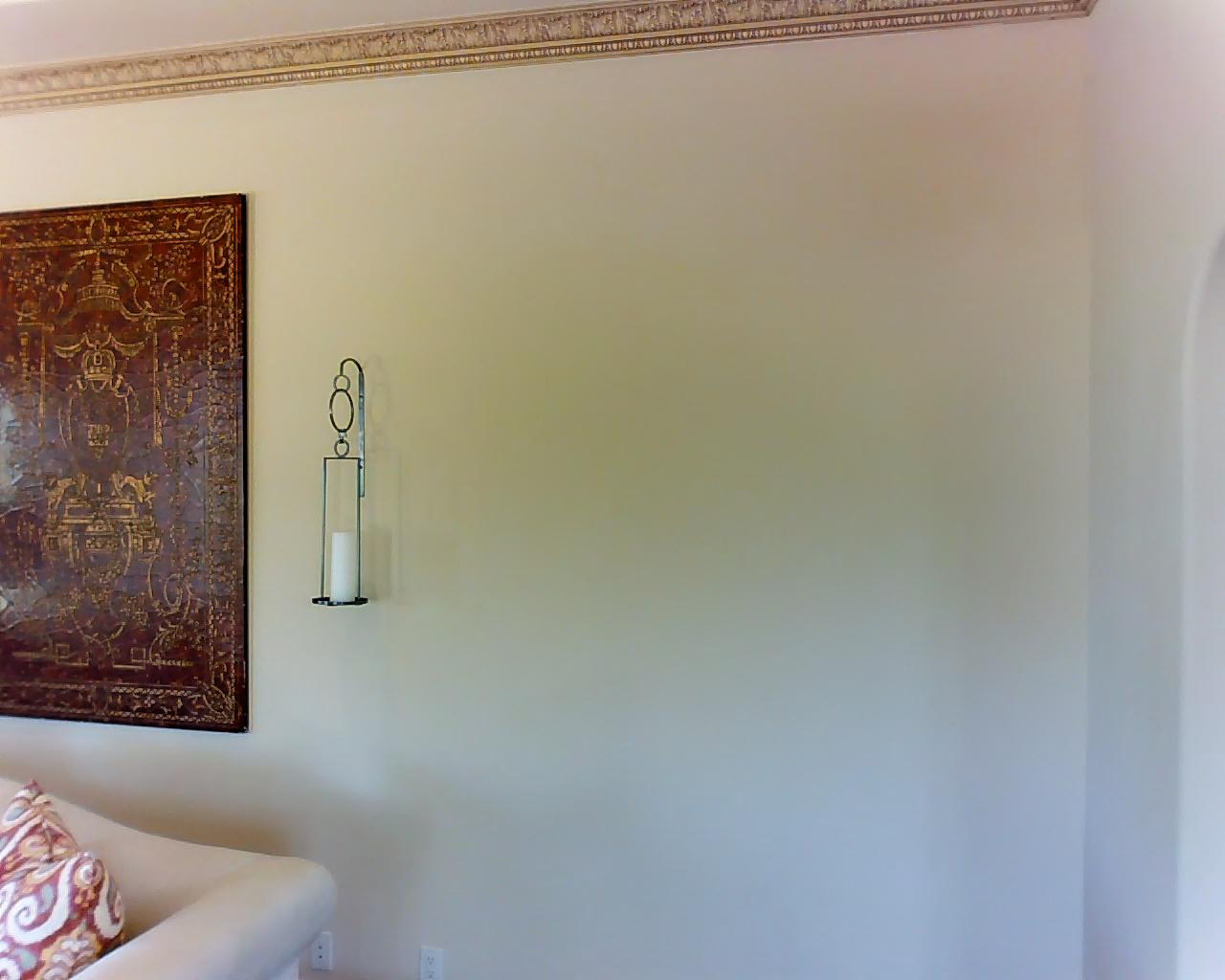}\\[1pt]{\scriptsize\color{exMute} input\_image\_6}\end{minipage}\hfill\begin{minipage}[t]{0.312\linewidth}\centering\includegraphics[width=\linewidth,height=0.22\textheight,keepaspectratio]{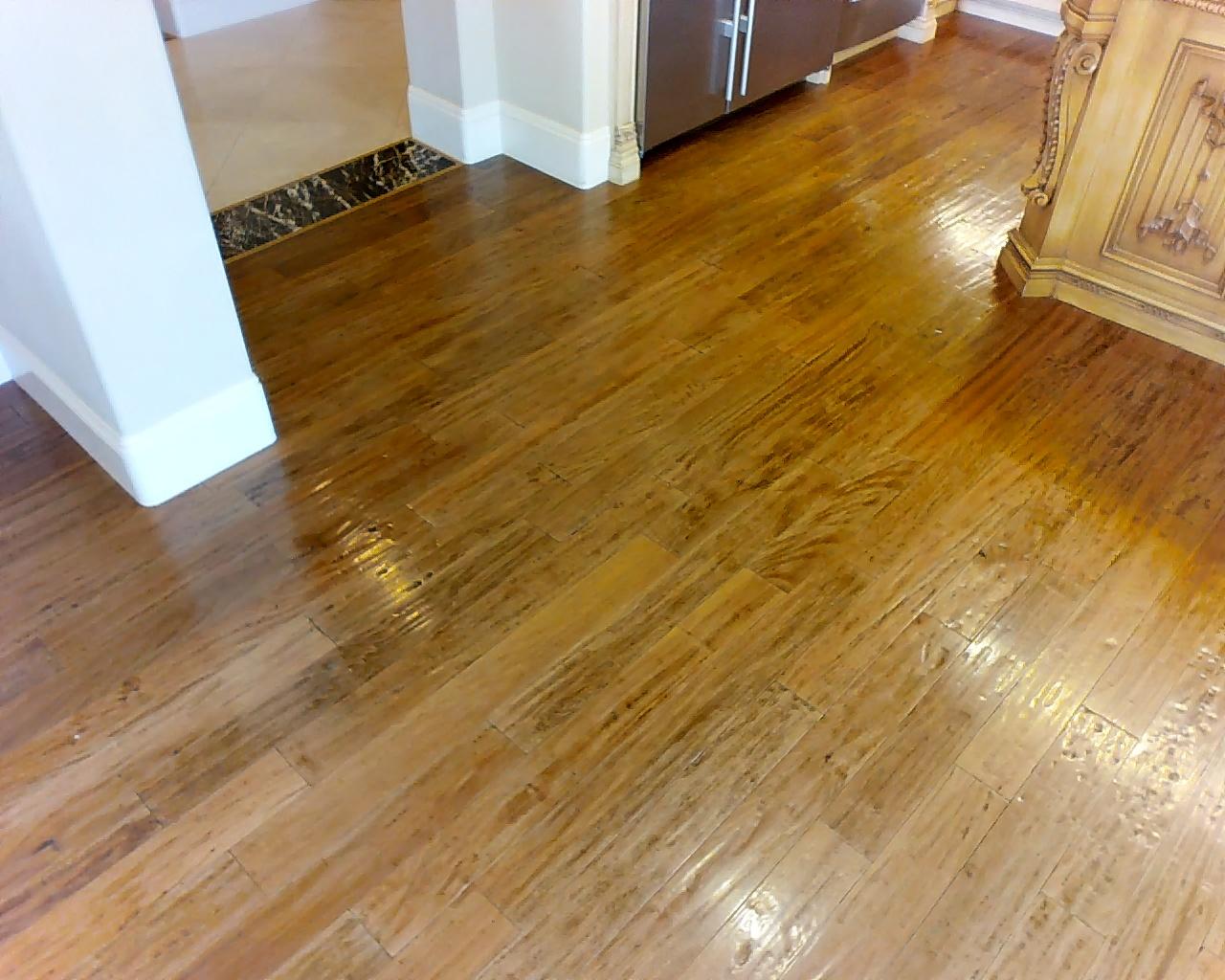}\\[1pt]{\scriptsize\color{exMute} input\_image\_7}\end{minipage}
\end{center}
\exLabel{Question}{as sent to the model}
\begin{exverb}
Question: The kitchen is located to the north relative to the sofa. Which of the following statements is correct?\\{}
Options: A: The mirror is to the southwest of the refrigerator,\allowbreak{} B: The radius of the vase in Figure 2 is larger than the radius of the white lampshade of the table lamp in Figure 5,\allowbreak{} C: There are a total of 16 throw pillows in the picture (including those partially covered by other objects),\allowbreak{} D: There are a total of 12 throw pillows in the picture (including those partially covered by other objects),\allowbreak{} \\{}
Answer with the option\textquotesingle{}s letter from the given choices directly. Enclose the option\textquotesingle{}s letter within \textasciigrave{}\textasciigrave{}.
\end{exverb}
\exLabel{Without tools}{GPT-6 Sol's full response to the same question}
\begin{exverb}
\textasciigrave{}D\textasciigrave{}
\end{exverb}

%% file: text/appendix_examples_steps/mmsi.tex
\begin{exharness}
\exLabel{Harness}{environment update, shown in full}
\begin{exverb}
Environment Update (not a new question):\\{}
Currently available image IDs for tool calls:\\{}
- input\_\allowbreak{}image\_\allowbreak{}1: user\_\allowbreak{}input\\{}
- input\_\allowbreak{}image\_\allowbreak{}2: user\_\allowbreak{}input\\{}
- input\_\allowbreak{}image\_\allowbreak{}3: user\_\allowbreak{}input\\{}
- input\_\allowbreak{}image\_\allowbreak{}4: user\_\allowbreak{}input\\{}
- input\_\allowbreak{}image\_\allowbreak{}5: user\_\allowbreak{}input\\{}
- input\_\allowbreak{}image\_\allowbreak{}6: user\_\allowbreak{}input\\{}
- input\_\allowbreak{}image\_\allowbreak{}7: user\_\allowbreak{}input\\{}
When a tool needs image input,\allowbreak{} pass one of these IDs via the *\_\allowbreak{}image\_\allowbreak{}id field.\\{}
Do NOT invent image file names or paths.
\end{exverb}
\end{exharness}
\begin{exstep}{1}{\toolVG}{}
\exWalk{Grounding in view~5, which shows the main sofa, returns 11 pillows.}
\exLabel{Arguments}{}
\begin{exverb}
\{\\{}
~~"\allowbreak{}query"\allowbreak{}: "\allowbreak{}throw pillows"\allowbreak{},\allowbreak{}\\{}
~~"\allowbreak{}image\_\allowbreak{}id"\allowbreak{}: "\allowbreak{}input\_\allowbreak{}image\_\allowbreak{}5"\allowbreak{}\\{}
\}
\end{exverb}
\exLabel{Tool output}{}
\begin{exverb}
Visual\_\allowbreak{}Grounding\_\allowbreak{}Tool grounded 11 objects for the prompt \textquotesingle{}throw pillows\textquotesingle{}. Their corresponding bounding boxes (on a 0-1000 scale,\allowbreak{} [x1,\allowbreak{} y1,\allowbreak{} x2,\allowbreak{} y2]) are: \textasciigrave{}\textasciigrave{}\textasciigrave{}json\\{}
[\\{}
~~~~\{"\allowbreak{}bbox\_\allowbreak{}2d"\allowbreak{}: [0,\allowbreak{} 684,\allowbreak{} 113,\allowbreak{} 866],\allowbreak{} "\allowbreak{}label"\allowbreak{}: "\allowbreak{}throw pillows"\allowbreak{}\},\allowbreak{}\\{}
~~~~\{"\allowbreak{}bbox\_\allowbreak{}2d"\allowbreak{}: [55,\allowbreak{} 717,\allowbreak{} 159,\allowbreak{} 861],\allowbreak{} "\allowbreak{}label"\allowbreak{}: "\allowbreak{}throw pillows"\allowbreak{}\},\allowbreak{}\\{}
~~~~\{"\allowbreak{}bbox\_\allowbreak{}2d"\allowbreak{}: [180,\allowbreak{} 677,\allowbreak{} 308,\allowbreak{} 825],\allowbreak{} "\allowbreak{}label"\allowbreak{}: "\allowbreak{}throw pillows"\allowbreak{}\},\allowbreak{}\\{}
~~~~\{"\allowbreak{}bbox\_\allowbreak{}2d"\allowbreak{}: [326,\allowbreak{} 709,\allowbreak{} 428,\allowbreak{} 830],\allowbreak{} "\allowbreak{}label"\allowbreak{}: "\allowbreak{}throw pillows"\allowbreak{}\},\allowbreak{}\\{}
~~~~\{"\allowbreak{}bbox\_\allowbreak{}2d"\allowbreak{}: [352,\allowbreak{} 664,\allowbreak{} 457,\allowbreak{} 817],\allowbreak{} "\allowbreak{}label"\allowbreak{}: "\allowbreak{}throw pillows"\allowbreak{}\},\allowbreak{}\\{}
~~~~\{"\allowbreak{}bbox\_\allowbreak{}2d"\allowbreak{}: [592,\allowbreak{} 787,\allowbreak{} 715,\allowbreak{} 908],\allowbreak{} "\allowbreak{}label"\allowbreak{}: "\allowbreak{}throw pillows"\allowbreak{}\},\allowbreak{}\\{}
~~~~\{"\allowbreak{}bbox\_\allowbreak{}2d"\allowbreak{}: [631,\allowbreak{} 707,\allowbreak{} 772,\allowbreak{} 841],\allowbreak{} "\allowbreak{}label"\allowbreak{}: "\allowbreak{}throw pillows"\allowbreak{}\},\allowbreak{}\\{}
~~~~\{"\allowbreak{}bbox\_\allowbreak{}2d"\allowbreak{}: [690,\allowbreak{} 819,\allowbreak{} 831,\allowbreak{} 1000],\allowbreak{} "\allowbreak{}label"\allowbreak{}: "\allowbreak{}throw pillows"\allowbreak{}\},\allowbreak{}\\{}
~~~~\{"\allowbreak{}bbox\_\allowbreak{}2d"\allowbreak{}: [755,\allowbreak{} 738,\allowbreak{} 898,\allowbreak{} 993],\allowbreak{} "\allowbreak{}label"\allowbreak{}: "\allowbreak{}throw pillows"\allowbreak{}\},\allowbreak{}\\{}
~~~~\{"\allowbreak{}bbox\_\allowbreak{}2d"\allowbreak{}: [841,\allowbreak{} 905,\allowbreak{} 937,\allowbreak{} 1000],\allowbreak{} "\allowbreak{}label"\allowbreak{}: "\allowbreak{}throw pillows"\allowbreak{}\},\allowbreak{}\\{}
~~~~\{"\allowbreak{}bbox\_\allowbreak{}2d"\allowbreak{}: [930,\allowbreak{} 816,\allowbreak{} 1000,\allowbreak{} 1000],\allowbreak{} "\allowbreak{}label"\allowbreak{}: "\allowbreak{}throw pillows"\allowbreak{}\}\\{}
]\textasciigrave{}\textasciigrave{}\textasciigrave{}. The object numbers correspond to the labeled masks rendered in the following image,\allowbreak{} which you can refer to as tool\_\allowbreak{}generated\_\allowbreak{}image\_\allowbreak{}1.
\end{exverb}
\begin{center}\includegraphics[width=0.45\linewidth,height=0.22\textheight,keepaspectratio]{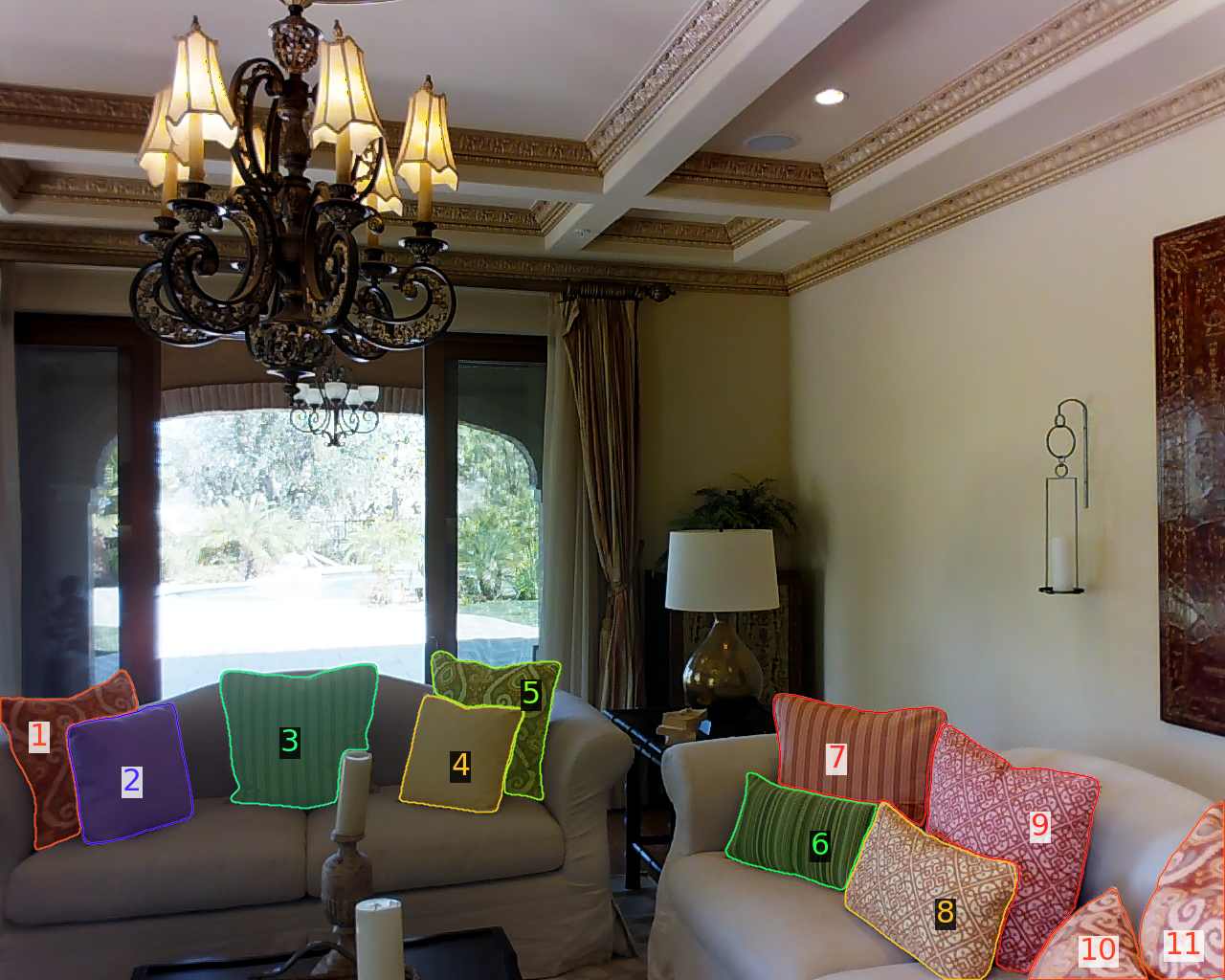}\\[2pt]
{\scriptsize\color{exMute} Image returned with this output.}\end{center}
\begin{exharness}
\exLabel{Harness}{environment update, summarized: new memory entries; new data bindings verbatim}
\begin{exverb}
Memory update: registers 1 new image in memory (tool\_\allowbreak{}generated\_\allowbreak{}image\_\allowbreak{}1,\allowbreak{} from Visual\_\allowbreak{}Grounding\_\allowbreak{}Tool).\\{}
Available data bindings (auto-injected into Python\_\allowbreak{}Coding\_\allowbreak{}Agent\_\allowbreak{}Tool):\\{}
- segment\_\allowbreak{}throw\_\allowbreak{}pillows: dict (keys: pred\_\allowbreak{}boxes,\allowbreak{} pred\_\allowbreak{}scores,\allowbreak{} pred\_\allowbreak{}masks,\allowbreak{} num\_\allowbreak{}masks,\allowbreak{} orig\_\allowbreak{}img\_\allowbreak{}h,\allowbreak{} orig\_\allowbreak{}img\_\allowbreak{}w)
\end{exverb}
\end{exharness}
\end{exstep}
\begin{exstep}{2}{\toolVG}{}
\exWalk{Grounding in view~3 returns 3 pillows.}
\exLabel{Arguments}{}
\begin{exverb}
\{\\{}
~~"\allowbreak{}query"\allowbreak{}: "\allowbreak{}throw pillows"\allowbreak{},\allowbreak{}\\{}
~~"\allowbreak{}image\_\allowbreak{}id"\allowbreak{}: "\allowbreak{}input\_\allowbreak{}image\_\allowbreak{}3"\allowbreak{}\\{}
\}
\end{exverb}
\exLabel{Tool output}{}
\begin{exverb}
Visual\_\allowbreak{}Grounding\_\allowbreak{}Tool grounded 3 objects for the prompt \textquotesingle{}throw pillows\textquotesingle{}. Their corresponding bounding boxes (on a 0-1000 scale,\allowbreak{} [x1,\allowbreak{} y1,\allowbreak{} x2,\allowbreak{} y2]) are: \textasciigrave{}\textasciigrave{}\textasciigrave{}json\\{}
[\\{}
~~~~\{"\allowbreak{}bbox\_\allowbreak{}2d"\allowbreak{}: [814,\allowbreak{} 671,\allowbreak{} 883,\allowbreak{} 806],\allowbreak{} "\allowbreak{}label"\allowbreak{}: "\allowbreak{}throw pillows"\allowbreak{}\},\allowbreak{}\\{}
~~~~\{"\allowbreak{}bbox\_\allowbreak{}2d"\allowbreak{}: [939,\allowbreak{} 663,\allowbreak{} 974,\allowbreak{} 738],\allowbreak{} "\allowbreak{}label"\allowbreak{}: "\allowbreak{}throw pillows"\allowbreak{}\},\allowbreak{}\\{}
~~~~\{"\allowbreak{}bbox\_\allowbreak{}2d"\allowbreak{}: [963,\allowbreak{} 669,\allowbreak{} 1000,\allowbreak{} 748],\allowbreak{} "\allowbreak{}label"\allowbreak{}: "\allowbreak{}throw pillows"\allowbreak{}\}\\{}
]\textasciigrave{}\textasciigrave{}\textasciigrave{}. The object numbers correspond to the labeled masks rendered in the following image,\allowbreak{} which you can refer to as tool\_\allowbreak{}generated\_\allowbreak{}image\_\allowbreak{}2.
\end{exverb}
\begin{center}\includegraphics[width=0.45\linewidth,height=0.22\textheight,keepaspectratio]{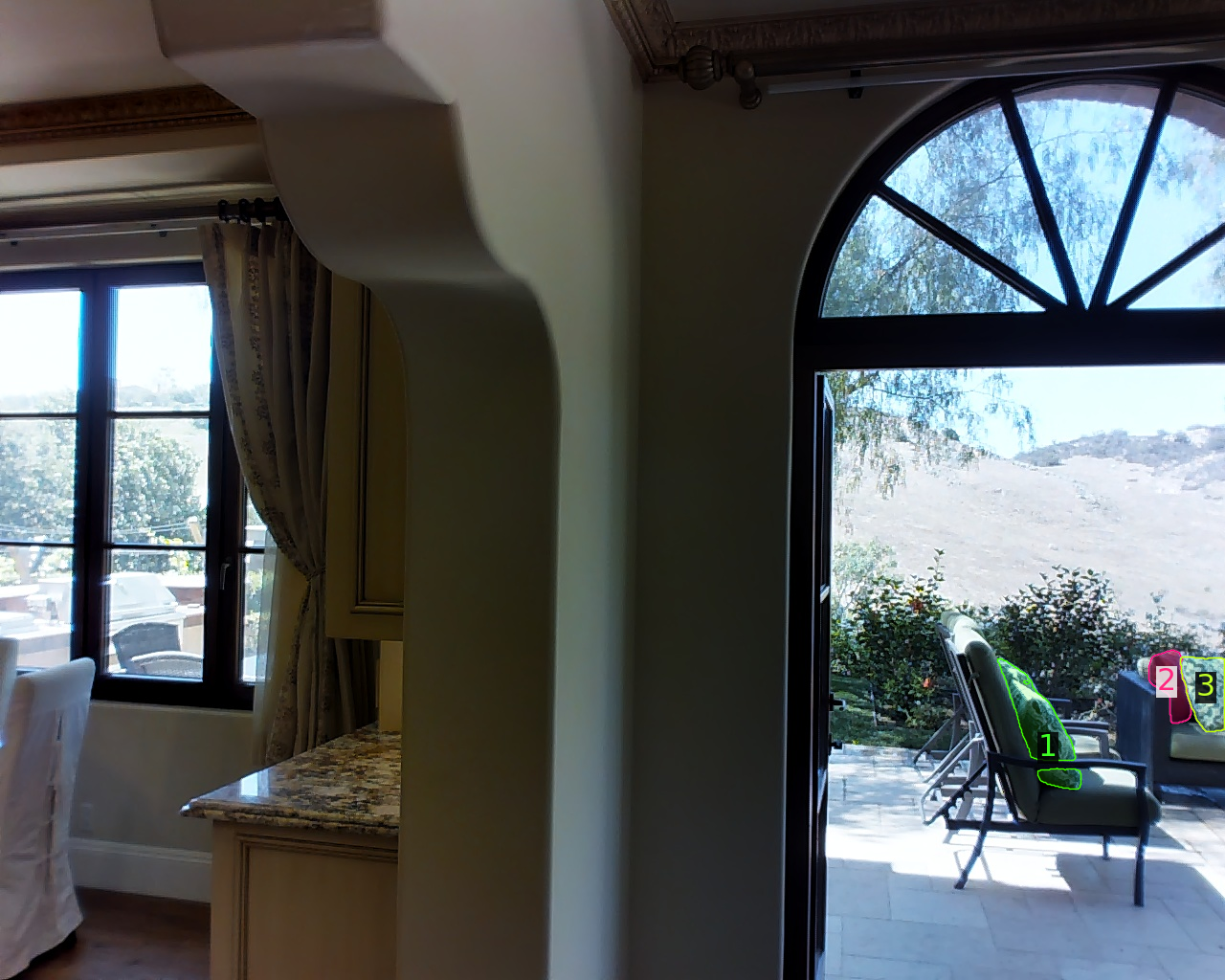}\\[2pt]
{\scriptsize\color{exMute} Image returned with this output.}\end{center}
\begin{exharness}
\exLabel{Harness}{environment update, summarized: new memory entries; new data bindings verbatim}
\begin{exverb}
Memory update: registers 1 new image in memory (tool\_\allowbreak{}generated\_\allowbreak{}image\_\allowbreak{}2,\allowbreak{} from Visual\_\allowbreak{}Grounding\_\allowbreak{}Tool).\\{}
- segment\_\allowbreak{}throw\_\allowbreak{}pillows\_\allowbreak{}2: dict (keys: pred\_\allowbreak{}boxes,\allowbreak{} pred\_\allowbreak{}scores,\allowbreak{} pred\_\allowbreak{}masks,\allowbreak{} num\_\allowbreak{}masks,\allowbreak{} orig\_\allowbreak{}img\_\allowbreak{}h,\allowbreak{} orig\_\allowbreak{}img\_\allowbreak{}w)
\end{exverb}
\end{exharness}
\end{exstep}
\begin{exstep}{3}{\toolVG}{}
\exWalk{Grounding in view~4 returns 3 pillows.}
\exLabel{Arguments}{}
\begin{exverb}
\{\\{}
~~"\allowbreak{}query"\allowbreak{}: "\allowbreak{}throw pillows"\allowbreak{},\allowbreak{}\\{}
~~"\allowbreak{}image\_\allowbreak{}id"\allowbreak{}: "\allowbreak{}input\_\allowbreak{}image\_\allowbreak{}4"\allowbreak{}\\{}
\}
\end{exverb}
\exLabel{Tool output}{}
\begin{exverb}
Visual\_\allowbreak{}Grounding\_\allowbreak{}Tool grounded 3 objects for the prompt \textquotesingle{}throw pillows\textquotesingle{}. Their corresponding bounding boxes (on a 0-1000 scale,\allowbreak{} [x1,\allowbreak{} y1,\allowbreak{} x2,\allowbreak{} y2]) are: \textasciigrave{}\textasciigrave{}\textasciigrave{}json\\{}
[\\{}
~~~~\{"\allowbreak{}bbox\_\allowbreak{}2d"\allowbreak{}: [0,\allowbreak{} 658,\allowbreak{} 46,\allowbreak{} 729],\allowbreak{} "\allowbreak{}label"\allowbreak{}: "\allowbreak{}throw pillows"\allowbreak{}\},\allowbreak{}\\{}
~~~~\{"\allowbreak{}bbox\_\allowbreak{}2d"\allowbreak{}: [144,\allowbreak{} 881,\allowbreak{} 277,\allowbreak{} 1000],\allowbreak{} "\allowbreak{}label"\allowbreak{}: "\allowbreak{}throw pillows"\allowbreak{}\},\allowbreak{}\\{}
~~~~\{"\allowbreak{}bbox\_\allowbreak{}2d"\allowbreak{}: [963,\allowbreak{} 724,\allowbreak{} 1000,\allowbreak{} 852],\allowbreak{} "\allowbreak{}label"\allowbreak{}: "\allowbreak{}throw pillows"\allowbreak{}\}\\{}
]\textasciigrave{}\textasciigrave{}\textasciigrave{}. The object numbers correspond to the labeled masks rendered in the following image,\allowbreak{} which you can refer to as tool\_\allowbreak{}generated\_\allowbreak{}image\_\allowbreak{}3.
\end{exverb}
\begin{center}\includegraphics[width=0.45\linewidth,height=0.22\textheight,keepaspectratio]{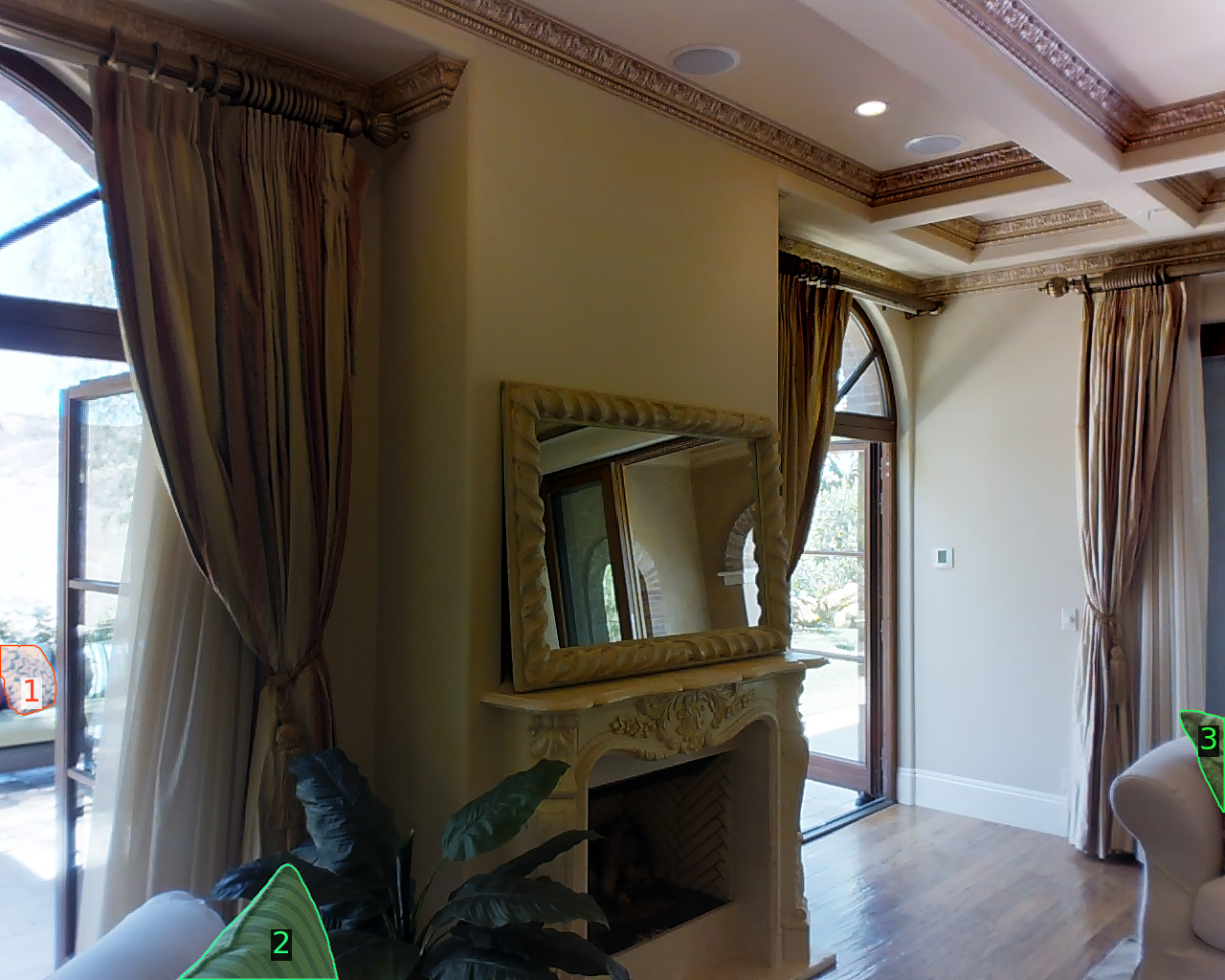}\\[2pt]
{\scriptsize\color{exMute} Image returned with this output.}\end{center}
\begin{exharness}
\exLabel{Harness}{environment update, summarized: new memory entries; new data bindings verbatim}
\begin{exverb}
Memory update: registers 1 new image in memory (tool\_\allowbreak{}generated\_\allowbreak{}image\_\allowbreak{}3,\allowbreak{} from Visual\_\allowbreak{}Grounding\_\allowbreak{}Tool).\\{}
- segment\_\allowbreak{}throw\_\allowbreak{}pillows\_\allowbreak{}3: dict (keys: pred\_\allowbreak{}boxes,\allowbreak{} pred\_\allowbreak{}scores,\allowbreak{} pred\_\allowbreak{}masks,\allowbreak{} num\_\allowbreak{}masks,\allowbreak{} orig\_\allowbreak{}img\_\allowbreak{}h,\allowbreak{} orig\_\allowbreak{}img\_\allowbreak{}w)
\end{exverb}
\end{exharness}
\end{exstep}
\begin{exstep}{4}{\toolVG}{}
\exWalk{Grounding in view~6 returns 1 pillow.}
\exLabel{Arguments}{}
\begin{exverb}
\{\\{}
~~"\allowbreak{}query"\allowbreak{}: "\allowbreak{}throw pillows"\allowbreak{},\allowbreak{}\\{}
~~"\allowbreak{}image\_\allowbreak{}id"\allowbreak{}: "\allowbreak{}input\_\allowbreak{}image\_\allowbreak{}6"\allowbreak{}\\{}
\}
\end{exverb}
\exLabel{Tool output}{}
\begin{exverb}
Visual\_\allowbreak{}Grounding\_\allowbreak{}Tool grounded 1 objects for the prompt \textquotesingle{}throw pillows\textquotesingle{}. Their corresponding bounding boxes (on a 0-1000 scale,\allowbreak{} [x1,\allowbreak{} y1,\allowbreak{} x2,\allowbreak{} y2]) are: \textasciigrave{}\textasciigrave{}\textasciigrave{}json\\{}
[\\{}
~~~~\{"\allowbreak{}bbox\_\allowbreak{}2d"\allowbreak{}: [0,\allowbreak{} 799,\allowbreak{} 102,\allowbreak{} 1000],\allowbreak{} "\allowbreak{}label"\allowbreak{}: "\allowbreak{}throw pillows"\allowbreak{}\}\\{}
]\textasciigrave{}\textasciigrave{}\textasciigrave{}. The object numbers correspond to the labeled masks rendered in the following image,\allowbreak{} which you can refer to as tool\_\allowbreak{}generated\_\allowbreak{}image\_\allowbreak{}4.
\end{exverb}
\begin{center}\includegraphics[width=0.45\linewidth,height=0.22\textheight,keepaspectratio]{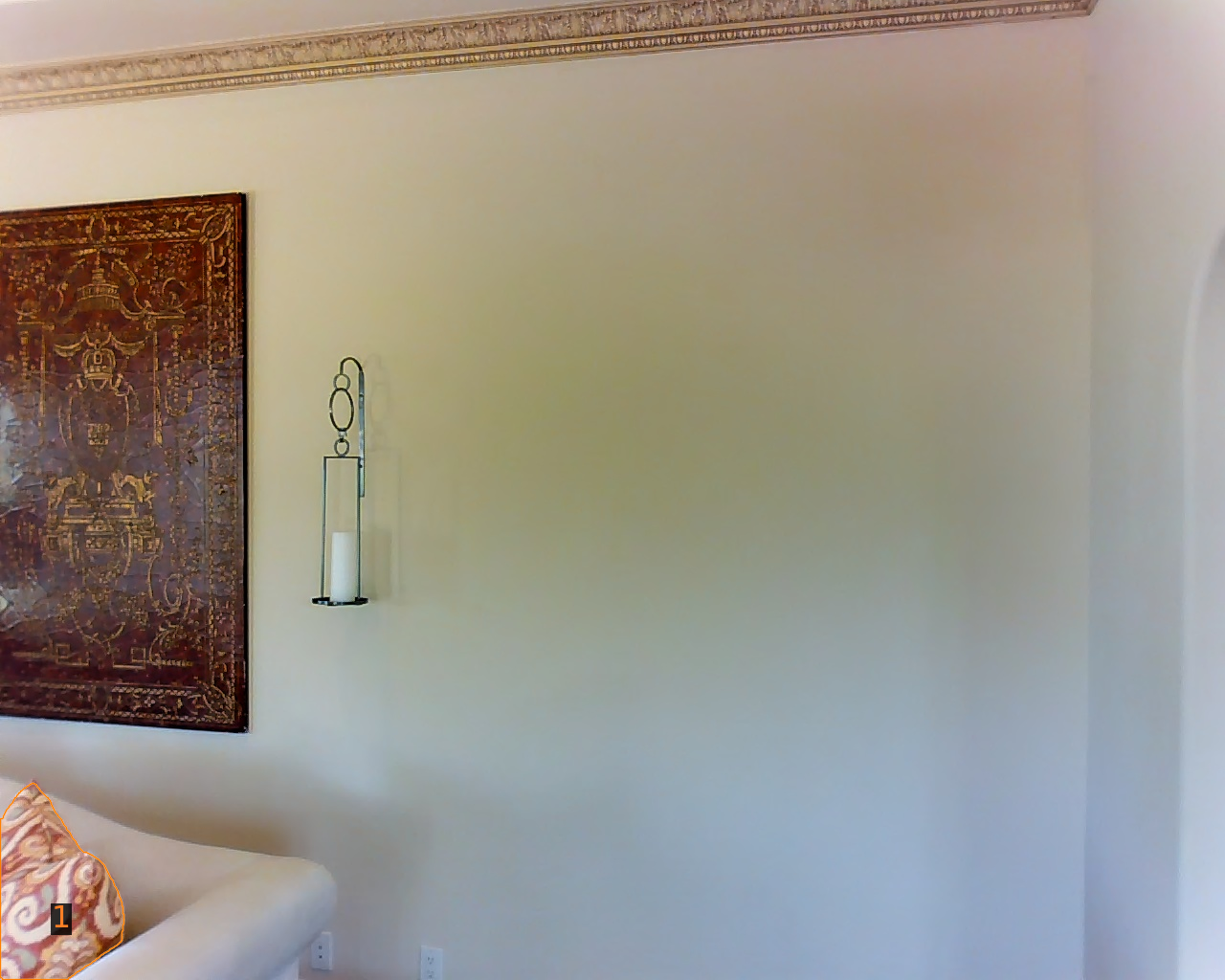}\\[2pt]
{\scriptsize\color{exMute} Image returned with this output.}\end{center}
\begin{exharness}
\exLabel{Harness}{environment update, summarized: new memory entries; new data bindings verbatim}
\begin{exverb}
Memory update: registers 1 new image in memory (tool\_\allowbreak{}generated\_\allowbreak{}image\_\allowbreak{}4,\allowbreak{} from Visual\_\allowbreak{}Grounding\_\allowbreak{}Tool).\\{}
- segment\_\allowbreak{}throw\_\allowbreak{}pillows\_\allowbreak{}4: dict (keys: pred\_\allowbreak{}boxes,\allowbreak{} pred\_\allowbreak{}scores,\allowbreak{} pred\_\allowbreak{}masks,\allowbreak{} num\_\allowbreak{}masks,\allowbreak{} orig\_\allowbreak{}img\_\allowbreak{}h,\allowbreak{} orig\_\allowbreak{}img\_\allowbreak{}w)
\end{exverb}
\end{exharness}
\end{exstep}
\begin{exstep}{5}{\toolZoom}{}
\exWalk{A zoom on the lower right of view~5, where pillows on the sofa overlap.}
\exLabel{Arguments}{}
\begin{exverb}
\{\\{}
~~"\allowbreak{}image\_\allowbreak{}id"\allowbreak{}: "\allowbreak{}input\_\allowbreak{}image\_\allowbreak{}5"\allowbreak{},\allowbreak{}\\{}
~~"\allowbreak{}coords"\allowbreak{}: [530,\allowbreak{} 660,\allowbreak{} 1000,\allowbreak{} 1000],\allowbreak{}\\{}
~~"\allowbreak{}padding"\allowbreak{}: 50\\{}
\}
\end{exverb}
\exLabel{Tool output}{}
\begin{exverb}
Tool \textasciigrave{}Zoom\_\allowbreak{}In\_\allowbreak{}Tool\textasciigrave{} (tool\_\allowbreak{}call\_\allowbreak{}id=\allowbreak{}call\_\allowbreak{}MTJTBg06TkGQbxJlDKAYPR3r) returned:\\{}
Zoomed in image rendered below; refer to it as tool\_\allowbreak{}generated\_\allowbreak{}image\_\allowbreak{}5.
\end{exverb}
\begin{center}\includegraphics[width=0.45\linewidth,height=0.22\textheight,keepaspectratio]{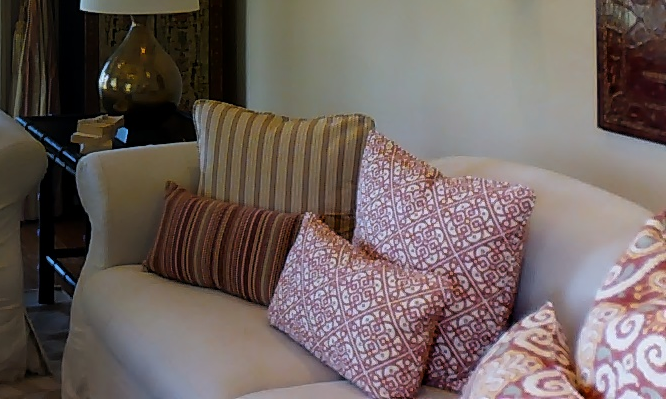}\\[2pt]
{\scriptsize\color{exMute} Image returned with this output.}\end{center}
\begin{exharness}
\exLabel{Harness}{environment update, summarized: new memory entries; new data bindings verbatim}
\begin{exverb}
Memory update: registers 1 new image in memory (tool\_\allowbreak{}generated\_\allowbreak{}image\_\allowbreak{}5,\allowbreak{} from Zoom\_\allowbreak{}In\_\allowbreak{}Tool).
\end{exverb}
\end{exharness}
\end{exstep}
\begin{exstep}{6}{\toolZoom}{}
\exWalk{A zoom on the lower left of view~5.}
\exLabel{Arguments}{}
\begin{exverb}
\{\\{}
~~"\allowbreak{}image\_\allowbreak{}id"\allowbreak{}: "\allowbreak{}input\_\allowbreak{}image\_\allowbreak{}5"\allowbreak{},\allowbreak{}\\{}
~~"\allowbreak{}coords"\allowbreak{}: [0,\allowbreak{} 650,\allowbreak{} 530,\allowbreak{} 1000],\allowbreak{}\\{}
~~"\allowbreak{}padding"\allowbreak{}: 50\\{}
\}
\end{exverb}
\exLabel{Tool output}{}
\begin{exverb}
Tool \textasciigrave{}Zoom\_\allowbreak{}In\_\allowbreak{}Tool\textasciigrave{} (tool\_\allowbreak{}call\_\allowbreak{}id=\allowbreak{}call\_\allowbreak{}Ybk2IO4dyyOZDI6yeGGdwGt8) returned:\\{}
Zoomed in image rendered below; refer to it as tool\_\allowbreak{}generated\_\allowbreak{}image\_\allowbreak{}6.
\end{exverb}
\begin{center}\includegraphics[width=0.45\linewidth,height=0.22\textheight,keepaspectratio]{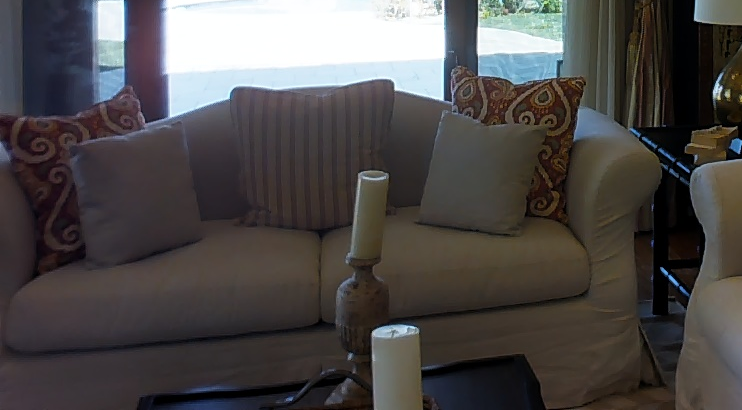}\\[2pt]
{\scriptsize\color{exMute} Image returned with this output.}\end{center}
\begin{exharness}
\exLabel{Harness}{environment update, summarized: new memory entries; new data bindings verbatim}
\begin{exverb}
Memory update: registers 1 new image in memory (tool\_\allowbreak{}generated\_\allowbreak{}image\_\allowbreak{}6,\allowbreak{} from Zoom\_\allowbreak{}In\_\allowbreak{}Tool).
\end{exverb}
\end{exharness}
\end{exstep}
\begin{exstep}{}{\exTool{exInk}{Final answer}}{}
\exWalk{The grounding calls found 18 pillows in total, but some are the same pillow seen from two views. The agent answers C, 16 pillows, which is correct; it does not write out which detections it treated as duplicates.}
\exLabel{Model}{}
\begin{exverb}
\textless{}answer\textgreater{}\textasciigrave{}C\textasciigrave{}\textless{}/\allowbreak{}answer\textgreater{}
\end{exverb}
\end{exstep}

%% file: text/appendix_examples_steps/qmmsi7_header.tex
\exOverview{\textbf{Task.} Two photographs show two rooms of a historic house: a parlour with a piano, and a dining room. Which is longer, the piano or the table under the mirror? Without tools the same model answers D, that the piano is longer. 

\textbf{How the agent approaches it.} It grounds both objects, estimates metric depth in both photos, and has the coding agent convert each box to a real-world length. 

\textbf{Caveat.} The answer is correct, but the evidence does not support it. The mirror hangs above the fireplace, so the ``table under the mirror'' is the mantel shelf. Grounding returns it as object~3, but the agent chooses the round dining table (object~2) instead, and concludes that the table, at 2.77\,m, is longer than the piano, at 2.21\,m. This example shows the harness's tool chain, not a sound comparison.}
\exLabel{Input images}{}
\begin{center}
\begin{minipage}[t]{0.475\linewidth}\centering\includegraphics[width=\linewidth,height=0.22\textheight,keepaspectratio]{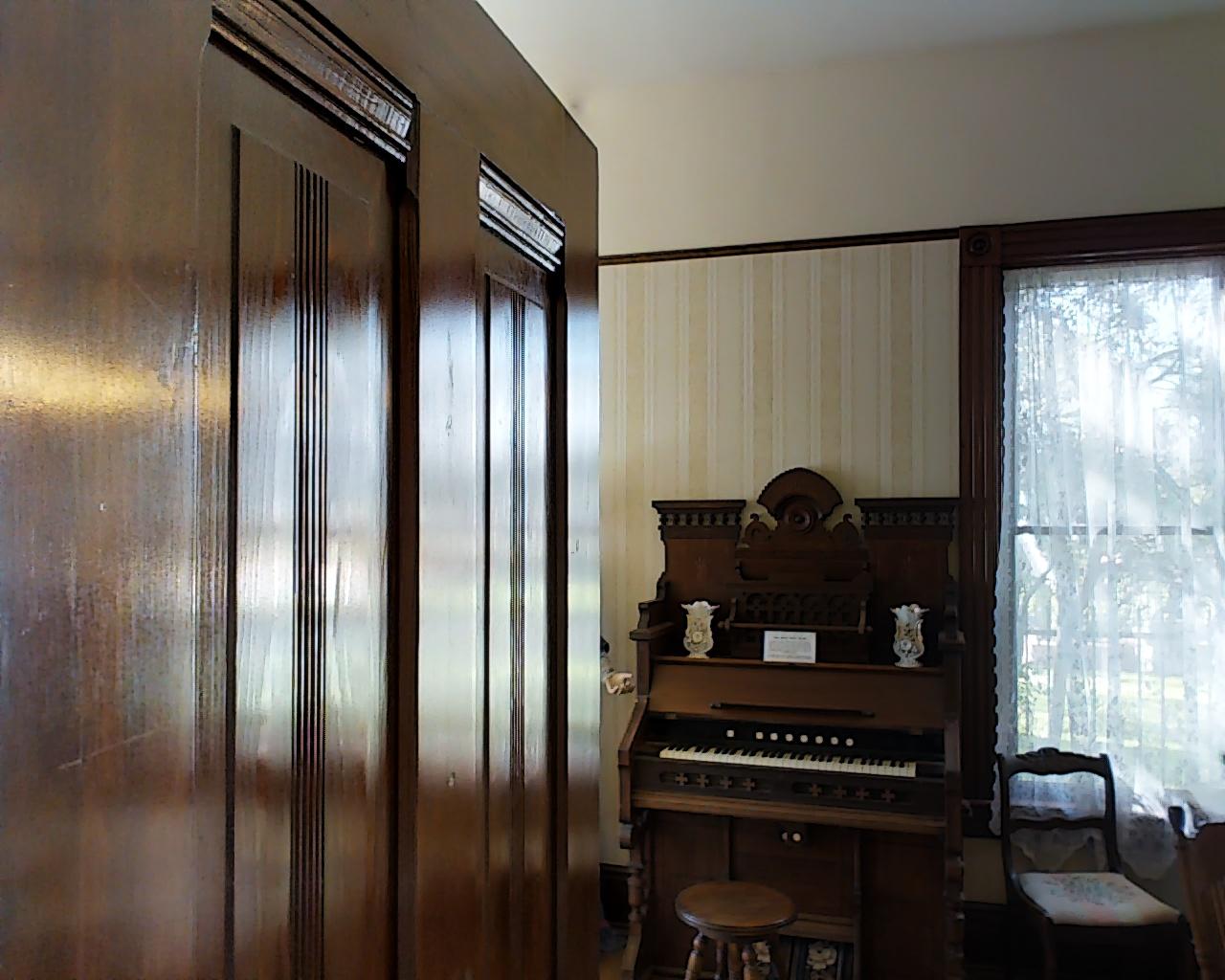}\\[1pt]{\scriptsize\color{exMute} input\_image\_1}\end{minipage}\hfill\begin{minipage}[t]{0.475\linewidth}\centering\includegraphics[width=\linewidth,height=0.22\textheight,keepaspectratio]{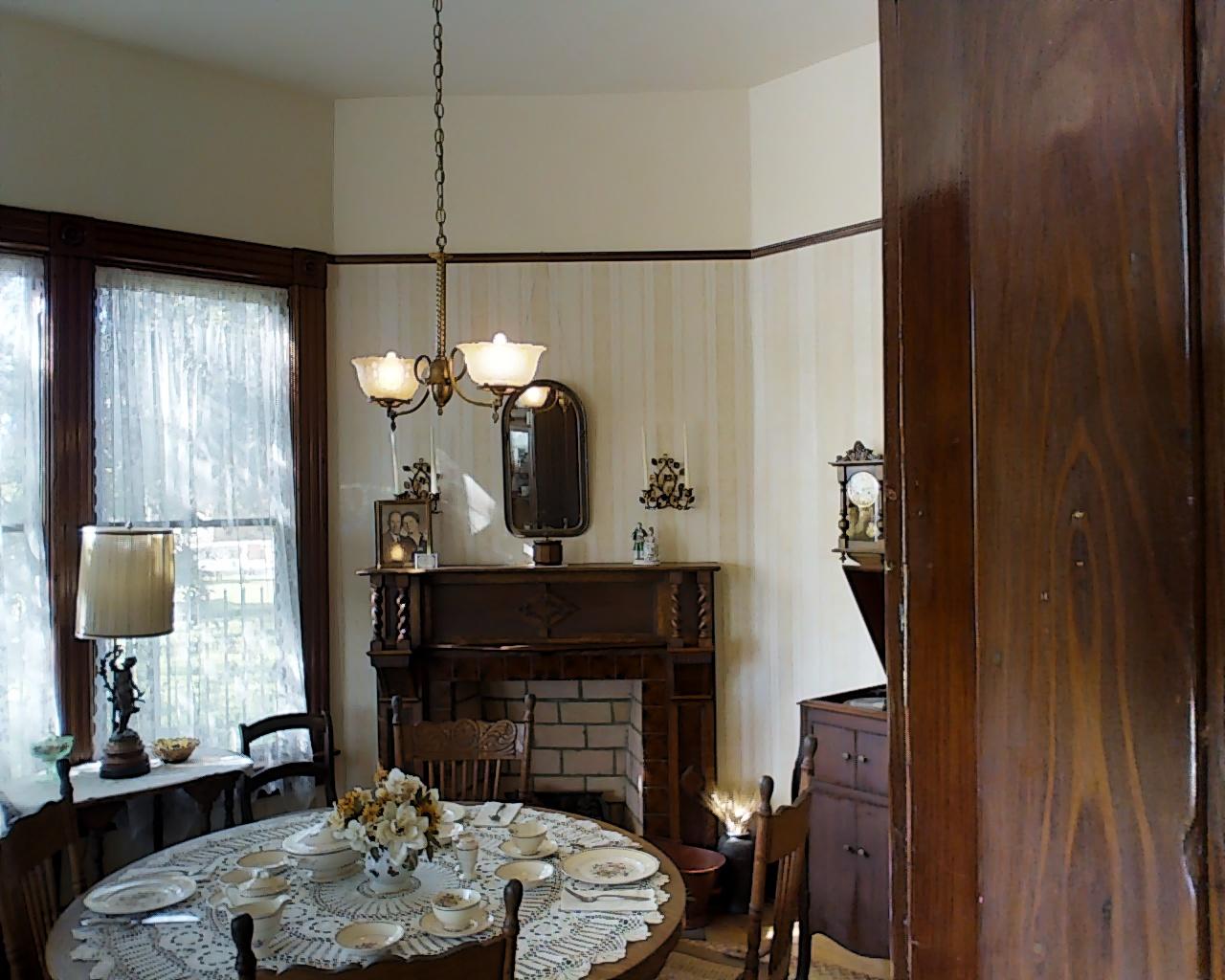}\\[1pt]{\scriptsize\color{exMute} input\_image\_2}\end{minipage}
\end{center}
\exLabel{Question}{as sent to the model}
\begin{exverb}
Question: Which is longer,\allowbreak{} the piano or the table under the mirror?\\{}
Options: A: The table under the mirror is longer,\allowbreak{} B: Sometimes the table under the mirror is longer,\allowbreak{} sometimes the piano is longer,\allowbreak{} C: They are the same length,\allowbreak{} D: The piano is longer,\allowbreak{} \\{}
Answer with the option\textquotesingle{}s letter from the given choices directly. Enclose the option\textquotesingle{}s letter within \textasciigrave{}\textasciigrave{}.
\end{exverb}

%% file: text/appendix_examples_steps/qmmsi7.tex
\begin{exharness}
\exLabel{Harness}{environment update, shown in full}
\begin{exverb}
Environment Update (not a new question):\\{}
Currently available image IDs for tool calls:\\{}
- input\_\allowbreak{}image\_\allowbreak{}1: user\_\allowbreak{}input\\{}
- input\_\allowbreak{}image\_\allowbreak{}2: user\_\allowbreak{}input\\{}
When a tool needs image input,\allowbreak{} pass one of these IDs via the *\_\allowbreak{}image\_\allowbreak{}id field.\\{}
Do NOT invent image file names or paths.
\end{exverb}
\end{exharness}
\begin{exstep}{1}{\toolVG}{}
\exLabel{Model}{text emitted with this call}
\exSayBlock{\textless{}thinking\textgreater{}To determine which is longer—the piano or the table under the mirror—I need to compare their lengths. The piano is visible in input\_\allowbreak{}image\_\allowbreak{}1,\allowbreak{} and the table under the mirror is visible in input\_\allowbreak{}image\_\allowbreak{}2. Since they are in different images,\allowbreak{} I need to use a tool to estimate their lengths relative to each other. I will first ground both objects in their respective images to get bounding boxes,\allowbreak{} then use a depth estimation tool to get relative scales,\allowbreak{} and finally compare them using Python\_\allowbreak{}Coding\_\allowbreak{}Agent\_\allowbreak{}Tool.\textless{}/\allowbreak{}thinking\textgreater{}}
\exLabel{Arguments}{}
\begin{exverb}
\{\\{}
~~"\allowbreak{}query"\allowbreak{}: "\allowbreak{}piano"\allowbreak{},\allowbreak{}\\{}
~~"\allowbreak{}image\_\allowbreak{}id"\allowbreak{}: "\allowbreak{}input\_\allowbreak{}image\_\allowbreak{}1"\allowbreak{}\\{}
\}
\end{exverb}
\exLabel{Tool output}{}
\begin{exverb}
Visual\_\allowbreak{}Grounding\_\allowbreak{}Tool grounded 1 objects for the prompt \textquotesingle{}piano\textquotesingle{}. Their corresponding bounding boxes (on a 0-1000 scale,\allowbreak{} [x1,\allowbreak{} y1,\allowbreak{} x2,\allowbreak{} y2]) are: \{\textquotesingle{}object 1\textquotesingle{}: [503.72,\allowbreak{} 477.32,\allowbreak{} 787.18,\allowbreak{} 999.63]\}. The object numbers correspond to the labeled masks rendered in the following image,\allowbreak{} which you can refer to as tool\_\allowbreak{}generated\_\allowbreak{}image\_\allowbreak{}1.
\end{exverb}
\begin{center}\includegraphics[width=0.45\linewidth,height=0.22\textheight,keepaspectratio]{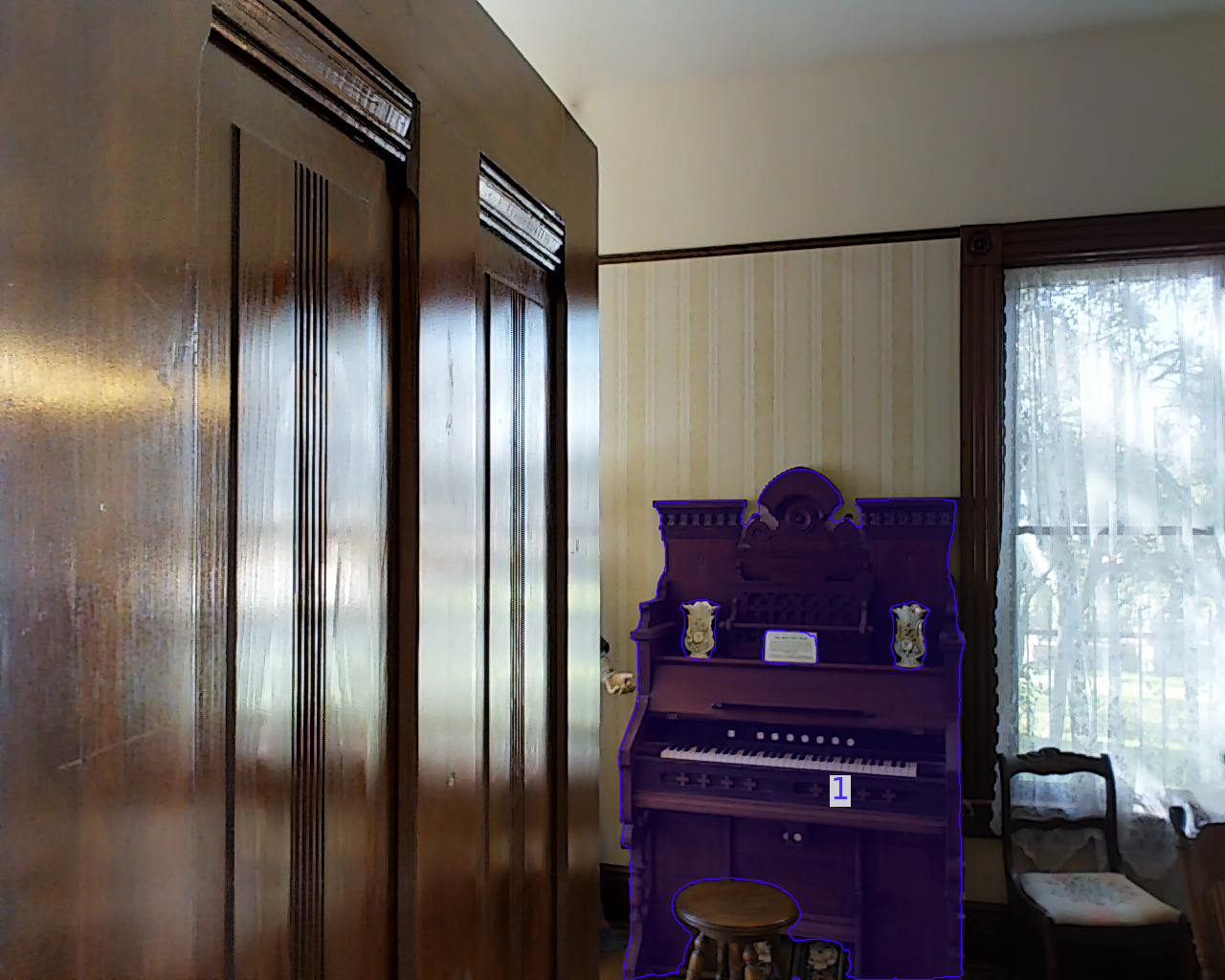}\\[2pt]
{\scriptsize\color{exMute} Image returned with this output.}\end{center}
\begin{exharness}
\exLabel{Harness}{environment update, summarized: new memory entries; new data bindings verbatim}
\begin{exverb}
Memory update: registers 1 new image in memory (tool\_\allowbreak{}generated\_\allowbreak{}image\_\allowbreak{}1,\allowbreak{} from Visual\_\allowbreak{}Grounding\_\allowbreak{}Tool).\\{}
Available data bindings (auto-injected into Python\_\allowbreak{}Coding\_\allowbreak{}Agent\_\allowbreak{}Tool):\\{}
- segment\_\allowbreak{}piano: dict (keys: pred\_\allowbreak{}boxes,\allowbreak{} pred\_\allowbreak{}scores,\allowbreak{} pred\_\allowbreak{}masks,\allowbreak{} num\_\allowbreak{}masks,\allowbreak{} orig\_\allowbreak{}img\_\allowbreak{}h,\allowbreak{} orig\_\allowbreak{}img\_\allowbreak{}w)
\end{exverb}
\end{exharness}
\end{exstep}
\begin{exstep}{2}{\toolVG}{}
\exLabel{Arguments}{}
\begin{exverb}
\{\\{}
~~"\allowbreak{}query"\allowbreak{}: "\allowbreak{}table under the mirror"\allowbreak{},\allowbreak{}\\{}
~~"\allowbreak{}image\_\allowbreak{}id"\allowbreak{}: "\allowbreak{}input\_\allowbreak{}image\_\allowbreak{}2"\allowbreak{}\\{}
\}
\end{exverb}
\exLabel{Tool output}{}
\begin{exverb}
Visual\_\allowbreak{}Grounding\_\allowbreak{}Tool grounded 3 objects for the prompt \textquotesingle{}table under the mirror\textquotesingle{}. Their corresponding bounding boxes (on a 0-1000 scale,\allowbreak{} [x1,\allowbreak{} y1,\allowbreak{} x2,\allowbreak{} y2]) are: \{\textquotesingle{}object 1\textquotesingle{}: [2.6,\allowbreak{} 762.44,\allowbreak{} 205.84,\allowbreak{} 901.39],\allowbreak{} \textquotesingle{}object 2\textquotesingle{}: [38.62,\allowbreak{} 823.36,\allowbreak{} 556.79,\allowbreak{} 1000],\allowbreak{} \textquotesingle{}object 3\textquotesingle{}: [294.14,\allowbreak{} 573.73,\allowbreak{} 587.94,\allowbreak{} 865.76]\}. The object numbers correspond to the labeled masks rendered in the following image,\allowbreak{} which you can refer to as tool\_\allowbreak{}generated\_\allowbreak{}image\_\allowbreak{}2.
\end{exverb}
\begin{center}\includegraphics[width=0.45\linewidth,height=0.22\textheight,keepaspectratio]{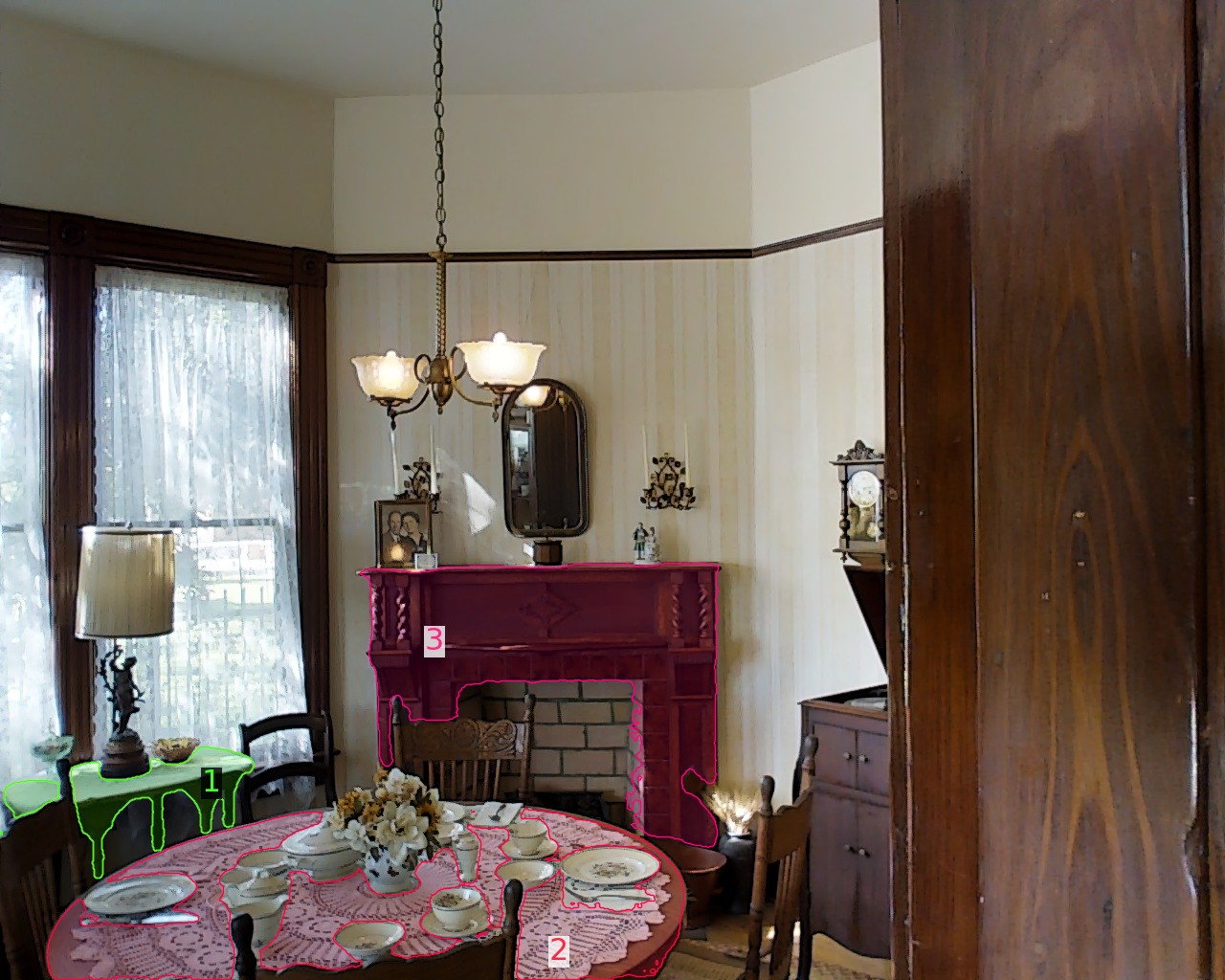}\\[2pt]
{\scriptsize\color{exMute} Image returned with this output.}\end{center}
\begin{exharness}
\exLabel{Harness}{environment update, summarized: new memory entries; new data bindings verbatim}
\begin{exverb}
Memory update: registers 1 new image in memory (tool\_\allowbreak{}generated\_\allowbreak{}image\_\allowbreak{}2,\allowbreak{} from Visual\_\allowbreak{}Grounding\_\allowbreak{}Tool).\\{}
- segment\_\allowbreak{}table\_\allowbreak{}under\_\allowbreak{}the\_\allowbreak{}mirror: dict (keys: pred\_\allowbreak{}boxes,\allowbreak{} pred\_\allowbreak{}scores,\allowbreak{} pred\_\allowbreak{}masks,\allowbreak{} num\_\allowbreak{}masks,\allowbreak{} orig\_\allowbreak{}img\_\allowbreak{}h,\allowbreak{} orig\_\allowbreak{}img\_\allowbreak{}w)
\end{exverb}
\end{exharness}
\end{exstep}
\begin{exstep}{3}{\toolDepth}{}
\exLabel{Model}{text emitted with this call}
\exSayBlock{\textless{}thinking\textgreater{}From the grounding results,\allowbreak{} I have identified the piano in input\_\allowbreak{}image\_\allowbreak{}1 with bounding box [503.72,\allowbreak{} 477.32,\allowbreak{} 787.18,\allowbreak{} 999.63]. In input\_\allowbreak{}image\_\allowbreak{}2,\allowbreak{} there are three objects grounded under the prompt \textquotesingle{}table under the mirror\textquotesingle{}: a small side table (object 1),\allowbreak{} a round dining table (object 2),\allowbreak{} and a fireplace mantel (object 3). The most relevant object for comparison is the round dining table (object 2) with bounding box [38.62,\allowbreak{} 823.36,\allowbreak{} 556.79,\allowbreak{} 1000].

To compare their lengths accurately,\allowbreak{} I need to account for perspective and depth. I will estimate the depth map for both images to get relative depth values at the centers of the piano and the dining table,\allowbreak{} then compute their projected lengths using the bounding box dimensions and depth information.\textless{}/\allowbreak{}thinking\textgreater{}}
\exLabel{Arguments}{}
\begin{exverb}
\{\\{}
~~"\allowbreak{}image\_\allowbreak{}id"\allowbreak{}: "\allowbreak{}input\_\allowbreak{}image\_\allowbreak{}1"\allowbreak{},\allowbreak{}\\{}
~~"\allowbreak{}colormap"\allowbreak{}: "\allowbreak{}grayscale"\allowbreak{}\\{}
\}
\end{exverb}
\exLabel{Tool output}{}
\begin{exverb}
Tool \textasciigrave{}Depth\_\allowbreak{}Estimation\_\allowbreak{}Tool\textasciigrave{} (tool\_\allowbreak{}call\_\allowbreak{}id=\allowbreak{}chatcmpl-tool-a16443994cabdb5b) returned:\\{}
Depth estimation complete. Shape: [1022,\allowbreak{} 1274],\allowbreak{} colormap: grayscale.\\{}
Metric depth in meters available as \textasciigrave{}depth\_\allowbreak{}meters\textasciigrave{} (float32 ndarray).\\{}
Colorized depth map rendered below; refer to it as tool\_\allowbreak{}generated\_\allowbreak{}image\_\allowbreak{}3.
\end{exverb}
\begin{center}\includegraphics[width=0.45\linewidth,height=0.22\textheight,keepaspectratio]{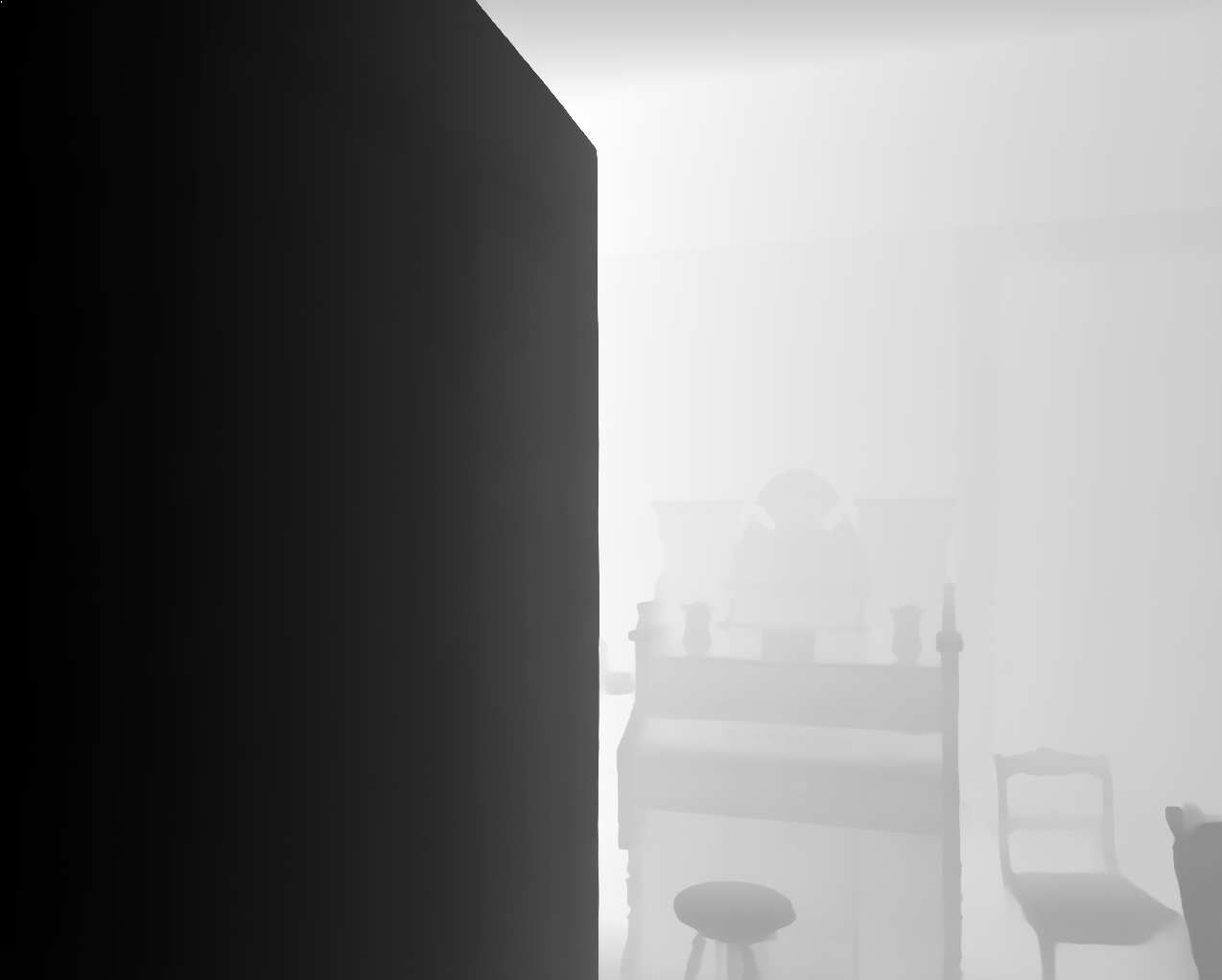}\\[2pt]
{\scriptsize\color{exMute} Image returned with this output.}\end{center}
\begin{exharness}
\exLabel{Harness}{environment update, summarized: new memory entries; new data bindings verbatim}
\begin{exverb}
Memory update: registers 1 new image in memory (tool\_\allowbreak{}generated\_\allowbreak{}image\_\allowbreak{}3,\allowbreak{} from Depth\_\allowbreak{}Estimation\_\allowbreak{}Tool).\\{}
- depth\_\allowbreak{}result: dict (keys: depth\_\allowbreak{}shape,\allowbreak{} colormap,\allowbreak{} size,\allowbreak{} depth\_\allowbreak{}meters)
\end{exverb}
\end{exharness}
\end{exstep}
\begin{exstep}{4}{\toolDepth}{}
\exLabel{Arguments}{}
\begin{exverb}
\{\\{}
~~"\allowbreak{}image\_\allowbreak{}id"\allowbreak{}: "\allowbreak{}input\_\allowbreak{}image\_\allowbreak{}2"\allowbreak{},\allowbreak{}\\{}
~~"\allowbreak{}colormap"\allowbreak{}: "\allowbreak{}grayscale"\allowbreak{}\\{}
\}
\end{exverb}
\exLabel{Tool output}{}
\begin{exverb}
Tool \textasciigrave{}Depth\_\allowbreak{}Estimation\_\allowbreak{}Tool\textasciigrave{} (tool\_\allowbreak{}call\_\allowbreak{}id=\allowbreak{}chatcmpl-tool-928a1ad88a9f8106) returned:\\{}
Depth estimation complete. Shape: [1022,\allowbreak{} 1274],\allowbreak{} colormap: grayscale.\\{}
Metric depth in meters available as \textasciigrave{}depth\_\allowbreak{}meters\textasciigrave{} (float32 ndarray).\\{}
Colorized depth map rendered below; refer to it as tool\_\allowbreak{}generated\_\allowbreak{}image\_\allowbreak{}4.
\end{exverb}
\begin{center}\includegraphics[width=0.45\linewidth,height=0.22\textheight,keepaspectratio]{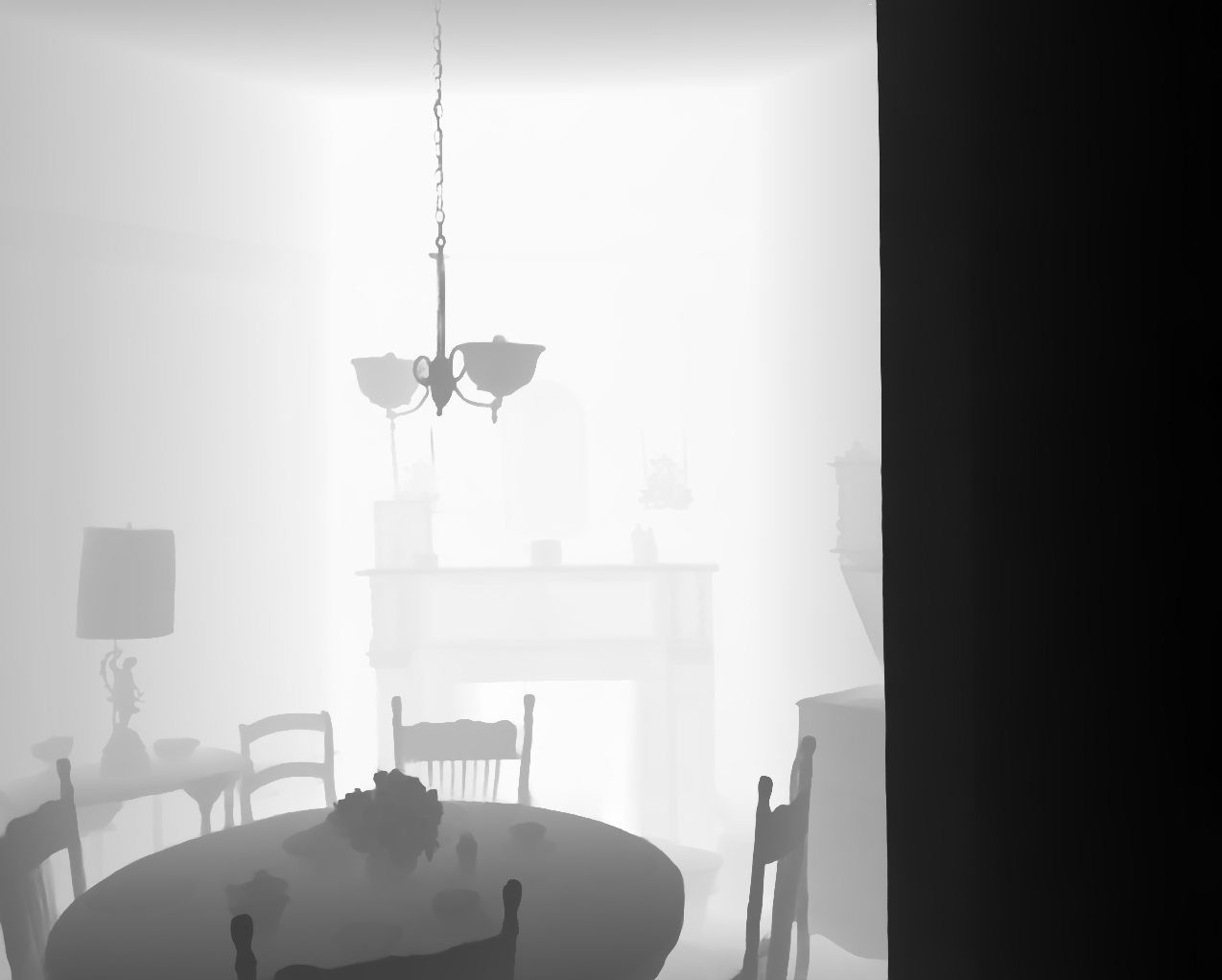}\\[2pt]
{\scriptsize\color{exMute} Image returned with this output.}\end{center}
\begin{exharness}
\exLabel{Harness}{environment update, summarized: new memory entries; new data bindings verbatim}
\begin{exverb}
Memory update: registers 1 new image in memory (tool\_\allowbreak{}generated\_\allowbreak{}image\_\allowbreak{}4,\allowbreak{} from Depth\_\allowbreak{}Estimation\_\allowbreak{}Tool).\\{}
- depth\_\allowbreak{}result\_\allowbreak{}2: dict (keys: depth\_\allowbreak{}shape,\allowbreak{} colormap,\allowbreak{} size,\allowbreak{} depth\_\allowbreak{}meters)
\end{exverb}
\end{exharness}
\end{exstep}
\begin{exstep}{5}{\toolCode}{}
\exLabel{Model}{text emitted with this call}
\exSayBlock{\textless{}thinking\textgreater{}Now I have depth maps for both images. I will use the Python\_\allowbreak{}Coding\_\allowbreak{}Agent\_\allowbreak{}Tool to compute the actual lengths of the piano and the dining table (object 2) by combining their bounding box dimensions with the depth information at their centers. This will allow me to compare their real-world lengths accurately.\textless{}/\allowbreak{}thinking\textgreater{}}
\exLabel{Arguments}{}
\begin{exverb}
\{\\{}
~~"\allowbreak{}objective"\allowbreak{}: "\allowbreak{}Compute the real-world length of the piano from input\_\allowbreak{}image\_\allowbreak{}1 and the dining table (object 2) from input\_\allowbreak{}image\_\allowbreak{}2 using their bounding boxes and depth values at their centers. The piano\textquotesingle{}s bounding box is [503.72,\allowbreak{} 477.32,\allowbreak{} 787.18,\allowbreak{} 999.63] and the dining table\textquotesingle{}s bounding box is [38.62,\allowbreak{} 823.36,\allowbreak{} 556.79,\allowbreak{} 1000]. Use the depth maps from the Depth\_\allowbreak{}Estimation\_\allowbreak{}Tool to get the depth at the center of each object. Then,\allowbreak{} calculate the projected length using the formula: real\_\allowbreak{}length =\allowbreak{} (bbox\_\allowbreak{}width /\allowbreak{} image\_\allowbreak{}width) * depth * 2 (assuming a rough perspective correction factor). Compare the two lengths and determine which is longer."\allowbreak{}\\{}
\}
\end{exverb}
\exLabel{Tool output}{}
\begin{exverb}
Coding agent generated and executed the following Python code:\\{}
\textasciigrave{}\textasciigrave{}\textasciigrave{}python\\{}
import numpy as np\\{}
\mbox{}\\{}
\# Given bounding boxes (from the objective)\\{}
piano\_\allowbreak{}bbox =\allowbreak{} [503.72,\allowbreak{} 477.32,\allowbreak{} 787.18,\allowbreak{} 999.63]  \# [x1,\allowbreak{} y1,\allowbreak{} x2,\allowbreak{} y2]\\{}
table\_\allowbreak{}bbox =\allowbreak{} [38.62,\allowbreak{} 823.36,\allowbreak{} 556.79,\allowbreak{} 1000]     \# [x1,\allowbreak{} y1,\allowbreak{} x2,\allowbreak{} y2]\\{}
\mbox{}\\{}
\# Image dimensions (from segment\_\allowbreak{}piano and depth\_\allowbreak{}result)\\{}
image\_\allowbreak{}width =\allowbreak{} segment\_\allowbreak{}piano["\allowbreak{}orig\_\allowbreak{}img\_\allowbreak{}w"\allowbreak{}]  \# 1280\\{}
image\_\allowbreak{}height =\allowbreak{} segment\_\allowbreak{}piano["\allowbreak{}orig\_\allowbreak{}img\_\allowbreak{}h"\allowbreak{}]  \# 1024\\{}
\mbox{}\\{}
\# Depth maps\\{}
depth\_\allowbreak{}map\_\allowbreak{}1 =\allowbreak{} depth\_\allowbreak{}result["\allowbreak{}depth\_\allowbreak{}meters"\allowbreak{}]  \# for piano (input\_\allowbreak{}image\_\allowbreak{}1)\\{}
depth\_\allowbreak{}map\_\allowbreak{}2 =\allowbreak{} depth\_\allowbreak{}result\_\allowbreak{}2["\allowbreak{}depth\_\allowbreak{}meters"\allowbreak{}]  \# for table (input\_\allowbreak{}image\_\allowbreak{}2)\\{}
\mbox{}\\{}
\# Compute center coordinates for each object\\{}
piano\_\allowbreak{}center\_\allowbreak{}x =\allowbreak{} (piano\_\allowbreak{}bbox[0] + piano\_\allowbreak{}bbox[2]) /\allowbreak{} 2\\{}
piano\_\allowbreak{}center\_\allowbreak{}y =\allowbreak{} (piano\_\allowbreak{}bbox[1] + piano\_\allowbreak{}bbox[3]) /\allowbreak{} 2\\{}
table\_\allowbreak{}center\_\allowbreak{}x =\allowbreak{} (table\_\allowbreak{}bbox[0] + table\_\allowbreak{}bbox[2]) /\allowbreak{} 2\\{}
table\_\allowbreak{}center\_\allowbreak{}y =\allowbreak{} (table\_\allowbreak{}bbox[1] + table\_\allowbreak{}bbox[3]) /\allowbreak{} 2\\{}
\mbox{}\\{}
\# Convert center coordinates to integer for depth map indexing\\{}
piano\_\allowbreak{}center\_\allowbreak{}x\_\allowbreak{}int =\allowbreak{} int(piano\_\allowbreak{}center\_\allowbreak{}x)\\{}
piano\_\allowbreak{}center\_\allowbreak{}y\_\allowbreak{}int =\allowbreak{} int(piano\_\allowbreak{}center\_\allowbreak{}y)\\{}
table\_\allowbreak{}center\_\allowbreak{}x\_\allowbreak{}int =\allowbreak{} int(table\_\allowbreak{}center\_\allowbreak{}x)\\{}
table\_\allowbreak{}center\_\allowbreak{}y\_\allowbreak{}int =\allowbreak{} int(table\_\allowbreak{}center\_\allowbreak{}y)\\{}
\mbox{}\\{}
\# Get depth at center points (clip to bounds if needed)\\{}
\# Ensure coordinates are within depth map bounds\\{}
piano\_\allowbreak{}center\_\allowbreak{}x\_\allowbreak{}int =\allowbreak{} max(0,\allowbreak{} min(piano\_\allowbreak{}center\_\allowbreak{}x\_\allowbreak{}int,\allowbreak{} depth\_\allowbreak{}map\_\allowbreak{}1.shape[1] - 1))\\{}
piano\_\allowbreak{}center\_\allowbreak{}y\_\allowbreak{}int =\allowbreak{} max(0,\allowbreak{} min(piano\_\allowbreak{}center\_\allowbreak{}y\_\allowbreak{}int,\allowbreak{} depth\_\allowbreak{}map\_\allowbreak{}1.shape[0] - 1))\\{}
table\_\allowbreak{}center\_\allowbreak{}x\_\allowbreak{}int =\allowbreak{} max(0,\allowbreak{} min(table\_\allowbreak{}center\_\allowbreak{}x\_\allowbreak{}int,\allowbreak{} depth\_\allowbreak{}map\_\allowbreak{}2.shape[1] - 1))\\{}
table\_\allowbreak{}center\_\allowbreak{}y\_\allowbreak{}int =\allowbreak{} max(0,\allowbreak{} min(table\_\allowbreak{}center\_\allowbreak{}y\_\allowbreak{}int,\allowbreak{} depth\_\allowbreak{}map\_\allowbreak{}2.shape[0] - 1))\\{}
\mbox{}\\{}
piano\_\allowbreak{}depth =\allowbreak{} depth\_\allowbreak{}map\_\allowbreak{}1[piano\_\allowbreak{}center\_\allowbreak{}y\_\allowbreak{}int,\allowbreak{} piano\_\allowbreak{}center\_\allowbreak{}x\_\allowbreak{}int]\\{}
table\_\allowbreak{}depth =\allowbreak{} depth\_\allowbreak{}map\_\allowbreak{}2[table\_\allowbreak{}center\_\allowbreak{}y\_\allowbreak{}int,\allowbreak{} table\_\allowbreak{}center\_\allowbreak{}x\_\allowbreak{}int]\\{}
\mbox{}\\{}
\# Compute projected real-world lengths\\{}
\# Formula: real\_\allowbreak{}length =\allowbreak{} (bbox\_\allowbreak{}width /\allowbreak{} image\_\allowbreak{}width) * depth * 2\\{}
piano\_\allowbreak{}width =\allowbreak{} piano\_\allowbreak{}bbox[2] - piano\_\allowbreak{}bbox[0]\\{}
table\_\allowbreak{}width =\allowbreak{} table\_\allowbreak{}bbox[2] - table\_\allowbreak{}bbox[0]\\{}
\mbox{}\\{}
piano\_\allowbreak{}real\_\allowbreak{}length =\allowbreak{} (piano\_\allowbreak{}width /\allowbreak{} image\_\allowbreak{}width) * piano\_\allowbreak{}depth * 2\\{}
table\_\allowbreak{}real\_\allowbreak{}length =\allowbreak{} (table\_\allowbreak{}width /\allowbreak{} image\_\allowbreak{}width) * table\_\allowbreak{}depth * 2\\{}
\mbox{}\\{}
\# Compare and output results\\{}
print(f"\allowbreak{}Piano real-world length: \{piano\_\allowbreak{}real\_\allowbreak{}length:.4f\} meters"\allowbreak{})\\{}
print(f"\allowbreak{}Dining table real-world length: \{table\_\allowbreak{}real\_\allowbreak{}length:.4f\} meters"\allowbreak{})\\{}
\mbox{}\\{}
if piano\_\allowbreak{}real\_\allowbreak{}length \textgreater{} table\_\allowbreak{}real\_\allowbreak{}length:\\{}
~~~~longer\_\allowbreak{}object =\allowbreak{} "\allowbreak{}piano"\allowbreak{}\\{}
else:\\{}
~~~~longer\_\allowbreak{}object =\allowbreak{} "\allowbreak{}dining table"\allowbreak{}\\{}
\mbox{}\\{}
print(f"\allowbreak{}The \{longer\_\allowbreak{}object\} is longer."\allowbreak{})\\{}
\mbox{}\\{}
\# Assign results to variables for result namespace\\{}
piano\_\allowbreak{}real\_\allowbreak{}length =\allowbreak{} piano\_\allowbreak{}real\_\allowbreak{}length\\{}
table\_\allowbreak{}real\_\allowbreak{}length =\allowbreak{} table\_\allowbreak{}real\_\allowbreak{}length\\{}
longer\_\allowbreak{}object =\allowbreak{} longer\_\allowbreak{}object\\{}
\textasciigrave{}\textasciigrave{}\textasciigrave{}\\{}
\mbox{}\\{}
\mbox{}\\{}
\mbox{}\\{}
Execution output:\\{}
Piano real-world length: 2.2109 meters\\{}
Dining table real-world length: 2.7731 meters\\{}
The dining table is longer.
\end{exverb}
\end{exstep}
\begin{exstep}{}{\exTool{exInk}{Final answer}}{}
\exLabel{Model}{}
\exSayBlock{\textless{}thinking\textgreater{}The Python\_\allowbreak{}Coding\_\allowbreak{}Agent\_\allowbreak{}Tool computed the real-world lengths of the piano and the dining table. The piano is approximately 2.21 meters long,\allowbreak{} while the dining table (the round table under the mirror) is approximately 2.77 meters long. Therefore,\allowbreak{} the table under the mirror is longer than the piano.\textless{}/\allowbreak{}thinking\textgreater{}
\textless{}answer\textgreater{}A\textless{}/\allowbreak{}answer\textgreater{}}
\end{exstep}

%% file: text/appendix_examples_steps/spar_header.tex
\exOverview{\textbf{Task.} Three photographs of a library are taken from nearby positions. A red point marks a bookshelf in the first photo and a blue point marks another bookshelf in the third. Give the 3D distance between them in metres. 

\textbf{Why it is hard.} The two points never appear in the same photo, so the distance cannot be judged from any single view; without tools the model answers 4.0\,m. 

\textbf{How the agent solves it.} It recovers where each photo was taken from, measures metric depth at each point, and has the coding agent place both points in one 3D frame and measure between them: 2.0\,m, against a ground truth of 1.9\,m.}
\exLabel{Input images}{}
\begin{center}
\begin{minipage}[t]{0.312\linewidth}\centering\includegraphics[width=\linewidth,height=0.22\textheight,keepaspectratio]{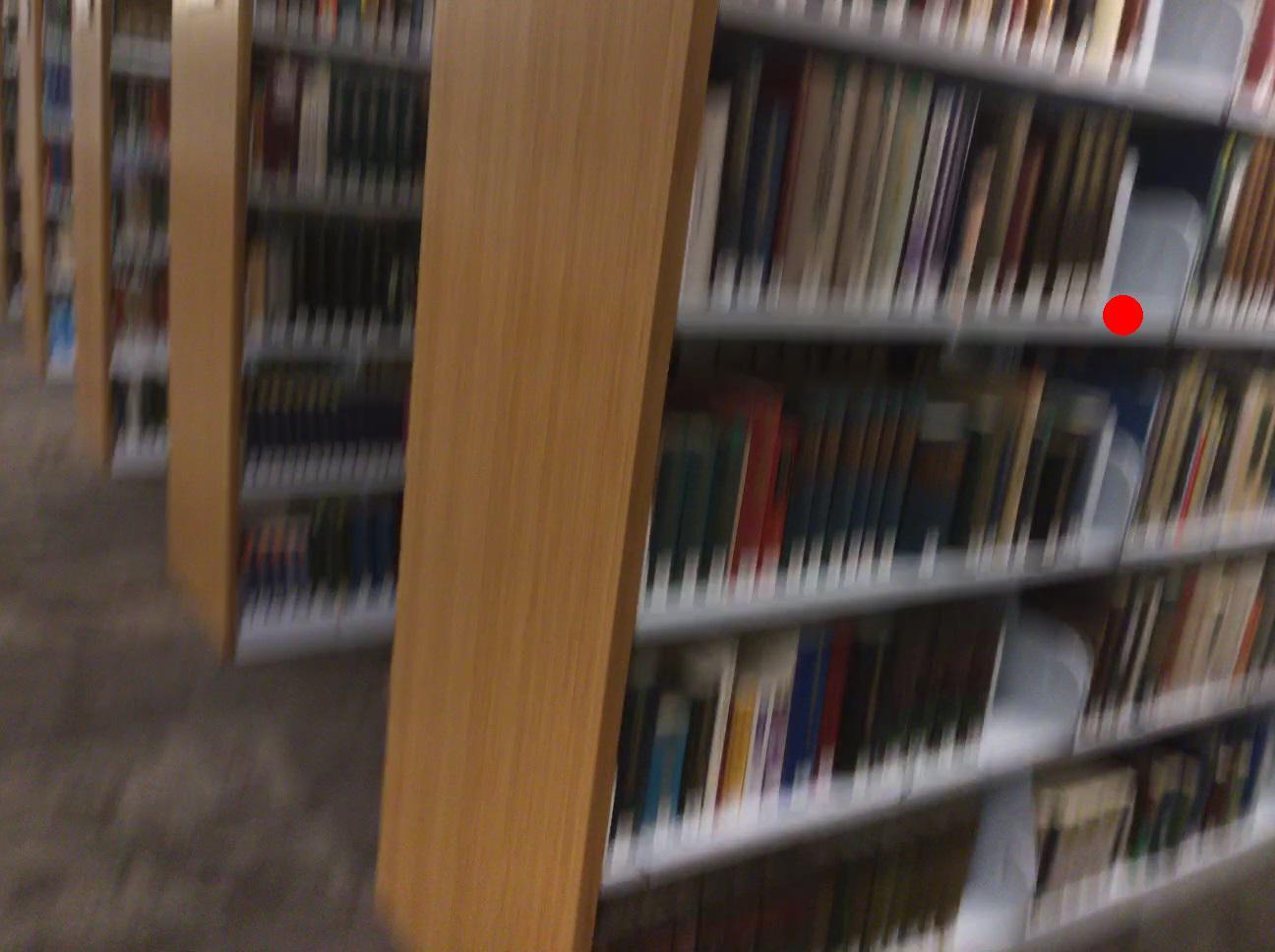}\\[1pt]{\scriptsize\color{exMute} input\_image\_1. The red point marks the first bookshelf.}\end{minipage}\hfill\begin{minipage}[t]{0.312\linewidth}\centering\includegraphics[width=\linewidth,height=0.22\textheight,keepaspectratio]{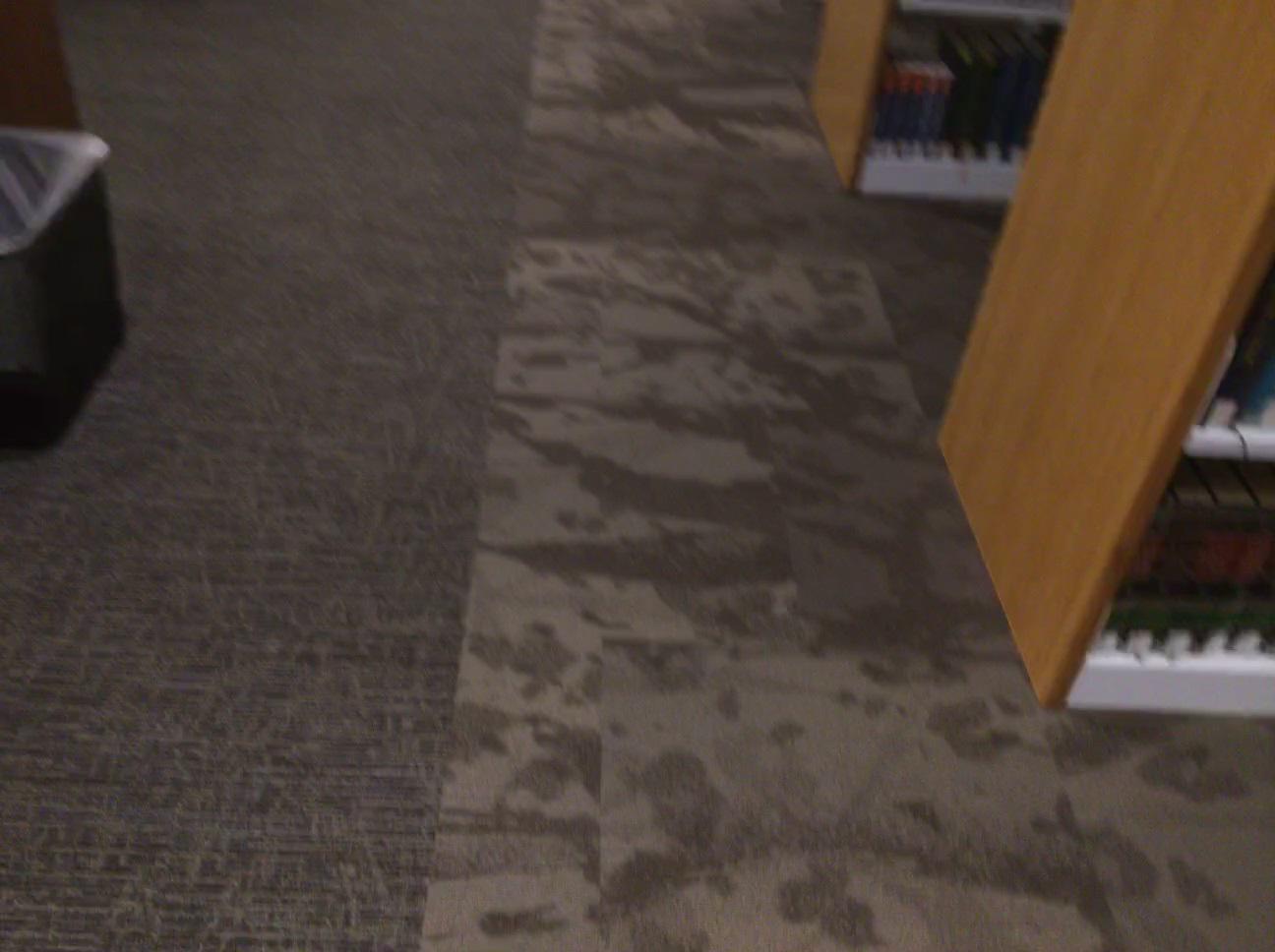}\\[1pt]{\scriptsize\color{exMute} input\_image\_2}\end{minipage}\hfill\begin{minipage}[t]{0.312\linewidth}\centering\includegraphics[width=\linewidth,height=0.22\textheight,keepaspectratio]{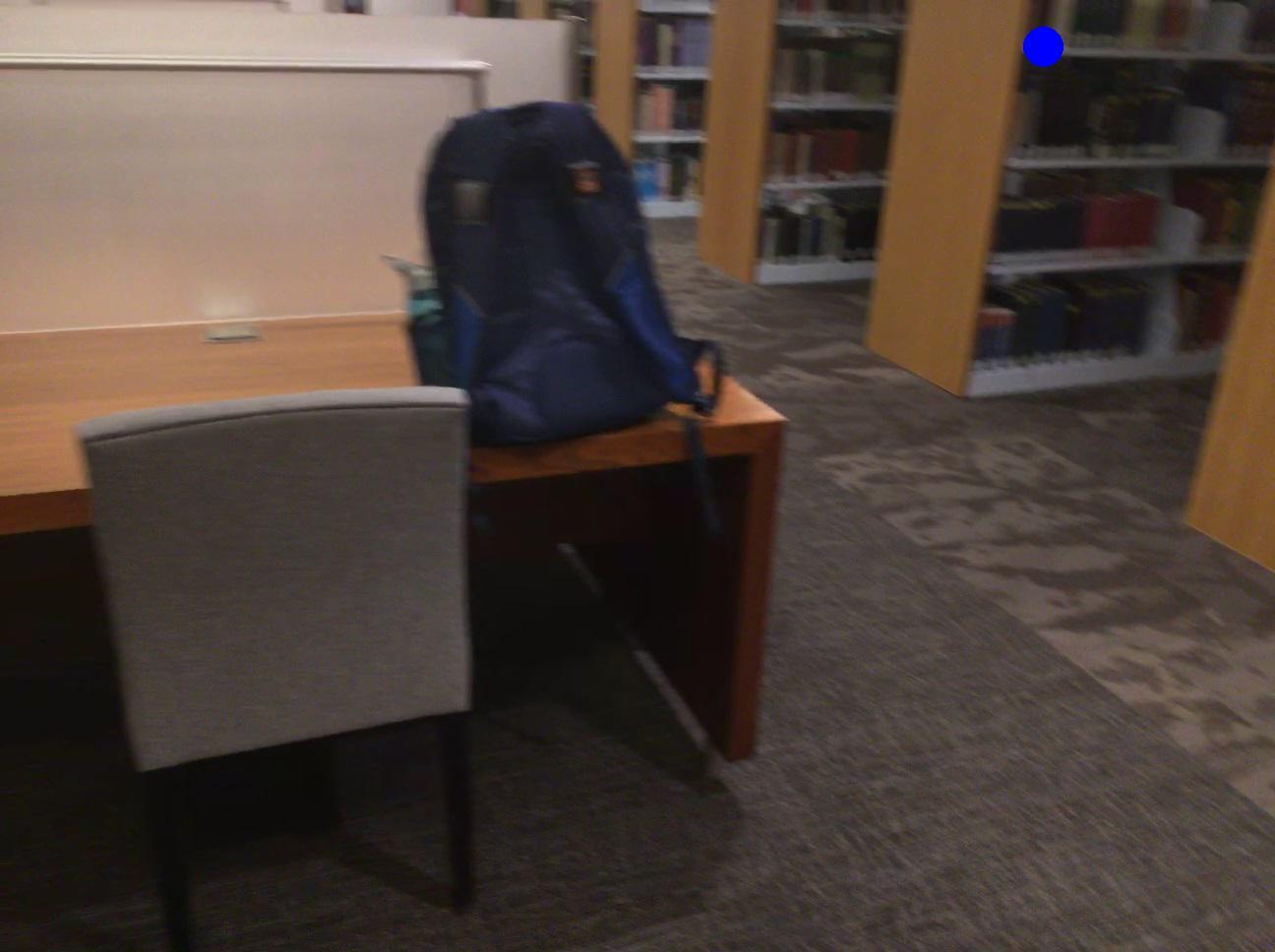}\\[1pt]{\scriptsize\color{exMute} input\_image\_3. The blue point marks the second bookshelf.}\end{minipage}
\end{center}
\exLabel{Question}{as sent to the model}
\begin{exverb}
Question: \\{}
State the Euclidean distance in meters from the center of bookshelf (red point) to the center of bookshelf (blue point). Calculate or judge based on the 3D center points of these objects. Provide only one number for your response.\\{}
Please answer the question using a single word or phrase.
\end{exverb}
\exLabel{Without tools}{GPT-6 Sol's full response to the same question}
\begin{exverb}
4.0
\end{exverb}

%% file: text/appendix_examples_steps/spar.tex
\begin{exharness}
\exLabel{Harness}{environment update, shown in full}
\begin{exverb}
Environment Update (not a new question):\\{}
Currently available image IDs for tool calls:\\{}
- input\_\allowbreak{}image\_\allowbreak{}1: user\_\allowbreak{}input\\{}
- input\_\allowbreak{}image\_\allowbreak{}2: user\_\allowbreak{}input\\{}
- input\_\allowbreak{}image\_\allowbreak{}3: user\_\allowbreak{}input\\{}
When a tool needs image input,\allowbreak{} pass one of these IDs via the *\_\allowbreak{}image\_\allowbreak{}id field.\\{}
Do NOT invent image file names or paths.
\end{exverb}
\end{exharness}
\begin{exstep}{1}{\toolCam}{}
\exWalk{Camera trajectory estimates each camera's position and orientation relative to the first photo. The third photo was taken about 1.5\,m to the left and 1.65\,m behind the first. The full camera parameters are kept as a data binding for code.}
\exLabel{Arguments}{}
\begin{exverb}
\{\\{}
~~"\allowbreak{}image\_\allowbreak{}ids"\allowbreak{}: ["\allowbreak{}input\_\allowbreak{}image\_\allowbreak{}1"\allowbreak{},\allowbreak{} "\allowbreak{}input\_\allowbreak{}image\_\allowbreak{}2"\allowbreak{},\allowbreak{} "\allowbreak{}input\_\allowbreak{}image\_\allowbreak{}3"\allowbreak{}]\\{}
\}
\end{exverb}
\exLabel{Tool output}{}
\begin{exverb}
Camera trajectory summary:\\{}
- input\_\allowbreak{}image\_\allowbreak{}1: reference camera and reference view.\\{}
- input\_\allowbreak{}image\_\allowbreak{}2:\\{}
~~~~camera movement: salient direction: none (all movements \textless{} 1 meter); detailed: right (+0.01),\allowbreak{} down (+0.16),\allowbreak{} backward (-0.08)\\{}
~~~~camera rotation: salient direction: left (yaw -27.4°),\allowbreak{} down (pitch +19.8°); detailed: left (yaw -27.4°),\allowbreak{} down (pitch +19.8°)\\{}
- input\_\allowbreak{}image\_\allowbreak{}3:\\{}
~~~~camera movement: salient direction: left (-1.49),\allowbreak{} backward (-1.65); detailed: left (-1.49),\allowbreak{} down (+0.27),\allowbreak{} backward (-1.65)\\{}
~~~~camera rotation: salient direction: left (yaw -12.1°); detailed: left (yaw -12.1°),\allowbreak{} down (pitch +6.1°)\\{}
If no salient movement or rotation is reported,\allowbreak{} use the magnitude of each axis to determine the most possible movement/\allowbreak{}rotation direction.\\{}
\mbox{}\\{}
Bird\textquotesingle{}s-eye view of the camera trajectory on input\_\allowbreak{}image\_\allowbreak{}1\textquotesingle{}s X-Z plane is rendered in the following image - refer as tool\_\allowbreak{}generated\_\allowbreak{}image\_\allowbreak{}1 (dots =\allowbreak{} camera positions,\allowbreak{} forward-pointing arrows =\allowbreak{} facing direction).\\{}
\mbox{}\\{}
All vectors are relative to input\_\allowbreak{}image\_\allowbreak{}1\textquotesingle{}s camera frame,\allowbreak{} using the coordinate system: +X =\allowbreak{} right,\allowbreak{} +Y =\allowbreak{} down,\allowbreak{} +Z =\allowbreak{} forward. With this convention,\allowbreak{} positive yaw (rotation around Y axis) =\allowbreak{} right and positive pitch (rotation around X axis) =\allowbreak{} down. If the original question assumes a different coordinate system (e.g. with +Y =\allowbreak{} up and -Z =\allowbreak{} forward,\allowbreak{} which flips the yaw/\allowbreak{}pitch signs),\allowbreak{} convert before answering.
\end{exverb}
\begin{center}\includegraphics[width=0.45\linewidth,height=0.22\textheight,keepaspectratio]{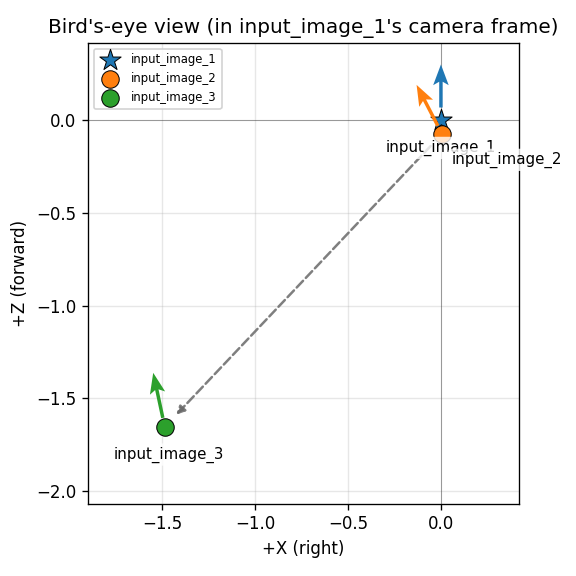}\\[2pt]
{\scriptsize\color{exMute} Bird's-eye view of the three camera positions in input\_image\_1's frame.}\end{center}
\begin{exharness}
\exLabel{Harness}{environment update, summarized: new memory entries; new data bindings verbatim}
\begin{exverb}
Memory update: registers 1 new image in memory (tool\_\allowbreak{}generated\_\allowbreak{}image\_\allowbreak{}1,\allowbreak{} from Camera\_\allowbreak{}Trajectory\_\allowbreak{}Tool).\\{}
Available data bindings (auto-injected into Python\_\allowbreak{}Coding\_\allowbreak{}Agent\_\allowbreak{}Tool):\\{}
- camera\_\allowbreak{}trajectory\_\allowbreak{}result: dict (keys: intrinsics,\allowbreak{} extrinsics,\allowbreak{} image\_\allowbreak{}ids,\allowbreak{} num\_\allowbreak{}images,\allowbreak{} cameras,\allowbreak{} coordinate\_\allowbreak{}system,\allowbreak{} reference\_\allowbreak{}image\_\allowbreak{}id)
\end{exverb}
\end{exharness}
\end{exstep}
\begin{exstep}{2}{\toolDepth}{}
\exWalk{Metric depth for the first photo, where the red point is.}
\exLabel{Arguments}{}
\begin{exverb}
\{\\{}
~~"\allowbreak{}image\_\allowbreak{}id"\allowbreak{}: "\allowbreak{}input\_\allowbreak{}image\_\allowbreak{}1"\allowbreak{},\allowbreak{}\\{}
~~"\allowbreak{}colormap"\allowbreak{}: "\allowbreak{}inferno"\allowbreak{}\\{}
\}
\end{exverb}
\exLabel{Tool output}{}
\begin{exverb}
Tool \textasciigrave{}Depth\_\allowbreak{}Estimation\_\allowbreak{}Tool\textasciigrave{} (tool\_\allowbreak{}call\_\allowbreak{}id=\allowbreak{}call\_\allowbreak{}qrfQNTtmO9TNFVBT4xdZiRyL) returned:\\{}
Depth estimation complete. Shape: [966,\allowbreak{} 1302],\allowbreak{} colormap: inferno.\\{}
Metric depth in meters available as \textasciigrave{}depth\_\allowbreak{}meters\textasciigrave{} (float32 ndarray).\\{}
Colorized depth map rendered below; refer to it as tool\_\allowbreak{}generated\_\allowbreak{}image\_\allowbreak{}2.
\end{exverb}
\begin{center}\includegraphics[width=0.45\linewidth,height=0.22\textheight,keepaspectratio]{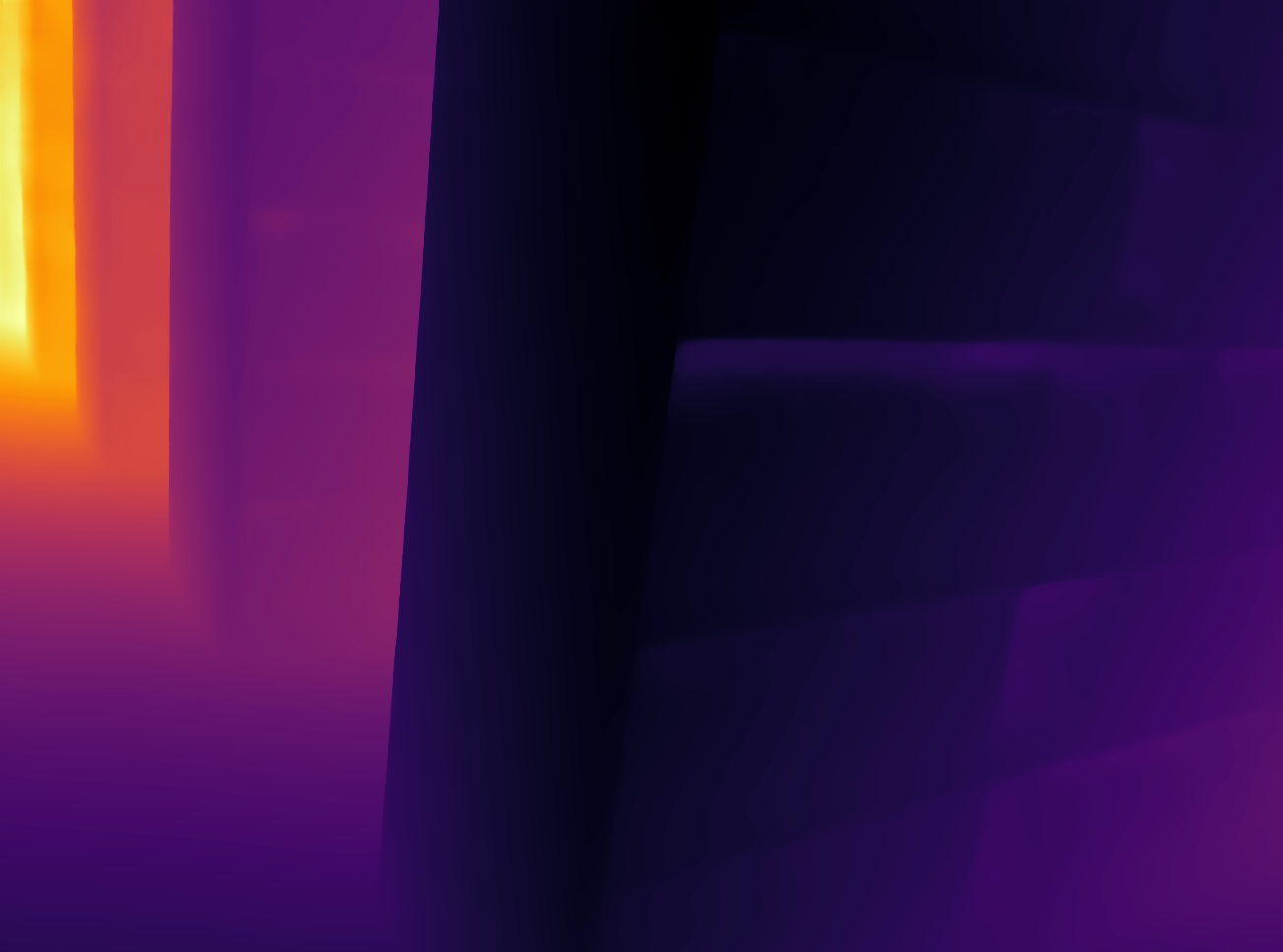}\\[2pt]
{\scriptsize\color{exMute} Colorized depth map of input\_image\_1.}\end{center}
\begin{exharness}
\exLabel{Harness}{environment update, summarized: new memory entries; new data bindings verbatim}
\begin{exverb}
Memory update: registers 1 new image in memory (tool\_\allowbreak{}generated\_\allowbreak{}image\_\allowbreak{}2,\allowbreak{} from Depth\_\allowbreak{}Estimation\_\allowbreak{}Tool).\\{}
- depth\_\allowbreak{}result: dict (keys: depth\_\allowbreak{}shape,\allowbreak{} colormap,\allowbreak{} size,\allowbreak{} depth\_\allowbreak{}meters)
\end{exverb}
\end{exharness}
\end{exstep}
\begin{exstep}{3}{\toolDepth}{}
\exWalk{Metric depth for the third photo, where the blue point is.}
\exLabel{Arguments}{}
\begin{exverb}
\{\\{}
~~"\allowbreak{}image\_\allowbreak{}id"\allowbreak{}: "\allowbreak{}input\_\allowbreak{}image\_\allowbreak{}3"\allowbreak{},\allowbreak{}\\{}
~~"\allowbreak{}colormap"\allowbreak{}: "\allowbreak{}inferno"\allowbreak{}\\{}
\}
\end{exverb}
\exLabel{Tool output}{}
\begin{exverb}
Tool \textasciigrave{}Depth\_\allowbreak{}Estimation\_\allowbreak{}Tool\textasciigrave{} (tool\_\allowbreak{}call\_\allowbreak{}id=\allowbreak{}call\_\allowbreak{}YwF06eDGu6x1VabL7jtqFml7) returned:\\{}
Depth estimation complete. Shape: [966,\allowbreak{} 1302],\allowbreak{} colormap: inferno.\\{}
Metric depth in meters available as \textasciigrave{}depth\_\allowbreak{}meters\textasciigrave{} (float32 ndarray).\\{}
Colorized depth map rendered below; refer to it as tool\_\allowbreak{}generated\_\allowbreak{}image\_\allowbreak{}3.
\end{exverb}
\begin{center}\includegraphics[width=0.45\linewidth,height=0.22\textheight,keepaspectratio]{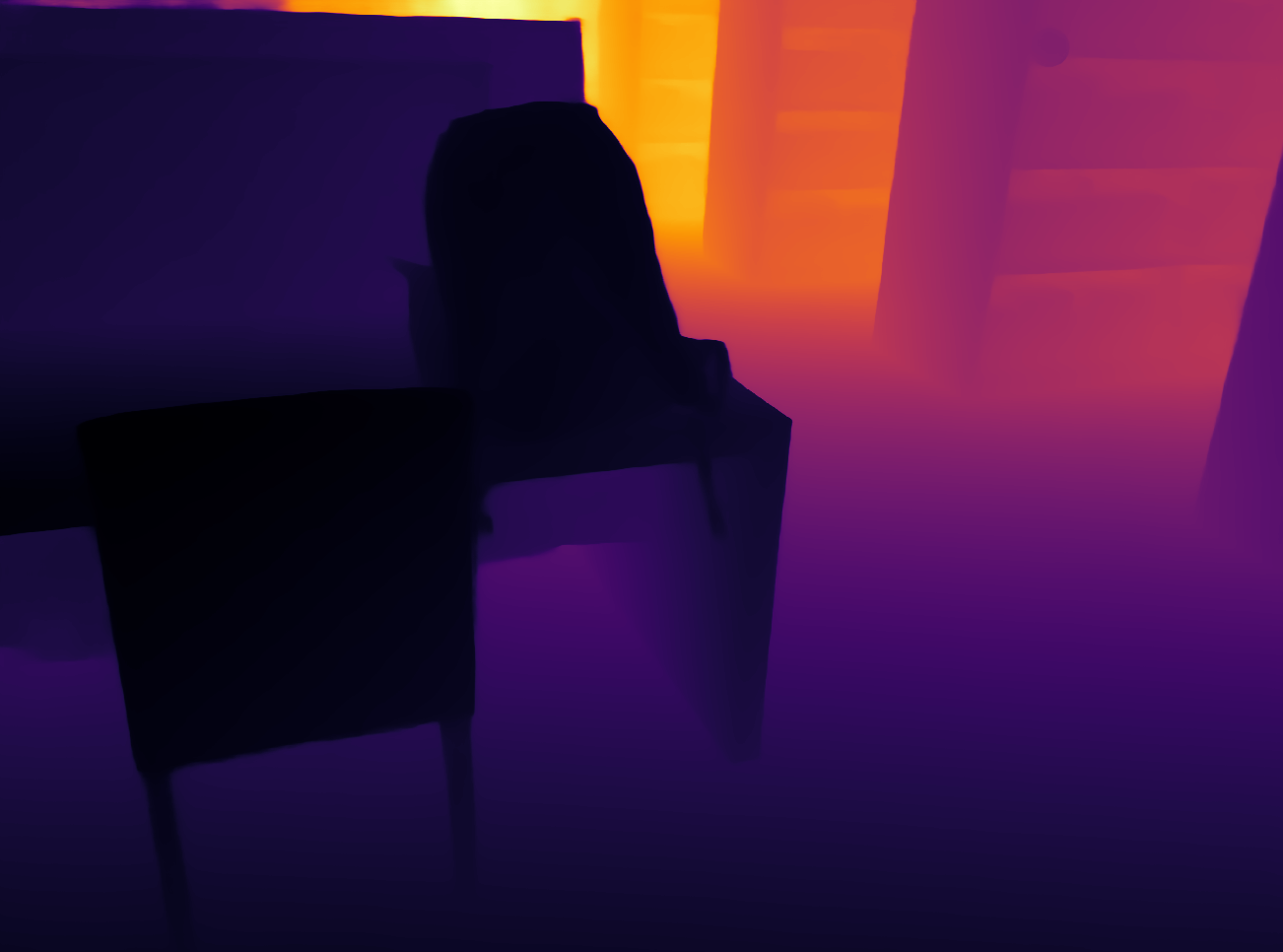}\\[2pt]
{\scriptsize\color{exMute} Colorized depth map of input\_image\_3.}\end{center}
\begin{exharness}
\exLabel{Harness}{environment update, summarized: new memory entries; new data bindings verbatim}
\begin{exverb}
Memory update: registers 1 new image in memory (tool\_\allowbreak{}generated\_\allowbreak{}image\_\allowbreak{}3,\allowbreak{} from Depth\_\allowbreak{}Estimation\_\allowbreak{}Tool).\\{}
- depth\_\allowbreak{}result\_\allowbreak{}2: dict (keys: depth\_\allowbreak{}shape,\allowbreak{} colormap,\allowbreak{} size,\allowbreak{} depth\_\allowbreak{}meters)
\end{exverb}
\end{exharness}
\end{exstep}
\begin{exstep}{4}{\toolCode}{}
\exWalk{The agent gives the coding agent the pixel positions of the two points, which it read off the images itself. The code reads the depth at each point (2.15\,m and 4.08\,m), turns each point into 3D coordinates using its camera's parameters, and measures the distance between them: 2.0076\,m.}
\exLabel{Arguments}{}
\begin{exverb}
\{\\{}
~~"\allowbreak{}objective"\allowbreak{}: "\allowbreak{}Estimate Euclidean distance between red dot at pixel (1141,\allowbreak{}320) in first 1302x966 image and blue dot at pixel (1060,\allowbreak{}47) in third image using respective estimated depth maps,\allowbreak{} camera intrinsics and world-to-camera extrinsics. Print intrinsics,\allowbreak{} camera transforms,\allowbreak{} depth local median at each point,\allowbreak{} derived 3D world coordinates and distance. Be careful depth\_\allowbreak{}result is first image and depth\_\allowbreak{}result\_\allowbreak{}2 is third."\allowbreak{}\\{}
\}
\end{exverb}
\exLabel{Tool output}{}
\begin{exverb}
Coding agent generated and executed the following Python code:\\{}
\textasciigrave{}\textasciigrave{}\textasciigrave{}python\\{}
import numpy as np\\{}
\mbox{}\\{}
\# Extract relevant data\\{}
image\_\allowbreak{}ids =\allowbreak{} camera\_\allowbreak{}trajectory\_\allowbreak{}result["\allowbreak{}image\_\allowbreak{}ids"\allowbreak{}]\\{}
num\_\allowbreak{}images =\allowbreak{} camera\_\allowbreak{}trajectory\_\allowbreak{}result["\allowbreak{}num\_\allowbreak{}images"\allowbreak{}]\\{}
intrinsics =\allowbreak{} camera\_\allowbreak{}trajectory\_\allowbreak{}result["\allowbreak{}intrinsics"\allowbreak{}]  \# shape: (3,\allowbreak{} 3,\allowbreak{} 3)\\{}
extrinsics =\allowbreak{} camera\_\allowbreak{}trajectory\_\allowbreak{}result["\allowbreak{}extrinsics"\allowbreak{}]  \# shape: (3,\allowbreak{} 3,\allowbreak{} 4)\\{}
depth\_\allowbreak{}meters\_\allowbreak{}1 =\allowbreak{} depth\_\allowbreak{}result["\allowbreak{}depth\_\allowbreak{}meters"\allowbreak{}]  \# first image depth map\\{}
depth\_\allowbreak{}meters\_\allowbreak{}3 =\allowbreak{} depth\_\allowbreak{}result\_\allowbreak{}2["\allowbreak{}depth\_\allowbreak{}meters"\allowbreak{}]  \# third image depth map\\{}
\mbox{}\\{}
\# Pixel coordinates (note: depth\_\allowbreak{}result is for first image,\allowbreak{} depth\_\allowbreak{}result\_\allowbreak{}2 for third)\\{}
\# Pixel (1141,\allowbreak{} 320) in first image (depth\_\allowbreak{}result)\\{}
\# Pixel (1060,\allowbreak{} 47) in third image (depth\_\allowbreak{}result\_\allowbreak{}2)\\{}
\# Note: image dimensions are [height,\allowbreak{} width] =\allowbreak{} [966,\allowbreak{} 1302] per depth\_\allowbreak{}result\\{}
\mbox{}\\{}
\# Convert pixel coordinates to 0-based indexing (if needed,\allowbreak{} but usually they are 0-based)\\{}
x1,\allowbreak{} y1 =\allowbreak{} 1141,\allowbreak{} 320  \# first image\\{}
x3,\allowbreak{} y3 =\allowbreak{} 1060,\allowbreak{} 47   \# third image\\{}
\mbox{}\\{}
\# Get depth at each pixel\\{}
depth\_\allowbreak{}1 =\allowbreak{} depth\_\allowbreak{}meters\_\allowbreak{}1[y1,\allowbreak{} x1]  \# depth at pixel (1141,\allowbreak{} 320) in first image\\{}
depth\_\allowbreak{}3 =\allowbreak{} depth\_\allowbreak{}meters\_\allowbreak{}3[y3,\allowbreak{} x3]  \# depth at pixel (1060,\allowbreak{} 47) in third image\\{}
\mbox{}\\{}
\# Compute local median depth for each point (for robustness,\allowbreak{} though not strictly needed for distance)\\{}
depth\_\allowbreak{}median\_\allowbreak{}1 =\allowbreak{} np.median(depth\_\allowbreak{}meters\_\allowbreak{}1[y1-1:y1+2,\allowbreak{} x1-1:x1+2])  \# 3x3 neighborhood\\{}
depth\_\allowbreak{}median\_\allowbreak{}3 =\allowbreak{} np.median(depth\_\allowbreak{}meters\_\allowbreak{}3[y3-1:y3+2,\allowbreak{} x3-1:x3+2])  \# 3x3 neighborhood\\{}
\mbox{}\\{}
\# Print intrinsics (first image)\\{}
print("\allowbreak{}Intrinsics (first image):"\allowbreak{})\\{}
print(intrinsics[0])\\{}
\mbox{}\\{}
\# Print camera transforms (first and third images)\\{}
print("\allowbreak{}\\{}
Camera transforms (first image):"\allowbreak{})\\{}
print(extrinsics[0])\\{}
print("\allowbreak{}\\{}
Camera transforms (third image):"\allowbreak{})\\{}
print(extrinsics[2])\\{}
\mbox{}\\{}
\# Print depth local medians\\{}
print(f"\allowbreak{}\\{}
Depth local median at pixel (1141,\allowbreak{} 320) in first image: \{depth\_\allowbreak{}median\_\allowbreak{}1:.4f\} meters"\allowbreak{})\\{}
print(f"\allowbreak{}Depth local median at pixel (1060,\allowbreak{} 47) in third image: \{depth\_\allowbreak{}median\_\allowbreak{}3:.4f\} meters"\allowbreak{})\\{}
\mbox{}\\{}
\# Convert 2D pixel coordinates to 3D world coordinates using camera model\\{}
\# For first image (image index 0)\\{}
K =\allowbreak{} intrinsics[0]  \# 3x3 intrinsic matrix\\{}
R1,\allowbreak{} t1 =\allowbreak{} extrinsics[0][:3,\allowbreak{} :3],\allowbreak{} extrinsics[0][:3,\allowbreak{} 3]  \# rotation and translation (world to camera)\\{}
\mbox{}\\{}
\# Convert pixel to normalized image coordinates (assuming 0-based indexing)\\{}
u1,\allowbreak{} v1 =\allowbreak{} x1,\allowbreak{} y1\\{}
\# Apply intrinsic matrix to get normalized coordinates\\{}
x\_\allowbreak{}norm1 =\allowbreak{} (u1 - K[0,\allowbreak{} 2]) /\allowbreak{} K[0,\allowbreak{} 0]  \# (u - cx) /\allowbreak{} fx\\{}
y\_\allowbreak{}norm1 =\allowbreak{} (v1 - K[1,\allowbreak{} 2]) /\allowbreak{} K[1,\allowbreak{} 1]  \# (v - cy) /\allowbreak{} fy\\{}
\mbox{}\\{}
\# Convert normalized image coordinates to 3D camera coordinates\\{}
\# [Xc,\allowbreak{} Yc,\allowbreak{} Zc] =\allowbreak{} depth * [x\_\allowbreak{}norm,\allowbreak{} y\_\allowbreak{}norm,\allowbreak{} 1]\\{}
Xc1 =\allowbreak{} depth\_\allowbreak{}median\_\allowbreak{}1 * x\_\allowbreak{}norm1\\{}
Yc1 =\allowbreak{} depth\_\allowbreak{}median\_\allowbreak{}1 * y\_\allowbreak{}norm1\\{}
Zc1 =\allowbreak{} depth\_\allowbreak{}median\_\allowbreak{}1\\{}
\mbox{}\\{}
\# Convert camera coordinates to world coordinates\\{}
\# Xw =\allowbreak{} R\^{}T * (Xc - t)\\{}
Xw1 =\allowbreak{} np.dot(R1.T,\allowbreak{} np.array([Xc1,\allowbreak{} Yc1,\allowbreak{} Zc1]) - t1)\\{}
Xw1 =\allowbreak{} Xw1.flatten()  \# ensure 1D\\{}
\mbox{}\\{}
\# For third image (image index 2)\\{}
K =\allowbreak{} intrinsics[2]  \# 3x3 intrinsic matrix\\{}
R3,\allowbreak{} t3 =\allowbreak{} extrinsics[2][:3,\allowbreak{} :3],\allowbreak{} extrinsics[2][:3,\allowbreak{} 3]  \# rotation and translation (world to camera)\\{}
\mbox{}\\{}
\# Convert pixel to normalized image coordinates\\{}
u3,\allowbreak{} v3 =\allowbreak{} x3,\allowbreak{} y3\\{}
x\_\allowbreak{}norm3 =\allowbreak{} (u3 - K[0,\allowbreak{} 2]) /\allowbreak{} K[0,\allowbreak{} 0]\\{}
y\_\allowbreak{}norm3 =\allowbreak{} (v3 - K[1,\allowbreak{} 2]) /\allowbreak{} K[1,\allowbreak{} 1]\\{}
\mbox{}\\{}
\# Convert normalized image coordinates to 3D camera coordinates\\{}
Xc3 =\allowbreak{} depth\_\allowbreak{}median\_\allowbreak{}3 * x\_\allowbreak{}norm3\\{}
Yc3 =\allowbreak{} depth\_\allowbreak{}median\_\allowbreak{}3 * y\_\allowbreak{}norm3\\{}
Zc3 =\allowbreak{} depth\_\allowbreak{}median\_\allowbreak{}3\\{}
\mbox{}\\{}
\# Convert camera coordinates to world coordinates\\{}
Xw3 =\allowbreak{} np.dot(R3.T,\allowbreak{} np.array([Xc3,\allowbreak{} Yc3,\allowbreak{} Zc3]) - t3)\\{}
Xw3 =\allowbreak{} Xw3.flatten()  \# ensure 1D\\{}
\mbox{}\\{}
\# Print derived 3D world coordinates\\{}
print(f"\allowbreak{}\\{}
3D world coordinates (first image point): (\{Xw1[0]:.4f\},\allowbreak{} \{Xw1[1]:.4f\},\allowbreak{} \{Xw1[2]:.4f\})"\allowbreak{})\\{}
print(f"\allowbreak{}3D world coordinates (third image point): (\{Xw3[0]:.4f\},\allowbreak{} \{Xw3[1]:.4f\},\allowbreak{} \{Xw3[2]:.4f\})"\allowbreak{})\\{}
\mbox{}\\{}
\# Compute Euclidean distance between the two 3D points\\{}
distance =\allowbreak{} np.linalg.norm(Xw1 - Xw3)\\{}
print(f"\allowbreak{}\\{}
Euclidean distance between the two points: \{distance:.4f\} meters"\allowbreak{})\\{}
\textasciigrave{}\textasciigrave{}\textasciigrave{}\\{}
\mbox{}\\{}
\mbox{}\\{}
\mbox{}\\{}
Execution output:\\{}
Intrinsics (first image):\\{}
[[1.2203016e+03 0.0000000e+00 6.5100000e+02]\\{}
~[0.0000000e+00 1.3876421e+03 4.8300000e+02]\\{}
~[0.0000000e+00 0.0000000e+00 1.0000000e+00]]\\{}
\mbox{}\\{}
Camera transforms (first image):\\{}
[[ 0.8591257   0.2705432  -0.43440697  0.01150304]\\{}
~[-0.12202131  0.93264997  0.33952138  0.15892358]\\{}
~[ 0.49700487 -0.23868464  0.8342756  -0.07446227]]\\{}
\mbox{}\\{}
Camera transforms (third image):\\{}
[[ 0.941648    0.22854781 -0.24711327  1.7913343 ]\\{}
~[-0.16772765  0.9551071   0.24420871 -0.2721306 ]\\{}
~[ 0.29183298 -0.18851092  0.93770844  1.2118572 ]]\\{}
\mbox{}\\{}
Depth local median at pixel (1141,\allowbreak{} 320) in first image: 2.1547 meters\\{}
Depth local median at pixel (1060,\allowbreak{} 47) in third image: 4.0840 meters\\{}
\mbox{}\\{}
3D world coordinates (first image point): (1.8916,\allowbreak{} -0.6854,\allowbreak{} 1.3490)\\{}
3D world coordinates (third image point): (0.5929,\allowbreak{} -1.6351,\allowbreak{} 2.5497)\\{}
\mbox{}\\{}
Euclidean distance between the two points: 2.0076 meters
\end{exverb}
\end{exstep}
\begin{exstep}{}{\exTool{exInk}{Final answer}}{}
\exWalk{The agent rounds the computed 2.0076\,m to 2.0.}
\exLabel{Model}{}
\begin{exverb}
\textless{}answer\textgreater{}2.0\textless{}/\allowbreak{}answer\textgreater{}
\end{exverb}
\end{exstep}

%% file: text/appendix_examples_steps/cab_header.tex
\exOverview{\textbf{Task.} Three photographs of a basement storage area are taken from nearby positions; the first is the observer's view. A red point marks a cabinet, visible only in the third photo. Give the distance from the observer to the cabinet in metres. 

\textbf{Why it is hard.} The cabinet is seen from a different position than the one the question asks about, so its depth in the third photo is not the answer; without tools the model answers 2.0\,m. 

\textbf{How the agent solves it.} It recovers the camera positions, reads the cabinet's depth in the third photo, and has the coding agent move that point into the observer's frame: 2.40\,m, exactly the ground truth.}
\exLabel{Input images}{}
\begin{center}
\begin{minipage}[t]{0.312\linewidth}\centering\includegraphics[width=\linewidth,height=0.22\textheight,keepaspectratio]{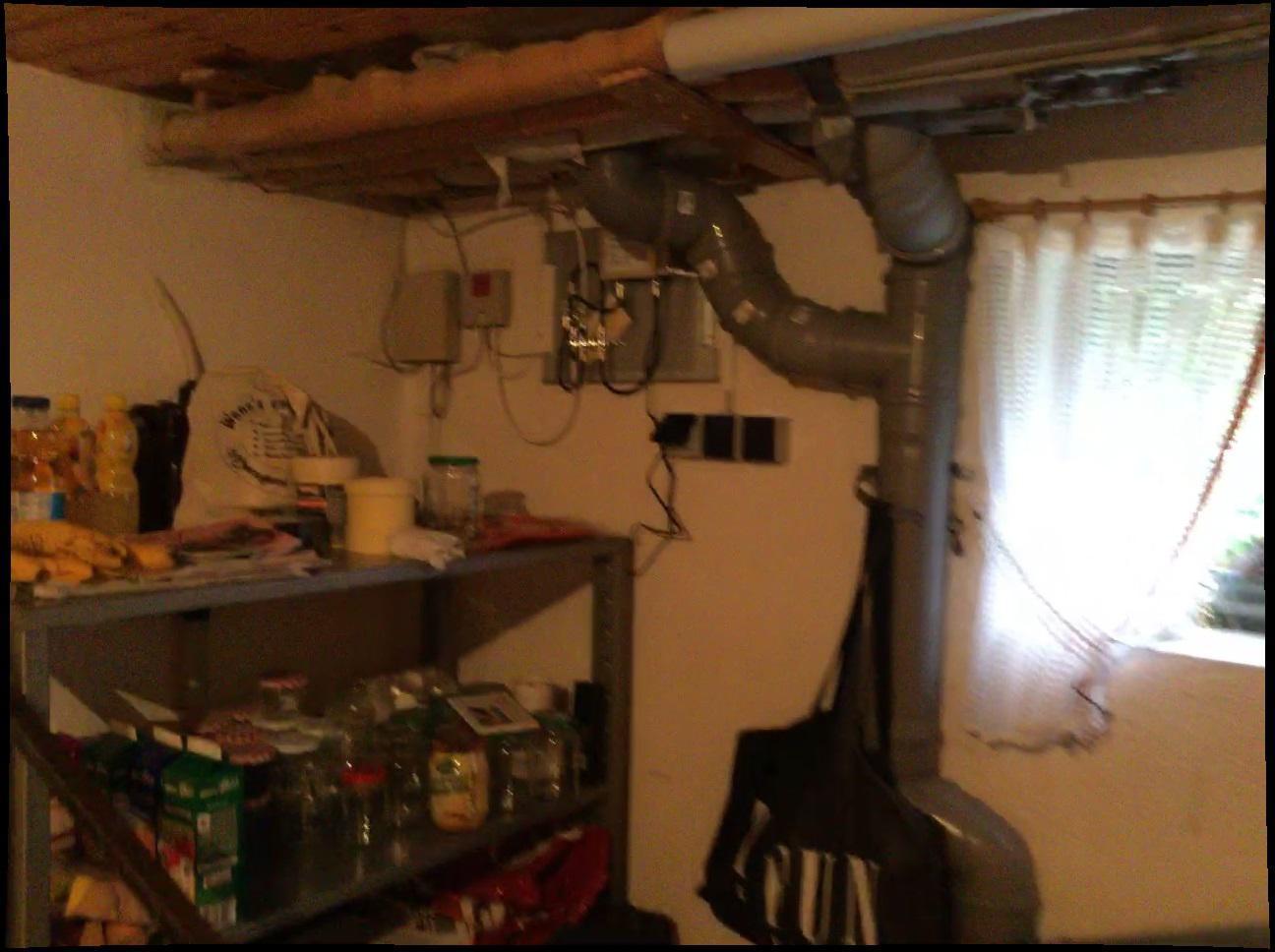}\\[1pt]{\scriptsize\color{exMute} input\_image\_1. The observer's view.}\end{minipage}\hfill\begin{minipage}[t]{0.312\linewidth}\centering\includegraphics[width=\linewidth,height=0.22\textheight,keepaspectratio]{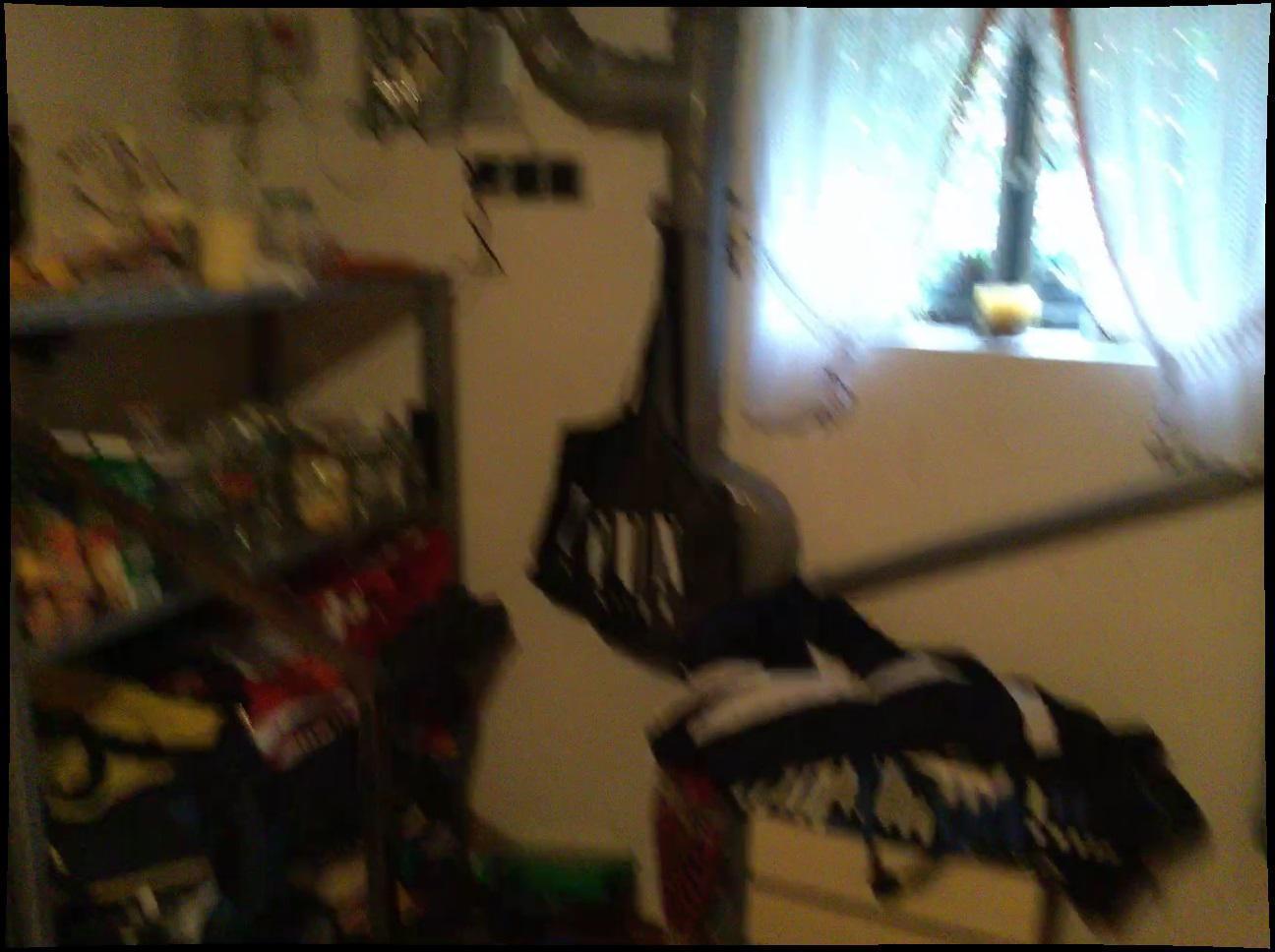}\\[1pt]{\scriptsize\color{exMute} input\_image\_2}\end{minipage}\hfill\begin{minipage}[t]{0.312\linewidth}\centering\includegraphics[width=\linewidth,height=0.22\textheight,keepaspectratio]{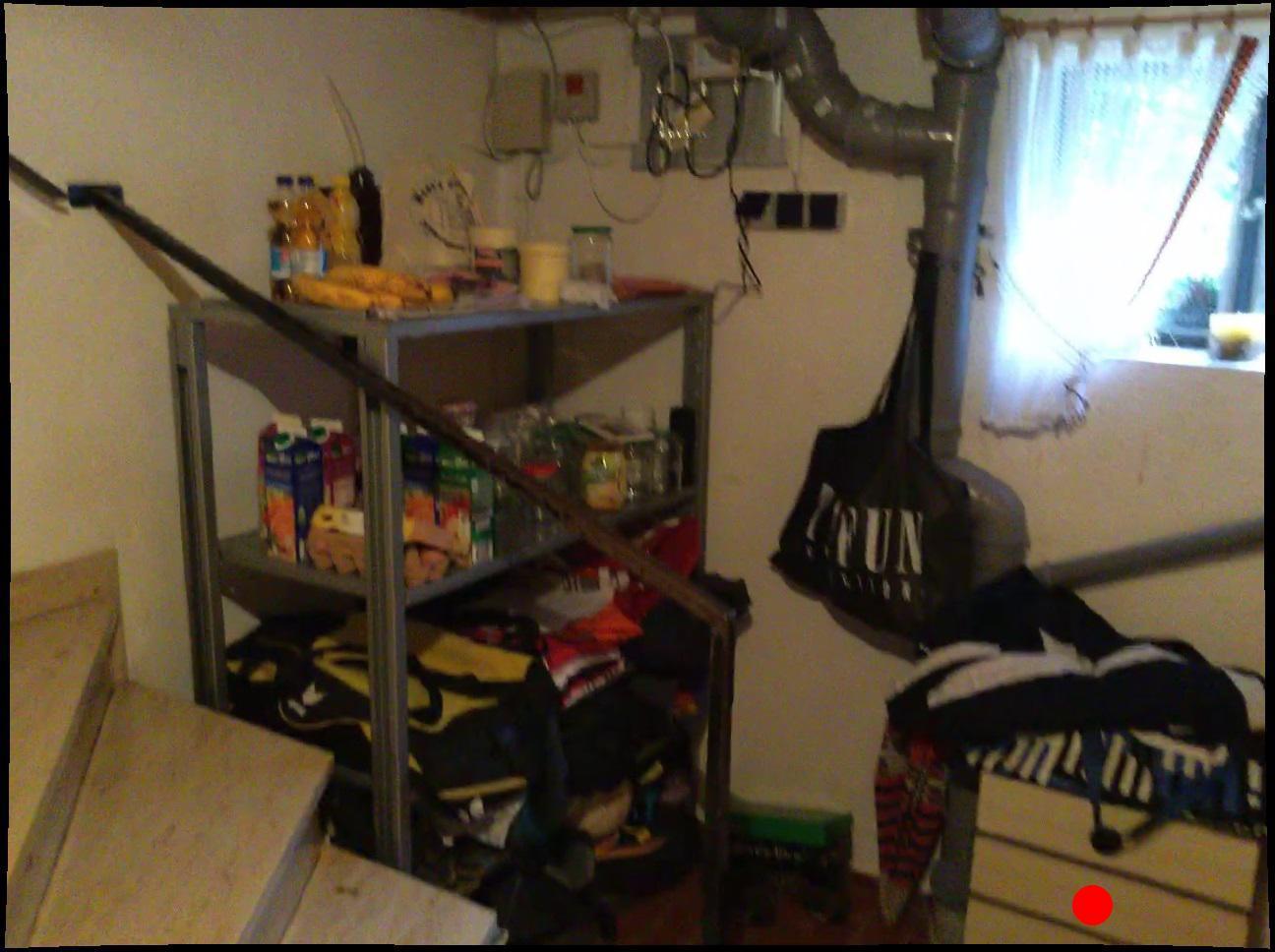}\\[1pt]{\scriptsize\color{exMute} input\_image\_3. The red point marks the cabinet, which appears only in this view.}\end{minipage}
\end{center}
\exLabel{Question}{as sent to the model}
\begin{exverb}
Question: \\{}
From the person’s position who took the photo,\allowbreak{} what is the distance to the center of cabinet (red point)? Express it in meters. Calculate or judge based on the 3D center points of these objects. The first image becomes the primary perspective,\allowbreak{} correlating with the observer\textquotesingle{}s view. Only one number should be included in your response.\\{}
Please answer the question using a single word or phrase.
\end{exverb}
\exLabel{Without tools}{GPT-6 Luna's full response to the same question}
\begin{exverb}
2.0 m
\end{exverb}

%% file: text/appendix_examples_steps/cab.tex
\begin{exharness}
\exLabel{Harness}{environment update, shown in full}
\begin{exverb}
Environment Update (not a new question):\\{}
Currently available image IDs for tool calls:\\{}
- input\_\allowbreak{}image\_\allowbreak{}1: user\_\allowbreak{}input\\{}
- input\_\allowbreak{}image\_\allowbreak{}2: user\_\allowbreak{}input\\{}
- input\_\allowbreak{}image\_\allowbreak{}3: user\_\allowbreak{}input\\{}
When a tool needs image input,\allowbreak{} pass one of these IDs via the *\_\allowbreak{}image\_\allowbreak{}id field.\\{}
Do NOT invent image file names or paths.
\end{exverb}
\end{exharness}
\begin{exstep}{1}{\toolCam}{}
\exWalk{Camera trajectory recovers where each photo was taken. The third was taken about 0.7\,m behind and 0.5\,m to the left of the observer.}
\exLabel{Arguments}{}
\begin{exverb}
\{\\{}
~~"\allowbreak{}image\_\allowbreak{}ids"\allowbreak{}: ["\allowbreak{}input\_\allowbreak{}image\_\allowbreak{}1"\allowbreak{},\allowbreak{} "\allowbreak{}input\_\allowbreak{}image\_\allowbreak{}2"\allowbreak{},\allowbreak{} "\allowbreak{}input\_\allowbreak{}image\_\allowbreak{}3"\allowbreak{}]\\{}
\}
\end{exverb}
\exLabel{Tool output}{}
\begin{exverb}
Camera trajectory summary:\\{}
- input\_\allowbreak{}image\_\allowbreak{}1: reference camera and reference view.\\{}
- input\_\allowbreak{}image\_\allowbreak{}2:\\{}
~~~~camera movement: salient direction: none (all movements \textless{} 1 meter); detailed: left (-0.22),\allowbreak{} down (+0.12),\allowbreak{} backward (-0.38)\\{}
~~~~camera rotation: salient direction: right (yaw +12.7°); detailed: right (yaw +12.7°),\allowbreak{} down (pitch +9.4°)\\{}
- input\_\allowbreak{}image\_\allowbreak{}3:\\{}
~~~~camera movement: salient direction: none (all movements \textless{} 1 meter); detailed: left (-0.53),\allowbreak{} down (+0.11),\allowbreak{} backward (-0.70)\\{}
~~~~camera rotation: salient direction: none (all rotations \textless{} 10°); detailed: right (yaw +5.7°),\allowbreak{} down (pitch +8.8°)\\{}
If no salient movement or rotation is reported,\allowbreak{} use the magnitude of each axis to determine the most possible movement/\allowbreak{}rotation direction.\\{}
\mbox{}\\{}
Bird\textquotesingle{}s-eye view of the camera trajectory on input\_\allowbreak{}image\_\allowbreak{}1\textquotesingle{}s X-Z plane is rendered in the following image - refer as tool\_\allowbreak{}generated\_\allowbreak{}image\_\allowbreak{}1 (dots =\allowbreak{} camera positions,\allowbreak{} forward-pointing arrows =\allowbreak{} facing direction).\\{}
\mbox{}\\{}
All vectors are relative to input\_\allowbreak{}image\_\allowbreak{}1\textquotesingle{}s camera frame,\allowbreak{} using the coordinate system: +X =\allowbreak{} right,\allowbreak{} +Y =\allowbreak{} down,\allowbreak{} +Z =\allowbreak{} forward. With this convention,\allowbreak{} positive yaw (rotation around Y axis) =\allowbreak{} right and positive pitch (rotation around X axis) =\allowbreak{} down. If the original question assumes a different coordinate system (e.g. with +Y =\allowbreak{} up and -Z =\allowbreak{} forward,\allowbreak{} which flips the yaw/\allowbreak{}pitch signs),\allowbreak{} convert before answering.
\end{exverb}
\begin{center}\includegraphics[width=0.45\linewidth,height=0.22\textheight,keepaspectratio]{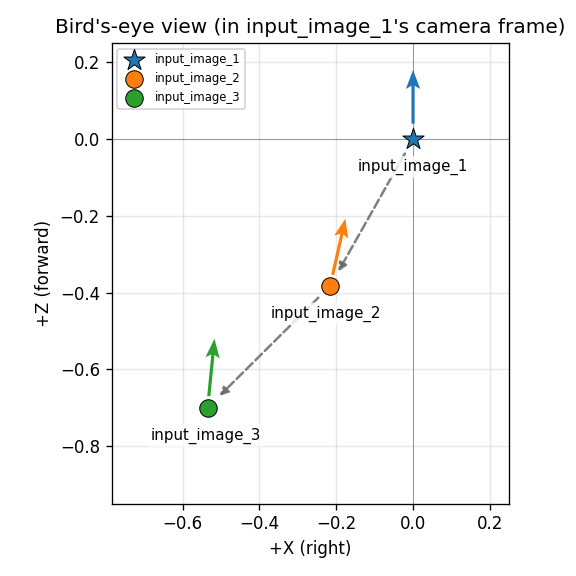}\\[2pt]
{\scriptsize\color{exMute} Bird's-eye view of the three camera positions in input\_image\_1's frame.}\end{center}
\begin{exharness}
\exLabel{Harness}{environment update, summarized: new memory entries; new data bindings verbatim}
\begin{exverb}
Memory update: registers 1 new image in memory (tool\_\allowbreak{}generated\_\allowbreak{}image\_\allowbreak{}1,\allowbreak{} from Camera\_\allowbreak{}Trajectory\_\allowbreak{}Tool).\\{}
Available data bindings (auto-injected into Python\_\allowbreak{}Coding\_\allowbreak{}Agent\_\allowbreak{}Tool):\\{}
- camera\_\allowbreak{}trajectory\_\allowbreak{}result: dict (keys: intrinsics,\allowbreak{} extrinsics,\allowbreak{} image\_\allowbreak{}ids,\allowbreak{} num\_\allowbreak{}images,\allowbreak{} cameras,\allowbreak{} coordinate\_\allowbreak{}system,\allowbreak{} reference\_\allowbreak{}image\_\allowbreak{}id)
\end{exverb}
\end{exharness}
\end{exstep}
\begin{exstep}{2}{\toolDepth}{}
\exWalk{Metric depth for the third photo, the only one showing the cabinet. The agent explains that it wants the depth at the red point.}
\exLabel{Arguments}{}
\begin{exverb}
\{\\{}
~~"\allowbreak{}image\_\allowbreak{}id"\allowbreak{}: "\allowbreak{}input\_\allowbreak{}image\_\allowbreak{}3"\allowbreak{},\allowbreak{}\\{}
~~"\allowbreak{}colormap"\allowbreak{}: "\allowbreak{}inferno"\allowbreak{}\\{}
\}
\end{exverb}
\exLabel{Model}{text emitted with this call}
\exSayBlock{I’ll estimate the marked cabinet’s depth relative to the camera using a depth map of the image containing the red point.}
\exLabel{Tool output}{}
\begin{exverb}
Tool \textasciigrave{}Depth\_\allowbreak{}Estimation\_\allowbreak{}Tool\textasciigrave{} (tool\_\allowbreak{}call\_\allowbreak{}id=\allowbreak{}call\_\allowbreak{}wl4V1aXn8L7m11iXLof958TR) returned:\\{}
Depth estimation complete. Shape: [966,\allowbreak{} 1302],\allowbreak{} colormap: inferno.\\{}
Metric depth in meters available as \textasciigrave{}depth\_\allowbreak{}meters\textasciigrave{} (float32 ndarray).\\{}
Colorized depth map rendered below; refer to it as tool\_\allowbreak{}generated\_\allowbreak{}image\_\allowbreak{}2.
\end{exverb}
\begin{center}\includegraphics[width=0.45\linewidth,height=0.22\textheight,keepaspectratio]{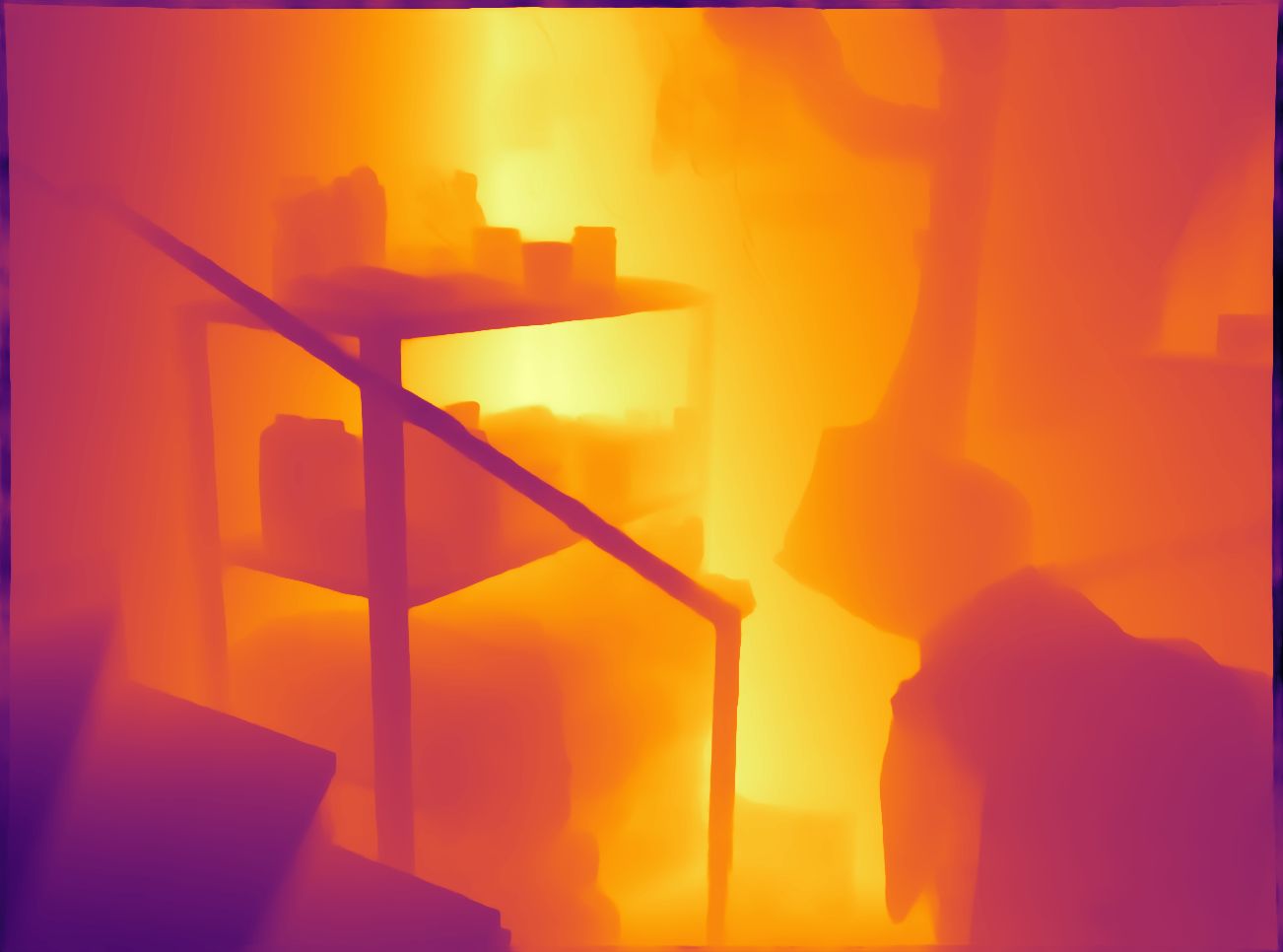}\\[2pt]
{\scriptsize\color{exMute} Colorized depth map of input\_image\_3.}\end{center}
\begin{exharness}
\exLabel{Harness}{environment update, summarized: new memory entries; new data bindings verbatim}
\begin{exverb}
Memory update: registers 1 new image in memory (tool\_\allowbreak{}generated\_\allowbreak{}image\_\allowbreak{}2,\allowbreak{} from Depth\_\allowbreak{}Estimation\_\allowbreak{}Tool).\\{}
- depth\_\allowbreak{}result: dict (keys: depth\_\allowbreak{}shape,\allowbreak{} colormap,\allowbreak{} size,\allowbreak{} depth\_\allowbreak{}meters)
\end{exverb}
\end{exharness}
\end{exstep}
\begin{exstep}{3}{\toolCode}{}
\exWalk{The coding agent is asked for the depth at the red point. The code fails because it uses a library it did not import.}
\exLabel{Arguments}{}
\begin{exverb}
\{\\{}
~~"\allowbreak{}objective"\allowbreak{}: "\allowbreak{}Extract the estimated depth in meters at the red point marking the cabinet in input\_\allowbreak{}image\_\allowbreak{}3 (roughly x=\allowbreak{}1110,\allowbreak{} y=\allowbreak{}923 in a 1302x966 image). Report the local median depth in a small surrounding patch,\allowbreak{} ignoring invalid values."\allowbreak{}\\{}
\}
\end{exverb}
\exLabel{Tool output}{}
\begin{exverb}
Coding agent generated and executed the following Python code:
\end{exverb}
\exCut{The generated code, 28 lines, is not shown: this run failed with the error below.}
\begin{exverb}
Execution error: name \textquotesingle{}math\textquotesingle{} is not defined
\end{exverb}
\end{exstep}
\begin{exstep}{4}{\toolCode}{}
\exWalk{A narrower retry reads the depth at the red point: 2.16\,m. This is the distance from the third camera, not from the observer.}
\exLabel{Arguments}{}
\begin{exverb}
\{\\{}
~~"\allowbreak{}objective"\allowbreak{}: "\allowbreak{}Read depth\_\allowbreak{}result depth\_\allowbreak{}meters at red point x$\approx$1110 y$\approx$923,\allowbreak{} image3 is 1302x966. Compute median finite values in the nearby 7x7 patch. Use no imports."\allowbreak{}\\{}
\}
\end{exverb}
\exLabel{Tool output}{}
\begin{exverb}
Coding agent generated and executed the following Python code:\\{}
\textasciigrave{}\textasciigrave{}\textasciigrave{}python\\{}
\# Get depth\_\allowbreak{}meters array and dimensions\\{}
depth\_\allowbreak{}map =\allowbreak{} depth\_\allowbreak{}result["\allowbreak{}depth\_\allowbreak{}meters"\allowbreak{}]\\{}
height,\allowbreak{} width =\allowbreak{} depth\_\allowbreak{}result["\allowbreak{}depth\_\allowbreak{}shape"\allowbreak{}]\\{}
\mbox{}\\{}
\# Target point (x$\approx$1110,\allowbreak{} y$\approx$923) - note: image coordinates are (width,\allowbreak{} height) =\allowbreak{} (1302,\allowbreak{} 966)\\{}
x,\allowbreak{} y =\allowbreak{} 1110,\allowbreak{} 923\\{}
\mbox{}\\{}
\# Define 7x7 patch around the point (centered at (x,\allowbreak{} y))\\{}
patch\_\allowbreak{}start\_\allowbreak{}row =\allowbreak{} max(0,\allowbreak{} y - 3)\\{}
patch\_\allowbreak{}end\_\allowbreak{}row =\allowbreak{} min(height,\allowbreak{} y + 4)\\{}
patch\_\allowbreak{}start\_\allowbreak{}col =\allowbreak{} max(0,\allowbreak{} x - 3)\\{}
patch\_\allowbreak{}end\_\allowbreak{}col =\allowbreak{} min(width,\allowbreak{} x + 4)\\{}
\mbox{}\\{}
\# Collect finite values in the 7x7 patch\\{}
finite\_\allowbreak{}values =\allowbreak{} []\\{}
for row in range(patch\_\allowbreak{}start\_\allowbreak{}row,\allowbreak{} patch\_\allowbreak{}end\_\allowbreak{}row):\\{}
~~~~for col in range(patch\_\allowbreak{}start\_\allowbreak{}col,\allowbreak{} patch\_\allowbreak{}end\_\allowbreak{}col):\\{}
~~~~~~~~val =\allowbreak{} depth\_\allowbreak{}map[row,\allowbreak{} col]\\{}
~~~~~~~~if not (val =\allowbreak{}=\allowbreak{} 0 or val =\allowbreak{}=\allowbreak{} float(\textquotesingle{}inf\textquotesingle{}) or val =\allowbreak{}=\allowbreak{} float(\textquotesingle{}-inf\textquotesingle{})):\\{}
~~~~~~~~~~~~finite\_\allowbreak{}values.append(val)\\{}
\mbox{}\\{}
\# Compute median of finite values\\{}
if len(finite\_\allowbreak{}values) \textgreater{} 0:\\{}
~~~~sorted\_\allowbreak{}values =\allowbreak{} sorted(finite\_\allowbreak{}values)\\{}
~~~~n =\allowbreak{} len(sorted\_\allowbreak{}values)\\{}
~~~~median =\allowbreak{} sorted\_\allowbreak{}values[n /\allowbreak{}/\allowbreak{} 2] if n \% 2 =\allowbreak{}=\allowbreak{} 1 else (sorted\_\allowbreak{}values[n /\allowbreak{}/\allowbreak{} 2 - 1] + sorted\_\allowbreak{}values[n /\allowbreak{}/\allowbreak{} 2]) /\allowbreak{} 2\\{}
else:\\{}
~~~~median =\allowbreak{} None\\{}
\mbox{}\\{}
print(median)\\{}
\textasciigrave{}\textasciigrave{}\textasciigrave{}\\{}
\mbox{}\\{}
\mbox{}\\{}
\mbox{}\\{}
Execution output:\\{}
2.1564598
\end{exverb}
\end{exstep}
\begin{exstep}{5}{\toolCode}{}
\exWalk{The agent asks for the distance from the observer. The code turns the red point into a 3D point in the third camera's frame and moves it into the first camera's frame using the recovered camera positions: 2.3995\,m.}
\exLabel{Arguments}{}
\begin{exverb}
\{\\{}
~~"\allowbreak{}objective"\allowbreak{}: "\allowbreak{}Estimate distance from reference camera (image1) to cabinet red-point center in image3. Use its depth (local median around x=\allowbreak{}1110,\allowbreak{}y=\allowbreak{}923),\allowbreak{} intrinsics and relative camera extrinsics/\allowbreak{}centers from camera\_\allowbreak{}trajectory\_\allowbreak{}result. Unproject the pixel into image3 camera coordinates,\allowbreak{} transform the 3D point into image1 camera coordinates,\allowbreak{} and calculate Euclidean distance. Report the computed value and relevant transform conventions."\allowbreak{}\\{}
\}
\end{exverb}
\exLabel{Tool output}{}
\begin{exverb}
Coding agent generated and executed the following Python code:\\{}
\textasciigrave{}\textasciigrave{}\textasciigrave{}python\\{}
import numpy as np\\{}
\mbox{}\\{}
\# Extract reference image index (image1) and target image index (image3)\\{}
ref\_\allowbreak{}idx =\allowbreak{} 0  \# \textquotesingle{}input\_\allowbreak{}image\_\allowbreak{}1\textquotesingle{}\\{}
target\_\allowbreak{}idx =\allowbreak{} 2  \# \textquotesingle{}input\_\allowbreak{}image\_\allowbreak{}3\textquotesingle{}\\{}
\mbox{}\\{}
\# Get depth at pixel (x=\allowbreak{}1110,\allowbreak{} y=\allowbreak{}923) in image3 (note: depth is in meters)\\{}
depth\_\allowbreak{}at\_\allowbreak{}pixel =\allowbreak{} depth\_\allowbreak{}result["\allowbreak{}depth\_\allowbreak{}meters"\allowbreak{}][923,\allowbreak{} 1110]  \# y,\allowbreak{} x indexing\\{}
\mbox{}\\{}
\# Compute local median depth around the pixel for robustness (3x3 neighborhood)\\{}
y\_\allowbreak{}center,\allowbreak{} x\_\allowbreak{}center =\allowbreak{} 923,\allowbreak{} 1110\\{}
y\_\allowbreak{}start,\allowbreak{} y\_\allowbreak{}end =\allowbreak{} max(0,\allowbreak{} y\_\allowbreak{}center - 1),\allowbreak{} min(depth\_\allowbreak{}result["\allowbreak{}depth\_\allowbreak{}shape"\allowbreak{}][0],\allowbreak{} y\_\allowbreak{}center + 2)\\{}
x\_\allowbreak{}start,\allowbreak{} x\_\allowbreak{}end =\allowbreak{} max(0,\allowbreak{} x\_\allowbreak{}center - 1),\allowbreak{} min(depth\_\allowbreak{}result["\allowbreak{}depth\_\allowbreak{}shape"\allowbreak{}][1],\allowbreak{} x\_\allowbreak{}center + 2)\\{}
local\_\allowbreak{}depth =\allowbreak{} depth\_\allowbreak{}result["\allowbreak{}depth\_\allowbreak{}meters"\allowbreak{}][y\_\allowbreak{}start:y\_\allowbreak{}end,\allowbreak{} x\_\allowbreak{}start:x\_\allowbreak{}end]\\{}
median\_\allowbreak{}depth =\allowbreak{} np.median(local\_\allowbreak{}depth)\\{}
\mbox{}\\{}
\# Use median depth for robustness\\{}
depth =\allowbreak{} median\_\allowbreak{}depth\\{}
\mbox{}\\{}
\# Unproject pixel (x=\allowbreak{}1110,\allowbreak{} y=\allowbreak{}923) in image3 to 3D point in image3 camera coordinates\\{}
\# Using intrinsics and depth\\{}
intrinsics =\allowbreak{} camera\_\allowbreak{}trajectory\_\allowbreak{}result["\allowbreak{}intrinsics"\allowbreak{}][target\_\allowbreak{}idx]  \# 3x3\\{}
K =\allowbreak{} intrinsics  \# camera matrix\\{}
\mbox{}\\{}
\# pixel coordinates (0-indexed,\allowbreak{} so subtract 1 if needed,\allowbreak{} but depth\_\allowbreak{}result uses 0-indexed)\\{}
x\_\allowbreak{}pixel =\allowbreak{} x\_\allowbreak{}center\\{}
y\_\allowbreak{}pixel =\allowbreak{} y\_\allowbreak{}center\\{}
\mbox{}\\{}
\# Unproject: (X,\allowbreak{} Y,\allowbreak{} Z) =\allowbreak{} depth * (u - cx) /\allowbreak{} fx,\allowbreak{} depth * (v - cy) /\allowbreak{} fy,\allowbreak{} depth\\{}
fx,\allowbreak{} fy =\allowbreak{} K[0,\allowbreak{} 0],\allowbreak{} K[1,\allowbreak{} 1]\\{}
cx,\allowbreak{} cy =\allowbreak{} K[0,\allowbreak{} 2],\allowbreak{} K[1,\allowbreak{} 2]\\{}
\mbox{}\\{}
X\_\allowbreak{}3d =\allowbreak{} depth * (x\_\allowbreak{}pixel - cx) /\allowbreak{} fx\\{}
Y\_\allowbreak{}3d =\allowbreak{} depth * (y\_\allowbreak{}pixel - cy) /\allowbreak{} fy\\{}
Z\_\allowbreak{}3d =\allowbreak{} depth\\{}
\mbox{}\\{}
point\_\allowbreak{}3d\_\allowbreak{}in\_\allowbreak{}target\_\allowbreak{}camera =\allowbreak{} np.array([X\_\allowbreak{}3d,\allowbreak{} Y\_\allowbreak{}3d,\allowbreak{} Z\_\allowbreak{}3d])\\{}
\mbox{}\\{}
\# Transform point from image3 camera coordinates to image1 camera coordinates\\{}
\# Use extrinsics: world\_\allowbreak{}to\_\allowbreak{}camera =\allowbreak{} [R \textbar{} t],\allowbreak{} so camera\_\allowbreak{}to\_\allowbreak{}world =\allowbreak{} R\^{}T * (-t)\\{}
\# We need to transform from image3 camera frame to image1 camera frame\\{}
\# So: point\_\allowbreak{}in\_\allowbreak{}ref\_\allowbreak{}camera =\allowbreak{} R\_\allowbreak{}ref\_\allowbreak{}to\_\allowbreak{}target @ point\_\allowbreak{}in\_\allowbreak{}target\_\allowbreak{}camera + t\_\allowbreak{}ref\_\allowbreak{}to\_\allowbreak{}target\\{}
\mbox{}\\{}
\# Get extrinsics for image3 (target) and image1 (reference)\\{}
extrinsics\_\allowbreak{}target =\allowbreak{} camera\_\allowbreak{}trajectory\_\allowbreak{}result["\allowbreak{}extrinsics"\allowbreak{}][target\_\allowbreak{}idx]  \# 3x4\\{}
extrinsics\_\allowbreak{}ref =\allowbreak{} camera\_\allowbreak{}trajectory\_\allowbreak{}result["\allowbreak{}extrinsics"\allowbreak{}][ref\_\allowbreak{}idx]  \# 3x4\\{}
\mbox{}\\{}
\# Extract rotation matrix and translation vector for target and reference\\{}
R\_\allowbreak{}target =\allowbreak{} extrinsics\_\allowbreak{}target[:,\allowbreak{} :3]  \# 3x3 rotation matrix\\{}
t\_\allowbreak{}target =\allowbreak{} extrinsics\_\allowbreak{}target[:,\allowbreak{} 3]   \# 3x1 translation vector\\{}
\mbox{}\\{}
R\_\allowbreak{}ref =\allowbreak{} extrinsics\_\allowbreak{}ref[:,\allowbreak{} :3]  \# 3x3 rotation matrix\\{}
t\_\allowbreak{}ref =\allowbreak{} extrinsics\_\allowbreak{}ref[:,\allowbreak{} 3]   \# 3x1 translation vector\\{}
\mbox{}\\{}
\# Transform point from target camera frame to reference camera frame\\{}
\# point\_\allowbreak{}in\_\allowbreak{}ref\_\allowbreak{}camera =\allowbreak{} R\_\allowbreak{}ref\_\allowbreak{}to\_\allowbreak{}target @ point\_\allowbreak{}in\_\allowbreak{}target\_\allowbreak{}camera + t\_\allowbreak{}ref\_\allowbreak{}to\_\allowbreak{}target\\{}
\# R\_\allowbreak{}ref\_\allowbreak{}to\_\allowbreak{}target =\allowbreak{} R\_\allowbreak{}target\^{}T (since extrinsics are world\_\allowbreak{}to\_\allowbreak{}camera)\\{}
\# t\_\allowbreak{}ref\_\allowbreak{}to\_\allowbreak{}target =\allowbreak{} -R\_\allowbreak{}target\^{}T @ t\_\allowbreak{}target + t\_\allowbreak{}ref\\{}
\mbox{}\\{}
R\_\allowbreak{}ref\_\allowbreak{}to\_\allowbreak{}target =\allowbreak{} R\_\allowbreak{}target.T  \# rotation from target to ref\\{}
t\_\allowbreak{}ref\_\allowbreak{}to\_\allowbreak{}target =\allowbreak{} -R\_\allowbreak{}ref\_\allowbreak{}to\_\allowbreak{}target @ t\_\allowbreak{}target + t\_\allowbreak{}ref\\{}
\mbox{}\\{}
\# Apply transformation\\{}
point\_\allowbreak{}3d\_\allowbreak{}in\_\allowbreak{}ref\_\allowbreak{}camera =\allowbreak{} R\_\allowbreak{}ref\_\allowbreak{}to\_\allowbreak{}target @ point\_\allowbreak{}3d\_\allowbreak{}in\_\allowbreak{}target\_\allowbreak{}camera + t\_\allowbreak{}ref\_\allowbreak{}to\_\allowbreak{}target\\{}
\mbox{}\\{}
\# Compute Euclidean distance from reference camera center to the point\\{}
camera\_\allowbreak{}center\_\allowbreak{}ref =\allowbreak{} t\_\allowbreak{}ref  \# camera center in reference frame is the translation vector\\{}
distance =\allowbreak{} np.linalg.norm(point\_\allowbreak{}3d\_\allowbreak{}in\_\allowbreak{}ref\_\allowbreak{}camera - camera\_\allowbreak{}center\_\allowbreak{}ref)\\{}
\mbox{}\\{}
\# Print results\\{}
print(f"\allowbreak{}Estimated distance from reference camera to cabinet red-point center in image3: \{distance:.4f\} meters"\allowbreak{})\\{}
print(f"\allowbreak{}Transform conventions: All vectors are relative to \{camera\_\allowbreak{}trajectory\_\allowbreak{}result[\textquotesingle{}reference\_\allowbreak{}image\_\allowbreak{}id\textquotesingle{}]\}\textquotesingle{}s camera frame,\allowbreak{} using the coordinate system: \{camera\_\allowbreak{}trajectory\_\allowbreak{}result[\textquotesingle{}coordinate\_\allowbreak{}system\textquotesingle{}]\}"\allowbreak{})\\{}
\textasciigrave{}\textasciigrave{}\textasciigrave{}\\{}
\mbox{}\\{}
\mbox{}\\{}
\mbox{}\\{}
Execution output:\\{}
Estimated distance from reference camera to cabinet red-point center in image3: 2.3995 meters\\{}
Transform conventions: All vectors are relative to input\_\allowbreak{}image\_\allowbreak{}1\textquotesingle{}s camera frame,\allowbreak{} using the coordinate system: All vectors are relative to input\_\allowbreak{}image\_\allowbreak{}1\textquotesingle{}s camera frame,\allowbreak{} using the coordinate system: +X =\allowbreak{} right,\allowbreak{} +Y =\allowbreak{} down,\allowbreak{} +Z =\allowbreak{} forward. With this convention,\allowbreak{} positive yaw (rotation around Y axis) =\allowbreak{} right and positive pitch (rotation around X axis) =\allowbreak{} down. If the original question assumes a different coordinate system (e.g. with +Y =\allowbreak{} up and -Z =\allowbreak{} forward,\allowbreak{} which flips the yaw/\allowbreak{}pitch signs),\allowbreak{} convert before answering.
\end{exverb}
\end{exstep}
\begin{exstep}{}{\exTool{exInk}{Final answer}}{}
\exWalk{The agent answers 2.4, which matches the ground truth.}
\exLabel{Model}{}
\begin{exverb}
\textless{}answer\textgreater{}2.4\textless{}/\allowbreak{}answer\textgreater{}
\end{exverb}
\end{exstep}

%% file: text/appendix_examples_steps/q558_header.tex
\exOverview{\textbf{Task.} Three photographs of an office corner are taken from nearby positions. A red point marks a window in the first photo and a blue point marks a cabinet in the third. Give the 3D distance between them in metres. Without tools the same model answers 1.7\,m. 

\textbf{How the agent approaches it.} It chains four kinds of tool: camera trajectory to relate the views, depth in all three views, grounding to locate the window and the cabinet, and the coding agent to back-project both points into one frame and measure between them. 

\textbf{Caveat.} The pipeline is the right one, but its last step does not use the marked points. The generated code reads each box as position plus width and height, although the boxes are corner coordinates, and it takes the first ``cabinet'' box, an upper wall cabinet at the top edge of the photo, instead of the lower cabinet with the blue point. The answer, 1.20\,m against a ground truth of 1.1\,m, is therefore close partly by coincidence.}
\exLabel{Input images}{}
\begin{center}
\begin{minipage}[t]{0.312\linewidth}\centering\includegraphics[width=\linewidth,height=0.22\textheight,keepaspectratio]{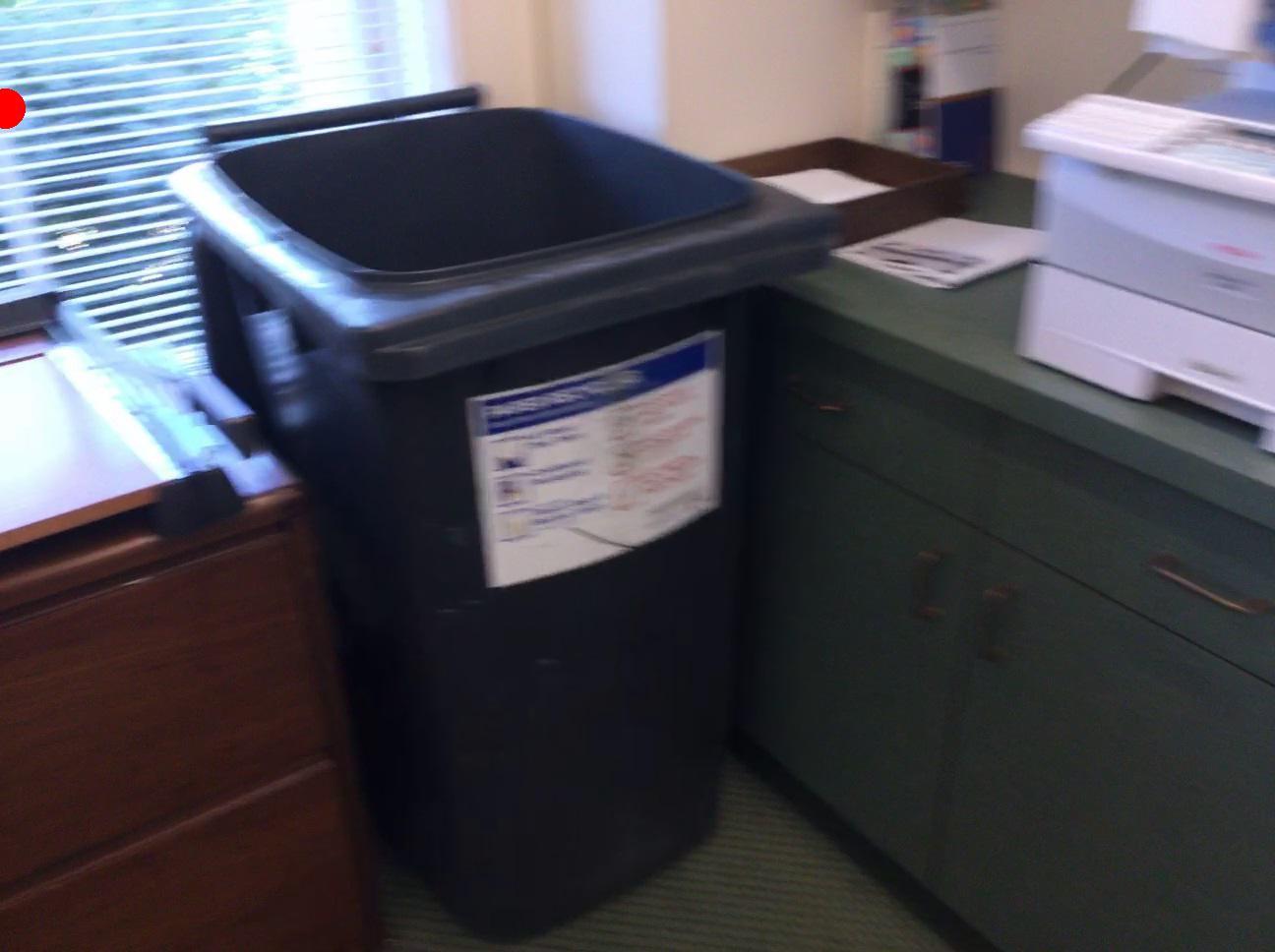}\\[1pt]{\scriptsize\color{exMute} input\_image\_1}\end{minipage}\hfill\begin{minipage}[t]{0.312\linewidth}\centering\includegraphics[width=\linewidth,height=0.22\textheight,keepaspectratio]{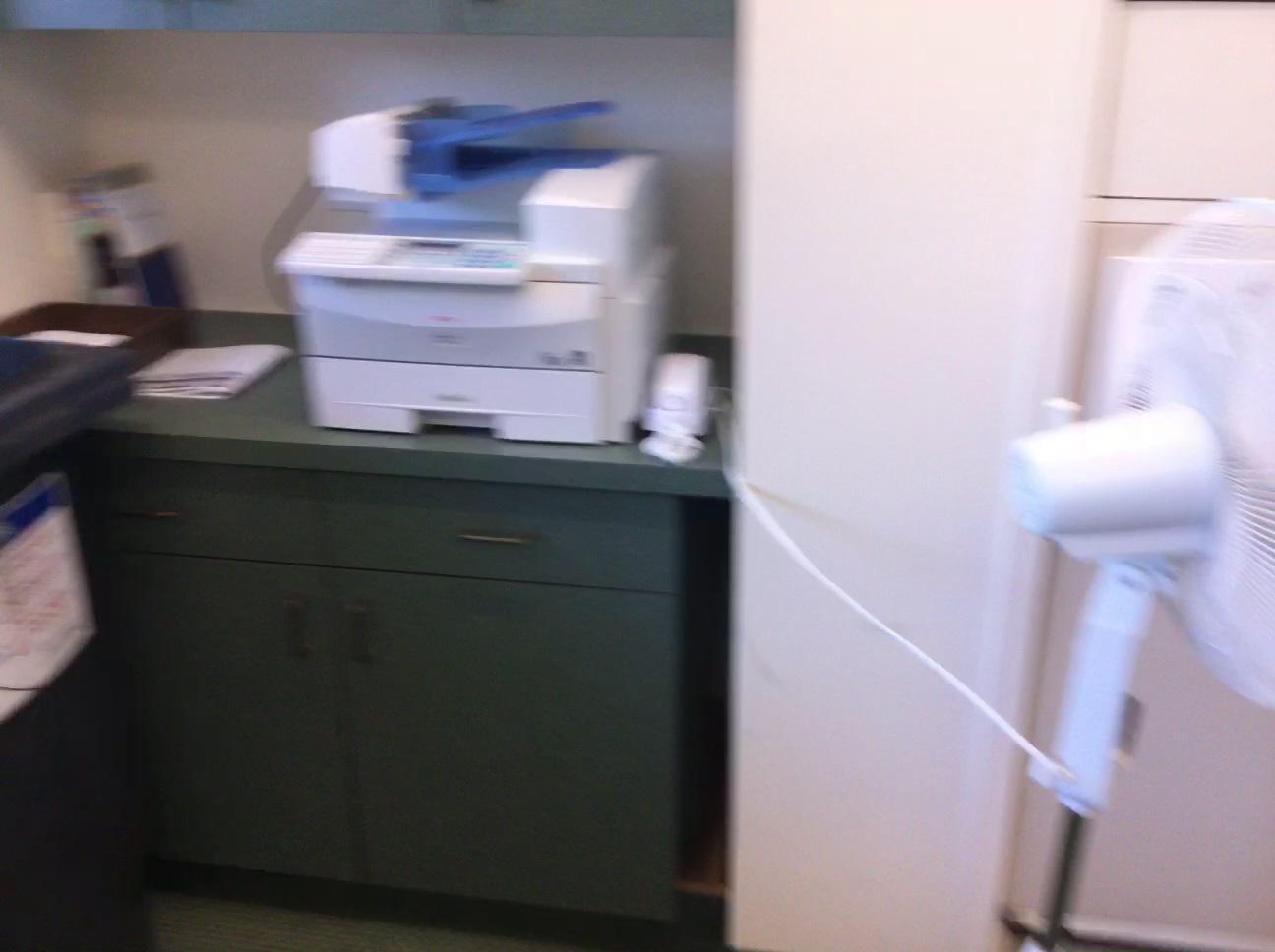}\\[1pt]{\scriptsize\color{exMute} input\_image\_2}\end{minipage}\hfill\begin{minipage}[t]{0.312\linewidth}\centering\includegraphics[width=\linewidth,height=0.22\textheight,keepaspectratio]{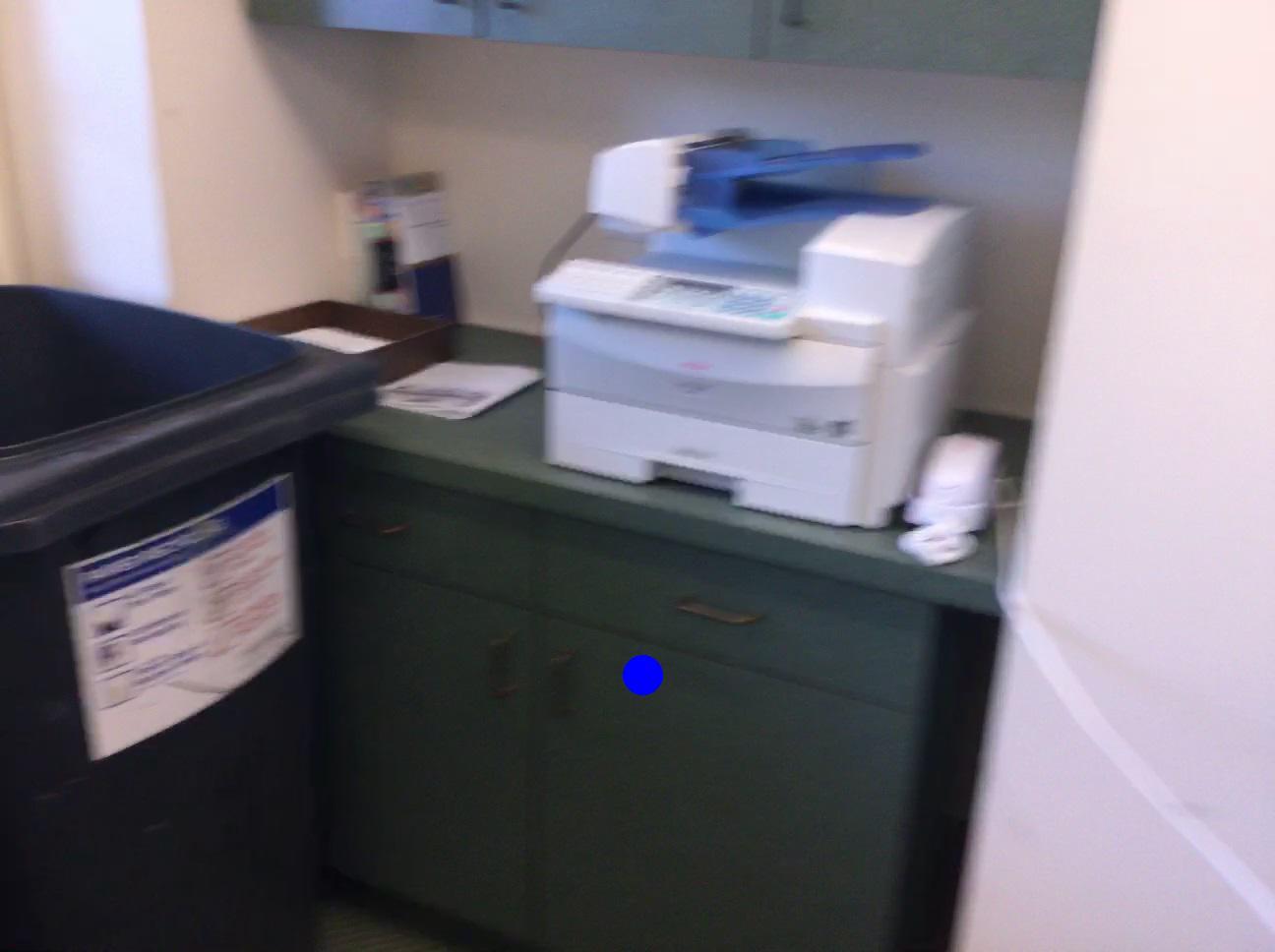}\\[1pt]{\scriptsize\color{exMute} input\_image\_3}\end{minipage}
\end{center}
\exLabel{Question}{as sent to the model}
\begin{exverb}
Question: \\{}
In meters,\allowbreak{} how far is the center of window (red point) from the center of cabinet (blue point)? Calculate or judge based on the 3D center points of these objects. Only one numeric value is required for your response.\\{}
Please answer the question using a single word or phrase.
\end{exverb}

%% file: text/appendix_examples_steps/q558.tex
\begin{exharness}
\exLabel{Harness}{environment update, shown in full}
\begin{exverb}
Environment Update (not a new question):\\{}
Currently available image IDs for tool calls:\\{}
- input\_\allowbreak{}image\_\allowbreak{}1: user\_\allowbreak{}input\\{}
- input\_\allowbreak{}image\_\allowbreak{}2: user\_\allowbreak{}input\\{}
- input\_\allowbreak{}image\_\allowbreak{}3: user\_\allowbreak{}input\\{}
When a tool needs image input,\allowbreak{} pass one of these IDs via the *\_\allowbreak{}image\_\allowbreak{}id field.\\{}
Do NOT invent image file names or paths.
\end{exverb}
\end{exharness}
\begin{exstep}{1}{\toolCam}{}
\exLabel{Model}{text emitted with this call}
\exSayBlock{\textless{}thinking\textgreater{}To calculate the distance between the center of the window (red point) and the center of the cabinet (blue point),\allowbreak{} I need to estimate their 3D positions. Since the images are from different viewpoints,\allowbreak{} I will first use the Camera\_\allowbreak{}Trajectory\_\allowbreak{}Tool to understand the camera motion between the frames. Then,\allowbreak{} I can use depth estimation and object grounding to get approximate 3D coordinates for both points. Finally,\allowbreak{} I\textquotesingle{}ll compute the Euclidean distance between them.\textless{}/\allowbreak{}thinking\textgreater{}}
\exLabel{Arguments}{}
\begin{exverb}
\{\\{}
~~"\allowbreak{}image\_\allowbreak{}ids"\allowbreak{}: ["\allowbreak{}input\_\allowbreak{}image\_\allowbreak{}1"\allowbreak{},\allowbreak{} "\allowbreak{}input\_\allowbreak{}image\_\allowbreak{}2"\allowbreak{},\allowbreak{} "\allowbreak{}input\_\allowbreak{}image\_\allowbreak{}3"\allowbreak{}]\\{}
\}
\end{exverb}
\exLabel{Tool output}{}
\begin{exverb}
Camera trajectory summary:\\{}
- input\_\allowbreak{}image\_\allowbreak{}1: reference camera and reference view.\\{}
- input\_\allowbreak{}image\_\allowbreak{}2:\\{}
~~~~camera movement: salient direction: none (all movements \textless{} 1 meter); detailed: left (-0.40),\allowbreak{} down (+0.07),\allowbreak{} backward (-0.29)\\{}
~~~~camera rotation: salient direction: right (yaw +40.7°); detailed: right (yaw +40.7°),\allowbreak{} down (pitch +1.0°)\\{}
- input\_\allowbreak{}image\_\allowbreak{}3:\\{}
~~~~camera movement: salient direction: none (all movements \textless{} 1 meter); detailed: left (-0.26),\allowbreak{} down (+0.03),\allowbreak{} backward (-0.15)\\{}
~~~~camera rotation: salient direction: right (yaw +28.8°); detailed: right (yaw +28.8°),\allowbreak{} up (pitch -4.1°)\\{}
If no salient movement or rotation is reported,\allowbreak{} use the magnitude of each axis to determine the most possible movement/\allowbreak{}rotation direction.\\{}
\mbox{}\\{}
Bird\textquotesingle{}s-eye view of the camera trajectory on input\_\allowbreak{}image\_\allowbreak{}1\textquotesingle{}s X-Z plane is rendered in the following image - refer as tool\_\allowbreak{}generated\_\allowbreak{}image\_\allowbreak{}1 (dots =\allowbreak{} camera positions,\allowbreak{} forward-pointing arrows =\allowbreak{} facing direction).\\{}
\mbox{}\\{}
All vectors are relative to input\_\allowbreak{}image\_\allowbreak{}1\textquotesingle{}s camera frame,\allowbreak{} using the coordinate system: +X =\allowbreak{} right,\allowbreak{} +Y =\allowbreak{} down,\allowbreak{} +Z =\allowbreak{} forward. With this convention,\allowbreak{} positive yaw (rotation around Y axis) =\allowbreak{} right and positive pitch (rotation around X axis) =\allowbreak{} down. If the original question assumes a different coordinate system (e.g. with +Y =\allowbreak{} up and -Z =\allowbreak{} forward,\allowbreak{} which flips the yaw/\allowbreak{}pitch signs),\allowbreak{} convert before answering.
\end{exverb}
\begin{center}\includegraphics[width=0.45\linewidth,height=0.22\textheight,keepaspectratio]{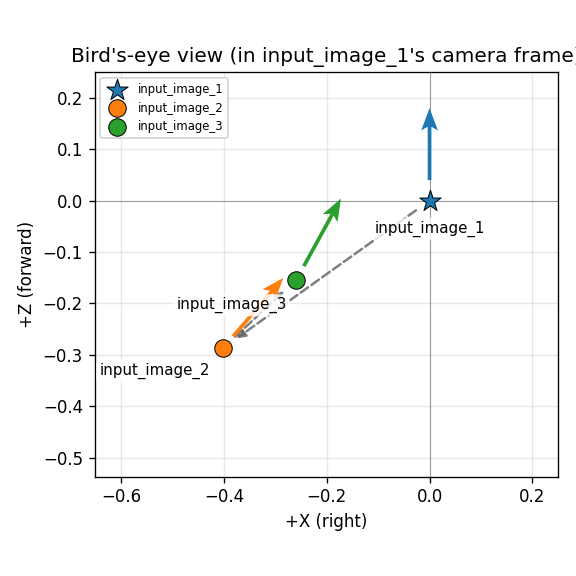}\\[2pt]
{\scriptsize\color{exMute} Image returned with this output.}\end{center}
\begin{exharness}
\exLabel{Harness}{environment update, summarized: new memory entries; new data bindings verbatim}
\begin{exverb}
Memory update: registers 1 new image in memory (tool\_\allowbreak{}generated\_\allowbreak{}image\_\allowbreak{}1,\allowbreak{} from Camera\_\allowbreak{}Trajectory\_\allowbreak{}Tool).\\{}
Available data bindings (auto-injected into Python\_\allowbreak{}Coding\_\allowbreak{}Agent\_\allowbreak{}Tool):\\{}
- camera\_\allowbreak{}trajectory\_\allowbreak{}result: dict (keys: intrinsics,\allowbreak{} extrinsics,\allowbreak{} image\_\allowbreak{}ids,\allowbreak{} num\_\allowbreak{}images,\allowbreak{} cameras,\allowbreak{} coordinate\_\allowbreak{}system,\allowbreak{} reference\_\allowbreak{}image\_\allowbreak{}id)
\end{exverb}
\end{exharness}
\end{exstep}
\begin{exstep}{2}{\toolDepth}{}
\exLabel{Model}{text emitted with this call}
\exSayBlock{\textless{}thinking\textgreater{}Now that I have the camera trajectory,\allowbreak{} I will estimate the 3D positions of the window (red point) and cabinet (blue point). I\textquotesingle{}ll use depth estimation on each image to get depth values at these points,\allowbreak{} then combine with the camera positions and orientations to compute their 3D coordinates. Finally,\allowbreak{} I\textquotesingle{}ll calculate the Euclidean distance between them.\textless{}/\allowbreak{}thinking\textgreater{}}
\exLabel{Arguments}{}
\begin{exverb}
\{\\{}
~~"\allowbreak{}image\_\allowbreak{}id"\allowbreak{}: "\allowbreak{}input\_\allowbreak{}image\_\allowbreak{}1"\allowbreak{},\allowbreak{}\\{}
~~"\allowbreak{}colormap"\allowbreak{}: "\allowbreak{}inferno"\allowbreak{}\\{}
\}
\end{exverb}
\exLabel{Tool output}{}
\begin{exverb}
Tool \textasciigrave{}Depth\_\allowbreak{}Estimation\_\allowbreak{}Tool\textasciigrave{} (tool\_\allowbreak{}call\_\allowbreak{}id=\allowbreak{}chatcmpl-tool-ba25e41bcbdd4895) returned:\\{}
Depth estimation complete. Shape: [966,\allowbreak{} 1302],\allowbreak{} colormap: inferno.\\{}
Metric depth in meters available as \textasciigrave{}depth\_\allowbreak{}meters\textasciigrave{} (float32 ndarray).\\{}
Colorized depth map rendered below; refer to it as tool\_\allowbreak{}generated\_\allowbreak{}image\_\allowbreak{}2.
\end{exverb}
\begin{center}\includegraphics[width=0.45\linewidth,height=0.22\textheight,keepaspectratio]{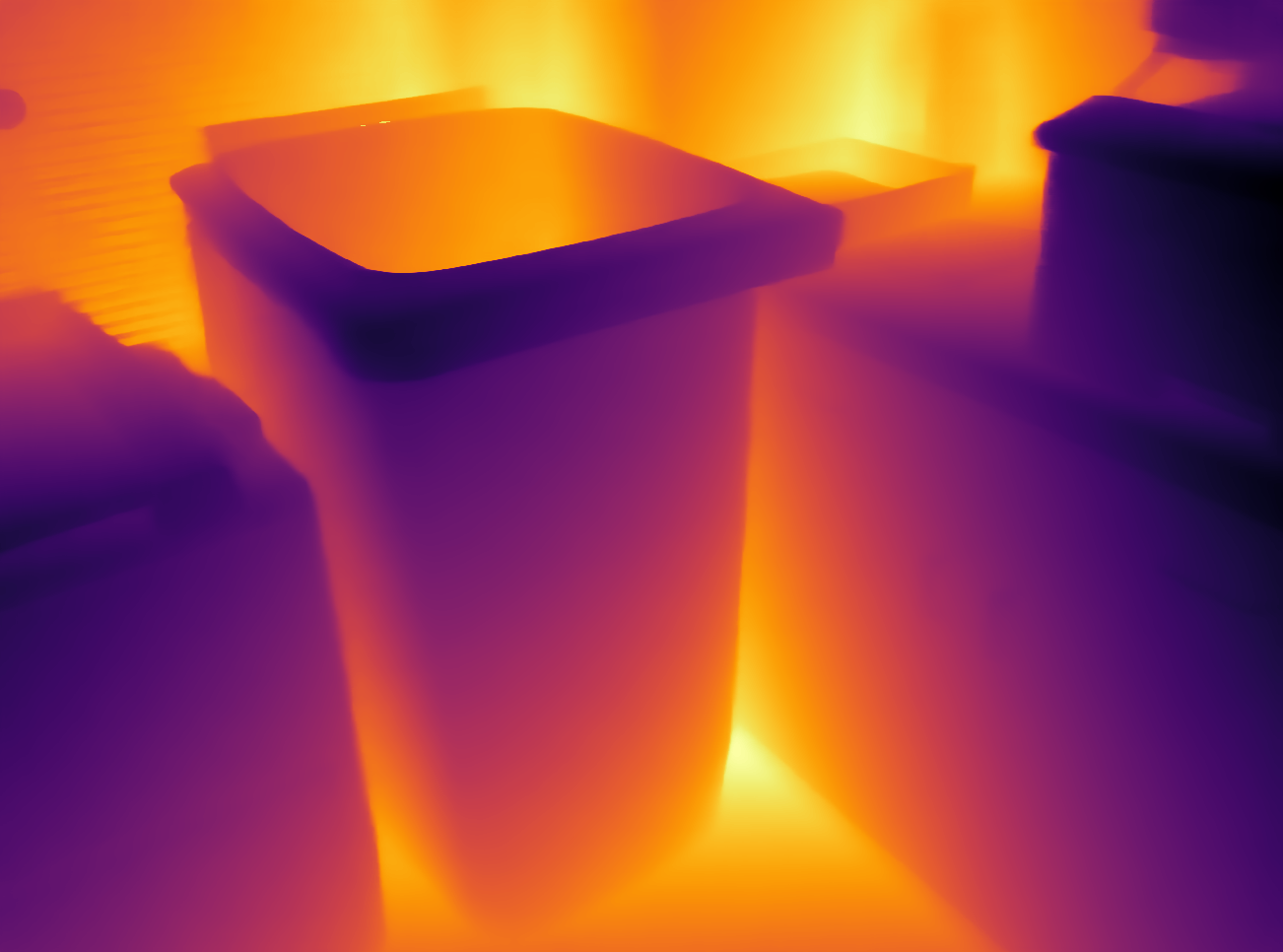}\\[2pt]
{\scriptsize\color{exMute} Image returned with this output.}\end{center}
\begin{exharness}
\exLabel{Harness}{environment update, summarized: new memory entries; new data bindings verbatim}
\begin{exverb}
Memory update: registers 1 new image in memory (tool\_\allowbreak{}generated\_\allowbreak{}image\_\allowbreak{}2,\allowbreak{} from Depth\_\allowbreak{}Estimation\_\allowbreak{}Tool).\\{}
- depth\_\allowbreak{}result: dict (keys: depth\_\allowbreak{}shape,\allowbreak{} colormap,\allowbreak{} size,\allowbreak{} depth\_\allowbreak{}meters)
\end{exverb}
\end{exharness}
\end{exstep}
\begin{exstep}{3}{\toolDepth}{}
\exLabel{Model}{text emitted with this call}
\exSayBlock{\textless{}thinking\textgreater{}Next,\allowbreak{} I will estimate the depth at the red point (window center) and blue point (cabinet center) in each image. Since the points are marked in different images,\allowbreak{} I\textquotesingle{}ll need to use the depth maps from each respective image to get their depths. Then,\allowbreak{} using the camera positions and orientations from the trajectory,\allowbreak{} I can compute the 3D coordinates of each point and calculate the distance between them.\textless{}/\allowbreak{}thinking\textgreater{}}
\exLabel{Arguments}{}
\begin{exverb}
\{\\{}
~~"\allowbreak{}image\_\allowbreak{}id"\allowbreak{}: "\allowbreak{}input\_\allowbreak{}image\_\allowbreak{}2"\allowbreak{},\allowbreak{}\\{}
~~"\allowbreak{}colormap"\allowbreak{}: "\allowbreak{}inferno"\allowbreak{}\\{}
\}
\end{exverb}
\exLabel{Tool output}{}
\begin{exverb}
Tool \textasciigrave{}Depth\_\allowbreak{}Estimation\_\allowbreak{}Tool\textasciigrave{} (tool\_\allowbreak{}call\_\allowbreak{}id=\allowbreak{}chatcmpl-tool-a034e48c2e99033e) returned:\\{}
Depth estimation complete. Shape: [966,\allowbreak{} 1302],\allowbreak{} colormap: inferno.\\{}
Metric depth in meters available as \textasciigrave{}depth\_\allowbreak{}meters\textasciigrave{} (float32 ndarray).\\{}
Colorized depth map rendered below; refer to it as tool\_\allowbreak{}generated\_\allowbreak{}image\_\allowbreak{}3.
\end{exverb}
\begin{center}\includegraphics[width=0.45\linewidth,height=0.22\textheight,keepaspectratio]{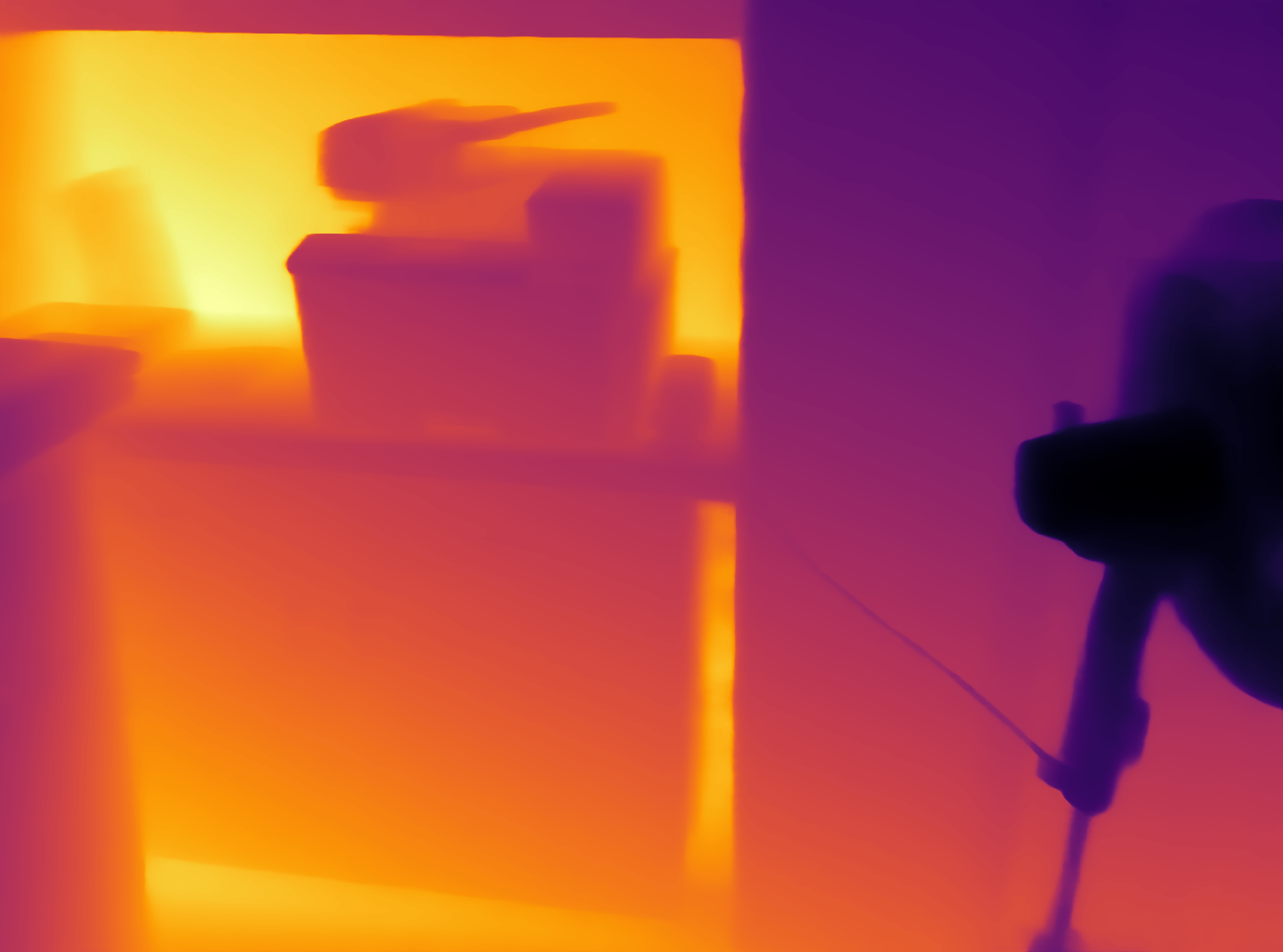}\\[2pt]
{\scriptsize\color{exMute} Image returned with this output.}\end{center}
\begin{exharness}
\exLabel{Harness}{environment update, summarized: new memory entries; new data bindings verbatim}
\begin{exverb}
Memory update: registers 1 new image in memory (tool\_\allowbreak{}generated\_\allowbreak{}image\_\allowbreak{}3,\allowbreak{} from Depth\_\allowbreak{}Estimation\_\allowbreak{}Tool).\\{}
- depth\_\allowbreak{}result\_\allowbreak{}2: dict (keys: depth\_\allowbreak{}shape,\allowbreak{} colormap,\allowbreak{} size,\allowbreak{} depth\_\allowbreak{}meters)
\end{exverb}
\end{exharness}
\end{exstep}
\begin{exstep}{4}{\toolDepth}{}
\exLabel{Model}{text emitted with this call}
\exSayBlock{\textless{}thinking\textgreater{}Now,\allowbreak{} I will estimate the depth at the red point (window center) and blue point (cabinet center) in each image. Since the points are marked in different images,\allowbreak{} I\textquotesingle{}ll need to use the depth maps from each respective image to get their depths. Then,\allowbreak{} using the camera positions and orientations from the trajectory,\allowbreak{} I can compute the 3D coordinates of each point and calculate the distance between them.\textless{}/\allowbreak{}thinking\textgreater{}}
\exLabel{Arguments}{}
\begin{exverb}
\{\\{}
~~"\allowbreak{}image\_\allowbreak{}id"\allowbreak{}: "\allowbreak{}input\_\allowbreak{}image\_\allowbreak{}3"\allowbreak{},\allowbreak{}\\{}
~~"\allowbreak{}colormap"\allowbreak{}: "\allowbreak{}inferno"\allowbreak{}\\{}
\}
\end{exverb}
\exLabel{Tool output}{}
\begin{exverb}
Tool \textasciigrave{}Depth\_\allowbreak{}Estimation\_\allowbreak{}Tool\textasciigrave{} (tool\_\allowbreak{}call\_\allowbreak{}id=\allowbreak{}chatcmpl-tool-956d2565d6a5d9fb) returned:\\{}
Depth estimation complete. Shape: [966,\allowbreak{} 1302],\allowbreak{} colormap: inferno.\\{}
Metric depth in meters available as \textasciigrave{}depth\_\allowbreak{}meters\textasciigrave{} (float32 ndarray).\\{}
Colorized depth map rendered below; refer to it as tool\_\allowbreak{}generated\_\allowbreak{}image\_\allowbreak{}4.
\end{exverb}
\begin{center}\includegraphics[width=0.45\linewidth,height=0.22\textheight,keepaspectratio]{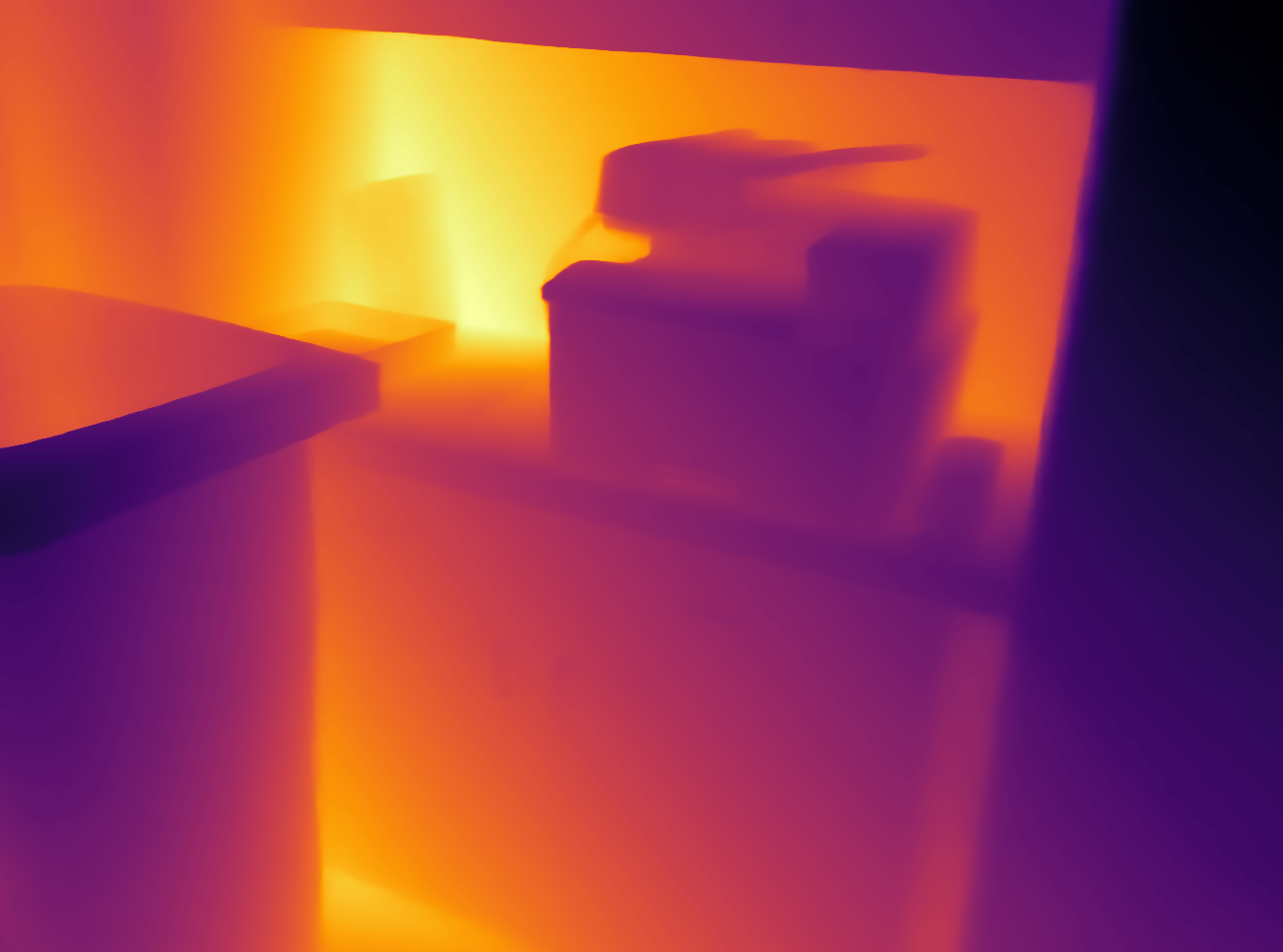}\\[2pt]
{\scriptsize\color{exMute} Image returned with this output.}\end{center}
\begin{exharness}
\exLabel{Harness}{environment update, summarized: new memory entries; new data bindings verbatim}
\begin{exverb}
Memory update: registers 1 new image in memory (tool\_\allowbreak{}generated\_\allowbreak{}image\_\allowbreak{}4,\allowbreak{} from Depth\_\allowbreak{}Estimation\_\allowbreak{}Tool).\\{}
- depth\_\allowbreak{}result\_\allowbreak{}3: dict (keys: depth\_\allowbreak{}shape,\allowbreak{} colormap,\allowbreak{} size,\allowbreak{} depth\_\allowbreak{}meters)
\end{exverb}
\end{exharness}
\end{exstep}
\begin{exstep}{5}{\toolVG}{}
\exLabel{Model}{text emitted with this call}
\exSayBlock{\textless{}thinking\textgreater{}Now I have depth maps for all three images. Next,\allowbreak{} I will use the Visual\_\allowbreak{}Grounding\_\allowbreak{}Tool to locate the window (red point) in input\_\allowbreak{}image\_\allowbreak{}1 and the cabinet (blue point) in input\_\allowbreak{}image\_\allowbreak{}3. This will give me the pixel coordinates of these points,\allowbreak{} which I can then use with the depth maps to get their 3D coordinates relative to each camera. After that,\allowbreak{} I\textquotesingle{}ll use the camera trajectory data to transform these points into a common coordinate system and compute the distance between them.\textless{}/\allowbreak{}thinking\textgreater{}}
\exLabel{Arguments}{}
\begin{exverb}
\{\\{}
~~"\allowbreak{}query"\allowbreak{}: "\allowbreak{}window"\allowbreak{},\allowbreak{}\\{}
~~"\allowbreak{}image\_\allowbreak{}id"\allowbreak{}: "\allowbreak{}input\_\allowbreak{}image\_\allowbreak{}1"\allowbreak{}\\{}
\}
\end{exverb}
\exLabel{Tool output}{}
\begin{exverb}
Visual\_\allowbreak{}Grounding\_\allowbreak{}Tool grounded 1 objects for the prompt \textquotesingle{}window\textquotesingle{}. Their corresponding bounding boxes (on a 0-1000 scale,\allowbreak{} [x1,\allowbreak{} y1,\allowbreak{} x2,\allowbreak{} y2]) are: \{\textquotesingle{}object 1\textquotesingle{}: [0,\allowbreak{} 0,\allowbreak{} 353.92,\allowbreak{} 389.56]\}. The object numbers correspond to the labeled masks rendered in the following image,\allowbreak{} which you can refer to as tool\_\allowbreak{}generated\_\allowbreak{}image\_\allowbreak{}5.
\end{exverb}
\begin{center}\includegraphics[width=0.45\linewidth,height=0.22\textheight,keepaspectratio]{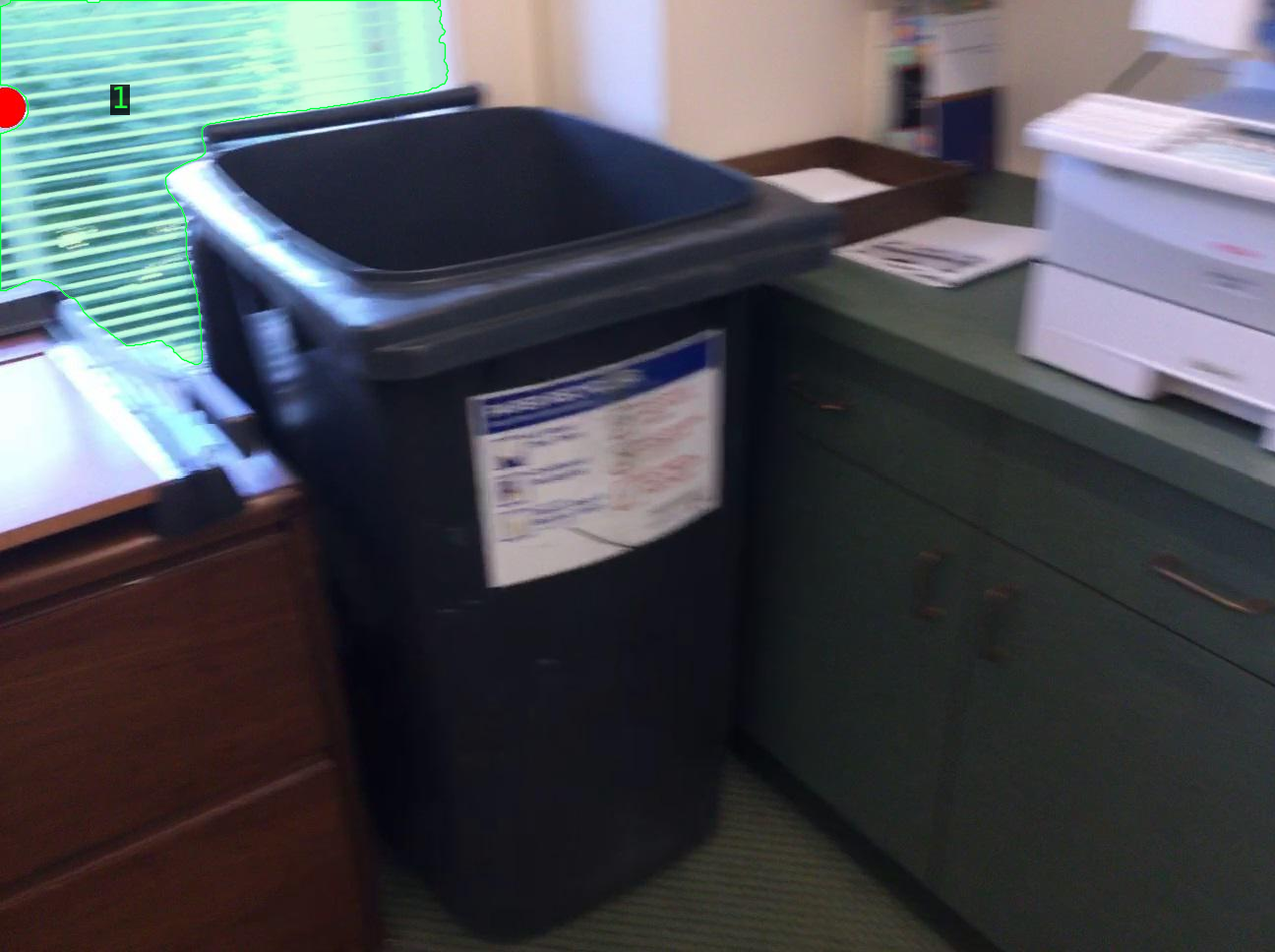}\\[2pt]
{\scriptsize\color{exMute} Image returned with this output.}\end{center}
\begin{exharness}
\exLabel{Harness}{environment update, summarized: new memory entries; new data bindings verbatim}
\begin{exverb}
Memory update: registers 1 new image in memory (tool\_\allowbreak{}generated\_\allowbreak{}image\_\allowbreak{}5,\allowbreak{} from Visual\_\allowbreak{}Grounding\_\allowbreak{}Tool).\\{}
- segment\_\allowbreak{}window: dict (keys: pred\_\allowbreak{}boxes,\allowbreak{} pred\_\allowbreak{}scores,\allowbreak{} pred\_\allowbreak{}masks,\allowbreak{} num\_\allowbreak{}masks,\allowbreak{} orig\_\allowbreak{}img\_\allowbreak{}h,\allowbreak{} orig\_\allowbreak{}img\_\allowbreak{}w)
\end{exverb}
\end{exharness}
\end{exstep}
\begin{exstep}{6}{\toolVG}{}
\exLabel{Model}{text emitted with this call}
\exSayBlock{\textless{}thinking\textgreater{}Now I will ground the cabinet (blue point) in input\_\allowbreak{}image\_\allowbreak{}3 to get its pixel coordinates. After that,\allowbreak{} I\textquotesingle{}ll use the depth maps and camera trajectory data to compute the 3D coordinates of both points and calculate the distance between them.\textless{}/\allowbreak{}thinking\textgreater{}}
\exLabel{Arguments}{}
\begin{exverb}
\{\\{}
~~"\allowbreak{}query"\allowbreak{}: "\allowbreak{}cabinet"\allowbreak{},\allowbreak{}\\{}
~~"\allowbreak{}image\_\allowbreak{}id"\allowbreak{}: "\allowbreak{}input\_\allowbreak{}image\_\allowbreak{}3"\allowbreak{}\\{}
\}
\end{exverb}
\exLabel{Tool output}{}
\begin{exverb}
Visual\_\allowbreak{}Grounding\_\allowbreak{}Tool grounded 5 objects for the prompt \textquotesingle{}cabinet\textquotesingle{}. Their corresponding bounding boxes (on a 0-1000 scale,\allowbreak{} [x1,\allowbreak{} y1,\allowbreak{} x2,\allowbreak{} y2]) are: \{\textquotesingle{}object 1\textquotesingle{}: [194.95,\allowbreak{} 0,\allowbreak{} 373.1,\allowbreak{} 39.21],\allowbreak{} \textquotesingle{}object 2\textquotesingle{}: [252.05,\allowbreak{} 590.15,\allowbreak{} 416.32,\allowbreak{} 983.2],\allowbreak{} \textquotesingle{}object 3\textquotesingle{}: [382.59,\allowbreak{} 0,\allowbreak{} 593.34,\allowbreak{} 61.06],\allowbreak{} \textquotesingle{}object 4\textquotesingle{}: [423.93,\allowbreak{} 649.71,\allowbreak{} 721.75,\allowbreak{} 1000],\allowbreak{} \textquotesingle{}object 5\textquotesingle{}: [599.26,\allowbreak{} 0,\allowbreak{} 864.26,\allowbreak{} 83.16]\}. The object numbers correspond to the labeled masks rendered in the following image,\allowbreak{} which you can refer to as tool\_\allowbreak{}generated\_\allowbreak{}image\_\allowbreak{}6.
\end{exverb}
\begin{center}\includegraphics[width=0.45\linewidth,height=0.22\textheight,keepaspectratio]{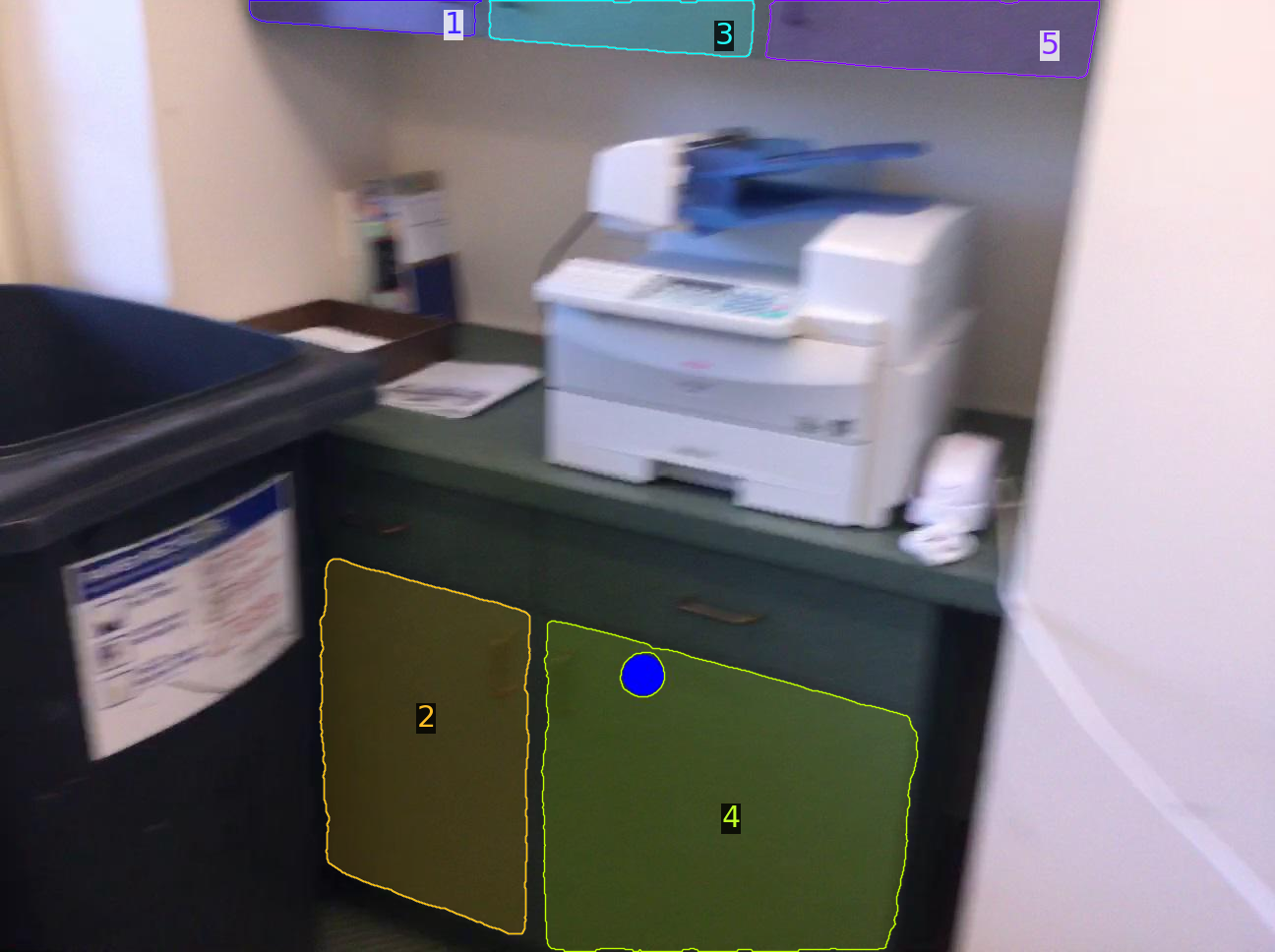}\\[2pt]
{\scriptsize\color{exMute} Image returned with this output.}\end{center}
\begin{exharness}
\exLabel{Harness}{environment update, summarized: new memory entries; new data bindings verbatim}
\begin{exverb}
Memory update: registers 1 new image in memory (tool\_\allowbreak{}generated\_\allowbreak{}image\_\allowbreak{}6,\allowbreak{} from Visual\_\allowbreak{}Grounding\_\allowbreak{}Tool).\\{}
- segment\_\allowbreak{}cabinet: dict (keys: pred\_\allowbreak{}boxes,\allowbreak{} pred\_\allowbreak{}scores,\allowbreak{} pred\_\allowbreak{}masks,\allowbreak{} num\_\allowbreak{}masks,\allowbreak{} orig\_\allowbreak{}img\_\allowbreak{}h,\allowbreak{} orig\_\allowbreak{}img\_\allowbreak{}w)
\end{exverb}
\end{exharness}
\end{exstep}
\begin{exstep}{7}{\toolCode}{}
\exLabel{Model}{text emitted with this call}
\exSayBlock{\textless{}thinking\textgreater{}Now I have the bounding boxes for both the window and cabinet. The red point (window center) is in input\_\allowbreak{}image\_\allowbreak{}1,\allowbreak{} and the blue point (cabinet center) is in input\_\allowbreak{}image\_\allowbreak{}3. I will use the depth maps for each image to get the depth at these points,\allowbreak{} then combine with the camera positions and orientations from the trajectory to compute their 3D coordinates. Finally,\allowbreak{} I\textquotesingle{}ll calculate the Euclidean distance between them.\textless{}/\allowbreak{}thinking\textgreater{}}
\exLabel{Arguments}{}
\begin{exverb}
\{\\{}
~~"\allowbreak{}objective"\allowbreak{}: "\allowbreak{}Compute the 3D coordinates of the window center (red point) in input\_\allowbreak{}image\_\allowbreak{}1 and the cabinet center (blue point) in input\_\allowbreak{}image\_\allowbreak{}3 using the depth maps and camera trajectory data. Then,\allowbreak{} calculate the Euclidean distance between these two 3D points."\allowbreak{}\\{}
\}
\end{exverb}
\exLabel{Tool output}{}
\begin{exverb}
Coding agent generated and executed the following Python code:
\end{exverb}
\exCut{The generated code, 126 lines, is not shown: this run failed with the error below.}
\begin{exverb}
Execution error: index 182417 is out of bounds for axis 0 with size 966
\end{exverb}
\end{exstep}
\begin{exstep}{8}{\toolCode}{}
\exLabel{Model}{text emitted with this call}
\exSayBlock{\textless{}thinking\textgreater{}The error occurred because the pixel coordinates for the window center exceeded the depth map dimensions. I need to adjust the pixel coordinates to fit within the depth map bounds. I\textquotesingle{}ll use the original image dimensions to compute the center,\allowbreak{} then clamp the coordinates to the depth map size before accessing the depth value.\textless{}/\allowbreak{}thinking\textgreater{}}
\exLabel{Arguments}{}
\begin{exverb}
\{\\{}
~~"\allowbreak{}objective"\allowbreak{}: "\allowbreak{}Compute the 3D coordinates of the window center (red point) in input\_\allowbreak{}image\_\allowbreak{}1 and the cabinet center (blue point) in input\_\allowbreak{}image\_\allowbreak{}3 using the depth maps and camera trajectory data. Clamp the pixel coordinates to the depth map dimensions to avoid out-of-bounds errors. Then,\allowbreak{} calculate the Euclidean distance between these two 3D points."\allowbreak{}\\{}
\}
\end{exverb}
\exLabel{Tool output}{}
\begin{exverb}
Coding agent generated and executed the following Python code:\\{}
\textasciigrave{}\textasciigrave{}\textasciigrave{}python\\{}
import numpy as np\\{}
from PIL import Image\\{}
import io\\{}
import base64\\{}
\mbox{}\\{}
\# Step 1: Extract camera intrinsics and extrinsics for input\_\allowbreak{}image\_\allowbreak{}1 and input\_\allowbreak{}image\_\allowbreak{}3\\{}
\# Note: camera\_\allowbreak{}trajectory\_\allowbreak{}result["\allowbreak{}image\_\allowbreak{}ids"\allowbreak{}] is [\textquotesingle{}input\_\allowbreak{}image\_\allowbreak{}1\textquotesingle{},\allowbreak{} \textquotesingle{}input\_\allowbreak{}image\_\allowbreak{}2\textquotesingle{},\allowbreak{} \textquotesingle{}input\_\allowbreak{}image\_\allowbreak{}3\textquotesingle{}]\\{}
\# So index 0 is input\_\allowbreak{}image\_\allowbreak{}1,\allowbreak{} index 2 is input\_\allowbreak{}image\_\allowbreak{}3\\{}
\mbox{}\\{}
K1 =\allowbreak{} camera\_\allowbreak{}trajectory\_\allowbreak{}result["\allowbreak{}intrinsics"\allowbreak{}][0]  \# 3x3 intrinsic matrix for input\_\allowbreak{}image\_\allowbreak{}1\\{}
K3 =\allowbreak{} camera\_\allowbreak{}trajectory\_\allowbreak{}result["\allowbreak{}intrinsics"\allowbreak{}][2]  \# 3x3 intrinsic matrix for input\_\allowbreak{}image\_\allowbreak{}3\\{}
\mbox{}\\{}
\# World-to-camera extrinsic matrices (OpenCV/\allowbreak{}COLMAP format: [R \textbar{} t])\\{}
\# Note: extrinsics are 3x4 matrices,\allowbreak{} where the first 3x3 is rotation and last column is translation\\{}
\# We need to convert to camera-to-world for 3D point reconstruction\\{}
\# But for now,\allowbreak{} we\textquotesingle{}ll use the extrinsics as is for projecting 3D points to camera space\\{}
\mbox{}\\{}
\# Extract extrinsics for input\_\allowbreak{}image\_\allowbreak{}1 and input\_\allowbreak{}image\_\allowbreak{}3\\{}
E1 =\allowbreak{} camera\_\allowbreak{}trajectory\_\allowbreak{}result["\allowbreak{}extrinsics"\allowbreak{}][0]  \# world-to-camera for input\_\allowbreak{}image\_\allowbreak{}1\\{}
E3 =\allowbreak{} camera\_\allowbreak{}trajectory\_\allowbreak{}result["\allowbreak{}extrinsics"\allowbreak{}][2]  \# world-to-camera for input\_\allowbreak{}image\_\allowbreak{}3\\{}
\mbox{}\\{}
\# Step 2: Get window bounding box from segment\_\allowbreak{}window (for input\_\allowbreak{}image\_\allowbreak{}1)\\{}
\# The bounding box is in [x1,\allowbreak{} y1,\allowbreak{} w,\allowbreak{} h] format,\allowbreak{} normalized? Let\textquotesingle{}s check the values\\{}
\# Preview: [[-0.7693090438842773,\allowbreak{} -0.09972977638244629,\allowbreak{} 458.6855163574219,\allowbreak{} 377.0949401855469]]\\{}
\# The x1,\allowbreak{} y1 are negative? That suggests they are normalized or relative to center?\\{}
\# But orig\_\allowbreak{}img\_\allowbreak{}h=\allowbreak{}968,\allowbreak{} orig\_\allowbreak{}img\_\allowbreak{}w=\allowbreak{}1296,\allowbreak{} and depth map is 966x1302 — very close.\\{}
\mbox{}\\{}
\# Let\textquotesingle{}s assume the bounding box is in pixel coordinates,\allowbreak{} but the negative values suggest it\textquotesingle{}s not.\\{}
\# Since the values are large (458,\allowbreak{} 377) and the image is 1296x968,\allowbreak{} it\textquotesingle{}s likely absolute pixel coordinates.\\{}
\# But x1 is negative? That\textquotesingle{}s odd. Let\textquotesingle{}s check if it\textquotesingle{}s normalized.\\{}
\mbox{}\\{}
\# Actually,\allowbreak{} looking at the values: x1=\allowbreak{}-0.769,\allowbreak{} y1=\allowbreak{}-0.099 — these are very small,\allowbreak{} likely normalized.\\{}
\# But then w=\allowbreak{}458,\allowbreak{} h=\allowbreak{}377 — these are large,\allowbreak{} so probably not normalized.\\{}
\mbox{}\\{}
\# Let\textquotesingle{}s assume the bounding box is in absolute pixel coordinates,\allowbreak{} and clamp to image size.\\{}
\# We\textquotesingle{}ll use the original image dimensions from segment\_\allowbreak{}window for clamping.\\{}
\mbox{}\\{}
orig\_\allowbreak{}h =\allowbreak{} segment\_\allowbreak{}window["\allowbreak{}orig\_\allowbreak{}img\_\allowbreak{}h"\allowbreak{}]  \# 968\\{}
orig\_\allowbreak{}w =\allowbreak{} segment\_\allowbreak{}window["\allowbreak{}orig\_\allowbreak{}img\_\allowbreak{}w"\allowbreak{}]  \# 1296\\{}
\mbox{}\\{}
\# Get window box\\{}
window\_\allowbreak{}box =\allowbreak{} segment\_\allowbreak{}window["\allowbreak{}pred\_\allowbreak{}boxes"\allowbreak{}][0]\\{}
x1,\allowbreak{} y1,\allowbreak{} w,\allowbreak{} h =\allowbreak{} window\_\allowbreak{}box\\{}
\mbox{}\\{}
\# Convert to center coordinates\\{}
window\_\allowbreak{}center\_\allowbreak{}x =\allowbreak{} x1 + w /\allowbreak{} 2\\{}
window\_\allowbreak{}center\_\allowbreak{}y =\allowbreak{} y1 + h /\allowbreak{} 2\\{}
\mbox{}\\{}
\# Clamp to image dimensions\\{}
window\_\allowbreak{}center\_\allowbreak{}x =\allowbreak{} np.clip(window\_\allowbreak{}center\_\allowbreak{}x,\allowbreak{} 0,\allowbreak{} orig\_\allowbreak{}w - 1)\\{}
window\_\allowbreak{}center\_\allowbreak{}y =\allowbreak{} np.clip(window\_\allowbreak{}center\_\allowbreak{}y,\allowbreak{} 0,\allowbreak{} orig\_\allowbreak{}h - 1)\\{}
\mbox{}\\{}
\# Step 3: Get depth for window center in input\_\allowbreak{}image\_\allowbreak{}1\\{}
\# depth\_\allowbreak{}result is for input\_\allowbreak{}image\_\allowbreak{}1 (since it\textquotesingle{}s the first depth result and we\textquotesingle{}re using it for input\_\allowbreak{}image\_\allowbreak{}1)\\{}
depth\_\allowbreak{}map\_\allowbreak{}1 =\allowbreak{} depth\_\allowbreak{}result["\allowbreak{}depth\_\allowbreak{}meters"\allowbreak{}]\\{}
\mbox{}\\{}
\# Clamp to depth map dimensions (966,\allowbreak{} 1302)\\{}
depth\_\allowbreak{}h,\allowbreak{} depth\_\allowbreak{}w =\allowbreak{} depth\_\allowbreak{}map\_\allowbreak{}1.shape\\{}
window\_\allowbreak{}center\_\allowbreak{}x\_\allowbreak{}clamped =\allowbreak{} np.clip(window\_\allowbreak{}center\_\allowbreak{}x,\allowbreak{} 0,\allowbreak{} depth\_\allowbreak{}w - 1)\\{}
window\_\allowbreak{}center\_\allowbreak{}y\_\allowbreak{}clamped =\allowbreak{} np.clip(window\_\allowbreak{}center\_\allowbreak{}y,\allowbreak{} 0,\allowbreak{} depth\_\allowbreak{}h - 1)\\{}
\mbox{}\\{}
\# Get depth at window center\\{}
depth\_\allowbreak{}window =\allowbreak{} depth\_\allowbreak{}map\_\allowbreak{}1[int(window\_\allowbreak{}center\_\allowbreak{}y\_\allowbreak{}clamped),\allowbreak{} int(window\_\allowbreak{}center\_\allowbreak{}x\_\allowbreak{}clamped)]\\{}
\mbox{}\\{}
\# Step 4: Convert 2D pixel + depth to 3D point in camera coordinate system (input\_\allowbreak{}image\_\allowbreak{}1)\\{}
\# We need to use the intrinsic matrix K1 to unproject\\{}
\mbox{}\\{}
\# Convert pixel to normalized coordinates\\{}
u =\allowbreak{} window\_\allowbreak{}center\_\allowbreak{}x\_\allowbreak{}clamped\\{}
v =\allowbreak{} window\_\allowbreak{}center\_\allowbreak{}y\_\allowbreak{}clamped\\{}
\mbox{}\\{}
\# Unproject using depth and intrinsic matrix\\{}
\# 3D point in camera coordinates: [X,\allowbreak{} Y,\allowbreak{} Z] =\allowbreak{} depth * [ (u - cx) /\allowbreak{} fx,\allowbreak{} (v - cy) /\allowbreak{} fy,\allowbreak{} 1 ]\\{}
\# But we need to invert the intrinsic matrix to get the 3D point\\{}
\mbox{}\\{}
\# Intrinsic matrix K1 is 3x3\\{}
fx,\allowbreak{} fy =\allowbreak{} K1[0,\allowbreak{} 0],\allowbreak{} K1[1,\allowbreak{} 1]\\{}
cx,\allowbreak{} cy =\allowbreak{} K1[0,\allowbreak{} 2],\allowbreak{} K1[1,\allowbreak{} 2]\\{}
\mbox{}\\{}
\# Compute 3D point in camera coordinates\\{}
X\_\allowbreak{}cam =\allowbreak{} (u - cx) * depth\_\allowbreak{}window /\allowbreak{} fx\\{}
Y\_\allowbreak{}cam =\allowbreak{} (v - cy) * depth\_\allowbreak{}window /\allowbreak{} fy\\{}
Z\_\allowbreak{}cam =\allowbreak{} depth\_\allowbreak{}window\\{}
\mbox{}\\{}
\# So 3D point in camera 1 coordinate system is [X\_\allowbreak{}cam,\allowbreak{} Y\_\allowbreak{}cam,\allowbreak{} Z\_\allowbreak{}cam]\\{}
point\_\allowbreak{}window\_\allowbreak{}cam1 =\allowbreak{} np.array([X\_\allowbreak{}cam,\allowbreak{} Y\_\allowbreak{}cam,\allowbreak{} Z\_\allowbreak{}cam])\\{}
\mbox{}\\{}
\# Step 5: Get cabinet bounding box from segment\_\allowbreak{}cabinet (for input\_\allowbreak{}image\_\allowbreak{}3)\\{}
\# There are 5 detections,\allowbreak{} we\textquotesingle{}ll use the first one (highest score)\\{}
cabinet\_\allowbreak{}box =\allowbreak{} segment\_\allowbreak{}cabinet["\allowbreak{}pred\_\allowbreak{}boxes"\allowbreak{}][0]\\{}
x1\_\allowbreak{}cab,\allowbreak{} y1\_\allowbreak{}cab,\allowbreak{} w\_\allowbreak{}cab,\allowbreak{} h\_\allowbreak{}cab =\allowbreak{} cabinet\_\allowbreak{}box\\{}
\mbox{}\\{}
\# Convert to center\\{}
cabinet\_\allowbreak{}center\_\allowbreak{}x =\allowbreak{} x1\_\allowbreak{}cab + w\_\allowbreak{}cab /\allowbreak{} 2\\{}
cabinet\_\allowbreak{}center\_\allowbreak{}y =\allowbreak{} y1\_\allowbreak{}cab + h\_\allowbreak{}cab /\allowbreak{} 2\\{}
\mbox{}\\{}
\# Clamp to image dimensions\\{}
cabinet\_\allowbreak{}center\_\allowbreak{}x =\allowbreak{} np.clip(cabinet\_\allowbreak{}center\_\allowbreak{}x,\allowbreak{} 0,\allowbreak{} orig\_\allowbreak{}w - 1)\\{}
cabinet\_\allowbreak{}center\_\allowbreak{}y =\allowbreak{} np.clip(cabinet\_\allowbreak{}center\_\allowbreak{}y,\allowbreak{} 0,\allowbreak{} orig\_\allowbreak{}h - 1)\\{}
\mbox{}\\{}
\# Step 6: Get depth for cabinet center in input\_\allowbreak{}image\_\allowbreak{}3\\{}
depth\_\allowbreak{}map\_\allowbreak{}3 =\allowbreak{} depth\_\allowbreak{}result\_\allowbreak{}3["\allowbreak{}depth\_\allowbreak{}meters"\allowbreak{}]\\{}
\mbox{}\\{}
\# Clamp to depth map dimensions\\{}
cabinet\_\allowbreak{}center\_\allowbreak{}x\_\allowbreak{}clamped =\allowbreak{} np.clip(cabinet\_\allowbreak{}center\_\allowbreak{}x,\allowbreak{} 0,\allowbreak{} depth\_\allowbreak{}w - 1)\\{}
cabinet\_\allowbreak{}center\_\allowbreak{}y\_\allowbreak{}clamped =\allowbreak{} np.clip(cabinet\_\allowbreak{}center\_\allowbreak{}y,\allowbreak{} 0,\allowbreak{} depth\_\allowbreak{}h - 1)\\{}
\mbox{}\\{}
\# Get depth at cabinet center\\{}
depth\_\allowbreak{}cabinet =\allowbreak{} depth\_\allowbreak{}map\_\allowbreak{}3[int(cabinet\_\allowbreak{}center\_\allowbreak{}y\_\allowbreak{}clamped),\allowbreak{} int(cabinet\_\allowbreak{}center\_\allowbreak{}x\_\allowbreak{}clamped)]\\{}
\mbox{}\\{}
\# Step 7: Convert 2D pixel + depth to 3D point in camera coordinate system (input\_\allowbreak{}image\_\allowbreak{}3)\\{}
\# Use intrinsic matrix K3\\{}
\mbox{}\\{}
fx3,\allowbreak{} fy3 =\allowbreak{} K3[0,\allowbreak{} 0],\allowbreak{} K3[1,\allowbreak{} 1]\\{}
cx3,\allowbreak{} cy3 =\allowbreak{} K3[0,\allowbreak{} 2],\allowbreak{} K3[1,\allowbreak{} 2]\\{}
\mbox{}\\{}
\# Unproject\\{}
X\_\allowbreak{}cam3 =\allowbreak{} (cabinet\_\allowbreak{}center\_\allowbreak{}x\_\allowbreak{}clamped - cx3) * depth\_\allowbreak{}cabinet /\allowbreak{} fx3\\{}
Y\_\allowbreak{}cam3 =\allowbreak{} (cabinet\_\allowbreak{}center\_\allowbreak{}y\_\allowbreak{}clamped - cy3) * depth\_\allowbreak{}cabinet /\allowbreak{} fy3\\{}
Z\_\allowbreak{}cam3 =\allowbreak{} depth\_\allowbreak{}cabinet\\{}
\mbox{}\\{}
point\_\allowbreak{}cabinet\_\allowbreak{}cam3 =\allowbreak{} np.array([X\_\allowbreak{}cam3,\allowbreak{} Y\_\allowbreak{}cam3,\allowbreak{} Z\_\allowbreak{}cam3])\\{}
\mbox{}\\{}
\# Step 8: Convert both 3D points to world coordinates using extrinsics\\{}
\# Extrinsics are world-to-camera: E =\allowbreak{} [R \textbar{} t],\allowbreak{} so to get world point from camera point:\\{}
\# P\_\allowbreak{}world =\allowbreak{} R\^{}T * (P\_\allowbreak{}cam - t)  [since E * P\_\allowbreak{}world =\allowbreak{} P\_\allowbreak{}cam,\allowbreak{} so P\_\allowbreak{}world =\allowbreak{} E\^{}\{-1\} * P\_\allowbreak{}cam]\\{}
\mbox{}\\{}
\# For input\_\allowbreak{}image\_\allowbreak{}1: E1 =\allowbreak{} [R1 \textbar{} t1],\allowbreak{} so E1\^{}\{-1\} =\allowbreak{} [R1\^{}T \textbar{} -R1\^{}T * t1]\\{}
R1 =\allowbreak{} E1[:,\allowbreak{} :3]  \# 3x3 rotation matrix\\{}
t1 =\allowbreak{} E1[:,\allowbreak{} 3]   \# 3x1 translation vector\\{}
\mbox{}\\{}
\# Invert E1: E1\_\allowbreak{}inv =\allowbreak{} [R1\^{}T,\allowbreak{} -R1\^{}T * t1]\\{}
R1\_\allowbreak{}T =\allowbreak{} R1.T\\{}
t1\_\allowbreak{}inv =\allowbreak{} -R1\_\allowbreak{}T @ t1\\{}
\mbox{}\\{}
\# Convert point\_\allowbreak{}window\_\allowbreak{}cam1 to world coordinates\\{}
point\_\allowbreak{}window\_\allowbreak{}world =\allowbreak{} R1\_\allowbreak{}T @ (point\_\allowbreak{}window\_\allowbreak{}cam1 - t1)\\{}
\mbox{}\\{}
\# For input\_\allowbreak{}image\_\allowbreak{}3: E3 =\allowbreak{} [R3 \textbar{} t3]\\{}
R3 =\allowbreak{} E3[:,\allowbreak{} :3]\\{}
t3 =\allowbreak{} E3[:,\allowbreak{} 3]\\{}
\mbox{}\\{}
R3\_\allowbreak{}T =\allowbreak{} R3.T\\{}
t3\_\allowbreak{}inv =\allowbreak{} -R3\_\allowbreak{}T @ t3\\{}
\mbox{}\\{}
\# Convert point\_\allowbreak{}cabinet\_\allowbreak{}cam3 to world coordinates\\{}
point\_\allowbreak{}cabinet\_\allowbreak{}world =\allowbreak{} R3\_\allowbreak{}T @ (point\_\allowbreak{}cabinet\_\allowbreak{}cam3 - t3)\\{}
\mbox{}\\{}
\# Step 9: Compute Euclidean distance between the two 3D points in world coordinates\\{}
distance =\allowbreak{} np.linalg.norm(point\_\allowbreak{}window\_\allowbreak{}world - point\_\allowbreak{}cabinet\_\allowbreak{}world)\\{}
\mbox{}\\{}
\# Output results\\{}
print(f"\allowbreak{}Window center 3D in world coordinates: \{point\_\allowbreak{}window\_\allowbreak{}world\}"\allowbreak{})\\{}
print(f"\allowbreak{}Cabinet center 3D in world coordinates: \{point\_\allowbreak{}cabinet\_\allowbreak{}world\}"\allowbreak{})\\{}
print(f"\allowbreak{}Euclidean distance between them: \{distance:.4f\} meters"\allowbreak{})\\{}
\mbox{}\\{}
\# Assign to variables for result namespace\\{}
window\_\allowbreak{}3d\_\allowbreak{}world =\allowbreak{} point\_\allowbreak{}window\_\allowbreak{}world\\{}
cabinet\_\allowbreak{}3d\_\allowbreak{}world =\allowbreak{} point\_\allowbreak{}cabinet\_\allowbreak{}world\\{}
distance\_\allowbreak{}3d =\allowbreak{} distance\\{}
\textasciigrave{}\textasciigrave{}\textasciigrave{}\\{}
\mbox{}\\{}
\mbox{}\\{}
\mbox{}\\{}
Execution output:\\{}
Window center 3D in world coordinates: [-0.59107007 -0.35877492  1.67238153]\\{}
Cabinet center 3D in world coordinates: [ 0.5192192  -0.77874663  1.52856489]\\{}
Euclidean distance between them: 1.1957 meters
\end{exverb}
\end{exstep}
\begin{exstep}{}{\exTool{exInk}{Final answer}}{}
\exLabel{Model}{}
\exSayBlock{\textless{}thinking\textgreater{}Based on the computation,\allowbreak{} the Euclidean distance between the center of the window (red point) and the center of the cabinet (blue point) is 1.1957 meters. This was calculated by estimating the 3D coordinates of both points using depth maps and camera trajectory data,\allowbreak{} then computing the distance in the world coordinate system.\textless{}/\allowbreak{}thinking\textgreater{}
\textless{}answer\textgreater{}1.1957\textless{}/\allowbreak{}answer\textgreater{}}
\end{exstep}

%% file: tables/std_main.tex
\begin{table}[h]
\centering
\small
\setlength{\tabcolsep}{5pt}
\renewcommand{\arraystretch}{1.2}
\begin{tabular}{cl|cc|cc}
\toprule
\multicolumn{2}{c|}{} & \multicolumn{2}{c|}{\textbf{Qwen3-VL-8B}} & \multicolumn{2}{c}{\textbf{Qwen3-VL-32B}} \\
\textbf{Domain} & \textbf{Dataset} & \textbf{Base} & \textbf{Ours} & \textbf{Base} & \textbf{Ours} \\
\midrule
& CountQA & 0.40 & 1.08 & 0.40 & 0.53 \\
& FSC-147 & 0.26 & 0.17 & 0.26 & 0.59 \\
& OdinW-13-All & 0.98 & 0.94 & 0.89 & 1.94 \\
& FSCD-147 & 0.14 & 0.06 & 0.14 & 0.10 \\
\multirow{-5}{*}{\rotatebox[origin=c]{90}{\shortstack[c]{Grounded\\Perception}}}
 & FSCD-LVIS & 0.15 & 0.14 & 0.15 & 0.22 \\
\midrule
& MMSearch & 0.00 & 4.94 & 0.00 & 1.08 \\
& MMSearch-Plus & 0.40 & 1.27 & 0.25 & 0.61 \\
& RealX-Bench & 1.10 & 1.60 & 1.10 & 2.01 \\
& FVQA & 1.41 & 3.93 & 1.41 & 0.67 \\
& HLE-VL & 0.00 & 0.87 & 0.00 & 1.37 \\
\multirow{-6}{*}{\rotatebox[origin=c]{90}{\shortstack[c]{Visual Search and\\Deep Research}}}
 & BC-VL & 1.41 & 0.77 & 1.41 & 0.29 \\
\midrule
& RealWorldQA & 0.80 & 2.27 & 0.80 & 0.57 \\
& SimpleVQA & 0.07 & 0.69 & 0.07 & 0.60 \\
& HallusionBench & 0.64 & 0.61 & 0.64 & 1.64 \\
& FREAK-VQA & 0.67 & 0.65 & 0.67 & 0.80 \\
& FREAK-MCQ & 0.70 & 0.61 & 0.70 & 0.35 \\
\multirow{-6}{*}{\rotatebox[origin=c]{90}{\shortstack[c]{General VQA and\\Hallucination}}}
 & V* & 1.11 & 0.91 & 1.11 & 2.12 \\
\midrule
& MMSI-Bench & 0.40 & 0.62 & 0.40 & 0.31 \\
& ERQA & 0.18 & 1.95 & 0.18 & 2.29 \\
& EmbSpatial & 0.10 & 0.34 & 0.10 & 0.22 \\
& SparBench & 0.36 & 1.47 & 0.35 & 1.67 \\
\multirow{-5}{*}{\rotatebox[origin=c]{90}{\shortstack[c]{Spatial\\Reasoning}}}
 & MindCube & 0.59 & 1.08 & 0.59 & 1.15 \\
\midrule
{} & Average & 0.16 & 0.35 & 0.16 & 0.24 \\
\bottomrule
\end{tabular}
\caption{Standard deviations across three runs for the Qwen3-VL Base and \ours results in Table~\ref{tab:harness-combined}. \emph{Average} is the standard deviation of the per-run $22$-dataset average.}
\label{tab:appendix_std_main}
\end{table}

%% file: tables/std_ablation.tex
\begin{table*}[h]
\centering
\small
\setlength{\tabcolsep}{4.5pt}
\resizebox{\textwidth}{!}{%
\begin{tabular}{l*{10}{c}}
\toprule
\multirow{2}{*}{\textbf{Variant}}
& \multicolumn{2}{c}{\textbf{Grounding}}
& \multicolumn{2}{c}{\textbf{VS \& DR}}
& \multicolumn{2}{c}{\textbf{General VQA}}
& \multicolumn{2}{c}{\textbf{Spatial}}
& \multicolumn{2}{c}{\textbf{Overall}} \\
\cmidrule(lr){2-3}\cmidrule(lr){4-5}\cmidrule(lr){6-7}\cmidrule(lr){8-9}\cmidrule(lr){10-11}
& \textbf{8B} & \textbf{32B} & \textbf{8B} & \textbf{32B} & \textbf{8B} & \textbf{32B} & \textbf{8B} & \textbf{32B} & \textbf{8B} & \textbf{32B} \\
\midrule
Baseline (no tool) & 0.64 & 0.63 & 0.46 & 0.46 & 0.36 & 0.36 & 0.24 & 0.24 & 0.22 & 0.22 \\
+ Tool & 0.23 & 0.19 & 1.11 & 0.75 & 0.40 & 0.43 & 0.59 & 0.73 & 0.34 & 0.29 \\
+ Capability layer & 0.27 & 0.31 & 0.71 & 0.64 & 0.36 & 0.42 & 0.55 & 1.01 & 0.25 & 0.32 \\
+ Textual memory & 0.22 & 0.20 & 0.74 & 0.56 & 0.56 & 0.46 & 0.31 & 0.39 & 0.25 & 0.21 \\
\textbf{OpenVLHarness} & 0.27 & 0.24 & 1.05 & 0.65 & 0.62 & 0.44 & 0.56 & 0.75 & 0.34 & 0.28 \\
\bottomrule
\end{tabular}%
}
\caption{Standard deviations across three runs for the ablation in Table~\ref{tab:ablation}. Domain and overall columns are standard deviations of the per-run averages.}
\label{tab:appendix_std_ablation}
\end{table*}